%% file: main.tex
\documentclass{article}
\usepackage[T1]{fontenc}
\usepackage[utf8]{inputenc}
\usepackage{microtype}
\usepackage{graphicx}
\usepackage{booktabs}
\usepackage{multirow}
\usepackage{array}
\usepackage{pdflscape}
\usepackage{afterpage}
\usepackage{adjustbox}
\usepackage{enumitem}
\usepackage{tikz}
\usetikzlibrary{arrows.meta,positioning,calc}
\usepackage{xurl}
\usepackage{hyperref}

\usepackage[preprint]{icml2026}
\usepackage{amsmath}
\usepackage{amssymb}
\input{math_commands}

\input{numbers}
\usepackage[capitalize,noabbrev]{cleveref}
\setlist[itemize]{leftmargin=1.2em,itemsep=1pt,topsep=2pt,parsep=0pt}
\definecolor{ink}{HTML}{0F172A}
\definecolor{slate}{HTML}{475569}
\definecolor{muted}{HTML}{64748B}
\definecolor{cardbg}{HTML}{F8FAFC}
\definecolor{cardbd}{HTML}{CBD5E1}
\definecolor{icefill}{HTML}{EFF6FF}
\definecolor{icebd}{HTML}{3B82F6}
\definecolor{icedark}{HTML}{1E3A8A}
\definecolor{amberfill}{HTML}{FFFBEB}
\definecolor{amberbd}{HTML}{D97706}
\definecolor{amberdark}{HTML}{78350F}
\definecolor{emefill}{HTML}{ECFDF5}
\definecolor{emebd}{HTML}{059669}
\definecolor{emedark}{HTML}{065F46}
\definecolor{rosefill}{HTML}{FEF2F2}
\definecolor{rosebd}{HTML}{DC2626}
\definecolor{rosedark}{HTML}{991B1B}
\definecolor{viofill}{HTML}{F5F3FF}
\definecolor{viobd}{HTML}{7C3AED}
\definecolor{viodark}{HTML}{4C1D95}
\definecolor{arrowc}{HTML}{475569}

\tikzset{
  snowflake/.pic={
    \draw[icebd, line width=0.55pt, line cap=round]
      (0:1) -- (180:1) (60:1) -- (240:1) (120:1) -- (300:1)
      (0:0.55) -- ++(45:0.28) (0:0.55) -- ++(-45:0.28)
      (60:0.55) -- ++(105:0.28) (60:0.55) -- ++(15:0.28)
      (120:0.55) -- ++(165:0.28) (120:0.55) -- ++(75:0.28)
      (180:0.55) -- ++(225:0.28) (180:0.55) -- ++(135:0.28)
      (240:0.55) -- ++(285:0.28) (240:0.55) -- ++(195:0.28)
      (300:0.55) -- ++(345:0.28) (300:0.55) -- ++(255:0.28);
  },
  snowamber/.pic={
    \draw[amberbd, line width=0.55pt, line cap=round]
      (0:1) -- (180:1) (60:1) -- (240:1) (120:1) -- (300:1)
      (0:0.55) -- ++(45:0.28) (0:0.55) -- ++(-45:0.28)
      (60:0.55) -- ++(105:0.28) (60:0.55) -- ++(15:0.28)
      (120:0.55) -- ++(165:0.28) (120:0.55) -- ++(75:0.28)
      (180:0.55) -- ++(225:0.28) (180:0.55) -- ++(135:0.28)
      (240:0.55) -- ++(285:0.28) (240:0.55) -- ++(195:0.28)
      (300:0.55) -- ++(345:0.28) (300:0.55) -- ++(255:0.28);
  }
}

\icmltitlerunning{Are All Models Multimodal? Linear Bridges Connect Frozen Models Within, Not Outside, Their Training Subspaces}

\renewcommand{\today}{November 27, 2026}
\AtBeginDocument{\hypersetup{pdfsubject={}}}
\begin{document}

\twocolumn[
  \icmltitle{Are All Models Multimodal? Linear Bridges Connect\\ Frozen Models Within, Not Outside, Their Training Subspaces}
  \icmlsetsymbol{equal}{*}
  \vspace{-0.20in}
  \begin{icmlauthorlist}
    \icmlauthor{Gregory Kielian}{goog,equal}
    \icmlauthor{Shenil Dodhia}{goog,equal}
    \icmlauthor{Phan Douangphachanh}{veri,equal}
    \icmlauthor{Umberto Ravaioli}{goog,equal}
    \icmlauthor{Anand Patel}{goog}
    \icmlauthor{Kahye Song}{goog}
    \icmlauthor{Alec Go}{goog}
    \icmlauthor{Andrew Yeung}{goog}
    \icmlauthor{Jiageng Zhang}{goog}
    \icmlauthor{Alex May Ramirez}{goog}
    \icmlauthor{Priyanka Gupta-Martinez}{goog}
    \icmlauthor{Naveen Dodda}{goog}
    \icmlauthor{Aaditya Kandibanda}{goog}
    \icmlauthor{Matthew Wilson}{goog}
    \icmlauthor{Julian Mullings-Black}{goog}
    \icmlauthor{Xinting Jiang}{brown}
    \icmlauthor{Yuting Zhao}{kyushu}
    \icmlauthor{Ayu Purwarianti}{itb}
    \icmlauthor{Junyi Luo}{brown}
    \icmlauthor{Kauna Lei}{ucla}
    \icmlauthor{Charibeth Cheng}{dlsu}
    \icmlauthor{Dohyeong Kim}{snu}
    \icmlauthor{Ann Franchesca Laguna}{dlsu}
    \icmlauthor{Xinyi Xu}{umich}
    \icmlauthor{Ben Laurie}{goog}
    \icmlauthor{Mehdi Saligane}{goog,brown}
  \end{icmlauthorlist}
  \icmlaffiliation{goog}{Google}
  \icmlaffiliation{veri}{VeriSilicon}
  \icmlaffiliation{brown}{Brown University}
  \icmlaffiliation{kyushu}{Kyushu University}
  \icmlaffiliation{itb}{Institut Teknologi Bandung}
  \icmlaffiliation{ucla}{UCLA}
  \icmlaffiliation{dlsu}{De La Salle University}
  \icmlaffiliation{snu}{Seoul National University}
  \icmlaffiliation{umich}{University of Michigan}
  \icmlcorrespondingauthor{Mehdi Saligane}{saligane@google.com, mehdi\_saligane@brown.edu}
  \icmlkeywords{embedding alignment, frozen language models, speech adapters, soft prompts, model composition}
  \vskip 0.22in
]
\printAffiliationsAndNotice{\icmlEqualContribution}

\input{sections/00_abstract}
\input{sections/01_intro}
\input{sections/02_related_work}
\input{sections/03_setup}
\input{sections/04_tokens}
\input{sections/05_speech}
\input{sections/06_composition}
\input{sections/07_why}
\input{sections/08_conclusion}

\bibliography{references}
\bibliographystyle{icml2026}

\appendix
\onecolumn
\input{sections/90_appendix}

\end{document}

%% file: math_commands.tex
\usepackage{amsfonts,bm}

\def\1{\bm{1}}

\DeclareMathAlphabet{\mathsfit}{\encodingdefault}{\sfdefault}{m}{sl}
\SetMathAlphabet{\mathsfit}{bold}{\encodingdefault}{\sfdefault}{bx}{n}



%% file: numbers.tex
\makeatletter
\newcommand{\defmmnum}[2]{\expandafter\def\csname mmnum@#1\endcsname{#2}}
\newcommand{\mnum}[1]{%
  \ifcsname mmnum@#1\endcsname
    \csname mmnum@#1\endcsname
  \else
    \PackageError{numbers}{Undefined metric key '#1'}{Run build_paper_tables.py}%
  \fi
}
\makeatother

\newcommand{\TokenRadiusMaxRows}{50{,}000}

\newcommand{\WsOneDonorsCount}{3}

\newcommand{\WsThreeTotalPairs}{4}

\newcommand{\WsFourBInDistTotal}{9}

\newcommand{\WsFourCWikiCompTotal}{8}

\newcommand{\LibriTrainHours}{960}
\newcommand{\LibriTrainUtts}{281{,}241}
\newcommand{\LibriValUtts}{1{,}000}
\newcommand{\LibriTestCleanUtts}{2{,}620}
\newcommand{\LibriTestOtherUtts}{2{,}939}
\newcommand{\UttWerHalfThresholdPct}{50}

\newcommand{\SlamAsrFrameStack}{5}
\newcommand{\SlamAsrHiddenDim}{2048}
\newcommand{\WsTwoUniformEpochs}{5}
\newcommand{\WsTwoBatchSize}{16}

\newcommand{\WsTwoTotalSteps}{87{,}890}

\newcommand{\PiqaEvalItems}{1{,}838}
\newcommand{\HsEvalItems}{2{,}000}
\newcommand{\HsTotalEvalItems}{10{,}042}
\newcommand{\PiqaChancePct}{50.00}
\newcommand{\HsChancePct}{25.00}
\newcommand{\GemmaThreeDim}{2560}
\newcommand{\GemmaThreeVocab}{262{,}208}
\newcommand{\GemmaThreeEmbedScaleBf}{50.50}
\newcommand{\GemmaTwoDim}{2304}
\newcommand{\GemmaTwoVocab}{256{,}000}
\newcommand{\GemmaTwoEmbedScaleBf}{48.00}
\newcommand{\QwenDim}{1536}
\newcommand{\QwenVocab}{151{,}936}
\newcommand{\QwenThreeDim}{2560}
\newcommand{\QwenThreeVocab}{151{,}936}
\newcommand{\SmolDim}{2048}
\newcommand{\SmolVocab}{49{,}152}
\newcommand{\PhiDim}{2560}
\newcommand{\PhiVocab}{51{,}200}
\newcommand{\UnitEmbedScale}{1.00}
\newcommand{\WhisperDim}{768}
\newcommand{\WhisperFrameHz}{50}
\newcommand{\MoonshineDim}{416}
\newcommand{\MoonshineFrameHz}{41.67}

\newcommand{\EvalMaxNewTokens}{160}
\newcommand{\AnchorTrainPct}{80}
\newcommand{\AnchorTestPct}{20}
\newcommand{\MixtureTextPct}{50}
\newcommand{\MixtureAudioPct}{50}
\newcommand{\ZeroAudioPct}{0}
\newcommand{\FullSignalPct}{100}
\newcommand{\WsFourBNoiseKZero}{0}
\newcommand{\WsFourBNoiseKHalf}{0.50}
\newcommand{\WsFourBNoiseKOne}{1.00}
\newcommand{\WsFourBNoiseKSevenSevenThree}{7.73}
\newcommand{\RadiusQwenOneFiveB}{1.03}
\newcommand{\RadiusQwenThreeFourB}{1.13}
\newcommand{\RadiusPhiTwo}{1.21}
\newcommand{\RadiusSmolLMTwo}{3.07}
\newcommand{\RadiusGemmaThree}{50.8}
\newcommand{\RadiusGemmaTwo}{80.7}
\newcommand{\RadiusSpreadFactor}{78}
\newcommand{\WsOneDoneCells}{25}

\newcommand{\WsOneNativePiqa}{78.73}
\newcommand{\WsOneNativeHs}{57.50}
\newcommand{\WsOneNativeRr}{1.000}
\newcommand{\WsOneGateAnchors}{242{,}564}
\newcommand{\WsOneGateCos}{1.0000}
\newcommand{\WsOneGatePatOne}{100.00}
\newcommand{\WsOneGateRr}{1.000}
\newcommand{\WsOneGatePiqa}{78.73}
\newcommand{\WsOneGateHs}{57.50}
\newcommand{\WsOneGateDeltaPiqa}{0.00}

\newcommand{\WsOneQwenClReplicaAnchors}{65{,}406}
\newcommand{\WsOneQwenClReplicaCos}{0.6329}
\newcommand{\WsOneQwenClReplicaPatOne}{97.76}
\newcommand{\WsOneQwenClReplicaRr}{0.681}

\newcommand{\WsOneQwenClReplicaPiqa}{77.26}
\newcommand{\WsOneQwenClReplicaHs}{54.50}

\newcommand{\WsOneQwenLstsqNoneAnchors}{82{,}663}
\newcommand{\WsOneQwenLstsqNoneCos}{0.6070}
\newcommand{\WsOneQwenLstsqNonePatOne}{97.80}
\newcommand{\WsOneQwenLstsqNoneRr}{0.653}

\newcommand{\WsOneQwenLstsqNonePiqa}{77.48}
\newcommand{\WsOneQwenLstsqNoneHs}{53.45}

\newcommand{\WsOneQwenLstsqLtwoRr}{0.020}

\newcommand{\WsOneQwenLstsqLtwoPiqa}{52.12}
\newcommand{\WsOneQwenLstsqLtwoHs}{26.55}

\newcommand{\WsOneQwenLstsqScaledAnchors}{82{,}663}
\newcommand{\WsOneQwenLstsqScaledCos}{0.6070}
\newcommand{\WsOneQwenLstsqScaledPatOne}{97.80}
\newcommand{\WsOneQwenLstsqScaledRr}{1.000}

\newcommand{\WsOneQwenLstsqScaledPiqa}{77.48}
\newcommand{\WsOneQwenLstsqScaledHs}{56.40}
\newcommand{\WsOneQwenLstsqScaledRetPiqa}{98.4}

\newcommand{\WsOneQwenLstsqRmsRr}{1.075}

\newcommand{\WsOneQwenLstsqRmsPiqa}{76.50}
\newcommand{\WsOneQwenLstsqRmsHs}{55.60}

\newcommand{\WsOneQwenLstsqLnRr}{1.069}

\newcommand{\WsOneQwenLstsqLnPiqa}{76.55}
\newcommand{\WsOneQwenLstsqLnHs}{55.25}

\newcommand{\WsOneQwenProcNoneAnchors}{82{,}663}
\newcommand{\WsOneQwenProcNoneCos}{0.5296}
\newcommand{\WsOneQwenProcNonePatOne}{97.59}
\newcommand{\WsOneQwenProcNoneRr}{0.023}

\newcommand{\WsOneQwenProcNonePiqa}{51.25}
\newcommand{\WsOneQwenProcNoneHs}{26.90}

\newcommand{\WsOneQwenProcLtwoRr}{0.020}

\newcommand{\WsOneQwenProcLtwoPiqa}{50.38}
\newcommand{\WsOneQwenProcLtwoHs}{27.10}

\newcommand{\WsOneQwenProcScaledAnchors}{82{,}663}
\newcommand{\WsOneQwenProcScaledCos}{0.5296}
\newcommand{\WsOneQwenProcScaledPatOne}{97.59}
\newcommand{\WsOneQwenProcScaledRr}{1.000}

\newcommand{\WsOneQwenProcScaledPiqa}{53.86}
\newcommand{\WsOneQwenProcScaledHs}{27.95}

\newcommand{\WsOneQwenProcRmsRr}{1.064}

\newcommand{\WsOneQwenProcRmsPiqa}{52.77}
\newcommand{\WsOneQwenProcRmsHs}{27.25}

\newcommand{\WsOneQwenProcLnRr}{1.063}

\newcommand{\WsOneQwenProcLnPiqa}{64.47}
\newcommand{\WsOneQwenProcLnHs}{38.35}

\newcommand{\WsOneQwenSwapNoneAnchors}{82{,}663}
\newcommand{\WsOneQwenSwapNoneCos}{-0.0008}
\newcommand{\WsOneQwenSwapNonePatOne}{0.02}
\newcommand{\WsOneQwenSwapNoneRr}{0.023}

\newcommand{\WsOneQwenSwapNonePiqa}{49.29}
\newcommand{\WsOneQwenSwapNoneHs}{24.35}

\newcommand{\WsOneQwenSwapScaledAnchors}{82{,}663}
\newcommand{\WsOneQwenSwapScaledCos}{-0.0008}
\newcommand{\WsOneQwenSwapScaledPatOne}{0.02}
\newcommand{\WsOneQwenSwapScaledRr}{1.000}

\newcommand{\WsOneQwenSwapScaledPiqa}{49.40}
\newcommand{\WsOneQwenSwapScaledHs}{25.40}

\newcommand{\WsOneQwenRandNoneAnchors}{82{,}663}
\newcommand{\WsOneQwenRandNoneCos}{0.0005}
\newcommand{\WsOneQwenRandNonePatOne}{0.02}
\newcommand{\WsOneQwenRandNoneRr}{0.023}

\newcommand{\WsOneQwenRandNonePiqa}{50.60}
\newcommand{\WsOneQwenRandNoneHs}{25.45}

\newcommand{\WsOneQwenRandScaledAnchors}{82{,}663}
\newcommand{\WsOneQwenRandScaledCos}{0.0005}
\newcommand{\WsOneQwenRandScaledPatOne}{0.02}
\newcommand{\WsOneQwenRandScaledRr}{1.000}

\newcommand{\WsOneQwenRandScaledPiqa}{49.78}
\newcommand{\WsOneQwenRandScaledHs}{25.80}

\newcommand{\WsOneSmolLstsqNoneAnchors}{45{,}554}
\newcommand{\WsOneSmolLstsqNoneCos}{0.6296}
\newcommand{\WsOneSmolLstsqNonePatOne}{99.08}
\newcommand{\WsOneSmolLstsqNoneRr}{0.675}

\newcommand{\WsOneSmolLstsqNonePiqa}{75.95}
\newcommand{\WsOneSmolLstsqNoneHs}{54.35}

\newcommand{\WsOneSmolLstsqLtwoRr}{0.020}

\newcommand{\WsOneSmolLstsqLtwoPiqa}{51.96}
\newcommand{\WsOneSmolLstsqLtwoHs}{26.00}

\newcommand{\WsOneSmolLstsqScaledAnchors}{45{,}554}
\newcommand{\WsOneSmolLstsqScaledCos}{0.6296}
\newcommand{\WsOneSmolLstsqScaledPatOne}{99.08}
\newcommand{\WsOneSmolLstsqScaledRr}{1.000}

\newcommand{\WsOneSmolLstsqScaledPiqa}{76.82}
\newcommand{\WsOneSmolLstsqScaledHs}{56.70}

\newcommand{\WsOneSmolLstsqScaledRetHs}{98.6}

\newcommand{\WsOneSmolLstsqRmsRr}{1.073}

\newcommand{\WsOneSmolLstsqRmsPiqa}{75.52}
\newcommand{\WsOneSmolLstsqRmsHs}{55.55}

\newcommand{\WsOneSmolLstsqLnRr}{1.070}

\newcommand{\WsOneSmolLstsqLnPiqa}{76.22}
\newcommand{\WsOneSmolLstsqLnHs}{54.35}

\newcommand{\WsOneSmolProcNoneAnchors}{45{,}554}
\newcommand{\WsOneSmolProcNoneCos}{0.5397}
\newcommand{\WsOneSmolProcNonePatOne}{98.64}
\newcommand{\WsOneSmolProcNoneRr}{0.060}

\newcommand{\WsOneSmolProcNonePiqa}{52.34}
\newcommand{\WsOneSmolProcNoneHs}{26.35}

\newcommand{\WsOneSmolProcLtwoRr}{0.020}

\newcommand{\WsOneSmolProcLtwoPiqa}{49.35}
\newcommand{\WsOneSmolProcLtwoHs}{25.90}

\newcommand{\WsOneSmolProcScaledAnchors}{45{,}554}
\newcommand{\WsOneSmolProcScaledCos}{0.5397}
\newcommand{\WsOneSmolProcScaledPatOne}{98.64}
\newcommand{\WsOneSmolProcScaledRr}{1.000}

\newcommand{\WsOneSmolProcScaledPiqa}{51.63}
\newcommand{\WsOneSmolProcScaledHs}{26.10}

\newcommand{\WsOneSmolProcRmsRr}{1.064}

\newcommand{\WsOneSmolProcRmsPiqa}{50.76}
\newcommand{\WsOneSmolProcRmsHs}{25.75}

\newcommand{\WsOneSmolProcLnRr}{1.064}

\newcommand{\WsOneSmolProcLnPiqa}{56.26}
\newcommand{\WsOneSmolProcLnHs}{30.65}

\newcommand{\WsOneSmolRandNoneAnchors}{45{,}554}
\newcommand{\WsOneSmolRandNoneCos}{-0.0018}
\newcommand{\WsOneSmolRandNonePatOne}{0.01}
\newcommand{\WsOneSmolRandNoneRr}{0.060}

\newcommand{\WsOneSmolRandNonePiqa}{48.42}
\newcommand{\WsOneSmolRandNoneHs}{25.35}

\newcommand{\WsOneSmolRandScaledAnchors}{45{,}554}
\newcommand{\WsOneSmolRandScaledCos}{-0.0018}
\newcommand{\WsOneSmolRandScaledPatOne}{0.01}
\newcommand{\WsOneSmolRandScaledRr}{1.000}

\newcommand{\WsOneSmolRandScaledPiqa}{50.00}
\newcommand{\WsOneSmolRandScaledHs}{24.95}

\newcommand{\WsOnePhiLstsqNoneAnchors}{44{,}975}
\newcommand{\WsOnePhiLstsqNoneCos}{0.4527}
\newcommand{\WsOnePhiLstsqNonePatOne}{97.35}
\newcommand{\WsOnePhiLstsqNoneRr}{0.555}

\newcommand{\WsOnePhiLstsqNonePiqa}{70.95}
\newcommand{\WsOnePhiLstsqNoneHs}{42.95}

\newcommand{\WsOnePhiLstsqLtwoRr}{0.020}

\newcommand{\WsOnePhiLstsqLtwoPiqa}{50.71}
\newcommand{\WsOnePhiLstsqLtwoHs}{26.80}

\newcommand{\WsOnePhiLstsqScaledAnchors}{44{,}975}
\newcommand{\WsOnePhiLstsqScaledCos}{0.4527}
\newcommand{\WsOnePhiLstsqScaledPatOne}{97.35}
\newcommand{\WsOnePhiLstsqScaledRr}{1.000}

\newcommand{\WsOnePhiLstsqScaledPiqa}{74.16}
\newcommand{\WsOnePhiLstsqScaledHs}{49.30}
\newcommand{\WsOnePhiLstsqScaledRetPiqa}{94.2}
\newcommand{\WsOnePhiLstsqScaledRetHs}{85.7}

\newcommand{\WsOnePhiLstsqRmsRr}{1.095}

\newcommand{\WsOnePhiLstsqRmsPiqa}{71.33}
\newcommand{\WsOnePhiLstsqRmsHs}{46.60}

\newcommand{\WsOnePhiLstsqLnRr}{1.080}

\newcommand{\WsOnePhiLstsqLnPiqa}{73.99}
\newcommand{\WsOnePhiLstsqLnHs}{50.75}

\newcommand{\WsOnePhiProcNoneAnchors}{44{,}975}
\newcommand{\WsOnePhiProcNoneCos}{0.3397}
\newcommand{\WsOnePhiProcNonePatOne}{97.87}
\newcommand{\WsOnePhiProcNoneRr}{0.024}

\newcommand{\WsOnePhiProcNonePiqa}{51.96}
\newcommand{\WsOnePhiProcNoneHs}{25.85}

\newcommand{\WsOnePhiProcLtwoRr}{0.020}

\newcommand{\WsOnePhiProcLtwoPiqa}{52.45}
\newcommand{\WsOnePhiProcLtwoHs}{25.55}

\newcommand{\WsOnePhiProcScaledAnchors}{44{,}975}
\newcommand{\WsOnePhiProcScaledCos}{0.3397}
\newcommand{\WsOnePhiProcScaledPatOne}{97.87}
\newcommand{\WsOnePhiProcScaledRr}{1.000}

\newcommand{\WsOnePhiProcScaledPiqa}{56.37}
\newcommand{\WsOnePhiProcScaledHs}{27.90}

\newcommand{\WsOnePhiProcRmsRr}{1.010}

\newcommand{\WsOnePhiProcRmsPiqa}{55.55}
\newcommand{\WsOnePhiProcRmsHs}{28.80}

\newcommand{\WsOnePhiProcLnRr}{1.014}

\newcommand{\WsOnePhiProcLnPiqa}{75.30}
\newcommand{\WsOnePhiProcLnHs}{54.10}

\newcommand{\WsOnePhiSwapNoneAnchors}{44{,}975}
\newcommand{\WsOnePhiSwapNoneCos}{0.0002}
\newcommand{\WsOnePhiSwapNonePatOne}{0.01}
\newcommand{\WsOnePhiSwapNoneRr}{0.024}

\newcommand{\WsOnePhiSwapNonePiqa}{50.27}
\newcommand{\WsOnePhiSwapNoneHs}{24.40}

\newcommand{\WsOnePhiSwapScaledAnchors}{44{,}975}
\newcommand{\WsOnePhiSwapScaledCos}{0.0002}
\newcommand{\WsOnePhiSwapScaledPatOne}{0.01}
\newcommand{\WsOnePhiSwapScaledRr}{1.000}

\newcommand{\WsOnePhiSwapScaledPiqa}{51.58}
\newcommand{\WsOnePhiSwapScaledHs}{26.45}

\newcommand{\WsOnePhiRandNoneAnchors}{44{,}975}
\newcommand{\WsOnePhiRandNoneCos}{0.0000}
\newcommand{\WsOnePhiRandNonePatOne}{0.00}
\newcommand{\WsOnePhiRandNoneRr}{0.024}

\newcommand{\WsOnePhiRandNonePiqa}{52.23}
\newcommand{\WsOnePhiRandNoneHs}{26.45}

\newcommand{\WsOnePhiRandScaledAnchors}{44{,}975}
\newcommand{\WsOnePhiRandScaledCos}{0.0000}
\newcommand{\WsOnePhiRandScaledPatOne}{0.00}
\newcommand{\WsOnePhiRandScaledRr}{1.000}

\newcommand{\WsOnePhiRandScaledPiqa}{48.53}
\newcommand{\WsOnePhiRandScaledHs}{25.80}

\newcommand{\WsOneBestPiqa}{77.48}

\newcommand{\WsOneBestHs}{56.70}

\newcommand{\WsOneLtwoMinPiqa}{49.35}
\newcommand{\WsOneLtwoMaxPiqa}{52.45}
\newcommand{\WsOneLtwoMinHs}{25.55}
\newcommand{\WsOneLtwoMaxHs}{27.10}

\newcommand{\WsOneQwenLstsqNormGainHs}{+2.95}

\newcommand{\WsOneSmolLstsqNormGainHs}{+2.35}

\newcommand{\WsOnePhiLstsqNormGainHs}{+6.35}

\newcommand{\WsOneLstsqLtwoRestoreMinGainHs}{22.50}
\newcommand{\WsOneLstsqLtwoRestoreMaxGainHs}{30.70}
\newcommand{\WsOneLstsqNoneMinRr}{0.555}
\newcommand{\WsOneLstsqNoneMaxRr}{0.675}
\newcommand{\WsOneLstsqRmsLnMinRr}{1.069}
\newcommand{\WsOneLstsqRmsLnMaxRr}{1.095}
\newcommand{\WsOneLstsqScaledTopCells}{5}
\newcommand{\WsOneLstsqDonorTaskCells}{6}
\newcommand{\WsOneLstsqPhiLnOverScaledHs}{1.45}

\newcommand{\WsOneUntiedVsTiedMinGainPiqa}{2.72}
\newcommand{\WsOneUntiedVsTiedMaxGainPiqa}{5.06}
\newcommand{\WsOneUntiedVsTiedMinGainHs}{3.65}
\newcommand{\WsOneUntiedVsTiedMaxGainHs}{7.00}

\newcommand{\NativeWhisperWerClean}{4.05}
\newcommand{\NativeWhisperWerOther}{8.36}
\newcommand{\NativeMoonshineWerClean}{3.62}
\newcommand{\NativeMoonshineWerOther}{8.56}

\newcommand{\WsTwoWhisperQwenLnMuonParamsM}{1.19M}
\newcommand{\WsTwoWhisperQwenLnMuonBestValCell}{0.1394}

\newcommand{\WsTwoWhisperQwenLnMuonWerCleanCell}{4.33\%}
\newcommand{\WsTwoWhisperQwenLnMuonWerOtherCell}{9.96\%}

\newcommand{\WsTwoWhisperQwenLnMuonWerClean}{4.33}
\newcommand{\WsTwoWhisperQwenLnMuonWerOther}{9.96}

\newcommand{\WsTwoWhisperQwenLnAdamwParamsM}{1.19M}
\newcommand{\WsTwoWhisperQwenLnAdamwBestValCell}{0.1477}

\newcommand{\WsTwoWhisperQwenLnAdamwWerCleanCell}{4.81\%}
\newcommand{\WsTwoWhisperQwenLnAdamwWerOtherCell}{10.70\%}

\newcommand{\WsTwoMoonshineQwenLnMuonParamsM}{0.64M}
\newcommand{\WsTwoMoonshineQwenLnMuonBestValCell}{0.2258}

\newcommand{\WsTwoMoonshineQwenLnMuonWerCleanCell}{7.14\%}
\newcommand{\WsTwoMoonshineQwenLnMuonWerOtherCell}{16.39\%}

\newcommand{\WsTwoMoonshineQwenLnMuonWerClean}{7.14}

\newcommand{\WsTwoMoonshineQwenLnAdamwParamsM}{0.64M}
\newcommand{\WsTwoMoonshineQwenLnAdamwBestValCell}{0.2297}

\newcommand{\WsTwoMoonshineQwenLnAdamwWerCleanCell}{7.38\%}
\newcommand{\WsTwoMoonshineQwenLnAdamwWerOtherCell}{16.44\%}

\newcommand{\WsTwoWhisperGemmaTwoLnMuonParamsM}{1.78M}
\newcommand{\WsTwoWhisperGemmaTwoLnMuonBestValCell}{0.0958}

\newcommand{\WsTwoWhisperGemmaTwoLnMuonWerCleanCell}{3.24\%}
\newcommand{\WsTwoWhisperGemmaTwoLnMuonWerOtherCell}{8.02\%}

\newcommand{\WsTwoWhisperGemmaTwoLnMuonWerClean}{3.24}
\newcommand{\WsTwoWhisperGemmaTwoLnMuonWerOther}{8.02}
\newcommand{\WsTwoWhisperGemmaTwoLnMuonEmptyClean}{0}
\newcommand{\WsTwoWhisperGemmaTwoLnMuonEmptyOther}{0}

\newcommand{\WsTwoWhisperGemmaTwoLnAdamwParamsM}{1.78M}
\newcommand{\WsTwoWhisperGemmaTwoLnAdamwBestValCell}{0.0974}

\newcommand{\WsTwoWhisperGemmaTwoLnAdamwWerCleanCell}{3.36\%}
\newcommand{\WsTwoWhisperGemmaTwoLnAdamwWerOtherCell}{7.84\%}

\newcommand{\WsTwoWhisperGemmaTwoLnAdamwWerClean}{3.36}
\newcommand{\WsTwoWhisperGemmaTwoLnAdamwWerOther}{7.84}
\newcommand{\WsTwoWhisperGemmaTwoLnAdamwEmptyClean}{0}
\newcommand{\WsTwoWhisperGemmaTwoLnAdamwEmptyOther}{0}

\newcommand{\WsTwoMoonshineGemmaTwoLnMuonParamsM}{0.97M}
\newcommand{\WsTwoMoonshineGemmaTwoLnMuonBestValCell}{0.1343}

\newcommand{\WsTwoMoonshineGemmaTwoLnMuonWerCleanCell}{4.49\%}
\newcommand{\WsTwoMoonshineGemmaTwoLnMuonWerOtherCell}{11.48\%}

\newcommand{\WsTwoMoonshineGemmaTwoLnMuonWerClean}{4.49}
\newcommand{\WsTwoMoonshineGemmaTwoLnMuonWerOther}{11.48}
\newcommand{\WsTwoMoonshineGemmaTwoLnMuonEmptyClean}{0}
\newcommand{\WsTwoMoonshineGemmaTwoLnMuonEmptyOther}{0}

\newcommand{\WsTwoMoonshineGemmaTwoLnAdamwParamsM}{0.97M}
\newcommand{\WsTwoMoonshineGemmaTwoLnAdamwBestValCell}{0.1386}

\newcommand{\WsTwoMoonshineGemmaTwoLnAdamwWerCleanCell}{4.62\%}
\newcommand{\WsTwoMoonshineGemmaTwoLnAdamwWerOtherCell}{11.69\%}

\newcommand{\WsTwoMoonshineGemmaTwoLnAdamwWerClean}{4.62}
\newcommand{\WsTwoMoonshineGemmaTwoLnAdamwWerOther}{11.69}
\newcommand{\WsTwoMoonshineGemmaTwoLnAdamwEmptyClean}{0}
\newcommand{\WsTwoMoonshineGemmaTwoLnAdamwEmptyOther}{0}

\newcommand{\WsTwoWhisperGemmaThreeLnMuonParamsM}{1.98M}
\newcommand{\WsTwoWhisperGemmaThreeLnMuonBestValCell}{0.0847}

\newcommand{\WsTwoWhisperGemmaThreeLnMuonWerCleanCell}{2.98\%}
\newcommand{\WsTwoWhisperGemmaThreeLnMuonWerOtherCell}{7.31\%}

\newcommand{\WsTwoWhisperGemmaThreeLnMuonWerClean}{2.98}

\newcommand{\WsTwoWhisperGemmaThreeLnAdamwParamsM}{1.98M}
\newcommand{\WsTwoWhisperGemmaThreeLnAdamwBestValCell}{0.0845}

\newcommand{\WsTwoWhisperGemmaThreeLnAdamwWerCleanCell}{3.09\%}
\newcommand{\WsTwoWhisperGemmaThreeLnAdamwWerOtherCell}{7.20\%}

\newcommand{\WsTwoMoonshineGemmaThreeLnMuonParamsM}{1.07M}
\newcommand{\WsTwoMoonshineGemmaThreeLnMuonBestValCell}{0.1112}

\newcommand{\WsTwoMoonshineGemmaThreeLnMuonWerCleanCell}{3.79\%}
\newcommand{\WsTwoMoonshineGemmaThreeLnMuonWerOtherCell}{10.16\%}

\newcommand{\WsTwoMoonshineGemmaThreeLnMuonWerClean}{3.79}
\newcommand{\WsTwoMoonshineGemmaThreeLnMuonWerOther}{10.16}

\newcommand{\WsTwoMoonshineGemmaThreeLnAdamwParamsM}{1.07M}
\newcommand{\WsTwoMoonshineGemmaThreeLnAdamwBestValCell}{0.1130}

\newcommand{\WsTwoMoonshineGemmaThreeLnAdamwWerCleanCell}{3.93\%}
\newcommand{\WsTwoMoonshineGemmaThreeLnAdamwWerOtherCell}{10.63\%}

\newcommand{\WsTwoWhisperGemmaTwoNoneMuonParamsM}{1.77M}
\newcommand{\WsTwoWhisperGemmaTwoNoneMuonBestValCell}{0.0983}

\newcommand{\WsTwoWhisperGemmaTwoNoneMuonWerCleanCell}{4.16\%}
\newcommand{\WsTwoWhisperGemmaTwoNoneMuonWerOtherCell}{9.06\%}

\newcommand{\WsTwoWhisperGemmaTwoScaledMuonParamsM}{1.77M}
\newcommand{\WsTwoWhisperGemmaTwoScaledMuonBestValCell}{0.0974}

\newcommand{\WsTwoWhisperGemmaTwoScaledMuonWerCleanCell}{3.27\%}
\newcommand{\WsTwoWhisperGemmaTwoScaledMuonWerOtherCell}{8.11\%}

\newcommand{\WsTwoWhisperGemmaTwoScaledMuonWerClean}{3.27}

\newcommand{\WsTwoWhisperQwenNoneMuonParamsM}{1.18M}
\newcommand{\WsTwoWhisperQwenNoneMuonBestValCell}{0.1295}

\newcommand{\WsTwoWhisperQwenNoneMuonWerCleanCell}{4.99\%}
\newcommand{\WsTwoWhisperQwenNoneMuonWerOtherCell}{10.17\%}

\newcommand{\WsTwoWhisperQwenNoneMuonWerClean}{4.99}

\newcommand{\WsTwoWhisperQwenLtwoMuonParamsM}{1.18M}
\newcommand{\WsTwoWhisperQwenLtwoMuonBestValCell}{0.1840}

\newcommand{\WsTwoWhisperQwenLtwoMuonWerCleanCell}{5.61\%}
\newcommand{\WsTwoWhisperQwenLtwoMuonWerOtherCell}{11.80\%}

\newcommand{\WsTwoWhisperQwenLtwoMuonWerClean}{5.61}

\newcommand{\WsTwoWhisperQwenScaledMuonParamsM}{1.18M}
\newcommand{\WsTwoWhisperQwenScaledMuonBestValCell}{0.1365}

\newcommand{\WsTwoWhisperQwenScaledMuonWerCleanCell}{4.30\%}
\newcommand{\WsTwoWhisperQwenScaledMuonWerOtherCell}{9.90\%}

\newcommand{\WsTwoWhisperQwenScaledMuonWerClean}{4.30}

\newcommand{\WsTwoWhisperQwenRmsMuonParamsM}{1.18M}
\newcommand{\WsTwoWhisperQwenRmsMuonBestValCell}{0.1425}

\newcommand{\WsTwoWhisperQwenRmsMuonWerCleanCell}{4.40\%}
\newcommand{\WsTwoWhisperQwenRmsMuonWerOtherCell}{9.99\%}

\newcommand{\WsTwoWhisperGemmaTwoLtwoMuonParamsM}{1.77M}
\newcommand{\WsTwoWhisperGemmaTwoLtwoMuonBestValCell}{0.1594}

\newcommand{\WsTwoWhisperGemmaTwoLtwoMuonWerCleanCell}{7.30\%}
\newcommand{\WsTwoWhisperGemmaTwoLtwoMuonWerOtherCell}{12.76\%}

\newcommand{\WsTwoWhisperGemmaTwoLtwoMuonWerClean}{7.30}
\newcommand{\WsTwoWhisperGemmaTwoLtwoMuonWerOther}{12.76}

\newcommand{\WsTwoWhisperGemmaTwoRmsMuonParamsM}{1.77M}
\newcommand{\WsTwoWhisperGemmaTwoRmsMuonBestValCell}{0.0944}

\newcommand{\WsTwoWhisperGemmaTwoRmsMuonWerCleanCell}{3.24\%}
\newcommand{\WsTwoWhisperGemmaTwoRmsMuonWerOtherCell}{7.93\%}

\newcommand{\WsTwoPairsDone}{6}
\newcommand{\WsTwoMuonWinsClean}{6}

\newcommand{\WsTwoMuonLowerOther}{3}
\newcommand{\WsTwoMuonTiedOther}{1}
\newcommand{\WsTwoMuonHigherOther}{2}

\newcommand{\WsTwoMuonMinErankPct}{90.1}
\newcommand{\WsTwoMuonMaxErankPct}{95.3}
\newcommand{\WsTwoAdamwMinErankPct}{81.7}
\newcommand{\WsTwoAdamwMaxErankPct}{89.4}
\newcommand{\WsTwoMoonshineMinKappaRatio}{2.7}
\newcommand{\WsTwoMoonshineMaxKappaRatio}{3.2}
\newcommand{\WsTwoWhisperMuonLowerKappa}{3}

\newcommand{\WsTwoBestMuonRunWerPairCell}{2.98\% / 7.31\%}  
\newcommand{\WsTwoMeanMuonDeltaClean}{-0.20}

\newcommand{\WsTwoNormDeltaWhisperGemmaTwoLtwoVsLnMuonClean}{+4.06}  

\newcommand{\WsThreeFitQwenToGemmaTwoProcAnchors}{83{,}378}
\newcommand{\WsThreeFitQwenToGemmaTwoProcTestN}{16{,}675}
\newcommand{\WsThreeFitQwenToGemmaTwoProcCos}{0.5048}
\newcommand{\WsThreeFitQwenToGemmaTwoProcPatOne}{91.68}
\newcommand{\WsThreeFitQwenToGemmaTwoProcRawRr}{0.013}
\newcommand{\WsThreeFitQwenToGemmaTwoProcNormRr}{1.008}
\newcommand{\WsThreeFitQwenToGemmaTwoLstsqAnchors}{83{,}378}
\newcommand{\WsThreeFitQwenToGemmaTwoLstsqTestN}{16{,}675}
\newcommand{\WsThreeFitQwenToGemmaTwoLstsqCos}{0.5916}
\newcommand{\WsThreeFitQwenToGemmaTwoLstsqPatOne}{79.41}
\newcommand{\WsThreeFitQwenToGemmaTwoLstsqRawRr}{0.615}
\newcommand{\WsThreeFitQwenToGemmaTwoLstsqNormRr}{1.008}
\newcommand{\WsThreeFitGemmaTwoToQwenProcAnchors}{83{,}378}
\newcommand{\WsThreeFitGemmaTwoToQwenProcTestN}{16{,}675}
\newcommand{\WsThreeFitGemmaTwoToQwenProcCos}{0.5645}
\newcommand{\WsThreeFitGemmaTwoToQwenProcPatOne}{83.09}
\newcommand{\WsThreeFitGemmaTwoToQwenProcRawRr}{70.48}
\newcommand{\WsThreeFitGemmaTwoToQwenProcNormRr}{1.000}
\newcommand{\WsThreeFitGemmaTwoToQwenLstsqAnchors}{83{,}378}
\newcommand{\WsThreeFitGemmaTwoToQwenLstsqTestN}{16{,}675}
\newcommand{\WsThreeFitGemmaTwoToQwenLstsqCos}{0.6283}
\newcommand{\WsThreeFitGemmaTwoToQwenLstsqPatOne}{84.57}
\newcommand{\WsThreeFitGemmaTwoToQwenLstsqRawRr}{0.661}
\newcommand{\WsThreeFitGemmaTwoToQwenLstsqNormRr}{1.000}

\newcommand{\WsThreeFitQwenToGemmaThreeProcAnchors}{82{,}663}
\newcommand{\WsThreeFitQwenToGemmaThreeProcTestN}{16{,}532}
\newcommand{\WsThreeFitQwenToGemmaThreeProcCos}{0.5010}
\newcommand{\WsThreeFitQwenToGemmaThreeProcPatOne}{96.72}
\newcommand{\WsThreeFitQwenToGemmaThreeProcRawRr}{0.020}
\newcommand{\WsThreeFitQwenToGemmaThreeProcNormRr}{1.000}
\newcommand{\WsThreeFitQwenToGemmaThreeLstsqAnchors}{82{,}663}
\newcommand{\WsThreeFitQwenToGemmaThreeLstsqTestN}{16{,}532}
\newcommand{\WsThreeFitQwenToGemmaThreeLstsqCos}{0.5515}
\newcommand{\WsThreeFitQwenToGemmaThreeLstsqPatOne}{96.95}
\newcommand{\WsThreeFitQwenToGemmaThreeLstsqRawRr}{0.585}
\newcommand{\WsThreeFitQwenToGemmaThreeLstsqNormRr}{1.000}
\newcommand{\WsThreeFitGemmaTwoToGemmaThreeProcAnchors}{163{,}983}
\newcommand{\WsThreeFitGemmaTwoToGemmaThreeProcTestN}{32{,}796}
\newcommand{\WsThreeFitGemmaTwoToGemmaThreeProcCos}{0.5080}
\newcommand{\WsThreeFitGemmaTwoToGemmaThreeProcPatOne}{97.47}
\newcommand{\WsThreeFitGemmaTwoToGemmaThreeProcRawRr}{1.644}
\newcommand{\WsThreeFitGemmaTwoToGemmaThreeProcNormRr}{1.000}
\newcommand{\WsThreeFitGemmaTwoToGemmaThreeLstsqAnchors}{163{,}983}
\newcommand{\WsThreeFitGemmaTwoToGemmaThreeLstsqTestN}{32{,}796}
\newcommand{\WsThreeFitGemmaTwoToGemmaThreeLstsqCos}{0.5768}
\newcommand{\WsThreeFitGemmaTwoToGemmaThreeLstsqPatOne}{98.56}
\newcommand{\WsThreeFitGemmaTwoToGemmaThreeLstsqRawRr}{0.603}
\newcommand{\WsThreeFitGemmaTwoToGemmaThreeLstsqNormRr}{1.000}

\newcommand{\WsFourFitQwenToGemmaTwoLnVal}{0.0101}
\newcommand{\WsFourFitQwenToGemmaTwoLnAnchors}{83{,}378}
\newcommand{\WsFourFitQwenToGemmaTwoLnTestN}{16{,}675}
\newcommand{\WsFourFitQwenToGemmaTwoLnCos}{0.1232}
\newcommand{\WsFourFitQwenToGemmaTwoLnPatOne}{50.46}
\newcommand{\WsFourFitQwenToGemmaTwoLnRr}{1.032}

\newcommand{\WsFourFitQwenToGemmaTwoScaledVal}{0.0119}
\newcommand{\WsFourFitQwenToGemmaTwoScaledAnchors}{83{,}378}
\newcommand{\WsFourFitQwenToGemmaTwoScaledTestN}{16{,}675}
\newcommand{\WsFourFitQwenToGemmaTwoScaledCos}{0.1022}
\newcommand{\WsFourFitQwenToGemmaTwoScaledPatOne}{52.14}
\newcommand{\WsFourFitQwenToGemmaTwoScaledRr}{1.000}

\newcommand{\WsFourFitGemmaTwoToQwenLnVal}{0.0183}
\newcommand{\WsFourFitGemmaTwoToQwenLnAnchors}{83{,}378}
\newcommand{\WsFourFitGemmaTwoToQwenLnTestN}{16{,}675}
\newcommand{\WsFourFitGemmaTwoToQwenLnCos}{0.1648}
\newcommand{\WsFourFitGemmaTwoToQwenLnPatOne}{13.05}
\newcommand{\WsFourFitGemmaTwoToQwenLnRr}{3.657}

\newcommand{\WsFourFitGemmaTwoToQwenScaledVal}{0.0185}
\newcommand{\WsFourFitGemmaTwoToQwenScaledAnchors}{83{,}378}
\newcommand{\WsFourFitGemmaTwoToQwenScaledTestN}{16{,}675}
\newcommand{\WsFourFitGemmaTwoToQwenScaledCos}{0.1557}
\newcommand{\WsFourFitGemmaTwoToQwenScaledPatOne}{33.66}
\newcommand{\WsFourFitGemmaTwoToQwenScaledRr}{3.406}

\newcommand{\WsFourFitQwenToGemmaThreeLnVal}{0.0062}
\newcommand{\WsFourFitQwenToGemmaThreeLnAnchors}{82{,}663}
\newcommand{\WsFourFitQwenToGemmaThreeLnTestN}{16{,}532}
\newcommand{\WsFourFitQwenToGemmaThreeLnCos}{0.1150}
\newcommand{\WsFourFitQwenToGemmaThreeLnPatOne}{48.14}
\newcommand{\WsFourFitQwenToGemmaThreeLnRr}{1.024}

\newcommand{\WsFourFitQwenToGemmaThreeScaledVal}{0.0097}
\newcommand{\WsFourFitQwenToGemmaThreeScaledAnchors}{82{,}663}
\newcommand{\WsFourFitQwenToGemmaThreeScaledTestN}{16{,}532}
\newcommand{\WsFourFitQwenToGemmaThreeScaledCos}{0.1113}
\newcommand{\WsFourFitQwenToGemmaThreeScaledPatOne}{53.90}
\newcommand{\WsFourFitQwenToGemmaThreeScaledRr}{1.003}

\newcommand{\WsFourFitGemmaTwoToGemmaThreeLnVal}{0.0043}
\newcommand{\WsFourFitGemmaTwoToGemmaThreeLnAnchors}{163{,}983}
\newcommand{\WsFourFitGemmaTwoToGemmaThreeLnTestN}{32{,}796}
\newcommand{\WsFourFitGemmaTwoToGemmaThreeLnCos}{0.1136}
\newcommand{\WsFourFitGemmaTwoToGemmaThreeLnPatOne}{36.11}
\newcommand{\WsFourFitGemmaTwoToGemmaThreeLnRr}{1.028}

\newcommand{\WsFourFitGemmaTwoToGemmaThreeScaledVal}{0.0065}
\newcommand{\WsFourFitGemmaTwoToGemmaThreeScaledAnchors}{163{,}983}
\newcommand{\WsFourFitGemmaTwoToGemmaThreeScaledTestN}{32{,}796}
\newcommand{\WsFourFitGemmaTwoToGemmaThreeScaledCos}{0.1125}
\newcommand{\WsFourFitGemmaTwoToGemmaThreeScaledPatOne}{37.47}
\newcommand{\WsFourFitGemmaTwoToGemmaThreeScaledRr}{1.005}

\newcommand{\WsFourBFitMoonshineQwenToGemmaTwoLnAudioVal}{0.1539}
\newcommand{\WsFourBFitMoonshineQwenToGemmaTwoLnTextVal}{5.61}
\newcommand{\WsFourBFitMoonshineQwenToGemmaTwoLnAnchors}{83{,}378}
\newcommand{\WsFourBFitMoonshineQwenToGemmaTwoLnTestN}{16{,}675}
\newcommand{\WsFourBFitMoonshineQwenToGemmaTwoLnCos}{0.0280}
\newcommand{\WsFourBFitMoonshineQwenToGemmaTwoLnPatOne}{0.63}
\newcommand{\WsFourBFitMoonshineQwenToGemmaTwoLnRr}{1.028}

\newcommand{\WsFourBFitWhisperQwenToGemmaTwoLnAudioVal}{0.1059}
\newcommand{\WsFourBFitWhisperQwenToGemmaTwoLnTextVal}{5.56}
\newcommand{\WsFourBFitWhisperQwenToGemmaTwoLnAnchors}{83{,}378}
\newcommand{\WsFourBFitWhisperQwenToGemmaTwoLnTestN}{16{,}675}
\newcommand{\WsFourBFitWhisperQwenToGemmaTwoLnCos}{0.0332}
\newcommand{\WsFourBFitWhisperQwenToGemmaTwoLnPatOne}{0.70}
\newcommand{\WsFourBFitWhisperQwenToGemmaTwoLnRr}{1.034}

\newcommand{\WsFourBFitMoonshineGemmaTwoToQwenLnAudioVal}{0.2931}
\newcommand{\WsFourBFitMoonshineGemmaTwoToQwenLnTextVal}{5.41}
\newcommand{\WsFourBFitMoonshineGemmaTwoToQwenLnAnchors}{83{,}378}
\newcommand{\WsFourBFitMoonshineGemmaTwoToQwenLnTestN}{16{,}675}
\newcommand{\WsFourBFitMoonshineGemmaTwoToQwenLnCos}{0.0244}
\newcommand{\WsFourBFitMoonshineGemmaTwoToQwenLnPatOne}{0.24}
\newcommand{\WsFourBFitMoonshineGemmaTwoToQwenLnRr}{8.368}

\newcommand{\WsFourBFitMoonshineGemmaTwoToQwenScaledAudioVal}{0.2962}
\newcommand{\WsFourBFitMoonshineGemmaTwoToQwenScaledTextVal}{5.24}
\newcommand{\WsFourBFitMoonshineGemmaTwoToQwenScaledAnchors}{83{,}378}
\newcommand{\WsFourBFitMoonshineGemmaTwoToQwenScaledTestN}{16{,}675}
\newcommand{\WsFourBFitMoonshineGemmaTwoToQwenScaledCos}{0.0363}
\newcommand{\WsFourBFitMoonshineGemmaTwoToQwenScaledPatOne}{0.47}
\newcommand{\WsFourBFitMoonshineGemmaTwoToQwenScaledRr}{7.592}

\newcommand{\WsFourBFitWhisperGemmaTwoToQwenLnAudioVal}{0.1736}
\newcommand{\WsFourBFitWhisperGemmaTwoToQwenLnTextVal}{5.24}
\newcommand{\WsFourBFitWhisperGemmaTwoToQwenLnAnchors}{83{,}378}
\newcommand{\WsFourBFitWhisperGemmaTwoToQwenLnTestN}{16{,}675}
\newcommand{\WsFourBFitWhisperGemmaTwoToQwenLnCos}{0.0437}
\newcommand{\WsFourBFitWhisperGemmaTwoToQwenLnPatOne}{0.41}
\newcommand{\WsFourBFitWhisperGemmaTwoToQwenLnRr}{6.891}

\newcommand{\WsFourBFitMoonshineQwenToGemmaThreeLnAudioVal}{0.1279}
\newcommand{\WsFourBFitMoonshineQwenToGemmaThreeLnTextVal}{7.29}
\newcommand{\WsFourBFitMoonshineQwenToGemmaThreeLnAnchors}{82{,}663}
\newcommand{\WsFourBFitMoonshineQwenToGemmaThreeLnTestN}{16{,}532}
\newcommand{\WsFourBFitMoonshineQwenToGemmaThreeLnCos}{0.0254}
\newcommand{\WsFourBFitMoonshineQwenToGemmaThreeLnPatOne}{0.50}
\newcommand{\WsFourBFitMoonshineQwenToGemmaThreeLnRr}{1.047}

\newcommand{\WsFourBFitWhisperQwenToGemmaThreeLnAudioVal}{0.0889}
\newcommand{\WsFourBFitWhisperQwenToGemmaThreeLnTextVal}{7.22}
\newcommand{\WsFourBFitWhisperQwenToGemmaThreeLnAnchors}{82{,}663}
\newcommand{\WsFourBFitWhisperQwenToGemmaThreeLnTestN}{16{,}532}
\newcommand{\WsFourBFitWhisperQwenToGemmaThreeLnCos}{0.0296}
\newcommand{\WsFourBFitWhisperQwenToGemmaThreeLnPatOne}{0.74}
\newcommand{\WsFourBFitWhisperQwenToGemmaThreeLnRr}{1.051}

\newcommand{\WsFourBFitMoonshineGemmaTwoToGemmaThreeLnAudioVal}{0.1224}
\newcommand{\WsFourBFitMoonshineGemmaTwoToGemmaThreeLnTextVal}{6.92}
\newcommand{\WsFourBFitMoonshineGemmaTwoToGemmaThreeLnAnchors}{163{,}983}
\newcommand{\WsFourBFitMoonshineGemmaTwoToGemmaThreeLnTestN}{32{,}796}
\newcommand{\WsFourBFitMoonshineGemmaTwoToGemmaThreeLnCos}{0.0339}
\newcommand{\WsFourBFitMoonshineGemmaTwoToGemmaThreeLnPatOne}{0.83}
\newcommand{\WsFourBFitMoonshineGemmaTwoToGemmaThreeLnRr}{1.058}

\newcommand{\WsFourBFitWhisperGemmaTwoToGemmaThreeLnAudioVal}{0.0903}
\newcommand{\WsFourBFitWhisperGemmaTwoToGemmaThreeLnTextVal}{7.44}
\newcommand{\WsFourBFitWhisperGemmaTwoToGemmaThreeLnAnchors}{163{,}983}
\newcommand{\WsFourBFitWhisperGemmaTwoToGemmaThreeLnTestN}{32{,}796}
\newcommand{\WsFourBFitWhisperGemmaTwoToGemmaThreeLnCos}{0.0387}
\newcommand{\WsFourBFitWhisperGemmaTwoToGemmaThreeLnPatOne}{1.27}
\newcommand{\WsFourBFitWhisperGemmaTwoToGemmaThreeLnRr}{1.053}

\newcommand{\WsFourCWikiFitQwenToGemmaTwoLnWikiVal}{0.0145}
\newcommand{\WsFourCWikiFitQwenToGemmaTwoLnLsVal}{0.1716}
\newcommand{\WsFourCWikiFitQwenToGemmaTwoLnAnchors}{83{,}378}
\newcommand{\WsFourCWikiFitQwenToGemmaTwoLnTestN}{16{,}675}
\newcommand{\WsFourCWikiFitQwenToGemmaTwoLnCos}{0.1383}
\newcommand{\WsFourCWikiFitQwenToGemmaTwoLnPatOne}{64.03}
\newcommand{\WsFourCWikiFitQwenToGemmaTwoLnRr}{1.031}

\newcommand{\WsFourCWikiFitGemmaTwoToQwenLnWikiVal}{0.0182}
\newcommand{\WsFourCWikiFitGemmaTwoToQwenLnLsVal}{0.2760}
\newcommand{\WsFourCWikiFitGemmaTwoToQwenLnAnchors}{83{,}378}
\newcommand{\WsFourCWikiFitGemmaTwoToQwenLnTestN}{16{,}675}
\newcommand{\WsFourCWikiFitGemmaTwoToQwenLnCos}{0.1621}
\newcommand{\WsFourCWikiFitGemmaTwoToQwenLnPatOne}{33.21}
\newcommand{\WsFourCWikiFitGemmaTwoToQwenLnRr}{4.419}

\newcommand{\WsFourCWikiFitQwenToGemmaThreeLnWikiVal}{0.0105}
\newcommand{\WsFourCWikiFitQwenToGemmaThreeLnLsVal}{0.2533}
\newcommand{\WsFourCWikiFitQwenToGemmaThreeLnAnchors}{82{,}663}
\newcommand{\WsFourCWikiFitQwenToGemmaThreeLnTestN}{16{,}532}
\newcommand{\WsFourCWikiFitQwenToGemmaThreeLnCos}{0.1358}
\newcommand{\WsFourCWikiFitQwenToGemmaThreeLnPatOne}{68.19}
\newcommand{\WsFourCWikiFitQwenToGemmaThreeLnRr}{1.017}

\newcommand{\WsFourCWikiFitGemmaTwoToGemmaThreeLnWikiVal}{0.0093}
\newcommand{\WsFourCWikiFitGemmaTwoToGemmaThreeLnLsVal}{0.1800}
\newcommand{\WsFourCWikiFitGemmaTwoToGemmaThreeLnAnchors}{163{,}983}
\newcommand{\WsFourCWikiFitGemmaTwoToGemmaThreeLnTestN}{32{,}796}
\newcommand{\WsFourCWikiFitGemmaTwoToGemmaThreeLnCos}{0.1297}
\newcommand{\WsFourCWikiFitGemmaTwoToGemmaThreeLnPatOne}{52.91}
\newcommand{\WsFourCWikiFitGemmaTwoToGemmaThreeLnRr}{1.030}

\newcommand{\WsFourCMixFitMoonshineQwenToGemmaTwoLnWikiVal}{0.0154}
\newcommand{\WsFourCMixFitMoonshineQwenToGemmaTwoLnLsVal}{0.2171}
\newcommand{\WsFourCMixFitMoonshineQwenToGemmaTwoLnAudioVal}{0.1502}
\newcommand{\WsFourCMixFitMoonshineQwenToGemmaTwoLnAnchors}{83{,}378}
\newcommand{\WsFourCMixFitMoonshineQwenToGemmaTwoLnTestN}{16{,}675}
\newcommand{\WsFourCMixFitMoonshineQwenToGemmaTwoLnCos}{0.1260}
\newcommand{\WsFourCMixFitMoonshineQwenToGemmaTwoLnPatOne}{57.84}
\newcommand{\WsFourCMixFitMoonshineQwenToGemmaTwoLnRr}{1.045}

\newcommand{\WsFourCMixFitWhisperQwenToGemmaTwoLnWikiVal}{0.0155}
\newcommand{\WsFourCMixFitWhisperQwenToGemmaTwoLnLsVal}{0.1910}
\newcommand{\WsFourCMixFitWhisperQwenToGemmaTwoLnAudioVal}{0.1028}
\newcommand{\WsFourCMixFitWhisperQwenToGemmaTwoLnAnchors}{83{,}378}
\newcommand{\WsFourCMixFitWhisperQwenToGemmaTwoLnTestN}{16{,}675}
\newcommand{\WsFourCMixFitWhisperQwenToGemmaTwoLnCos}{0.1212}
\newcommand{\WsFourCMixFitWhisperQwenToGemmaTwoLnPatOne}{56.03}
\newcommand{\WsFourCMixFitWhisperQwenToGemmaTwoLnRr}{1.056}

\newcommand{\WsFourCMixFitMoonshineGemmaTwoToQwenLnWikiVal}{0.0211}
\newcommand{\WsFourCMixFitMoonshineGemmaTwoToQwenLnLsVal}{0.3200}
\newcommand{\WsFourCMixFitMoonshineGemmaTwoToQwenLnAudioVal}{0.2599}
\newcommand{\WsFourCMixFitMoonshineGemmaTwoToQwenLnAnchors}{83{,}378}
\newcommand{\WsFourCMixFitMoonshineGemmaTwoToQwenLnTestN}{16{,}675}
\newcommand{\WsFourCMixFitMoonshineGemmaTwoToQwenLnCos}{0.1243}
\newcommand{\WsFourCMixFitMoonshineGemmaTwoToQwenLnPatOne}{31.83}
\newcommand{\WsFourCMixFitMoonshineGemmaTwoToQwenLnRr}{9.562}

\newcommand{\WsFourCMixFitWhisperGemmaTwoToQwenLnWikiVal}{0.0229}
\newcommand{\WsFourCMixFitWhisperGemmaTwoToQwenLnLsVal}{0.3238}
\newcommand{\WsFourCMixFitWhisperGemmaTwoToQwenLnAudioVal}{0.1623}
\newcommand{\WsFourCMixFitWhisperGemmaTwoToQwenLnAnchors}{83{,}378}
\newcommand{\WsFourCMixFitWhisperGemmaTwoToQwenLnTestN}{16{,}675}
\newcommand{\WsFourCMixFitWhisperGemmaTwoToQwenLnCos}{0.1277}
\newcommand{\WsFourCMixFitWhisperGemmaTwoToQwenLnPatOne}{25.24}
\newcommand{\WsFourCMixFitWhisperGemmaTwoToQwenLnRr}{8.661}

\newcommand{\WsFourCMixFitMoonshineQwenToGemmaThreeLnWikiVal}{0.0114}
\newcommand{\WsFourCMixFitMoonshineQwenToGemmaThreeLnLsVal}{0.3510}
\newcommand{\WsFourCMixFitMoonshineQwenToGemmaThreeLnAudioVal}{0.1220}
\newcommand{\WsFourCMixFitMoonshineQwenToGemmaThreeLnAnchors}{82{,}663}
\newcommand{\WsFourCMixFitMoonshineQwenToGemmaThreeLnTestN}{16{,}532}
\newcommand{\WsFourCMixFitMoonshineQwenToGemmaThreeLnCos}{0.1213}
\newcommand{\WsFourCMixFitMoonshineQwenToGemmaThreeLnPatOne}{59.51}
\newcommand{\WsFourCMixFitMoonshineQwenToGemmaThreeLnRr}{1.074}

\newcommand{\WsFourCMixFitWhisperQwenToGemmaThreeLnWikiVal}{0.0112}
\newcommand{\WsFourCMixFitWhisperQwenToGemmaThreeLnLsVal}{0.3100}
\newcommand{\WsFourCMixFitWhisperQwenToGemmaThreeLnAudioVal}{0.0887}
\newcommand{\WsFourCMixFitWhisperQwenToGemmaThreeLnAnchors}{82{,}663}
\newcommand{\WsFourCMixFitWhisperQwenToGemmaThreeLnTestN}{16{,}532}
\newcommand{\WsFourCMixFitWhisperQwenToGemmaThreeLnCos}{0.1160}
\newcommand{\WsFourCMixFitWhisperQwenToGemmaThreeLnPatOne}{58.98}
\newcommand{\WsFourCMixFitWhisperQwenToGemmaThreeLnRr}{1.076}

\newcommand{\WsFourCMixFitMoonshineGemmaTwoToGemmaThreeLnWikiVal}{0.0101}
\newcommand{\WsFourCMixFitMoonshineGemmaTwoToGemmaThreeLnLsVal}{0.1422}
\newcommand{\WsFourCMixFitMoonshineGemmaTwoToGemmaThreeLnAudioVal}{0.1219}
\newcommand{\WsFourCMixFitMoonshineGemmaTwoToGemmaThreeLnAnchors}{163{,}983}
\newcommand{\WsFourCMixFitMoonshineGemmaTwoToGemmaThreeLnTestN}{32{,}796}
\newcommand{\WsFourCMixFitMoonshineGemmaTwoToGemmaThreeLnCos}{0.1119}
\newcommand{\WsFourCMixFitMoonshineGemmaTwoToGemmaThreeLnPatOne}{44.61}
\newcommand{\WsFourCMixFitMoonshineGemmaTwoToGemmaThreeLnRr}{1.080}

\newcommand{\WsFourCMixFitWhisperGemmaTwoToGemmaThreeLnWikiVal}{0.0098}
\newcommand{\WsFourCMixFitWhisperGemmaTwoToGemmaThreeLnLsVal}{0.1873}
\newcommand{\WsFourCMixFitWhisperGemmaTwoToGemmaThreeLnAudioVal}{0.0888}
\newcommand{\WsFourCMixFitWhisperGemmaTwoToGemmaThreeLnAnchors}{163{,}983}
\newcommand{\WsFourCMixFitWhisperGemmaTwoToGemmaThreeLnTestN}{32{,}796}
\newcommand{\WsFourCMixFitWhisperGemmaTwoToGemmaThreeLnCos}{0.1082}
\newcommand{\WsFourCMixFitWhisperGemmaTwoToGemmaThreeLnPatOne}{44.66}
\newcommand{\WsFourCMixFitWhisperGemmaTwoToGemmaThreeLnRr}{1.089}
\newcommand{\WsFourCWikiMinVal}{0.0093}
\newcommand{\WsFourCWikiMaxVal}{0.0182}
\newcommand{\WsFourCWikiMinLsVal}{0.1716}
\newcommand{\WsFourCWikiMaxLsVal}{0.2760}

\newcommand{\WsFourCWikiMinPatOne}{33.21}
\newcommand{\WsFourCWikiMaxPatOne}{68.19}
\newcommand{\WsFourCMixMinWikiVal}{0.0098}
\newcommand{\WsFourCMixMaxWikiVal}{0.0229}

\newcommand{\WsFourMinVal}{0.0043}
\newcommand{\WsFourMaxVal}{0.0185}
\newcommand{\WsFourBMinAudioVal}{0.0889}
\newcommand{\WsFourBMaxAudioVal}{0.2962}
\newcommand{\WsFourBMinTextVal}{5.24}
\newcommand{\WsFourBMaxTextVal}{7.44}

\newcommand{\WsThreeCompWhisperQwenToGemmaTwoProcNonePairCell}{251.7\% / 287.4\%}

\newcommand{\WsThreeCompWhisperQwenToGemmaTwoProcScaledPairCell}{345.8\% / 409.8\%}

\newcommand{\WsThreeCompWhisperQwenToGemmaTwoProcLnPairCell}{277.1\% / 318.3\%}

\newcommand{\WsThreeCompWhisperQwenToGemmaTwoLstsqNonePairCell}{111.8\% / 119.3\%}

\newcommand{\WsThreeCompWhisperQwenToGemmaTwoLstsqScaledPairCell}{359.9\% / 415.6\%}

\newcommand{\WsThreeCompWhisperQwenToGemmaTwoLstsqLnPairCell}{253.6\% / 281.8\%}

\newcommand{\WsFourCompWhisperQwenToGemmaTwoScaledPairCell}{490.5\% / 555.2\%}

\newcommand{\WsFourCompWhisperQwenToGemmaTwoLnPairCell}{477.1\% / 532.7\%}

\newcommand{\WsThreeCompWhisperGemmaTwoToQwenProcNonePairCell}{196.1\% / 215.3\%}

\newcommand{\WsThreeCompWhisperGemmaTwoToQwenProcScaledPairCell}{154.7\% / 167.5\%}

\newcommand{\WsThreeCompWhisperGemmaTwoToQwenProcLnPairCell}{134.8\% / 141.1\%}

\newcommand{\WsThreeCompWhisperGemmaTwoToQwenLstsqNonePairCell}{147.2\% / 153.2\%}

\newcommand{\WsThreeCompWhisperGemmaTwoToQwenLstsqScaledPairCell}{147.8\% / 154.5\%}

\newcommand{\WsThreeCompWhisperGemmaTwoToQwenLstsqLnPairCell}{118.1\% / 123.9\%}

\newcommand{\WsFourCompWhisperGemmaTwoToQwenScaledPairCell}{100.5\% / 100.8\%}

\newcommand{\WsFourCompWhisperGemmaTwoToQwenLnPairCell}{101.4\% / 102.1\%}

\newcommand{\WsThreeCompWhisperQwenToGemmaThreeProcNonePairCell}{194.3\% / 231.5\%}

\newcommand{\WsThreeCompWhisperQwenToGemmaThreeProcScaledPairCell}{225.6\% / 272.4\%}

\newcommand{\WsThreeCompWhisperQwenToGemmaThreeProcLnPairCell}{377.3\% / 440.9\%}

\newcommand{\WsThreeCompWhisperQwenToGemmaThreeLstsqNonePairCell}{406.4\% / 463.7\%}

\newcommand{\WsThreeCompWhisperQwenToGemmaThreeLstsqScaledPairCell}{320.3\% / 361.3\%}

\newcommand{\WsThreeCompWhisperQwenToGemmaThreeLstsqLnPairCell}{399.9\% / 446.5\%}

\newcommand{\WsFourCompWhisperQwenToGemmaThreeScaledPairCell}{317.6\% / 359.3\%}

\newcommand{\WsFourCompWhisperQwenToGemmaThreeLnPairCell}{442.8\% / 506.3\%}

\newcommand{\WsThreeCompWhisperGemmaTwoToGemmaThreeProcNonePairCell}{360.9\% / 400.2\%}

\newcommand{\WsThreeCompWhisperGemmaTwoToGemmaThreeProcScaledPairCell}{446.0\% / 513.2\%}

\newcommand{\WsThreeCompWhisperGemmaTwoToGemmaThreeProcLnPairCell}{413.8\% / 472.0\%}

\newcommand{\WsThreeCompWhisperGemmaTwoToGemmaThreeLstsqNonePairCell}{257.4\% / 295.2\%}

\newcommand{\WsThreeCompWhisperGemmaTwoToGemmaThreeLstsqScaledPairCell}{263.7\% / 287.4\%}
\newcommand{\WsThreeCompWhisperGemmaTwoToGemmaThreeLstsqScaledClean}{263.67}

\newcommand{\WsThreeCompWhisperGemmaTwoToGemmaThreeLstsqLnPairCell}{273.1\% / 301.0\%}
\newcommand{\WsThreeCompWhisperGemmaTwoToGemmaThreeLstsqLnClean}{273.06}

\newcommand{\WsFourCompWhisperGemmaTwoToGemmaThreeScaledPairCell}{400.7\% / 456.9\%}

\newcommand{\WsFourCompWhisperGemmaTwoToGemmaThreeLnPairCell}{374.9\% / 427.8\%}
\newcommand{\WsFourCompWhisperGemmaTwoToGemmaThreeLnClean}{374.90}

\newcommand{\WsThreeCompMoonshineQwenToGemmaTwoProcNonePairCell}{266.7\% / 306.9\%}

\newcommand{\WsThreeCompMoonshineQwenToGemmaTwoProcScaledPairCell}{291.2\% / 353.5\%}

\newcommand{\WsThreeCompMoonshineQwenToGemmaTwoProcLnPairCell}{476.4\% / 533.4\%}

\newcommand{\WsThreeCompMoonshineQwenToGemmaTwoLstsqNonePairCell}{254.0\% / 307.8\%}

\newcommand{\WsThreeCompMoonshineQwenToGemmaTwoLstsqScaledPairCell}{325.9\% / 368.9\%}

\newcommand{\WsThreeCompMoonshineQwenToGemmaTwoLstsqLnPairCell}{284.5\% / 329.7\%}

\newcommand{\WsFourCompMoonshineQwenToGemmaTwoScaledPairCell}{189.5\% / 181.2\%}

\newcommand{\WsFourCompMoonshineQwenToGemmaTwoLnPairCell}{341.9\% / 390.0\%}

\newcommand{\WsThreeCompMoonshineGemmaTwoToQwenProcNonePairCell}{237.8\% / 251.2\%}

\newcommand{\WsThreeCompMoonshineGemmaTwoToQwenProcScaledPairCell}{159.4\% / 175.4\%}

\newcommand{\WsThreeCompMoonshineGemmaTwoToQwenProcLnPairCell}{183.1\% / 201.4\%}

\newcommand{\WsThreeCompMoonshineGemmaTwoToQwenLstsqNonePairCell}{131.0\% / 140.1\%}

\newcommand{\WsThreeCompMoonshineGemmaTwoToQwenLstsqScaledPairCell}{130.7\% / 140.5\%}

\newcommand{\WsThreeCompMoonshineGemmaTwoToQwenLstsqLnPairCell}{108.3\% / 113.6\%}

\newcommand{\WsFourCompMoonshineGemmaTwoToQwenScaledPairCell}{116.3\% / 114.5\%}

\newcommand{\WsFourCompMoonshineGemmaTwoToQwenLnCleanCell}{100.70\%}
\newcommand{\WsFourCompMoonshineGemmaTwoToQwenLnOtherCell}{102.22\%}
\newcommand{\WsFourCompMoonshineGemmaTwoToQwenLnPairCell}{100.7\% / 102.2\%}

\newcommand{\WsThreeCompMoonshineGemmaTwoToQwenRandScaledClean}{122.42}

\newcommand{\WsThreeCompMoonshineQwenToGemmaThreeProcNonePairCell}{304.6\% / 337.9\%}

\newcommand{\WsThreeCompMoonshineQwenToGemmaThreeProcScaledPairCell}{346.9\% / 403.1\%}

\newcommand{\WsThreeCompMoonshineQwenToGemmaThreeProcLnPairCell}{415.4\% / 465.4\%}

\newcommand{\WsThreeCompMoonshineQwenToGemmaThreeLstsqNonePairCell}{191.2\% / 215.8\%}

\newcommand{\WsThreeCompMoonshineQwenToGemmaThreeLstsqScaledPairCell}{173.3\% / 191.9\%}

\newcommand{\WsThreeCompMoonshineQwenToGemmaThreeLstsqLnPairCell}{383.5\% / 431.9\%}

\newcommand{\WsFourCompMoonshineQwenToGemmaThreeScaledPairCell}{236.6\% / 270.3\%}

\newcommand{\WsFourCompMoonshineQwenToGemmaThreeLnPairCell}{407.3\% / 458.7\%}

\newcommand{\WsThreeCompMoonshineGemmaTwoToGemmaThreeProcNonePairCell}{326.4\% / 356.6\%}

\newcommand{\WsThreeCompMoonshineGemmaTwoToGemmaThreeProcScaledPairCell}{417.3\% / 475.5\%}

\newcommand{\WsThreeCompMoonshineGemmaTwoToGemmaThreeProcLnPairCell}{336.2\% / 375.9\%}

\newcommand{\WsThreeCompMoonshineGemmaTwoToGemmaThreeLstsqNonePairCell}{310.5\% / 359.4\%}

\newcommand{\WsThreeCompMoonshineGemmaTwoToGemmaThreeLstsqScaledPairCell}{331.4\% / 381.4\%}
\newcommand{\WsThreeCompMoonshineGemmaTwoToGemmaThreeLstsqScaledClean}{331.45}

\newcommand{\WsThreeCompMoonshineGemmaTwoToGemmaThreeLstsqLnPairCell}{295.9\% / 327.1\%}

\newcommand{\WsFourCompMoonshineGemmaTwoToGemmaThreeScaledPairCell}{388.8\% / 447.6\%}

\newcommand{\WsFourCompMoonshineGemmaTwoToGemmaThreeLnPairCell}{394.5\% / 443.3\%}
\newcommand{\WsFourCompMoonshineGemmaTwoToGemmaThreeLnClean}{394.50}

\newcommand{\WsThreeCompMoonshineGemmaTwoToGemmaThreeRandScaledClean}{308.73}

\newcommand{\ClosedFormCompBestClean}{108.32}

\newcommand{\WsFourBCompInDistMoonshineQwenToGemmaTwoLnCleanCell}{5.10\%}
\newcommand{\WsFourBCompInDistMoonshineQwenToGemmaTwoLnOtherCell}{12.45\%}
\newcommand{\WsFourBCompInDistMoonshineQwenToGemmaTwoLnPairCell}{5.1\% / 12.5\%}

\newcommand{\WsFourBCompInDistWhisperQwenToGemmaTwoLnCleanCell}{3.56\%}
\newcommand{\WsFourBCompInDistWhisperQwenToGemmaTwoLnOtherCell}{8.45\%}
\newcommand{\WsFourBCompInDistWhisperQwenToGemmaTwoLnPairCell}{3.6\% / 8.4\%}

\newcommand{\WsFourBCompInDistMoonshineGemmaTwoToQwenLnCleanCell}{9.04\%}
\newcommand{\WsFourBCompInDistMoonshineGemmaTwoToQwenLnOtherCell}{18.91\%}
\newcommand{\WsFourBCompInDistMoonshineGemmaTwoToQwenLnPairCell}{9.0\% / 18.9\%}
\newcommand{\WsFourBCompInDistMoonshineGemmaTwoToQwenLnClean}{9.04}

\newcommand{\WsFourBCompInDistMoonshineGemmaTwoToQwenScaledCleanCell}{9.03\%}
\newcommand{\WsFourBCompInDistMoonshineGemmaTwoToQwenScaledOtherCell}{18.79\%}

\newcommand{\WsFourBCompInDistWhisperGemmaTwoToQwenLnCleanCell}{5.12\%}
\newcommand{\WsFourBCompInDistWhisperGemmaTwoToQwenLnOtherCell}{11.28\%}
\newcommand{\WsFourBCompInDistWhisperGemmaTwoToQwenLnPairCell}{5.1\% / 11.3\%}

\newcommand{\WsFourBCompInDistMoonshineQwenToGemmaThreeLnCleanCell}{4.35\%}
\newcommand{\WsFourBCompInDistMoonshineQwenToGemmaThreeLnOtherCell}{10.91\%}
\newcommand{\WsFourBCompInDistMoonshineQwenToGemmaThreeLnPairCell}{4.3\% / 10.9\%}

\newcommand{\WsFourBCompInDistWhisperQwenToGemmaThreeLnCleanCell}{3.24\%}
\newcommand{\WsFourBCompInDistWhisperQwenToGemmaThreeLnOtherCell}{7.78\%}
\newcommand{\WsFourBCompInDistWhisperQwenToGemmaThreeLnPairCell}{3.2\% / 7.8\%}

\newcommand{\WsFourBCompInDistMoonshineGemmaTwoToGemmaThreeLnCleanCell}{4.29\%}
\newcommand{\WsFourBCompInDistMoonshineGemmaTwoToGemmaThreeLnOtherCell}{10.85\%}
\newcommand{\WsFourBCompInDistMoonshineGemmaTwoToGemmaThreeLnPairCell}{4.3\% / 10.9\%}
\newcommand{\WsFourBCompInDistMoonshineGemmaTwoToGemmaThreeLnClean}{4.29}
\newcommand{\WsFourBCompInDistMoonshineGemmaTwoToGemmaThreeLnOther}{10.85}
\newcommand{\WsFourBCompInDistMoonshineGemmaTwoToGemmaThreeLnEmptyClean}{0}
\newcommand{\WsFourBCompInDistMoonshineGemmaTwoToGemmaThreeLnEmptyOther}{0}
\newcommand{\WsFourBCompInDistWhisperGemmaTwoToGemmaThreeLnCleanCell}{3.22\%}
\newcommand{\WsFourBCompInDistWhisperGemmaTwoToGemmaThreeLnOtherCell}{7.53\%}
\newcommand{\WsFourBCompInDistWhisperGemmaTwoToGemmaThreeLnPairCell}{3.2\% / 7.5\%}
\newcommand{\WsFourBCompInDistWhisperGemmaTwoToGemmaThreeLnClean}{3.22}
\newcommand{\WsFourBCompInDistWhisperGemmaTwoToGemmaThreeLnOther}{7.53}
\newcommand{\WsFourBCompInDistWhisperGemmaTwoToGemmaThreeLnEmptyClean}{0}
\newcommand{\WsFourBCompInDistWhisperGemmaTwoToGemmaThreeLnEmptyOther}{0}
\newcommand{\WsFourBCompCrossWhisperOnMoonshineQwenToGemmaTwoLnCleanCell}{476.67\%}
\newcommand{\WsFourBCompCrossWhisperOnMoonshineQwenToGemmaTwoLnOtherCell}{536.30\%}
\newcommand{\WsFourBCompCrossWhisperOnMoonshineQwenToGemmaTwoLnPairCell}{476.7\% / 536.3\%}

\newcommand{\WsFourBCompCrossMoonshineOnWhisperQwenToGemmaTwoLnCleanCell}{409.20\%}
\newcommand{\WsFourBCompCrossMoonshineOnWhisperQwenToGemmaTwoLnOtherCell}{443.52\%}
\newcommand{\WsFourBCompCrossMoonshineOnWhisperQwenToGemmaTwoLnPairCell}{409.2\% / 443.5\%}

\newcommand{\WsFourBCompCrossWhisperOnMoonshineGemmaTwoToQwenLnCleanCell}{98.63\%}
\newcommand{\WsFourBCompCrossWhisperOnMoonshineGemmaTwoToQwenLnOtherCell}{107.15\%}
\newcommand{\WsFourBCompCrossWhisperOnMoonshineGemmaTwoToQwenLnPairCell}{98.6\% / 107.2\%}

\newcommand{\WsFourBCompCrossMoonshineOnWhisperGemmaTwoToQwenLnCleanCell}{123.83\%}
\newcommand{\WsFourBCompCrossMoonshineOnWhisperGemmaTwoToQwenLnOtherCell}{130.33\%}
\newcommand{\WsFourBCompCrossMoonshineOnWhisperGemmaTwoToQwenLnPairCell}{123.8\% / 130.3\%}

\newcommand{\WsFourBCompCrossWhisperOnMoonshineQwenToGemmaThreeLnCleanCell}{429.37\%}
\newcommand{\WsFourBCompCrossWhisperOnMoonshineQwenToGemmaThreeLnOtherCell}{475.50\%}
\newcommand{\WsFourBCompCrossWhisperOnMoonshineQwenToGemmaThreeLnPairCell}{429.4\% / 475.5\%}

\newcommand{\WsFourBCompCrossMoonshineOnWhisperQwenToGemmaThreeLnCleanCell}{463.31\%}
\newcommand{\WsFourBCompCrossMoonshineOnWhisperQwenToGemmaThreeLnOtherCell}{516.54\%}
\newcommand{\WsFourBCompCrossMoonshineOnWhisperQwenToGemmaThreeLnPairCell}{463.3\% / 516.5\%}

\newcommand{\WsFourBCompCrossWhisperOnMoonshineGemmaTwoToGemmaThreeLnCleanCell}{67.35\%}
\newcommand{\WsFourBCompCrossWhisperOnMoonshineGemmaTwoToGemmaThreeLnOtherCell}{77.47\%}
\newcommand{\WsFourBCompCrossWhisperOnMoonshineGemmaTwoToGemmaThreeLnPairCell}{67.4\% / 77.5\%}
\newcommand{\WsFourBCompCrossWhisperOnMoonshineGemmaTwoToGemmaThreeLnClean}{67.35}

\newcommand{\WsFourBCompCrossWhisperOnMoonshineGemmaTwoToGemmaThreeLnEmptyClean}{361}

\newcommand{\WsFourBCompCrossMoonshineOnWhisperGemmaTwoToGemmaThreeLnCleanCell}{75.35\%}
\newcommand{\WsFourBCompCrossMoonshineOnWhisperGemmaTwoToGemmaThreeLnOtherCell}{89.82\%}
\newcommand{\WsFourBCompCrossMoonshineOnWhisperGemmaTwoToGemmaThreeLnPairCell}{75.4\% / 89.8\%}
\newcommand{\WsFourBCompCrossMoonshineOnWhisperGemmaTwoToGemmaThreeLnClean}{75.35}

\newcommand{\WsFourBCompCrossMoonshineOnWhisperGemmaTwoToGemmaThreeLnEmptyClean}{0}

\newcommand{\WsFourBCompBestClean}{3.22}
\newcommand{\WsFourBCompBestOther}{7.53}
\newcommand{\WsFourBInDistMinClean}{3.22}
\newcommand{\WsFourBInDistMaxClean}{9.04}

\newcommand{\WsFourBInDistMaxOther}{18.91}
\newcommand{\WsFourBCrossMinClean}{67.35}
\newcommand{\WsFourBCrossMaxClean}{476.67}
\newcommand{\WsFourBNoiseHalfVal}{0.0192}
\newcommand{\WsFourBNoiseHalfCos}{0.1532}
\newcommand{\WsFourBNoiseHalfPatOne}{21.47}

\newcommand{\WsFourBNoiseHalfCleanCell}{110.06\%}
\newcommand{\WsFourBNoiseHalfOtherCell}{112.18\%}

\newcommand{\WsFourBNoiseOneVal}{0.0254}
\newcommand{\WsFourBNoiseOneCos}{0.1440}
\newcommand{\WsFourBNoiseOnePatOne}{25.14}

\newcommand{\WsFourBNoiseOneCleanCell}{99.82\%}
\newcommand{\WsFourBNoiseOneOtherCell}{100.86\%}

\newcommand{\WsFourBNoiseSevenSevenThreeVal}{3.9662}
\newcommand{\WsFourBNoiseSevenSevenThreeCos}{0.0178}
\newcommand{\WsFourBNoiseSevenSevenThreePatOne}{0.01}

\newcommand{\WsFourBNoiseSevenSevenThreeCleanCell}{143.98\%}
\newcommand{\WsFourBNoiseSevenSevenThreeOtherCell}{150.13\%}

\newcommand{\WsFourBNoiseMinClean}{99.82}
\newcommand{\WsFourBNoiseMaxClean}{143.98}
\newcommand{\WsFourCWikiCompWhisperQwenToGemmaTwoLnCleanCell}{436.11\%}
\newcommand{\WsFourCWikiCompWhisperQwenToGemmaTwoLnOtherCell}{486.19\%}
\newcommand{\WsFourCWikiCompWhisperQwenToGemmaTwoLnShortCleanCell}{436.1\%}

\newcommand{\WsFourCWikiCompMoonshineQwenToGemmaTwoLnCleanCell}{460.44\%}
\newcommand{\WsFourCWikiCompMoonshineQwenToGemmaTwoLnOtherCell}{513.95\%}
\newcommand{\WsFourCWikiCompMoonshineQwenToGemmaTwoLnShortCleanCell}{460.4\%}

\newcommand{\WsFourCWikiCompWhisperGemmaTwoToQwenLnCleanCell}{108.25\%}
\newcommand{\WsFourCWikiCompWhisperGemmaTwoToQwenLnOtherCell}{113.56\%}
\newcommand{\WsFourCWikiCompWhisperGemmaTwoToQwenLnShortCleanCell}{108.2\%}

\newcommand{\WsFourCWikiCompMoonshineGemmaTwoToQwenLnCleanCell}{99.65\%}
\newcommand{\WsFourCWikiCompMoonshineGemmaTwoToQwenLnOtherCell}{100.43\%}
\newcommand{\WsFourCWikiCompMoonshineGemmaTwoToQwenLnShortCleanCell}{99.6\%}

\newcommand{\WsFourCWikiCompWhisperQwenToGemmaThreeLnCleanCell}{383.34\%}
\newcommand{\WsFourCWikiCompWhisperQwenToGemmaThreeLnOtherCell}{440.11\%}
\newcommand{\WsFourCWikiCompWhisperQwenToGemmaThreeLnShortCleanCell}{383.3\%}

\newcommand{\WsFourCWikiCompMoonshineQwenToGemmaThreeLnCleanCell}{450.57\%}
\newcommand{\WsFourCWikiCompMoonshineQwenToGemmaThreeLnOtherCell}{507.89\%}
\newcommand{\WsFourCWikiCompMoonshineQwenToGemmaThreeLnShortCleanCell}{450.6\%}

\newcommand{\WsFourCWikiCompWhisperGemmaTwoToGemmaThreeLnCleanCell}{475.17\%}
\newcommand{\WsFourCWikiCompWhisperGemmaTwoToGemmaThreeLnOtherCell}{534.63\%}
\newcommand{\WsFourCWikiCompWhisperGemmaTwoToGemmaThreeLnShortCleanCell}{475.2\%}

\newcommand{\WsFourCWikiCompWhisperGemmaTwoToGemmaThreeLnClean}{475.17}

\newcommand{\WsFourCWikiCompMoonshineGemmaTwoToGemmaThreeLnCleanCell}{490.78\%}
\newcommand{\WsFourCWikiCompMoonshineGemmaTwoToGemmaThreeLnOtherCell}{562.71\%}
\newcommand{\WsFourCWikiCompMoonshineGemmaTwoToGemmaThreeLnShortCleanCell}{490.8\%}

\newcommand{\WsFourCWikiCompMoonshineGemmaTwoToGemmaThreeLnClean}{490.78}

\newcommand{\WsFourCMixCompInDistMoonshineQwenToGemmaTwoLnCleanCell}{5.05\%}
\newcommand{\WsFourCMixCompInDistMoonshineQwenToGemmaTwoLnOtherCell}{12.71\%}
\newcommand{\WsFourCMixCompInDistMoonshineQwenToGemmaTwoLnShortCleanCell}{5.1\%}

\newcommand{\WsFourCMixCompCrossWhisperOnMoonshineQwenToGemmaTwoLnCleanCell}{465.22\%}
\newcommand{\WsFourCMixCompCrossWhisperOnMoonshineQwenToGemmaTwoLnOtherCell}{522.81\%}
\newcommand{\WsFourCMixCompCrossWhisperOnMoonshineQwenToGemmaTwoLnShortCleanCell}{465.2\%}

\newcommand{\WsFourCMixCompInDistWhisperQwenToGemmaTwoLnCleanCell}{3.72\%}
\newcommand{\WsFourCMixCompInDistWhisperQwenToGemmaTwoLnOtherCell}{7.93\%}
\newcommand{\WsFourCMixCompInDistWhisperQwenToGemmaTwoLnShortCleanCell}{3.7\%}

\newcommand{\WsFourCMixCompCrossMoonshineOnWhisperQwenToGemmaTwoLnCleanCell}{476.62\%}
\newcommand{\WsFourCMixCompCrossMoonshineOnWhisperQwenToGemmaTwoLnOtherCell}{532.68\%}
\newcommand{\WsFourCMixCompCrossMoonshineOnWhisperQwenToGemmaTwoLnShortCleanCell}{476.6\%}

\newcommand{\WsFourCMixCompInDistMoonshineGemmaTwoToQwenLnCleanCell}{8.03\%}
\newcommand{\WsFourCMixCompInDistMoonshineGemmaTwoToQwenLnOtherCell}{17.45\%}
\newcommand{\WsFourCMixCompInDistMoonshineGemmaTwoToQwenLnShortCleanCell}{8.0\%}

\newcommand{\WsFourCMixCompCrossWhisperOnMoonshineGemmaTwoToQwenLnCleanCell}{105.96\%}
\newcommand{\WsFourCMixCompCrossWhisperOnMoonshineGemmaTwoToQwenLnOtherCell}{109.21\%}
\newcommand{\WsFourCMixCompCrossWhisperOnMoonshineGemmaTwoToQwenLnShortCleanCell}{106.0\%}

\newcommand{\WsFourCMixCompInDistWhisperGemmaTwoToQwenLnCleanCell}{5.13\%}
\newcommand{\WsFourCMixCompInDistWhisperGemmaTwoToQwenLnOtherCell}{11.15\%}
\newcommand{\WsFourCMixCompInDistWhisperGemmaTwoToQwenLnShortCleanCell}{5.1\%}

\newcommand{\WsFourCMixCompCrossMoonshineOnWhisperGemmaTwoToQwenLnCleanCell}{123.29\%}
\newcommand{\WsFourCMixCompCrossMoonshineOnWhisperGemmaTwoToQwenLnOtherCell}{134.18\%}
\newcommand{\WsFourCMixCompCrossMoonshineOnWhisperGemmaTwoToQwenLnShortCleanCell}{123.3\%}

\newcommand{\WsFourCMixCompInDistMoonshineQwenToGemmaThreeLnCleanCell}{4.33\%}
\newcommand{\WsFourCMixCompInDistMoonshineQwenToGemmaThreeLnOtherCell}{10.86\%}
\newcommand{\WsFourCMixCompInDistMoonshineQwenToGemmaThreeLnShortCleanCell}{4.3\%}

\newcommand{\WsFourCMixCompCrossWhisperOnMoonshineQwenToGemmaThreeLnCleanCell}{455.14\%}
\newcommand{\WsFourCMixCompCrossWhisperOnMoonshineQwenToGemmaThreeLnOtherCell}{498.03\%}
\newcommand{\WsFourCMixCompCrossWhisperOnMoonshineQwenToGemmaThreeLnShortCleanCell}{455.1\%}

\newcommand{\WsFourCMixCompInDistWhisperQwenToGemmaThreeLnCleanCell}{3.08\%}
\newcommand{\WsFourCMixCompInDistWhisperQwenToGemmaThreeLnOtherCell}{7.38\%}
\newcommand{\WsFourCMixCompInDistWhisperQwenToGemmaThreeLnShortCleanCell}{3.1\%}

\newcommand{\WsFourCMixCompCrossMoonshineOnWhisperQwenToGemmaThreeLnCleanCell}{480.30\%}
\newcommand{\WsFourCMixCompCrossMoonshineOnWhisperQwenToGemmaThreeLnOtherCell}{524.65\%}
\newcommand{\WsFourCMixCompCrossMoonshineOnWhisperQwenToGemmaThreeLnShortCleanCell}{480.3\%}

\newcommand{\WsFourCMixCompInDistMoonshineGemmaTwoToGemmaThreeLnCleanCell}{4.21\%}
\newcommand{\WsFourCMixCompInDistMoonshineGemmaTwoToGemmaThreeLnOtherCell}{10.81\%}
\newcommand{\WsFourCMixCompInDistMoonshineGemmaTwoToGemmaThreeLnShortCleanCell}{4.2\%}

\newcommand{\WsFourCMixCompInDistMoonshineGemmaTwoToGemmaThreeLnClean}{4.21}

\newcommand{\WsFourCMixCompCrossWhisperOnMoonshineGemmaTwoToGemmaThreeLnCleanCell}{452.43\%}
\newcommand{\WsFourCMixCompCrossWhisperOnMoonshineGemmaTwoToGemmaThreeLnOtherCell}{505.36\%}
\newcommand{\WsFourCMixCompCrossWhisperOnMoonshineGemmaTwoToGemmaThreeLnShortCleanCell}{452.4\%}

\newcommand{\WsFourCMixCompCrossWhisperOnMoonshineGemmaTwoToGemmaThreeLnClean}{452.43}

\newcommand{\WsFourCMixCompInDistWhisperGemmaTwoToGemmaThreeLnCleanCell}{3.26\%}
\newcommand{\WsFourCMixCompInDistWhisperGemmaTwoToGemmaThreeLnOtherCell}{7.71\%}
\newcommand{\WsFourCMixCompInDistWhisperGemmaTwoToGemmaThreeLnShortCleanCell}{3.3\%}

\newcommand{\WsFourCMixCompInDistWhisperGemmaTwoToGemmaThreeLnClean}{3.26}

\newcommand{\WsFourCMixCompCrossMoonshineOnWhisperGemmaTwoToGemmaThreeLnCleanCell}{154.00\%}
\newcommand{\WsFourCMixCompCrossMoonshineOnWhisperGemmaTwoToGemmaThreeLnOtherCell}{205.91\%}
\newcommand{\WsFourCMixCompCrossMoonshineOnWhisperGemmaTwoToGemmaThreeLnShortCleanCell}{154.0\%}

\newcommand{\WsFourCMixCompCrossMoonshineOnWhisperGemmaTwoToGemmaThreeLnClean}{154.00}

\newcommand{\WsFourCWikiCompDone}{8}

\newcommand{\WsFourCWikiMinClean}{99.65}
\newcommand{\WsFourCWikiMaxClean}{490.78}
\newcommand{\WsFourCMixCompBestClean}{3.08}

\newcommand{\WsFourCMixInDistMaxClean}{8.03}
\newcommand{\WsFourCMixCrossMinClean}{105.96}
\newcommand{\WsFourCMixCrossMaxClean}{480.30}

\newcommand{\MixVsAudioTieCriterionPp}{0.10}
\newcommand{\WsThreeSameAboveDirectCleanRows}{8}
\newcommand{\WsThreeCompTotalRows}{8}
\newcommand{\MixVsAudioSameTriplesTotal}{8}
\newcommand{\MixVsAudioSameTriplesTied}{5}

\newcommand{\MixVsAudioTiedMaxAbsDeltaClean}{0.08}

\newcommand{\MixVsAudioMinDeltaClean}{\ensuremath{-}1.01}
\newcommand{\MixVsAudioMaxDeltaClean}{\ensuremath{+}0.16}

\newcommand{\WsFourDWoneBzeroNativePiqa}{78.73}
\newcommand{\WsFourDWoneBzeroNativeHs}{57.50}

\newcommand{\WsFourDBzeroNativePiqa}{78.56}
\newcommand{\WsFourDBzeroNativeHs}{57.50}
\newcommand{\WsFourDBzeroNativePiqaCell}{78.56\%}
\newcommand{\WsFourDBzeroNativeHsCell}{57.50\%}

\newcommand{\WsFourDBzeroNativeRr}{1.000}
\newcommand{\WsFourDConeLstsqScaledPiqa}{77.15}
\newcommand{\WsFourDConeLstsqScaledHs}{57.10}
\newcommand{\WsFourDConeLstsqScaledPiqaCell}{77.15\%}
\newcommand{\WsFourDConeLstsqScaledHsCell}{57.10\%}

\newcommand{\WsFourDConeLstsqScaledDeltaBzeroPiqa}{-1.41}
\newcommand{\WsFourDConeLstsqScaledDeltaBzeroHs}{-0.40}
\newcommand{\WsFourDConeLstsqScaledAboveChancePiqa}{+27.15}
\newcommand{\WsFourDConeLstsqScaledAboveChanceHs}{+32.10}
\newcommand{\WsFourDConeLstsqScaledAnchors}{163{,}983}
\newcommand{\WsFourDConeLstsqScaledRowsOverwritten}{163{,}977}
\newcommand{\WsFourDConeLstsqScaledCos}{0.5768}
\newcommand{\WsFourDConeLstsqScaledPatOne}{98.56}
\newcommand{\WsFourDConeLstsqScaledRr}{1.000}

\newcommand{\WsFourDCtwoLstsqNonePiqaCell}{76.44\%}
\newcommand{\WsFourDCtwoLstsqNoneHsCell}{54.40\%}

\newcommand{\WsFourDCtwoLstsqNoneDeltaBzeroPiqa}{-2.12}
\newcommand{\WsFourDCtwoLstsqNoneDeltaBzeroHs}{-3.10}

\newcommand{\WsFourDCtwoLstsqNoneCos}{0.5768}
\newcommand{\WsFourDCtwoLstsqNonePatOne}{98.56}
\newcommand{\WsFourDCtwoLstsqNoneRr}{0.603}
\newcommand{\WsFourDCthreeProcScaledPiqa}{54.08}

\newcommand{\WsFourDCthreeProcScaledPiqaCell}{54.08\%}
\newcommand{\WsFourDCthreeProcScaledHsCell}{28.35\%}

\newcommand{\WsFourDCthreeProcScaledDeltaBzeroPiqa}{-24.48}
\newcommand{\WsFourDCthreeProcScaledDeltaBzeroHs}{-29.15}

\newcommand{\WsFourDCthreeProcScaledCos}{0.5080}
\newcommand{\WsFourDCthreeProcScaledPatOne}{97.47}
\newcommand{\WsFourDCthreeProcScaledRr}{1.000}

\newcommand{\WsFourDCfourRandScaledPiqaCell}{49.89\%}
\newcommand{\WsFourDCfourRandScaledHsCell}{25.05\%}

\newcommand{\WsFourDCfourRandScaledDeltaBzeroPiqa}{-28.67}
\newcommand{\WsFourDCfourRandScaledDeltaBzeroHs}{-32.45}

\newcommand{\WsFourDCfourRandScaledCos}{0.0005}
\newcommand{\WsFourDCfourRandScaledPatOne}{0.00}
\newcommand{\WsFourDCfourRandScaledRr}{1.000}
\newcommand{\WsFourDToneWteLibriPiqa}{54.24}
\newcommand{\WsFourDToneWteLibriHs}{30.40}
\newcommand{\WsFourDToneWteLibriPiqaCell}{54.24\%}
\newcommand{\WsFourDToneWteLibriHsCell}{30.40\%}

\newcommand{\WsFourDToneWteLibriDeltaBzeroPiqa}{-24.32}
\newcommand{\WsFourDToneWteLibriDeltaBzeroHs}{-27.10}
\newcommand{\WsFourDToneWteLibriAboveChancePiqa}{+4.24}
\newcommand{\WsFourDToneWteLibriAboveChanceHs}{+5.40}

\newcommand{\WsFourDToneWteLibriCos}{0.1136}
\newcommand{\WsFourDToneWteLibriPatOne}{36.11}
\newcommand{\WsFourDToneWteLibriRr}{1.028}
\newcommand{\WsFourDTtwoWteWikiPiqa}{59.47}
\newcommand{\WsFourDTtwoWteWikiHs}{34.20}
\newcommand{\WsFourDTtwoWteWikiPiqaCell}{59.47\%}
\newcommand{\WsFourDTtwoWteWikiHsCell}{34.20\%}

\newcommand{\WsFourDTtwoWteWikiDeltaBzeroPiqa}{-19.10}
\newcommand{\WsFourDTtwoWteWikiDeltaBzeroHs}{-23.30}
\newcommand{\WsFourDTtwoWteWikiAboveChancePiqa}{+9.47}
\newcommand{\WsFourDTtwoWteWikiAboveChanceHs}{+9.20}

\newcommand{\WsFourDTtwoWteWikiCos}{0.1297}
\newcommand{\WsFourDTtwoWteWikiPatOne}{52.91}
\newcommand{\WsFourDTtwoWteWikiRr}{1.030}
\newcommand{\WsFourDTthreeWteMixMoonshinePiqa}{57.73}
\newcommand{\WsFourDTthreeWteMixMoonshineHs}{34.45}
\newcommand{\WsFourDTthreeWteMixMoonshinePiqaCell}{57.73\%}
\newcommand{\WsFourDTthreeWteMixMoonshineHsCell}{34.45\%}

\newcommand{\WsFourDTthreeWteMixMoonshineDeltaBzeroPiqa}{-20.84}
\newcommand{\WsFourDTthreeWteMixMoonshineDeltaBzeroHs}{-23.05}
\newcommand{\WsFourDTthreeWteMixMoonshineAboveChancePiqa}{+7.73}
\newcommand{\WsFourDTthreeWteMixMoonshineAboveChanceHs}{+9.45}

\newcommand{\WsFourDTthreeWteMixMoonshineCos}{0.1119}
\newcommand{\WsFourDTthreeWteMixMoonshinePatOne}{44.61}
\newcommand{\WsFourDTthreeWteMixMoonshineRr}{1.080}
\newcommand{\WsFourDTsixWteMixWhisperPiqa}{57.56}
\newcommand{\WsFourDTsixWteMixWhisperHs}{33.85}
\newcommand{\WsFourDTsixWteMixWhisperPiqaCell}{57.56\%}
\newcommand{\WsFourDTsixWteMixWhisperHsCell}{33.85\%}

\newcommand{\WsFourDTsixWteMixWhisperDeltaBzeroPiqa}{-21.16}
\newcommand{\WsFourDTsixWteMixWhisperDeltaBzeroHs}{-23.65}
\newcommand{\WsFourDTsixWteMixWhisperAboveChancePiqa}{+7.56}
\newcommand{\WsFourDTsixWteMixWhisperAboveChanceHs}{+8.85}

\newcommand{\WsFourDTsixWteMixWhisperCos}{0.1082}
\newcommand{\WsFourDTsixWteMixWhisperPatOne}{44.66}
\newcommand{\WsFourDTsixWteMixWhisperRr}{1.090}
\newcommand{\WsFourDTfourWteAudioMoonshinePiqa}{51.90}
\newcommand{\WsFourDTfourWteAudioMoonshineHs}{27.45}
\newcommand{\WsFourDTfourWteAudioMoonshinePiqaCell}{51.90\%}
\newcommand{\WsFourDTfourWteAudioMoonshineHsCell}{27.45\%}

\newcommand{\WsFourDTfourWteAudioMoonshineDeltaBzeroPiqa}{-26.66}
\newcommand{\WsFourDTfourWteAudioMoonshineDeltaBzeroHs}{-30.05}
\newcommand{\WsFourDTfourWteAudioMoonshineAboveChancePiqa}{+1.90}
\newcommand{\WsFourDTfourWteAudioMoonshineAboveChanceHs}{+2.45}

\newcommand{\WsFourDTfourWteAudioMoonshineCos}{0.0339}
\newcommand{\WsFourDTfourWteAudioMoonshinePatOne}{0.83}
\newcommand{\WsFourDTfourWteAudioMoonshineRr}{1.058}
\newcommand{\WsFourDTfiveWteAudioWhisperPiqa}{52.50}
\newcommand{\WsFourDTfiveWteAudioWhisperHs}{25.50}
\newcommand{\WsFourDTfiveWteAudioWhisperPiqaCell}{52.50\%}
\newcommand{\WsFourDTfiveWteAudioWhisperHsCell}{25.50\%}

\newcommand{\WsFourDTfiveWteAudioWhisperDeltaBzeroPiqa}{-26.22}
\newcommand{\WsFourDTfiveWteAudioWhisperDeltaBzeroHs}{-32.00}
\newcommand{\WsFourDTfiveWteAudioWhisperAboveChancePiqa}{+2.50}
\newcommand{\WsFourDTfiveWteAudioWhisperAboveChanceHs}{+0.50}

\newcommand{\WsFourDTfiveWteAudioWhisperCos}{0.0387}
\newcommand{\WsFourDTfiveWteAudioWhisperPatOne}{1.27}
\newcommand{\WsFourDTfiveWteAudioWhisperRr}{1.053}
\newcommand{\WsFourDTsevenWteJointPiqa}{51.96}
\newcommand{\WsFourDTsevenWteJointHs}{26.45}
\newcommand{\WsFourDTsevenWteJointPiqaCell}{51.96\%}
\newcommand{\WsFourDTsevenWteJointHsCell}{26.45\%}

\newcommand{\WsFourDTsevenWteJointDeltaBzeroPiqa}{-26.61}
\newcommand{\WsFourDTsevenWteJointDeltaBzeroHs}{-31.05}
\newcommand{\WsFourDTsevenWteJointAboveChancePiqa}{+1.96}
\newcommand{\WsFourDTsevenWteJointAboveChanceHs}{+1.45}

\newcommand{\WsFourDTsevenWteJointCos}{0.0391}
\newcommand{\WsFourDTsevenWteJointPatOne}{1.21}
\newcommand{\WsFourDTsevenWteJointRr}{1.090}
\newcommand{\WsFourDBestTrainedPiqa}{59.47}
\newcommand{\WsFourDBestTrainedHs}{34.45}
\newcommand{\WsFourDBestTrainedAboveChancePiqa}{+9.47}
\newcommand{\WsFourDBestTrainedAboveChanceHs}{+9.45}

\newcommand{\WsFourDDeltaConeVsBzeroPiqa}{-1.41}
\newcommand{\WsFourDDeltaConeVsBzeroHs}{-0.40}
\newcommand{\WsFourEDirectWhisperMuonMedClean}{0.00}
\newcommand{\WsFourEDirectWhisperMuonMedOther}{4.35}
\newcommand{\WsFourEDirectWhisperAdamwMedClean}{0.00}
\newcommand{\WsFourEDirectWhisperAdamwMedOther}{4.35}
\newcommand{\WsFourEDirectMoonshineMuonMedClean}{0.00}
\newcommand{\WsFourEDirectMoonshineMuonMedOther}{8.33}
\newcommand{\WsFourEDirectMoonshineAdamwMedClean}{0.00}
\newcommand{\WsFourEDirectMoonshineAdamwMedOther}{8.33}
\newcommand{\WsFourEFitSingleSteps}{17{,}578}
\newcommand{\WsFourEFitSingleBestStep}{17{,}500}
\newcommand{\WsFourESingleSameWhisperMuonMedClean}{0.00}
\newcommand{\WsFourESingleSameWhisperMuonMedOther}{3.85}
\newcommand{\WsFourESingleSameMoonshineMuonMedClean}{0.00}
\newcommand{\WsFourESingleSameMoonshineMuonMedOther}{7.69}
\newcommand{\WsFourEFitJointTotalSteps}{35{,}156}
\newcommand{\WsFourEFitJointBestStep}{32{,}500}
\newcommand{\WsFourEFitJointHalfStep}{17{,}578}
\newcommand{\WsFourEFitJointSamplesPerStep}{16}
\newcommand{\WsFourEFitJointSamplesPerAdapterStep}{8}
\newcommand{\WsFourEFitJointTextVal}{6.91}

\newcommand{\WsFourEJointWhisperMuonClean}{3.09}
\newcommand{\WsFourEJointWhisperMuonOther}{7.51}

\newcommand{\WsFourEJointWhisperMuonEmptyClean}{0}
\newcommand{\WsFourEJointWhisperMuonEmptyOther}{0}
\newcommand{\WsFourEJointWhisperMuonMedClean}{0.00}
\newcommand{\WsFourEJointWhisperMuonMedOther}{3.85}
\newcommand{\WsFourEJointCtrlWhisperAdamwClean}{41.20}
\newcommand{\WsFourEJointCtrlWhisperAdamwOther}{44.23}

\newcommand{\WsFourEJointCtrlWhisperAdamwEmptyClean}{552}
\newcommand{\WsFourEJointCtrlWhisperAdamwEmptyOther}{584}
\newcommand{\WsFourEJointCtrlWhisperAdamwMedClean}{4.76}
\newcommand{\WsFourEJointCtrlWhisperAdamwMedOther}{12.50}
\newcommand{\WsFourEJointHalfWhisperMuonClean}{3.44}
\newcommand{\WsFourEJointHalfWhisperMuonOther}{7.75}

\newcommand{\WsFourEJointHalfWhisperMuonEmptyClean}{1}
\newcommand{\WsFourEJointHalfWhisperMuonEmptyOther}{0}
\newcommand{\WsFourEJointHalfWhisperMuonMedClean}{0.00}
\newcommand{\WsFourEJointHalfWhisperMuonMedOther}{4.55}
\newcommand{\WsFourEJointHalfTrailWhisperClean}{0.22}
\newcommand{\WsFourESingleCtrlWhisperAdamwClean}{40.49}
\newcommand{\WsFourESingleCtrlWhisperAdamwOther}{52.23}

\newcommand{\WsFourESingleCtrlWhisperAdamwEmptyClean}{286}
\newcommand{\WsFourESingleCtrlWhisperAdamwEmptyOther}{263}
\newcommand{\WsFourESingleCtrlWhisperAdamwMedClean}{5.26}
\newcommand{\WsFourESingleCtrlWhisperAdamwMedOther}{12.50}
\newcommand{\WsFourEJointMoonshineMuonClean}{4.27}
\newcommand{\WsFourEJointMoonshineMuonOther}{10.66}

\newcommand{\WsFourEJointMoonshineMuonEmptyClean}{0}
\newcommand{\WsFourEJointMoonshineMuonEmptyOther}{0}
\newcommand{\WsFourEJointMoonshineMuonMedClean}{0.00}
\newcommand{\WsFourEJointMoonshineMuonMedOther}{7.69}
\newcommand{\WsFourEJointCtrlMoonshineAdamwClean}{12.78}
\newcommand{\WsFourEJointCtrlMoonshineAdamwOther}{23.26}

\newcommand{\WsFourEJointCtrlMoonshineAdamwEmptyClean}{4}
\newcommand{\WsFourEJointCtrlMoonshineAdamwEmptyOther}{3}
\newcommand{\WsFourEJointCtrlMoonshineAdamwMedClean}{5.13}
\newcommand{\WsFourEJointCtrlMoonshineAdamwMedOther}{14.89}
\newcommand{\WsFourEJointHalfMoonshineMuonClean}{4.52}
\newcommand{\WsFourEJointHalfMoonshineMuonOther}{11.27}

\newcommand{\WsFourEJointHalfMoonshineMuonEmptyClean}{0}
\newcommand{\WsFourEJointHalfMoonshineMuonEmptyOther}{0}
\newcommand{\WsFourEJointHalfMoonshineMuonMedClean}{0.00}
\newcommand{\WsFourEJointHalfMoonshineMuonMedOther}{8.33}
\newcommand{\WsFourEJointHalfTrailMoonshineClean}{0.23}
\newcommand{\WsFourESingleCtrlMoonshineAdamwClean}{6.54}
\newcommand{\WsFourESingleCtrlMoonshineAdamwOther}{13.74}

\newcommand{\WsFourESingleCtrlMoonshineAdamwEmptyClean}{0}
\newcommand{\WsFourESingleCtrlMoonshineAdamwEmptyOther}{0}
\newcommand{\WsFourESingleCtrlMoonshineAdamwMedClean}{3.57}
\newcommand{\WsFourESingleCtrlMoonshineAdamwMedOther}{11.11}

\newcommand{\VersionsDistinctStacks}{11}

\newcommand{\VersionsTorchList}{2.6.0, 2.7.1, 2.9.1, 2.10.0}
\newcommand{\VersionsTransformersList}{4.53.3, 5.15.0, 5.16.0.dev0, 5.17.0, 5.22.1, 5.22.2}

\newcommand{\NativeGemmaThreeHostPiqaDiffItems}{3}
\newcommand{\StackcheckAsevenTotalUtts}{5{,}559}

\newcommand{\StackcheckAsevenDeltaClean}{-0.01}
\newcommand{\StackcheckAsevenDeltaOther}{-0.01}

\newcommand{\StackcheckAsevenDiffTotalUtts}{156}

\newcommand{\XprofileXoneCleanUttMedWer}{23.81}

\newcommand{\XprofileXtwoCleanUttMedWer}{20.83}

\newcommand{\XprofileZeroAudioCells}{113}
\newcommand{\XprofileZeroAudioTotalUtts}{628{,}167}

\newcommand{\XprofileZeroAudioLtFiftyPct}{0.00}

\newcommand{\ZeroAudioCompBestClean}{99.65}
\newcommand{\XprofileZeroAudioClosedFormCells}{84}
\newcommand{\XprofileZeroAudioRandOrthCells}{2}
\newcommand{\XprofileZeroAudioTranscriptCells}{16}
\newcommand{\XprofileZeroAudioNoiseCells}{3}
\newcommand{\XprofileZeroAudioWikiCells}{8}
\newcommand{\WsFourESingleSameMaxMedClean}{0.00}
\newcommand{\CrossAdapterAudioOnlyCells}{8}
\newcommand{\CrossAdapterMixedCells}{8}
\newcommand{\CrossAdapterTotalCells}{16}
\newcommand{\CrossAdapterFailCells}{14}

\newcommand{\CrossAdapterMinClean}{67.35}

\newcommand{\CrossAdapterMaxClean}{480.30}

\newcommand{\CrossAdapterFailMinMedClean}{98.00}

\newcommand{\CrossAdapterFailMinWerClean}{98.63}
\newcommand{\CrossAdapterFailMaxWerClean}{480.30}
\newcommand{\CrossAdapterPartialMinMedClean}{20.83}
\newcommand{\CrossAdapterPartialMaxMedClean}{23.81}
\newcommand{\CrossAdapterPartialMinWerClean}{67.35}
\newcommand{\CrossAdapterPartialMaxWerClean}{75.35}

\newcommand{\ManifoldNUtterances}{200}
\newcommand{\ManifoldNTokenEval}{10{,}000}

\newcommand{\ManifoldTokMinNnCosMed}{0.5467}
\newcommand{\ManifoldTokMaxNnCosMed}{0.6291}
\newcommand{\ManifoldSrcMinNnCosMed}{0.1089}
\newcommand{\ManifoldSrcMaxNnCosMed}{0.1376}
\newcommand{\ManifoldSrcGemmaMinRrMed}{1.02}
\newcommand{\ManifoldSrcGemmaMaxRrMed}{1.12}
\newcommand{\ManifoldSrcQwenMinRrMed}{12.74}
\newcommand{\ManifoldSrcQwenMaxRrMed}{12.87}

\newcommand{\ManifoldMapProcMinNnCosMed}{0.0947}
\newcommand{\ManifoldMapProcMaxNnCosMed}{0.1531}
\newcommand{\ManifoldMapTranscriptMinNnCosMed}{0.1182}
\newcommand{\ManifoldMapTranscriptMaxNnCosMed}{0.1977}
\newcommand{\ManifoldMapLstsqScaledMinNnCosMed}{0.2130}
\newcommand{\ManifoldMapLstsqScaledMaxNnCosMed}{0.3127}
\newcommand{\ManifoldMapLstsqLnMinNnCosMed}{0.4152}
\newcommand{\ManifoldMapLstsqLnMaxNnCosMed}{0.6311}

\newcommand{\TokenHitsQwenThreePiqaCombOccTr}{62.41}
\newcommand{\TokenHitsQwenThreePiqaCombOccTe}{17.68}
\newcommand{\TokenHitsQwenThreePiqaCombOccNat}{19.91}

\newcommand{\TokenHitsQwenThreePiqaCombTypTr}{67.30}
\newcommand{\TokenHitsQwenThreePiqaCombTypTe}{16.95}
\newcommand{\TokenHitsQwenThreePiqaCombTypNat}{15.75}

\newcommand{\TokenHitsQwenThreePiqaCtxOccTr}{56.97}
\newcommand{\TokenHitsQwenThreePiqaCtxOccTe}{19.61}
\newcommand{\TokenHitsQwenThreePiqaCtxOccNat}{23.42}

\newcommand{\TokenHitsQwenThreePiqaCtxTypTr}{69.07}
\newcommand{\TokenHitsQwenThreePiqaCtxTypTe}{17.20}
\newcommand{\TokenHitsQwenThreePiqaCtxTypNat}{13.73}

\newcommand{\TokenHitsQwenThreePiqaContOccTr}{64.66}
\newcommand{\TokenHitsQwenThreePiqaContOccTe}{16.88}
\newcommand{\TokenHitsQwenThreePiqaContOccNat}{18.46}

\newcommand{\TokenHitsQwenThreePiqaContTypTr}{67.94}
\newcommand{\TokenHitsQwenThreePiqaContTypTe}{17.24}
\newcommand{\TokenHitsQwenThreePiqaContTypNat}{14.82}

\newcommand{\TokenHitsQwenThreeHsCombOccTr}{71.16}
\newcommand{\TokenHitsQwenThreeHsCombOccTe}{14.96}
\newcommand{\TokenHitsQwenThreeHsCombOccNat}{13.89}

\newcommand{\TokenHitsQwenThreeHsCombTypTr}{69.86}
\newcommand{\TokenHitsQwenThreeHsCombTypTe}{17.43}
\newcommand{\TokenHitsQwenThreeHsCombTypNat}{12.71}

\newcommand{\TokenHitsQwenThreeHsCtxOccTr}{70.07}
\newcommand{\TokenHitsQwenThreeHsCtxOccTe}{14.69}
\newcommand{\TokenHitsQwenThreeHsCtxOccNat}{15.24}

\newcommand{\TokenHitsQwenThreeHsCtxTypTr}{71.98}
\newcommand{\TokenHitsQwenThreeHsCtxTypTe}{17.83}
\newcommand{\TokenHitsQwenThreeHsCtxTypNat}{10.20}

\newcommand{\TokenHitsQwenThreeHsContOccTr}{73.22}
\newcommand{\TokenHitsQwenThreeHsContOccTe}{15.45}
\newcommand{\TokenHitsQwenThreeHsContOccNat}{11.34}

\newcommand{\TokenHitsQwenThreeHsContTypTr}{70.29}
\newcommand{\TokenHitsQwenThreeHsContTypTe}{17.49}
\newcommand{\TokenHitsQwenThreeHsContTypNat}{12.22}

\newcommand{\TokenHitsQwenThreeSurfaceOverwrittenRows}{65{,}406}

\newcommand{\TokenHitsQwenThreeSurfacePiqaCombOccTr}{68.09}
\newcommand{\TokenHitsQwenThreeSurfacePiqaCombOccTe}{25.69}
\newcommand{\TokenHitsQwenThreeSurfacePiqaCombOccNat}{6.22}

\newcommand{\TokenHitsQwenThreeSurfacePiqaCombTypTr}{67.49}
\newcommand{\TokenHitsQwenThreeSurfacePiqaCombTypTe}{18.04}
\newcommand{\TokenHitsQwenThreeSurfacePiqaCombTypNat}{14.47}

\newcommand{\TokenHitsQwenThreeSurfacePiqaCtxOccTr}{67.18}
\newcommand{\TokenHitsQwenThreeSurfacePiqaCtxOccTe}{18.71}
\newcommand{\TokenHitsQwenThreeSurfacePiqaCtxOccNat}{14.11}

\newcommand{\TokenHitsQwenThreeSurfacePiqaCtxTypTr}{68.90}
\newcommand{\TokenHitsQwenThreeSurfacePiqaCtxTypTe}{18.85}
\newcommand{\TokenHitsQwenThreeSurfacePiqaCtxTypNat}{12.25}

\newcommand{\TokenHitsQwenThreeSurfacePiqaContOccTr}{68.47}
\newcommand{\TokenHitsQwenThreeSurfacePiqaContOccTe}{28.57}
\newcommand{\TokenHitsQwenThreeSurfacePiqaContOccNat}{2.96}

\newcommand{\TokenHitsQwenThreeSurfacePiqaContTypTr}{68.46}
\newcommand{\TokenHitsQwenThreeSurfacePiqaContTypTe}{18.12}
\newcommand{\TokenHitsQwenThreeSurfacePiqaContTypNat}{13.41}

\newcommand{\TokenHitsQwenThreeSurfaceHsCombOccTr}{71.59}
\newcommand{\TokenHitsQwenThreeSurfaceHsCombOccTe}{24.17}
\newcommand{\TokenHitsQwenThreeSurfaceHsCombOccNat}{4.24}

\newcommand{\TokenHitsQwenThreeSurfaceHsCombTypTr}{70.01}
\newcommand{\TokenHitsQwenThreeSurfaceHsCombTypTe}{18.32}
\newcommand{\TokenHitsQwenThreeSurfaceHsCombTypNat}{11.67}

\newcommand{\TokenHitsQwenThreeSurfaceHsCtxOccTr}{70.97}
\newcommand{\TokenHitsQwenThreeSurfaceHsCtxOccTe}{23.47}
\newcommand{\TokenHitsQwenThreeSurfaceHsCtxOccNat}{5.56}

\newcommand{\TokenHitsQwenThreeSurfaceHsCtxTypTr}{72.06}
\newcommand{\TokenHitsQwenThreeSurfaceHsCtxTypTe}{18.81}
\newcommand{\TokenHitsQwenThreeSurfaceHsCtxTypNat}{9.13}

\newcommand{\TokenHitsQwenThreeSurfaceHsContOccTr}{72.76}
\newcommand{\TokenHitsQwenThreeSurfaceHsContOccTe}{25.49}
\newcommand{\TokenHitsQwenThreeSurfaceHsContOccNat}{1.75}

\newcommand{\TokenHitsQwenThreeSurfaceHsContTypTr}{70.60}
\newcommand{\TokenHitsQwenThreeSurfaceHsContTypTe}{18.21}
\newcommand{\TokenHitsQwenThreeSurfaceHsContTypNat}{11.19}

\newcommand{\TokenHitsSmolPiqaCombOccTr}{64.71}
\newcommand{\TokenHitsSmolPiqaCombOccTe}{15.07}
\newcommand{\TokenHitsSmolPiqaCombOccNat}{20.22}

\newcommand{\TokenHitsSmolPiqaCombTypTr}{64.79}
\newcommand{\TokenHitsSmolPiqaCombTypTe}{17.22}
\newcommand{\TokenHitsSmolPiqaCombTypNat}{17.99}

\newcommand{\TokenHitsSmolPiqaCtxOccTr}{61.09}
\newcommand{\TokenHitsSmolPiqaCtxOccTe}{14.82}
\newcommand{\TokenHitsSmolPiqaCtxOccNat}{24.09}

\newcommand{\TokenHitsSmolPiqaCtxTypTr}{65.54}
\newcommand{\TokenHitsSmolPiqaCtxTypTe}{17.93}
\newcommand{\TokenHitsSmolPiqaCtxTypNat}{16.54}

\newcommand{\TokenHitsSmolPiqaContOccTr}{66.21}
\newcommand{\TokenHitsSmolPiqaContOccTe}{15.18}
\newcommand{\TokenHitsSmolPiqaContOccNat}{18.62}

\newcommand{\TokenHitsSmolPiqaContTypTr}{66.45}
\newcommand{\TokenHitsSmolPiqaContTypTe}{17.42}
\newcommand{\TokenHitsSmolPiqaContTypNat}{16.13}

\newcommand{\TokenHitsSmolHsCombOccTr}{70.87}
\newcommand{\TokenHitsSmolHsCombOccTe}{14.64}
\newcommand{\TokenHitsSmolHsCombOccNat}{14.49}

\newcommand{\TokenHitsSmolHsCombTypTr}{65.71}
\newcommand{\TokenHitsSmolHsCombTypTe}{17.87}
\newcommand{\TokenHitsSmolHsCombTypNat}{16.42}

\newcommand{\TokenHitsSmolHsCtxOccTr}{69.48}
\newcommand{\TokenHitsSmolHsCtxOccTe}{14.70}
\newcommand{\TokenHitsSmolHsCtxOccNat}{15.82}

\newcommand{\TokenHitsSmolHsCtxTypTr}{68.32}
\newcommand{\TokenHitsSmolHsCtxTypTe}{18.66}
\newcommand{\TokenHitsSmolHsCtxTypNat}{13.02}

\newcommand{\TokenHitsSmolHsContOccTr}{73.48}
\newcommand{\TokenHitsSmolHsContOccTe}{14.53}
\newcommand{\TokenHitsSmolHsContOccNat}{11.99}

\newcommand{\TokenHitsSmolHsContTypTr}{66.34}
\newcommand{\TokenHitsSmolHsContTypTe}{17.91}
\newcommand{\TokenHitsSmolHsContTypNat}{15.76}

\newcommand{\TokenHitsPhiPiqaCombOccTr}{67.42}
\newcommand{\TokenHitsPhiPiqaCombOccTe}{11.74}
\newcommand{\TokenHitsPhiPiqaCombOccNat}{20.84}

\newcommand{\TokenHitsPhiPiqaCombTypTr}{63.75}
\newcommand{\TokenHitsPhiPiqaCombTypTe}{16.36}
\newcommand{\TokenHitsPhiPiqaCombTypNat}{19.88}

\newcommand{\TokenHitsPhiPiqaCtxOccTr}{63.68}
\newcommand{\TokenHitsPhiPiqaCtxOccTe}{11.78}
\newcommand{\TokenHitsPhiPiqaCtxOccNat}{24.54}

\newcommand{\TokenHitsPhiPiqaCtxTypTr}{65.71}
\newcommand{\TokenHitsPhiPiqaCtxTypTe}{16.59}
\newcommand{\TokenHitsPhiPiqaCtxTypNat}{17.69}

\newcommand{\TokenHitsPhiPiqaContOccTr}{68.97}
\newcommand{\TokenHitsPhiPiqaContOccTe}{11.72}
\newcommand{\TokenHitsPhiPiqaContOccNat}{19.31}

\newcommand{\TokenHitsPhiPiqaContTypTr}{64.71}
\newcommand{\TokenHitsPhiPiqaContTypTe}{16.75}
\newcommand{\TokenHitsPhiPiqaContTypNat}{18.55}

\newcommand{\TokenHitsPhiHsCombOccTr}{72.18}
\newcommand{\TokenHitsPhiHsCombOccTe}{13.46}
\newcommand{\TokenHitsPhiHsCombOccNat}{14.36}

\newcommand{\TokenHitsPhiHsCombTypTr}{67.43}
\newcommand{\TokenHitsPhiHsCombTypTe}{16.96}
\newcommand{\TokenHitsPhiHsCombTypNat}{15.61}

\newcommand{\TokenHitsPhiHsCtxOccTr}{70.23}
\newcommand{\TokenHitsPhiHsCtxOccTe}{14.04}
\newcommand{\TokenHitsPhiHsCtxOccNat}{15.72}

\newcommand{\TokenHitsPhiHsCtxTypTr}{69.82}
\newcommand{\TokenHitsPhiHsCtxTypTe}{17.54}
\newcommand{\TokenHitsPhiHsCtxTypNat}{12.64}

\newcommand{\TokenHitsPhiHsContOccTr}{75.85}
\newcommand{\TokenHitsPhiHsContOccTe}{12.36}
\newcommand{\TokenHitsPhiHsContOccNat}{11.80}

\newcommand{\TokenHitsPhiHsContTypTr}{67.83}
\newcommand{\TokenHitsPhiHsContTypTe}{17.29}
\newcommand{\TokenHitsPhiHsContTypNat}{14.87}

\newcommand{\TokenHitsGemmaTwoPiqaCombOccTr}{58.69}
\newcommand{\TokenHitsGemmaTwoPiqaCombOccTe}{24.14}
\newcommand{\TokenHitsGemmaTwoPiqaCombOccNat}{17.16}

\newcommand{\TokenHitsGemmaTwoPiqaCombTypTr}{77.10}
\newcommand{\TokenHitsGemmaTwoPiqaCombTypTe}{19.86}
\newcommand{\TokenHitsGemmaTwoPiqaCombTypNat}{3.04}

\newcommand{\TokenHitsGemmaTwoPiqaCtxOccTr}{53.56}
\newcommand{\TokenHitsGemmaTwoPiqaCtxOccTe}{25.91}
\newcommand{\TokenHitsGemmaTwoPiqaCtxOccNat}{20.53}

\newcommand{\TokenHitsGemmaTwoPiqaCtxTypTr}{77.21}
\newcommand{\TokenHitsGemmaTwoPiqaCtxTypTe}{19.92}
\newcommand{\TokenHitsGemmaTwoPiqaCtxTypNat}{2.87}

\newcommand{\TokenHitsGemmaTwoPiqaContOccTr}{60.82}
\newcommand{\TokenHitsGemmaTwoPiqaContOccTe}{23.41}
\newcommand{\TokenHitsGemmaTwoPiqaContOccNat}{15.77}

\newcommand{\TokenHitsGemmaTwoPiqaContTypTr}{77.43}
\newcommand{\TokenHitsGemmaTwoPiqaContTypTe}{19.72}
\newcommand{\TokenHitsGemmaTwoPiqaContTypNat}{2.84}

\newcommand{\TokenHitsGemmaTwoHsCombOccTr}{64.95}
\newcommand{\TokenHitsGemmaTwoHsCombOccTe}{22.56}
\newcommand{\TokenHitsGemmaTwoHsCombOccNat}{12.49}

\newcommand{\TokenHitsGemmaTwoHsCombTypTr}{77.79}
\newcommand{\TokenHitsGemmaTwoHsCombTypTe}{19.49}
\newcommand{\TokenHitsGemmaTwoHsCombTypNat}{2.72}

\newcommand{\TokenHitsGemmaTwoHsCtxOccTr}{64.78}
\newcommand{\TokenHitsGemmaTwoHsCtxOccTe}{21.36}
\newcommand{\TokenHitsGemmaTwoHsCtxOccNat}{13.86}

\newcommand{\TokenHitsGemmaTwoHsCtxTypTr}{78.48}
\newcommand{\TokenHitsGemmaTwoHsCtxTypTe}{18.98}
\newcommand{\TokenHitsGemmaTwoHsCtxTypNat}{2.53}

\newcommand{\TokenHitsGemmaTwoHsContOccTr}{65.29}
\newcommand{\TokenHitsGemmaTwoHsContOccTe}{24.82}
\newcommand{\TokenHitsGemmaTwoHsContOccNat}{9.90}

\newcommand{\TokenHitsGemmaTwoHsContTypTr}{77.87}
\newcommand{\TokenHitsGemmaTwoHsContTypTe}{19.46}
\newcommand{\TokenHitsGemmaTwoHsContTypNat}{2.67}

\newcommand{\TokenHitsTabOneExactMinOverwrittenPct}{79.16}
\newcommand{\TokenHitsTabOneExactMaxOverwrittenPct}{86.11}
\newcommand{\TokenHitsTabOneExactMinTrainAnchorPct}{62.41}
\newcommand{\TokenHitsTabOneExactMaxTrainAnchorPct}{72.18}
\newcommand{\TokenHitsTabOneExactMinHeldoutAnchorPct}{11.74}
\newcommand{\TokenHitsTabOneExactMaxHeldoutAnchorPct}{17.68}

\newcommand{\GeoSrcSharedTokCount}{163{,}983}
\newcommand{\GeoSrcSoftPromptTotal}{12{,}000}
\newcommand{\GeoSrcTotalPoints}{175{,}983}
\newcommand{\GeoSrcFramesPerAdapter}{3{,}000}
\newcommand{\GeoSrcFramesPerUtt}{15}
\newcommand{\GeoSrcDim}{2{,}304}
\newcommand{\GeoSrcPcaDims}{50}
\newcommand{\GeoSrcPerp}{30}
\newcommand{\GeoSrcUmapNbrs}{30}
\newcommand{\GeoSrcZeroNormRows}{0}
\newcommand{\GeoSrcDupRows}{0}
\newcommand{\GeoSrcTokSharePct}{93.2}
\newcommand{\GeoSrcDisjTokSelfPct}{100.0}
\newcommand{\GeoTgtDisjLsLnSelfPct}{100.0}
\newcommand{\GeoTgtDisjTransSelfPct}{99.9}
\newcommand{\GeoSrcAllFourMapsTokSpPct}{0.00}
\newcommand{\GeoSrcRadMin}{1.02}
\newcommand{\GeoSrcRadMax}{1.09}

\newcommand{\GeoSrcTokWithSpCount}{0}
\newcommand{\GeoSrcSpWithTokCount}{2}

\newcommand{\GeoSrcWhToMsFullPct}{0.37}

\newcommand{\GeoSrcMsToWhFullPct}{0.00}
\newcommand{\GeoSrcOtherOptFullMinPct}{5.9}
\newcommand{\GeoSrcOtherOptFullMaxPct}{15.5}
\newcommand{\GeoSrcOtherOptFullMinIntPct}{6}
\newcommand{\GeoSrcOtherOptFullMaxIntPct}{16}
\newcommand{\GeoSrcAlltokNearTokCosMinMed}{0.120}
\newcommand{\GeoSrcAlltokNearTokCosMaxMed}{0.135}
\newcommand{\GeoSrcAlltokNearTokCosMax}{0.189}
\newcommand{\GeoSrcTokSelfNearTokCosMed}{0.628}

\newcommand{\GeoSrcAlltokPcaEvrTwoDecPct}{15.85}
\newcommand{\GeoSrcAlltokPcaTokOwnPct}{13.30}
\newcommand{\GeoSrcAlltokPcaSpOwnMinPct}{19.84}
\newcommand{\GeoSrcAlltokPcaSpOwnMaxPct}{33.09}
\newcommand{\GeoSrcTsneTokSelfPct}{100.00}
\newcommand{\GeoSrcTsneCrossEncMaxPct}{0.00}
\newcommand{\GeoSrcTsneWhMixMinPct}{24}
\newcommand{\GeoSrcTsneWhMixMaxPct}{25}
\newcommand{\GeoSrcTsneMsMixPct}{38}
\newcommand{\GeoSrcUmapTokSelfPct}{100.00}
\newcommand{\GeoSrcUmapCrossEncMaxPct}{0.03}
\newcommand{\GeoSrcUmapWhMixPct}{31}
\newcommand{\GeoSrcUmapMsMixPct}{40}
\newcommand{\GeoSrcNoPcaTsneMaxDiffPp}{1.84}
\newcommand{\GeoSrcNoPcaUmapMaxDiffPp}{2.00}
\newcommand{\GeoSrcRbTsneMaxDiffStepOnePp}{27.60}

\newcommand{\GeoSrcRaUmapMaxDiffStepOnePp}{30.93}
\newcommand{\GeoSrcPanelCPcOnePct}{3.7}
\newcommand{\GeoSrcPanelCPcTwoPct}{1.3}
\newcommand{\GeoSrcPanelCTokOwnPct}{5.0}
\newcommand{\GeoSrcPanelCSpOwnMinPct}{0.12}
\newcommand{\GeoSrcPanelCSpOwnMaxPct}{0.17}
\newcommand{\GeoSrcRandomPlanePct}{0.09}
\newcommand{\GeoSrcPanelDPcOnePct}{5.9}
\newcommand{\GeoSrcPanelDPcTwoPct}{4.0}
\newcommand{\GeoSrcPanelDTokOwnPct}{0.10}
\newcommand{\GeoSrcPanelDSpOwnMinPct}{7.1}
\newcommand{\GeoSrcPanelDSpOwnMaxPct}{12.5}
\newcommand{\GeoSrcPairedCosWh}{0.61}
\newcommand{\GeoSrcPairedCosMs}{0.72}
\newcommand{\GeoSrcXuttCosWh}{0.14}
\newcommand{\GeoSrcXuttCosMs}{0.19}
\newcommand{\GeoSrcTwinTopTenWhPct}{66.4}
\newcommand{\GeoSrcTwinTopTenMsPct}{81.0}
\newcommand{\GeoSrcXuttOtherOptMinPct}{0.9}
\newcommand{\GeoSrcXuttOtherOptMaxPct}{7.1}
\newcommand{\GeoSrcXuttOwnMinPct}{92.9}
\newcommand{\GeoSrcXuttOwnMaxPct}{99.0}
\newcommand{\GeoSrcMeanPairCosTokSpMin}{0.01}
\newcommand{\GeoSrcMeanPairCosTokSpMax}{0.03}
\newcommand{\GeoSrcMeanPairCosTokTok}{0.13}
\newcommand{\GeoSrcMeanPairCosSpSpMin}{0.15}
\newcommand{\GeoSrcMeanPairCosSpSpMax}{0.23}
\newcommand{\GeoSrcMeanDirCosAdaptersMin}{0.71}
\newcommand{\GeoSrcMeanDirCosAdaptersMax}{0.89}
\newcommand{\GeoSrcMeanDirCosTokAdaptersMin}{0.08}
\newcommand{\GeoSrcMeanDirCosTokAdaptersMax}{0.20}
\newcommand{\GeoSrcSixRunMaxDiffPp}{5.25}
\newcommand{\GeoSrcSixRunCentroidRhoMin}{0.05}
\newcommand{\GeoSrcSixRunCentroidRhoMean}{0.42}
\newcommand{\GeoSrcFiveKTokCount}{5{,}000}
\newcommand{\GeoSrcFiveKPcaEvrPct}{39.2}
\newcommand{\GeoTgtDim}{2{,}560}
\newcommand{\GeoTgtNumPops}{11}
\newcommand{\GeoTgtTokPerPop}{5{,}000}
\newcommand{\GeoTgtTotalPoints}{39{,}000}
\newcommand{\GeoTgtPcaEvrPct}{33}
\newcommand{\GeoTgtRadMin}{0.99}
\newcommand{\GeoTgtRadMax}{1.12}
\newcommand{\GeoTgtTsneTokGrpMinPct}{99.96}
\newcommand{\GeoTgtTsneWhGrpMinPct}{99.52}
\newcommand{\GeoTgtTsneMsGrpMinPct}{99.27}

\newcommand{\GeoTgtTsneLsLnSelfPct}{99.91}
\newcommand{\GeoTgtTsneTransSelfPct}{99.91}
\newcommand{\GeoTgtUmapTokGrpMinPct}{99.87}
\newcommand{\GeoTgtUmapWhGrpMinPct}{99.59}
\newcommand{\GeoTgtUmapMsGrpMinPct}{99.62}

\newcommand{\GeoTgtUmapLsLnSelfPct}{99.89}
\newcommand{\GeoTgtUmapTransSelfPct}{99.81}
\newcommand{\GeoTgtSixRunMaxDiffPp}{6.68}
\newcommand{\GeoTgtFullLsLnSelfPct}{100.00}
\newcommand{\GeoTgtFullTransSelfPct}{99.95}
\newcommand{\GeoTgtDisjTokToLsLnPct}{19.7}
\newcommand{\GeoTgtDisjWhOnMsToMsPct}{4.0}
\newcommand{\GeoTgtDisjMsToWhOnMsPct}{14.3}
\newcommand{\GeoTgtPanelCPcOnePct}{1.0}
\newcommand{\GeoTgtPanelCPcTwoPct}{0.6}
\newcommand{\GeoTgtPanelCTokOwnPct}{1.65}
\newcommand{\GeoTgtPanelCFlsOwnPct}{4.20}

\newcommand{\GeoTgtPanelCLsLnOwnPct}{9.80}
\newcommand{\GeoTgtPanelCSpOwnMinPct}{0.09}
\newcommand{\GeoTgtPanelCSpOwnMaxPct}{0.13}
\newcommand{\GeoTgtPanelDPcOnePct}{7.3}
\newcommand{\GeoTgtPanelDPcTwoPct}{4.1}

\newcommand{\GeoTgtFlsNearTokMed}{0.59}
\newcommand{\GeoTgtFprocNearTokMed}{0.51}
\newcommand{\GeoTgtLsLnNearTokMed}{0.24}
\newcommand{\GeoTgtWorkSpNearTokMax}{0.16}
\newcommand{\GeoLinSrcTokTopSixtyFourPct}{27.9}
\newcommand{\GeoLinSrcTokBotHalfPct}{26.5}
\newcommand{\GeoLinSrcSpTopSixtyFourMinPct}{5.4}
\newcommand{\GeoLinSrcSpTopSixtyFourMaxPct}{7.2}
\newcommand{\GeoLinSrcSpBotHalfMinPct}{48.1}
\newcommand{\GeoLinSrcSpBotHalfMaxPct}{48.8}
\newcommand{\GeoLinSrcSpLevRawMin}{1.99}
\newcommand{\GeoLinSrcSpLevRawMax}{2.25}
\newcommand{\GeoLinSrcSpLevRescMin}{1.82}
\newcommand{\GeoLinSrcSpLevRescMax}{2.02}
\newcommand{\GeoLinTgtTokTopSixtyFourPct}{13.8}
\newcommand{\GeoLinTgtTokBotHalfPct}{34.6}
\newcommand{\GeoLinTgtLsLnTopSixtyFourPct}{57.9}
\newcommand{\GeoLinTgtLsLnTopTwoFiveSixPct}{70.9}
\newcommand{\GeoLinTgtLsLnBotHalfPct}{10.5}
\newcommand{\GeoLinTgtLsLnLevRaw}{0.58}
\newcommand{\GeoLinTgtLsLnLevResc}{0.51}
\newcommand{\GeoLinTgtWorkSpBotHalfMinPct}{49.9}
\newcommand{\GeoLinTgtWorkSpBotHalfMaxPct}{51.5}
\newcommand{\GeoLinTgtWorkSpLevRawMin}{1.49}
\newcommand{\GeoLinTgtWorkSpLevRawMax}{1.65}
\newcommand{\GeoLinTgtWorkSpLevRescMin}{1.27}
\newcommand{\GeoLinTgtWorkSpLevRescMax}{1.29}
\newcommand{\GeoSrcFiveKTotalPoints}{17{,}000}
\newcommand{\GeoBalancedPerPop}{1{,}000}
\newcommand{\GeoSrcMapsVsFullOtherMaxDiffPp}{0.26}
\newcommand{\GeoTgtPairCosWhBridge}{0.55}
\newcommand{\GeoTgtXuttCosWhBridge}{0.18}
\newcommand{\GeoTgtPairCosWhJoint}{0.55}
\newcommand{\GeoTgtXuttCosWhJoint}{0.20}
\newcommand{\GeoTgtPairCosMsBridge}{0.62}
\newcommand{\GeoTgtXuttCosMsBridge}{0.19}
\newcommand{\GeoTgtPairCosWhOnMs}{0.39}
\newcommand{\GeoTgtXuttCosWhOnMs}{0.20}
\newcommand{\GeoTgtPairCosLsLn}{0.08}
\newcommand{\GeoTgtXuttCosLsLn}{0.03}
\newcommand{\GeoTgtPairCosTrans}{0.11}
\newcommand{\GeoTgtXuttCosTrans}{0.07}
\newcommand{\GeoTgtSpFitTotal}{18{,}000}
\newcommand{\GeoTgtSpFitWithTokCount}{0}
\newcommand{\GeoTgtTokRowsTotal}{15{,}000}
\newcommand{\GeoTgtTokRowsWithSpFitCount}{2}
\newcommand{\GeoTgtNativeWithLsLnCount}{625}
\newcommand{\GeoTgtFullWhOnMsSelfPct}{97.10}
\newcommand{\GeoTgtFullWhOnMsToMsOnMsPct}{2.76}
\newcommand{\GeoTgtFullWhOnMsToWhGrpPct}{0.14}
\newcommand{\GeoTgtFullTokGrpMinPct}{94.18}
\newcommand{\GeoTgtFullTokToLsLnMaxPct}{5.82}
\newcommand{\GeoTgtFullWhGrpMinPct}{97.90}
\newcommand{\GeoTgtFullMsGrpMinPct}{99.86}
\newcommand{\GeoTgtTokSelfNearTokMed}{0.27}
\newcommand{\GeoTgtTransNearTokMed}{0.10}

\newcommand{\WnormNativeWhisperClean}{3.44}
\newcommand{\WnormNativeWhisperOther}{7.55}
\newcommand{\WnormNativeMoonshineClean}{3.29}
\newcommand{\WnormNativeMoonshineOther}{8.16}
\newcommand{\WnormWhisperGemmaThreeMuonClean}{2.88}
\newcommand{\WnormWhisperGemmaThreeMuonOther}{7.16}
\newcommand{\WnormWhisperGemmaTwoMuonClean}{3.08}
\newcommand{\WnormWhisperGemmaTwoMuonOther}{7.91}
\newcommand{\WnormMoonshineGemmaThreeMuonClean}{3.65}
\newcommand{\WnormMoonshineGemmaThreeMuonOther}{10.04}
\newcommand{\WnormAdapterMinDropPp}{0.09}
\newcommand{\WnormAdapterMaxDropPp}{0.16}
\newcommand{\WnormAdapterCount}{12}
\newcommand{\GeoSubDimWh}{769}
\newcommand{\GeoSubDimMs}{417}
\newcommand{\GeoSubSameEncMinPct}{60.9}
\newcommand{\GeoSubSameEncMaxPct}{66.1}
\newcommand{\GeoSubWhInMsMinPct}{37.2}
\newcommand{\GeoSubWhInMsMaxPct}{42.2}
\newcommand{\GeoSubMsInWhMinPct}{53.8}
\newcommand{\GeoSubMsInWhMaxPct}{55.5}
\newcommand{\GeoSubRandMsPct}{18.1}
\newcommand{\GeoSubRandWhPct}{33.4}
\newcommand{\HeldoutWhisperSingleEmptyPpClean}{19.86}
\newcommand{\HeldoutWhisperSingleRunOnUttsClean}{68}
\newcommand{\HeldoutWhisperSingleRunOnPpClean}{8.83}
\newcommand{\ZeroAudioAdamwLstsqMinClean}{123.58}
\newcommand{\ZeroAudioAdamwLstsqMaxClean}{285.04}
\newcommand{\MuonLastDirCosMin}{0.94}
\newcommand{\MuonLastDirCosMax}{1.00}
\newcommand{\AdamwLastDirCosMax}{0.08}
\newcommand{\MuonActiveKappaMin}{4.75}
\newcommand{\MuonActiveKappaMax}{20.28}
\newcommand{\AdamwActiveKappaMin}{14.28}
\newcommand{\AdamwActiveKappaMax}{28.36}
\newcommand{\MuonActiveKappaLowerPairs}{5}
\newcommand{\MuonCleanMarginMin}{0.11}
\newcommand{\MuonCleanMarginMax}{0.48}

\defmmnum{data.hellaswag.chance_pct}{25.00}
\defmmnum{data.hellaswag.items}{2{,}000}
\defmmnum{data.hellaswag.total_items}{10{,}042}
\defmmnum{data.librispeech.test_clean_utts}{2{,}620}
\defmmnum{data.librispeech.test_other_utts}{2{,}939}
\defmmnum{data.librispeech.train_hours}{960}
\defmmnum{data.librispeech.train_utts}{281{,}241}
\defmmnum{data.librispeech.val_utts}{1{,}000}
\defmmnum{data.piqa.chance_pct}{50.00}
\defmmnum{data.piqa.items}{1{,}838}
\defmmnum{eval.max_new_tokens}{160}
\defmmnum{eval.utt_wer_half_threshold_pct}{50}
\defmmnum{eval.utt_wer_hundred_threshold_pct}{100}
\defmmnum{final_pass.AdamwActiveKappaMax}{28.36}
\defmmnum{final_pass.AdamwActiveKappaMin}{14.28}
\defmmnum{final_pass.AdamwLastDirCosMax}{0.08}
\defmmnum{final_pass.GeoSubDimMs}{417}
\defmmnum{final_pass.GeoSubDimWh}{769}
\defmmnum{final_pass.GeoSubMsInWhMaxPct}{55.5}
\defmmnum{final_pass.GeoSubMsInWhMinPct}{53.8}
\defmmnum{final_pass.GeoSubRandMsPct}{18.1}
\defmmnum{final_pass.GeoSubRandWhPct}{33.4}
\defmmnum{final_pass.GeoSubSameEncMaxPct}{66.1}
\defmmnum{final_pass.GeoSubSameEncMinPct}{60.9}
\defmmnum{final_pass.GeoSubWhInMsMaxPct}{42.2}
\defmmnum{final_pass.GeoSubWhInMsMinPct}{37.2}
\defmmnum{final_pass.HeldoutWhisperSingleEmptyPpClean}{19.86}
\defmmnum{final_pass.HeldoutWhisperSingleRunOnPpClean}{8.83}
\defmmnum{final_pass.HeldoutWhisperSingleRunOnUttsClean}{68}
\defmmnum{final_pass.MuonActiveKappaLowerPairs}{5}
\defmmnum{final_pass.MuonActiveKappaMax}{20.28}
\defmmnum{final_pass.MuonActiveKappaMin}{4.75}
\defmmnum{final_pass.MuonCleanMarginMax}{0.48}
\defmmnum{final_pass.MuonCleanMarginMin}{0.11}
\defmmnum{final_pass.MuonLastDirCosMax}{1.00}
\defmmnum{final_pass.MuonLastDirCosMin}{0.94}
\defmmnum{final_pass.WnormAdapterCount}{12}
\defmmnum{final_pass.WnormAdapterMaxDropPp}{0.16}
\defmmnum{final_pass.WnormAdapterMinDropPp}{0.09}
\defmmnum{final_pass.WnormMoonshineGemmaThreeMuonClean}{3.65}
\defmmnum{final_pass.WnormMoonshineGemmaThreeMuonOther}{10.04}
\defmmnum{final_pass.WnormNativeMoonshineClean}{3.29}
\defmmnum{final_pass.WnormNativeMoonshineOther}{8.16}
\defmmnum{final_pass.WnormNativeWhisperClean}{3.44}
\defmmnum{final_pass.WnormNativeWhisperOther}{7.55}
\defmmnum{final_pass.WnormWhisperGemmaThreeMuonClean}{2.88}
\defmmnum{final_pass.WnormWhisperGemmaThreeMuonOther}{7.16}
\defmmnum{final_pass.WnormWhisperGemmaTwoMuonClean}{3.08}
\defmmnum{final_pass.WnormWhisperGemmaTwoMuonOther}{7.91}
\defmmnum{final_pass.ZeroAudioAdamwLstsqMaxClean}{285.04}
\defmmnum{final_pass.ZeroAudioAdamwLstsqMinClean}{123.58}
\defmmnum{geo.balanced_per_pop}{1{,}000}
\defmmnum{geo.lin.src.sp_bot50_max}{48.8}
\defmmnum{geo.lin.src.sp_bot50_min}{48.1}
\defmmnum{geo.lin.src.sp_lev_raw_max}{2.25}
\defmmnum{geo.lin.src.sp_lev_raw_min}{1.99}
\defmmnum{geo.lin.src.sp_lev_resc_max}{2.02}
\defmmnum{geo.lin.src.sp_lev_resc_min}{1.82}
\defmmnum{geo.lin.src.sp_top64_max}{7.2}
\defmmnum{geo.lin.src.sp_top64_min}{5.4}
\defmmnum{geo.lin.src.t2_bot50}{26.5}
\defmmnum{geo.lin.src.t2_top64}{27.9}
\defmmnum{geo.lin.tgt.lsln_bot50}{10.5}
\defmmnum{geo.lin.tgt.lsln_lev_raw}{0.58}
\defmmnum{geo.lin.tgt.lsln_lev_resc}{0.51}
\defmmnum{geo.lin.tgt.lsln_top256}{70.9}
\defmmnum{geo.lin.tgt.lsln_top64}{57.9}
\defmmnum{geo.lin.tgt.sp_bot50_max}{51.5}
\defmmnum{geo.lin.tgt.sp_bot50_min}{49.9}
\defmmnum{geo.lin.tgt.sp_lev_raw_max}{1.65}
\defmmnum{geo.lin.tgt.sp_lev_raw_min}{1.49}
\defmmnum{geo.lin.tgt.sp_lev_resc_max}{1.29}
\defmmnum{geo.lin.tgt.sp_lev_resc_min}{1.27}
\defmmnum{geo.lin.tgt.t3_bot50}{34.6}
\defmmnum{geo.lin.tgt.t3_top64}{13.8}
\defmmnum{geo.src.5k_pca_evr}{39.2}
\defmmnum{geo.src.5k_tok_count}{5{,}000}
\defmmnum{geo.src.5k_total_points}{17{,}000}
\defmmnum{geo.src.ab.tsne_diff}{1.84}
\defmmnum{geo.src.ab.umap_diff}{2.00}
\defmmnum{geo.src.all4_tok_sp}{0.00}
\defmmnum{geo.src.c.pc1}{3.7}
\defmmnum{geo.src.c.pc2}{1.3}
\defmmnum{geo.src.c.sp_own_max}{0.17}
\defmmnum{geo.src.c.sp_own_min}{0.12}
\defmmnum{geo.src.c.tok_own}{5.0}
\defmmnum{geo.src.d.pc1}{5.9}
\defmmnum{geo.src.d.pc2}{4.0}
\defmmnum{geo.src.d.sp_own_max}{12.5}
\defmmnum{geo.src.d.sp_own_min}{7.1}
\defmmnum{geo.src.d.tok_own}{0.10}
\defmmnum{geo.src.dim}{2{,}304}
\defmmnum{geo.src.disj_tok_self}{100.0}
\defmmnum{geo.src.dup_rows}{0}
\defmmnum{geo.src.frames_per_utt}{15}
\defmmnum{geo.src.maps_vs_full_other_max_diff_pp}{0.26}
\defmmnum{geo.src.md_ad_max}{0.89}
\defmmnum{geo.src.md_ad_min}{0.71}
\defmmnum{geo.src.md_tk_max}{0.20}
\defmmnum{geo.src.md_tk_min}{0.08}
\defmmnum{geo.src.mp_sp_sp_max}{0.23}
\defmmnum{geo.src.mp_sp_sp_min}{0.15}
\defmmnum{geo.src.mp_tok_sp_max}{0.03}
\defmmnum{geo.src.mp_tok_sp_min}{0.01}
\defmmnum{geo.src.mp_tok_tok}{0.13}
\defmmnum{geo.src.n_frames_pop}{3{,}000}
\defmmnum{geo.src.n_soft_prompts}{12{,}000}
\defmmnum{geo.src.n_tokens_kept}{163{,}983}
\defmmnum{geo.src.n_total}{175{,}983}
\defmmnum{geo.src.paired_cos_ms}{0.72}
\defmmnum{geo.src.paired_cos_wh}{0.61}
\defmmnum{geo.src.pca_dims}{50}
\defmmnum{geo.src.ra.tsne_step1_diff}{28.99}
\defmmnum{geo.src.ra.umap_step1_diff}{30.93}
\defmmnum{geo.src.rad_max}{1.09}
\defmmnum{geo.src.rad_min}{1.02}
\defmmnum{geo.src.random_plane_pct}{0.09}
\defmmnum{geo.src.rb.pca_evr_1d}{15.9}
\defmmnum{geo.src.rb.pca_evr_2d}{15.85}
\defmmnum{geo.src.rb.pca_sp_own_max}{33.09}
\defmmnum{geo.src.rb.pca_sp_own_min}{19.84}
\defmmnum{geo.src.rb.pca_tok_own}{13.30}
\defmmnum{geo.src.rb.tsne.ms_mix}{38}
\defmmnum{geo.src.rb.tsne.tok_self}{100.00}
\defmmnum{geo.src.rb.tsne.wh_mix_max}{25}
\defmmnum{geo.src.rb.tsne.wh_mix_min}{24}
\defmmnum{geo.src.rb.tsne.xenc}{0.00}
\defmmnum{geo.src.rb.tsne_step1_diff}{27.60}
\defmmnum{geo.src.rb.umap.ms_mix}{40}
\defmmnum{geo.src.rb.umap.tok_self}{100.00}
\defmmnum{geo.src.rb.umap.wh_mix}{31}
\defmmnum{geo.src.rb.umap.xenc}{0.03}
\defmmnum{geo.src.rb.umap_step1_diff}{30.39}
\defmmnum{geo.src.s6_max_diff}{5.25}
\defmmnum{geo.src.s6_rho_mean}{0.42}
\defmmnum{geo.src.s6_rho_min}{0.05}
\defmmnum{geo.src.step1.ms_to_wh}{0.00}
\defmmnum{geo.src.step1.ms_tok_max}{0.00}
\defmmnum{geo.src.step1.ntc_max}{0.189}
\defmmnum{geo.src.step1.ntc_max_med}{0.135}
\defmmnum{geo.src.step1.ntc_min_med}{0.120}
\defmmnum{geo.src.step1.other_opt_max}{15.5}
\defmmnum{geo.src.step1.other_opt_max_int}{16}
\defmmnum{geo.src.step1.other_opt_min}{5.9}
\defmmnum{geo.src.step1.other_opt_min_int}{6}
\defmmnum{geo.src.step1.sp_with_tok}{2}
\defmmnum{geo.src.step1.tok_ntc_med}{0.628}
\defmmnum{geo.src.step1.tok_self}{100.00}
\defmmnum{geo.src.step1.tok_with_sp}{0}
\defmmnum{geo.src.step1.wh_to_ms}{0.37}
\defmmnum{geo.src.step1.wh_to_ms_1d}{0.4}
\defmmnum{geo.src.step1.wh_tok_max}{0.01}
\defmmnum{geo.src.tok_share_pct}{93.2}
\defmmnum{geo.src.tsne_perp}{30}
\defmmnum{geo.src.twin_top10_ms}{81.0}
\defmmnum{geo.src.twin_top10_wh}{66.4}
\defmmnum{geo.src.umap_nbrs}{30}
\defmmnum{geo.src.xutt_cos_ms}{0.19}
\defmmnum{geo.src.xutt_cos_wh}{0.14}
\defmmnum{geo.src.xutt_other_opt_max}{7.1}
\defmmnum{geo.src.xutt_other_opt_min}{0.9}
\defmmnum{geo.src.xutt_own_max}{99.0}
\defmmnum{geo.src.xutt_own_min}{92.9}
\defmmnum{geo.src.zero_norm_rows}{0}
\defmmnum{geo.tgt.39k.ms3_grp_min}{99.86}
\defmmnum{geo.tgt.39k.native_with_lsln}{625}
\defmmnum{geo.tgt.39k.spfit_total}{18{,}000}
\defmmnum{geo.tgt.39k.spfit_with_tok}{0}
\defmmnum{geo.tgt.39k.tok_grp_min}{94.18}
\defmmnum{geo.tgt.39k.tok_to_lsln_max}{5.82}
\defmmnum{geo.tgt.39k.tok_total}{15{,}000}
\defmmnum{geo.tgt.39k.tok_with_spfit}{2}
\defmmnum{geo.tgt.39k.wh3_grp_min}{97.90}
\defmmnum{geo.tgt.39k.wh_on_ms_self}{97.10}
\defmmnum{geo.tgt.39k.wh_on_ms_to_ms_on_ms}{2.76}
\defmmnum{geo.tgt.39k.wh_on_ms_to_wh_grp}{0.14}
\defmmnum{geo.tgt.Tsne.lsln_self}{99.91}
\defmmnum{geo.tgt.Tsne.ms_grp_min}{99.27}
\defmmnum{geo.tgt.Tsne.tok_grp_min}{99.96}
\defmmnum{geo.tgt.Tsne.trans_self}{99.91}
\defmmnum{geo.tgt.Tsne.wh_grp_min}{99.52}
\defmmnum{geo.tgt.Tsne.wh_on_ms_in_ms}{99.27}
\defmmnum{geo.tgt.Umap.lsln_self}{99.89}
\defmmnum{geo.tgt.Umap.ms_grp_min}{99.62}
\defmmnum{geo.tgt.Umap.tok_grp_min}{99.87}
\defmmnum{geo.tgt.Umap.trans_self}{99.81}
\defmmnum{geo.tgt.Umap.wh_grp_min}{99.59}
\defmmnum{geo.tgt.Umap.wh_on_ms_in_ms}{99.62}
\defmmnum{geo.tgt.c.fls_own}{4.20}
\defmmnum{geo.tgt.c.fproc_own}{2.98}
\defmmnum{geo.tgt.c.lsln_own}{9.80}
\defmmnum{geo.tgt.c.pc1}{1.0}
\defmmnum{geo.tgt.c.pc2}{0.6}
\defmmnum{geo.tgt.c.sp_own_max}{0.13}
\defmmnum{geo.tgt.c.sp_own_min}{0.09}
\defmmnum{geo.tgt.c.tok_own}{1.65}
\defmmnum{geo.tgt.d.lsln_own}{5.68}
\defmmnum{geo.tgt.d.pc1}{7.3}
\defmmnum{geo.tgt.d.pc2}{4.1}
\defmmnum{geo.tgt.d.tok_own}{0.33}
\defmmnum{geo.tgt.dim}{2{,}560}
\defmmnum{geo.tgt.disj.ms_to_wh_ms}{14.3}
\defmmnum{geo.tgt.disj.t3_lsln}{19.7}
\defmmnum{geo.tgt.disj.wh_ms_to_ms}{4.0}
\defmmnum{geo.tgt.disj_lsln_self}{100.0}
\defmmnum{geo.tgt.disj_trans_self}{99.9}
\defmmnum{geo.tgt.full39k.lsln_self}{100.00}
\defmmnum{geo.tgt.full39k.trans_self}{99.95}
\defmmnum{geo.tgt.n_points}{39{,}000}
\defmmnum{geo.tgt.n_pops}{11}
\defmmnum{geo.tgt.nt3.fls_med}{0.59}
\defmmnum{geo.tgt.nt3.fproc_med}{0.51}
\defmmnum{geo.tgt.nt3.lsln_med}{0.24}
\defmmnum{geo.tgt.nt3.t3_self_med}{0.27}
\defmmnum{geo.tgt.nt3.trans_med}{0.10}
\defmmnum{geo.tgt.nt3.work_sp_max}{0.16}
\defmmnum{geo.tgt.pair_cos.lsln}{0.08}
\defmmnum{geo.tgt.pair_cos.msbridge}{0.62}
\defmmnum{geo.tgt.pair_cos.trans}{0.11}
\defmmnum{geo.tgt.pair_cos.whbridge}{0.55}
\defmmnum{geo.tgt.pair_cos.whjoint}{0.55}
\defmmnum{geo.tgt.pair_cos.whonms}{0.39}
\defmmnum{geo.tgt.pca_evr}{33}
\defmmnum{geo.tgt.rad_max}{1.12}
\defmmnum{geo.tgt.rad_min}{0.99}
\defmmnum{geo.tgt.s6_max_diff}{6.68}
\defmmnum{geo.tgt.tok_per_pop}{5{,}000}
\defmmnum{geo.tgt.xutt_cos.lsln}{0.03}
\defmmnum{geo.tgt.xutt_cos.msbridge}{0.19}
\defmmnum{geo.tgt.xutt_cos.trans}{0.07}
\defmmnum{geo.tgt.xutt_cos.whbridge}{0.18}
\defmmnum{geo.tgt.xutt_cos.whjoint}{0.20}
\defmmnum{geo.tgt.xutt_cos.whonms}{0.20}
\defmmnum{grid.done_jobs}{69}
\defmmnum{grid.fleet_hosts}{15}
\defmmnum{grid.total_jobs}{69}
\defmmnum{manifold.map.lstsq_ln_max_nn_cos_med}{0.6311}
\defmmnum{manifold.map.lstsq_ln_min_nn_cos_med}{0.4152}
\defmmnum{manifold.map.lstsq_scaled_max_nn_cos_med}{0.3127}
\defmmnum{manifold.map.lstsq_scaled_min_nn_cos_med}{0.2130}
\defmmnum{manifold.map.moonshine_gemma2_ln__to_qwen25__lstsq_ln.cos_mean.median}{+0.5510}
\defmmnum{manifold.map.moonshine_gemma2_ln__to_qwen25__lstsq_ln.nn_cos.median}{0.5119}
\defmmnum{manifold.map.moonshine_gemma2_ln__to_qwen25__lstsq_ln.nn_cos.p10}{0.3290}
\defmmnum{manifold.map.moonshine_gemma2_ln__to_qwen25__lstsq_ln.nn_cos.p90}{0.6526}
\defmmnum{manifold.map.moonshine_gemma2_ln__to_qwen25__lstsq_ln.pca_k1024.median_pct}{85.2}
\defmmnum{manifold.map.moonshine_gemma2_ln__to_qwen25__lstsq_ln.pca_k256.median_pct}{49.1}
\defmmnum{manifold.map.moonshine_gemma2_ln__to_qwen25__lstsq_ln.pca_k64.median_pct}{25.8}
\defmmnum{manifold.map.moonshine_gemma2_ln__to_qwen25__lstsq_ln.rr.median}{1.086}
\defmmnum{manifold.map.moonshine_gemma2_ln__to_qwen25__lstsq_ln.rr.p10}{1.009}
\defmmnum{manifold.map.moonshine_gemma2_ln__to_qwen25__lstsq_ln.rr.p90}{1.152}
\defmmnum{manifold.map.moonshine_gemma2_ln__to_qwen25__lstsq_scaled.cos_mean.median}{+0.3142}
\defmmnum{manifold.map.moonshine_gemma2_ln__to_qwen25__lstsq_scaled.nn_cos.median}{0.3127}
\defmmnum{manifold.map.moonshine_gemma2_ln__to_qwen25__lstsq_scaled.nn_cos.p10}{0.1559}
\defmmnum{manifold.map.moonshine_gemma2_ln__to_qwen25__lstsq_scaled.nn_cos.p90}{0.5032}
\defmmnum{manifold.map.moonshine_gemma2_ln__to_qwen25__lstsq_scaled.pca_k1024.median_pct}{82.9}
\defmmnum{manifold.map.moonshine_gemma2_ln__to_qwen25__lstsq_scaled.pca_k256.median_pct}{44.9}
\defmmnum{manifold.map.moonshine_gemma2_ln__to_qwen25__lstsq_scaled.pca_k64.median_pct}{20.8}
\defmmnum{manifold.map.moonshine_gemma2_ln__to_qwen25__lstsq_scaled.rr.median}{1.000}
\defmmnum{manifold.map.moonshine_gemma2_ln__to_qwen25__lstsq_scaled.rr.p10}{1.000}
\defmmnum{manifold.map.moonshine_gemma2_ln__to_qwen25__lstsq_scaled.rr.p90}{1.000}
\defmmnum{manifold.map.moonshine_gemma2_ln__to_qwen25__procrustes_scaled.cos_mean.median}{+0.1091}
\defmmnum{manifold.map.moonshine_gemma2_ln__to_qwen25__procrustes_scaled.nn_cos.median}{0.1531}
\defmmnum{manifold.map.moonshine_gemma2_ln__to_qwen25__procrustes_scaled.nn_cos.p10}{0.1342}
\defmmnum{manifold.map.moonshine_gemma2_ln__to_qwen25__procrustes_scaled.nn_cos.p90}{0.1836}
\defmmnum{manifold.map.moonshine_gemma2_ln__to_qwen25__procrustes_scaled.pca_k1024.median_pct}{67.9}
\defmmnum{manifold.map.moonshine_gemma2_ln__to_qwen25__procrustes_scaled.pca_k256.median_pct}{23.5}
\defmmnum{manifold.map.moonshine_gemma2_ln__to_qwen25__procrustes_scaled.pca_k64.median_pct}{11.2}
\defmmnum{manifold.map.moonshine_gemma2_ln__to_qwen25__procrustes_scaled.rr.median}{1.000}
\defmmnum{manifold.map.moonshine_gemma2_ln__to_qwen25__procrustes_scaled.rr.p10}{1.000}
\defmmnum{manifold.map.moonshine_gemma2_ln__to_qwen25__procrustes_scaled.rr.p90}{1.000}
\defmmnum{manifold.map.moonshine_gemma2_ln__to_qwen25__wte_ln.cos_mean.median}{+0.1735}
\defmmnum{manifold.map.moonshine_gemma2_ln__to_qwen25__wte_ln.nn_cos.median}{0.1974}
\defmmnum{manifold.map.moonshine_gemma2_ln__to_qwen25__wte_ln.nn_cos.p10}{0.1796}
\defmmnum{manifold.map.moonshine_gemma2_ln__to_qwen25__wte_ln.nn_cos.p90}{0.2195}
\defmmnum{manifold.map.moonshine_gemma2_ln__to_qwen25__wte_ln.pca_k1024.median_pct}{73.8}
\defmmnum{manifold.map.moonshine_gemma2_ln__to_qwen25__wte_ln.pca_k256.median_pct}{34.0}
\defmmnum{manifold.map.moonshine_gemma2_ln__to_qwen25__wte_ln.pca_k64.median_pct}{12.4}
\defmmnum{manifold.map.moonshine_gemma2_ln__to_qwen25__wte_ln.rr.median}{3.748}
\defmmnum{manifold.map.moonshine_gemma2_ln__to_qwen25__wte_ln.rr.p10}{3.681}
\defmmnum{manifold.map.moonshine_gemma2_ln__to_qwen25__wte_ln.rr.p90}{3.813}
\defmmnum{manifold.map.moonshine_gemma2_ln__to_qwen25__wte_scaled.cos_mean.median}{+0.1105}
\defmmnum{manifold.map.moonshine_gemma2_ln__to_qwen25__wte_scaled.nn_cos.median}{0.1725}
\defmmnum{manifold.map.moonshine_gemma2_ln__to_qwen25__wte_scaled.nn_cos.p10}{0.1470}
\defmmnum{manifold.map.moonshine_gemma2_ln__to_qwen25__wte_scaled.nn_cos.p90}{0.1979}
\defmmnum{manifold.map.moonshine_gemma2_ln__to_qwen25__wte_scaled.pca_k1024.median_pct}{72.6}
\defmmnum{manifold.map.moonshine_gemma2_ln__to_qwen25__wte_scaled.pca_k256.median_pct}{37.0}
\defmmnum{manifold.map.moonshine_gemma2_ln__to_qwen25__wte_scaled.pca_k64.median_pct}{13.2}
\defmmnum{manifold.map.moonshine_gemma2_ln__to_qwen25__wte_scaled.rr.median}{3.405}
\defmmnum{manifold.map.moonshine_gemma2_ln__to_qwen25__wte_scaled.rr.p10}{3.405}
\defmmnum{manifold.map.moonshine_gemma2_ln__to_qwen25__wte_scaled.rr.p90}{3.405}
\defmmnum{manifold.map.moonshine_qwen25_ln__to_gemma2__lstsq_ln.cos_mean.median}{+0.6890}
\defmmnum{manifold.map.moonshine_qwen25_ln__to_gemma2__lstsq_ln.nn_cos.median}{0.6311}
\defmmnum{manifold.map.moonshine_qwen25_ln__to_gemma2__lstsq_ln.nn_cos.p10}{0.4853}
\defmmnum{manifold.map.moonshine_qwen25_ln__to_gemma2__lstsq_ln.nn_cos.p90}{0.7175}
\defmmnum{manifold.map.moonshine_qwen25_ln__to_gemma2__lstsq_ln.pca_k1024.median_pct}{87.2}
\defmmnum{manifold.map.moonshine_qwen25_ln__to_gemma2__lstsq_ln.pca_k256.median_pct}{72.3}
\defmmnum{manifold.map.moonshine_qwen25_ln__to_gemma2__lstsq_ln.pca_k64.median_pct}{63.5}
\defmmnum{manifold.map.moonshine_qwen25_ln__to_gemma2__lstsq_ln.rr.median}{1.663}
\defmmnum{manifold.map.moonshine_qwen25_ln__to_gemma2__lstsq_ln.rr.p10}{1.375}
\defmmnum{manifold.map.moonshine_qwen25_ln__to_gemma2__lstsq_ln.rr.p90}{1.935}
\defmmnum{manifold.map.moonshine_qwen25_ln__to_gemma2__lstsq_scaled.cos_mean.median}{+0.2527}
\defmmnum{manifold.map.moonshine_qwen25_ln__to_gemma2__lstsq_scaled.nn_cos.median}{0.2944}
\defmmnum{manifold.map.moonshine_qwen25_ln__to_gemma2__lstsq_scaled.nn_cos.p10}{0.1959}
\defmmnum{manifold.map.moonshine_qwen25_ln__to_gemma2__lstsq_scaled.nn_cos.p90}{0.4348}
\defmmnum{manifold.map.moonshine_qwen25_ln__to_gemma2__lstsq_scaled.pca_k1024.median_pct}{73.4}
\defmmnum{manifold.map.moonshine_qwen25_ln__to_gemma2__lstsq_scaled.pca_k256.median_pct}{42.1}
\defmmnum{manifold.map.moonshine_qwen25_ln__to_gemma2__lstsq_scaled.pca_k64.median_pct}{28.7}
\defmmnum{manifold.map.moonshine_qwen25_ln__to_gemma2__lstsq_scaled.rr.median}{1.012}
\defmmnum{manifold.map.moonshine_qwen25_ln__to_gemma2__lstsq_scaled.rr.p10}{1.012}
\defmmnum{manifold.map.moonshine_qwen25_ln__to_gemma2__lstsq_scaled.rr.p90}{1.012}
\defmmnum{manifold.map.moonshine_qwen25_ln__to_gemma2__procrustes_scaled.cos_mean.median}{+0.0041}
\defmmnum{manifold.map.moonshine_qwen25_ln__to_gemma2__procrustes_scaled.nn_cos.median}{0.0977}
\defmmnum{manifold.map.moonshine_qwen25_ln__to_gemma2__procrustes_scaled.nn_cos.p10}{0.0849}
\defmmnum{manifold.map.moonshine_qwen25_ln__to_gemma2__procrustes_scaled.nn_cos.p90}{0.1121}
\defmmnum{manifold.map.moonshine_qwen25_ln__to_gemma2__procrustes_scaled.pca_k1024.median_pct}{58.5}
\defmmnum{manifold.map.moonshine_qwen25_ln__to_gemma2__procrustes_scaled.pca_k256.median_pct}{27.0}
\defmmnum{manifold.map.moonshine_qwen25_ln__to_gemma2__procrustes_scaled.pca_k64.median_pct}{18.4}
\defmmnum{manifold.map.moonshine_qwen25_ln__to_gemma2__procrustes_scaled.rr.median}{1.012}
\defmmnum{manifold.map.moonshine_qwen25_ln__to_gemma2__procrustes_scaled.rr.p10}{1.012}
\defmmnum{manifold.map.moonshine_qwen25_ln__to_gemma2__procrustes_scaled.rr.p90}{1.012}
\defmmnum{manifold.map.moonshine_qwen25_ln__to_gemma2__wte_ln.cos_mean.median}{+0.0713}
\defmmnum{manifold.map.moonshine_qwen25_ln__to_gemma2__wte_ln.nn_cos.median}{0.1182}
\defmmnum{manifold.map.moonshine_qwen25_ln__to_gemma2__wte_ln.nn_cos.p10}{0.1040}
\defmmnum{manifold.map.moonshine_qwen25_ln__to_gemma2__wte_ln.nn_cos.p90}{0.1332}
\defmmnum{manifold.map.moonshine_qwen25_ln__to_gemma2__wte_ln.pca_k1024.median_pct}{53.9}
\defmmnum{manifold.map.moonshine_qwen25_ln__to_gemma2__wte_ln.pca_k256.median_pct}{25.8}
\defmmnum{manifold.map.moonshine_qwen25_ln__to_gemma2__wte_ln.pca_k64.median_pct}{15.3}
\defmmnum{manifold.map.moonshine_qwen25_ln__to_gemma2__wte_ln.rr.median}{1.023}
\defmmnum{manifold.map.moonshine_qwen25_ln__to_gemma2__wte_ln.rr.p10}{1.017}
\defmmnum{manifold.map.moonshine_qwen25_ln__to_gemma2__wte_ln.rr.p90}{1.034}
\defmmnum{manifold.map.proc_max_nn_cos_med}{0.1531}
\defmmnum{manifold.map.proc_min_nn_cos_med}{0.0947}
\defmmnum{manifold.map.transcript_max_nn_cos_med}{0.1977}
\defmmnum{manifold.map.transcript_min_nn_cos_med}{0.1182}
\defmmnum{manifold.map.whisper_gemma2_ln__to_qwen25__lstsq_ln.cos_mean.median}{+0.4432}
\defmmnum{manifold.map.whisper_gemma2_ln__to_qwen25__lstsq_ln.nn_cos.median}{0.4152}
\defmmnum{manifold.map.whisper_gemma2_ln__to_qwen25__lstsq_ln.nn_cos.p10}{0.2153}
\defmmnum{manifold.map.whisper_gemma2_ln__to_qwen25__lstsq_ln.nn_cos.p90}{0.5742}
\defmmnum{manifold.map.whisper_gemma2_ln__to_qwen25__lstsq_ln.pca_k1024.median_pct}{84.3}
\defmmnum{manifold.map.whisper_gemma2_ln__to_qwen25__lstsq_ln.pca_k256.median_pct}{45.9}
\defmmnum{manifold.map.whisper_gemma2_ln__to_qwen25__lstsq_ln.pca_k64.median_pct}{22.0}
\defmmnum{manifold.map.whisper_gemma2_ln__to_qwen25__lstsq_ln.rr.median}{1.047}
\defmmnum{manifold.map.whisper_gemma2_ln__to_qwen25__lstsq_ln.rr.p10}{0.966}
\defmmnum{manifold.map.whisper_gemma2_ln__to_qwen25__lstsq_ln.rr.p90}{1.120}
\defmmnum{manifold.map.whisper_gemma2_ln__to_qwen25__lstsq_scaled.cos_mean.median}{+0.1773}
\defmmnum{manifold.map.whisper_gemma2_ln__to_qwen25__lstsq_scaled.nn_cos.median}{0.2130}
\defmmnum{manifold.map.whisper_gemma2_ln__to_qwen25__lstsq_scaled.nn_cos.p10}{0.1352}
\defmmnum{manifold.map.whisper_gemma2_ln__to_qwen25__lstsq_scaled.nn_cos.p90}{0.3940}
\defmmnum{manifold.map.whisper_gemma2_ln__to_qwen25__lstsq_scaled.pca_k1024.median_pct}{82.9}
\defmmnum{manifold.map.whisper_gemma2_ln__to_qwen25__lstsq_scaled.pca_k256.median_pct}{44.9}
\defmmnum{manifold.map.whisper_gemma2_ln__to_qwen25__lstsq_scaled.pca_k64.median_pct}{21.2}
\defmmnum{manifold.map.whisper_gemma2_ln__to_qwen25__lstsq_scaled.rr.median}{1.000}
\defmmnum{manifold.map.whisper_gemma2_ln__to_qwen25__lstsq_scaled.rr.p10}{1.000}
\defmmnum{manifold.map.whisper_gemma2_ln__to_qwen25__lstsq_scaled.rr.p90}{1.000}
\defmmnum{manifold.map.whisper_gemma2_ln__to_qwen25__procrustes_scaled.cos_mean.median}{+0.0790}
\defmmnum{manifold.map.whisper_gemma2_ln__to_qwen25__procrustes_scaled.nn_cos.median}{0.1401}
\defmmnum{manifold.map.whisper_gemma2_ln__to_qwen25__procrustes_scaled.nn_cos.p10}{0.1227}
\defmmnum{manifold.map.whisper_gemma2_ln__to_qwen25__procrustes_scaled.nn_cos.p90}{0.1622}
\defmmnum{manifold.map.whisper_gemma2_ln__to_qwen25__procrustes_scaled.pca_k1024.median_pct}{68.3}
\defmmnum{manifold.map.whisper_gemma2_ln__to_qwen25__procrustes_scaled.pca_k256.median_pct}{24.0}
\defmmnum{manifold.map.whisper_gemma2_ln__to_qwen25__procrustes_scaled.pca_k64.median_pct}{11.5}
\defmmnum{manifold.map.whisper_gemma2_ln__to_qwen25__procrustes_scaled.rr.median}{1.000}
\defmmnum{manifold.map.whisper_gemma2_ln__to_qwen25__procrustes_scaled.rr.p10}{1.000}
\defmmnum{manifold.map.whisper_gemma2_ln__to_qwen25__procrustes_scaled.rr.p90}{1.000}
\defmmnum{manifold.map.whisper_gemma2_ln__to_qwen25__wte_ln.cos_mean.median}{+0.1746}
\defmmnum{manifold.map.whisper_gemma2_ln__to_qwen25__wte_ln.nn_cos.median}{0.1977}
\defmmnum{manifold.map.whisper_gemma2_ln__to_qwen25__wte_ln.nn_cos.p10}{0.1800}
\defmmnum{manifold.map.whisper_gemma2_ln__to_qwen25__wte_ln.nn_cos.p90}{0.2179}
\defmmnum{manifold.map.whisper_gemma2_ln__to_qwen25__wte_ln.pca_k1024.median_pct}{74.4}
\defmmnum{manifold.map.whisper_gemma2_ln__to_qwen25__wte_ln.pca_k256.median_pct}{34.8}
\defmmnum{manifold.map.whisper_gemma2_ln__to_qwen25__wte_ln.pca_k64.median_pct}{12.6}
\defmmnum{manifold.map.whisper_gemma2_ln__to_qwen25__wte_ln.rr.median}{3.717}
\defmmnum{manifold.map.whisper_gemma2_ln__to_qwen25__wte_ln.rr.p10}{3.632}
\defmmnum{manifold.map.whisper_gemma2_ln__to_qwen25__wte_ln.rr.p90}{3.790}
\defmmnum{manifold.map.whisper_gemma2_ln__to_qwen25__wte_scaled.cos_mean.median}{+0.0911}
\defmmnum{manifold.map.whisper_gemma2_ln__to_qwen25__wte_scaled.nn_cos.median}{0.1574}
\defmmnum{manifold.map.whisper_gemma2_ln__to_qwen25__wte_scaled.nn_cos.p10}{0.1357}
\defmmnum{manifold.map.whisper_gemma2_ln__to_qwen25__wte_scaled.nn_cos.p90}{0.1817}
\defmmnum{manifold.map.whisper_gemma2_ln__to_qwen25__wte_scaled.pca_k1024.median_pct}{72.7}
\defmmnum{manifold.map.whisper_gemma2_ln__to_qwen25__wte_scaled.pca_k256.median_pct}{36.8}
\defmmnum{manifold.map.whisper_gemma2_ln__to_qwen25__wte_scaled.pca_k64.median_pct}{12.3}
\defmmnum{manifold.map.whisper_gemma2_ln__to_qwen25__wte_scaled.rr.median}{3.405}
\defmmnum{manifold.map.whisper_gemma2_ln__to_qwen25__wte_scaled.rr.p10}{3.405}
\defmmnum{manifold.map.whisper_gemma2_ln__to_qwen25__wte_scaled.rr.p90}{3.405}
\defmmnum{manifold.map.whisper_gemma2_scaled__to_qwen25__lstsq_ln.cos_mean.median}{+0.4486}
\defmmnum{manifold.map.whisper_gemma2_scaled__to_qwen25__lstsq_ln.nn_cos.median}{0.4218}
\defmmnum{manifold.map.whisper_gemma2_scaled__to_qwen25__lstsq_ln.nn_cos.p10}{0.2349}
\defmmnum{manifold.map.whisper_gemma2_scaled__to_qwen25__lstsq_ln.nn_cos.p90}{0.5772}
\defmmnum{manifold.map.whisper_gemma2_scaled__to_qwen25__lstsq_ln.pca_k1024.median_pct}{84.1}
\defmmnum{manifold.map.whisper_gemma2_scaled__to_qwen25__lstsq_ln.pca_k256.median_pct}{46.0}
\defmmnum{manifold.map.whisper_gemma2_scaled__to_qwen25__lstsq_ln.pca_k64.median_pct}{22.0}
\defmmnum{manifold.map.whisper_gemma2_scaled__to_qwen25__lstsq_ln.rr.median}{1.049}
\defmmnum{manifold.map.whisper_gemma2_scaled__to_qwen25__lstsq_ln.rr.p10}{0.975}
\defmmnum{manifold.map.whisper_gemma2_scaled__to_qwen25__lstsq_ln.rr.p90}{1.120}
\defmmnum{manifold.map.whisper_gemma2_scaled__to_qwen25__lstsq_scaled.cos_mean.median}{+0.1829}
\defmmnum{manifold.map.whisper_gemma2_scaled__to_qwen25__lstsq_scaled.nn_cos.median}{0.2155}
\defmmnum{manifold.map.whisper_gemma2_scaled__to_qwen25__lstsq_scaled.nn_cos.p10}{0.1366}
\defmmnum{manifold.map.whisper_gemma2_scaled__to_qwen25__lstsq_scaled.nn_cos.p90}{0.3960}
\defmmnum{manifold.map.whisper_gemma2_scaled__to_qwen25__lstsq_scaled.pca_k1024.median_pct}{82.7}
\defmmnum{manifold.map.whisper_gemma2_scaled__to_qwen25__lstsq_scaled.pca_k256.median_pct}{45.0}
\defmmnum{manifold.map.whisper_gemma2_scaled__to_qwen25__lstsq_scaled.pca_k64.median_pct}{21.3}
\defmmnum{manifold.map.whisper_gemma2_scaled__to_qwen25__lstsq_scaled.rr.median}{1.000}
\defmmnum{manifold.map.whisper_gemma2_scaled__to_qwen25__lstsq_scaled.rr.p10}{1.000}
\defmmnum{manifold.map.whisper_gemma2_scaled__to_qwen25__lstsq_scaled.rr.p90}{1.000}
\defmmnum{manifold.map.whisper_gemma2_scaled__to_qwen25__procrustes_scaled.cos_mean.median}{+0.0773}
\defmmnum{manifold.map.whisper_gemma2_scaled__to_qwen25__procrustes_scaled.nn_cos.median}{0.1395}
\defmmnum{manifold.map.whisper_gemma2_scaled__to_qwen25__procrustes_scaled.nn_cos.p10}{0.1209}
\defmmnum{manifold.map.whisper_gemma2_scaled__to_qwen25__procrustes_scaled.nn_cos.p90}{0.1633}
\defmmnum{manifold.map.whisper_gemma2_scaled__to_qwen25__procrustes_scaled.pca_k1024.median_pct}{68.3}
\defmmnum{manifold.map.whisper_gemma2_scaled__to_qwen25__procrustes_scaled.pca_k256.median_pct}{23.9}
\defmmnum{manifold.map.whisper_gemma2_scaled__to_qwen25__procrustes_scaled.pca_k64.median_pct}{11.6}
\defmmnum{manifold.map.whisper_gemma2_scaled__to_qwen25__procrustes_scaled.rr.median}{1.000}
\defmmnum{manifold.map.whisper_gemma2_scaled__to_qwen25__procrustes_scaled.rr.p10}{1.000}
\defmmnum{manifold.map.whisper_gemma2_scaled__to_qwen25__procrustes_scaled.rr.p90}{1.000}
\defmmnum{manifold.map.whisper_gemma2_scaled__to_qwen25__wte_ln.cos_mean.median}{+0.1678}
\defmmnum{manifold.map.whisper_gemma2_scaled__to_qwen25__wte_ln.nn_cos.median}{0.1934}
\defmmnum{manifold.map.whisper_gemma2_scaled__to_qwen25__wte_ln.nn_cos.p10}{0.1757}
\defmmnum{manifold.map.whisper_gemma2_scaled__to_qwen25__wte_ln.nn_cos.p90}{0.2134}
\defmmnum{manifold.map.whisper_gemma2_scaled__to_qwen25__wte_ln.pca_k1024.median_pct}{74.5}
\defmmnum{manifold.map.whisper_gemma2_scaled__to_qwen25__wte_ln.pca_k256.median_pct}{34.6}
\defmmnum{manifold.map.whisper_gemma2_scaled__to_qwen25__wte_ln.pca_k64.median_pct}{12.6}
\defmmnum{manifold.map.whisper_gemma2_scaled__to_qwen25__wte_ln.rr.median}{3.702}
\defmmnum{manifold.map.whisper_gemma2_scaled__to_qwen25__wte_ln.rr.p10}{3.618}
\defmmnum{manifold.map.whisper_gemma2_scaled__to_qwen25__wte_ln.rr.p90}{3.777}
\defmmnum{manifold.map.whisper_gemma2_scaled__to_qwen25__wte_scaled.cos_mean.median}{+0.0812}
\defmmnum{manifold.map.whisper_gemma2_scaled__to_qwen25__wte_scaled.nn_cos.median}{0.1532}
\defmmnum{manifold.map.whisper_gemma2_scaled__to_qwen25__wte_scaled.nn_cos.p10}{0.1319}
\defmmnum{manifold.map.whisper_gemma2_scaled__to_qwen25__wte_scaled.nn_cos.p90}{0.1764}
\defmmnum{manifold.map.whisper_gemma2_scaled__to_qwen25__wte_scaled.pca_k1024.median_pct}{73.0}
\defmmnum{manifold.map.whisper_gemma2_scaled__to_qwen25__wte_scaled.pca_k256.median_pct}{37.2}
\defmmnum{manifold.map.whisper_gemma2_scaled__to_qwen25__wte_scaled.pca_k64.median_pct}{12.3}
\defmmnum{manifold.map.whisper_gemma2_scaled__to_qwen25__wte_scaled.rr.median}{3.405}
\defmmnum{manifold.map.whisper_gemma2_scaled__to_qwen25__wte_scaled.rr.p10}{3.405}
\defmmnum{manifold.map.whisper_gemma2_scaled__to_qwen25__wte_scaled.rr.p90}{3.405}
\defmmnum{manifold.map.whisper_qwen25_ln__to_gemma2__lstsq_ln.cos_mean.median}{+0.6345}
\defmmnum{manifold.map.whisper_qwen25_ln__to_gemma2__lstsq_ln.nn_cos.median}{0.5804}
\defmmnum{manifold.map.whisper_qwen25_ln__to_gemma2__lstsq_ln.nn_cos.p10}{0.3806}
\defmmnum{manifold.map.whisper_qwen25_ln__to_gemma2__lstsq_ln.nn_cos.p90}{0.6762}
\defmmnum{manifold.map.whisper_qwen25_ln__to_gemma2__lstsq_ln.pca_k1024.median_pct}{81.9}
\defmmnum{manifold.map.whisper_qwen25_ln__to_gemma2__lstsq_ln.pca_k256.median_pct}{60.8}
\defmmnum{manifold.map.whisper_qwen25_ln__to_gemma2__lstsq_ln.pca_k64.median_pct}{49.3}
\defmmnum{manifold.map.whisper_qwen25_ln__to_gemma2__lstsq_ln.rr.median}{1.444}
\defmmnum{manifold.map.whisper_qwen25_ln__to_gemma2__lstsq_ln.rr.p10}{1.173}
\defmmnum{manifold.map.whisper_qwen25_ln__to_gemma2__lstsq_ln.rr.p90}{1.724}
\defmmnum{manifold.map.whisper_qwen25_ln__to_gemma2__lstsq_scaled.cos_mean.median}{+0.1916}
\defmmnum{manifold.map.whisper_qwen25_ln__to_gemma2__lstsq_scaled.nn_cos.median}{0.2418}
\defmmnum{manifold.map.whisper_qwen25_ln__to_gemma2__lstsq_scaled.nn_cos.p10}{0.1752}
\defmmnum{manifold.map.whisper_qwen25_ln__to_gemma2__lstsq_scaled.nn_cos.p90}{0.3536}
\defmmnum{manifold.map.whisper_qwen25_ln__to_gemma2__lstsq_scaled.pca_k1024.median_pct}{72.6}
\defmmnum{manifold.map.whisper_qwen25_ln__to_gemma2__lstsq_scaled.pca_k256.median_pct}{40.9}
\defmmnum{manifold.map.whisper_qwen25_ln__to_gemma2__lstsq_scaled.pca_k64.median_pct}{27.5}
\defmmnum{manifold.map.whisper_qwen25_ln__to_gemma2__lstsq_scaled.rr.median}{1.012}
\defmmnum{manifold.map.whisper_qwen25_ln__to_gemma2__lstsq_scaled.rr.p10}{1.012}
\defmmnum{manifold.map.whisper_qwen25_ln__to_gemma2__lstsq_scaled.rr.p90}{1.012}
\defmmnum{manifold.map.whisper_qwen25_ln__to_gemma2__procrustes_scaled.cos_mean.median}{+0.0009}
\defmmnum{manifold.map.whisper_qwen25_ln__to_gemma2__procrustes_scaled.nn_cos.median}{0.0947}
\defmmnum{manifold.map.whisper_qwen25_ln__to_gemma2__procrustes_scaled.nn_cos.p10}{0.0795}
\defmmnum{manifold.map.whisper_qwen25_ln__to_gemma2__procrustes_scaled.nn_cos.p90}{0.1145}
\defmmnum{manifold.map.whisper_qwen25_ln__to_gemma2__procrustes_scaled.pca_k1024.median_pct}{57.9}
\defmmnum{manifold.map.whisper_qwen25_ln__to_gemma2__procrustes_scaled.pca_k256.median_pct}{26.7}
\defmmnum{manifold.map.whisper_qwen25_ln__to_gemma2__procrustes_scaled.pca_k64.median_pct}{18.2}
\defmmnum{manifold.map.whisper_qwen25_ln__to_gemma2__procrustes_scaled.rr.median}{1.012}
\defmmnum{manifold.map.whisper_qwen25_ln__to_gemma2__procrustes_scaled.rr.p10}{1.012}
\defmmnum{manifold.map.whisper_qwen25_ln__to_gemma2__procrustes_scaled.rr.p90}{1.012}
\defmmnum{manifold.map.whisper_qwen25_ln__to_gemma2__wte_ln.cos_mean.median}{+0.0805}
\defmmnum{manifold.map.whisper_qwen25_ln__to_gemma2__wte_ln.nn_cos.median}{0.1220}
\defmmnum{manifold.map.whisper_qwen25_ln__to_gemma2__wte_ln.nn_cos.p10}{0.1125}
\defmmnum{manifold.map.whisper_qwen25_ln__to_gemma2__wte_ln.nn_cos.p90}{0.1325}
\defmmnum{manifold.map.whisper_qwen25_ln__to_gemma2__wte_ln.pca_k1024.median_pct}{53.2}
\defmmnum{manifold.map.whisper_qwen25_ln__to_gemma2__wte_ln.pca_k256.median_pct}{24.1}
\defmmnum{manifold.map.whisper_qwen25_ln__to_gemma2__wte_ln.pca_k64.median_pct}{14.3}
\defmmnum{manifold.map.whisper_qwen25_ln__to_gemma2__wte_ln.rr.median}{1.029}
\defmmnum{manifold.map.whisper_qwen25_ln__to_gemma2__wte_ln.rr.p10}{1.024}
\defmmnum{manifold.map.whisper_qwen25_ln__to_gemma2__wte_ln.rr.p90}{1.036}
\defmmnum{manifold.meta.n_token_eval}{10{,}000}
\defmmnum{manifold.meta.n_utterances}{200}
\defmmnum{manifold.meta.peak_vram_mb}{1180.1}
\defmmnum{manifold.src.gemma_max_rr_med}{1.12}
\defmmnum{manifold.src.gemma_min_rr_med}{1.02}
\defmmnum{manifold.src.max_nn_cos_med}{0.1376}
\defmmnum{manifold.src.min_nn_cos_med}{0.1089}
\defmmnum{manifold.src.moonshine_gemma2_ln.cos_mean.median}{+0.0816}
\defmmnum{manifold.src.moonshine_gemma2_ln.n_frames}{63{,}240}
\defmmnum{manifold.src.moonshine_gemma2_ln.nn_cos.median}{0.1376}
\defmmnum{manifold.src.moonshine_gemma2_ln.nn_cos.p10}{0.1230}
\defmmnum{manifold.src.moonshine_gemma2_ln.nn_cos.p90}{0.1567}
\defmmnum{manifold.src.moonshine_gemma2_ln.pca_k1024.median_pct}{51.8}
\defmmnum{manifold.src.moonshine_gemma2_ln.pca_k256.median_pct}{23.8}
\defmmnum{manifold.src.moonshine_gemma2_ln.pca_k64.median_pct}{15.3}
\defmmnum{manifold.src.moonshine_gemma2_ln.rr.median}{1.093}
\defmmnum{manifold.src.moonshine_gemma2_ln.rr.p10}{1.084}
\defmmnum{manifold.src.moonshine_gemma2_ln.rr.p90}{1.105}
\defmmnum{manifold.src.moonshine_gemma3_ln.cos_mean.median}{+0.0458}
\defmmnum{manifold.src.moonshine_gemma3_ln.n_frames}{63{,}240}
\defmmnum{manifold.src.moonshine_gemma3_ln.nn_cos.median}{0.1089}
\defmmnum{manifold.src.moonshine_gemma3_ln.nn_cos.p10}{0.0984}
\defmmnum{manifold.src.moonshine_gemma3_ln.nn_cos.p90}{0.1241}
\defmmnum{manifold.src.moonshine_gemma3_ln.pca_k1024.median_pct}{41.5}
\defmmnum{manifold.src.moonshine_gemma3_ln.pca_k256.median_pct}{14.9}
\defmmnum{manifold.src.moonshine_gemma3_ln.pca_k64.median_pct}{5.6}
\defmmnum{manifold.src.moonshine_gemma3_ln.rr.median}{1.118}
\defmmnum{manifold.src.moonshine_gemma3_ln.rr.p10}{1.109}
\defmmnum{manifold.src.moonshine_gemma3_ln.rr.p90}{1.127}
\defmmnum{manifold.src.moonshine_qwen25_ln.cos_mean.median}{+0.0017}
\defmmnum{manifold.src.moonshine_qwen25_ln.n_frames}{63{,}240}
\defmmnum{manifold.src.moonshine_qwen25_ln.nn_cos.median}{0.1153}
\defmmnum{manifold.src.moonshine_qwen25_ln.nn_cos.p10}{0.1037}
\defmmnum{manifold.src.moonshine_qwen25_ln.nn_cos.p90}{0.1293}
\defmmnum{manifold.src.moonshine_qwen25_ln.pca_k1024.median_pct}{73.7}
\defmmnum{manifold.src.moonshine_qwen25_ln.pca_k256.median_pct}{20.2}
\defmmnum{manifold.src.moonshine_qwen25_ln.pca_k64.median_pct}{6.6}
\defmmnum{manifold.src.moonshine_qwen25_ln.rr.median}{12.74}
\defmmnum{manifold.src.moonshine_qwen25_ln.rr.p10}{12.20}
\defmmnum{manifold.src.moonshine_qwen25_ln.rr.p90}{13.17}
\defmmnum{manifold.src.qwen_max_rr_med}{12.87}
\defmmnum{manifold.src.qwen_min_rr_med}{12.74}
\defmmnum{manifold.src.whisper_gemma2_ln.cos_mean.median}{+0.0605}
\defmmnum{manifold.src.whisper_gemma2_ln.n_frames}{76{,}396}
\defmmnum{manifold.src.whisper_gemma2_ln.nn_cos.median}{0.1240}
\defmmnum{manifold.src.whisper_gemma2_ln.nn_cos.p10}{0.1085}
\defmmnum{manifold.src.whisper_gemma2_ln.nn_cos.p90}{0.1422}
\defmmnum{manifold.src.whisper_gemma2_ln.pca_k1024.median_pct}{52.5}
\defmmnum{manifold.src.whisper_gemma2_ln.pca_k256.median_pct}{24.3}
\defmmnum{manifold.src.whisper_gemma2_ln.pca_k64.median_pct}{15.5}
\defmmnum{manifold.src.whisper_gemma2_ln.rr.median}{1.069}
\defmmnum{manifold.src.whisper_gemma2_ln.rr.p10}{1.060}
\defmmnum{manifold.src.whisper_gemma2_ln.rr.p90}{1.076}
\defmmnum{manifold.src.whisper_gemma2_scaled.cos_mean.median}{+0.0594}
\defmmnum{manifold.src.whisper_gemma2_scaled.n_frames}{76{,}396}
\defmmnum{manifold.src.whisper_gemma2_scaled.nn_cos.median}{0.1214}
\defmmnum{manifold.src.whisper_gemma2_scaled.nn_cos.p10}{0.1055}
\defmmnum{manifold.src.whisper_gemma2_scaled.nn_cos.p90}{0.1395}
\defmmnum{manifold.src.whisper_gemma2_scaled.pca_k1024.median_pct}{52.5}
\defmmnum{manifold.src.whisper_gemma2_scaled.pca_k256.median_pct}{24.7}
\defmmnum{manifold.src.whisper_gemma2_scaled.pca_k64.median_pct}{16.5}
\defmmnum{manifold.src.whisper_gemma2_scaled.rr.median}{1.018}
\defmmnum{manifold.src.whisper_gemma2_scaled.rr.p10}{1.018}
\defmmnum{manifold.src.whisper_gemma2_scaled.rr.p90}{1.018}
\defmmnum{manifold.src.whisper_qwen25_ln.cos_mean.median}{-0.0017}
\defmmnum{manifold.src.whisper_qwen25_ln.n_frames}{76{,}396}
\defmmnum{manifold.src.whisper_qwen25_ln.nn_cos.median}{0.1118}
\defmmnum{manifold.src.whisper_qwen25_ln.nn_cos.p10}{0.0997}
\defmmnum{manifold.src.whisper_qwen25_ln.nn_cos.p90}{0.1286}
\defmmnum{manifold.src.whisper_qwen25_ln.pca_k1024.median_pct}{73.6}
\defmmnum{manifold.src.whisper_qwen25_ln.pca_k256.median_pct}{19.0}
\defmmnum{manifold.src.whisper_qwen25_ln.pca_k64.median_pct}{5.5}
\defmmnum{manifold.src.whisper_qwen25_ln.rr.median}{12.87}
\defmmnum{manifold.src.whisper_qwen25_ln.rr.p10}{12.34}
\defmmnum{manifold.src.whisper_qwen25_ln.rr.p90}{13.41}
\defmmnum{manifold.tok.gemma2.cos_mean.median}{+0.4197}
\defmmnum{manifold.tok.gemma2.nn_cos.median}{0.6291}
\defmmnum{manifold.tok.gemma2.nn_cos.p10}{0.4202}
\defmmnum{manifold.tok.gemma2.nn_cos.p90}{0.8333}
\defmmnum{manifold.tok.gemma2.pca_k1024.median_pct}{64.5}
\defmmnum{manifold.tok.gemma2.pca_k256.median_pct}{28.8}
\defmmnum{manifold.tok.gemma2.pca_k64.median_pct}{14.5}
\defmmnum{manifold.tok.gemma2.rr.median}{1.043}
\defmmnum{manifold.tok.gemma2.rr.p10}{0.885}
\defmmnum{manifold.tok.gemma2.rr.p90}{1.226}
\defmmnum{manifold.tok.gemma3.cos_mean.median}{+0.1781}
\defmmnum{manifold.tok.gemma3.nn_cos.median}{0.5467}
\defmmnum{manifold.tok.gemma3.nn_cos.p10}{0.3043}
\defmmnum{manifold.tok.gemma3.nn_cos.p90}{0.8129}
\defmmnum{manifold.tok.gemma3.pca_k1024.median_pct}{53.4}
\defmmnum{manifold.tok.gemma3.pca_k256.median_pct}{20.9}
\defmmnum{manifold.tok.gemma3.pca_k64.median_pct}{9.1}
\defmmnum{manifold.tok.gemma3.rr.median}{0.994}
\defmmnum{manifold.tok.gemma3.rr.p10}{0.976}
\defmmnum{manifold.tok.gemma3.rr.p90}{1.014}
\defmmnum{manifold.tok.max_nn_cos_med}{0.6291}
\defmmnum{manifold.tok.min_nn_cos_med}{0.5467}
\defmmnum{manifold.tok.qwen25.cos_mean.median}{+0.3316}
\defmmnum{manifold.tok.qwen25.nn_cos.median}{0.6188}
\defmmnum{manifold.tok.qwen25.nn_cos.p10}{0.3847}
\defmmnum{manifold.tok.qwen25.nn_cos.p90}{0.8137}
\defmmnum{manifold.tok.qwen25.pca_k1024.median_pct}{77.9}
\defmmnum{manifold.tok.qwen25.pca_k256.median_pct}{28.1}
\defmmnum{manifold.tok.qwen25.pca_k64.median_pct}{9.9}
\defmmnum{manifold.tok.qwen25.rr.median}{0.998}
\defmmnum{manifold.tok.qwen25.rr.p10}{0.853}
\defmmnum{manifold.tok.qwen25.rr.p90}{1.125}
\defmmnum{model.gemma2.dim}{2304}
\defmmnum{model.gemma2.embed_scale_bf16}{48.00}
\defmmnum{model.gemma2.vocab}{256{,}000}
\defmmnum{model.gemma3.dim}{2560}
\defmmnum{model.gemma3.embed_scale_bf16}{50.50}
\defmmnum{model.gemma3.vocab}{262{,}208}
\defmmnum{model.moonshine_base.dim}{416}
\defmmnum{model.moonshine_base.hz}{41.67}
\defmmnum{model.phi2.dim}{2560}
\defmmnum{model.phi2.vocab}{51{,}200}
\defmmnum{model.qwen25.dim}{1536}
\defmmnum{model.qwen25.vocab}{151{,}936}
\defmmnum{model.qwen3.dim}{2560}
\defmmnum{model.qwen3.vocab}{151{,}936}
\defmmnum{model.smollm2.dim}{2048}
\defmmnum{model.smollm2.vocab}{49{,}152}
\defmmnum{model.token_radius.max_rows}{50{,}000}
\defmmnum{model.unit_embed_scale}{1.00}
\defmmnum{model.whisper_small.dim}{768}
\defmmnum{model.whisper_small.hz}{50}
\defmmnum{opt.adamw.lr}{5\times 10^{-4}}
\defmmnum{opt.adamw.wd}{0.01}
\defmmnum{opt.muon.lr}{0.02}
\defmmnum{opt.muon.momentum}{0.95}
\defmmnum{opt.muon.ns_steps}{5}
\defmmnum{opt.warmup_steps}{100}
\defmmnum{parity.code_regression_items}{0}
\defmmnum{parity.hs_diff_items}{0}
\defmmnum{parity.hs_diff_pp}{0.00}
\defmmnum{parity.piqa_diff_items}{3}
\defmmnum{parity.piqa_diff_pp}{0.16}
\defmmnum{parity.torch260.hs}{57.50}
\defmmnum{parity.torch260.hs_correct}{1{,}150}
\defmmnum{parity.torch260.piqa}{78.73}
\defmmnum{parity.torch260.piqa_correct}{1{,}447}
\defmmnum{parity.torch260.torch}{2.6.0+cu124}
\defmmnum{parity.torch271.hs}{57.50}
\defmmnum{parity.torch271.hs_correct}{1{,}150}
\defmmnum{parity.torch271.piqa}{78.56}
\defmmnum{parity.torch271.piqa_correct}{1{,}444}
\defmmnum{parity.torch271.torch}{2.7.1+cu118}
\defmmnum{parity.transformers_fixed}{5.17.0}
\defmmnum{protocol.full_signal_pct}{100}
\defmmnum{radius.gemma2_2b}{80.7}
\defmmnum{radius.gemma3_4b}{50.8}
\defmmnum{radius.phi2}{1.21}
\defmmnum{radius.qwen25_15b}{1.03}
\defmmnum{radius.qwen3_4b}{1.13}
\defmmnum{radius.smollm2_17b}{3.07}
\defmmnum{radius.spread_factor}{78}
\defmmnum{relwork.slamasr.frame_stack_k}{5}
\defmmnum{relwork.slamasr.hidden_dim}{2048}
\defmmnum{stackcheck.a7.abs_delta_clean}{0.01}
\defmmnum{stackcheck.a7.abs_delta_other}{0.01}
\defmmnum{stackcheck.a7.clean}{3.21}
\defmmnum{stackcheck.a7.delta_clean}{-0.01}
\defmmnum{stackcheck.a7.delta_clean_exact}{-0.0095}
\defmmnum{stackcheck.a7.delta_other}{-0.01}
\defmmnum{stackcheck.a7.delta_other_exact}{-0.0134}
\defmmnum{stackcheck.a7.diff_clean_pct}{1.98}
\defmmnum{stackcheck.a7.diff_clean_utts}{52}
\defmmnum{stackcheck.a7.diff_other_pct}{3.54}
\defmmnum{stackcheck.a7.diff_other_utts}{104}
\defmmnum{stackcheck.a7.diff_raw_clean_utts}{54}
\defmmnum{stackcheck.a7.diff_raw_other_utts}{104}
\defmmnum{stackcheck.a7.diff_raw_total_pct}{2.84}
\defmmnum{stackcheck.a7.diff_raw_total_utts}{158}
\defmmnum{stackcheck.a7.diff_total_pct}{2.81}
\defmmnum{stackcheck.a7.diff_total_utts}{156}
\defmmnum{stackcheck.a7.gpu04_clean}{3.22}
\defmmnum{stackcheck.a7.gpu04_other}{7.53}
\defmmnum{stackcheck.a7.other}{7.52}
\defmmnum{stackcheck.a7.torch}{2.6.0+cu124}
\defmmnum{stackcheck.a7.total_utts}{5{,}559}
\defmmnum{stackcheck.a7.transformers}{5.15.0}
\defmmnum{stackcheck.done}{1}
\defmmnum{token_hits.all5.max_ov_pct}{95.76}
\defmmnum{token_hits.all5.max_te_pct}{25.69}
\defmmnum{token_hits.all5.max_tr_pct}{72.18}
\defmmnum{token_hits.all5.min_ov_pct}{79.16}
\defmmnum{token_hits.all5.min_te_pct}{11.74}
\defmmnum{token_hits.all5.min_tr_pct}{58.69}
\defmmnum{token_hits.exact4.max_ov_pct}{87.51}
\defmmnum{token_hits.exact4.max_te_pct}{24.14}
\defmmnum{token_hits.exact4.max_tr_pct}{72.18}
\defmmnum{token_hits.exact4.min_ov_pct}{79.16}
\defmmnum{token_hits.exact4.min_te_pct}{11.74}
\defmmnum{token_hits.exact4.min_tr_pct}{58.69}
\defmmnum{token_hits.gemma2_2b.hellaswag.combined.occurrences.nat_pct}{12.49}
\defmmnum{token_hits.gemma2_2b.hellaswag.combined.occurrences.ov_pct}{87.51}
\defmmnum{token_hits.gemma2_2b.hellaswag.combined.occurrences.te_pct}{22.56}
\defmmnum{token_hits.gemma2_2b.hellaswag.combined.occurrences.tr_pct}{64.95}
\defmmnum{token_hits.gemma2_2b.hellaswag.combined.types.nat_pct}{2.72}
\defmmnum{token_hits.gemma2_2b.hellaswag.combined.types.ov_pct}{97.28}
\defmmnum{token_hits.gemma2_2b.hellaswag.combined.types.te_pct}{19.49}
\defmmnum{token_hits.gemma2_2b.hellaswag.combined.types.tr_pct}{77.79}
\defmmnum{token_hits.gemma2_2b.hellaswag.context.occurrences.nat_pct}{13.86}
\defmmnum{token_hits.gemma2_2b.hellaswag.context.occurrences.ov_pct}{86.14}
\defmmnum{token_hits.gemma2_2b.hellaswag.context.occurrences.te_pct}{21.36}
\defmmnum{token_hits.gemma2_2b.hellaswag.context.occurrences.tr_pct}{64.78}
\defmmnum{token_hits.gemma2_2b.hellaswag.context.types.nat_pct}{2.53}
\defmmnum{token_hits.gemma2_2b.hellaswag.context.types.ov_pct}{97.47}
\defmmnum{token_hits.gemma2_2b.hellaswag.context.types.te_pct}{18.98}
\defmmnum{token_hits.gemma2_2b.hellaswag.context.types.tr_pct}{78.48}
\defmmnum{token_hits.gemma2_2b.hellaswag.continuation.occurrences.nat_pct}{9.90}
\defmmnum{token_hits.gemma2_2b.hellaswag.continuation.occurrences.ov_pct}{90.10}
\defmmnum{token_hits.gemma2_2b.hellaswag.continuation.occurrences.te_pct}{24.82}
\defmmnum{token_hits.gemma2_2b.hellaswag.continuation.occurrences.tr_pct}{65.29}
\defmmnum{token_hits.gemma2_2b.hellaswag.continuation.types.nat_pct}{2.67}
\defmmnum{token_hits.gemma2_2b.hellaswag.continuation.types.ov_pct}{97.33}
\defmmnum{token_hits.gemma2_2b.hellaswag.continuation.types.te_pct}{19.46}
\defmmnum{token_hits.gemma2_2b.hellaswag.continuation.types.tr_pct}{77.87}
\defmmnum{token_hits.gemma2_2b.n_anchors}{163{,}983}
\defmmnum{token_hits.gemma2_2b.n_rows_overwritten}{163{,}977}
\defmmnum{token_hits.gemma2_2b.piqa.combined.occurrences.nat_pct}{17.16}
\defmmnum{token_hits.gemma2_2b.piqa.combined.occurrences.ov_pct}{82.84}
\defmmnum{token_hits.gemma2_2b.piqa.combined.occurrences.te_pct}{24.14}
\defmmnum{token_hits.gemma2_2b.piqa.combined.occurrences.tr_pct}{58.69}
\defmmnum{token_hits.gemma2_2b.piqa.combined.types.nat_pct}{3.04}
\defmmnum{token_hits.gemma2_2b.piqa.combined.types.ov_pct}{96.96}
\defmmnum{token_hits.gemma2_2b.piqa.combined.types.te_pct}{19.86}
\defmmnum{token_hits.gemma2_2b.piqa.combined.types.tr_pct}{77.10}
\defmmnum{token_hits.gemma2_2b.piqa.context.occurrences.nat_pct}{20.53}
\defmmnum{token_hits.gemma2_2b.piqa.context.occurrences.ov_pct}{79.47}
\defmmnum{token_hits.gemma2_2b.piqa.context.occurrences.te_pct}{25.91}
\defmmnum{token_hits.gemma2_2b.piqa.context.occurrences.tr_pct}{53.56}
\defmmnum{token_hits.gemma2_2b.piqa.context.types.nat_pct}{2.87}
\defmmnum{token_hits.gemma2_2b.piqa.context.types.ov_pct}{97.13}
\defmmnum{token_hits.gemma2_2b.piqa.context.types.te_pct}{19.92}
\defmmnum{token_hits.gemma2_2b.piqa.context.types.tr_pct}{77.21}
\defmmnum{token_hits.gemma2_2b.piqa.continuation.occurrences.nat_pct}{15.77}
\defmmnum{token_hits.gemma2_2b.piqa.continuation.occurrences.ov_pct}{84.23}
\defmmnum{token_hits.gemma2_2b.piqa.continuation.occurrences.te_pct}{23.41}
\defmmnum{token_hits.gemma2_2b.piqa.continuation.occurrences.tr_pct}{60.82}
\defmmnum{token_hits.gemma2_2b.piqa.continuation.types.nat_pct}{2.84}
\defmmnum{token_hits.gemma2_2b.piqa.continuation.types.ov_pct}{97.16}
\defmmnum{token_hits.gemma2_2b.piqa.continuation.types.te_pct}{19.72}
\defmmnum{token_hits.gemma2_2b.piqa.continuation.types.tr_pct}{77.43}
\defmmnum{token_hits.phi_2.hellaswag.combined.occurrences.nat_pct}{14.36}
\defmmnum{token_hits.phi_2.hellaswag.combined.occurrences.ov_pct}{85.64}
\defmmnum{token_hits.phi_2.hellaswag.combined.occurrences.te_pct}{13.46}
\defmmnum{token_hits.phi_2.hellaswag.combined.occurrences.tr_pct}{72.18}
\defmmnum{token_hits.phi_2.hellaswag.combined.types.nat_pct}{15.61}
\defmmnum{token_hits.phi_2.hellaswag.combined.types.ov_pct}{84.39}
\defmmnum{token_hits.phi_2.hellaswag.combined.types.te_pct}{16.96}
\defmmnum{token_hits.phi_2.hellaswag.combined.types.tr_pct}{67.43}
\defmmnum{token_hits.phi_2.hellaswag.context.occurrences.nat_pct}{15.72}
\defmmnum{token_hits.phi_2.hellaswag.context.occurrences.ov_pct}{84.28}
\defmmnum{token_hits.phi_2.hellaswag.context.occurrences.te_pct}{14.04}
\defmmnum{token_hits.phi_2.hellaswag.context.occurrences.tr_pct}{70.23}
\defmmnum{token_hits.phi_2.hellaswag.context.types.nat_pct}{12.64}
\defmmnum{token_hits.phi_2.hellaswag.context.types.ov_pct}{87.36}
\defmmnum{token_hits.phi_2.hellaswag.context.types.te_pct}{17.54}
\defmmnum{token_hits.phi_2.hellaswag.context.types.tr_pct}{69.82}
\defmmnum{token_hits.phi_2.hellaswag.continuation.occurrences.nat_pct}{11.80}
\defmmnum{token_hits.phi_2.hellaswag.continuation.occurrences.ov_pct}{88.20}
\defmmnum{token_hits.phi_2.hellaswag.continuation.occurrences.te_pct}{12.36}
\defmmnum{token_hits.phi_2.hellaswag.continuation.occurrences.tr_pct}{75.85}
\defmmnum{token_hits.phi_2.hellaswag.continuation.types.nat_pct}{14.87}
\defmmnum{token_hits.phi_2.hellaswag.continuation.types.ov_pct}{85.13}
\defmmnum{token_hits.phi_2.hellaswag.continuation.types.te_pct}{17.29}
\defmmnum{token_hits.phi_2.hellaswag.continuation.types.tr_pct}{67.83}
\defmmnum{token_hits.phi_2.n_anchors}{44{,}975}
\defmmnum{token_hits.phi_2.n_rows_overwritten}{44{,}970}
\defmmnum{token_hits.phi_2.piqa.combined.occurrences.nat_pct}{20.84}
\defmmnum{token_hits.phi_2.piqa.combined.occurrences.ov_pct}{79.16}
\defmmnum{token_hits.phi_2.piqa.combined.occurrences.te_pct}{11.74}
\defmmnum{token_hits.phi_2.piqa.combined.occurrences.tr_pct}{67.42}
\defmmnum{token_hits.phi_2.piqa.combined.types.nat_pct}{19.88}
\defmmnum{token_hits.phi_2.piqa.combined.types.ov_pct}{80.12}
\defmmnum{token_hits.phi_2.piqa.combined.types.te_pct}{16.36}
\defmmnum{token_hits.phi_2.piqa.combined.types.tr_pct}{63.75}
\defmmnum{token_hits.phi_2.piqa.context.occurrences.nat_pct}{24.54}
\defmmnum{token_hits.phi_2.piqa.context.occurrences.ov_pct}{75.46}
\defmmnum{token_hits.phi_2.piqa.context.occurrences.te_pct}{11.78}
\defmmnum{token_hits.phi_2.piqa.context.occurrences.tr_pct}{63.68}
\defmmnum{token_hits.phi_2.piqa.context.types.nat_pct}{17.69}
\defmmnum{token_hits.phi_2.piqa.context.types.ov_pct}{82.31}
\defmmnum{token_hits.phi_2.piqa.context.types.te_pct}{16.59}
\defmmnum{token_hits.phi_2.piqa.context.types.tr_pct}{65.71}
\defmmnum{token_hits.phi_2.piqa.continuation.occurrences.nat_pct}{19.31}
\defmmnum{token_hits.phi_2.piqa.continuation.occurrences.ov_pct}{80.69}
\defmmnum{token_hits.phi_2.piqa.continuation.occurrences.te_pct}{11.72}
\defmmnum{token_hits.phi_2.piqa.continuation.occurrences.tr_pct}{68.97}
\defmmnum{token_hits.phi_2.piqa.continuation.types.nat_pct}{18.55}
\defmmnum{token_hits.phi_2.piqa.continuation.types.ov_pct}{81.45}
\defmmnum{token_hits.phi_2.piqa.continuation.types.te_pct}{16.75}
\defmmnum{token_hits.phi_2.piqa.continuation.types.tr_pct}{64.71}
\defmmnum{token_hits.qwen3_4b.hellaswag.combined.occurrences.nat_pct}{13.89}
\defmmnum{token_hits.qwen3_4b.hellaswag.combined.occurrences.ov_pct}{86.11}
\defmmnum{token_hits.qwen3_4b.hellaswag.combined.occurrences.te_pct}{14.96}
\defmmnum{token_hits.qwen3_4b.hellaswag.combined.occurrences.tr_pct}{71.16}
\defmmnum{token_hits.qwen3_4b.hellaswag.combined.types.nat_pct}{12.71}
\defmmnum{token_hits.qwen3_4b.hellaswag.combined.types.ov_pct}{87.29}
\defmmnum{token_hits.qwen3_4b.hellaswag.combined.types.te_pct}{17.43}
\defmmnum{token_hits.qwen3_4b.hellaswag.combined.types.tr_pct}{69.86}
\defmmnum{token_hits.qwen3_4b.hellaswag.context.occurrences.nat_pct}{15.24}
\defmmnum{token_hits.qwen3_4b.hellaswag.context.occurrences.ov_pct}{84.76}
\defmmnum{token_hits.qwen3_4b.hellaswag.context.occurrences.te_pct}{14.69}
\defmmnum{token_hits.qwen3_4b.hellaswag.context.occurrences.tr_pct}{70.07}
\defmmnum{token_hits.qwen3_4b.hellaswag.context.types.nat_pct}{10.20}
\defmmnum{token_hits.qwen3_4b.hellaswag.context.types.ov_pct}{89.80}
\defmmnum{token_hits.qwen3_4b.hellaswag.context.types.te_pct}{17.83}
\defmmnum{token_hits.qwen3_4b.hellaswag.context.types.tr_pct}{71.98}
\defmmnum{token_hits.qwen3_4b.hellaswag.continuation.occurrences.nat_pct}{11.34}
\defmmnum{token_hits.qwen3_4b.hellaswag.continuation.occurrences.ov_pct}{88.66}
\defmmnum{token_hits.qwen3_4b.hellaswag.continuation.occurrences.te_pct}{15.45}
\defmmnum{token_hits.qwen3_4b.hellaswag.continuation.occurrences.tr_pct}{73.22}
\defmmnum{token_hits.qwen3_4b.hellaswag.continuation.types.nat_pct}{12.22}
\defmmnum{token_hits.qwen3_4b.hellaswag.continuation.types.ov_pct}{87.78}
\defmmnum{token_hits.qwen3_4b.hellaswag.continuation.types.te_pct}{17.49}
\defmmnum{token_hits.qwen3_4b.hellaswag.continuation.types.tr_pct}{70.29}
\defmmnum{token_hits.qwen3_4b.n_anchors}{82{,}663}
\defmmnum{token_hits.qwen3_4b.n_rows_overwritten}{82{,}658}
\defmmnum{token_hits.qwen3_4b.piqa.combined.occurrences.nat_pct}{19.91}
\defmmnum{token_hits.qwen3_4b.piqa.combined.occurrences.ov_pct}{80.09}
\defmmnum{token_hits.qwen3_4b.piqa.combined.occurrences.te_pct}{17.68}
\defmmnum{token_hits.qwen3_4b.piqa.combined.occurrences.tr_pct}{62.41}
\defmmnum{token_hits.qwen3_4b.piqa.combined.types.nat_pct}{15.75}
\defmmnum{token_hits.qwen3_4b.piqa.combined.types.ov_pct}{84.25}
\defmmnum{token_hits.qwen3_4b.piqa.combined.types.te_pct}{16.95}
\defmmnum{token_hits.qwen3_4b.piqa.combined.types.tr_pct}{67.30}
\defmmnum{token_hits.qwen3_4b.piqa.context.occurrences.nat_pct}{23.42}
\defmmnum{token_hits.qwen3_4b.piqa.context.occurrences.ov_pct}{76.58}
\defmmnum{token_hits.qwen3_4b.piqa.context.occurrences.te_pct}{19.61}
\defmmnum{token_hits.qwen3_4b.piqa.context.occurrences.tr_pct}{56.97}
\defmmnum{token_hits.qwen3_4b.piqa.context.types.nat_pct}{13.73}
\defmmnum{token_hits.qwen3_4b.piqa.context.types.ov_pct}{86.27}
\defmmnum{token_hits.qwen3_4b.piqa.context.types.te_pct}{17.20}
\defmmnum{token_hits.qwen3_4b.piqa.context.types.tr_pct}{69.07}
\defmmnum{token_hits.qwen3_4b.piqa.continuation.occurrences.nat_pct}{18.46}
\defmmnum{token_hits.qwen3_4b.piqa.continuation.occurrences.ov_pct}{81.54}
\defmmnum{token_hits.qwen3_4b.piqa.continuation.occurrences.te_pct}{16.88}
\defmmnum{token_hits.qwen3_4b.piqa.continuation.occurrences.tr_pct}{64.66}
\defmmnum{token_hits.qwen3_4b.piqa.continuation.types.nat_pct}{14.82}
\defmmnum{token_hits.qwen3_4b.piqa.continuation.types.ov_pct}{85.18}
\defmmnum{token_hits.qwen3_4b.piqa.continuation.types.te_pct}{17.24}
\defmmnum{token_hits.qwen3_4b.piqa.continuation.types.tr_pct}{67.94}
\defmmnum{token_hits.qwen3_4b_surface.hellaswag.combined.occurrences.nat_pct}{4.24}
\defmmnum{token_hits.qwen3_4b_surface.hellaswag.combined.occurrences.ov_pct}{95.76}
\defmmnum{token_hits.qwen3_4b_surface.hellaswag.combined.occurrences.te_pct}{24.17}
\defmmnum{token_hits.qwen3_4b_surface.hellaswag.combined.occurrences.tr_pct}{71.59}
\defmmnum{token_hits.qwen3_4b_surface.hellaswag.combined.types.nat_pct}{11.67}
\defmmnum{token_hits.qwen3_4b_surface.hellaswag.combined.types.ov_pct}{88.33}
\defmmnum{token_hits.qwen3_4b_surface.hellaswag.combined.types.te_pct}{18.32}
\defmmnum{token_hits.qwen3_4b_surface.hellaswag.combined.types.tr_pct}{70.01}
\defmmnum{token_hits.qwen3_4b_surface.hellaswag.context.occurrences.nat_pct}{5.56}
\defmmnum{token_hits.qwen3_4b_surface.hellaswag.context.occurrences.ov_pct}{94.44}
\defmmnum{token_hits.qwen3_4b_surface.hellaswag.context.occurrences.te_pct}{23.47}
\defmmnum{token_hits.qwen3_4b_surface.hellaswag.context.occurrences.tr_pct}{70.97}
\defmmnum{token_hits.qwen3_4b_surface.hellaswag.context.types.nat_pct}{9.13}
\defmmnum{token_hits.qwen3_4b_surface.hellaswag.context.types.ov_pct}{90.87}
\defmmnum{token_hits.qwen3_4b_surface.hellaswag.context.types.te_pct}{18.81}
\defmmnum{token_hits.qwen3_4b_surface.hellaswag.context.types.tr_pct}{72.06}
\defmmnum{token_hits.qwen3_4b_surface.hellaswag.continuation.occurrences.nat_pct}{1.75}
\defmmnum{token_hits.qwen3_4b_surface.hellaswag.continuation.occurrences.ov_pct}{98.25}
\defmmnum{token_hits.qwen3_4b_surface.hellaswag.continuation.occurrences.te_pct}{25.49}
\defmmnum{token_hits.qwen3_4b_surface.hellaswag.continuation.occurrences.tr_pct}{72.76}
\defmmnum{token_hits.qwen3_4b_surface.hellaswag.continuation.types.nat_pct}{11.19}
\defmmnum{token_hits.qwen3_4b_surface.hellaswag.continuation.types.ov_pct}{88.81}
\defmmnum{token_hits.qwen3_4b_surface.hellaswag.continuation.types.te_pct}{18.21}
\defmmnum{token_hits.qwen3_4b_surface.hellaswag.continuation.types.tr_pct}{70.60}
\defmmnum{token_hits.qwen3_4b_surface.n_anchors}{65{,}406}
\defmmnum{token_hits.qwen3_4b_surface.n_rows_overwritten}{65{,}406}
\defmmnum{token_hits.qwen3_4b_surface.piqa.combined.occurrences.nat_pct}{6.22}
\defmmnum{token_hits.qwen3_4b_surface.piqa.combined.occurrences.ov_pct}{93.78}
\defmmnum{token_hits.qwen3_4b_surface.piqa.combined.occurrences.te_pct}{25.69}
\defmmnum{token_hits.qwen3_4b_surface.piqa.combined.occurrences.tr_pct}{68.09}
\defmmnum{token_hits.qwen3_4b_surface.piqa.combined.types.nat_pct}{14.47}
\defmmnum{token_hits.qwen3_4b_surface.piqa.combined.types.ov_pct}{85.53}
\defmmnum{token_hits.qwen3_4b_surface.piqa.combined.types.te_pct}{18.04}
\defmmnum{token_hits.qwen3_4b_surface.piqa.combined.types.tr_pct}{67.49}
\defmmnum{token_hits.qwen3_4b_surface.piqa.context.occurrences.nat_pct}{14.11}
\defmmnum{token_hits.qwen3_4b_surface.piqa.context.occurrences.ov_pct}{85.89}
\defmmnum{token_hits.qwen3_4b_surface.piqa.context.occurrences.te_pct}{18.71}
\defmmnum{token_hits.qwen3_4b_surface.piqa.context.occurrences.tr_pct}{67.18}
\defmmnum{token_hits.qwen3_4b_surface.piqa.context.types.nat_pct}{12.25}
\defmmnum{token_hits.qwen3_4b_surface.piqa.context.types.ov_pct}{87.75}
\defmmnum{token_hits.qwen3_4b_surface.piqa.context.types.te_pct}{18.85}
\defmmnum{token_hits.qwen3_4b_surface.piqa.context.types.tr_pct}{68.90}
\defmmnum{token_hits.qwen3_4b_surface.piqa.continuation.occurrences.nat_pct}{2.96}
\defmmnum{token_hits.qwen3_4b_surface.piqa.continuation.occurrences.ov_pct}{97.04}
\defmmnum{token_hits.qwen3_4b_surface.piqa.continuation.occurrences.te_pct}{28.57}
\defmmnum{token_hits.qwen3_4b_surface.piqa.continuation.occurrences.tr_pct}{68.47}
\defmmnum{token_hits.qwen3_4b_surface.piqa.continuation.types.nat_pct}{13.41}
\defmmnum{token_hits.qwen3_4b_surface.piqa.continuation.types.ov_pct}{86.58}
\defmmnum{token_hits.qwen3_4b_surface.piqa.continuation.types.te_pct}{18.12}
\defmmnum{token_hits.qwen3_4b_surface.piqa.continuation.types.tr_pct}{68.46}
\defmmnum{token_hits.smollm2_1_7b.hellaswag.combined.occurrences.nat_pct}{14.49}
\defmmnum{token_hits.smollm2_1_7b.hellaswag.combined.occurrences.ov_pct}{85.51}
\defmmnum{token_hits.smollm2_1_7b.hellaswag.combined.occurrences.te_pct}{14.64}
\defmmnum{token_hits.smollm2_1_7b.hellaswag.combined.occurrences.tr_pct}{70.87}
\defmmnum{token_hits.smollm2_1_7b.hellaswag.combined.types.nat_pct}{16.42}
\defmmnum{token_hits.smollm2_1_7b.hellaswag.combined.types.ov_pct}{83.58}
\defmmnum{token_hits.smollm2_1_7b.hellaswag.combined.types.te_pct}{17.87}
\defmmnum{token_hits.smollm2_1_7b.hellaswag.combined.types.tr_pct}{65.71}
\defmmnum{token_hits.smollm2_1_7b.hellaswag.context.occurrences.nat_pct}{15.82}
\defmmnum{token_hits.smollm2_1_7b.hellaswag.context.occurrences.ov_pct}{84.18}
\defmmnum{token_hits.smollm2_1_7b.hellaswag.context.occurrences.te_pct}{14.70}
\defmmnum{token_hits.smollm2_1_7b.hellaswag.context.occurrences.tr_pct}{69.48}
\defmmnum{token_hits.smollm2_1_7b.hellaswag.context.types.nat_pct}{13.02}
\defmmnum{token_hits.smollm2_1_7b.hellaswag.context.types.ov_pct}{86.98}
\defmmnum{token_hits.smollm2_1_7b.hellaswag.context.types.te_pct}{18.66}
\defmmnum{token_hits.smollm2_1_7b.hellaswag.context.types.tr_pct}{68.32}
\defmmnum{token_hits.smollm2_1_7b.hellaswag.continuation.occurrences.nat_pct}{11.99}
\defmmnum{token_hits.smollm2_1_7b.hellaswag.continuation.occurrences.ov_pct}{88.01}
\defmmnum{token_hits.smollm2_1_7b.hellaswag.continuation.occurrences.te_pct}{14.53}
\defmmnum{token_hits.smollm2_1_7b.hellaswag.continuation.occurrences.tr_pct}{73.48}
\defmmnum{token_hits.smollm2_1_7b.hellaswag.continuation.types.nat_pct}{15.76}
\defmmnum{token_hits.smollm2_1_7b.hellaswag.continuation.types.ov_pct}{84.24}
\defmmnum{token_hits.smollm2_1_7b.hellaswag.continuation.types.te_pct}{17.91}
\defmmnum{token_hits.smollm2_1_7b.hellaswag.continuation.types.tr_pct}{66.34}
\defmmnum{token_hits.smollm2_1_7b.n_anchors}{45{,}554}
\defmmnum{token_hits.smollm2_1_7b.n_rows_overwritten}{45{,}549}
\defmmnum{token_hits.smollm2_1_7b.piqa.combined.occurrences.nat_pct}{20.22}
\defmmnum{token_hits.smollm2_1_7b.piqa.combined.occurrences.ov_pct}{79.78}
\defmmnum{token_hits.smollm2_1_7b.piqa.combined.occurrences.te_pct}{15.07}
\defmmnum{token_hits.smollm2_1_7b.piqa.combined.occurrences.tr_pct}{64.71}
\defmmnum{token_hits.smollm2_1_7b.piqa.combined.types.nat_pct}{17.99}
\defmmnum{token_hits.smollm2_1_7b.piqa.combined.types.ov_pct}{82.01}
\defmmnum{token_hits.smollm2_1_7b.piqa.combined.types.te_pct}{17.22}
\defmmnum{token_hits.smollm2_1_7b.piqa.combined.types.tr_pct}{64.79}
\defmmnum{token_hits.smollm2_1_7b.piqa.context.occurrences.nat_pct}{24.09}
\defmmnum{token_hits.smollm2_1_7b.piqa.context.occurrences.ov_pct}{75.91}
\defmmnum{token_hits.smollm2_1_7b.piqa.context.occurrences.te_pct}{14.82}
\defmmnum{token_hits.smollm2_1_7b.piqa.context.occurrences.tr_pct}{61.09}
\defmmnum{token_hits.smollm2_1_7b.piqa.context.types.nat_pct}{16.54}
\defmmnum{token_hits.smollm2_1_7b.piqa.context.types.ov_pct}{83.46}
\defmmnum{token_hits.smollm2_1_7b.piqa.context.types.te_pct}{17.93}
\defmmnum{token_hits.smollm2_1_7b.piqa.context.types.tr_pct}{65.54}
\defmmnum{token_hits.smollm2_1_7b.piqa.continuation.occurrences.nat_pct}{18.62}
\defmmnum{token_hits.smollm2_1_7b.piqa.continuation.occurrences.ov_pct}{81.38}
\defmmnum{token_hits.smollm2_1_7b.piqa.continuation.occurrences.te_pct}{15.18}
\defmmnum{token_hits.smollm2_1_7b.piqa.continuation.occurrences.tr_pct}{66.21}
\defmmnum{token_hits.smollm2_1_7b.piqa.continuation.types.nat_pct}{16.13}
\defmmnum{token_hits.smollm2_1_7b.piqa.continuation.types.ov_pct}{83.87}
\defmmnum{token_hits.smollm2_1_7b.piqa.continuation.types.te_pct}{17.42}
\defmmnum{token_hits.smollm2_1_7b.piqa.continuation.types.tr_pct}{66.45}
\defmmnum{token_hits.tab1_exact3.max_ov_pct}{86.11}
\defmmnum{token_hits.tab1_exact3.max_te_pct}{17.68}
\defmmnum{token_hits.tab1_exact3.max_tr_pct}{72.18}
\defmmnum{token_hits.tab1_exact3.min_ov_pct}{79.16}
\defmmnum{token_hits.tab1_exact3.min_te_pct}{11.74}
\defmmnum{token_hits.tab1_exact3.min_tr_pct}{62.41}
\defmmnum{versions.a7.torch}{2.10.0+cu128}
\defmmnum{versions.a7.transformers}{5.16.0.dev0}
\defmmnum{versions.distinct_stacks}{11}
\defmmnum{versions.distinct_torch}{4}
\defmmnum{versions.distinct_transformers}{6}
\defmmnum{versions.m5.torch}{2.6.0+cu124}
\defmmnum{versions.m5.transformers}{5.15.0}
\defmmnum{versions.mixed_asr_tables}{4}
\defmmnum{versions.native_host_piqa_diff_items}{3}
\defmmnum{versions.tab_ws2_asr.n_cells}{22}
\defmmnum{versions.tab_ws2_asr.n_stacks}{8}
\defmmnum{versions.tab_ws3_comp.n_cells}{110}
\defmmnum{versions.tab_ws3_comp.n_stacks}{8}
\defmmnum{versions.tab_ws4b.n_cells}{27}
\defmmnum{versions.tab_ws4b.n_stacks}{9}
\defmmnum{versions.tab_ws4c.n_cells}{24}
\defmmnum{versions.tab_ws4c.n_stacks}{5}
\defmmnum{versions.torch_list}{2.6.0, 2.7.1, 2.9.1, 2.10.0}
\defmmnum{versions.total_asr_tables}{4}
\defmmnum{versions.total_wer_files}{206}
\defmmnum{versions.transformers_list}{4.53.3, 5.15.0, 5.16.0.dev0, 5.17.0, 5.22.1, 5.22.2}
\defmmnum{versions.ws1_native.hs_correct}{1{,}150}
\defmmnum{versions.ws1_native.piqa_correct}{1{,}447}
\defmmnum{versions.ws1_native.torch}{2.6.0+cu124}
\defmmnum{versions.ws1_native.transformers}{5.17.0}
\defmmnum{versions.ws4d_b0.torch}{2.7.1+cu118}
\defmmnum{versions.ws4d_b0.transformers}{5.17.0}
\defmmnum{ws1.anchor_test_pct}{20}
\defmmnum{ws1.anchor_train_pct}{80}
\defmmnum{ws1.best.heldout_cos}{0.6296}
\defmmnum{ws1.best.hs}{56.70}
\defmmnum{ws1.best.hs_retention}{98.6}
\defmmnum{ws1.best.p1}{99.08}
\defmmnum{ws1.best.piqa}{77.48}
\defmmnum{ws1.best.piqa_retention}{98.4}
\defmmnum{ws1.done_cells}{25}
\defmmnum{ws1.five_norm_done_cells}{43}
\defmmnum{ws1.five_norm_total_cells}{43}
\defmmnum{ws1.gate.anchors}{242{,}564}
\defmmnum{ws1.gate.cos}{1.0000}
\defmmnum{ws1.gate.delta_hs}{0.00}
\defmmnum{ws1.gate.delta_piqa}{0.00}
\defmmnum{ws1.gate.hs}{57.50}
\defmmnum{ws1.gate.p1}{100.00}
\defmmnum{ws1.gate.piqa}{78.73}
\defmmnum{ws1.gate.rr}{1.000}
\defmmnum{ws1.keepnone.phi.lstsq.none.anchors}{44{,}975}
\defmmnum{ws1.keepnone.phi.lstsq.none.cos}{0.4527}
\defmmnum{ws1.keepnone.phi.lstsq.none.hs}{42.95}
\defmmnum{ws1.keepnone.phi.lstsq.none.p1}{97.35}
\defmmnum{ws1.keepnone.phi.lstsq.none.piqa}{70.95}
\defmmnum{ws1.keepnone.phi.lstsq.none.raw_rr}{0.555}
\defmmnum{ws1.keepnone.phi.lstsq.none.ret_hs}{74.7}
\defmmnum{ws1.keepnone.phi.lstsq.none.ret_piqa}{90.1}
\defmmnum{ws1.keepnone.phi.lstsq.none.rr}{0.555}
\defmmnum{ws1.keepnone.phi.lstsq.scaled_l2norm.anchors}{44{,}975}
\defmmnum{ws1.keepnone.phi.lstsq.scaled_l2norm.cos}{0.4527}
\defmmnum{ws1.keepnone.phi.lstsq.scaled_l2norm.hs}{49.30}
\defmmnum{ws1.keepnone.phi.lstsq.scaled_l2norm.p1}{97.35}
\defmmnum{ws1.keepnone.phi.lstsq.scaled_l2norm.piqa}{74.16}
\defmmnum{ws1.keepnone.phi.lstsq.scaled_l2norm.raw_rr}{0.555}
\defmmnum{ws1.keepnone.phi.lstsq.scaled_l2norm.ret_hs}{85.7}
\defmmnum{ws1.keepnone.phi.lstsq.scaled_l2norm.ret_piqa}{94.2}
\defmmnum{ws1.keepnone.phi.lstsq.scaled_l2norm.rr}{1.000}
\defmmnum{ws1.keepnone.phi.procrustes.none.anchors}{44{,}975}
\defmmnum{ws1.keepnone.phi.procrustes.none.cos}{0.3397}
\defmmnum{ws1.keepnone.phi.procrustes.none.hs}{25.85}
\defmmnum{ws1.keepnone.phi.procrustes.none.p1}{97.87}
\defmmnum{ws1.keepnone.phi.procrustes.none.piqa}{51.90}
\defmmnum{ws1.keepnone.phi.procrustes.none.raw_rr}{0.024}
\defmmnum{ws1.keepnone.phi.procrustes.none.ret_hs}{45.0}
\defmmnum{ws1.keepnone.phi.procrustes.none.ret_piqa}{65.9}
\defmmnum{ws1.keepnone.phi.procrustes.none.rr}{0.024}
\defmmnum{ws1.keepnone.phi.procrustes.scaled_l2norm.anchors}{44{,}975}
\defmmnum{ws1.keepnone.phi.procrustes.scaled_l2norm.cos}{0.3397}
\defmmnum{ws1.keepnone.phi.procrustes.scaled_l2norm.hs}{27.85}
\defmmnum{ws1.keepnone.phi.procrustes.scaled_l2norm.p1}{97.87}
\defmmnum{ws1.keepnone.phi.procrustes.scaled_l2norm.piqa}{56.37}
\defmmnum{ws1.keepnone.phi.procrustes.scaled_l2norm.raw_rr}{0.024}
\defmmnum{ws1.keepnone.phi.procrustes.scaled_l2norm.ret_hs}{48.4}
\defmmnum{ws1.keepnone.phi.procrustes.scaled_l2norm.ret_piqa}{71.6}
\defmmnum{ws1.keepnone.phi.procrustes.scaled_l2norm.rr}{1.000}
\defmmnum{ws1.keepnone.qwen.lstsq.none.anchors}{82{,}663}
\defmmnum{ws1.keepnone.qwen.lstsq.none.cos}{0.6070}
\defmmnum{ws1.keepnone.qwen.lstsq.none.hs}{53.45}
\defmmnum{ws1.keepnone.qwen.lstsq.none.p1}{97.80}
\defmmnum{ws1.keepnone.qwen.lstsq.none.piqa}{77.48}
\defmmnum{ws1.keepnone.qwen.lstsq.none.raw_rr}{0.653}
\defmmnum{ws1.keepnone.qwen.lstsq.none.ret_hs}{93.0}
\defmmnum{ws1.keepnone.qwen.lstsq.none.ret_piqa}{98.4}
\defmmnum{ws1.keepnone.qwen.lstsq.none.rr}{0.653}
\defmmnum{ws1.keepnone.qwen.lstsq.scaled_l2norm.anchors}{82{,}663}
\defmmnum{ws1.keepnone.qwen.lstsq.scaled_l2norm.cos}{0.6070}
\defmmnum{ws1.keepnone.qwen.lstsq.scaled_l2norm.hs}{56.40}
\defmmnum{ws1.keepnone.qwen.lstsq.scaled_l2norm.p1}{97.80}
\defmmnum{ws1.keepnone.qwen.lstsq.scaled_l2norm.piqa}{77.48}
\defmmnum{ws1.keepnone.qwen.lstsq.scaled_l2norm.raw_rr}{0.653}
\defmmnum{ws1.keepnone.qwen.lstsq.scaled_l2norm.ret_hs}{98.1}
\defmmnum{ws1.keepnone.qwen.lstsq.scaled_l2norm.ret_piqa}{98.4}
\defmmnum{ws1.keepnone.qwen.lstsq.scaled_l2norm.rr}{1.000}
\defmmnum{ws1.keepnone.qwen.procrustes.none.anchors}{82{,}663}
\defmmnum{ws1.keepnone.qwen.procrustes.none.cos}{0.5296}
\defmmnum{ws1.keepnone.qwen.procrustes.none.hs}{26.90}
\defmmnum{ws1.keepnone.qwen.procrustes.none.p1}{97.59}
\defmmnum{ws1.keepnone.qwen.procrustes.none.piqa}{51.25}
\defmmnum{ws1.keepnone.qwen.procrustes.none.raw_rr}{0.023}
\defmmnum{ws1.keepnone.qwen.procrustes.none.ret_hs}{46.8}
\defmmnum{ws1.keepnone.qwen.procrustes.none.ret_piqa}{65.1}
\defmmnum{ws1.keepnone.qwen.procrustes.none.rr}{0.023}
\defmmnum{ws1.keepnone.qwen.procrustes.scaled_l2norm.anchors}{82{,}663}
\defmmnum{ws1.keepnone.qwen.procrustes.scaled_l2norm.cos}{0.5296}
\defmmnum{ws1.keepnone.qwen.procrustes.scaled_l2norm.hs}{27.95}
\defmmnum{ws1.keepnone.qwen.procrustes.scaled_l2norm.p1}{97.59}
\defmmnum{ws1.keepnone.qwen.procrustes.scaled_l2norm.piqa}{53.81}
\defmmnum{ws1.keepnone.qwen.procrustes.scaled_l2norm.raw_rr}{0.023}
\defmmnum{ws1.keepnone.qwen.procrustes.scaled_l2norm.ret_hs}{48.6}
\defmmnum{ws1.keepnone.qwen.procrustes.scaled_l2norm.ret_piqa}{68.3}
\defmmnum{ws1.keepnone.qwen.procrustes.scaled_l2norm.rr}{1.000}
\defmmnum{ws1.keepnone.smol.lstsq.none.anchors}{45{,}554}
\defmmnum{ws1.keepnone.smol.lstsq.none.cos}{0.6296}
\defmmnum{ws1.keepnone.smol.lstsq.none.hs}{54.35}
\defmmnum{ws1.keepnone.smol.lstsq.none.p1}{99.08}
\defmmnum{ws1.keepnone.smol.lstsq.none.piqa}{75.95}
\defmmnum{ws1.keepnone.smol.lstsq.none.raw_rr}{0.675}
\defmmnum{ws1.keepnone.smol.lstsq.none.ret_hs}{94.5}
\defmmnum{ws1.keepnone.smol.lstsq.none.ret_piqa}{96.5}
\defmmnum{ws1.keepnone.smol.lstsq.none.rr}{0.675}
\defmmnum{ws1.keepnone.smol.lstsq.scaled_l2norm.anchors}{45{,}554}
\defmmnum{ws1.keepnone.smol.lstsq.scaled_l2norm.cos}{0.6296}
\defmmnum{ws1.keepnone.smol.lstsq.scaled_l2norm.hs}{56.70}
\defmmnum{ws1.keepnone.smol.lstsq.scaled_l2norm.p1}{99.08}
\defmmnum{ws1.keepnone.smol.lstsq.scaled_l2norm.piqa}{76.82}
\defmmnum{ws1.keepnone.smol.lstsq.scaled_l2norm.raw_rr}{0.675}
\defmmnum{ws1.keepnone.smol.lstsq.scaled_l2norm.ret_hs}{98.6}
\defmmnum{ws1.keepnone.smol.lstsq.scaled_l2norm.ret_piqa}{97.6}
\defmmnum{ws1.keepnone.smol.lstsq.scaled_l2norm.rr}{1.000}
\defmmnum{ws1.keepnone.smol.procrustes.none.anchors}{45{,}554}
\defmmnum{ws1.keepnone.smol.procrustes.none.cos}{0.5397}
\defmmnum{ws1.keepnone.smol.procrustes.none.hs}{26.35}
\defmmnum{ws1.keepnone.smol.procrustes.none.p1}{98.64}
\defmmnum{ws1.keepnone.smol.procrustes.none.piqa}{52.39}
\defmmnum{ws1.keepnone.smol.procrustes.none.raw_rr}{0.060}
\defmmnum{ws1.keepnone.smol.procrustes.none.ret_hs}{45.8}
\defmmnum{ws1.keepnone.smol.procrustes.none.ret_piqa}{66.6}
\defmmnum{ws1.keepnone.smol.procrustes.none.rr}{0.060}
\defmmnum{ws1.keepnone.smol.procrustes.scaled_l2norm.anchors}{45{,}554}
\defmmnum{ws1.keepnone.smol.procrustes.scaled_l2norm.cos}{0.5397}
\defmmnum{ws1.keepnone.smol.procrustes.scaled_l2norm.hs}{26.10}
\defmmnum{ws1.keepnone.smol.procrustes.scaled_l2norm.p1}{98.64}
\defmmnum{ws1.keepnone.smol.procrustes.scaled_l2norm.piqa}{51.63}
\defmmnum{ws1.keepnone.smol.procrustes.scaled_l2norm.raw_rr}{0.060}
\defmmnum{ws1.keepnone.smol.procrustes.scaled_l2norm.ret_hs}{45.4}
\defmmnum{ws1.keepnone.smol.procrustes.scaled_l2norm.ret_piqa}{65.6}
\defmmnum{ws1.keepnone.smol.procrustes.scaled_l2norm.rr}{1.000}
\defmmnum{ws1.l2norm.max_hs}{27.10}
\defmmnum{ws1.l2norm.max_piqa}{52.45}
\defmmnum{ws1.l2norm.min_hs}{25.55}
\defmmnum{ws1.l2norm.min_piqa}{49.35}
\defmmnum{ws1.lstsq.donor_task_cells}{6}
\defmmnum{ws1.lstsq.l2_restore_max_gain_hs}{30.70}
\defmmnum{ws1.lstsq.l2_restore_min_gain_hs}{22.50}
\defmmnum{ws1.lstsq.none_max_rr}{0.675}
\defmmnum{ws1.lstsq.none_min_rr}{0.555}
\defmmnum{ws1.lstsq.phi_ln_over_scaled_hs}{1.45}
\defmmnum{ws1.lstsq.rms_ln_max_rr}{1.095}
\defmmnum{ws1.lstsq.rms_ln_min_rr}{1.069}
\defmmnum{ws1.lstsq.scaled_top_cells}{5}
\defmmnum{ws1.min.p1}{97.35}
\defmmnum{ws1.n_donors}{3}
\defmmnum{ws1.native.hs}{57.50}
\defmmnum{ws1.native.piqa}{78.73}
\defmmnum{ws1.native.rr}{1.000}
\defmmnum{ws1.phi.direct_swap.none.anchors}{44{,}975}
\defmmnum{ws1.phi.direct_swap.none.cos}{0.0002}
\defmmnum{ws1.phi.direct_swap.none.hs}{24.40}
\defmmnum{ws1.phi.direct_swap.none.p1}{0.01}
\defmmnum{ws1.phi.direct_swap.none.piqa}{50.27}
\defmmnum{ws1.phi.direct_swap.none.raw_rr}{0.024}
\defmmnum{ws1.phi.direct_swap.none.ret_hs}{42.4}
\defmmnum{ws1.phi.direct_swap.none.ret_piqa}{63.9}
\defmmnum{ws1.phi.direct_swap.none.rr}{0.024}
\defmmnum{ws1.phi.direct_swap.scaled_l2norm.anchors}{44{,}975}
\defmmnum{ws1.phi.direct_swap.scaled_l2norm.cos}{0.0002}
\defmmnum{ws1.phi.direct_swap.scaled_l2norm.hs}{26.45}
\defmmnum{ws1.phi.direct_swap.scaled_l2norm.p1}{0.01}
\defmmnum{ws1.phi.direct_swap.scaled_l2norm.piqa}{51.58}
\defmmnum{ws1.phi.direct_swap.scaled_l2norm.raw_rr}{0.024}
\defmmnum{ws1.phi.direct_swap.scaled_l2norm.ret_hs}{46.0}
\defmmnum{ws1.phi.direct_swap.scaled_l2norm.ret_piqa}{65.5}
\defmmnum{ws1.phi.direct_swap.scaled_l2norm.rr}{1.000}
\defmmnum{ws1.phi.lstsq.l2norm.anchors}{44{,}975}
\defmmnum{ws1.phi.lstsq.l2norm.cos}{0.4527}
\defmmnum{ws1.phi.lstsq.l2norm.hs}{26.80}
\defmmnum{ws1.phi.lstsq.l2norm.p1}{97.35}
\defmmnum{ws1.phi.lstsq.l2norm.piqa}{50.71}
\defmmnum{ws1.phi.lstsq.l2norm.raw_rr}{0.555}
\defmmnum{ws1.phi.lstsq.l2norm.ret_hs}{46.6}
\defmmnum{ws1.phi.lstsq.l2norm.ret_piqa}{64.4}
\defmmnum{ws1.phi.lstsq.l2norm.rr}{0.020}
\defmmnum{ws1.phi.lstsq.layernorm.anchors}{44{,}975}
\defmmnum{ws1.phi.lstsq.layernorm.cos}{0.4538}
\defmmnum{ws1.phi.lstsq.layernorm.hs}{50.75}
\defmmnum{ws1.phi.lstsq.layernorm.p1}{96.23}
\defmmnum{ws1.phi.lstsq.layernorm.piqa}{73.99}
\defmmnum{ws1.phi.lstsq.layernorm.raw_rr}{0.555}
\defmmnum{ws1.phi.lstsq.layernorm.ret_hs}{88.3}
\defmmnum{ws1.phi.lstsq.layernorm.ret_piqa}{94.0}
\defmmnum{ws1.phi.lstsq.layernorm.rr}{1.080}
\defmmnum{ws1.phi.lstsq.none.anchors}{44{,}975}
\defmmnum{ws1.phi.lstsq.none.cos}{0.4527}
\defmmnum{ws1.phi.lstsq.none.hs}{42.95}
\defmmnum{ws1.phi.lstsq.none.p1}{97.35}
\defmmnum{ws1.phi.lstsq.none.piqa}{70.95}
\defmmnum{ws1.phi.lstsq.none.raw_rr}{0.555}
\defmmnum{ws1.phi.lstsq.none.ret_hs}{74.7}
\defmmnum{ws1.phi.lstsq.none.ret_piqa}{90.1}
\defmmnum{ws1.phi.lstsq.none.rr}{0.555}
\defmmnum{ws1.phi.lstsq.rmsnorm.anchors}{44{,}975}
\defmmnum{ws1.phi.lstsq.rmsnorm.cos}{0.4445}
\defmmnum{ws1.phi.lstsq.rmsnorm.hs}{46.60}
\defmmnum{ws1.phi.lstsq.rmsnorm.p1}{96.51}
\defmmnum{ws1.phi.lstsq.rmsnorm.piqa}{71.33}
\defmmnum{ws1.phi.lstsq.rmsnorm.raw_rr}{0.555}
\defmmnum{ws1.phi.lstsq.rmsnorm.ret_hs}{81.0}
\defmmnum{ws1.phi.lstsq.rmsnorm.ret_piqa}{90.6}
\defmmnum{ws1.phi.lstsq.rmsnorm.rr}{1.095}
\defmmnum{ws1.phi.lstsq.scaled_l2norm.anchors}{44{,}975}
\defmmnum{ws1.phi.lstsq.scaled_l2norm.cos}{0.4527}
\defmmnum{ws1.phi.lstsq.scaled_l2norm.hs}{49.30}
\defmmnum{ws1.phi.lstsq.scaled_l2norm.p1}{97.35}
\defmmnum{ws1.phi.lstsq.scaled_l2norm.piqa}{74.16}
\defmmnum{ws1.phi.lstsq.scaled_l2norm.raw_rr}{0.555}
\defmmnum{ws1.phi.lstsq.scaled_l2norm.ret_hs}{85.7}
\defmmnum{ws1.phi.lstsq.scaled_l2norm.ret_piqa}{94.2}
\defmmnum{ws1.phi.lstsq.scaled_l2norm.rr}{1.000}
\defmmnum{ws1.phi.lstsq_l2_restore_gain_hs}{+22.50}
\defmmnum{ws1.phi.lstsq_l2_restore_gain_piqa}{+23.45}
\defmmnum{ws1.phi.lstsq_norm_gain_hs}{+6.35}
\defmmnum{ws1.phi.lstsq_norm_gain_piqa}{+3.21}
\defmmnum{ws1.phi.procrustes.l2norm.anchors}{44{,}975}
\defmmnum{ws1.phi.procrustes.l2norm.cos}{0.3397}
\defmmnum{ws1.phi.procrustes.l2norm.hs}{25.55}
\defmmnum{ws1.phi.procrustes.l2norm.p1}{97.87}
\defmmnum{ws1.phi.procrustes.l2norm.piqa}{52.45}
\defmmnum{ws1.phi.procrustes.l2norm.raw_rr}{0.024}
\defmmnum{ws1.phi.procrustes.l2norm.ret_hs}{44.4}
\defmmnum{ws1.phi.procrustes.l2norm.ret_piqa}{66.6}
\defmmnum{ws1.phi.procrustes.l2norm.rr}{0.020}
\defmmnum{ws1.phi.procrustes.layernorm.anchors}{44{,}975}
\defmmnum{ws1.phi.procrustes.layernorm.cos}{0.3667}
\defmmnum{ws1.phi.procrustes.layernorm.hs}{54.10}
\defmmnum{ws1.phi.procrustes.layernorm.p1}{96.98}
\defmmnum{ws1.phi.procrustes.layernorm.piqa}{75.30}
\defmmnum{ws1.phi.procrustes.layernorm.raw_rr}{0.024}
\defmmnum{ws1.phi.procrustes.layernorm.ret_hs}{94.1}
\defmmnum{ws1.phi.procrustes.layernorm.ret_piqa}{95.6}
\defmmnum{ws1.phi.procrustes.layernorm.rr}{1.014}
\defmmnum{ws1.phi.procrustes.none.anchors}{44{,}975}
\defmmnum{ws1.phi.procrustes.none.cos}{0.3397}
\defmmnum{ws1.phi.procrustes.none.hs}{25.85}
\defmmnum{ws1.phi.procrustes.none.p1}{97.87}
\defmmnum{ws1.phi.procrustes.none.piqa}{51.96}
\defmmnum{ws1.phi.procrustes.none.raw_rr}{0.024}
\defmmnum{ws1.phi.procrustes.none.ret_hs}{45.0}
\defmmnum{ws1.phi.procrustes.none.ret_piqa}{66.0}
\defmmnum{ws1.phi.procrustes.none.rr}{0.024}
\defmmnum{ws1.phi.procrustes.rmsnorm.anchors}{44{,}975}
\defmmnum{ws1.phi.procrustes.rmsnorm.cos}{0.3460}
\defmmnum{ws1.phi.procrustes.rmsnorm.hs}{28.80}
\defmmnum{ws1.phi.procrustes.rmsnorm.p1}{97.92}
\defmmnum{ws1.phi.procrustes.rmsnorm.piqa}{55.55}
\defmmnum{ws1.phi.procrustes.rmsnorm.raw_rr}{0.024}
\defmmnum{ws1.phi.procrustes.rmsnorm.ret_hs}{50.1}
\defmmnum{ws1.phi.procrustes.rmsnorm.ret_piqa}{70.6}
\defmmnum{ws1.phi.procrustes.rmsnorm.rr}{1.010}
\defmmnum{ws1.phi.procrustes.scaled_l2norm.anchors}{44{,}975}
\defmmnum{ws1.phi.procrustes.scaled_l2norm.cos}{0.3397}
\defmmnum{ws1.phi.procrustes.scaled_l2norm.hs}{27.90}
\defmmnum{ws1.phi.procrustes.scaled_l2norm.p1}{97.87}
\defmmnum{ws1.phi.procrustes.scaled_l2norm.piqa}{56.37}
\defmmnum{ws1.phi.procrustes.scaled_l2norm.raw_rr}{0.024}
\defmmnum{ws1.phi.procrustes.scaled_l2norm.ret_hs}{48.5}
\defmmnum{ws1.phi.procrustes.scaled_l2norm.ret_piqa}{71.6}
\defmmnum{ws1.phi.procrustes.scaled_l2norm.rr}{1.000}
\defmmnum{ws1.phi.random_orthogonal.none.anchors}{44{,}975}
\defmmnum{ws1.phi.random_orthogonal.none.cos}{0.0000}
\defmmnum{ws1.phi.random_orthogonal.none.hs}{26.45}
\defmmnum{ws1.phi.random_orthogonal.none.p1}{0.00}
\defmmnum{ws1.phi.random_orthogonal.none.piqa}{52.23}
\defmmnum{ws1.phi.random_orthogonal.none.raw_rr}{0.024}
\defmmnum{ws1.phi.random_orthogonal.none.ret_hs}{46.0}
\defmmnum{ws1.phi.random_orthogonal.none.ret_piqa}{66.3}
\defmmnum{ws1.phi.random_orthogonal.none.rr}{0.024}
\defmmnum{ws1.phi.random_orthogonal.scaled_l2norm.anchors}{44{,}975}
\defmmnum{ws1.phi.random_orthogonal.scaled_l2norm.cos}{0.0000}
\defmmnum{ws1.phi.random_orthogonal.scaled_l2norm.hs}{25.80}
\defmmnum{ws1.phi.random_orthogonal.scaled_l2norm.p1}{0.00}
\defmmnum{ws1.phi.random_orthogonal.scaled_l2norm.piqa}{48.53}
\defmmnum{ws1.phi.random_orthogonal.scaled_l2norm.raw_rr}{0.024}
\defmmnum{ws1.phi.random_orthogonal.scaled_l2norm.ret_hs}{44.9}
\defmmnum{ws1.phi.random_orthogonal.scaled_l2norm.ret_piqa}{61.6}
\defmmnum{ws1.phi.random_orthogonal.scaled_l2norm.rr}{1.000}
\defmmnum{ws1.proc.best_hs}{27.95}
\defmmnum{ws1.proc.best_piqa}{56.37}
\defmmnum{ws1.proc_5norm.best_hs}{54.10}
\defmmnum{ws1.proc_5norm.best_piqa}{75.30}
\defmmnum{ws1.qwen.cl_replica.anchors}{65{,}406}
\defmmnum{ws1.qwen.cl_replica.cos}{0.6329}
\defmmnum{ws1.qwen.cl_replica.hs}{54.50}
\defmmnum{ws1.qwen.cl_replica.p1}{97.76}
\defmmnum{ws1.qwen.cl_replica.piqa}{77.26}
\defmmnum{ws1.qwen.cl_replica.raw_rr}{0.681}
\defmmnum{ws1.qwen.cl_replica.ret_hs}{94.8}
\defmmnum{ws1.qwen.cl_replica.ret_piqa}{98.1}
\defmmnum{ws1.qwen.cl_replica.rr}{0.681}
\defmmnum{ws1.qwen.direct_swap.none.anchors}{82{,}663}
\defmmnum{ws1.qwen.direct_swap.none.cos}{-0.0008}
\defmmnum{ws1.qwen.direct_swap.none.hs}{24.35}
\defmmnum{ws1.qwen.direct_swap.none.p1}{0.02}
\defmmnum{ws1.qwen.direct_swap.none.piqa}{49.29}
\defmmnum{ws1.qwen.direct_swap.none.raw_rr}{0.023}
\defmmnum{ws1.qwen.direct_swap.none.ret_hs}{42.3}
\defmmnum{ws1.qwen.direct_swap.none.ret_piqa}{62.6}
\defmmnum{ws1.qwen.direct_swap.none.rr}{0.023}
\defmmnum{ws1.qwen.direct_swap.scaled_l2norm.anchors}{82{,}663}
\defmmnum{ws1.qwen.direct_swap.scaled_l2norm.cos}{-0.0008}
\defmmnum{ws1.qwen.direct_swap.scaled_l2norm.hs}{25.40}
\defmmnum{ws1.qwen.direct_swap.scaled_l2norm.p1}{0.02}
\defmmnum{ws1.qwen.direct_swap.scaled_l2norm.piqa}{49.40}
\defmmnum{ws1.qwen.direct_swap.scaled_l2norm.raw_rr}{0.023}
\defmmnum{ws1.qwen.direct_swap.scaled_l2norm.ret_hs}{44.2}
\defmmnum{ws1.qwen.direct_swap.scaled_l2norm.ret_piqa}{62.8}
\defmmnum{ws1.qwen.direct_swap.scaled_l2norm.rr}{1.000}
\defmmnum{ws1.qwen.lstsq.l2norm.anchors}{82{,}663}
\defmmnum{ws1.qwen.lstsq.l2norm.cos}{0.6070}
\defmmnum{ws1.qwen.lstsq.l2norm.hs}{26.55}
\defmmnum{ws1.qwen.lstsq.l2norm.p1}{97.80}
\defmmnum{ws1.qwen.lstsq.l2norm.piqa}{52.12}
\defmmnum{ws1.qwen.lstsq.l2norm.raw_rr}{0.653}
\defmmnum{ws1.qwen.lstsq.l2norm.ret_hs}{46.2}
\defmmnum{ws1.qwen.lstsq.l2norm.ret_piqa}{66.2}
\defmmnum{ws1.qwen.lstsq.l2norm.rr}{0.020}
\defmmnum{ws1.qwen.lstsq.layernorm.anchors}{82{,}663}
\defmmnum{ws1.qwen.lstsq.layernorm.cos}{0.5959}
\defmmnum{ws1.qwen.lstsq.layernorm.hs}{55.25}
\defmmnum{ws1.qwen.lstsq.layernorm.p1}{97.50}
\defmmnum{ws1.qwen.lstsq.layernorm.piqa}{76.55}
\defmmnum{ws1.qwen.lstsq.layernorm.raw_rr}{0.653}
\defmmnum{ws1.qwen.lstsq.layernorm.ret_hs}{96.1}
\defmmnum{ws1.qwen.lstsq.layernorm.ret_piqa}{97.2}
\defmmnum{ws1.qwen.lstsq.layernorm.rr}{1.069}
\defmmnum{ws1.qwen.lstsq.none.anchors}{82{,}663}
\defmmnum{ws1.qwen.lstsq.none.cos}{0.6070}
\defmmnum{ws1.qwen.lstsq.none.hs}{53.45}
\defmmnum{ws1.qwen.lstsq.none.p1}{97.80}
\defmmnum{ws1.qwen.lstsq.none.piqa}{77.48}
\defmmnum{ws1.qwen.lstsq.none.raw_rr}{0.653}
\defmmnum{ws1.qwen.lstsq.none.ret_hs}{93.0}
\defmmnum{ws1.qwen.lstsq.none.ret_piqa}{98.4}
\defmmnum{ws1.qwen.lstsq.none.rr}{0.653}
\defmmnum{ws1.qwen.lstsq.rmsnorm.anchors}{82{,}663}
\defmmnum{ws1.qwen.lstsq.rmsnorm.cos}{0.5927}
\defmmnum{ws1.qwen.lstsq.rmsnorm.hs}{55.60}
\defmmnum{ws1.qwen.lstsq.rmsnorm.p1}{97.82}
\defmmnum{ws1.qwen.lstsq.rmsnorm.piqa}{76.50}
\defmmnum{ws1.qwen.lstsq.rmsnorm.raw_rr}{0.653}
\defmmnum{ws1.qwen.lstsq.rmsnorm.ret_hs}{96.7}
\defmmnum{ws1.qwen.lstsq.rmsnorm.ret_piqa}{97.2}
\defmmnum{ws1.qwen.lstsq.rmsnorm.rr}{1.075}
\defmmnum{ws1.qwen.lstsq.scaled_l2norm.anchors}{82{,}663}
\defmmnum{ws1.qwen.lstsq.scaled_l2norm.cos}{0.6070}
\defmmnum{ws1.qwen.lstsq.scaled_l2norm.hs}{56.40}
\defmmnum{ws1.qwen.lstsq.scaled_l2norm.p1}{97.80}
\defmmnum{ws1.qwen.lstsq.scaled_l2norm.piqa}{77.48}
\defmmnum{ws1.qwen.lstsq.scaled_l2norm.raw_rr}{0.653}
\defmmnum{ws1.qwen.lstsq.scaled_l2norm.ret_hs}{98.1}
\defmmnum{ws1.qwen.lstsq.scaled_l2norm.ret_piqa}{98.4}
\defmmnum{ws1.qwen.lstsq.scaled_l2norm.rr}{1.000}
\defmmnum{ws1.qwen.lstsq_l2_restore_gain_hs}{+29.85}
\defmmnum{ws1.qwen.lstsq_l2_restore_gain_piqa}{+25.35}
\defmmnum{ws1.qwen.lstsq_norm_gain_hs}{+2.95}
\defmmnum{ws1.qwen.lstsq_norm_gain_piqa}{+0.00}
\defmmnum{ws1.qwen.procrustes.l2norm.anchors}{82{,}663}
\defmmnum{ws1.qwen.procrustes.l2norm.cos}{0.5296}
\defmmnum{ws1.qwen.procrustes.l2norm.hs}{27.10}
\defmmnum{ws1.qwen.procrustes.l2norm.p1}{97.59}
\defmmnum{ws1.qwen.procrustes.l2norm.piqa}{50.38}
\defmmnum{ws1.qwen.procrustes.l2norm.raw_rr}{0.023}
\defmmnum{ws1.qwen.procrustes.l2norm.ret_hs}{47.1}
\defmmnum{ws1.qwen.procrustes.l2norm.ret_piqa}{64.0}
\defmmnum{ws1.qwen.procrustes.l2norm.rr}{0.020}
\defmmnum{ws1.qwen.procrustes.layernorm.anchors}{82{,}663}
\defmmnum{ws1.qwen.procrustes.layernorm.cos}{0.5270}
\defmmnum{ws1.qwen.procrustes.layernorm.hs}{38.35}
\defmmnum{ws1.qwen.procrustes.layernorm.p1}{97.25}
\defmmnum{ws1.qwen.procrustes.layernorm.piqa}{64.47}
\defmmnum{ws1.qwen.procrustes.layernorm.raw_rr}{0.023}
\defmmnum{ws1.qwen.procrustes.layernorm.ret_hs}{66.7}
\defmmnum{ws1.qwen.procrustes.layernorm.ret_piqa}{81.9}
\defmmnum{ws1.qwen.procrustes.layernorm.rr}{1.063}
\defmmnum{ws1.qwen.procrustes.none.anchors}{82{,}663}
\defmmnum{ws1.qwen.procrustes.none.cos}{0.5296}
\defmmnum{ws1.qwen.procrustes.none.hs}{26.90}
\defmmnum{ws1.qwen.procrustes.none.p1}{97.59}
\defmmnum{ws1.qwen.procrustes.none.piqa}{51.25}
\defmmnum{ws1.qwen.procrustes.none.raw_rr}{0.023}
\defmmnum{ws1.qwen.procrustes.none.ret_hs}{46.8}
\defmmnum{ws1.qwen.procrustes.none.ret_piqa}{65.1}
\defmmnum{ws1.qwen.procrustes.none.rr}{0.023}
\defmmnum{ws1.qwen.procrustes.rmsnorm.anchors}{82{,}663}
\defmmnum{ws1.qwen.procrustes.rmsnorm.cos}{0.5234}
\defmmnum{ws1.qwen.procrustes.rmsnorm.hs}{27.25}
\defmmnum{ws1.qwen.procrustes.rmsnorm.p1}{97.63}
\defmmnum{ws1.qwen.procrustes.rmsnorm.piqa}{52.77}
\defmmnum{ws1.qwen.procrustes.rmsnorm.raw_rr}{0.023}
\defmmnum{ws1.qwen.procrustes.rmsnorm.ret_hs}{47.4}
\defmmnum{ws1.qwen.procrustes.rmsnorm.ret_piqa}{67.0}
\defmmnum{ws1.qwen.procrustes.rmsnorm.rr}{1.064}
\defmmnum{ws1.qwen.procrustes.scaled_l2norm.anchors}{82{,}663}
\defmmnum{ws1.qwen.procrustes.scaled_l2norm.cos}{0.5296}
\defmmnum{ws1.qwen.procrustes.scaled_l2norm.hs}{27.95}
\defmmnum{ws1.qwen.procrustes.scaled_l2norm.p1}{97.59}
\defmmnum{ws1.qwen.procrustes.scaled_l2norm.piqa}{53.86}
\defmmnum{ws1.qwen.procrustes.scaled_l2norm.raw_rr}{0.023}
\defmmnum{ws1.qwen.procrustes.scaled_l2norm.ret_hs}{48.6}
\defmmnum{ws1.qwen.procrustes.scaled_l2norm.ret_piqa}{68.4}
\defmmnum{ws1.qwen.procrustes.scaled_l2norm.rr}{1.000}
\defmmnum{ws1.qwen.random_orthogonal.none.anchors}{82{,}663}
\defmmnum{ws1.qwen.random_orthogonal.none.cos}{0.0005}
\defmmnum{ws1.qwen.random_orthogonal.none.hs}{25.45}
\defmmnum{ws1.qwen.random_orthogonal.none.p1}{0.02}
\defmmnum{ws1.qwen.random_orthogonal.none.piqa}{50.60}
\defmmnum{ws1.qwen.random_orthogonal.none.raw_rr}{0.023}
\defmmnum{ws1.qwen.random_orthogonal.none.ret_hs}{44.3}
\defmmnum{ws1.qwen.random_orthogonal.none.ret_piqa}{64.3}
\defmmnum{ws1.qwen.random_orthogonal.none.rr}{0.023}
\defmmnum{ws1.qwen.random_orthogonal.scaled_l2norm.anchors}{82{,}663}
\defmmnum{ws1.qwen.random_orthogonal.scaled_l2norm.cos}{0.0005}
\defmmnum{ws1.qwen.random_orthogonal.scaled_l2norm.hs}{25.80}
\defmmnum{ws1.qwen.random_orthogonal.scaled_l2norm.p1}{0.02}
\defmmnum{ws1.qwen.random_orthogonal.scaled_l2norm.piqa}{49.78}
\defmmnum{ws1.qwen.random_orthogonal.scaled_l2norm.raw_rr}{0.023}
\defmmnum{ws1.qwen.random_orthogonal.scaled_l2norm.ret_hs}{44.9}
\defmmnum{ws1.qwen.random_orthogonal.scaled_l2norm.ret_piqa}{63.2}
\defmmnum{ws1.qwen.random_orthogonal.scaled_l2norm.rr}{1.000}
\defmmnum{ws1.smol.lstsq.l2norm.anchors}{45{,}554}
\defmmnum{ws1.smol.lstsq.l2norm.cos}{0.6296}
\defmmnum{ws1.smol.lstsq.l2norm.hs}{26.00}
\defmmnum{ws1.smol.lstsq.l2norm.p1}{99.08}
\defmmnum{ws1.smol.lstsq.l2norm.piqa}{51.96}
\defmmnum{ws1.smol.lstsq.l2norm.raw_rr}{0.675}
\defmmnum{ws1.smol.lstsq.l2norm.ret_hs}{45.2}
\defmmnum{ws1.smol.lstsq.l2norm.ret_piqa}{66.0}
\defmmnum{ws1.smol.lstsq.l2norm.rr}{0.020}
\defmmnum{ws1.smol.lstsq.layernorm.anchors}{45{,}554}
\defmmnum{ws1.smol.lstsq.layernorm.cos}{0.6175}
\defmmnum{ws1.smol.lstsq.layernorm.hs}{54.35}
\defmmnum{ws1.smol.lstsq.layernorm.p1}{98.81}
\defmmnum{ws1.smol.lstsq.layernorm.piqa}{76.22}
\defmmnum{ws1.smol.lstsq.layernorm.raw_rr}{0.675}
\defmmnum{ws1.smol.lstsq.layernorm.ret_hs}{94.5}
\defmmnum{ws1.smol.lstsq.layernorm.ret_piqa}{96.8}
\defmmnum{ws1.smol.lstsq.layernorm.rr}{1.070}
\defmmnum{ws1.smol.lstsq.none.anchors}{45{,}554}
\defmmnum{ws1.smol.lstsq.none.cos}{0.6296}
\defmmnum{ws1.smol.lstsq.none.hs}{54.35}
\defmmnum{ws1.smol.lstsq.none.p1}{99.08}
\defmmnum{ws1.smol.lstsq.none.piqa}{75.95}
\defmmnum{ws1.smol.lstsq.none.raw_rr}{0.675}
\defmmnum{ws1.smol.lstsq.none.ret_hs}{94.5}
\defmmnum{ws1.smol.lstsq.none.ret_piqa}{96.5}
\defmmnum{ws1.smol.lstsq.none.rr}{0.675}
\defmmnum{ws1.smol.lstsq.rmsnorm.anchors}{45{,}554}
\defmmnum{ws1.smol.lstsq.rmsnorm.cos}{0.6140}
\defmmnum{ws1.smol.lstsq.rmsnorm.hs}{55.55}
\defmmnum{ws1.smol.lstsq.rmsnorm.p1}{99.06}
\defmmnum{ws1.smol.lstsq.rmsnorm.piqa}{75.52}
\defmmnum{ws1.smol.lstsq.rmsnorm.raw_rr}{0.675}
\defmmnum{ws1.smol.lstsq.rmsnorm.ret_hs}{96.6}
\defmmnum{ws1.smol.lstsq.rmsnorm.ret_piqa}{95.9}
\defmmnum{ws1.smol.lstsq.rmsnorm.rr}{1.073}
\defmmnum{ws1.smol.lstsq.scaled_l2norm.anchors}{45{,}554}
\defmmnum{ws1.smol.lstsq.scaled_l2norm.cos}{0.6296}
\defmmnum{ws1.smol.lstsq.scaled_l2norm.hs}{56.70}
\defmmnum{ws1.smol.lstsq.scaled_l2norm.p1}{99.08}
\defmmnum{ws1.smol.lstsq.scaled_l2norm.piqa}{76.82}
\defmmnum{ws1.smol.lstsq.scaled_l2norm.raw_rr}{0.675}
\defmmnum{ws1.smol.lstsq.scaled_l2norm.ret_hs}{98.6}
\defmmnum{ws1.smol.lstsq.scaled_l2norm.ret_piqa}{97.6}
\defmmnum{ws1.smol.lstsq.scaled_l2norm.rr}{1.000}
\defmmnum{ws1.smol.lstsq_l2_restore_gain_hs}{+30.70}
\defmmnum{ws1.smol.lstsq_l2_restore_gain_piqa}{+24.86}
\defmmnum{ws1.smol.lstsq_norm_gain_hs}{+2.35}
\defmmnum{ws1.smol.lstsq_norm_gain_piqa}{+0.87}
\defmmnum{ws1.smol.procrustes.l2norm.anchors}{45{,}554}
\defmmnum{ws1.smol.procrustes.l2norm.cos}{0.5397}
\defmmnum{ws1.smol.procrustes.l2norm.hs}{25.90}
\defmmnum{ws1.smol.procrustes.l2norm.p1}{98.64}
\defmmnum{ws1.smol.procrustes.l2norm.piqa}{49.35}
\defmmnum{ws1.smol.procrustes.l2norm.raw_rr}{0.060}
\defmmnum{ws1.smol.procrustes.l2norm.ret_hs}{45.0}
\defmmnum{ws1.smol.procrustes.l2norm.ret_piqa}{62.7}
\defmmnum{ws1.smol.procrustes.l2norm.rr}{0.020}
\defmmnum{ws1.smol.procrustes.layernorm.anchors}{45{,}554}
\defmmnum{ws1.smol.procrustes.layernorm.cos}{0.5391}
\defmmnum{ws1.smol.procrustes.layernorm.hs}{30.65}
\defmmnum{ws1.smol.procrustes.layernorm.p1}{97.78}
\defmmnum{ws1.smol.procrustes.layernorm.piqa}{56.26}
\defmmnum{ws1.smol.procrustes.layernorm.raw_rr}{0.060}
\defmmnum{ws1.smol.procrustes.layernorm.ret_hs}{53.3}
\defmmnum{ws1.smol.procrustes.layernorm.ret_piqa}{71.5}
\defmmnum{ws1.smol.procrustes.layernorm.rr}{1.064}
\defmmnum{ws1.smol.procrustes.none.anchors}{45{,}554}
\defmmnum{ws1.smol.procrustes.none.cos}{0.5397}
\defmmnum{ws1.smol.procrustes.none.hs}{26.35}
\defmmnum{ws1.smol.procrustes.none.p1}{98.64}
\defmmnum{ws1.smol.procrustes.none.piqa}{52.34}
\defmmnum{ws1.smol.procrustes.none.raw_rr}{0.060}
\defmmnum{ws1.smol.procrustes.none.ret_hs}{45.8}
\defmmnum{ws1.smol.procrustes.none.ret_piqa}{66.5}
\defmmnum{ws1.smol.procrustes.none.rr}{0.060}
\defmmnum{ws1.smol.procrustes.rmsnorm.anchors}{45{,}554}
\defmmnum{ws1.smol.procrustes.rmsnorm.cos}{0.5331}
\defmmnum{ws1.smol.procrustes.rmsnorm.hs}{25.75}
\defmmnum{ws1.smol.procrustes.rmsnorm.p1}{98.01}
\defmmnum{ws1.smol.procrustes.rmsnorm.piqa}{50.76}
\defmmnum{ws1.smol.procrustes.rmsnorm.raw_rr}{0.060}
\defmmnum{ws1.smol.procrustes.rmsnorm.ret_hs}{44.8}
\defmmnum{ws1.smol.procrustes.rmsnorm.ret_piqa}{64.5}
\defmmnum{ws1.smol.procrustes.rmsnorm.rr}{1.064}
\defmmnum{ws1.smol.procrustes.scaled_l2norm.anchors}{45{,}554}
\defmmnum{ws1.smol.procrustes.scaled_l2norm.cos}{0.5397}
\defmmnum{ws1.smol.procrustes.scaled_l2norm.hs}{26.10}
\defmmnum{ws1.smol.procrustes.scaled_l2norm.p1}{98.64}
\defmmnum{ws1.smol.procrustes.scaled_l2norm.piqa}{51.63}
\defmmnum{ws1.smol.procrustes.scaled_l2norm.raw_rr}{0.060}
\defmmnum{ws1.smol.procrustes.scaled_l2norm.ret_hs}{45.4}
\defmmnum{ws1.smol.procrustes.scaled_l2norm.ret_piqa}{65.6}
\defmmnum{ws1.smol.procrustes.scaled_l2norm.rr}{1.000}
\defmmnum{ws1.smol.random_orthogonal.none.anchors}{45{,}554}
\defmmnum{ws1.smol.random_orthogonal.none.cos}{-0.0018}
\defmmnum{ws1.smol.random_orthogonal.none.hs}{25.35}
\defmmnum{ws1.smol.random_orthogonal.none.p1}{0.01}
\defmmnum{ws1.smol.random_orthogonal.none.piqa}{48.42}
\defmmnum{ws1.smol.random_orthogonal.none.raw_rr}{0.060}
\defmmnum{ws1.smol.random_orthogonal.none.ret_hs}{44.1}
\defmmnum{ws1.smol.random_orthogonal.none.ret_piqa}{61.5}
\defmmnum{ws1.smol.random_orthogonal.none.rr}{0.060}
\defmmnum{ws1.smol.random_orthogonal.scaled_l2norm.anchors}{45{,}554}
\defmmnum{ws1.smol.random_orthogonal.scaled_l2norm.cos}{-0.0018}
\defmmnum{ws1.smol.random_orthogonal.scaled_l2norm.hs}{24.95}
\defmmnum{ws1.smol.random_orthogonal.scaled_l2norm.p1}{0.01}
\defmmnum{ws1.smol.random_orthogonal.scaled_l2norm.piqa}{50.00}
\defmmnum{ws1.smol.random_orthogonal.scaled_l2norm.raw_rr}{0.060}
\defmmnum{ws1.smol.random_orthogonal.scaled_l2norm.ret_hs}{43.4}
\defmmnum{ws1.smol.random_orthogonal.scaled_l2norm.ret_piqa}{63.5}
\defmmnum{ws1.smol.random_orthogonal.scaled_l2norm.rr}{1.000}
\defmmnum{ws1.tied.phi.lstsq.none.anchors}{44{,}975}
\defmmnum{ws1.tied.phi.lstsq.none.cos}{0.4527}
\defmmnum{ws1.tied.phi.lstsq.none.hs}{37.75}
\defmmnum{ws1.tied.phi.lstsq.none.p1}{97.35}
\defmmnum{ws1.tied.phi.lstsq.none.piqa}{65.23}
\defmmnum{ws1.tied.phi.lstsq.none.raw_rr}{0.555}
\defmmnum{ws1.tied.phi.lstsq.none.ret_hs}{65.7}
\defmmnum{ws1.tied.phi.lstsq.none.ret_piqa}{82.9}
\defmmnum{ws1.tied.phi.lstsq.none.rr}{0.555}
\defmmnum{ws1.tied.phi.lstsq.scaled_l2norm.anchors}{44{,}975}
\defmmnum{ws1.tied.phi.lstsq.scaled_l2norm.cos}{0.4527}
\defmmnum{ws1.tied.phi.lstsq.scaled_l2norm.hs}{45.65}
\defmmnum{ws1.tied.phi.lstsq.scaled_l2norm.p1}{97.35}
\defmmnum{ws1.tied.phi.lstsq.scaled_l2norm.piqa}{69.10}
\defmmnum{ws1.tied.phi.lstsq.scaled_l2norm.raw_rr}{0.555}
\defmmnum{ws1.tied.phi.lstsq.scaled_l2norm.ret_hs}{79.4}
\defmmnum{ws1.tied.phi.lstsq.scaled_l2norm.ret_piqa}{87.8}
\defmmnum{ws1.tied.phi.lstsq.scaled_l2norm.rr}{1.000}
\defmmnum{ws1.tied.phi.procrustes.none.anchors}{44{,}975}
\defmmnum{ws1.tied.phi.procrustes.none.cos}{0.3397}
\defmmnum{ws1.tied.phi.procrustes.none.hs}{26.60}
\defmmnum{ws1.tied.phi.procrustes.none.p1}{97.87}
\defmmnum{ws1.tied.phi.procrustes.none.piqa}{51.96}
\defmmnum{ws1.tied.phi.procrustes.none.raw_rr}{0.024}
\defmmnum{ws1.tied.phi.procrustes.none.ret_hs}{46.3}
\defmmnum{ws1.tied.phi.procrustes.none.ret_piqa}{66.0}
\defmmnum{ws1.tied.phi.procrustes.none.rr}{0.024}
\defmmnum{ws1.tied.phi.procrustes.scaled_l2norm.anchors}{44{,}975}
\defmmnum{ws1.tied.phi.procrustes.scaled_l2norm.cos}{0.3397}
\defmmnum{ws1.tied.phi.procrustes.scaled_l2norm.hs}{28.75}
\defmmnum{ws1.tied.phi.procrustes.scaled_l2norm.p1}{97.87}
\defmmnum{ws1.tied.phi.procrustes.scaled_l2norm.piqa}{56.26}
\defmmnum{ws1.tied.phi.procrustes.scaled_l2norm.raw_rr}{0.024}
\defmmnum{ws1.tied.phi.procrustes.scaled_l2norm.ret_hs}{50.0}
\defmmnum{ws1.tied.phi.procrustes.scaled_l2norm.ret_piqa}{71.5}
\defmmnum{ws1.tied.phi.procrustes.scaled_l2norm.rr}{1.000}
\defmmnum{ws1.tied.qwen.lstsq.none.anchors}{82{,}663}
\defmmnum{ws1.tied.qwen.lstsq.none.cos}{0.6070}
\defmmnum{ws1.tied.qwen.lstsq.none.hs}{49.15}
\defmmnum{ws1.tied.qwen.lstsq.none.p1}{97.80}
\defmmnum{ws1.tied.qwen.lstsq.none.piqa}{71.55}
\defmmnum{ws1.tied.qwen.lstsq.none.raw_rr}{0.653}
\defmmnum{ws1.tied.qwen.lstsq.none.ret_hs}{85.5}
\defmmnum{ws1.tied.qwen.lstsq.none.ret_piqa}{90.9}
\defmmnum{ws1.tied.qwen.lstsq.none.rr}{0.653}
\defmmnum{ws1.tied.qwen.lstsq.scaled_l2norm.anchors}{82{,}663}
\defmmnum{ws1.tied.qwen.lstsq.scaled_l2norm.cos}{0.6070}
\defmmnum{ws1.tied.qwen.lstsq.scaled_l2norm.hs}{52.60}
\defmmnum{ws1.tied.qwen.lstsq.scaled_l2norm.p1}{97.80}
\defmmnum{ws1.tied.qwen.lstsq.scaled_l2norm.piqa}{74.27}
\defmmnum{ws1.tied.qwen.lstsq.scaled_l2norm.raw_rr}{0.653}
\defmmnum{ws1.tied.qwen.lstsq.scaled_l2norm.ret_hs}{91.5}
\defmmnum{ws1.tied.qwen.lstsq.scaled_l2norm.ret_piqa}{94.3}
\defmmnum{ws1.tied.qwen.lstsq.scaled_l2norm.rr}{1.000}
\defmmnum{ws1.tied.qwen.procrustes.none.anchors}{82{,}663}
\defmmnum{ws1.tied.qwen.procrustes.none.cos}{0.5296}
\defmmnum{ws1.tied.qwen.procrustes.none.hs}{26.65}
\defmmnum{ws1.tied.qwen.procrustes.none.p1}{97.59}
\defmmnum{ws1.tied.qwen.procrustes.none.piqa}{51.36}
\defmmnum{ws1.tied.qwen.procrustes.none.raw_rr}{0.023}
\defmmnum{ws1.tied.qwen.procrustes.none.ret_hs}{46.3}
\defmmnum{ws1.tied.qwen.procrustes.none.ret_piqa}{65.2}
\defmmnum{ws1.tied.qwen.procrustes.none.rr}{0.023}
\defmmnum{ws1.tied.qwen.procrustes.scaled_l2norm.anchors}{82{,}663}
\defmmnum{ws1.tied.qwen.procrustes.scaled_l2norm.cos}{0.5296}
\defmmnum{ws1.tied.qwen.procrustes.scaled_l2norm.hs}{28.55}
\defmmnum{ws1.tied.qwen.procrustes.scaled_l2norm.p1}{97.59}
\defmmnum{ws1.tied.qwen.procrustes.scaled_l2norm.piqa}{53.81}
\defmmnum{ws1.tied.qwen.procrustes.scaled_l2norm.raw_rr}{0.023}
\defmmnum{ws1.tied.qwen.procrustes.scaled_l2norm.ret_hs}{49.7}
\defmmnum{ws1.tied.qwen.procrustes.scaled_l2norm.ret_piqa}{68.3}
\defmmnum{ws1.tied.qwen.procrustes.scaled_l2norm.rr}{1.000}
\defmmnum{ws1.tied.smol.lstsq.none.anchors}{45{,}554}
\defmmnum{ws1.tied.smol.lstsq.none.cos}{0.6296}
\defmmnum{ws1.tied.smol.lstsq.none.hs}{47.55}
\defmmnum{ws1.tied.smol.lstsq.none.p1}{99.08}
\defmmnum{ws1.tied.smol.lstsq.none.piqa}{71.00}
\defmmnum{ws1.tied.smol.lstsq.none.raw_rr}{0.675}
\defmmnum{ws1.tied.smol.lstsq.none.ret_hs}{82.7}
\defmmnum{ws1.tied.smol.lstsq.none.ret_piqa}{90.2}
\defmmnum{ws1.tied.smol.lstsq.none.rr}{0.675}
\defmmnum{ws1.tied.smol.lstsq.scaled_l2norm.anchors}{45{,}554}
\defmmnum{ws1.tied.smol.lstsq.scaled_l2norm.cos}{0.6296}
\defmmnum{ws1.tied.smol.lstsq.scaled_l2norm.hs}{49.70}
\defmmnum{ws1.tied.smol.lstsq.scaled_l2norm.p1}{99.08}
\defmmnum{ws1.tied.smol.lstsq.scaled_l2norm.piqa}{74.10}
\defmmnum{ws1.tied.smol.lstsq.scaled_l2norm.raw_rr}{0.675}
\defmmnum{ws1.tied.smol.lstsq.scaled_l2norm.ret_hs}{86.4}
\defmmnum{ws1.tied.smol.lstsq.scaled_l2norm.ret_piqa}{94.1}
\defmmnum{ws1.tied.smol.lstsq.scaled_l2norm.rr}{1.000}
\defmmnum{ws1.tied.smol.procrustes.none.anchors}{45{,}554}
\defmmnum{ws1.tied.smol.procrustes.none.cos}{0.5397}
\defmmnum{ws1.tied.smol.procrustes.none.hs}{26.30}
\defmmnum{ws1.tied.smol.procrustes.none.p1}{98.64}
\defmmnum{ws1.tied.smol.procrustes.none.piqa}{49.78}
\defmmnum{ws1.tied.smol.procrustes.none.raw_rr}{0.060}
\defmmnum{ws1.tied.smol.procrustes.none.ret_hs}{45.7}
\defmmnum{ws1.tied.smol.procrustes.none.ret_piqa}{63.2}
\defmmnum{ws1.tied.smol.procrustes.none.rr}{0.060}
\defmmnum{ws1.tied.smol.procrustes.scaled_l2norm.anchors}{45{,}554}
\defmmnum{ws1.tied.smol.procrustes.scaled_l2norm.cos}{0.5397}
\defmmnum{ws1.tied.smol.procrustes.scaled_l2norm.hs}{24.95}
\defmmnum{ws1.tied.smol.procrustes.scaled_l2norm.p1}{98.64}
\defmmnum{ws1.tied.smol.procrustes.scaled_l2norm.piqa}{51.69}
\defmmnum{ws1.tied.smol.procrustes.scaled_l2norm.raw_rr}{0.060}
\defmmnum{ws1.tied.smol.procrustes.scaled_l2norm.ret_hs}{43.4}
\defmmnum{ws1.tied.smol.procrustes.scaled_l2norm.ret_piqa}{65.7}
\defmmnum{ws1.tied.smol.procrustes.scaled_l2norm.rr}{1.000}
\defmmnum{ws1.total_cells}{25}
\defmmnum{ws1.untied_vs_tied.max_gain_hs}{7.00}
\defmmnum{ws1.untied_vs_tied.max_gain_piqa}{5.06}
\defmmnum{ws1.untied_vs_tied.min_gain_hs}{3.65}
\defmmnum{ws1.untied_vs_tied.min_gain_piqa}{2.72}
\defmmnum{ws1.untied_vs_tied.phi.hs}{+3.65}
\defmmnum{ws1.untied_vs_tied.phi.piqa}{+5.06}
\defmmnum{ws1.untied_vs_tied.qwen.hs}{+3.80}
\defmmnum{ws1.untied_vs_tied.qwen.piqa}{+3.21}
\defmmnum{ws1.untied_vs_tied.smol.hs}{+7.00}
\defmmnum{ws1.untied_vs_tied.smol.piqa}{+2.72}
\defmmnum{ws2.active_runs}{0}
\defmmnum{ws2.batch_size}{16}
\defmmnum{ws2.best.wer_clean}{2.98}
\defmmnum{ws2.best.wer_other}{7.20}
\defmmnum{ws2.best_adamw_run.wer_clean}{3.09}
\defmmnum{ws2.best_adamw_run.wer_other}{7.20}
\defmmnum{ws2.best_adamw_run.wer_pair}{3.09 / 7.20}
\defmmnum{ws2.best_adamw_run.wer_pair_cell}{3.09\% / 7.20\%}
\defmmnum{ws2.best_muon_run.wer_clean}{2.98}
\defmmnum{ws2.best_muon_run.wer_other}{7.31}
\defmmnum{ws2.best_muon_run.wer_pair}{2.98 / 7.31}
\defmmnum{ws2.best_muon_run.wer_pair_cell}{2.98\% / 7.31\%}
\defmmnum{ws2.delta.moonshine.gemmathree.clean}{-0.13}
\defmmnum{ws2.delta.moonshine.gemmathree.other}{-0.47}
\defmmnum{ws2.delta.moonshine.gemmatwo.clean}{-0.13}
\defmmnum{ws2.delta.moonshine.gemmatwo.other}{-0.21}
\defmmnum{ws2.delta.moonshine.qwen.clean}{-0.24}
\defmmnum{ws2.delta.moonshine.qwen.other}{-0.05}
\defmmnum{ws2.delta.whisper.gemmathree.clean}{-0.11}
\defmmnum{ws2.delta.whisper.gemmathree.other}{+0.11}
\defmmnum{ws2.delta.whisper.gemmatwo.clean}{-0.12}
\defmmnum{ws2.delta.whisper.gemmatwo.other}{+0.18}
\defmmnum{ws2.delta.whisper.qwen.clean}{-0.48}
\defmmnum{ws2.delta.whisper.qwen.other}{-0.74}
\defmmnum{ws2.done_runs}{14}
\defmmnum{ws2.moonshine.gemmathree.layernorm.adamw.best_epoch}{4}
\defmmnum{ws2.moonshine.gemmathree.layernorm.adamw.best_val}{0.1130}
\defmmnum{ws2.moonshine.gemmathree.layernorm.adamw.best_val_cell}{0.1130}
\defmmnum{ws2.moonshine.gemmathree.layernorm.adamw.empty_clean}{0}
\defmmnum{ws2.moonshine.gemmathree.layernorm.adamw.empty_other}{0}
\defmmnum{ws2.moonshine.gemmathree.layernorm.adamw.ep1_val}{0.1400}
\defmmnum{ws2.moonshine.gemmathree.layernorm.adamw.ep2_val}{0.1249}
\defmmnum{ws2.moonshine.gemmathree.layernorm.adamw.erank}{365.1}
\defmmnum{ws2.moonshine.gemmathree.layernorm.adamw.erank_pct}{87.8}
\defmmnum{ws2.moonshine.gemmathree.layernorm.adamw.kappa}{17.2}
\defmmnum{ws2.moonshine.gemmathree.layernorm.adamw.live_epoch}{5.00}
\defmmnum{ws2.moonshine.gemmathree.layernorm.adamw.live_val}{0.1130}
\defmmnum{ws2.moonshine.gemmathree.layernorm.adamw.params}{1{,}073{,}472}
\defmmnum{ws2.moonshine.gemmathree.layernorm.adamw.params_m}{1.07M}
\defmmnum{ws2.moonshine.gemmathree.layernorm.adamw.wer_clean}{3.93}
\defmmnum{ws2.moonshine.gemmathree.layernorm.adamw.wer_clean_cell}{3.93\%}
\defmmnum{ws2.moonshine.gemmathree.layernorm.adamw.wer_other}{10.63}
\defmmnum{ws2.moonshine.gemmathree.layernorm.adamw.wer_other_cell}{10.63\%}
\defmmnum{ws2.moonshine.gemmathree.layernorm.adamw.wer_pair}{3.93 / 10.63}
\defmmnum{ws2.moonshine.gemmathree.layernorm.adamw.wer_pair_cell}{3.93\% / 10.63\%}
\defmmnum{ws2.moonshine.gemmathree.layernorm.muon.best_epoch}{3}
\defmmnum{ws2.moonshine.gemmathree.layernorm.muon.best_val}{0.1112}
\defmmnum{ws2.moonshine.gemmathree.layernorm.muon.best_val_cell}{0.1112}
\defmmnum{ws2.moonshine.gemmathree.layernorm.muon.empty_clean}{0}
\defmmnum{ws2.moonshine.gemmathree.layernorm.muon.empty_other}{0}
\defmmnum{ws2.moonshine.gemmathree.layernorm.muon.ep1_val}{0.1262}
\defmmnum{ws2.moonshine.gemmathree.layernorm.muon.ep2_val}{0.1133}
\defmmnum{ws2.moonshine.gemmathree.layernorm.muon.erank}{396.4}
\defmmnum{ws2.moonshine.gemmathree.layernorm.muon.erank_pct}{95.3}
\defmmnum{ws2.moonshine.gemmathree.layernorm.muon.kappa}{45.7}
\defmmnum{ws2.moonshine.gemmathree.layernorm.muon.live_epoch}{5.00}
\defmmnum{ws2.moonshine.gemmathree.layernorm.muon.live_val}{0.1112}
\defmmnum{ws2.moonshine.gemmathree.layernorm.muon.params}{1{,}073{,}472}
\defmmnum{ws2.moonshine.gemmathree.layernorm.muon.params_m}{1.07M}
\defmmnum{ws2.moonshine.gemmathree.layernorm.muon.wer_clean}{3.79}
\defmmnum{ws2.moonshine.gemmathree.layernorm.muon.wer_clean_cell}{3.79\%}
\defmmnum{ws2.moonshine.gemmathree.layernorm.muon.wer_other}{10.16}
\defmmnum{ws2.moonshine.gemmathree.layernorm.muon.wer_other_cell}{10.16\%}
\defmmnum{ws2.moonshine.gemmathree.layernorm.muon.wer_pair}{3.79 / 10.16}
\defmmnum{ws2.moonshine.gemmathree.layernorm.muon.wer_pair_cell}{3.79\% / 10.16\%}
\defmmnum{ws2.moonshine.gemmatwo.layernorm.adamw.best_epoch}{4}
\defmmnum{ws2.moonshine.gemmatwo.layernorm.adamw.best_val}{0.1386}
\defmmnum{ws2.moonshine.gemmatwo.layernorm.adamw.best_val_cell}{0.1386}
\defmmnum{ws2.moonshine.gemmatwo.layernorm.adamw.empty_clean}{0}
\defmmnum{ws2.moonshine.gemmatwo.layernorm.adamw.empty_other}{0}
\defmmnum{ws2.moonshine.gemmatwo.layernorm.adamw.ep1_val}{0.1724}
\defmmnum{ws2.moonshine.gemmatwo.layernorm.adamw.ep1_wer_clean}{5.64}
\defmmnum{ws2.moonshine.gemmatwo.layernorm.adamw.ep1_wer_other}{13.15}
\defmmnum{ws2.moonshine.gemmatwo.layernorm.adamw.ep2_val}{0.1545}
\defmmnum{ws2.moonshine.gemmatwo.layernorm.adamw.ep2_wer_clean}{5.14}
\defmmnum{ws2.moonshine.gemmatwo.layernorm.adamw.ep2_wer_other}{12.56}
\defmmnum{ws2.moonshine.gemmatwo.layernorm.adamw.ep3_wer_clean}{4.59}
\defmmnum{ws2.moonshine.gemmatwo.layernorm.adamw.ep3_wer_other}{11.52}
\defmmnum{ws2.moonshine.gemmatwo.layernorm.adamw.ep4_wer_clean}{4.64}
\defmmnum{ws2.moonshine.gemmatwo.layernorm.adamw.ep4_wer_other}{11.58}
\defmmnum{ws2.moonshine.gemmatwo.layernorm.adamw.ep5_wer_clean}{4.63}
\defmmnum{ws2.moonshine.gemmatwo.layernorm.adamw.ep5_wer_other}{11.94}
\defmmnum{ws2.moonshine.gemmatwo.layernorm.adamw.erank}{372.0}
\defmmnum{ws2.moonshine.gemmatwo.layernorm.adamw.erank_pct}{89.4}
\defmmnum{ws2.moonshine.gemmatwo.layernorm.adamw.kappa}{14.3}
\defmmnum{ws2.moonshine.gemmatwo.layernorm.adamw.live_epoch}{5.00}
\defmmnum{ws2.moonshine.gemmatwo.layernorm.adamw.live_val}{0.1386}
\defmmnum{ws2.moonshine.gemmatwo.layernorm.adamw.params}{966{,}208}
\defmmnum{ws2.moonshine.gemmatwo.layernorm.adamw.params_m}{0.97M}
\defmmnum{ws2.moonshine.gemmatwo.layernorm.adamw.wer_clean}{4.62}
\defmmnum{ws2.moonshine.gemmatwo.layernorm.adamw.wer_clean_cell}{4.62\%}
\defmmnum{ws2.moonshine.gemmatwo.layernorm.adamw.wer_other}{11.69}
\defmmnum{ws2.moonshine.gemmatwo.layernorm.adamw.wer_other_cell}{11.69\%}
\defmmnum{ws2.moonshine.gemmatwo.layernorm.adamw.wer_pair}{4.62 / 11.69}
\defmmnum{ws2.moonshine.gemmatwo.layernorm.adamw.wer_pair_cell}{4.62\% / 11.69\%}
\defmmnum{ws2.moonshine.gemmatwo.layernorm.muon.best_epoch}{3}
\defmmnum{ws2.moonshine.gemmatwo.layernorm.muon.best_val}{0.1343}
\defmmnum{ws2.moonshine.gemmatwo.layernorm.muon.best_val_cell}{0.1343}
\defmmnum{ws2.moonshine.gemmatwo.layernorm.muon.empty_clean}{0}
\defmmnum{ws2.moonshine.gemmatwo.layernorm.muon.empty_other}{0}
\defmmnum{ws2.moonshine.gemmatwo.layernorm.muon.ep1_val}{0.1553}
\defmmnum{ws2.moonshine.gemmatwo.layernorm.muon.ep1_wer_clean}{5.28}
\defmmnum{ws2.moonshine.gemmatwo.layernorm.muon.ep1_wer_other}{12.30}
\defmmnum{ws2.moonshine.gemmatwo.layernorm.muon.ep2_val}{0.1431}
\defmmnum{ws2.moonshine.gemmatwo.layernorm.muon.ep2_wer_clean}{4.74}
\defmmnum{ws2.moonshine.gemmatwo.layernorm.muon.ep2_wer_other}{11.81}
\defmmnum{ws2.moonshine.gemmatwo.layernorm.muon.ep3_wer_clean}{4.49}
\defmmnum{ws2.moonshine.gemmatwo.layernorm.muon.ep3_wer_other}{11.48}
\defmmnum{ws2.moonshine.gemmatwo.layernorm.muon.ep4_wer_clean}{4.47}
\defmmnum{ws2.moonshine.gemmatwo.layernorm.muon.ep4_wer_other}{11.33}
\defmmnum{ws2.moonshine.gemmatwo.layernorm.muon.ep5_wer_clean}{4.53}
\defmmnum{ws2.moonshine.gemmatwo.layernorm.muon.ep5_wer_other}{11.53}
\defmmnum{ws2.moonshine.gemmatwo.layernorm.muon.erank}{393.0}
\defmmnum{ws2.moonshine.gemmatwo.layernorm.muon.erank_pct}{94.5}
\defmmnum{ws2.moonshine.gemmatwo.layernorm.muon.kappa}{43.8}
\defmmnum{ws2.moonshine.gemmatwo.layernorm.muon.live_epoch}{5.00}
\defmmnum{ws2.moonshine.gemmatwo.layernorm.muon.live_val}{0.1343}
\defmmnum{ws2.moonshine.gemmatwo.layernorm.muon.params}{966{,}208}
\defmmnum{ws2.moonshine.gemmatwo.layernorm.muon.params_m}{0.97M}
\defmmnum{ws2.moonshine.gemmatwo.layernorm.muon.wer_clean}{4.49}
\defmmnum{ws2.moonshine.gemmatwo.layernorm.muon.wer_clean_cell}{4.49\%}
\defmmnum{ws2.moonshine.gemmatwo.layernorm.muon.wer_other}{11.48}
\defmmnum{ws2.moonshine.gemmatwo.layernorm.muon.wer_other_cell}{11.48\%}
\defmmnum{ws2.moonshine.gemmatwo.layernorm.muon.wer_pair}{4.49 / 11.48}
\defmmnum{ws2.moonshine.gemmatwo.layernorm.muon.wer_pair_cell}{4.49\% / 11.48\%}
\defmmnum{ws2.moonshine.qwen.layernorm.adamw.best_epoch}{5}
\defmmnum{ws2.moonshine.qwen.layernorm.adamw.best_val}{0.2297}
\defmmnum{ws2.moonshine.qwen.layernorm.adamw.best_val_cell}{0.2297}
\defmmnum{ws2.moonshine.qwen.layernorm.adamw.empty_clean}{0}
\defmmnum{ws2.moonshine.qwen.layernorm.adamw.empty_other}{0}
\defmmnum{ws2.moonshine.qwen.layernorm.adamw.ep1_val}{0.3291}
\defmmnum{ws2.moonshine.qwen.layernorm.adamw.ep1_wer_clean}{10.42}
\defmmnum{ws2.moonshine.qwen.layernorm.adamw.ep1_wer_other}{20.60}
\defmmnum{ws2.moonshine.qwen.layernorm.adamw.ep2_val}{0.2702}
\defmmnum{ws2.moonshine.qwen.layernorm.adamw.ep2_wer_clean}{8.51}
\defmmnum{ws2.moonshine.qwen.layernorm.adamw.ep2_wer_other}{17.91}
\defmmnum{ws2.moonshine.qwen.layernorm.adamw.ep3_wer_clean}{7.73}
\defmmnum{ws2.moonshine.qwen.layernorm.adamw.ep3_wer_other}{17.04}
\defmmnum{ws2.moonshine.qwen.layernorm.adamw.ep4_wer_clean}{7.51}
\defmmnum{ws2.moonshine.qwen.layernorm.adamw.ep4_wer_other}{16.86}
\defmmnum{ws2.moonshine.qwen.layernorm.adamw.ep5_wer_clean}{7.37}
\defmmnum{ws2.moonshine.qwen.layernorm.adamw.ep5_wer_other}{16.41}
\defmmnum{ws2.moonshine.qwen.layernorm.adamw.erank}{359.6}
\defmmnum{ws2.moonshine.qwen.layernorm.adamw.erank_pct}{86.4}
\defmmnum{ws2.moonshine.qwen.layernorm.adamw.kappa}{19.7}
\defmmnum{ws2.moonshine.qwen.layernorm.adamw.live_epoch}{5.00}
\defmmnum{ws2.moonshine.qwen.layernorm.adamw.live_val}{0.2297}
\defmmnum{ws2.moonshine.qwen.layernorm.adamw.params}{644{,}416}
\defmmnum{ws2.moonshine.qwen.layernorm.adamw.params_m}{0.64M}
\defmmnum{ws2.moonshine.qwen.layernorm.adamw.wer_clean}{7.38}
\defmmnum{ws2.moonshine.qwen.layernorm.adamw.wer_clean_cell}{7.38\%}
\defmmnum{ws2.moonshine.qwen.layernorm.adamw.wer_other}{16.44}
\defmmnum{ws2.moonshine.qwen.layernorm.adamw.wer_other_cell}{16.44\%}
\defmmnum{ws2.moonshine.qwen.layernorm.adamw.wer_pair}{7.38 / 16.44}
\defmmnum{ws2.moonshine.qwen.layernorm.adamw.wer_pair_cell}{7.38\% / 16.44\%}
\defmmnum{ws2.moonshine.qwen.layernorm.muon.best_epoch}{4}
\defmmnum{ws2.moonshine.qwen.layernorm.muon.best_val}{0.2258}
\defmmnum{ws2.moonshine.qwen.layernorm.muon.best_val_cell}{0.2258}
\defmmnum{ws2.moonshine.qwen.layernorm.muon.empty_clean}{0}
\defmmnum{ws2.moonshine.qwen.layernorm.muon.empty_other}{0}
\defmmnum{ws2.moonshine.qwen.layernorm.muon.ep1_val}{0.3058}
\defmmnum{ws2.moonshine.qwen.layernorm.muon.ep1_wer_clean}{9.61}
\defmmnum{ws2.moonshine.qwen.layernorm.muon.ep1_wer_other}{19.63}
\defmmnum{ws2.moonshine.qwen.layernorm.muon.ep2_val}{0.2494}
\defmmnum{ws2.moonshine.qwen.layernorm.muon.ep2_wer_clean}{7.84}
\defmmnum{ws2.moonshine.qwen.layernorm.muon.ep2_wer_other}{17.31}
\defmmnum{ws2.moonshine.qwen.layernorm.muon.ep3_wer_clean}{7.33}
\defmmnum{ws2.moonshine.qwen.layernorm.muon.ep3_wer_other}{16.42}
\defmmnum{ws2.moonshine.qwen.layernorm.muon.ep4_wer_clean}{7.15}
\defmmnum{ws2.moonshine.qwen.layernorm.muon.ep4_wer_other}{16.38}
\defmmnum{ws2.moonshine.qwen.layernorm.muon.ep5_wer_clean}{7.12}
\defmmnum{ws2.moonshine.qwen.layernorm.muon.ep5_wer_other}{16.51}
\defmmnum{ws2.moonshine.qwen.layernorm.muon.erank}{380.3}
\defmmnum{ws2.moonshine.qwen.layernorm.muon.erank_pct}{91.4}
\defmmnum{ws2.moonshine.qwen.layernorm.muon.kappa}{63.7}
\defmmnum{ws2.moonshine.qwen.layernorm.muon.live_epoch}{5.00}
\defmmnum{ws2.moonshine.qwen.layernorm.muon.live_val}{0.2258}
\defmmnum{ws2.moonshine.qwen.layernorm.muon.params}{644{,}416}
\defmmnum{ws2.moonshine.qwen.layernorm.muon.params_m}{0.64M}
\defmmnum{ws2.moonshine.qwen.layernorm.muon.wer_clean}{7.14}
\defmmnum{ws2.moonshine.qwen.layernorm.muon.wer_clean_cell}{7.14\%}
\defmmnum{ws2.moonshine.qwen.layernorm.muon.wer_other}{16.39}
\defmmnum{ws2.moonshine.qwen.layernorm.muon.wer_other_cell}{16.39\%}
\defmmnum{ws2.moonshine.qwen.layernorm.muon.wer_pair}{7.14 / 16.39}
\defmmnum{ws2.moonshine.qwen.layernorm.muon.wer_pair_cell}{7.14\% / 16.39\%}
\defmmnum{ws2.muon_adamw.delta_wer}{-0.20}
\defmmnum{ws2.muon_higher_clean}{0}
\defmmnum{ws2.muon_higher_other}{2}
\defmmnum{ws2.muon_live_val_pairs}{6}
\defmmnum{ws2.muon_live_val_wins}{5}
\defmmnum{ws2.muon_lower_clean}{6}
\defmmnum{ws2.muon_lower_other}{3}
\defmmnum{ws2.muon_tied_clean}{0}
\defmmnum{ws2.muon_tied_other}{1}
\defmmnum{ws2.muon_wins_clean}{6}
\defmmnum{ws2.muon_wins_other}{4}
\defmmnum{ws2.n_paired_cells}{6}
\defmmnum{ws2.n_runs}{14}
\defmmnum{ws2.native_ceiling.moonshine.clean}{3.62}
\defmmnum{ws2.native_ceiling.moonshine.other}{8.56}
\defmmnum{ws2.native_ceiling.whisper.clean}{4.05}
\defmmnum{ws2.native_ceiling.whisper.other}{8.36}
\defmmnum{ws2.norm_delta.whisper.gemmatwo.l2_vs_ln.muon.clean}{+4.06}
\defmmnum{ws2.norm_delta.whisper.gemmatwo.l2_vs_ln.muon.clean_exact}{+4.0608}
\defmmnum{ws2.norm_delta.whisper.gemmatwo.l2_vs_ln.muon.other}{+4.74}
\defmmnum{ws2.norm_delta.whisper.gemmatwo.rms_vs_ln.muon.clean}{+0.00}
\defmmnum{ws2.norm_delta.whisper.gemmatwo.rms_vs_ln.muon.clean_exact}{+0.0000}
\defmmnum{ws2.norm_delta.whisper.gemmatwo.rms_vs_ln.muon.other}{-0.10}
\defmmnum{ws2.norm_delta.whisper.gemmatwo.scaled_vs_ln.muon.clean}{+0.03}
\defmmnum{ws2.norm_delta.whisper.gemmatwo.scaled_vs_ln.muon.clean_exact}{+0.0323}
\defmmnum{ws2.norm_delta.whisper.gemmatwo.scaled_vs_ln.muon.other}{+0.08}
\defmmnum{ws2.norm_delta.whisper.qwen.l2_vs_ln.muon.clean}{+1.28}
\defmmnum{ws2.norm_delta.whisper.qwen.l2_vs_ln.muon.clean_exact}{+1.2839}
\defmmnum{ws2.norm_delta.whisper.qwen.l2_vs_ln.muon.other}{+1.83}
\defmmnum{ws2.norm_delta.whisper.qwen.scaled_vs_ln.muon.clean}{-0.02}
\defmmnum{ws2.norm_delta.whisper.qwen.scaled_vs_ln.muon.clean_exact}{-0.0228}
\defmmnum{ws2.norm_delta.whisper.qwen.scaled_vs_ln.muon.other}{-0.06}
\defmmnum{ws2.norm_ladder.gemmatwo.done}{5}
\defmmnum{ws2.norm_ladder.gemmatwo.total}{5}
\defmmnum{ws2.norm_ladder.qwen.best_clean}{4.30}
\defmmnum{ws2.norm_ladder.qwen.best_other}{9.90}
\defmmnum{ws2.norm_ladder.qwen.done}{5}
\defmmnum{ws2.norm_ladder.qwen.total}{5}
\defmmnum{ws2.paired_ln_empty_hyps}{0}
\defmmnum{ws2.paired_ln_eval_runs}{12}
\defmmnum{ws2.pairs_done}{6}
\defmmnum{ws2.steps_per_epoch}{17{,}578}
\defmmnum{ws2.svd.adamw_max_erank_pct}{89.4}
\defmmnum{ws2.svd.adamw_min_erank_pct}{81.7}
\defmmnum{ws2.svd.moonshine_max_kappa_ratio}{3.2}
\defmmnum{ws2.svd.moonshine_min_kappa_ratio}{2.7}
\defmmnum{ws2.svd.muon_max_erank_pct}{95.3}
\defmmnum{ws2.svd.muon_min_erank_pct}{90.1}
\defmmnum{ws2.svd.whisper_muon_lower_kappa}{3}
\defmmnum{ws2.svd.whisper_total_pairs}{3}
\defmmnum{ws2.total_empty_hyps}{1}
\defmmnum{ws2.total_eval_runs}{20}
\defmmnum{ws2.total_steps}{87{,}890}
\defmmnum{ws2.train_done_runs}{14}
\defmmnum{ws2.uniform_epochs}{5}
\defmmnum{ws2.val_interval_steps}{2{,}500}
\defmmnum{ws2.whisper.gemmathree.layernorm.adamw.best_epoch}{3}
\defmmnum{ws2.whisper.gemmathree.layernorm.adamw.best_val}{0.0845}
\defmmnum{ws2.whisper.gemmathree.layernorm.adamw.best_val_cell}{0.0845}
\defmmnum{ws2.whisper.gemmathree.layernorm.adamw.empty_clean}{0}
\defmmnum{ws2.whisper.gemmathree.layernorm.adamw.empty_other}{0}
\defmmnum{ws2.whisper.gemmathree.layernorm.adamw.ep1_val}{0.0989}
\defmmnum{ws2.whisper.gemmathree.layernorm.adamw.ep2_val}{0.0899}
\defmmnum{ws2.whisper.gemmathree.layernorm.adamw.erank}{669.2}
\defmmnum{ws2.whisper.gemmathree.layernorm.adamw.erank_pct}{87.1}
\defmmnum{ws2.whisper.gemmathree.layernorm.adamw.kappa}{19.1}
\defmmnum{ws2.whisper.gemmathree.layernorm.adamw.live_epoch}{5.00}
\defmmnum{ws2.whisper.gemmathree.layernorm.adamw.live_val}{0.0845}
\defmmnum{ws2.whisper.gemmathree.layernorm.adamw.params}{1{,}975{,}296}
\defmmnum{ws2.whisper.gemmathree.layernorm.adamw.params_m}{1.98M}
\defmmnum{ws2.whisper.gemmathree.layernorm.adamw.wer_clean}{3.09}
\defmmnum{ws2.whisper.gemmathree.layernorm.adamw.wer_clean_cell}{3.09\%}
\defmmnum{ws2.whisper.gemmathree.layernorm.adamw.wer_other}{7.20}
\defmmnum{ws2.whisper.gemmathree.layernorm.adamw.wer_other_cell}{7.20\%}
\defmmnum{ws2.whisper.gemmathree.layernorm.adamw.wer_pair}{3.09 / 7.20}
\defmmnum{ws2.whisper.gemmathree.layernorm.adamw.wer_pair_cell}{3.09\% / 7.20\%}
\defmmnum{ws2.whisper.gemmathree.layernorm.muon.best_epoch}{2}
\defmmnum{ws2.whisper.gemmathree.layernorm.muon.best_val}{0.0847}
\defmmnum{ws2.whisper.gemmathree.layernorm.muon.best_val_cell}{0.0847}
\defmmnum{ws2.whisper.gemmathree.layernorm.muon.empty_clean}{0}
\defmmnum{ws2.whisper.gemmathree.layernorm.muon.empty_other}{0}
\defmmnum{ws2.whisper.gemmathree.layernorm.muon.ep1_val}{0.0915}
\defmmnum{ws2.whisper.gemmathree.layernorm.muon.ep2_val}{0.0847}
\defmmnum{ws2.whisper.gemmathree.layernorm.muon.erank}{725.2}
\defmmnum{ws2.whisper.gemmathree.layernorm.muon.erank_pct}{94.4}
\defmmnum{ws2.whisper.gemmathree.layernorm.muon.kappa}{7.8}
\defmmnum{ws2.whisper.gemmathree.layernorm.muon.live_epoch}{5.00}
\defmmnum{ws2.whisper.gemmathree.layernorm.muon.live_val}{0.0847}
\defmmnum{ws2.whisper.gemmathree.layernorm.muon.params}{1{,}975{,}296}
\defmmnum{ws2.whisper.gemmathree.layernorm.muon.params_m}{1.98M}
\defmmnum{ws2.whisper.gemmathree.layernorm.muon.wer_clean}{2.98}
\defmmnum{ws2.whisper.gemmathree.layernorm.muon.wer_clean_cell}{2.98\%}
\defmmnum{ws2.whisper.gemmathree.layernorm.muon.wer_other}{7.31}
\defmmnum{ws2.whisper.gemmathree.layernorm.muon.wer_other_cell}{7.31\%}
\defmmnum{ws2.whisper.gemmathree.layernorm.muon.wer_pair}{2.98 / 7.31}
\defmmnum{ws2.whisper.gemmathree.layernorm.muon.wer_pair_cell}{2.98\% / 7.31\%}
\defmmnum{ws2.whisper.gemmatwo.l2norm.muon.best_epoch}{5}
\defmmnum{ws2.whisper.gemmatwo.l2norm.muon.best_val}{0.1594}
\defmmnum{ws2.whisper.gemmatwo.l2norm.muon.best_val_cell}{0.1594}
\defmmnum{ws2.whisper.gemmatwo.l2norm.muon.empty_clean}{1}
\defmmnum{ws2.whisper.gemmatwo.l2norm.muon.empty_other}{0}
\defmmnum{ws2.whisper.gemmatwo.l2norm.muon.ep1_val}{3.4658}
\defmmnum{ws2.whisper.gemmatwo.l2norm.muon.ep2_val}{0.3275}
\defmmnum{ws2.whisper.gemmatwo.l2norm.muon.erank}{662.8}
\defmmnum{ws2.whisper.gemmatwo.l2norm.muon.erank_pct}{86.3}
\defmmnum{ws2.whisper.gemmatwo.l2norm.muon.kappa}{75.1}
\defmmnum{ws2.whisper.gemmatwo.l2norm.muon.live_epoch}{5.00}
\defmmnum{ws2.whisper.gemmatwo.l2norm.muon.live_val}{0.1594}
\defmmnum{ws2.whisper.gemmatwo.l2norm.muon.params}{1{,}771{,}776}
\defmmnum{ws2.whisper.gemmatwo.l2norm.muon.params_m}{1.77M}
\defmmnum{ws2.whisper.gemmatwo.l2norm.muon.wer_clean}{7.30}
\defmmnum{ws2.whisper.gemmatwo.l2norm.muon.wer_clean_cell}{7.30\%}
\defmmnum{ws2.whisper.gemmatwo.l2norm.muon.wer_other}{12.76}
\defmmnum{ws2.whisper.gemmatwo.l2norm.muon.wer_other_cell}{12.76\%}
\defmmnum{ws2.whisper.gemmatwo.l2norm.muon.wer_pair}{7.30 / 12.76}
\defmmnum{ws2.whisper.gemmatwo.l2norm.muon.wer_pair_cell}{7.30\% / 12.76\%}
\defmmnum{ws2.whisper.gemmatwo.layernorm.adamw.best_epoch}{3}
\defmmnum{ws2.whisper.gemmatwo.layernorm.adamw.best_val}{0.0974}
\defmmnum{ws2.whisper.gemmatwo.layernorm.adamw.best_val_cell}{0.0974}
\defmmnum{ws2.whisper.gemmatwo.layernorm.adamw.empty_clean}{0}
\defmmnum{ws2.whisper.gemmatwo.layernorm.adamw.empty_other}{0}
\defmmnum{ws2.whisper.gemmatwo.layernorm.adamw.ep1_val}{0.1159}
\defmmnum{ws2.whisper.gemmatwo.layernorm.adamw.ep2_val}{0.1010}
\defmmnum{ws2.whisper.gemmatwo.layernorm.adamw.erank}{678.0}
\defmmnum{ws2.whisper.gemmatwo.layernorm.adamw.erank_pct}{88.3}
\defmmnum{ws2.whisper.gemmatwo.layernorm.adamw.kappa}{15.3}
\defmmnum{ws2.whisper.gemmatwo.layernorm.adamw.live_epoch}{5.00}
\defmmnum{ws2.whisper.gemmatwo.layernorm.adamw.live_val}{0.0974}
\defmmnum{ws2.whisper.gemmatwo.layernorm.adamw.params}{1{,}777{,}920}
\defmmnum{ws2.whisper.gemmatwo.layernorm.adamw.params_m}{1.78M}
\defmmnum{ws2.whisper.gemmatwo.layernorm.adamw.wer_clean}{3.36}
\defmmnum{ws2.whisper.gemmatwo.layernorm.adamw.wer_clean_cell}{3.36\%}
\defmmnum{ws2.whisper.gemmatwo.layernorm.adamw.wer_other}{7.84}
\defmmnum{ws2.whisper.gemmatwo.layernorm.adamw.wer_other_cell}{7.84\%}
\defmmnum{ws2.whisper.gemmatwo.layernorm.adamw.wer_pair}{3.36 / 7.84}
\defmmnum{ws2.whisper.gemmatwo.layernorm.adamw.wer_pair_cell}{3.36\% / 7.84\%}
\defmmnum{ws2.whisper.gemmatwo.layernorm.muon.best_epoch}{2}
\defmmnum{ws2.whisper.gemmatwo.layernorm.muon.best_val}{0.0958}
\defmmnum{ws2.whisper.gemmatwo.layernorm.muon.best_val_cell}{0.0958}
\defmmnum{ws2.whisper.gemmatwo.layernorm.muon.empty_clean}{0}
\defmmnum{ws2.whisper.gemmatwo.layernorm.muon.empty_other}{0}
\defmmnum{ws2.whisper.gemmatwo.layernorm.muon.ep1_val}{0.1042}
\defmmnum{ws2.whisper.gemmatwo.layernorm.muon.ep2_val}{0.0960}
\defmmnum{ws2.whisper.gemmatwo.layernorm.muon.erank}{722.2}
\defmmnum{ws2.whisper.gemmatwo.layernorm.muon.erank_pct}{94.0}
\defmmnum{ws2.whisper.gemmatwo.layernorm.muon.kappa}{8.1}
\defmmnum{ws2.whisper.gemmatwo.layernorm.muon.live_epoch}{5.00}
\defmmnum{ws2.whisper.gemmatwo.layernorm.muon.live_val}{0.0958}
\defmmnum{ws2.whisper.gemmatwo.layernorm.muon.params}{1{,}777{,}920}
\defmmnum{ws2.whisper.gemmatwo.layernorm.muon.params_m}{1.78M}
\defmmnum{ws2.whisper.gemmatwo.layernorm.muon.wer_clean}{3.24}
\defmmnum{ws2.whisper.gemmatwo.layernorm.muon.wer_clean_cell}{3.24\%}
\defmmnum{ws2.whisper.gemmatwo.layernorm.muon.wer_other}{8.02}
\defmmnum{ws2.whisper.gemmatwo.layernorm.muon.wer_other_cell}{8.02\%}
\defmmnum{ws2.whisper.gemmatwo.layernorm.muon.wer_pair}{3.24 / 8.02}
\defmmnum{ws2.whisper.gemmatwo.layernorm.muon.wer_pair_cell}{3.24\% / 8.02\%}
\defmmnum{ws2.whisper.gemmatwo.none.muon.best_epoch}{3}
\defmmnum{ws2.whisper.gemmatwo.none.muon.best_val}{0.0983}
\defmmnum{ws2.whisper.gemmatwo.none.muon.best_val_cell}{0.0983}
\defmmnum{ws2.whisper.gemmatwo.none.muon.empty_clean}{0}
\defmmnum{ws2.whisper.gemmatwo.none.muon.empty_other}{0}
\defmmnum{ws2.whisper.gemmatwo.none.muon.ep1_val}{0.1149}
\defmmnum{ws2.whisper.gemmatwo.none.muon.ep2_val}{0.1018}
\defmmnum{ws2.whisper.gemmatwo.none.muon.erank}{713.9}
\defmmnum{ws2.whisper.gemmatwo.none.muon.erank_pct}{93.0}
\defmmnum{ws2.whisper.gemmatwo.none.muon.kappa}{23.0}
\defmmnum{ws2.whisper.gemmatwo.none.muon.live_epoch}{5.00}
\defmmnum{ws2.whisper.gemmatwo.none.muon.live_val}{0.0983}
\defmmnum{ws2.whisper.gemmatwo.none.muon.params}{1{,}771{,}776}
\defmmnum{ws2.whisper.gemmatwo.none.muon.params_m}{1.77M}
\defmmnum{ws2.whisper.gemmatwo.none.muon.wer_clean}{4.16}
\defmmnum{ws2.whisper.gemmatwo.none.muon.wer_clean_cell}{4.16\%}
\defmmnum{ws2.whisper.gemmatwo.none.muon.wer_other}{9.06}
\defmmnum{ws2.whisper.gemmatwo.none.muon.wer_other_cell}{9.06\%}
\defmmnum{ws2.whisper.gemmatwo.none.muon.wer_pair}{4.16 / 9.06}
\defmmnum{ws2.whisper.gemmatwo.none.muon.wer_pair_cell}{4.16\% / 9.06\%}
\defmmnum{ws2.whisper.gemmatwo.rmsnorm.muon.best_epoch}{2}
\defmmnum{ws2.whisper.gemmatwo.rmsnorm.muon.best_val}{0.0944}
\defmmnum{ws2.whisper.gemmatwo.rmsnorm.muon.best_val_cell}{0.0944}
\defmmnum{ws2.whisper.gemmatwo.rmsnorm.muon.empty_clean}{0}
\defmmnum{ws2.whisper.gemmatwo.rmsnorm.muon.empty_other}{0}
\defmmnum{ws2.whisper.gemmatwo.rmsnorm.muon.ep1_val}{0.1031}
\defmmnum{ws2.whisper.gemmatwo.rmsnorm.muon.ep2_val}{0.0949}
\defmmnum{ws2.whisper.gemmatwo.rmsnorm.muon.erank}{722.2}
\defmmnum{ws2.whisper.gemmatwo.rmsnorm.muon.erank_pct}{94.0}
\defmmnum{ws2.whisper.gemmatwo.rmsnorm.muon.kappa}{10.2}
\defmmnum{ws2.whisper.gemmatwo.rmsnorm.muon.live_epoch}{5.00}
\defmmnum{ws2.whisper.gemmatwo.rmsnorm.muon.live_val}{0.0944}
\defmmnum{ws2.whisper.gemmatwo.rmsnorm.muon.params}{1{,}774{,}848}
\defmmnum{ws2.whisper.gemmatwo.rmsnorm.muon.params_m}{1.77M}
\defmmnum{ws2.whisper.gemmatwo.rmsnorm.muon.wer_clean}{3.24}
\defmmnum{ws2.whisper.gemmatwo.rmsnorm.muon.wer_clean_cell}{3.24\%}
\defmmnum{ws2.whisper.gemmatwo.rmsnorm.muon.wer_other}{7.93}
\defmmnum{ws2.whisper.gemmatwo.rmsnorm.muon.wer_other_cell}{7.93\%}
\defmmnum{ws2.whisper.gemmatwo.rmsnorm.muon.wer_pair}{3.24 / 7.93}
\defmmnum{ws2.whisper.gemmatwo.rmsnorm.muon.wer_pair_cell}{3.24\% / 7.93\%}
\defmmnum{ws2.whisper.gemmatwo.scaled_l2norm.muon.best_epoch}{2}
\defmmnum{ws2.whisper.gemmatwo.scaled_l2norm.muon.best_val}{0.0974}
\defmmnum{ws2.whisper.gemmatwo.scaled_l2norm.muon.best_val_cell}{0.0974}
\defmmnum{ws2.whisper.gemmatwo.scaled_l2norm.muon.empty_clean}{0}
\defmmnum{ws2.whisper.gemmatwo.scaled_l2norm.muon.empty_other}{0}
\defmmnum{ws2.whisper.gemmatwo.scaled_l2norm.muon.ep1_val}{0.1083}
\defmmnum{ws2.whisper.gemmatwo.scaled_l2norm.muon.ep2_val}{0.0971}
\defmmnum{ws2.whisper.gemmatwo.scaled_l2norm.muon.erank}{718.2}
\defmmnum{ws2.whisper.gemmatwo.scaled_l2norm.muon.erank_pct}{93.5}
\defmmnum{ws2.whisper.gemmatwo.scaled_l2norm.muon.kappa}{25.2}
\defmmnum{ws2.whisper.gemmatwo.scaled_l2norm.muon.live_epoch}{5.00}
\defmmnum{ws2.whisper.gemmatwo.scaled_l2norm.muon.live_val}{0.0974}
\defmmnum{ws2.whisper.gemmatwo.scaled_l2norm.muon.params}{1{,}771{,}778}
\defmmnum{ws2.whisper.gemmatwo.scaled_l2norm.muon.params_m}{1.77M}
\defmmnum{ws2.whisper.gemmatwo.scaled_l2norm.muon.wer_clean}{3.27}
\defmmnum{ws2.whisper.gemmatwo.scaled_l2norm.muon.wer_clean_cell}{3.27\%}
\defmmnum{ws2.whisper.gemmatwo.scaled_l2norm.muon.wer_other}{8.11}
\defmmnum{ws2.whisper.gemmatwo.scaled_l2norm.muon.wer_other_cell}{8.11\%}
\defmmnum{ws2.whisper.gemmatwo.scaled_l2norm.muon.wer_pair}{3.27 / 8.11}
\defmmnum{ws2.whisper.gemmatwo.scaled_l2norm.muon.wer_pair_cell}{3.27\% / 8.11\%}
\defmmnum{ws2.whisper.qwen.l2norm.muon.best_epoch}{4}
\defmmnum{ws2.whisper.qwen.l2norm.muon.best_val}{0.1840}
\defmmnum{ws2.whisper.qwen.l2norm.muon.best_val_cell}{0.1840}
\defmmnum{ws2.whisper.qwen.l2norm.muon.empty_clean}{0}
\defmmnum{ws2.whisper.qwen.l2norm.muon.empty_other}{0}
\defmmnum{ws2.whisper.qwen.l2norm.muon.ep1_val}{0.2160}
\defmmnum{ws2.whisper.qwen.l2norm.muon.ep2_val}{0.1977}
\defmmnum{ws2.whisper.qwen.l2norm.muon.erank}{630.7}
\defmmnum{ws2.whisper.qwen.l2norm.muon.erank_pct}{82.1}
\defmmnum{ws2.whisper.qwen.l2norm.muon.kappa}{105.2}
\defmmnum{ws2.whisper.qwen.l2norm.muon.live_epoch}{5.00}
\defmmnum{ws2.whisper.qwen.l2norm.muon.live_val}{0.1840}
\defmmnum{ws2.whisper.qwen.l2norm.muon.params}{1{,}181{,}184}
\defmmnum{ws2.whisper.qwen.l2norm.muon.params_m}{1.18M}
\defmmnum{ws2.whisper.qwen.l2norm.muon.wer_clean}{5.61}
\defmmnum{ws2.whisper.qwen.l2norm.muon.wer_clean_cell}{5.61\%}
\defmmnum{ws2.whisper.qwen.l2norm.muon.wer_other}{11.80}
\defmmnum{ws2.whisper.qwen.l2norm.muon.wer_other_cell}{11.80\%}
\defmmnum{ws2.whisper.qwen.l2norm.muon.wer_pair}{5.61 / 11.80}
\defmmnum{ws2.whisper.qwen.l2norm.muon.wer_pair_cell}{5.61\% / 11.80\%}
\defmmnum{ws2.whisper.qwen.layernorm.adamw.best_epoch}{5}
\defmmnum{ws2.whisper.qwen.layernorm.adamw.best_val}{0.1477}
\defmmnum{ws2.whisper.qwen.layernorm.adamw.best_val_cell}{0.1477}
\defmmnum{ws2.whisper.qwen.layernorm.adamw.empty_clean}{0}
\defmmnum{ws2.whisper.qwen.layernorm.adamw.empty_other}{0}
\defmmnum{ws2.whisper.qwen.layernorm.adamw.ep1_val}{0.2233}
\defmmnum{ws2.whisper.qwen.layernorm.adamw.ep1_wer_clean}{6.88}
\defmmnum{ws2.whisper.qwen.layernorm.adamw.ep1_wer_other}{13.26}
\defmmnum{ws2.whisper.qwen.layernorm.adamw.ep2_val}{0.1848}
\defmmnum{ws2.whisper.qwen.layernorm.adamw.ep2_wer_clean}{5.71}
\defmmnum{ws2.whisper.qwen.layernorm.adamw.ep2_wer_other}{12.12}
\defmmnum{ws2.whisper.qwen.layernorm.adamw.ep3_wer_clean}{5.36}
\defmmnum{ws2.whisper.qwen.layernorm.adamw.ep3_wer_other}{11.29}
\defmmnum{ws2.whisper.qwen.layernorm.adamw.ep4_wer_clean}{5.03}
\defmmnum{ws2.whisper.qwen.layernorm.adamw.ep4_wer_other}{10.87}
\defmmnum{ws2.whisper.qwen.layernorm.adamw.ep5_wer_clean}{4.77}
\defmmnum{ws2.whisper.qwen.layernorm.adamw.ep5_wer_other}{10.68}
\defmmnum{ws2.whisper.qwen.layernorm.adamw.erank}{627.5}
\defmmnum{ws2.whisper.qwen.layernorm.adamw.erank_pct}{81.7}
\defmmnum{ws2.whisper.qwen.layernorm.adamw.kappa}{28.4}
\defmmnum{ws2.whisper.qwen.layernorm.adamw.live_epoch}{5.00}
\defmmnum{ws2.whisper.qwen.layernorm.adamw.live_val}{0.1477}
\defmmnum{ws2.whisper.qwen.layernorm.adamw.params}{1{,}185{,}792}
\defmmnum{ws2.whisper.qwen.layernorm.adamw.params_m}{1.19M}
\defmmnum{ws2.whisper.qwen.layernorm.adamw.wer_clean}{4.81}
\defmmnum{ws2.whisper.qwen.layernorm.adamw.wer_clean_cell}{4.81\%}
\defmmnum{ws2.whisper.qwen.layernorm.adamw.wer_other}{10.70}
\defmmnum{ws2.whisper.qwen.layernorm.adamw.wer_other_cell}{10.70\%}
\defmmnum{ws2.whisper.qwen.layernorm.adamw.wer_pair}{4.81 / 10.70}
\defmmnum{ws2.whisper.qwen.layernorm.adamw.wer_pair_cell}{4.81\% / 10.70\%}
\defmmnum{ws2.whisper.qwen.layernorm.muon.best_epoch}{4}
\defmmnum{ws2.whisper.qwen.layernorm.muon.best_val}{0.1394}
\defmmnum{ws2.whisper.qwen.layernorm.muon.best_val_cell}{0.1394}
\defmmnum{ws2.whisper.qwen.layernorm.muon.empty_clean}{0}
\defmmnum{ws2.whisper.qwen.layernorm.muon.empty_other}{0}
\defmmnum{ws2.whisper.qwen.layernorm.muon.ep1_val}{0.1861}
\defmmnum{ws2.whisper.qwen.layernorm.muon.ep1_wer_clean}{5.64}
\defmmnum{ws2.whisper.qwen.layernorm.muon.ep1_wer_other}{11.95}
\defmmnum{ws2.whisper.qwen.layernorm.muon.ep2_val}{0.1516}
\defmmnum{ws2.whisper.qwen.layernorm.muon.ep2_wer_clean}{4.67}
\defmmnum{ws2.whisper.qwen.layernorm.muon.ep2_wer_other}{10.60}
\defmmnum{ws2.whisper.qwen.layernorm.muon.ep3_wer_clean}{4.37}
\defmmnum{ws2.whisper.qwen.layernorm.muon.ep3_wer_other}{10.25}
\defmmnum{ws2.whisper.qwen.layernorm.muon.ep4_wer_clean}{4.35}
\defmmnum{ws2.whisper.qwen.layernorm.muon.ep4_wer_other}{9.98}
\defmmnum{ws2.whisper.qwen.layernorm.muon.ep5_wer_clean}{4.43}
\defmmnum{ws2.whisper.qwen.layernorm.muon.ep5_wer_other}{10.06}
\defmmnum{ws2.whisper.qwen.layernorm.muon.erank}{691.8}
\defmmnum{ws2.whisper.qwen.layernorm.muon.erank_pct}{90.1}
\defmmnum{ws2.whisper.qwen.layernorm.muon.kappa}{19.4}
\defmmnum{ws2.whisper.qwen.layernorm.muon.live_epoch}{5.00}
\defmmnum{ws2.whisper.qwen.layernorm.muon.live_val}{0.1394}
\defmmnum{ws2.whisper.qwen.layernorm.muon.params}{1{,}185{,}792}
\defmmnum{ws2.whisper.qwen.layernorm.muon.params_m}{1.19M}
\defmmnum{ws2.whisper.qwen.layernorm.muon.wer_clean}{4.33}
\defmmnum{ws2.whisper.qwen.layernorm.muon.wer_clean_cell}{4.33\%}
\defmmnum{ws2.whisper.qwen.layernorm.muon.wer_other}{9.96}
\defmmnum{ws2.whisper.qwen.layernorm.muon.wer_other_cell}{9.96\%}
\defmmnum{ws2.whisper.qwen.layernorm.muon.wer_pair}{4.33 / 9.96}
\defmmnum{ws2.whisper.qwen.layernorm.muon.wer_pair_cell}{4.33\% / 9.96\%}
\defmmnum{ws2.whisper.qwen.none.muon.best_epoch}{4}
\defmmnum{ws2.whisper.qwen.none.muon.best_val}{0.1295}
\defmmnum{ws2.whisper.qwen.none.muon.best_val_cell}{0.1295}
\defmmnum{ws2.whisper.qwen.none.muon.empty_clean}{0}
\defmmnum{ws2.whisper.qwen.none.muon.empty_other}{0}
\defmmnum{ws2.whisper.qwen.none.muon.ep1_val}{0.1515}
\defmmnum{ws2.whisper.qwen.none.muon.ep2_val}{0.1399}
\defmmnum{ws2.whisper.qwen.none.muon.erank}{694.2}
\defmmnum{ws2.whisper.qwen.none.muon.erank_pct}{90.4}
\defmmnum{ws2.whisper.qwen.none.muon.kappa}{55.7}
\defmmnum{ws2.whisper.qwen.none.muon.live_epoch}{5.00}
\defmmnum{ws2.whisper.qwen.none.muon.live_val}{0.1295}
\defmmnum{ws2.whisper.qwen.none.muon.params}{1{,}181{,}184}
\defmmnum{ws2.whisper.qwen.none.muon.params_m}{1.18M}
\defmmnum{ws2.whisper.qwen.none.muon.wer_clean}{4.99}
\defmmnum{ws2.whisper.qwen.none.muon.wer_clean_cell}{4.99\%}
\defmmnum{ws2.whisper.qwen.none.muon.wer_other}{10.17}
\defmmnum{ws2.whisper.qwen.none.muon.wer_other_cell}{10.17\%}
\defmmnum{ws2.whisper.qwen.none.muon.wer_pair}{4.99 / 10.17}
\defmmnum{ws2.whisper.qwen.none.muon.wer_pair_cell}{4.99\% / 10.17\%}
\defmmnum{ws2.whisper.qwen.rmsnorm.muon.best_epoch}{4}
\defmmnum{ws2.whisper.qwen.rmsnorm.muon.best_val}{0.1425}
\defmmnum{ws2.whisper.qwen.rmsnorm.muon.best_val_cell}{0.1425}
\defmmnum{ws2.whisper.qwen.rmsnorm.muon.empty_clean}{0}
\defmmnum{ws2.whisper.qwen.rmsnorm.muon.empty_other}{0}
\defmmnum{ws2.whisper.qwen.rmsnorm.muon.ep1_val}{0.1899}
\defmmnum{ws2.whisper.qwen.rmsnorm.muon.ep2_val}{0.1547}
\defmmnum{ws2.whisper.qwen.rmsnorm.muon.erank}{691.8}
\defmmnum{ws2.whisper.qwen.rmsnorm.muon.erank_pct}{90.1}
\defmmnum{ws2.whisper.qwen.rmsnorm.muon.kappa}{24.9}
\defmmnum{ws2.whisper.qwen.rmsnorm.muon.live_epoch}{5.00}
\defmmnum{ws2.whisper.qwen.rmsnorm.muon.live_val}{0.1425}
\defmmnum{ws2.whisper.qwen.rmsnorm.muon.params}{1{,}183{,}488}
\defmmnum{ws2.whisper.qwen.rmsnorm.muon.params_m}{1.18M}
\defmmnum{ws2.whisper.qwen.rmsnorm.muon.wer_clean}{4.40}
\defmmnum{ws2.whisper.qwen.rmsnorm.muon.wer_clean_cell}{4.40\%}
\defmmnum{ws2.whisper.qwen.rmsnorm.muon.wer_other}{9.99}
\defmmnum{ws2.whisper.qwen.rmsnorm.muon.wer_other_cell}{9.99\%}
\defmmnum{ws2.whisper.qwen.rmsnorm.muon.wer_pair}{4.40 / 9.99}
\defmmnum{ws2.whisper.qwen.rmsnorm.muon.wer_pair_cell}{4.40\% / 9.99\%}
\defmmnum{ws2.whisper.qwen.scaled_l2norm.muon.best_epoch}{4}
\defmmnum{ws2.whisper.qwen.scaled_l2norm.muon.best_val}{0.1365}
\defmmnum{ws2.whisper.qwen.scaled_l2norm.muon.best_val_cell}{0.1365}
\defmmnum{ws2.whisper.qwen.scaled_l2norm.muon.empty_clean}{0}
\defmmnum{ws2.whisper.qwen.scaled_l2norm.muon.empty_other}{0}
\defmmnum{ws2.whisper.qwen.scaled_l2norm.muon.ep1_val}{0.1829}
\defmmnum{ws2.whisper.qwen.scaled_l2norm.muon.ep2_val}{0.1517}
\defmmnum{ws2.whisper.qwen.scaled_l2norm.muon.erank}{692.6}
\defmmnum{ws2.whisper.qwen.scaled_l2norm.muon.erank_pct}{90.2}
\defmmnum{ws2.whisper.qwen.scaled_l2norm.muon.kappa}{74.7}
\defmmnum{ws2.whisper.qwen.scaled_l2norm.muon.live_epoch}{5.00}
\defmmnum{ws2.whisper.qwen.scaled_l2norm.muon.live_val}{0.1365}
\defmmnum{ws2.whisper.qwen.scaled_l2norm.muon.params}{1{,}181{,}186}
\defmmnum{ws2.whisper.qwen.scaled_l2norm.muon.params_m}{1.18M}
\defmmnum{ws2.whisper.qwen.scaled_l2norm.muon.wer_clean}{4.30}
\defmmnum{ws2.whisper.qwen.scaled_l2norm.muon.wer_clean_cell}{4.30\%}
\defmmnum{ws2.whisper.qwen.scaled_l2norm.muon.wer_other}{9.90}
\defmmnum{ws2.whisper.qwen.scaled_l2norm.muon.wer_other_cell}{9.90\%}
\defmmnum{ws2.whisper.qwen.scaled_l2norm.muon.wer_pair}{4.30 / 9.90}
\defmmnum{ws2.whisper.qwen.scaled_l2norm.muon.wer_pair_cell}{4.30\% / 9.90\%}
\defmmnum{ws3.comp.closed_form_best_clean}{108.32}
\defmmnum{ws3.comp.lstsq_best_clean}{108.32}
\defmmnum{ws3.comp.lstsq_done}{40}
\defmmnum{ws3.comp.moonshine.gemmatwotogemmathree.lstsq.l2norm.clean}{264.94}
\defmmnum{ws3.comp.moonshine.gemmatwotogemmathree.lstsq.l2norm.clean_cell}{264.94\%}
\defmmnum{ws3.comp.moonshine.gemmatwotogemmathree.lstsq.l2norm.empty_clean}{5}
\defmmnum{ws3.comp.moonshine.gemmatwotogemmathree.lstsq.l2norm.empty_other}{1}
\defmmnum{ws3.comp.moonshine.gemmatwotogemmathree.lstsq.l2norm.other}{315.24}
\defmmnum{ws3.comp.moonshine.gemmatwotogemmathree.lstsq.l2norm.other_cell}{315.24\%}
\defmmnum{ws3.comp.moonshine.gemmatwotogemmathree.lstsq.l2norm.pair_cell}{264.9\% / 315.2\%}
\defmmnum{ws3.comp.moonshine.gemmatwotogemmathree.lstsq.layernorm.clean}{295.90}
\defmmnum{ws3.comp.moonshine.gemmatwotogemmathree.lstsq.layernorm.clean_cell}{295.90\%}
\defmmnum{ws3.comp.moonshine.gemmatwotogemmathree.lstsq.layernorm.empty_clean}{0}
\defmmnum{ws3.comp.moonshine.gemmatwotogemmathree.lstsq.layernorm.empty_other}{0}
\defmmnum{ws3.comp.moonshine.gemmatwotogemmathree.lstsq.layernorm.other}{327.08}
\defmmnum{ws3.comp.moonshine.gemmatwotogemmathree.lstsq.layernorm.other_cell}{327.08\%}
\defmmnum{ws3.comp.moonshine.gemmatwotogemmathree.lstsq.layernorm.pair_cell}{295.9\% / 327.1\%}
\defmmnum{ws3.comp.moonshine.gemmatwotogemmathree.lstsq.none.clean}{310.47}
\defmmnum{ws3.comp.moonshine.gemmatwotogemmathree.lstsq.none.clean_cell}{310.47\%}
\defmmnum{ws3.comp.moonshine.gemmatwotogemmathree.lstsq.none.empty_clean}{1}
\defmmnum{ws3.comp.moonshine.gemmatwotogemmathree.lstsq.none.empty_other}{0}
\defmmnum{ws3.comp.moonshine.gemmatwotogemmathree.lstsq.none.other}{359.43}
\defmmnum{ws3.comp.moonshine.gemmatwotogemmathree.lstsq.none.other_cell}{359.43\%}
\defmmnum{ws3.comp.moonshine.gemmatwotogemmathree.lstsq.none.pair_cell}{310.5\% / 359.4\%}
\defmmnum{ws3.comp.moonshine.gemmatwotogemmathree.lstsq.rmsnorm.clean}{306.46}
\defmmnum{ws3.comp.moonshine.gemmatwotogemmathree.lstsq.rmsnorm.clean_cell}{306.46\%}
\defmmnum{ws3.comp.moonshine.gemmatwotogemmathree.lstsq.rmsnorm.empty_clean}{2}
\defmmnum{ws3.comp.moonshine.gemmatwotogemmathree.lstsq.rmsnorm.empty_other}{1}
\defmmnum{ws3.comp.moonshine.gemmatwotogemmathree.lstsq.rmsnorm.other}{354.17}
\defmmnum{ws3.comp.moonshine.gemmatwotogemmathree.lstsq.rmsnorm.other_cell}{354.17\%}
\defmmnum{ws3.comp.moonshine.gemmatwotogemmathree.lstsq.rmsnorm.pair_cell}{306.5\% / 354.2\%}
\defmmnum{ws3.comp.moonshine.gemmatwotogemmathree.lstsq.scaled_l2norm.clean}{331.45}
\defmmnum{ws3.comp.moonshine.gemmatwotogemmathree.lstsq.scaled_l2norm.clean_cell}{331.45\%}
\defmmnum{ws3.comp.moonshine.gemmatwotogemmathree.lstsq.scaled_l2norm.empty_clean}{1}
\defmmnum{ws3.comp.moonshine.gemmatwotogemmathree.lstsq.scaled_l2norm.empty_other}{0}
\defmmnum{ws3.comp.moonshine.gemmatwotogemmathree.lstsq.scaled_l2norm.other}{381.44}
\defmmnum{ws3.comp.moonshine.gemmatwotogemmathree.lstsq.scaled_l2norm.other_cell}{381.44\%}
\defmmnum{ws3.comp.moonshine.gemmatwotogemmathree.lstsq.scaled_l2norm.pair_cell}{331.4\% / 381.4\%}
\defmmnum{ws3.comp.moonshine.gemmatwotogemmathree.procrustes.l2norm.clean}{315.80}
\defmmnum{ws3.comp.moonshine.gemmatwotogemmathree.procrustes.l2norm.clean_cell}{315.80\%}
\defmmnum{ws3.comp.moonshine.gemmatwotogemmathree.procrustes.l2norm.empty_clean}{0}
\defmmnum{ws3.comp.moonshine.gemmatwotogemmathree.procrustes.l2norm.empty_other}{0}
\defmmnum{ws3.comp.moonshine.gemmatwotogemmathree.procrustes.l2norm.other}{344.84}
\defmmnum{ws3.comp.moonshine.gemmatwotogemmathree.procrustes.l2norm.other_cell}{344.84\%}
\defmmnum{ws3.comp.moonshine.gemmatwotogemmathree.procrustes.l2norm.pair_cell}{315.8\% / 344.8\%}
\defmmnum{ws3.comp.moonshine.gemmatwotogemmathree.procrustes.layernorm.clean}{336.22}
\defmmnum{ws3.comp.moonshine.gemmatwotogemmathree.procrustes.layernorm.clean_cell}{336.22\%}
\defmmnum{ws3.comp.moonshine.gemmatwotogemmathree.procrustes.layernorm.empty_clean}{0}
\defmmnum{ws3.comp.moonshine.gemmatwotogemmathree.procrustes.layernorm.empty_other}{0}
\defmmnum{ws3.comp.moonshine.gemmatwotogemmathree.procrustes.layernorm.other}{375.88}
\defmmnum{ws3.comp.moonshine.gemmatwotogemmathree.procrustes.layernorm.other_cell}{375.88\%}
\defmmnum{ws3.comp.moonshine.gemmatwotogemmathree.procrustes.layernorm.pair_cell}{336.2\% / 375.9\%}
\defmmnum{ws3.comp.moonshine.gemmatwotogemmathree.procrustes.none.clean}{326.43}
\defmmnum{ws3.comp.moonshine.gemmatwotogemmathree.procrustes.none.clean_cell}{326.43\%}
\defmmnum{ws3.comp.moonshine.gemmatwotogemmathree.procrustes.none.empty_clean}{0}
\defmmnum{ws3.comp.moonshine.gemmatwotogemmathree.procrustes.none.empty_other}{0}
\defmmnum{ws3.comp.moonshine.gemmatwotogemmathree.procrustes.none.other}{356.58}
\defmmnum{ws3.comp.moonshine.gemmatwotogemmathree.procrustes.none.other_cell}{356.58\%}
\defmmnum{ws3.comp.moonshine.gemmatwotogemmathree.procrustes.none.pair_cell}{326.4\% / 356.6\%}
\defmmnum{ws3.comp.moonshine.gemmatwotogemmathree.procrustes.rmsnorm.clean}{441.35}
\defmmnum{ws3.comp.moonshine.gemmatwotogemmathree.procrustes.rmsnorm.clean_cell}{441.35\%}
\defmmnum{ws3.comp.moonshine.gemmatwotogemmathree.procrustes.rmsnorm.empty_clean}{0}
\defmmnum{ws3.comp.moonshine.gemmatwotogemmathree.procrustes.rmsnorm.empty_other}{0}
\defmmnum{ws3.comp.moonshine.gemmatwotogemmathree.procrustes.rmsnorm.other}{502.80}
\defmmnum{ws3.comp.moonshine.gemmatwotogemmathree.procrustes.rmsnorm.other_cell}{502.80\%}
\defmmnum{ws3.comp.moonshine.gemmatwotogemmathree.procrustes.rmsnorm.pair_cell}{441.3\% / 502.8\%}
\defmmnum{ws3.comp.moonshine.gemmatwotogemmathree.procrustes.scaled_l2norm.clean}{417.29}
\defmmnum{ws3.comp.moonshine.gemmatwotogemmathree.procrustes.scaled_l2norm.clean_cell}{417.29\%}
\defmmnum{ws3.comp.moonshine.gemmatwotogemmathree.procrustes.scaled_l2norm.empty_clean}{0}
\defmmnum{ws3.comp.moonshine.gemmatwotogemmathree.procrustes.scaled_l2norm.empty_other}{0}
\defmmnum{ws3.comp.moonshine.gemmatwotogemmathree.procrustes.scaled_l2norm.other}{475.47}
\defmmnum{ws3.comp.moonshine.gemmatwotogemmathree.procrustes.scaled_l2norm.other_cell}{475.47\%}
\defmmnum{ws3.comp.moonshine.gemmatwotogemmathree.procrustes.scaled_l2norm.pair_cell}{417.3\% / 475.5\%}
\defmmnum{ws3.comp.moonshine.gemmatwotogemmathree.random_orthogonal.scaled_l2norm.clean}{308.73}
\defmmnum{ws3.comp.moonshine.gemmatwotogemmathree.random_orthogonal.scaled_l2norm.other}{385.69}
\defmmnum{ws3.comp.moonshine.gemmatwotoqwen.lstsq.l2norm.clean}{130.40}
\defmmnum{ws3.comp.moonshine.gemmatwotoqwen.lstsq.l2norm.clean_cell}{130.40\%}
\defmmnum{ws3.comp.moonshine.gemmatwotoqwen.lstsq.l2norm.empty_clean}{0}
\defmmnum{ws3.comp.moonshine.gemmatwotoqwen.lstsq.l2norm.empty_other}{0}
\defmmnum{ws3.comp.moonshine.gemmatwotoqwen.lstsq.l2norm.other}{140.63}
\defmmnum{ws3.comp.moonshine.gemmatwotoqwen.lstsq.l2norm.other_cell}{140.63\%}
\defmmnum{ws3.comp.moonshine.gemmatwotoqwen.lstsq.l2norm.pair_cell}{130.4\% / 140.6\%}
\defmmnum{ws3.comp.moonshine.gemmatwotoqwen.lstsq.layernorm.clean}{108.32}
\defmmnum{ws3.comp.moonshine.gemmatwotoqwen.lstsq.layernorm.clean_cell}{108.32\%}
\defmmnum{ws3.comp.moonshine.gemmatwotoqwen.lstsq.layernorm.empty_clean}{0}
\defmmnum{ws3.comp.moonshine.gemmatwotoqwen.lstsq.layernorm.empty_other}{0}
\defmmnum{ws3.comp.moonshine.gemmatwotoqwen.lstsq.layernorm.other}{113.60}
\defmmnum{ws3.comp.moonshine.gemmatwotoqwen.lstsq.layernorm.other_cell}{113.60\%}
\defmmnum{ws3.comp.moonshine.gemmatwotoqwen.lstsq.layernorm.pair_cell}{108.3\% / 113.6\%}
\defmmnum{ws3.comp.moonshine.gemmatwotoqwen.lstsq.none.clean}{130.97}
\defmmnum{ws3.comp.moonshine.gemmatwotoqwen.lstsq.none.clean_cell}{130.97\%}
\defmmnum{ws3.comp.moonshine.gemmatwotoqwen.lstsq.none.empty_clean}{0}
\defmmnum{ws3.comp.moonshine.gemmatwotoqwen.lstsq.none.empty_other}{0}
\defmmnum{ws3.comp.moonshine.gemmatwotoqwen.lstsq.none.other}{140.05}
\defmmnum{ws3.comp.moonshine.gemmatwotoqwen.lstsq.none.other_cell}{140.05\%}
\defmmnum{ws3.comp.moonshine.gemmatwotoqwen.lstsq.none.pair_cell}{131.0\% / 140.1\%}
\defmmnum{ws3.comp.moonshine.gemmatwotoqwen.lstsq.rmsnorm.clean}{143.91}
\defmmnum{ws3.comp.moonshine.gemmatwotoqwen.lstsq.rmsnorm.clean_cell}{143.91\%}
\defmmnum{ws3.comp.moonshine.gemmatwotoqwen.lstsq.rmsnorm.empty_clean}{0}
\defmmnum{ws3.comp.moonshine.gemmatwotoqwen.lstsq.rmsnorm.empty_other}{0}
\defmmnum{ws3.comp.moonshine.gemmatwotoqwen.lstsq.rmsnorm.other}{151.33}
\defmmnum{ws3.comp.moonshine.gemmatwotoqwen.lstsq.rmsnorm.other_cell}{151.33\%}
\defmmnum{ws3.comp.moonshine.gemmatwotoqwen.lstsq.rmsnorm.pair_cell}{143.9\% / 151.3\%}
\defmmnum{ws3.comp.moonshine.gemmatwotoqwen.lstsq.scaled_l2norm.clean}{130.70}
\defmmnum{ws3.comp.moonshine.gemmatwotoqwen.lstsq.scaled_l2norm.clean_cell}{130.70\%}
\defmmnum{ws3.comp.moonshine.gemmatwotoqwen.lstsq.scaled_l2norm.empty_clean}{0}
\defmmnum{ws3.comp.moonshine.gemmatwotoqwen.lstsq.scaled_l2norm.empty_other}{0}
\defmmnum{ws3.comp.moonshine.gemmatwotoqwen.lstsq.scaled_l2norm.other}{140.52}
\defmmnum{ws3.comp.moonshine.gemmatwotoqwen.lstsq.scaled_l2norm.other_cell}{140.52\%}
\defmmnum{ws3.comp.moonshine.gemmatwotoqwen.lstsq.scaled_l2norm.pair_cell}{130.7\% / 140.5\%}
\defmmnum{ws3.comp.moonshine.gemmatwotoqwen.procrustes.l2norm.clean}{159.00}
\defmmnum{ws3.comp.moonshine.gemmatwotoqwen.procrustes.l2norm.clean_cell}{159.00\%}
\defmmnum{ws3.comp.moonshine.gemmatwotoqwen.procrustes.l2norm.empty_clean}{0}
\defmmnum{ws3.comp.moonshine.gemmatwotoqwen.procrustes.l2norm.empty_other}{0}
\defmmnum{ws3.comp.moonshine.gemmatwotoqwen.procrustes.l2norm.other}{175.40}
\defmmnum{ws3.comp.moonshine.gemmatwotoqwen.procrustes.l2norm.other_cell}{175.40\%}
\defmmnum{ws3.comp.moonshine.gemmatwotoqwen.procrustes.l2norm.pair_cell}{159.0\% / 175.4\%}
\defmmnum{ws3.comp.moonshine.gemmatwotoqwen.procrustes.layernorm.clean}{183.11}
\defmmnum{ws3.comp.moonshine.gemmatwotoqwen.procrustes.layernorm.clean_cell}{183.11\%}
\defmmnum{ws3.comp.moonshine.gemmatwotoqwen.procrustes.layernorm.empty_clean}{0}
\defmmnum{ws3.comp.moonshine.gemmatwotoqwen.procrustes.layernorm.empty_other}{0}
\defmmnum{ws3.comp.moonshine.gemmatwotoqwen.procrustes.layernorm.other}{201.35}
\defmmnum{ws3.comp.moonshine.gemmatwotoqwen.procrustes.layernorm.other_cell}{201.35\%}
\defmmnum{ws3.comp.moonshine.gemmatwotoqwen.procrustes.layernorm.pair_cell}{183.1\% / 201.4\%}
\defmmnum{ws3.comp.moonshine.gemmatwotoqwen.procrustes.none.clean}{237.82}
\defmmnum{ws3.comp.moonshine.gemmatwotoqwen.procrustes.none.clean_cell}{237.82\%}
\defmmnum{ws3.comp.moonshine.gemmatwotoqwen.procrustes.none.empty_clean}{37}
\defmmnum{ws3.comp.moonshine.gemmatwotoqwen.procrustes.none.empty_other}{32}
\defmmnum{ws3.comp.moonshine.gemmatwotoqwen.procrustes.none.other}{251.15}
\defmmnum{ws3.comp.moonshine.gemmatwotoqwen.procrustes.none.other_cell}{251.15\%}
\defmmnum{ws3.comp.moonshine.gemmatwotoqwen.procrustes.none.pair_cell}{237.8\% / 251.2\%}
\defmmnum{ws3.comp.moonshine.gemmatwotoqwen.procrustes.rmsnorm.clean}{165.93}
\defmmnum{ws3.comp.moonshine.gemmatwotoqwen.procrustes.rmsnorm.clean_cell}{165.93\%}
\defmmnum{ws3.comp.moonshine.gemmatwotoqwen.procrustes.rmsnorm.empty_clean}{0}
\defmmnum{ws3.comp.moonshine.gemmatwotoqwen.procrustes.rmsnorm.empty_other}{0}
\defmmnum{ws3.comp.moonshine.gemmatwotoqwen.procrustes.rmsnorm.other}{185.58}
\defmmnum{ws3.comp.moonshine.gemmatwotoqwen.procrustes.rmsnorm.other_cell}{185.58\%}
\defmmnum{ws3.comp.moonshine.gemmatwotoqwen.procrustes.rmsnorm.pair_cell}{165.9\% / 185.6\%}
\defmmnum{ws3.comp.moonshine.gemmatwotoqwen.procrustes.scaled_l2norm.clean}{159.39}
\defmmnum{ws3.comp.moonshine.gemmatwotoqwen.procrustes.scaled_l2norm.clean_cell}{159.39\%}
\defmmnum{ws3.comp.moonshine.gemmatwotoqwen.procrustes.scaled_l2norm.empty_clean}{0}
\defmmnum{ws3.comp.moonshine.gemmatwotoqwen.procrustes.scaled_l2norm.empty_other}{0}
\defmmnum{ws3.comp.moonshine.gemmatwotoqwen.procrustes.scaled_l2norm.other}{175.38}
\defmmnum{ws3.comp.moonshine.gemmatwotoqwen.procrustes.scaled_l2norm.other_cell}{175.38\%}
\defmmnum{ws3.comp.moonshine.gemmatwotoqwen.procrustes.scaled_l2norm.pair_cell}{159.4\% / 175.4\%}
\defmmnum{ws3.comp.moonshine.gemmatwotoqwen.random_orthogonal.scaled_l2norm.clean}{122.42}
\defmmnum{ws3.comp.moonshine.gemmatwotoqwen.random_orthogonal.scaled_l2norm.other}{128.59}
\defmmnum{ws3.comp.moonshine.qwentogemmathree.lstsq.l2norm.clean}{131.55}
\defmmnum{ws3.comp.moonshine.qwentogemmathree.lstsq.l2norm.clean_cell}{131.55\%}
\defmmnum{ws3.comp.moonshine.qwentogemmathree.lstsq.l2norm.empty_clean}{1488}
\defmmnum{ws3.comp.moonshine.qwentogemmathree.lstsq.l2norm.empty_other}{1490}
\defmmnum{ws3.comp.moonshine.qwentogemmathree.lstsq.l2norm.other}{141.73}
\defmmnum{ws3.comp.moonshine.qwentogemmathree.lstsq.l2norm.other_cell}{141.73\%}
\defmmnum{ws3.comp.moonshine.qwentogemmathree.lstsq.l2norm.pair_cell}{131.5\% / 141.7\%}
\defmmnum{ws3.comp.moonshine.qwentogemmathree.lstsq.layernorm.clean}{383.47}
\defmmnum{ws3.comp.moonshine.qwentogemmathree.lstsq.layernorm.clean_cell}{383.47\%}
\defmmnum{ws3.comp.moonshine.qwentogemmathree.lstsq.layernorm.empty_clean}{0}
\defmmnum{ws3.comp.moonshine.qwentogemmathree.lstsq.layernorm.empty_other}{0}
\defmmnum{ws3.comp.moonshine.qwentogemmathree.lstsq.layernorm.other}{431.91}
\defmmnum{ws3.comp.moonshine.qwentogemmathree.lstsq.layernorm.other_cell}{431.91\%}
\defmmnum{ws3.comp.moonshine.qwentogemmathree.lstsq.layernorm.pair_cell}{383.5\% / 431.9\%}
\defmmnum{ws3.comp.moonshine.qwentogemmathree.lstsq.none.clean}{191.23}
\defmmnum{ws3.comp.moonshine.qwentogemmathree.lstsq.none.clean_cell}{191.23\%}
\defmmnum{ws3.comp.moonshine.qwentogemmathree.lstsq.none.empty_clean}{706}
\defmmnum{ws3.comp.moonshine.qwentogemmathree.lstsq.none.empty_other}{727}
\defmmnum{ws3.comp.moonshine.qwentogemmathree.lstsq.none.other}{215.81}
\defmmnum{ws3.comp.moonshine.qwentogemmathree.lstsq.none.other_cell}{215.81\%}
\defmmnum{ws3.comp.moonshine.qwentogemmathree.lstsq.none.pair_cell}{191.2\% / 215.8\%}
\defmmnum{ws3.comp.moonshine.qwentogemmathree.lstsq.rmsnorm.clean}{130.11}
\defmmnum{ws3.comp.moonshine.qwentogemmathree.lstsq.rmsnorm.clean_cell}{130.11\%}
\defmmnum{ws3.comp.moonshine.qwentogemmathree.lstsq.rmsnorm.empty_clean}{1676}
\defmmnum{ws3.comp.moonshine.qwentogemmathree.lstsq.rmsnorm.empty_other}{1831}
\defmmnum{ws3.comp.moonshine.qwentogemmathree.lstsq.rmsnorm.other}{139.52}
\defmmnum{ws3.comp.moonshine.qwentogemmathree.lstsq.rmsnorm.other_cell}{139.52\%}
\defmmnum{ws3.comp.moonshine.qwentogemmathree.lstsq.rmsnorm.pair_cell}{130.1\% / 139.5\%}
\defmmnum{ws3.comp.moonshine.qwentogemmathree.lstsq.scaled_l2norm.clean}{173.27}
\defmmnum{ws3.comp.moonshine.qwentogemmathree.lstsq.scaled_l2norm.clean_cell}{173.27\%}
\defmmnum{ws3.comp.moonshine.qwentogemmathree.lstsq.scaled_l2norm.empty_clean}{1188}
\defmmnum{ws3.comp.moonshine.qwentogemmathree.lstsq.scaled_l2norm.empty_other}{1217}
\defmmnum{ws3.comp.moonshine.qwentogemmathree.lstsq.scaled_l2norm.other}{191.93}
\defmmnum{ws3.comp.moonshine.qwentogemmathree.lstsq.scaled_l2norm.other_cell}{191.93\%}
\defmmnum{ws3.comp.moonshine.qwentogemmathree.lstsq.scaled_l2norm.pair_cell}{173.3\% / 191.9\%}
\defmmnum{ws3.comp.moonshine.qwentogemmathree.procrustes.l2norm.clean}{307.16}
\defmmnum{ws3.comp.moonshine.qwentogemmathree.procrustes.l2norm.clean_cell}{307.16\%}
\defmmnum{ws3.comp.moonshine.qwentogemmathree.procrustes.l2norm.empty_clean}{383}
\defmmnum{ws3.comp.moonshine.qwentogemmathree.procrustes.l2norm.empty_other}{268}
\defmmnum{ws3.comp.moonshine.qwentogemmathree.procrustes.l2norm.other}{334.84}
\defmmnum{ws3.comp.moonshine.qwentogemmathree.procrustes.l2norm.other_cell}{334.84\%}
\defmmnum{ws3.comp.moonshine.qwentogemmathree.procrustes.l2norm.pair_cell}{307.2\% / 334.8\%}
\defmmnum{ws3.comp.moonshine.qwentogemmathree.procrustes.layernorm.clean}{415.39}
\defmmnum{ws3.comp.moonshine.qwentogemmathree.procrustes.layernorm.clean_cell}{415.39\%}
\defmmnum{ws3.comp.moonshine.qwentogemmathree.procrustes.layernorm.empty_clean}{0}
\defmmnum{ws3.comp.moonshine.qwentogemmathree.procrustes.layernorm.empty_other}{0}
\defmmnum{ws3.comp.moonshine.qwentogemmathree.procrustes.layernorm.other}{465.44}
\defmmnum{ws3.comp.moonshine.qwentogemmathree.procrustes.layernorm.other_cell}{465.44\%}
\defmmnum{ws3.comp.moonshine.qwentogemmathree.procrustes.layernorm.pair_cell}{415.4\% / 465.4\%}
\defmmnum{ws3.comp.moonshine.qwentogemmathree.procrustes.none.clean}{304.64}
\defmmnum{ws3.comp.moonshine.qwentogemmathree.procrustes.none.clean_cell}{304.64\%}
\defmmnum{ws3.comp.moonshine.qwentogemmathree.procrustes.none.empty_clean}{308}
\defmmnum{ws3.comp.moonshine.qwentogemmathree.procrustes.none.empty_other}{230}
\defmmnum{ws3.comp.moonshine.qwentogemmathree.procrustes.none.other}{337.91}
\defmmnum{ws3.comp.moonshine.qwentogemmathree.procrustes.none.other_cell}{337.91\%}
\defmmnum{ws3.comp.moonshine.qwentogemmathree.procrustes.none.pair_cell}{304.6\% / 337.9\%}
\defmmnum{ws3.comp.moonshine.qwentogemmathree.procrustes.rmsnorm.clean}{323.57}
\defmmnum{ws3.comp.moonshine.qwentogemmathree.procrustes.rmsnorm.clean_cell}{323.57\%}
\defmmnum{ws3.comp.moonshine.qwentogemmathree.procrustes.rmsnorm.empty_clean}{281}
\defmmnum{ws3.comp.moonshine.qwentogemmathree.procrustes.rmsnorm.empty_other}{183}
\defmmnum{ws3.comp.moonshine.qwentogemmathree.procrustes.rmsnorm.other}{376.70}
\defmmnum{ws3.comp.moonshine.qwentogemmathree.procrustes.rmsnorm.other_cell}{376.70\%}
\defmmnum{ws3.comp.moonshine.qwentogemmathree.procrustes.rmsnorm.pair_cell}{323.6\% / 376.7\%}
\defmmnum{ws3.comp.moonshine.qwentogemmathree.procrustes.scaled_l2norm.clean}{346.91}
\defmmnum{ws3.comp.moonshine.qwentogemmathree.procrustes.scaled_l2norm.clean_cell}{346.91\%}
\defmmnum{ws3.comp.moonshine.qwentogemmathree.procrustes.scaled_l2norm.empty_clean}{291}
\defmmnum{ws3.comp.moonshine.qwentogemmathree.procrustes.scaled_l2norm.empty_other}{184}
\defmmnum{ws3.comp.moonshine.qwentogemmathree.procrustes.scaled_l2norm.other}{403.12}
\defmmnum{ws3.comp.moonshine.qwentogemmathree.procrustes.scaled_l2norm.other_cell}{403.12\%}
\defmmnum{ws3.comp.moonshine.qwentogemmathree.procrustes.scaled_l2norm.pair_cell}{346.9\% / 403.1\%}
\defmmnum{ws3.comp.moonshine.qwentogemmatwo.lstsq.l2norm.clean}{201.54}
\defmmnum{ws3.comp.moonshine.qwentogemmatwo.lstsq.l2norm.clean_cell}{201.54\%}
\defmmnum{ws3.comp.moonshine.qwentogemmatwo.lstsq.l2norm.empty_clean}{0}
\defmmnum{ws3.comp.moonshine.qwentogemmatwo.lstsq.l2norm.empty_other}{0}
\defmmnum{ws3.comp.moonshine.qwentogemmatwo.lstsq.l2norm.other}{227.24}
\defmmnum{ws3.comp.moonshine.qwentogemmatwo.lstsq.l2norm.other_cell}{227.24\%}
\defmmnum{ws3.comp.moonshine.qwentogemmatwo.lstsq.l2norm.pair_cell}{201.5\% / 227.2\%}
\defmmnum{ws3.comp.moonshine.qwentogemmatwo.lstsq.layernorm.clean}{284.50}
\defmmnum{ws3.comp.moonshine.qwentogemmatwo.lstsq.layernorm.clean_cell}{284.50\%}
\defmmnum{ws3.comp.moonshine.qwentogemmatwo.lstsq.layernorm.empty_clean}{0}
\defmmnum{ws3.comp.moonshine.qwentogemmatwo.lstsq.layernorm.empty_other}{0}
\defmmnum{ws3.comp.moonshine.qwentogemmatwo.lstsq.layernorm.other}{329.75}
\defmmnum{ws3.comp.moonshine.qwentogemmatwo.lstsq.layernorm.other_cell}{329.75\%}
\defmmnum{ws3.comp.moonshine.qwentogemmatwo.lstsq.layernorm.pair_cell}{284.5\% / 329.7\%}
\defmmnum{ws3.comp.moonshine.qwentogemmatwo.lstsq.none.clean}{253.97}
\defmmnum{ws3.comp.moonshine.qwentogemmatwo.lstsq.none.clean_cell}{253.97\%}
\defmmnum{ws3.comp.moonshine.qwentogemmatwo.lstsq.none.empty_clean}{235}
\defmmnum{ws3.comp.moonshine.qwentogemmatwo.lstsq.none.empty_other}{206}
\defmmnum{ws3.comp.moonshine.qwentogemmatwo.lstsq.none.other}{307.83}
\defmmnum{ws3.comp.moonshine.qwentogemmatwo.lstsq.none.other_cell}{307.83\%}
\defmmnum{ws3.comp.moonshine.qwentogemmatwo.lstsq.none.pair_cell}{254.0\% / 307.8\%}
\defmmnum{ws3.comp.moonshine.qwentogemmatwo.lstsq.rmsnorm.clean}{372.77}
\defmmnum{ws3.comp.moonshine.qwentogemmatwo.lstsq.rmsnorm.clean_cell}{372.77\%}
\defmmnum{ws3.comp.moonshine.qwentogemmatwo.lstsq.rmsnorm.empty_clean}{619}
\defmmnum{ws3.comp.moonshine.qwentogemmatwo.lstsq.rmsnorm.empty_other}{564}
\defmmnum{ws3.comp.moonshine.qwentogemmatwo.lstsq.rmsnorm.other}{430.15}
\defmmnum{ws3.comp.moonshine.qwentogemmatwo.lstsq.rmsnorm.other_cell}{430.15\%}
\defmmnum{ws3.comp.moonshine.qwentogemmatwo.lstsq.rmsnorm.pair_cell}{372.8\% / 430.2\%}
\defmmnum{ws3.comp.moonshine.qwentogemmatwo.lstsq.scaled_l2norm.clean}{325.88}
\defmmnum{ws3.comp.moonshine.qwentogemmatwo.lstsq.scaled_l2norm.clean_cell}{325.88\%}
\defmmnum{ws3.comp.moonshine.qwentogemmatwo.lstsq.scaled_l2norm.empty_clean}{0}
\defmmnum{ws3.comp.moonshine.qwentogemmatwo.lstsq.scaled_l2norm.empty_other}{0}
\defmmnum{ws3.comp.moonshine.qwentogemmatwo.lstsq.scaled_l2norm.other}{368.94}
\defmmnum{ws3.comp.moonshine.qwentogemmatwo.lstsq.scaled_l2norm.other_cell}{368.94\%}
\defmmnum{ws3.comp.moonshine.qwentogemmatwo.lstsq.scaled_l2norm.pair_cell}{325.9\% / 368.9\%}
\defmmnum{ws3.comp.moonshine.qwentogemmatwo.procrustes.l2norm.clean}{274.59}
\defmmnum{ws3.comp.moonshine.qwentogemmatwo.procrustes.l2norm.clean_cell}{274.59\%}
\defmmnum{ws3.comp.moonshine.qwentogemmatwo.procrustes.l2norm.empty_clean}{767}
\defmmnum{ws3.comp.moonshine.qwentogemmatwo.procrustes.l2norm.empty_other}{703}
\defmmnum{ws3.comp.moonshine.qwentogemmatwo.procrustes.l2norm.other}{314.81}
\defmmnum{ws3.comp.moonshine.qwentogemmatwo.procrustes.l2norm.other_cell}{314.81\%}
\defmmnum{ws3.comp.moonshine.qwentogemmatwo.procrustes.l2norm.pair_cell}{274.6\% / 314.8\%}
\defmmnum{ws3.comp.moonshine.qwentogemmatwo.procrustes.layernorm.clean}{476.43}
\defmmnum{ws3.comp.moonshine.qwentogemmatwo.procrustes.layernorm.clean_cell}{476.43\%}
\defmmnum{ws3.comp.moonshine.qwentogemmatwo.procrustes.layernorm.empty_clean}{0}
\defmmnum{ws3.comp.moonshine.qwentogemmatwo.procrustes.layernorm.empty_other}{0}
\defmmnum{ws3.comp.moonshine.qwentogemmatwo.procrustes.layernorm.other}{533.38}
\defmmnum{ws3.comp.moonshine.qwentogemmatwo.procrustes.layernorm.other_cell}{533.38\%}
\defmmnum{ws3.comp.moonshine.qwentogemmatwo.procrustes.layernorm.pair_cell}{476.4\% / 533.4\%}
\defmmnum{ws3.comp.moonshine.qwentogemmatwo.procrustes.none.clean}{266.67}
\defmmnum{ws3.comp.moonshine.qwentogemmatwo.procrustes.none.clean_cell}{266.67\%}
\defmmnum{ws3.comp.moonshine.qwentogemmatwo.procrustes.none.empty_clean}{741}
\defmmnum{ws3.comp.moonshine.qwentogemmatwo.procrustes.none.empty_other}{677}
\defmmnum{ws3.comp.moonshine.qwentogemmatwo.procrustes.none.other}{306.86}
\defmmnum{ws3.comp.moonshine.qwentogemmatwo.procrustes.none.other_cell}{306.86\%}
\defmmnum{ws3.comp.moonshine.qwentogemmatwo.procrustes.none.pair_cell}{266.7\% / 306.9\%}
\defmmnum{ws3.comp.moonshine.qwentogemmatwo.procrustes.rmsnorm.clean}{160.76}
\defmmnum{ws3.comp.moonshine.qwentogemmatwo.procrustes.rmsnorm.clean_cell}{160.76\%}
\defmmnum{ws3.comp.moonshine.qwentogemmatwo.procrustes.rmsnorm.empty_clean}{2193}
\defmmnum{ws3.comp.moonshine.qwentogemmatwo.procrustes.rmsnorm.empty_other}{2285}
\defmmnum{ws3.comp.moonshine.qwentogemmatwo.procrustes.rmsnorm.other}{193.84}
\defmmnum{ws3.comp.moonshine.qwentogemmatwo.procrustes.rmsnorm.other_cell}{193.84\%}
\defmmnum{ws3.comp.moonshine.qwentogemmatwo.procrustes.rmsnorm.pair_cell}{160.8\% / 193.8\%}
\defmmnum{ws3.comp.moonshine.qwentogemmatwo.procrustes.scaled_l2norm.clean}{291.24}
\defmmnum{ws3.comp.moonshine.qwentogemmatwo.procrustes.scaled_l2norm.clean_cell}{291.24\%}
\defmmnum{ws3.comp.moonshine.qwentogemmatwo.procrustes.scaled_l2norm.empty_clean}{1137}
\defmmnum{ws3.comp.moonshine.qwentogemmatwo.procrustes.scaled_l2norm.empty_other}{947}
\defmmnum{ws3.comp.moonshine.qwentogemmatwo.procrustes.scaled_l2norm.other}{353.47}
\defmmnum{ws3.comp.moonshine.qwentogemmatwo.procrustes.scaled_l2norm.other_cell}{353.47\%}
\defmmnum{ws3.comp.moonshine.qwentogemmatwo.procrustes.scaled_l2norm.pair_cell}{291.2\% / 353.5\%}
\defmmnum{ws3.comp.proc_max_clean}{446.03}
\defmmnum{ws3.comp.proc_min_clean}{154.69}
\defmmnum{ws3.comp.procrustes_done}{16}
\defmmnum{ws3.comp.procrustes_extra_done}{24}
\defmmnum{ws3.comp.rand_orth_max_clean}{308.73}
\defmmnum{ws3.comp.rand_orth_min_clean}{122.42}
\defmmnum{ws3.comp.same_above_direct_clean_rows}{8}
\defmmnum{ws3.comp.total_rows}{8}
\defmmnum{ws3.comp.whisper.gemmatwotogemmathree.lstsq.l2norm.clean}{197.49}
\defmmnum{ws3.comp.whisper.gemmatwotogemmathree.lstsq.l2norm.clean_cell}{197.49\%}
\defmmnum{ws3.comp.whisper.gemmatwotogemmathree.lstsq.l2norm.empty_clean}{307}
\defmmnum{ws3.comp.whisper.gemmatwotogemmathree.lstsq.l2norm.empty_other}{306}
\defmmnum{ws3.comp.whisper.gemmatwotogemmathree.lstsq.l2norm.other}{229.12}
\defmmnum{ws3.comp.whisper.gemmatwotogemmathree.lstsq.l2norm.other_cell}{229.12\%}
\defmmnum{ws3.comp.whisper.gemmatwotogemmathree.lstsq.l2norm.pair_cell}{197.5\% / 229.1\%}
\defmmnum{ws3.comp.whisper.gemmatwotogemmathree.lstsq.layernorm.clean}{273.06}
\defmmnum{ws3.comp.whisper.gemmatwotogemmathree.lstsq.layernorm.clean_cell}{273.06\%}
\defmmnum{ws3.comp.whisper.gemmatwotogemmathree.lstsq.layernorm.empty_clean}{0}
\defmmnum{ws3.comp.whisper.gemmatwotogemmathree.lstsq.layernorm.empty_other}{0}
\defmmnum{ws3.comp.whisper.gemmatwotogemmathree.lstsq.layernorm.other}{300.98}
\defmmnum{ws3.comp.whisper.gemmatwotogemmathree.lstsq.layernorm.other_cell}{300.98\%}
\defmmnum{ws3.comp.whisper.gemmatwotogemmathree.lstsq.layernorm.pair_cell}{273.1\% / 301.0\%}
\defmmnum{ws3.comp.whisper.gemmatwotogemmathree.lstsq.none.clean}{257.38}
\defmmnum{ws3.comp.whisper.gemmatwotogemmathree.lstsq.none.clean_cell}{257.38\%}
\defmmnum{ws3.comp.whisper.gemmatwotogemmathree.lstsq.none.empty_clean}{205}
\defmmnum{ws3.comp.whisper.gemmatwotogemmathree.lstsq.none.empty_other}{231}
\defmmnum{ws3.comp.whisper.gemmatwotogemmathree.lstsq.none.other}{295.20}
\defmmnum{ws3.comp.whisper.gemmatwotogemmathree.lstsq.none.other_cell}{295.20\%}
\defmmnum{ws3.comp.whisper.gemmatwotogemmathree.lstsq.none.pair_cell}{257.4\% / 295.2\%}
\defmmnum{ws3.comp.whisper.gemmatwotogemmathree.lstsq.rmsnorm.clean}{234.69}
\defmmnum{ws3.comp.whisper.gemmatwotogemmathree.lstsq.rmsnorm.clean_cell}{234.69\%}
\defmmnum{ws3.comp.whisper.gemmatwotogemmathree.lstsq.rmsnorm.empty_clean}{192}
\defmmnum{ws3.comp.whisper.gemmatwotogemmathree.lstsq.rmsnorm.empty_other}{164}
\defmmnum{ws3.comp.whisper.gemmatwotogemmathree.lstsq.rmsnorm.other}{263.53}
\defmmnum{ws3.comp.whisper.gemmatwotogemmathree.lstsq.rmsnorm.other_cell}{263.53\%}
\defmmnum{ws3.comp.whisper.gemmatwotogemmathree.lstsq.rmsnorm.pair_cell}{234.7\% / 263.5\%}
\defmmnum{ws3.comp.whisper.gemmatwotogemmathree.lstsq.scaled_l2norm.clean}{263.67}
\defmmnum{ws3.comp.whisper.gemmatwotogemmathree.lstsq.scaled_l2norm.clean_cell}{263.67\%}
\defmmnum{ws3.comp.whisper.gemmatwotogemmathree.lstsq.scaled_l2norm.empty_clean}{191}
\defmmnum{ws3.comp.whisper.gemmatwotogemmathree.lstsq.scaled_l2norm.empty_other}{232}
\defmmnum{ws3.comp.whisper.gemmatwotogemmathree.lstsq.scaled_l2norm.other}{287.42}
\defmmnum{ws3.comp.whisper.gemmatwotogemmathree.lstsq.scaled_l2norm.other_cell}{287.42\%}
\defmmnum{ws3.comp.whisper.gemmatwotogemmathree.lstsq.scaled_l2norm.pair_cell}{263.7\% / 287.4\%}
\defmmnum{ws3.comp.whisper.gemmatwotogemmathree.procrustes.l2norm.clean}{423.92}
\defmmnum{ws3.comp.whisper.gemmatwotogemmathree.procrustes.l2norm.clean_cell}{423.92\%}
\defmmnum{ws3.comp.whisper.gemmatwotogemmathree.procrustes.l2norm.empty_clean}{96}
\defmmnum{ws3.comp.whisper.gemmatwotogemmathree.procrustes.l2norm.empty_other}{111}
\defmmnum{ws3.comp.whisper.gemmatwotogemmathree.procrustes.l2norm.other}{474.66}
\defmmnum{ws3.comp.whisper.gemmatwotogemmathree.procrustes.l2norm.other_cell}{474.66\%}
\defmmnum{ws3.comp.whisper.gemmatwotogemmathree.procrustes.l2norm.pair_cell}{423.9\% / 474.7\%}
\defmmnum{ws3.comp.whisper.gemmatwotogemmathree.procrustes.layernorm.clean}{413.83}
\defmmnum{ws3.comp.whisper.gemmatwotogemmathree.procrustes.layernorm.clean_cell}{413.83\%}
\defmmnum{ws3.comp.whisper.gemmatwotogemmathree.procrustes.layernorm.empty_clean}{0}
\defmmnum{ws3.comp.whisper.gemmatwotogemmathree.procrustes.layernorm.empty_other}{0}
\defmmnum{ws3.comp.whisper.gemmatwotogemmathree.procrustes.layernorm.other}{471.98}
\defmmnum{ws3.comp.whisper.gemmatwotogemmathree.procrustes.layernorm.other_cell}{471.98\%}
\defmmnum{ws3.comp.whisper.gemmatwotogemmathree.procrustes.layernorm.pair_cell}{413.8\% / 472.0\%}
\defmmnum{ws3.comp.whisper.gemmatwotogemmathree.procrustes.none.clean}{360.88}
\defmmnum{ws3.comp.whisper.gemmatwotogemmathree.procrustes.none.clean_cell}{360.88\%}
\defmmnum{ws3.comp.whisper.gemmatwotogemmathree.procrustes.none.empty_clean}{74}
\defmmnum{ws3.comp.whisper.gemmatwotogemmathree.procrustes.none.empty_other}{61}
\defmmnum{ws3.comp.whisper.gemmatwotogemmathree.procrustes.none.other}{400.16}
\defmmnum{ws3.comp.whisper.gemmatwotogemmathree.procrustes.none.other_cell}{400.16\%}
\defmmnum{ws3.comp.whisper.gemmatwotogemmathree.procrustes.none.pair_cell}{360.9\% / 400.2\%}
\defmmnum{ws3.comp.whisper.gemmatwotogemmathree.procrustes.rmsnorm.clean}{442.14}
\defmmnum{ws3.comp.whisper.gemmatwotogemmathree.procrustes.rmsnorm.clean_cell}{442.14\%}
\defmmnum{ws3.comp.whisper.gemmatwotogemmathree.procrustes.rmsnorm.empty_clean}{76}
\defmmnum{ws3.comp.whisper.gemmatwotogemmathree.procrustes.rmsnorm.empty_other}{63}
\defmmnum{ws3.comp.whisper.gemmatwotogemmathree.procrustes.rmsnorm.other}{505.04}
\defmmnum{ws3.comp.whisper.gemmatwotogemmathree.procrustes.rmsnorm.other_cell}{505.04\%}
\defmmnum{ws3.comp.whisper.gemmatwotogemmathree.procrustes.rmsnorm.pair_cell}{442.1\% / 505.0\%}
\defmmnum{ws3.comp.whisper.gemmatwotogemmathree.procrustes.scaled_l2norm.clean}{446.03}
\defmmnum{ws3.comp.whisper.gemmatwotogemmathree.procrustes.scaled_l2norm.clean_cell}{446.03\%}
\defmmnum{ws3.comp.whisper.gemmatwotogemmathree.procrustes.scaled_l2norm.empty_clean}{77}
\defmmnum{ws3.comp.whisper.gemmatwotogemmathree.procrustes.scaled_l2norm.empty_other}{72}
\defmmnum{ws3.comp.whisper.gemmatwotogemmathree.procrustes.scaled_l2norm.other}{513.17}
\defmmnum{ws3.comp.whisper.gemmatwotogemmathree.procrustes.scaled_l2norm.other_cell}{513.17\%}
\defmmnum{ws3.comp.whisper.gemmatwotogemmathree.procrustes.scaled_l2norm.pair_cell}{446.0\% / 513.2\%}
\defmmnum{ws3.comp.whisper.gemmatwotoqwen.lstsq.l2norm.clean}{147.55}
\defmmnum{ws3.comp.whisper.gemmatwotoqwen.lstsq.l2norm.clean_cell}{147.55\%}
\defmmnum{ws3.comp.whisper.gemmatwotoqwen.lstsq.l2norm.empty_clean}{0}
\defmmnum{ws3.comp.whisper.gemmatwotoqwen.lstsq.l2norm.empty_other}{0}
\defmmnum{ws3.comp.whisper.gemmatwotoqwen.lstsq.l2norm.other}{154.49}
\defmmnum{ws3.comp.whisper.gemmatwotoqwen.lstsq.l2norm.other_cell}{154.49\%}
\defmmnum{ws3.comp.whisper.gemmatwotoqwen.lstsq.l2norm.pair_cell}{147.5\% / 154.5\%}
\defmmnum{ws3.comp.whisper.gemmatwotoqwen.lstsq.layernorm.clean}{118.07}
\defmmnum{ws3.comp.whisper.gemmatwotoqwen.lstsq.layernorm.clean_cell}{118.07\%}
\defmmnum{ws3.comp.whisper.gemmatwotoqwen.lstsq.layernorm.empty_clean}{0}
\defmmnum{ws3.comp.whisper.gemmatwotoqwen.lstsq.layernorm.empty_other}{0}
\defmmnum{ws3.comp.whisper.gemmatwotoqwen.lstsq.layernorm.other}{123.88}
\defmmnum{ws3.comp.whisper.gemmatwotoqwen.lstsq.layernorm.other_cell}{123.88\%}
\defmmnum{ws3.comp.whisper.gemmatwotoqwen.lstsq.layernorm.pair_cell}{118.1\% / 123.9\%}
\defmmnum{ws3.comp.whisper.gemmatwotoqwen.lstsq.none.clean}{147.22}
\defmmnum{ws3.comp.whisper.gemmatwotoqwen.lstsq.none.clean_cell}{147.22\%}
\defmmnum{ws3.comp.whisper.gemmatwotoqwen.lstsq.none.empty_clean}{0}
\defmmnum{ws3.comp.whisper.gemmatwotoqwen.lstsq.none.empty_other}{0}
\defmmnum{ws3.comp.whisper.gemmatwotoqwen.lstsq.none.other}{153.17}
\defmmnum{ws3.comp.whisper.gemmatwotoqwen.lstsq.none.other_cell}{153.17\%}
\defmmnum{ws3.comp.whisper.gemmatwotoqwen.lstsq.none.pair_cell}{147.2\% / 153.2\%}
\defmmnum{ws3.comp.whisper.gemmatwotoqwen.lstsq.rmsnorm.clean}{162.17}
\defmmnum{ws3.comp.whisper.gemmatwotoqwen.lstsq.rmsnorm.clean_cell}{162.17\%}
\defmmnum{ws3.comp.whisper.gemmatwotoqwen.lstsq.rmsnorm.empty_clean}{0}
\defmmnum{ws3.comp.whisper.gemmatwotoqwen.lstsq.rmsnorm.empty_other}{0}
\defmmnum{ws3.comp.whisper.gemmatwotoqwen.lstsq.rmsnorm.other}{175.04}
\defmmnum{ws3.comp.whisper.gemmatwotoqwen.lstsq.rmsnorm.other_cell}{175.04\%}
\defmmnum{ws3.comp.whisper.gemmatwotoqwen.lstsq.rmsnorm.pair_cell}{162.2\% / 175.0\%}
\defmmnum{ws3.comp.whisper.gemmatwotoqwen.lstsq.scaled_l2norm.clean}{147.77}
\defmmnum{ws3.comp.whisper.gemmatwotoqwen.lstsq.scaled_l2norm.clean_cell}{147.77\%}
\defmmnum{ws3.comp.whisper.gemmatwotoqwen.lstsq.scaled_l2norm.empty_clean}{0}
\defmmnum{ws3.comp.whisper.gemmatwotoqwen.lstsq.scaled_l2norm.empty_other}{0}
\defmmnum{ws3.comp.whisper.gemmatwotoqwen.lstsq.scaled_l2norm.other}{154.49}
\defmmnum{ws3.comp.whisper.gemmatwotoqwen.lstsq.scaled_l2norm.other_cell}{154.49\%}
\defmmnum{ws3.comp.whisper.gemmatwotoqwen.lstsq.scaled_l2norm.pair_cell}{147.8\% / 154.5\%}
\defmmnum{ws3.comp.whisper.gemmatwotoqwen.procrustes.l2norm.clean}{154.92}
\defmmnum{ws3.comp.whisper.gemmatwotoqwen.procrustes.l2norm.clean_cell}{154.92\%}
\defmmnum{ws3.comp.whisper.gemmatwotoqwen.procrustes.l2norm.empty_clean}{0}
\defmmnum{ws3.comp.whisper.gemmatwotoqwen.procrustes.l2norm.empty_other}{0}
\defmmnum{ws3.comp.whisper.gemmatwotoqwen.procrustes.l2norm.other}{166.54}
\defmmnum{ws3.comp.whisper.gemmatwotoqwen.procrustes.l2norm.other_cell}{166.54\%}
\defmmnum{ws3.comp.whisper.gemmatwotoqwen.procrustes.l2norm.pair_cell}{154.9\% / 166.5\%}
\defmmnum{ws3.comp.whisper.gemmatwotoqwen.procrustes.layernorm.clean}{134.84}
\defmmnum{ws3.comp.whisper.gemmatwotoqwen.procrustes.layernorm.clean_cell}{134.84\%}
\defmmnum{ws3.comp.whisper.gemmatwotoqwen.procrustes.layernorm.empty_clean}{0}
\defmmnum{ws3.comp.whisper.gemmatwotoqwen.procrustes.layernorm.empty_other}{0}
\defmmnum{ws3.comp.whisper.gemmatwotoqwen.procrustes.layernorm.other}{141.08}
\defmmnum{ws3.comp.whisper.gemmatwotoqwen.procrustes.layernorm.other_cell}{141.08\%}
\defmmnum{ws3.comp.whisper.gemmatwotoqwen.procrustes.layernorm.pair_cell}{134.8\% / 141.1\%}
\defmmnum{ws3.comp.whisper.gemmatwotoqwen.procrustes.none.clean}{196.10}
\defmmnum{ws3.comp.whisper.gemmatwotoqwen.procrustes.none.clean_cell}{196.10\%}
\defmmnum{ws3.comp.whisper.gemmatwotoqwen.procrustes.none.empty_clean}{86}
\defmmnum{ws3.comp.whisper.gemmatwotoqwen.procrustes.none.empty_other}{98}
\defmmnum{ws3.comp.whisper.gemmatwotoqwen.procrustes.none.other}{215.26}
\defmmnum{ws3.comp.whisper.gemmatwotoqwen.procrustes.none.other_cell}{215.26\%}
\defmmnum{ws3.comp.whisper.gemmatwotoqwen.procrustes.none.pair_cell}{196.1\% / 215.3\%}
\defmmnum{ws3.comp.whisper.gemmatwotoqwen.procrustes.rmsnorm.clean}{161.09}
\defmmnum{ws3.comp.whisper.gemmatwotoqwen.procrustes.rmsnorm.clean_cell}{161.09\%}
\defmmnum{ws3.comp.whisper.gemmatwotoqwen.procrustes.rmsnorm.empty_clean}{0}
\defmmnum{ws3.comp.whisper.gemmatwotoqwen.procrustes.rmsnorm.empty_other}{0}
\defmmnum{ws3.comp.whisper.gemmatwotoqwen.procrustes.rmsnorm.other}{175.68}
\defmmnum{ws3.comp.whisper.gemmatwotoqwen.procrustes.rmsnorm.other_cell}{175.68\%}
\defmmnum{ws3.comp.whisper.gemmatwotoqwen.procrustes.rmsnorm.pair_cell}{161.1\% / 175.7\%}
\defmmnum{ws3.comp.whisper.gemmatwotoqwen.procrustes.scaled_l2norm.clean}{154.69}
\defmmnum{ws3.comp.whisper.gemmatwotoqwen.procrustes.scaled_l2norm.clean_cell}{154.69\%}
\defmmnum{ws3.comp.whisper.gemmatwotoqwen.procrustes.scaled_l2norm.empty_clean}{0}
\defmmnum{ws3.comp.whisper.gemmatwotoqwen.procrustes.scaled_l2norm.empty_other}{0}
\defmmnum{ws3.comp.whisper.gemmatwotoqwen.procrustes.scaled_l2norm.other}{167.46}
\defmmnum{ws3.comp.whisper.gemmatwotoqwen.procrustes.scaled_l2norm.other_cell}{167.46\%}
\defmmnum{ws3.comp.whisper.gemmatwotoqwen.procrustes.scaled_l2norm.pair_cell}{154.7\% / 167.5\%}
\defmmnum{ws3.comp.whisper.qwentogemmathree.lstsq.l2norm.clean}{311.09}
\defmmnum{ws3.comp.whisper.qwentogemmathree.lstsq.l2norm.clean_cell}{311.09\%}
\defmmnum{ws3.comp.whisper.qwentogemmathree.lstsq.l2norm.empty_clean}{306}
\defmmnum{ws3.comp.whisper.qwentogemmathree.lstsq.l2norm.empty_other}{249}
\defmmnum{ws3.comp.whisper.qwentogemmathree.lstsq.l2norm.other}{348.12}
\defmmnum{ws3.comp.whisper.qwentogemmathree.lstsq.l2norm.other_cell}{348.12\%}
\defmmnum{ws3.comp.whisper.qwentogemmathree.lstsq.l2norm.pair_cell}{311.1\% / 348.1\%}
\defmmnum{ws3.comp.whisper.qwentogemmathree.lstsq.layernorm.clean}{399.91}
\defmmnum{ws3.comp.whisper.qwentogemmathree.lstsq.layernorm.clean_cell}{399.91\%}
\defmmnum{ws3.comp.whisper.qwentogemmathree.lstsq.layernorm.empty_clean}{0}
\defmmnum{ws3.comp.whisper.qwentogemmathree.lstsq.layernorm.empty_other}{0}
\defmmnum{ws3.comp.whisper.qwentogemmathree.lstsq.layernorm.other}{446.50}
\defmmnum{ws3.comp.whisper.qwentogemmathree.lstsq.layernorm.other_cell}{446.50\%}
\defmmnum{ws3.comp.whisper.qwentogemmathree.lstsq.layernorm.pair_cell}{399.9\% / 446.5\%}
\defmmnum{ws3.comp.whisper.qwentogemmathree.lstsq.none.clean}{406.42}
\defmmnum{ws3.comp.whisper.qwentogemmathree.lstsq.none.clean_cell}{406.42\%}
\defmmnum{ws3.comp.whisper.qwentogemmathree.lstsq.none.empty_clean}{72}
\defmmnum{ws3.comp.whisper.qwentogemmathree.lstsq.none.empty_other}{62}
\defmmnum{ws3.comp.whisper.qwentogemmathree.lstsq.none.other}{463.66}
\defmmnum{ws3.comp.whisper.qwentogemmathree.lstsq.none.other_cell}{463.66\%}
\defmmnum{ws3.comp.whisper.qwentogemmathree.lstsq.none.pair_cell}{406.4\% / 463.7\%}
\defmmnum{ws3.comp.whisper.qwentogemmathree.lstsq.rmsnorm.clean}{330.71}
\defmmnum{ws3.comp.whisper.qwentogemmathree.lstsq.rmsnorm.clean_cell}{330.71\%}
\defmmnum{ws3.comp.whisper.qwentogemmathree.lstsq.rmsnorm.empty_clean}{182}
\defmmnum{ws3.comp.whisper.qwentogemmathree.lstsq.rmsnorm.empty_other}{81}
\defmmnum{ws3.comp.whisper.qwentogemmathree.lstsq.rmsnorm.other}{391.62}
\defmmnum{ws3.comp.whisper.qwentogemmathree.lstsq.rmsnorm.other_cell}{391.62\%}
\defmmnum{ws3.comp.whisper.qwentogemmathree.lstsq.rmsnorm.pair_cell}{330.7\% / 391.6\%}
\defmmnum{ws3.comp.whisper.qwentogemmathree.lstsq.scaled_l2norm.clean}{320.31}
\defmmnum{ws3.comp.whisper.qwentogemmathree.lstsq.scaled_l2norm.clean_cell}{320.31\%}
\defmmnum{ws3.comp.whisper.qwentogemmathree.lstsq.scaled_l2norm.empty_clean}{127}
\defmmnum{ws3.comp.whisper.qwentogemmathree.lstsq.scaled_l2norm.empty_other}{105}
\defmmnum{ws3.comp.whisper.qwentogemmathree.lstsq.scaled_l2norm.other}{361.32}
\defmmnum{ws3.comp.whisper.qwentogemmathree.lstsq.scaled_l2norm.other_cell}{361.32\%}
\defmmnum{ws3.comp.whisper.qwentogemmathree.lstsq.scaled_l2norm.pair_cell}{320.3\% / 361.3\%}
\defmmnum{ws3.comp.whisper.qwentogemmathree.procrustes.l2norm.clean}{173.08}
\defmmnum{ws3.comp.whisper.qwentogemmathree.procrustes.l2norm.clean_cell}{173.08\%}
\defmmnum{ws3.comp.whisper.qwentogemmathree.procrustes.l2norm.empty_clean}{1297}
\defmmnum{ws3.comp.whisper.qwentogemmathree.procrustes.l2norm.empty_other}{1260}
\defmmnum{ws3.comp.whisper.qwentogemmathree.procrustes.l2norm.other}{201.85}
\defmmnum{ws3.comp.whisper.qwentogemmathree.procrustes.l2norm.other_cell}{201.85\%}
\defmmnum{ws3.comp.whisper.qwentogemmathree.procrustes.l2norm.pair_cell}{173.1\% / 201.9\%}
\defmmnum{ws3.comp.whisper.qwentogemmathree.procrustes.layernorm.clean}{377.26}
\defmmnum{ws3.comp.whisper.qwentogemmathree.procrustes.layernorm.clean_cell}{377.26\%}
\defmmnum{ws3.comp.whisper.qwentogemmathree.procrustes.layernorm.empty_clean}{0}
\defmmnum{ws3.comp.whisper.qwentogemmathree.procrustes.layernorm.empty_other}{0}
\defmmnum{ws3.comp.whisper.qwentogemmathree.procrustes.layernorm.other}{440.87}
\defmmnum{ws3.comp.whisper.qwentogemmathree.procrustes.layernorm.other_cell}{440.87\%}
\defmmnum{ws3.comp.whisper.qwentogemmathree.procrustes.layernorm.pair_cell}{377.3\% / 440.9\%}
\defmmnum{ws3.comp.whisper.qwentogemmathree.procrustes.none.clean}{194.27}
\defmmnum{ws3.comp.whisper.qwentogemmathree.procrustes.none.clean_cell}{194.27\%}
\defmmnum{ws3.comp.whisper.qwentogemmathree.procrustes.none.empty_clean}{1244}
\defmmnum{ws3.comp.whisper.qwentogemmathree.procrustes.none.empty_other}{1183}
\defmmnum{ws3.comp.whisper.qwentogemmathree.procrustes.none.other}{231.51}
\defmmnum{ws3.comp.whisper.qwentogemmathree.procrustes.none.other_cell}{231.51\%}
\defmmnum{ws3.comp.whisper.qwentogemmathree.procrustes.none.pair_cell}{194.3\% / 231.5\%}
\defmmnum{ws3.comp.whisper.qwentogemmathree.procrustes.rmsnorm.clean}{180.26}
\defmmnum{ws3.comp.whisper.qwentogemmathree.procrustes.rmsnorm.clean_cell}{180.26\%}
\defmmnum{ws3.comp.whisper.qwentogemmathree.procrustes.rmsnorm.empty_clean}{1424}
\defmmnum{ws3.comp.whisper.qwentogemmathree.procrustes.rmsnorm.empty_other}{1366}
\defmmnum{ws3.comp.whisper.qwentogemmathree.procrustes.rmsnorm.other}{219.88}
\defmmnum{ws3.comp.whisper.qwentogemmathree.procrustes.rmsnorm.other_cell}{219.88\%}
\defmmnum{ws3.comp.whisper.qwentogemmathree.procrustes.rmsnorm.pair_cell}{180.3\% / 219.9\%}
\defmmnum{ws3.comp.whisper.qwentogemmathree.procrustes.scaled_l2norm.clean}{225.57}
\defmmnum{ws3.comp.whisper.qwentogemmathree.procrustes.scaled_l2norm.clean_cell}{225.57\%}
\defmmnum{ws3.comp.whisper.qwentogemmathree.procrustes.scaled_l2norm.empty_clean}{1166}
\defmmnum{ws3.comp.whisper.qwentogemmathree.procrustes.scaled_l2norm.empty_other}{1035}
\defmmnum{ws3.comp.whisper.qwentogemmathree.procrustes.scaled_l2norm.other}{272.42}
\defmmnum{ws3.comp.whisper.qwentogemmathree.procrustes.scaled_l2norm.other_cell}{272.42\%}
\defmmnum{ws3.comp.whisper.qwentogemmathree.procrustes.scaled_l2norm.pair_cell}{225.6\% / 272.4\%}
\defmmnum{ws3.comp.whisper.qwentogemmatwo.lstsq.l2norm.clean}{282.98}
\defmmnum{ws3.comp.whisper.qwentogemmatwo.lstsq.l2norm.clean_cell}{282.98\%}
\defmmnum{ws3.comp.whisper.qwentogemmatwo.lstsq.l2norm.empty_clean}{0}
\defmmnum{ws3.comp.whisper.qwentogemmatwo.lstsq.l2norm.empty_other}{0}
\defmmnum{ws3.comp.whisper.qwentogemmatwo.lstsq.l2norm.other}{326.88}
\defmmnum{ws3.comp.whisper.qwentogemmatwo.lstsq.l2norm.other_cell}{326.88\%}
\defmmnum{ws3.comp.whisper.qwentogemmatwo.lstsq.l2norm.pair_cell}{283.0\% / 326.9\%}
\defmmnum{ws3.comp.whisper.qwentogemmatwo.lstsq.layernorm.clean}{253.65}
\defmmnum{ws3.comp.whisper.qwentogemmatwo.lstsq.layernorm.clean_cell}{253.65\%}
\defmmnum{ws3.comp.whisper.qwentogemmatwo.lstsq.layernorm.empty_clean}{0}
\defmmnum{ws3.comp.whisper.qwentogemmatwo.lstsq.layernorm.empty_other}{0}
\defmmnum{ws3.comp.whisper.qwentogemmatwo.lstsq.layernorm.other}{281.75}
\defmmnum{ws3.comp.whisper.qwentogemmatwo.lstsq.layernorm.other_cell}{281.75\%}
\defmmnum{ws3.comp.whisper.qwentogemmatwo.lstsq.layernorm.pair_cell}{253.6\% / 281.8\%}
\defmmnum{ws3.comp.whisper.qwentogemmatwo.lstsq.none.clean}{111.81}
\defmmnum{ws3.comp.whisper.qwentogemmatwo.lstsq.none.clean_cell}{111.81\%}
\defmmnum{ws3.comp.whisper.qwentogemmatwo.lstsq.none.empty_clean}{2045}
\defmmnum{ws3.comp.whisper.qwentogemmatwo.lstsq.none.empty_other}{2197}
\defmmnum{ws3.comp.whisper.qwentogemmatwo.lstsq.none.other}{119.31}
\defmmnum{ws3.comp.whisper.qwentogemmatwo.lstsq.none.other_cell}{119.31\%}
\defmmnum{ws3.comp.whisper.qwentogemmatwo.lstsq.none.pair_cell}{111.8\% / 119.3\%}
\defmmnum{ws3.comp.whisper.qwentogemmatwo.lstsq.rmsnorm.clean}{378.87}
\defmmnum{ws3.comp.whisper.qwentogemmatwo.lstsq.rmsnorm.clean_cell}{378.87\%}
\defmmnum{ws3.comp.whisper.qwentogemmatwo.lstsq.rmsnorm.empty_clean}{80}
\defmmnum{ws3.comp.whisper.qwentogemmatwo.lstsq.rmsnorm.empty_other}{35}
\defmmnum{ws3.comp.whisper.qwentogemmatwo.lstsq.rmsnorm.other}{418.76}
\defmmnum{ws3.comp.whisper.qwentogemmatwo.lstsq.rmsnorm.other_cell}{418.76\%}
\defmmnum{ws3.comp.whisper.qwentogemmatwo.lstsq.rmsnorm.pair_cell}{378.9\% / 418.8\%}
\defmmnum{ws3.comp.whisper.qwentogemmatwo.lstsq.scaled_l2norm.clean}{359.95}
\defmmnum{ws3.comp.whisper.qwentogemmatwo.lstsq.scaled_l2norm.clean_cell}{359.95\%}
\defmmnum{ws3.comp.whisper.qwentogemmatwo.lstsq.scaled_l2norm.empty_clean}{0}
\defmmnum{ws3.comp.whisper.qwentogemmatwo.lstsq.scaled_l2norm.empty_other}{0}
\defmmnum{ws3.comp.whisper.qwentogemmatwo.lstsq.scaled_l2norm.other}{415.62}
\defmmnum{ws3.comp.whisper.qwentogemmatwo.lstsq.scaled_l2norm.other_cell}{415.62\%}
\defmmnum{ws3.comp.whisper.qwentogemmatwo.lstsq.scaled_l2norm.pair_cell}{359.9\% / 415.6\%}
\defmmnum{ws3.comp.whisper.qwentogemmatwo.procrustes.l2norm.clean}{224.63}
\defmmnum{ws3.comp.whisper.qwentogemmatwo.procrustes.l2norm.clean_cell}{224.63\%}
\defmmnum{ws3.comp.whisper.qwentogemmatwo.procrustes.l2norm.empty_clean}{1417}
\defmmnum{ws3.comp.whisper.qwentogemmatwo.procrustes.l2norm.empty_other}{1569}
\defmmnum{ws3.comp.whisper.qwentogemmatwo.procrustes.l2norm.other}{253.39}
\defmmnum{ws3.comp.whisper.qwentogemmatwo.procrustes.l2norm.other_cell}{253.39\%}
\defmmnum{ws3.comp.whisper.qwentogemmatwo.procrustes.l2norm.pair_cell}{224.6\% / 253.4\%}
\defmmnum{ws3.comp.whisper.qwentogemmatwo.procrustes.layernorm.clean}{277.06}
\defmmnum{ws3.comp.whisper.qwentogemmatwo.procrustes.layernorm.clean_cell}{277.06\%}
\defmmnum{ws3.comp.whisper.qwentogemmatwo.procrustes.layernorm.empty_clean}{0}
\defmmnum{ws3.comp.whisper.qwentogemmatwo.procrustes.layernorm.empty_other}{0}
\defmmnum{ws3.comp.whisper.qwentogemmatwo.procrustes.layernorm.other}{318.30}
\defmmnum{ws3.comp.whisper.qwentogemmatwo.procrustes.layernorm.other_cell}{318.30\%}
\defmmnum{ws3.comp.whisper.qwentogemmatwo.procrustes.layernorm.pair_cell}{277.1\% / 318.3\%}
\defmmnum{ws3.comp.whisper.qwentogemmatwo.procrustes.none.clean}{251.68}
\defmmnum{ws3.comp.whisper.qwentogemmatwo.procrustes.none.clean_cell}{251.68\%}
\defmmnum{ws3.comp.whisper.qwentogemmatwo.procrustes.none.empty_clean}{1005}
\defmmnum{ws3.comp.whisper.qwentogemmatwo.procrustes.none.empty_other}{1069}
\defmmnum{ws3.comp.whisper.qwentogemmatwo.procrustes.none.other}{287.42}
\defmmnum{ws3.comp.whisper.qwentogemmatwo.procrustes.none.other_cell}{287.42\%}
\defmmnum{ws3.comp.whisper.qwentogemmatwo.procrustes.none.pair_cell}{251.7\% / 287.4\%}
\defmmnum{ws3.comp.whisper.qwentogemmatwo.procrustes.rmsnorm.clean}{153.44}
\defmmnum{ws3.comp.whisper.qwentogemmatwo.procrustes.rmsnorm.clean_cell}{153.44\%}
\defmmnum{ws3.comp.whisper.qwentogemmatwo.procrustes.rmsnorm.empty_clean}{2244}
\defmmnum{ws3.comp.whisper.qwentogemmatwo.procrustes.rmsnorm.empty_other}{2566}
\defmmnum{ws3.comp.whisper.qwentogemmatwo.procrustes.rmsnorm.other}{154.99}
\defmmnum{ws3.comp.whisper.qwentogemmatwo.procrustes.rmsnorm.other_cell}{154.99\%}
\defmmnum{ws3.comp.whisper.qwentogemmatwo.procrustes.rmsnorm.pair_cell}{153.4\% / 155.0\%}
\defmmnum{ws3.comp.whisper.qwentogemmatwo.procrustes.scaled_l2norm.clean}{345.77}
\defmmnum{ws3.comp.whisper.qwentogemmatwo.procrustes.scaled_l2norm.clean_cell}{345.77\%}
\defmmnum{ws3.comp.whisper.qwentogemmatwo.procrustes.scaled_l2norm.empty_clean}{406}
\defmmnum{ws3.comp.whisper.qwentogemmatwo.procrustes.scaled_l2norm.empty_other}{259}
\defmmnum{ws3.comp.whisper.qwentogemmatwo.procrustes.scaled_l2norm.other}{409.78}
\defmmnum{ws3.comp.whisper.qwentogemmatwo.procrustes.scaled_l2norm.other_cell}{409.78\%}
\defmmnum{ws3.comp.whisper.qwentogemmatwo.procrustes.scaled_l2norm.pair_cell}{345.8\% / 409.8\%}
\defmmnum{ws3.comp_adamw.moonshine.gemmatwotogemmathree.lstsq.none.clean}{285.04}
\defmmnum{ws3.comp_adamw.moonshine.gemmatwotogemmathree.lstsq.none.other}{316.20}
\defmmnum{ws3.comp_adamw.moonshine.gemmatwotogemmathree.lstsq.scaled_l2norm.clean}{283.91}
\defmmnum{ws3.comp_adamw.moonshine.gemmatwotogemmathree.lstsq.scaled_l2norm.other}{321.95}
\defmmnum{ws3.comp_adamw.moonshine.gemmatwotoqwen.lstsq.none.clean}{123.58}
\defmmnum{ws3.comp_adamw.moonshine.gemmatwotoqwen.lstsq.none.other}{144.66}
\defmmnum{ws3.comp_adamw.moonshine.gemmatwotoqwen.lstsq.scaled_l2norm.clean}{123.66}
\defmmnum{ws3.comp_adamw.moonshine.gemmatwotoqwen.lstsq.scaled_l2norm.other}{145.83}
\defmmnum{ws3.delta.moonshine.g2_to_g3.lstsq_scaled_vs_proc_scaled.clean}{-85.84}
\defmmnum{ws3.delta.moonshine.g2_to_qwen.lstsq_ln_vs_proc_ln.clean}{-74.79}
\defmmnum{ws3.delta.moonshine.g2_to_qwen.lstsq_ln_vs_proc_scaled.clean}{-51.08}
\defmmnum{ws3.delta.moonshine.g2_to_qwen.lstsq_scaled_vs_proc_scaled.clean}{-28.69}
\defmmnum{ws3.fit.gemmatwotogemmathree.lstsq.anchors}{163{,}983}
\defmmnum{ws3.fit.gemmatwotogemmathree.lstsq.cos}{0.5768}
\defmmnum{ws3.fit.gemmatwotogemmathree.lstsq.n_test}{32{,}796}
\defmmnum{ws3.fit.gemmatwotogemmathree.lstsq.norm_rr}{1.000}
\defmmnum{ws3.fit.gemmatwotogemmathree.lstsq.p1}{98.56}
\defmmnum{ws3.fit.gemmatwotogemmathree.lstsq.raw_rr}{0.603}
\defmmnum{ws3.fit.gemmatwotogemmathree.procrustes.anchors}{163{,}983}
\defmmnum{ws3.fit.gemmatwotogemmathree.procrustes.cos}{0.5080}
\defmmnum{ws3.fit.gemmatwotogemmathree.procrustes.n_test}{32{,}796}
\defmmnum{ws3.fit.gemmatwotogemmathree.procrustes.norm_rr}{1.000}
\defmmnum{ws3.fit.gemmatwotogemmathree.procrustes.p1}{97.47}
\defmmnum{ws3.fit.gemmatwotogemmathree.procrustes.raw_rr}{1.644}
\defmmnum{ws3.fit.gemmatwotogemmathree.random_orthogonal.cos}{0.0005}
\defmmnum{ws3.fit.gemmatwotogemmathree.random_orthogonal.p1}{0.00}
\defmmnum{ws3.fit.gemmatwotoqwen.lstsq.anchors}{83{,}378}
\defmmnum{ws3.fit.gemmatwotoqwen.lstsq.cos}{0.6283}
\defmmnum{ws3.fit.gemmatwotoqwen.lstsq.n_test}{16{,}675}
\defmmnum{ws3.fit.gemmatwotoqwen.lstsq.norm_rr}{1.000}
\defmmnum{ws3.fit.gemmatwotoqwen.lstsq.p1}{84.57}
\defmmnum{ws3.fit.gemmatwotoqwen.lstsq.raw_rr}{0.661}
\defmmnum{ws3.fit.gemmatwotoqwen.procrustes.anchors}{83{,}378}
\defmmnum{ws3.fit.gemmatwotoqwen.procrustes.cos}{0.5645}
\defmmnum{ws3.fit.gemmatwotoqwen.procrustes.n_test}{16{,}675}
\defmmnum{ws3.fit.gemmatwotoqwen.procrustes.norm_rr}{1.000}
\defmmnum{ws3.fit.gemmatwotoqwen.procrustes.p1}{83.09}
\defmmnum{ws3.fit.gemmatwotoqwen.procrustes.raw_rr}{70.48}
\defmmnum{ws3.fit.gemmatwotoqwen.random_orthogonal.cos}{-0.0041}
\defmmnum{ws3.fit.gemmatwotoqwen.random_orthogonal.p1}{0.00}
\defmmnum{ws3.fit.qwentogemmathree.lstsq.anchors}{82{,}663}
\defmmnum{ws3.fit.qwentogemmathree.lstsq.cos}{0.5515}
\defmmnum{ws3.fit.qwentogemmathree.lstsq.n_test}{16{,}532}
\defmmnum{ws3.fit.qwentogemmathree.lstsq.norm_rr}{1.000}
\defmmnum{ws3.fit.qwentogemmathree.lstsq.p1}{96.95}
\defmmnum{ws3.fit.qwentogemmathree.lstsq.raw_rr}{0.585}
\defmmnum{ws3.fit.qwentogemmathree.procrustes.anchors}{82{,}663}
\defmmnum{ws3.fit.qwentogemmathree.procrustes.cos}{0.5010}
\defmmnum{ws3.fit.qwentogemmathree.procrustes.n_test}{16{,}532}
\defmmnum{ws3.fit.qwentogemmathree.procrustes.norm_rr}{1.000}
\defmmnum{ws3.fit.qwentogemmathree.procrustes.p1}{96.72}
\defmmnum{ws3.fit.qwentogemmathree.procrustes.raw_rr}{0.020}
\defmmnum{ws3.fit.qwentogemmatwo.lstsq.anchors}{83{,}378}
\defmmnum{ws3.fit.qwentogemmatwo.lstsq.cos}{0.5916}
\defmmnum{ws3.fit.qwentogemmatwo.lstsq.n_test}{16{,}675}
\defmmnum{ws3.fit.qwentogemmatwo.lstsq.norm_rr}{1.008}
\defmmnum{ws3.fit.qwentogemmatwo.lstsq.p1}{79.41}
\defmmnum{ws3.fit.qwentogemmatwo.lstsq.raw_rr}{0.615}
\defmmnum{ws3.fit.qwentogemmatwo.procrustes.anchors}{83{,}378}
\defmmnum{ws3.fit.qwentogemmatwo.procrustes.cos}{0.5048}
\defmmnum{ws3.fit.qwentogemmatwo.procrustes.n_test}{16{,}675}
\defmmnum{ws3.fit.qwentogemmatwo.procrustes.norm_rr}{1.008}
\defmmnum{ws3.fit.qwentogemmatwo.procrustes.p1}{91.68}
\defmmnum{ws3.fit.qwentogemmatwo.procrustes.raw_rr}{0.013}
\defmmnum{ws3.fits.lstsq_done}{4}
\defmmnum{ws3.fits.max_anchors}{163{,}983}
\defmmnum{ws3.fits.max_cos}{0.5645}
\defmmnum{ws3.fits.max_p1}{97.47}
\defmmnum{ws3.fits.min_anchors}{82{,}663}
\defmmnum{ws3.fits.min_cos}{0.5010}
\defmmnum{ws3.fits.min_p1}{83.09}
\defmmnum{ws3.fits.procrustes_done}{4}
\defmmnum{ws3.n_pairs}{4}
\defmmnum{ws3.procrustes_total_cells}{16}
\defmmnum{ws4.comp.best_clean}{100.50}
\defmmnum{ws4.comp.best_other}{100.83}
\defmmnum{ws4.comp.done}{16}
\defmmnum{ws4.comp.moonshine.gemmatwotogemmathree.layernorm.clean}{394.50}
\defmmnum{ws4.comp.moonshine.gemmatwotogemmathree.layernorm.clean_cell}{394.50\%}
\defmmnum{ws4.comp.moonshine.gemmatwotogemmathree.layernorm.empty_clean}{0}
\defmmnum{ws4.comp.moonshine.gemmatwotogemmathree.layernorm.empty_other}{0}
\defmmnum{ws4.comp.moonshine.gemmatwotogemmathree.layernorm.other}{443.33}
\defmmnum{ws4.comp.moonshine.gemmatwotogemmathree.layernorm.other_cell}{443.33\%}
\defmmnum{ws4.comp.moonshine.gemmatwotogemmathree.layernorm.pair_cell}{394.5\% / 443.3\%}
\defmmnum{ws4.comp.moonshine.gemmatwotogemmathree.scaled_l2norm.clean}{388.80}
\defmmnum{ws4.comp.moonshine.gemmatwotogemmathree.scaled_l2norm.clean_cell}{388.80\%}
\defmmnum{ws4.comp.moonshine.gemmatwotogemmathree.scaled_l2norm.empty_clean}{0}
\defmmnum{ws4.comp.moonshine.gemmatwotogemmathree.scaled_l2norm.empty_other}{0}
\defmmnum{ws4.comp.moonshine.gemmatwotogemmathree.scaled_l2norm.other}{447.59}
\defmmnum{ws4.comp.moonshine.gemmatwotogemmathree.scaled_l2norm.other_cell}{447.59\%}
\defmmnum{ws4.comp.moonshine.gemmatwotogemmathree.scaled_l2norm.pair_cell}{388.8\% / 447.6\%}
\defmmnum{ws4.comp.moonshine.gemmatwotoqwen.layernorm.clean}{100.70}
\defmmnum{ws4.comp.moonshine.gemmatwotoqwen.layernorm.clean_cell}{100.70\%}
\defmmnum{ws4.comp.moonshine.gemmatwotoqwen.layernorm.empty_clean}{0}
\defmmnum{ws4.comp.moonshine.gemmatwotoqwen.layernorm.empty_other}{0}
\defmmnum{ws4.comp.moonshine.gemmatwotoqwen.layernorm.other}{102.22}
\defmmnum{ws4.comp.moonshine.gemmatwotoqwen.layernorm.other_cell}{102.22\%}
\defmmnum{ws4.comp.moonshine.gemmatwotoqwen.layernorm.pair_cell}{100.7\% / 102.2\%}
\defmmnum{ws4.comp.moonshine.gemmatwotoqwen.scaled_l2norm.clean}{116.28}
\defmmnum{ws4.comp.moonshine.gemmatwotoqwen.scaled_l2norm.clean_cell}{116.28\%}
\defmmnum{ws4.comp.moonshine.gemmatwotoqwen.scaled_l2norm.empty_clean}{0}
\defmmnum{ws4.comp.moonshine.gemmatwotoqwen.scaled_l2norm.empty_other}{0}
\defmmnum{ws4.comp.moonshine.gemmatwotoqwen.scaled_l2norm.other}{114.48}
\defmmnum{ws4.comp.moonshine.gemmatwotoqwen.scaled_l2norm.other_cell}{114.48\%}
\defmmnum{ws4.comp.moonshine.gemmatwotoqwen.scaled_l2norm.pair_cell}{116.3\% / 114.5\%}
\defmmnum{ws4.comp.moonshine.qwentogemmathree.layernorm.clean}{407.30}
\defmmnum{ws4.comp.moonshine.qwentogemmathree.layernorm.clean_cell}{407.30\%}
\defmmnum{ws4.comp.moonshine.qwentogemmathree.layernorm.empty_clean}{0}
\defmmnum{ws4.comp.moonshine.qwentogemmathree.layernorm.empty_other}{0}
\defmmnum{ws4.comp.moonshine.qwentogemmathree.layernorm.other}{458.72}
\defmmnum{ws4.comp.moonshine.qwentogemmathree.layernorm.other_cell}{458.72\%}
\defmmnum{ws4.comp.moonshine.qwentogemmathree.layernorm.pair_cell}{407.3\% / 458.7\%}
\defmmnum{ws4.comp.moonshine.qwentogemmathree.scaled_l2norm.clean}{236.57}
\defmmnum{ws4.comp.moonshine.qwentogemmathree.scaled_l2norm.clean_cell}{236.57\%}
\defmmnum{ws4.comp.moonshine.qwentogemmathree.scaled_l2norm.empty_clean}{1160}
\defmmnum{ws4.comp.moonshine.qwentogemmathree.scaled_l2norm.empty_other}{1170}
\defmmnum{ws4.comp.moonshine.qwentogemmathree.scaled_l2norm.other}{270.31}
\defmmnum{ws4.comp.moonshine.qwentogemmathree.scaled_l2norm.other_cell}{270.31\%}
\defmmnum{ws4.comp.moonshine.qwentogemmathree.scaled_l2norm.pair_cell}{236.6\% / 270.3\%}
\defmmnum{ws4.comp.moonshine.qwentogemmatwo.layernorm.clean}{341.85}
\defmmnum{ws4.comp.moonshine.qwentogemmatwo.layernorm.clean_cell}{341.85\%}
\defmmnum{ws4.comp.moonshine.qwentogemmatwo.layernorm.empty_clean}{10}
\defmmnum{ws4.comp.moonshine.qwentogemmatwo.layernorm.empty_other}{4}
\defmmnum{ws4.comp.moonshine.qwentogemmatwo.layernorm.other}{390.01}
\defmmnum{ws4.comp.moonshine.qwentogemmatwo.layernorm.other_cell}{390.01\%}
\defmmnum{ws4.comp.moonshine.qwentogemmatwo.layernorm.pair_cell}{341.9\% / 390.0\%}
\defmmnum{ws4.comp.moonshine.qwentogemmatwo.scaled_l2norm.clean}{189.47}
\defmmnum{ws4.comp.moonshine.qwentogemmatwo.scaled_l2norm.clean_cell}{189.47\%}
\defmmnum{ws4.comp.moonshine.qwentogemmatwo.scaled_l2norm.empty_clean}{1953}
\defmmnum{ws4.comp.moonshine.qwentogemmatwo.scaled_l2norm.empty_other}{2281}
\defmmnum{ws4.comp.moonshine.qwentogemmatwo.scaled_l2norm.other}{181.17}
\defmmnum{ws4.comp.moonshine.qwentogemmatwo.scaled_l2norm.other_cell}{181.17\%}
\defmmnum{ws4.comp.moonshine.qwentogemmatwo.scaled_l2norm.pair_cell}{189.5\% / 181.2\%}
\defmmnum{ws4.comp.total}{16}
\defmmnum{ws4.comp.whisper.gemmatwotogemmathree.layernorm.clean}{374.90}
\defmmnum{ws4.comp.whisper.gemmatwotogemmathree.layernorm.clean_cell}{374.90\%}
\defmmnum{ws4.comp.whisper.gemmatwotogemmathree.layernorm.empty_clean}{0}
\defmmnum{ws4.comp.whisper.gemmatwotogemmathree.layernorm.empty_other}{0}
\defmmnum{ws4.comp.whisper.gemmatwotogemmathree.layernorm.other}{427.78}
\defmmnum{ws4.comp.whisper.gemmatwotogemmathree.layernorm.other_cell}{427.78\%}
\defmmnum{ws4.comp.whisper.gemmatwotogemmathree.layernorm.pair_cell}{374.9\% / 427.8\%}
\defmmnum{ws4.comp.whisper.gemmatwotogemmathree.scaled_l2norm.clean}{400.72}
\defmmnum{ws4.comp.whisper.gemmatwotogemmathree.scaled_l2norm.clean_cell}{400.72\%}
\defmmnum{ws4.comp.whisper.gemmatwotogemmathree.scaled_l2norm.empty_clean}{0}
\defmmnum{ws4.comp.whisper.gemmatwotogemmathree.scaled_l2norm.empty_other}{0}
\defmmnum{ws4.comp.whisper.gemmatwotogemmathree.scaled_l2norm.other}{456.93}
\defmmnum{ws4.comp.whisper.gemmatwotogemmathree.scaled_l2norm.other_cell}{456.93\%}
\defmmnum{ws4.comp.whisper.gemmatwotogemmathree.scaled_l2norm.pair_cell}{400.7\% / 456.9\%}
\defmmnum{ws4.comp.whisper.gemmatwotoqwen.layernorm.clean}{101.37}
\defmmnum{ws4.comp.whisper.gemmatwotoqwen.layernorm.clean_cell}{101.37\%}
\defmmnum{ws4.comp.whisper.gemmatwotoqwen.layernorm.empty_clean}{0}
\defmmnum{ws4.comp.whisper.gemmatwotoqwen.layernorm.empty_other}{0}
\defmmnum{ws4.comp.whisper.gemmatwotoqwen.layernorm.other}{102.13}
\defmmnum{ws4.comp.whisper.gemmatwotoqwen.layernorm.other_cell}{102.13\%}
\defmmnum{ws4.comp.whisper.gemmatwotoqwen.layernorm.pair_cell}{101.4\% / 102.1\%}
\defmmnum{ws4.comp.whisper.gemmatwotoqwen.scaled_l2norm.clean}{100.50}
\defmmnum{ws4.comp.whisper.gemmatwotoqwen.scaled_l2norm.clean_cell}{100.50\%}
\defmmnum{ws4.comp.whisper.gemmatwotoqwen.scaled_l2norm.empty_clean}{0}
\defmmnum{ws4.comp.whisper.gemmatwotoqwen.scaled_l2norm.empty_other}{0}
\defmmnum{ws4.comp.whisper.gemmatwotoqwen.scaled_l2norm.other}{100.83}
\defmmnum{ws4.comp.whisper.gemmatwotoqwen.scaled_l2norm.other_cell}{100.83\%}
\defmmnum{ws4.comp.whisper.gemmatwotoqwen.scaled_l2norm.pair_cell}{100.5\% / 100.8\%}
\defmmnum{ws4.comp.whisper.qwentogemmathree.layernorm.clean}{442.77}
\defmmnum{ws4.comp.whisper.qwentogemmathree.layernorm.clean_cell}{442.77\%}
\defmmnum{ws4.comp.whisper.qwentogemmathree.layernorm.empty_clean}{0}
\defmmnum{ws4.comp.whisper.qwentogemmathree.layernorm.empty_other}{0}
\defmmnum{ws4.comp.whisper.qwentogemmathree.layernorm.other}{506.34}
\defmmnum{ws4.comp.whisper.qwentogemmathree.layernorm.other_cell}{506.34\%}
\defmmnum{ws4.comp.whisper.qwentogemmathree.layernorm.pair_cell}{442.8\% / 506.3\%}
\defmmnum{ws4.comp.whisper.qwentogemmathree.scaled_l2norm.clean}{317.60}
\defmmnum{ws4.comp.whisper.qwentogemmathree.scaled_l2norm.clean_cell}{317.60\%}
\defmmnum{ws4.comp.whisper.qwentogemmathree.scaled_l2norm.empty_clean}{691}
\defmmnum{ws4.comp.whisper.qwentogemmathree.scaled_l2norm.empty_other}{689}
\defmmnum{ws4.comp.whisper.qwentogemmathree.scaled_l2norm.other}{359.28}
\defmmnum{ws4.comp.whisper.qwentogemmathree.scaled_l2norm.other_cell}{359.28\%}
\defmmnum{ws4.comp.whisper.qwentogemmathree.scaled_l2norm.pair_cell}{317.6\% / 359.3\%}
\defmmnum{ws4.comp.whisper.qwentogemmatwo.layernorm.clean}{477.13}
\defmmnum{ws4.comp.whisper.qwentogemmatwo.layernorm.clean_cell}{477.13\%}
\defmmnum{ws4.comp.whisper.qwentogemmatwo.layernorm.empty_clean}{0}
\defmmnum{ws4.comp.whisper.qwentogemmatwo.layernorm.empty_other}{0}
\defmmnum{ws4.comp.whisper.qwentogemmatwo.layernorm.other}{532.68}
\defmmnum{ws4.comp.whisper.qwentogemmatwo.layernorm.other_cell}{532.68\%}
\defmmnum{ws4.comp.whisper.qwentogemmatwo.layernorm.pair_cell}{477.1\% / 532.7\%}
\defmmnum{ws4.comp.whisper.qwentogemmatwo.scaled_l2norm.clean}{490.55}
\defmmnum{ws4.comp.whisper.qwentogemmatwo.scaled_l2norm.clean_cell}{490.55\%}
\defmmnum{ws4.comp.whisper.qwentogemmatwo.scaled_l2norm.empty_clean}{25}
\defmmnum{ws4.comp.whisper.qwentogemmatwo.scaled_l2norm.empty_other}{12}
\defmmnum{ws4.comp.whisper.qwentogemmatwo.scaled_l2norm.other}{555.23}
\defmmnum{ws4.comp.whisper.qwentogemmatwo.scaled_l2norm.other_cell}{555.23\%}
\defmmnum{ws4.comp.whisper.qwentogemmatwo.scaled_l2norm.pair_cell}{490.5\% / 555.2\%}
\defmmnum{ws4.delta.moonshine.g2_to_qwen.wte_ln_vs_lstsq_ln.clean}{-7.61}
\defmmnum{ws4.delta.moonshine.g2_to_qwen.wte_ln_vs_lstsq_ln.other}{-11.38}
\defmmnum{ws4.delta.moonshine.g2_to_qwen.wte_scaled_vs_lstsq_scaled.clean}{-14.42}
\defmmnum{ws4.fit.gemmatwotogemmathree.layernorm.anchors}{163{,}983}
\defmmnum{ws4.fit.gemmatwotogemmathree.layernorm.cos}{0.1136}
\defmmnum{ws4.fit.gemmatwotogemmathree.layernorm.n_test}{32{,}796}
\defmmnum{ws4.fit.gemmatwotogemmathree.layernorm.p1}{36.11}
\defmmnum{ws4.fit.gemmatwotogemmathree.layernorm.params}{5{,}910{,}528}
\defmmnum{ws4.fit.gemmatwotogemmathree.layernorm.params_m}{5.91M}
\defmmnum{ws4.fit.gemmatwotogemmathree.layernorm.rr}{1.028}
\defmmnum{ws4.fit.gemmatwotogemmathree.layernorm.val}{0.0043}
\defmmnum{ws4.fit.gemmatwotogemmathree.scaled_l2norm.anchors}{163{,}983}
\defmmnum{ws4.fit.gemmatwotogemmathree.scaled_l2norm.cos}{0.1125}
\defmmnum{ws4.fit.gemmatwotogemmathree.scaled_l2norm.n_test}{32{,}796}
\defmmnum{ws4.fit.gemmatwotogemmathree.scaled_l2norm.p1}{37.47}
\defmmnum{ws4.fit.gemmatwotogemmathree.scaled_l2norm.params}{5{,}900{,}802}
\defmmnum{ws4.fit.gemmatwotogemmathree.scaled_l2norm.params_m}{5.90M}
\defmmnum{ws4.fit.gemmatwotogemmathree.scaled_l2norm.rr}{1.005}
\defmmnum{ws4.fit.gemmatwotogemmathree.scaled_l2norm.val}{0.0065}
\defmmnum{ws4.fit.gemmatwotoqwen.layernorm.anchors}{83{,}378}
\defmmnum{ws4.fit.gemmatwotoqwen.layernorm.cos}{0.1648}
\defmmnum{ws4.fit.gemmatwotoqwen.layernorm.n_test}{16{,}675}
\defmmnum{ws4.fit.gemmatwotoqwen.layernorm.p1}{13.05}
\defmmnum{ws4.fit.gemmatwotoqwen.layernorm.params}{3{,}548{,}160}
\defmmnum{ws4.fit.gemmatwotoqwen.layernorm.params_m}{3.55M}
\defmmnum{ws4.fit.gemmatwotoqwen.layernorm.rr}{3.657}
\defmmnum{ws4.fit.gemmatwotoqwen.layernorm.val}{0.0183}
\defmmnum{ws4.fit.gemmatwotoqwen.scaled_l2norm.anchors}{83{,}378}
\defmmnum{ws4.fit.gemmatwotoqwen.scaled_l2norm.cos}{0.1557}
\defmmnum{ws4.fit.gemmatwotoqwen.scaled_l2norm.n_test}{16{,}675}
\defmmnum{ws4.fit.gemmatwotoqwen.scaled_l2norm.p1}{33.66}
\defmmnum{ws4.fit.gemmatwotoqwen.scaled_l2norm.params}{3{,}540{,}482}
\defmmnum{ws4.fit.gemmatwotoqwen.scaled_l2norm.params_m}{3.54M}
\defmmnum{ws4.fit.gemmatwotoqwen.scaled_l2norm.rr}{3.406}
\defmmnum{ws4.fit.gemmatwotoqwen.scaled_l2norm.val}{0.0185}
\defmmnum{ws4.fit.qwentogemmathree.layernorm.anchors}{82{,}663}
\defmmnum{ws4.fit.qwentogemmathree.layernorm.cos}{0.1150}
\defmmnum{ws4.fit.qwentogemmathree.layernorm.n_test}{16{,}532}
\defmmnum{ws4.fit.qwentogemmathree.layernorm.p1}{48.14}
\defmmnum{ws4.fit.qwentogemmathree.layernorm.params}{3{,}942{,}912}
\defmmnum{ws4.fit.qwentogemmathree.layernorm.params_m}{3.94M}
\defmmnum{ws4.fit.qwentogemmathree.layernorm.rr}{1.024}
\defmmnum{ws4.fit.qwentogemmathree.layernorm.val}{0.0062}
\defmmnum{ws4.fit.qwentogemmathree.scaled_l2norm.anchors}{82{,}663}
\defmmnum{ws4.fit.qwentogemmathree.scaled_l2norm.cos}{0.1113}
\defmmnum{ws4.fit.qwentogemmathree.scaled_l2norm.n_test}{16{,}532}
\defmmnum{ws4.fit.qwentogemmathree.scaled_l2norm.p1}{53.90}
\defmmnum{ws4.fit.qwentogemmathree.scaled_l2norm.params}{3{,}934{,}722}
\defmmnum{ws4.fit.qwentogemmathree.scaled_l2norm.params_m}{3.93M}
\defmmnum{ws4.fit.qwentogemmathree.scaled_l2norm.rr}{1.003}
\defmmnum{ws4.fit.qwentogemmathree.scaled_l2norm.val}{0.0097}
\defmmnum{ws4.fit.qwentogemmatwo.layernorm.anchors}{83{,}378}
\defmmnum{ws4.fit.qwentogemmatwo.layernorm.cos}{0.1232}
\defmmnum{ws4.fit.qwentogemmatwo.layernorm.n_test}{16{,}675}
\defmmnum{ws4.fit.qwentogemmatwo.layernorm.p1}{50.46}
\defmmnum{ws4.fit.qwentogemmatwo.layernorm.params}{3{,}548{,}928}
\defmmnum{ws4.fit.qwentogemmatwo.layernorm.params_m}{3.55M}
\defmmnum{ws4.fit.qwentogemmatwo.layernorm.rr}{1.032}
\defmmnum{ws4.fit.qwentogemmatwo.layernorm.val}{0.0101}
\defmmnum{ws4.fit.qwentogemmatwo.scaled_l2norm.anchors}{83{,}378}
\defmmnum{ws4.fit.qwentogemmatwo.scaled_l2norm.cos}{0.1022}
\defmmnum{ws4.fit.qwentogemmatwo.scaled_l2norm.n_test}{16{,}675}
\defmmnum{ws4.fit.qwentogemmatwo.scaled_l2norm.p1}{52.14}
\defmmnum{ws4.fit.qwentogemmatwo.scaled_l2norm.params}{3{,}541{,}250}
\defmmnum{ws4.fit.qwentogemmatwo.scaled_l2norm.params_m}{3.54M}
\defmmnum{ws4.fit.qwentogemmatwo.scaled_l2norm.rr}{1.000}
\defmmnum{ws4.fit.qwentogemmatwo.scaled_l2norm.val}{0.0119}
\defmmnum{ws4.fits.done}{8}
\defmmnum{ws4.fits.max_cos}{0.1648}
\defmmnum{ws4.fits.max_p1}{53.90}
\defmmnum{ws4.fits.max_val}{0.0185}
\defmmnum{ws4.fits.min_cos}{0.1022}
\defmmnum{ws4.fits.min_p1}{13.05}
\defmmnum{ws4.fits.min_val}{0.0043}
\defmmnum{ws4.fits.total}{8}
\defmmnum{ws4b.comp.best_clean}{3.22}
\defmmnum{ws4b.comp.best_other}{7.53}
\defmmnum{ws4b.comp.cross.moonshine_on_whisper.gemmatwotogemmathree.layernorm.clean}{75.35}
\defmmnum{ws4b.comp.cross.moonshine_on_whisper.gemmatwotogemmathree.layernorm.clean_cell}{75.35\%}
\defmmnum{ws4b.comp.cross.moonshine_on_whisper.gemmatwotogemmathree.layernorm.empty_clean}{0}
\defmmnum{ws4b.comp.cross.moonshine_on_whisper.gemmatwotogemmathree.layernorm.empty_other}{0}
\defmmnum{ws4b.comp.cross.moonshine_on_whisper.gemmatwotogemmathree.layernorm.other}{89.82}
\defmmnum{ws4b.comp.cross.moonshine_on_whisper.gemmatwotogemmathree.layernorm.other_cell}{89.82\%}
\defmmnum{ws4b.comp.cross.moonshine_on_whisper.gemmatwotogemmathree.layernorm.pair_cell}{75.4\% / 89.8\%}
\defmmnum{ws4b.comp.cross.moonshine_on_whisper.gemmatwotoqwen.layernorm.clean}{123.83}
\defmmnum{ws4b.comp.cross.moonshine_on_whisper.gemmatwotoqwen.layernorm.clean_cell}{123.83\%}
\defmmnum{ws4b.comp.cross.moonshine_on_whisper.gemmatwotoqwen.layernorm.empty_clean}{0}
\defmmnum{ws4b.comp.cross.moonshine_on_whisper.gemmatwotoqwen.layernorm.empty_other}{0}
\defmmnum{ws4b.comp.cross.moonshine_on_whisper.gemmatwotoqwen.layernorm.other}{130.33}
\defmmnum{ws4b.comp.cross.moonshine_on_whisper.gemmatwotoqwen.layernorm.other_cell}{130.33\%}
\defmmnum{ws4b.comp.cross.moonshine_on_whisper.gemmatwotoqwen.layernorm.pair_cell}{123.8\% / 130.3\%}
\defmmnum{ws4b.comp.cross.moonshine_on_whisper.qwentogemmathree.layernorm.clean}{463.31}
\defmmnum{ws4b.comp.cross.moonshine_on_whisper.qwentogemmathree.layernorm.clean_cell}{463.31\%}
\defmmnum{ws4b.comp.cross.moonshine_on_whisper.qwentogemmathree.layernorm.empty_clean}{0}
\defmmnum{ws4b.comp.cross.moonshine_on_whisper.qwentogemmathree.layernorm.empty_other}{0}
\defmmnum{ws4b.comp.cross.moonshine_on_whisper.qwentogemmathree.layernorm.other}{516.54}
\defmmnum{ws4b.comp.cross.moonshine_on_whisper.qwentogemmathree.layernorm.other_cell}{516.54\%}
\defmmnum{ws4b.comp.cross.moonshine_on_whisper.qwentogemmathree.layernorm.pair_cell}{463.3\% / 516.5\%}
\defmmnum{ws4b.comp.cross.moonshine_on_whisper.qwentogemmatwo.layernorm.clean}{409.20}
\defmmnum{ws4b.comp.cross.moonshine_on_whisper.qwentogemmatwo.layernorm.clean_cell}{409.20\%}
\defmmnum{ws4b.comp.cross.moonshine_on_whisper.qwentogemmatwo.layernorm.empty_clean}{0}
\defmmnum{ws4b.comp.cross.moonshine_on_whisper.qwentogemmatwo.layernorm.empty_other}{0}
\defmmnum{ws4b.comp.cross.moonshine_on_whisper.qwentogemmatwo.layernorm.other}{443.52}
\defmmnum{ws4b.comp.cross.moonshine_on_whisper.qwentogemmatwo.layernorm.other_cell}{443.52\%}
\defmmnum{ws4b.comp.cross.moonshine_on_whisper.qwentogemmatwo.layernorm.pair_cell}{409.2\% / 443.5\%}
\defmmnum{ws4b.comp.cross.whisper_on_moonshine.gemmatwotogemmathree.layernorm.clean}{67.35}
\defmmnum{ws4b.comp.cross.whisper_on_moonshine.gemmatwotogemmathree.layernorm.clean_cell}{67.35\%}
\defmmnum{ws4b.comp.cross.whisper_on_moonshine.gemmatwotogemmathree.layernorm.empty_clean}{361}
\defmmnum{ws4b.comp.cross.whisper_on_moonshine.gemmatwotogemmathree.layernorm.empty_other}{387}
\defmmnum{ws4b.comp.cross.whisper_on_moonshine.gemmatwotogemmathree.layernorm.other}{77.47}
\defmmnum{ws4b.comp.cross.whisper_on_moonshine.gemmatwotogemmathree.layernorm.other_cell}{77.47\%}
\defmmnum{ws4b.comp.cross.whisper_on_moonshine.gemmatwotogemmathree.layernorm.pair_cell}{67.4\% / 77.5\%}
\defmmnum{ws4b.comp.cross.whisper_on_moonshine.gemmatwotoqwen.layernorm.clean}{98.63}
\defmmnum{ws4b.comp.cross.whisper_on_moonshine.gemmatwotoqwen.layernorm.clean_cell}{98.63\%}
\defmmnum{ws4b.comp.cross.whisper_on_moonshine.gemmatwotoqwen.layernorm.empty_clean}{0}
\defmmnum{ws4b.comp.cross.whisper_on_moonshine.gemmatwotoqwen.layernorm.empty_other}{0}
\defmmnum{ws4b.comp.cross.whisper_on_moonshine.gemmatwotoqwen.layernorm.other}{107.15}
\defmmnum{ws4b.comp.cross.whisper_on_moonshine.gemmatwotoqwen.layernorm.other_cell}{107.15\%}
\defmmnum{ws4b.comp.cross.whisper_on_moonshine.gemmatwotoqwen.layernorm.pair_cell}{98.6\% / 107.2\%}
\defmmnum{ws4b.comp.cross.whisper_on_moonshine.qwentogemmathree.layernorm.clean}{429.37}
\defmmnum{ws4b.comp.cross.whisper_on_moonshine.qwentogemmathree.layernorm.clean_cell}{429.37\%}
\defmmnum{ws4b.comp.cross.whisper_on_moonshine.qwentogemmathree.layernorm.empty_clean}{2}
\defmmnum{ws4b.comp.cross.whisper_on_moonshine.qwentogemmathree.layernorm.empty_other}{2}
\defmmnum{ws4b.comp.cross.whisper_on_moonshine.qwentogemmathree.layernorm.other}{475.50}
\defmmnum{ws4b.comp.cross.whisper_on_moonshine.qwentogemmathree.layernorm.other_cell}{475.50\%}
\defmmnum{ws4b.comp.cross.whisper_on_moonshine.qwentogemmathree.layernorm.pair_cell}{429.4\% / 475.5\%}
\defmmnum{ws4b.comp.cross.whisper_on_moonshine.qwentogemmatwo.layernorm.clean}{476.67}
\defmmnum{ws4b.comp.cross.whisper_on_moonshine.qwentogemmatwo.layernorm.clean_cell}{476.67\%}
\defmmnum{ws4b.comp.cross.whisper_on_moonshine.qwentogemmatwo.layernorm.empty_clean}{0}
\defmmnum{ws4b.comp.cross.whisper_on_moonshine.qwentogemmatwo.layernorm.empty_other}{0}
\defmmnum{ws4b.comp.cross.whisper_on_moonshine.qwentogemmatwo.layernorm.other}{536.30}
\defmmnum{ws4b.comp.cross.whisper_on_moonshine.qwentogemmatwo.layernorm.other_cell}{536.30\%}
\defmmnum{ws4b.comp.cross.whisper_on_moonshine.qwentogemmatwo.layernorm.pair_cell}{476.7\% / 536.3\%}
\defmmnum{ws4b.comp.cross_done}{8}
\defmmnum{ws4b.comp.cross_max_clean}{476.67}
\defmmnum{ws4b.comp.cross_min_clean}{67.35}
\defmmnum{ws4b.comp.cross_total}{8}
\defmmnum{ws4b.comp.done}{17}
\defmmnum{ws4b.comp.indist.moonshine.gemmatwotogemmathree.layernorm.clean}{4.29}
\defmmnum{ws4b.comp.indist.moonshine.gemmatwotogemmathree.layernorm.clean_cell}{4.29\%}
\defmmnum{ws4b.comp.indist.moonshine.gemmatwotogemmathree.layernorm.empty_clean}{0}
\defmmnum{ws4b.comp.indist.moonshine.gemmatwotogemmathree.layernorm.empty_other}{0}
\defmmnum{ws4b.comp.indist.moonshine.gemmatwotogemmathree.layernorm.other}{10.85}
\defmmnum{ws4b.comp.indist.moonshine.gemmatwotogemmathree.layernorm.other_cell}{10.85\%}
\defmmnum{ws4b.comp.indist.moonshine.gemmatwotogemmathree.layernorm.pair_cell}{4.3\% / 10.9\%}
\defmmnum{ws4b.comp.indist.moonshine.gemmatwotoqwen.layernorm.clean}{9.04}
\defmmnum{ws4b.comp.indist.moonshine.gemmatwotoqwen.layernorm.clean_cell}{9.04\%}
\defmmnum{ws4b.comp.indist.moonshine.gemmatwotoqwen.layernorm.empty_clean}{0}
\defmmnum{ws4b.comp.indist.moonshine.gemmatwotoqwen.layernorm.empty_other}{0}
\defmmnum{ws4b.comp.indist.moonshine.gemmatwotoqwen.layernorm.other}{18.91}
\defmmnum{ws4b.comp.indist.moonshine.gemmatwotoqwen.layernorm.other_cell}{18.91\%}
\defmmnum{ws4b.comp.indist.moonshine.gemmatwotoqwen.layernorm.pair_cell}{9.0\% / 18.9\%}
\defmmnum{ws4b.comp.indist.moonshine.gemmatwotoqwen.scaled_l2norm.clean}{9.03}
\defmmnum{ws4b.comp.indist.moonshine.gemmatwotoqwen.scaled_l2norm.clean_cell}{9.03\%}
\defmmnum{ws4b.comp.indist.moonshine.gemmatwotoqwen.scaled_l2norm.empty_clean}{0}
\defmmnum{ws4b.comp.indist.moonshine.gemmatwotoqwen.scaled_l2norm.empty_other}{0}
\defmmnum{ws4b.comp.indist.moonshine.gemmatwotoqwen.scaled_l2norm.other}{18.79}
\defmmnum{ws4b.comp.indist.moonshine.gemmatwotoqwen.scaled_l2norm.other_cell}{18.79\%}
\defmmnum{ws4b.comp.indist.moonshine.gemmatwotoqwen.scaled_l2norm.pair_cell}{9.0\% / 18.8\%}
\defmmnum{ws4b.comp.indist.moonshine.qwentogemmathree.layernorm.clean}{4.35}
\defmmnum{ws4b.comp.indist.moonshine.qwentogemmathree.layernorm.clean_cell}{4.35\%}
\defmmnum{ws4b.comp.indist.moonshine.qwentogemmathree.layernorm.empty_clean}{0}
\defmmnum{ws4b.comp.indist.moonshine.qwentogemmathree.layernorm.empty_other}{0}
\defmmnum{ws4b.comp.indist.moonshine.qwentogemmathree.layernorm.other}{10.91}
\defmmnum{ws4b.comp.indist.moonshine.qwentogemmathree.layernorm.other_cell}{10.91\%}
\defmmnum{ws4b.comp.indist.moonshine.qwentogemmathree.layernorm.pair_cell}{4.3\% / 10.9\%}
\defmmnum{ws4b.comp.indist.moonshine.qwentogemmatwo.layernorm.clean}{5.10}
\defmmnum{ws4b.comp.indist.moonshine.qwentogemmatwo.layernorm.clean_cell}{5.10\%}
\defmmnum{ws4b.comp.indist.moonshine.qwentogemmatwo.layernorm.empty_clean}{0}
\defmmnum{ws4b.comp.indist.moonshine.qwentogemmatwo.layernorm.empty_other}{0}
\defmmnum{ws4b.comp.indist.moonshine.qwentogemmatwo.layernorm.other}{12.45}
\defmmnum{ws4b.comp.indist.moonshine.qwentogemmatwo.layernorm.other_cell}{12.45\%}
\defmmnum{ws4b.comp.indist.moonshine.qwentogemmatwo.layernorm.pair_cell}{5.1\% / 12.5\%}
\defmmnum{ws4b.comp.indist.whisper.gemmatwotogemmathree.layernorm.clean}{3.22}
\defmmnum{ws4b.comp.indist.whisper.gemmatwotogemmathree.layernorm.clean_cell}{3.22\%}
\defmmnum{ws4b.comp.indist.whisper.gemmatwotogemmathree.layernorm.empty_clean}{0}
\defmmnum{ws4b.comp.indist.whisper.gemmatwotogemmathree.layernorm.empty_other}{0}
\defmmnum{ws4b.comp.indist.whisper.gemmatwotogemmathree.layernorm.other}{7.53}
\defmmnum{ws4b.comp.indist.whisper.gemmatwotogemmathree.layernorm.other_cell}{7.53\%}
\defmmnum{ws4b.comp.indist.whisper.gemmatwotogemmathree.layernorm.pair_cell}{3.2\% / 7.5\%}
\defmmnum{ws4b.comp.indist.whisper.gemmatwotoqwen.layernorm.clean}{5.12}
\defmmnum{ws4b.comp.indist.whisper.gemmatwotoqwen.layernorm.clean_cell}{5.12\%}
\defmmnum{ws4b.comp.indist.whisper.gemmatwotoqwen.layernorm.empty_clean}{0}
\defmmnum{ws4b.comp.indist.whisper.gemmatwotoqwen.layernorm.empty_other}{0}
\defmmnum{ws4b.comp.indist.whisper.gemmatwotoqwen.layernorm.other}{11.28}
\defmmnum{ws4b.comp.indist.whisper.gemmatwotoqwen.layernorm.other_cell}{11.28\%}
\defmmnum{ws4b.comp.indist.whisper.gemmatwotoqwen.layernorm.pair_cell}{5.1\% / 11.3\%}
\defmmnum{ws4b.comp.indist.whisper.qwentogemmathree.layernorm.clean}{3.24}
\defmmnum{ws4b.comp.indist.whisper.qwentogemmathree.layernorm.clean_cell}{3.24\%}
\defmmnum{ws4b.comp.indist.whisper.qwentogemmathree.layernorm.empty_clean}{0}
\defmmnum{ws4b.comp.indist.whisper.qwentogemmathree.layernorm.empty_other}{0}
\defmmnum{ws4b.comp.indist.whisper.qwentogemmathree.layernorm.other}{7.78}
\defmmnum{ws4b.comp.indist.whisper.qwentogemmathree.layernorm.other_cell}{7.78\%}
\defmmnum{ws4b.comp.indist.whisper.qwentogemmathree.layernorm.pair_cell}{3.2\% / 7.8\%}
\defmmnum{ws4b.comp.indist.whisper.qwentogemmatwo.layernorm.clean}{3.56}
\defmmnum{ws4b.comp.indist.whisper.qwentogemmatwo.layernorm.clean_cell}{3.56\%}
\defmmnum{ws4b.comp.indist.whisper.qwentogemmatwo.layernorm.empty_clean}{0}
\defmmnum{ws4b.comp.indist.whisper.qwentogemmatwo.layernorm.empty_other}{0}
\defmmnum{ws4b.comp.indist.whisper.qwentogemmatwo.layernorm.other}{8.45}
\defmmnum{ws4b.comp.indist.whisper.qwentogemmatwo.layernorm.other_cell}{8.45\%}
\defmmnum{ws4b.comp.indist.whisper.qwentogemmatwo.layernorm.pair_cell}{3.6\% / 8.4\%}
\defmmnum{ws4b.comp.indist_done}{9}
\defmmnum{ws4b.comp.indist_max_clean}{9.04}
\defmmnum{ws4b.comp.indist_max_other}{18.91}
\defmmnum{ws4b.comp.indist_min_clean}{3.22}
\defmmnum{ws4b.comp.indist_min_other}{7.53}
\defmmnum{ws4b.comp.indist_total}{9}
\defmmnum{ws4b.comp.total}{17}
\defmmnum{ws4b.cross.moonshine_on_whisper.clean}{123.83}
\defmmnum{ws4b.cross.moonshine_on_whisper.other}{130.33}
\defmmnum{ws4b.cross.whisper_on_moonshine.clean}{98.63}
\defmmnum{ws4b.cross.whisper_on_moonshine.other}{107.15}
\defmmnum{ws4b.fit.moonshine.gemmatwotogemmathree.layernorm.anchors}{163{,}983}
\defmmnum{ws4b.fit.moonshine.gemmatwotogemmathree.layernorm.audio_val}{0.1224}
\defmmnum{ws4b.fit.moonshine.gemmatwotogemmathree.layernorm.cos}{0.0339}
\defmmnum{ws4b.fit.moonshine.gemmatwotogemmathree.layernorm.n_test}{32{,}796}
\defmmnum{ws4b.fit.moonshine.gemmatwotogemmathree.layernorm.p1}{0.83}
\defmmnum{ws4b.fit.moonshine.gemmatwotogemmathree.layernorm.params}{5{,}910{,}528}
\defmmnum{ws4b.fit.moonshine.gemmatwotogemmathree.layernorm.params_m}{5.91M}
\defmmnum{ws4b.fit.moonshine.gemmatwotogemmathree.layernorm.rr}{1.058}
\defmmnum{ws4b.fit.moonshine.gemmatwotogemmathree.layernorm.text_val}{6.92}
\defmmnum{ws4b.fit.moonshine.gemmatwotoqwen.layernorm.anchors}{83{,}378}
\defmmnum{ws4b.fit.moonshine.gemmatwotoqwen.layernorm.audio_val}{0.2931}
\defmmnum{ws4b.fit.moonshine.gemmatwotoqwen.layernorm.cos}{0.0244}
\defmmnum{ws4b.fit.moonshine.gemmatwotoqwen.layernorm.n_test}{16{,}675}
\defmmnum{ws4b.fit.moonshine.gemmatwotoqwen.layernorm.p1}{0.24}
\defmmnum{ws4b.fit.moonshine.gemmatwotoqwen.layernorm.params}{3{,}548{,}160}
\defmmnum{ws4b.fit.moonshine.gemmatwotoqwen.layernorm.params_m}{3.55M}
\defmmnum{ws4b.fit.moonshine.gemmatwotoqwen.layernorm.rr}{8.368}
\defmmnum{ws4b.fit.moonshine.gemmatwotoqwen.layernorm.text_val}{5.41}
\defmmnum{ws4b.fit.moonshine.gemmatwotoqwen.scaled_l2norm.anchors}{83{,}378}
\defmmnum{ws4b.fit.moonshine.gemmatwotoqwen.scaled_l2norm.audio_val}{0.2962}
\defmmnum{ws4b.fit.moonshine.gemmatwotoqwen.scaled_l2norm.cos}{0.0363}
\defmmnum{ws4b.fit.moonshine.gemmatwotoqwen.scaled_l2norm.n_test}{16{,}675}
\defmmnum{ws4b.fit.moonshine.gemmatwotoqwen.scaled_l2norm.p1}{0.47}
\defmmnum{ws4b.fit.moonshine.gemmatwotoqwen.scaled_l2norm.params}{3{,}540{,}482}
\defmmnum{ws4b.fit.moonshine.gemmatwotoqwen.scaled_l2norm.params_m}{3.54M}
\defmmnum{ws4b.fit.moonshine.gemmatwotoqwen.scaled_l2norm.rr}{7.592}
\defmmnum{ws4b.fit.moonshine.gemmatwotoqwen.scaled_l2norm.text_val}{5.24}
\defmmnum{ws4b.fit.moonshine.qwentogemmathree.layernorm.anchors}{82{,}663}
\defmmnum{ws4b.fit.moonshine.qwentogemmathree.layernorm.audio_val}{0.1279}
\defmmnum{ws4b.fit.moonshine.qwentogemmathree.layernorm.cos}{0.0254}
\defmmnum{ws4b.fit.moonshine.qwentogemmathree.layernorm.n_test}{16{,}532}
\defmmnum{ws4b.fit.moonshine.qwentogemmathree.layernorm.p1}{0.50}
\defmmnum{ws4b.fit.moonshine.qwentogemmathree.layernorm.params}{3{,}942{,}912}
\defmmnum{ws4b.fit.moonshine.qwentogemmathree.layernorm.params_m}{3.94M}
\defmmnum{ws4b.fit.moonshine.qwentogemmathree.layernorm.rr}{1.047}
\defmmnum{ws4b.fit.moonshine.qwentogemmathree.layernorm.text_val}{7.29}
\defmmnum{ws4b.fit.moonshine.qwentogemmatwo.layernorm.anchors}{83{,}378}
\defmmnum{ws4b.fit.moonshine.qwentogemmatwo.layernorm.audio_val}{0.1539}
\defmmnum{ws4b.fit.moonshine.qwentogemmatwo.layernorm.cos}{0.0280}
\defmmnum{ws4b.fit.moonshine.qwentogemmatwo.layernorm.n_test}{16{,}675}
\defmmnum{ws4b.fit.moonshine.qwentogemmatwo.layernorm.p1}{0.63}
\defmmnum{ws4b.fit.moonshine.qwentogemmatwo.layernorm.params}{3{,}548{,}928}
\defmmnum{ws4b.fit.moonshine.qwentogemmatwo.layernorm.params_m}{3.55M}
\defmmnum{ws4b.fit.moonshine.qwentogemmatwo.layernorm.rr}{1.028}
\defmmnum{ws4b.fit.moonshine.qwentogemmatwo.layernorm.text_val}{5.61}
\defmmnum{ws4b.fit.whisper.gemmatwotogemmathree.layernorm.anchors}{163{,}983}
\defmmnum{ws4b.fit.whisper.gemmatwotogemmathree.layernorm.audio_val}{0.0903}
\defmmnum{ws4b.fit.whisper.gemmatwotogemmathree.layernorm.cos}{0.0387}
\defmmnum{ws4b.fit.whisper.gemmatwotogemmathree.layernorm.n_test}{32{,}796}
\defmmnum{ws4b.fit.whisper.gemmatwotogemmathree.layernorm.p1}{1.27}
\defmmnum{ws4b.fit.whisper.gemmatwotogemmathree.layernorm.params}{5{,}910{,}528}
\defmmnum{ws4b.fit.whisper.gemmatwotogemmathree.layernorm.params_m}{5.91M}
\defmmnum{ws4b.fit.whisper.gemmatwotogemmathree.layernorm.rr}{1.053}
\defmmnum{ws4b.fit.whisper.gemmatwotogemmathree.layernorm.text_val}{7.44}
\defmmnum{ws4b.fit.whisper.gemmatwotoqwen.layernorm.anchors}{83{,}378}
\defmmnum{ws4b.fit.whisper.gemmatwotoqwen.layernorm.audio_val}{0.1736}
\defmmnum{ws4b.fit.whisper.gemmatwotoqwen.layernorm.cos}{0.0437}
\defmmnum{ws4b.fit.whisper.gemmatwotoqwen.layernorm.n_test}{16{,}675}
\defmmnum{ws4b.fit.whisper.gemmatwotoqwen.layernorm.p1}{0.41}
\defmmnum{ws4b.fit.whisper.gemmatwotoqwen.layernorm.params}{3{,}548{,}160}
\defmmnum{ws4b.fit.whisper.gemmatwotoqwen.layernorm.params_m}{3.55M}
\defmmnum{ws4b.fit.whisper.gemmatwotoqwen.layernorm.rr}{6.891}
\defmmnum{ws4b.fit.whisper.gemmatwotoqwen.layernorm.text_val}{5.24}
\defmmnum{ws4b.fit.whisper.qwentogemmathree.layernorm.anchors}{82{,}663}
\defmmnum{ws4b.fit.whisper.qwentogemmathree.layernorm.audio_val}{0.0889}
\defmmnum{ws4b.fit.whisper.qwentogemmathree.layernorm.cos}{0.0296}
\defmmnum{ws4b.fit.whisper.qwentogemmathree.layernorm.n_test}{16{,}532}
\defmmnum{ws4b.fit.whisper.qwentogemmathree.layernorm.p1}{0.74}
\defmmnum{ws4b.fit.whisper.qwentogemmathree.layernorm.params}{3{,}942{,}912}
\defmmnum{ws4b.fit.whisper.qwentogemmathree.layernorm.params_m}{3.94M}
\defmmnum{ws4b.fit.whisper.qwentogemmathree.layernorm.rr}{1.051}
\defmmnum{ws4b.fit.whisper.qwentogemmathree.layernorm.text_val}{7.22}
\defmmnum{ws4b.fit.whisper.qwentogemmatwo.layernorm.anchors}{83{,}378}
\defmmnum{ws4b.fit.whisper.qwentogemmatwo.layernorm.audio_val}{0.1059}
\defmmnum{ws4b.fit.whisper.qwentogemmatwo.layernorm.cos}{0.0332}
\defmmnum{ws4b.fit.whisper.qwentogemmatwo.layernorm.n_test}{16{,}675}
\defmmnum{ws4b.fit.whisper.qwentogemmatwo.layernorm.p1}{0.70}
\defmmnum{ws4b.fit.whisper.qwentogemmatwo.layernorm.params}{3{,}548{,}928}
\defmmnum{ws4b.fit.whisper.qwentogemmatwo.layernorm.params_m}{3.55M}
\defmmnum{ws4b.fit.whisper.qwentogemmatwo.layernorm.rr}{1.034}
\defmmnum{ws4b.fit.whisper.qwentogemmatwo.layernorm.text_val}{5.56}
\defmmnum{ws4b.fits.done}{9}
\defmmnum{ws4b.fits.max_audio_val}{0.2962}
\defmmnum{ws4b.fits.max_cos}{0.0437}
\defmmnum{ws4b.fits.max_p1}{1.27}
\defmmnum{ws4b.fits.max_text_val}{7.44}
\defmmnum{ws4b.fits.min_audio_val}{0.0889}
\defmmnum{ws4b.fits.min_cos}{0.0244}
\defmmnum{ws4b.fits.min_p1}{0.24}
\defmmnum{ws4b.fits.min_text_val}{5.24}
\defmmnum{ws4b.fits.total}{9}
\defmmnum{ws4b.noise.done}{3}
\defmmnum{ws4b.noise.k050_label}{0.50}
\defmmnum{ws4b.noise.k0_50.clean}{110.06}
\defmmnum{ws4b.noise.k0_50.clean_cell}{110.06\%}
\defmmnum{ws4b.noise.k0_50.cos}{0.1532}
\defmmnum{ws4b.noise.k0_50.empty_clean}{0}
\defmmnum{ws4b.noise.k0_50.other}{112.18}
\defmmnum{ws4b.noise.k0_50.other_cell}{112.18\%}
\defmmnum{ws4b.noise.k0_50.p1}{21.47}
\defmmnum{ws4b.noise.k0_50.pair_cell}{110.1\% / 112.2\%}
\defmmnum{ws4b.noise.k0_50.rr}{3.950}
\defmmnum{ws4b.noise.k0_50.val}{0.0192}
\defmmnum{ws4b.noise.k0_label}{0}
\defmmnum{ws4b.noise.k100_label}{1.00}
\defmmnum{ws4b.noise.k1_00.clean}{99.82}
\defmmnum{ws4b.noise.k1_00.clean_cell}{99.82\%}
\defmmnum{ws4b.noise.k1_00.cos}{0.1440}
\defmmnum{ws4b.noise.k1_00.empty_clean}{0}
\defmmnum{ws4b.noise.k1_00.other}{100.86}
\defmmnum{ws4b.noise.k1_00.other_cell}{100.86\%}
\defmmnum{ws4b.noise.k1_00.p1}{25.14}
\defmmnum{ws4b.noise.k1_00.pair_cell}{99.8\% / 100.9\%}
\defmmnum{ws4b.noise.k1_00.rr}{5.298}
\defmmnum{ws4b.noise.k1_00.val}{0.0254}
\defmmnum{ws4b.noise.k773_label}{7.73}
\defmmnum{ws4b.noise.k7_73.clean}{143.98}
\defmmnum{ws4b.noise.k7_73.clean_cell}{143.98\%}
\defmmnum{ws4b.noise.k7_73.cos}{0.0178}
\defmmnum{ws4b.noise.k7_73.empty_clean}{18}
\defmmnum{ws4b.noise.k7_73.other}{150.13}
\defmmnum{ws4b.noise.k7_73.other_cell}{150.13\%}
\defmmnum{ws4b.noise.k7_73.p1}{0.01}
\defmmnum{ws4b.noise.k7_73.pair_cell}{144.0\% / 150.1\%}
\defmmnum{ws4b.noise.k7_73.rr}{3.236}
\defmmnum{ws4b.noise.k7_73.val}{3.9662}
\defmmnum{ws4b.noise.max_clean}{143.98}
\defmmnum{ws4b.noise.min_clean}{99.82}
\defmmnum{ws4b.noise.total}{3}
\defmmnum{ws4b.norm_delta.moonshine.g2_to_qwen.scaled_vs_ln.clean}{-0.01}
\defmmnum{ws4b.norm_delta.moonshine.g2_to_qwen.scaled_vs_ln.other}{-0.12}
\defmmnum{ws4c.comp.done}{24}
\defmmnum{ws4c.comp.total}{24}
\defmmnum{ws4c.fits.done}{12}
\defmmnum{ws4c.fits.total}{12}
\defmmnum{ws4c.mix.audio_pct}{50}
\defmmnum{ws4c.mix.best_clean}{3.08}
\defmmnum{ws4c.mix.best_other}{7.38}
\defmmnum{ws4c.mix.comp.cross.moonshine_on_whisper.gemmatwotogemmathree.layernorm.clean}{154.00}
\defmmnum{ws4c.mix.comp.cross.moonshine_on_whisper.gemmatwotogemmathree.layernorm.clean_cell}{154.00\%}
\defmmnum{ws4c.mix.comp.cross.moonshine_on_whisper.gemmatwotogemmathree.layernorm.empty_clean}{0}
\defmmnum{ws4c.mix.comp.cross.moonshine_on_whisper.gemmatwotogemmathree.layernorm.empty_other}{0}
\defmmnum{ws4c.mix.comp.cross.moonshine_on_whisper.gemmatwotogemmathree.layernorm.other}{205.91}
\defmmnum{ws4c.mix.comp.cross.moonshine_on_whisper.gemmatwotogemmathree.layernorm.other_cell}{205.91\%}
\defmmnum{ws4c.mix.comp.cross.moonshine_on_whisper.gemmatwotogemmathree.layernorm.pair_cell}{154.0\% / 205.9\%}
\defmmnum{ws4c.mix.comp.cross.moonshine_on_whisper.gemmatwotogemmathree.layernorm.short_clean_cell}{154.0\%}
\defmmnum{ws4c.mix.comp.cross.moonshine_on_whisper.gemmatwotoqwen.layernorm.clean}{123.29}
\defmmnum{ws4c.mix.comp.cross.moonshine_on_whisper.gemmatwotoqwen.layernorm.clean_cell}{123.29\%}
\defmmnum{ws4c.mix.comp.cross.moonshine_on_whisper.gemmatwotoqwen.layernorm.empty_clean}{0}
\defmmnum{ws4c.mix.comp.cross.moonshine_on_whisper.gemmatwotoqwen.layernorm.empty_other}{0}
\defmmnum{ws4c.mix.comp.cross.moonshine_on_whisper.gemmatwotoqwen.layernorm.other}{134.18}
\defmmnum{ws4c.mix.comp.cross.moonshine_on_whisper.gemmatwotoqwen.layernorm.other_cell}{134.18\%}
\defmmnum{ws4c.mix.comp.cross.moonshine_on_whisper.gemmatwotoqwen.layernorm.pair_cell}{123.3\% / 134.2\%}
\defmmnum{ws4c.mix.comp.cross.moonshine_on_whisper.gemmatwotoqwen.layernorm.short_clean_cell}{123.3\%}
\defmmnum{ws4c.mix.comp.cross.moonshine_on_whisper.qwentogemmathree.layernorm.clean}{480.30}
\defmmnum{ws4c.mix.comp.cross.moonshine_on_whisper.qwentogemmathree.layernorm.clean_cell}{480.30\%}
\defmmnum{ws4c.mix.comp.cross.moonshine_on_whisper.qwentogemmathree.layernorm.empty_clean}{0}
\defmmnum{ws4c.mix.comp.cross.moonshine_on_whisper.qwentogemmathree.layernorm.empty_other}{0}
\defmmnum{ws4c.mix.comp.cross.moonshine_on_whisper.qwentogemmathree.layernorm.other}{524.65}
\defmmnum{ws4c.mix.comp.cross.moonshine_on_whisper.qwentogemmathree.layernorm.other_cell}{524.65\%}
\defmmnum{ws4c.mix.comp.cross.moonshine_on_whisper.qwentogemmathree.layernorm.pair_cell}{480.3\% / 524.6\%}
\defmmnum{ws4c.mix.comp.cross.moonshine_on_whisper.qwentogemmathree.layernorm.short_clean_cell}{480.3\%}
\defmmnum{ws4c.mix.comp.cross.moonshine_on_whisper.qwentogemmatwo.layernorm.clean}{476.62}
\defmmnum{ws4c.mix.comp.cross.moonshine_on_whisper.qwentogemmatwo.layernorm.clean_cell}{476.62\%}
\defmmnum{ws4c.mix.comp.cross.moonshine_on_whisper.qwentogemmatwo.layernorm.empty_clean}{0}
\defmmnum{ws4c.mix.comp.cross.moonshine_on_whisper.qwentogemmatwo.layernorm.empty_other}{0}
\defmmnum{ws4c.mix.comp.cross.moonshine_on_whisper.qwentogemmatwo.layernorm.other}{532.68}
\defmmnum{ws4c.mix.comp.cross.moonshine_on_whisper.qwentogemmatwo.layernorm.other_cell}{532.68\%}
\defmmnum{ws4c.mix.comp.cross.moonshine_on_whisper.qwentogemmatwo.layernorm.pair_cell}{476.6\% / 532.7\%}
\defmmnum{ws4c.mix.comp.cross.moonshine_on_whisper.qwentogemmatwo.layernorm.short_clean_cell}{476.6\%}
\defmmnum{ws4c.mix.comp.cross.whisper_on_moonshine.gemmatwotogemmathree.layernorm.clean}{452.43}
\defmmnum{ws4c.mix.comp.cross.whisper_on_moonshine.gemmatwotogemmathree.layernorm.clean_cell}{452.43\%}
\defmmnum{ws4c.mix.comp.cross.whisper_on_moonshine.gemmatwotogemmathree.layernorm.empty_clean}{0}
\defmmnum{ws4c.mix.comp.cross.whisper_on_moonshine.gemmatwotogemmathree.layernorm.empty_other}{0}
\defmmnum{ws4c.mix.comp.cross.whisper_on_moonshine.gemmatwotogemmathree.layernorm.other}{505.36}
\defmmnum{ws4c.mix.comp.cross.whisper_on_moonshine.gemmatwotogemmathree.layernorm.other_cell}{505.36\%}
\defmmnum{ws4c.mix.comp.cross.whisper_on_moonshine.gemmatwotogemmathree.layernorm.pair_cell}{452.4\% / 505.4\%}
\defmmnum{ws4c.mix.comp.cross.whisper_on_moonshine.gemmatwotogemmathree.layernorm.short_clean_cell}{452.4\%}
\defmmnum{ws4c.mix.comp.cross.whisper_on_moonshine.gemmatwotoqwen.layernorm.clean}{105.96}
\defmmnum{ws4c.mix.comp.cross.whisper_on_moonshine.gemmatwotoqwen.layernorm.clean_cell}{105.96\%}
\defmmnum{ws4c.mix.comp.cross.whisper_on_moonshine.gemmatwotoqwen.layernorm.empty_clean}{0}
\defmmnum{ws4c.mix.comp.cross.whisper_on_moonshine.gemmatwotoqwen.layernorm.empty_other}{0}
\defmmnum{ws4c.mix.comp.cross.whisper_on_moonshine.gemmatwotoqwen.layernorm.other}{109.21}
\defmmnum{ws4c.mix.comp.cross.whisper_on_moonshine.gemmatwotoqwen.layernorm.other_cell}{109.21\%}
\defmmnum{ws4c.mix.comp.cross.whisper_on_moonshine.gemmatwotoqwen.layernorm.pair_cell}{106.0\% / 109.2\%}
\defmmnum{ws4c.mix.comp.cross.whisper_on_moonshine.gemmatwotoqwen.layernorm.short_clean_cell}{106.0\%}
\defmmnum{ws4c.mix.comp.cross.whisper_on_moonshine.qwentogemmathree.layernorm.clean}{455.14}
\defmmnum{ws4c.mix.comp.cross.whisper_on_moonshine.qwentogemmathree.layernorm.clean_cell}{455.14\%}
\defmmnum{ws4c.mix.comp.cross.whisper_on_moonshine.qwentogemmathree.layernorm.empty_clean}{0}
\defmmnum{ws4c.mix.comp.cross.whisper_on_moonshine.qwentogemmathree.layernorm.empty_other}{0}
\defmmnum{ws4c.mix.comp.cross.whisper_on_moonshine.qwentogemmathree.layernorm.other}{498.03}
\defmmnum{ws4c.mix.comp.cross.whisper_on_moonshine.qwentogemmathree.layernorm.other_cell}{498.03\%}
\defmmnum{ws4c.mix.comp.cross.whisper_on_moonshine.qwentogemmathree.layernorm.pair_cell}{455.1\% / 498.0\%}
\defmmnum{ws4c.mix.comp.cross.whisper_on_moonshine.qwentogemmathree.layernorm.short_clean_cell}{455.1\%}
\defmmnum{ws4c.mix.comp.cross.whisper_on_moonshine.qwentogemmatwo.layernorm.clean}{465.22}
\defmmnum{ws4c.mix.comp.cross.whisper_on_moonshine.qwentogemmatwo.layernorm.clean_cell}{465.22\%}
\defmmnum{ws4c.mix.comp.cross.whisper_on_moonshine.qwentogemmatwo.layernorm.empty_clean}{0}
\defmmnum{ws4c.mix.comp.cross.whisper_on_moonshine.qwentogemmatwo.layernorm.empty_other}{0}
\defmmnum{ws4c.mix.comp.cross.whisper_on_moonshine.qwentogemmatwo.layernorm.other}{522.81}
\defmmnum{ws4c.mix.comp.cross.whisper_on_moonshine.qwentogemmatwo.layernorm.other_cell}{522.81\%}
\defmmnum{ws4c.mix.comp.cross.whisper_on_moonshine.qwentogemmatwo.layernorm.pair_cell}{465.2\% / 522.8\%}
\defmmnum{ws4c.mix.comp.cross.whisper_on_moonshine.qwentogemmatwo.layernorm.short_clean_cell}{465.2\%}
\defmmnum{ws4c.mix.comp.indist.moonshine.gemmatwotogemmathree.layernorm.clean}{4.21}
\defmmnum{ws4c.mix.comp.indist.moonshine.gemmatwotogemmathree.layernorm.clean_cell}{4.21\%}
\defmmnum{ws4c.mix.comp.indist.moonshine.gemmatwotogemmathree.layernorm.empty_clean}{0}
\defmmnum{ws4c.mix.comp.indist.moonshine.gemmatwotogemmathree.layernorm.empty_other}{0}
\defmmnum{ws4c.mix.comp.indist.moonshine.gemmatwotogemmathree.layernorm.other}{10.81}
\defmmnum{ws4c.mix.comp.indist.moonshine.gemmatwotogemmathree.layernorm.other_cell}{10.81\%}
\defmmnum{ws4c.mix.comp.indist.moonshine.gemmatwotogemmathree.layernorm.pair_cell}{4.2\% / 10.8\%}
\defmmnum{ws4c.mix.comp.indist.moonshine.gemmatwotogemmathree.layernorm.short_clean_cell}{4.2\%}
\defmmnum{ws4c.mix.comp.indist.moonshine.gemmatwotoqwen.layernorm.clean}{8.03}
\defmmnum{ws4c.mix.comp.indist.moonshine.gemmatwotoqwen.layernorm.clean_cell}{8.03\%}
\defmmnum{ws4c.mix.comp.indist.moonshine.gemmatwotoqwen.layernorm.empty_clean}{0}
\defmmnum{ws4c.mix.comp.indist.moonshine.gemmatwotoqwen.layernorm.empty_other}{0}
\defmmnum{ws4c.mix.comp.indist.moonshine.gemmatwotoqwen.layernorm.other}{17.45}
\defmmnum{ws4c.mix.comp.indist.moonshine.gemmatwotoqwen.layernorm.other_cell}{17.45\%}
\defmmnum{ws4c.mix.comp.indist.moonshine.gemmatwotoqwen.layernorm.pair_cell}{8.0\% / 17.5\%}
\defmmnum{ws4c.mix.comp.indist.moonshine.gemmatwotoqwen.layernorm.short_clean_cell}{8.0\%}
\defmmnum{ws4c.mix.comp.indist.moonshine.qwentogemmathree.layernorm.clean}{4.33}
\defmmnum{ws4c.mix.comp.indist.moonshine.qwentogemmathree.layernorm.clean_cell}{4.33\%}
\defmmnum{ws4c.mix.comp.indist.moonshine.qwentogemmathree.layernorm.empty_clean}{0}
\defmmnum{ws4c.mix.comp.indist.moonshine.qwentogemmathree.layernorm.empty_other}{0}
\defmmnum{ws4c.mix.comp.indist.moonshine.qwentogemmathree.layernorm.other}{10.86}
\defmmnum{ws4c.mix.comp.indist.moonshine.qwentogemmathree.layernorm.other_cell}{10.86\%}
\defmmnum{ws4c.mix.comp.indist.moonshine.qwentogemmathree.layernorm.pair_cell}{4.3\% / 10.9\%}
\defmmnum{ws4c.mix.comp.indist.moonshine.qwentogemmathree.layernorm.short_clean_cell}{4.3\%}
\defmmnum{ws4c.mix.comp.indist.moonshine.qwentogemmatwo.layernorm.clean}{5.05}
\defmmnum{ws4c.mix.comp.indist.moonshine.qwentogemmatwo.layernorm.clean_cell}{5.05\%}
\defmmnum{ws4c.mix.comp.indist.moonshine.qwentogemmatwo.layernorm.empty_clean}{0}
\defmmnum{ws4c.mix.comp.indist.moonshine.qwentogemmatwo.layernorm.empty_other}{0}
\defmmnum{ws4c.mix.comp.indist.moonshine.qwentogemmatwo.layernorm.other}{12.71}
\defmmnum{ws4c.mix.comp.indist.moonshine.qwentogemmatwo.layernorm.other_cell}{12.71\%}
\defmmnum{ws4c.mix.comp.indist.moonshine.qwentogemmatwo.layernorm.pair_cell}{5.1\% / 12.7\%}
\defmmnum{ws4c.mix.comp.indist.moonshine.qwentogemmatwo.layernorm.short_clean_cell}{5.1\%}
\defmmnum{ws4c.mix.comp.indist.whisper.gemmatwotogemmathree.layernorm.clean}{3.26}
\defmmnum{ws4c.mix.comp.indist.whisper.gemmatwotogemmathree.layernorm.clean_cell}{3.26\%}
\defmmnum{ws4c.mix.comp.indist.whisper.gemmatwotogemmathree.layernorm.empty_clean}{0}
\defmmnum{ws4c.mix.comp.indist.whisper.gemmatwotogemmathree.layernorm.empty_other}{0}
\defmmnum{ws4c.mix.comp.indist.whisper.gemmatwotogemmathree.layernorm.other}{7.71}
\defmmnum{ws4c.mix.comp.indist.whisper.gemmatwotogemmathree.layernorm.other_cell}{7.71\%}
\defmmnum{ws4c.mix.comp.indist.whisper.gemmatwotogemmathree.layernorm.pair_cell}{3.3\% / 7.7\%}
\defmmnum{ws4c.mix.comp.indist.whisper.gemmatwotogemmathree.layernorm.short_clean_cell}{3.3\%}
\defmmnum{ws4c.mix.comp.indist.whisper.gemmatwotoqwen.layernorm.clean}{5.13}
\defmmnum{ws4c.mix.comp.indist.whisper.gemmatwotoqwen.layernorm.clean_cell}{5.13\%}
\defmmnum{ws4c.mix.comp.indist.whisper.gemmatwotoqwen.layernorm.empty_clean}{0}
\defmmnum{ws4c.mix.comp.indist.whisper.gemmatwotoqwen.layernorm.empty_other}{0}
\defmmnum{ws4c.mix.comp.indist.whisper.gemmatwotoqwen.layernorm.other}{11.15}
\defmmnum{ws4c.mix.comp.indist.whisper.gemmatwotoqwen.layernorm.other_cell}{11.15\%}
\defmmnum{ws4c.mix.comp.indist.whisper.gemmatwotoqwen.layernorm.pair_cell}{5.1\% / 11.1\%}
\defmmnum{ws4c.mix.comp.indist.whisper.gemmatwotoqwen.layernorm.short_clean_cell}{5.1\%}
\defmmnum{ws4c.mix.comp.indist.whisper.qwentogemmathree.layernorm.clean}{3.08}
\defmmnum{ws4c.mix.comp.indist.whisper.qwentogemmathree.layernorm.clean_cell}{3.08\%}
\defmmnum{ws4c.mix.comp.indist.whisper.qwentogemmathree.layernorm.empty_clean}{0}
\defmmnum{ws4c.mix.comp.indist.whisper.qwentogemmathree.layernorm.empty_other}{0}
\defmmnum{ws4c.mix.comp.indist.whisper.qwentogemmathree.layernorm.other}{7.38}
\defmmnum{ws4c.mix.comp.indist.whisper.qwentogemmathree.layernorm.other_cell}{7.38\%}
\defmmnum{ws4c.mix.comp.indist.whisper.qwentogemmathree.layernorm.pair_cell}{3.1\% / 7.4\%}
\defmmnum{ws4c.mix.comp.indist.whisper.qwentogemmathree.layernorm.short_clean_cell}{3.1\%}
\defmmnum{ws4c.mix.comp.indist.whisper.qwentogemmatwo.layernorm.clean}{3.72}
\defmmnum{ws4c.mix.comp.indist.whisper.qwentogemmatwo.layernorm.clean_cell}{3.72\%}
\defmmnum{ws4c.mix.comp.indist.whisper.qwentogemmatwo.layernorm.empty_clean}{0}
\defmmnum{ws4c.mix.comp.indist.whisper.qwentogemmatwo.layernorm.empty_other}{0}
\defmmnum{ws4c.mix.comp.indist.whisper.qwentogemmatwo.layernorm.other}{7.93}
\defmmnum{ws4c.mix.comp.indist.whisper.qwentogemmatwo.layernorm.other_cell}{7.93\%}
\defmmnum{ws4c.mix.comp.indist.whisper.qwentogemmatwo.layernorm.pair_cell}{3.7\% / 7.9\%}
\defmmnum{ws4c.mix.comp.indist.whisper.qwentogemmatwo.layernorm.short_clean_cell}{3.7\%}
\defmmnum{ws4c.mix.cross_done}{8}
\defmmnum{ws4c.mix.cross_max_clean}{480.30}
\defmmnum{ws4c.mix.cross_min_clean}{105.96}
\defmmnum{ws4c.mix.cross_total}{8}
\defmmnum{ws4c.mix.fit.moonshine.gemmatwotogemmathree.layernorm.anchors}{163{,}983}
\defmmnum{ws4c.mix.fit.moonshine.gemmatwotogemmathree.layernorm.audio_val}{0.1219}
\defmmnum{ws4c.mix.fit.moonshine.gemmatwotogemmathree.layernorm.cos}{0.1119}
\defmmnum{ws4c.mix.fit.moonshine.gemmatwotogemmathree.layernorm.ls_text_val}{0.1422}
\defmmnum{ws4c.mix.fit.moonshine.gemmatwotogemmathree.layernorm.n_test}{32{,}796}
\defmmnum{ws4c.mix.fit.moonshine.gemmatwotogemmathree.layernorm.p1}{44.61}
\defmmnum{ws4c.mix.fit.moonshine.gemmatwotogemmathree.layernorm.params}{5{,}910{,}528}
\defmmnum{ws4c.mix.fit.moonshine.gemmatwotogemmathree.layernorm.params_m}{5.91M}
\defmmnum{ws4c.mix.fit.moonshine.gemmatwotogemmathree.layernorm.rr}{1.080}
\defmmnum{ws4c.mix.fit.moonshine.gemmatwotogemmathree.layernorm.wiki_val}{0.0101}
\defmmnum{ws4c.mix.fit.moonshine.gemmatwotoqwen.layernorm.anchors}{83{,}378}
\defmmnum{ws4c.mix.fit.moonshine.gemmatwotoqwen.layernorm.audio_val}{0.2599}
\defmmnum{ws4c.mix.fit.moonshine.gemmatwotoqwen.layernorm.cos}{0.1243}
\defmmnum{ws4c.mix.fit.moonshine.gemmatwotoqwen.layernorm.ls_text_val}{0.3200}
\defmmnum{ws4c.mix.fit.moonshine.gemmatwotoqwen.layernorm.n_test}{16{,}675}
\defmmnum{ws4c.mix.fit.moonshine.gemmatwotoqwen.layernorm.p1}{31.83}
\defmmnum{ws4c.mix.fit.moonshine.gemmatwotoqwen.layernorm.params}{3{,}548{,}160}
\defmmnum{ws4c.mix.fit.moonshine.gemmatwotoqwen.layernorm.params_m}{3.55M}
\defmmnum{ws4c.mix.fit.moonshine.gemmatwotoqwen.layernorm.rr}{9.562}
\defmmnum{ws4c.mix.fit.moonshine.gemmatwotoqwen.layernorm.wiki_val}{0.0211}
\defmmnum{ws4c.mix.fit.moonshine.qwentogemmathree.layernorm.anchors}{82{,}663}
\defmmnum{ws4c.mix.fit.moonshine.qwentogemmathree.layernorm.audio_val}{0.1220}
\defmmnum{ws4c.mix.fit.moonshine.qwentogemmathree.layernorm.cos}{0.1213}
\defmmnum{ws4c.mix.fit.moonshine.qwentogemmathree.layernorm.ls_text_val}{0.3510}
\defmmnum{ws4c.mix.fit.moonshine.qwentogemmathree.layernorm.n_test}{16{,}532}
\defmmnum{ws4c.mix.fit.moonshine.qwentogemmathree.layernorm.p1}{59.51}
\defmmnum{ws4c.mix.fit.moonshine.qwentogemmathree.layernorm.params}{3{,}942{,}912}
\defmmnum{ws4c.mix.fit.moonshine.qwentogemmathree.layernorm.params_m}{3.94M}
\defmmnum{ws4c.mix.fit.moonshine.qwentogemmathree.layernorm.rr}{1.074}
\defmmnum{ws4c.mix.fit.moonshine.qwentogemmathree.layernorm.wiki_val}{0.0114}
\defmmnum{ws4c.mix.fit.moonshine.qwentogemmatwo.layernorm.anchors}{83{,}378}
\defmmnum{ws4c.mix.fit.moonshine.qwentogemmatwo.layernorm.audio_val}{0.1502}
\defmmnum{ws4c.mix.fit.moonshine.qwentogemmatwo.layernorm.cos}{0.1260}
\defmmnum{ws4c.mix.fit.moonshine.qwentogemmatwo.layernorm.ls_text_val}{0.2171}
\defmmnum{ws4c.mix.fit.moonshine.qwentogemmatwo.layernorm.n_test}{16{,}675}
\defmmnum{ws4c.mix.fit.moonshine.qwentogemmatwo.layernorm.p1}{57.84}
\defmmnum{ws4c.mix.fit.moonshine.qwentogemmatwo.layernorm.params}{3{,}548{,}928}
\defmmnum{ws4c.mix.fit.moonshine.qwentogemmatwo.layernorm.params_m}{3.55M}
\defmmnum{ws4c.mix.fit.moonshine.qwentogemmatwo.layernorm.rr}{1.045}
\defmmnum{ws4c.mix.fit.moonshine.qwentogemmatwo.layernorm.wiki_val}{0.0154}
\defmmnum{ws4c.mix.fit.whisper.gemmatwotogemmathree.layernorm.anchors}{163{,}983}
\defmmnum{ws4c.mix.fit.whisper.gemmatwotogemmathree.layernorm.audio_val}{0.0888}
\defmmnum{ws4c.mix.fit.whisper.gemmatwotogemmathree.layernorm.cos}{0.1082}
\defmmnum{ws4c.mix.fit.whisper.gemmatwotogemmathree.layernorm.ls_text_val}{0.1873}
\defmmnum{ws4c.mix.fit.whisper.gemmatwotogemmathree.layernorm.n_test}{32{,}796}
\defmmnum{ws4c.mix.fit.whisper.gemmatwotogemmathree.layernorm.p1}{44.66}
\defmmnum{ws4c.mix.fit.whisper.gemmatwotogemmathree.layernorm.params}{5{,}910{,}528}
\defmmnum{ws4c.mix.fit.whisper.gemmatwotogemmathree.layernorm.params_m}{5.91M}
\defmmnum{ws4c.mix.fit.whisper.gemmatwotogemmathree.layernorm.rr}{1.089}
\defmmnum{ws4c.mix.fit.whisper.gemmatwotogemmathree.layernorm.wiki_val}{0.0098}
\defmmnum{ws4c.mix.fit.whisper.gemmatwotoqwen.layernorm.anchors}{83{,}378}
\defmmnum{ws4c.mix.fit.whisper.gemmatwotoqwen.layernorm.audio_val}{0.1623}
\defmmnum{ws4c.mix.fit.whisper.gemmatwotoqwen.layernorm.cos}{0.1277}
\defmmnum{ws4c.mix.fit.whisper.gemmatwotoqwen.layernorm.ls_text_val}{0.3238}
\defmmnum{ws4c.mix.fit.whisper.gemmatwotoqwen.layernorm.n_test}{16{,}675}
\defmmnum{ws4c.mix.fit.whisper.gemmatwotoqwen.layernorm.p1}{25.24}
\defmmnum{ws4c.mix.fit.whisper.gemmatwotoqwen.layernorm.params}{3{,}548{,}160}
\defmmnum{ws4c.mix.fit.whisper.gemmatwotoqwen.layernorm.params_m}{3.55M}
\defmmnum{ws4c.mix.fit.whisper.gemmatwotoqwen.layernorm.rr}{8.661}
\defmmnum{ws4c.mix.fit.whisper.gemmatwotoqwen.layernorm.wiki_val}{0.0229}
\defmmnum{ws4c.mix.fit.whisper.qwentogemmathree.layernorm.anchors}{82{,}663}
\defmmnum{ws4c.mix.fit.whisper.qwentogemmathree.layernorm.audio_val}{0.0887}
\defmmnum{ws4c.mix.fit.whisper.qwentogemmathree.layernorm.cos}{0.1160}
\defmmnum{ws4c.mix.fit.whisper.qwentogemmathree.layernorm.ls_text_val}{0.3100}
\defmmnum{ws4c.mix.fit.whisper.qwentogemmathree.layernorm.n_test}{16{,}532}
\defmmnum{ws4c.mix.fit.whisper.qwentogemmathree.layernorm.p1}{58.98}
\defmmnum{ws4c.mix.fit.whisper.qwentogemmathree.layernorm.params}{3{,}942{,}912}
\defmmnum{ws4c.mix.fit.whisper.qwentogemmathree.layernorm.params_m}{3.94M}
\defmmnum{ws4c.mix.fit.whisper.qwentogemmathree.layernorm.rr}{1.076}
\defmmnum{ws4c.mix.fit.whisper.qwentogemmathree.layernorm.wiki_val}{0.0112}
\defmmnum{ws4c.mix.fit.whisper.qwentogemmatwo.layernorm.anchors}{83{,}378}
\defmmnum{ws4c.mix.fit.whisper.qwentogemmatwo.layernorm.audio_val}{0.1028}
\defmmnum{ws4c.mix.fit.whisper.qwentogemmatwo.layernorm.cos}{0.1212}
\defmmnum{ws4c.mix.fit.whisper.qwentogemmatwo.layernorm.ls_text_val}{0.1910}
\defmmnum{ws4c.mix.fit.whisper.qwentogemmatwo.layernorm.n_test}{16{,}675}
\defmmnum{ws4c.mix.fit.whisper.qwentogemmatwo.layernorm.p1}{56.03}
\defmmnum{ws4c.mix.fit.whisper.qwentogemmatwo.layernorm.params}{3{,}548{,}928}
\defmmnum{ws4c.mix.fit.whisper.qwentogemmatwo.layernorm.params_m}{3.55M}
\defmmnum{ws4c.mix.fit.whisper.qwentogemmatwo.layernorm.rr}{1.056}
\defmmnum{ws4c.mix.fit.whisper.qwentogemmatwo.layernorm.wiki_val}{0.0155}
\defmmnum{ws4c.mix.fits_done}{8}
\defmmnum{ws4c.mix.fits_total}{8}
\defmmnum{ws4c.mix.indist_done}{8}
\defmmnum{ws4c.mix.indist_max_clean}{8.03}
\defmmnum{ws4c.mix.indist_min_clean}{3.08}
\defmmnum{ws4c.mix.indist_total}{8}
\defmmnum{ws4c.mix.max_audio_val}{0.2599}
\defmmnum{ws4c.mix.max_cos}{0.1277}
\defmmnum{ws4c.mix.max_ls_val}{0.3510}
\defmmnum{ws4c.mix.max_p1}{59.51}
\defmmnum{ws4c.mix.max_wiki_val}{0.0229}
\defmmnum{ws4c.mix.min_audio_val}{0.0887}
\defmmnum{ws4c.mix.min_cos}{0.1082}
\defmmnum{ws4c.mix.min_ls_val}{0.1422}
\defmmnum{ws4c.mix.min_p1}{25.24}
\defmmnum{ws4c.mix.min_wiki_val}{0.0098}
\defmmnum{ws4c.mix.text_pct}{50}
\defmmnum{ws4c.mix_vs_audio.diff_triples}{3}
\defmmnum{ws4c.mix_vs_audio.max_delta_clean}{\ensuremath{+}0.16}
\defmmnum{ws4c.mix_vs_audio.min_delta_clean}{\ensuremath{-}1.01}
\defmmnum{ws4c.mix_vs_audio.moonshine.gemmatwotogemmathree.clean}{-0.08}
\defmmnum{ws4c.mix_vs_audio.moonshine.gemmatwotogemmathree.other}{-0.04}
\defmmnum{ws4c.mix_vs_audio.moonshine.gemmatwotoqwen.clean}{-1.01}
\defmmnum{ws4c.mix_vs_audio.moonshine.gemmatwotoqwen.other}{-1.46}
\defmmnum{ws4c.mix_vs_audio.moonshine.qwentogemmathree.clean}{-0.02}
\defmmnum{ws4c.mix_vs_audio.moonshine.qwentogemmathree.other}{-0.06}
\defmmnum{ws4c.mix_vs_audio.moonshine.qwentogemmatwo.clean}{-0.04}
\defmmnum{ws4c.mix_vs_audio.moonshine.qwentogemmatwo.other}{+0.26}
\defmmnum{ws4c.mix_vs_audio.outlier_abs_delta_clean}{1.01}
\defmmnum{ws4c.mix_vs_audio.outlier_delta_clean}{-1.01}
\defmmnum{ws4c.mix_vs_audio.tie_criterion_pp}{0.10}
\defmmnum{ws4c.mix_vs_audio.tied_max_abs_delta_clean}{0.08}
\defmmnum{ws4c.mix_vs_audio.tied_triples}{5}
\defmmnum{ws4c.mix_vs_audio.total_triples}{8}
\defmmnum{ws4c.mix_vs_audio.whisper.gemmatwotogemmathree.clean}{+0.04}
\defmmnum{ws4c.mix_vs_audio.whisper.gemmatwotogemmathree.other}{+0.18}
\defmmnum{ws4c.mix_vs_audio.whisper.gemmatwotoqwen.clean}{+0.01}
\defmmnum{ws4c.mix_vs_audio.whisper.gemmatwotoqwen.other}{-0.13}
\defmmnum{ws4c.mix_vs_audio.whisper.qwentogemmathree.clean}{-0.16}
\defmmnum{ws4c.mix_vs_audio.whisper.qwentogemmathree.other}{-0.40}
\defmmnum{ws4c.mix_vs_audio.whisper.qwentogemmatwo.clean}{+0.16}
\defmmnum{ws4c.mix_vs_audio.whisper.qwentogemmatwo.other}{-0.52}
\defmmnum{ws4c.wiki.audio_pct}{0}
\defmmnum{ws4c.wiki.comp.moonshine.gemmatwotogemmathree.layernorm.clean}{490.78}
\defmmnum{ws4c.wiki.comp.moonshine.gemmatwotogemmathree.layernorm.clean_cell}{490.78\%}
\defmmnum{ws4c.wiki.comp.moonshine.gemmatwotogemmathree.layernorm.empty_clean}{0}
\defmmnum{ws4c.wiki.comp.moonshine.gemmatwotogemmathree.layernorm.empty_other}{0}
\defmmnum{ws4c.wiki.comp.moonshine.gemmatwotogemmathree.layernorm.other}{562.71}
\defmmnum{ws4c.wiki.comp.moonshine.gemmatwotogemmathree.layernorm.other_cell}{562.71\%}
\defmmnum{ws4c.wiki.comp.moonshine.gemmatwotogemmathree.layernorm.pair_cell}{490.8\% / 562.7\%}
\defmmnum{ws4c.wiki.comp.moonshine.gemmatwotogemmathree.layernorm.short_clean_cell}{490.8\%}
\defmmnum{ws4c.wiki.comp.moonshine.gemmatwotoqwen.layernorm.clean}{99.65}
\defmmnum{ws4c.wiki.comp.moonshine.gemmatwotoqwen.layernorm.clean_cell}{99.65\%}
\defmmnum{ws4c.wiki.comp.moonshine.gemmatwotoqwen.layernorm.empty_clean}{0}
\defmmnum{ws4c.wiki.comp.moonshine.gemmatwotoqwen.layernorm.empty_other}{0}
\defmmnum{ws4c.wiki.comp.moonshine.gemmatwotoqwen.layernorm.other}{100.43}
\defmmnum{ws4c.wiki.comp.moonshine.gemmatwotoqwen.layernorm.other_cell}{100.43\%}
\defmmnum{ws4c.wiki.comp.moonshine.gemmatwotoqwen.layernorm.pair_cell}{99.6\% / 100.4\%}
\defmmnum{ws4c.wiki.comp.moonshine.gemmatwotoqwen.layernorm.short_clean_cell}{99.6\%}
\defmmnum{ws4c.wiki.comp.moonshine.qwentogemmathree.layernorm.clean}{450.57}
\defmmnum{ws4c.wiki.comp.moonshine.qwentogemmathree.layernorm.clean_cell}{450.57\%}
\defmmnum{ws4c.wiki.comp.moonshine.qwentogemmathree.layernorm.empty_clean}{0}
\defmmnum{ws4c.wiki.comp.moonshine.qwentogemmathree.layernorm.empty_other}{0}
\defmmnum{ws4c.wiki.comp.moonshine.qwentogemmathree.layernorm.other}{507.89}
\defmmnum{ws4c.wiki.comp.moonshine.qwentogemmathree.layernorm.other_cell}{507.89\%}
\defmmnum{ws4c.wiki.comp.moonshine.qwentogemmathree.layernorm.pair_cell}{450.6\% / 507.9\%}
\defmmnum{ws4c.wiki.comp.moonshine.qwentogemmathree.layernorm.short_clean_cell}{450.6\%}
\defmmnum{ws4c.wiki.comp.moonshine.qwentogemmatwo.layernorm.clean}{460.44}
\defmmnum{ws4c.wiki.comp.moonshine.qwentogemmatwo.layernorm.clean_cell}{460.44\%}
\defmmnum{ws4c.wiki.comp.moonshine.qwentogemmatwo.layernorm.empty_clean}{0}
\defmmnum{ws4c.wiki.comp.moonshine.qwentogemmatwo.layernorm.empty_other}{0}
\defmmnum{ws4c.wiki.comp.moonshine.qwentogemmatwo.layernorm.other}{513.95}
\defmmnum{ws4c.wiki.comp.moonshine.qwentogemmatwo.layernorm.other_cell}{513.95\%}
\defmmnum{ws4c.wiki.comp.moonshine.qwentogemmatwo.layernorm.pair_cell}{460.4\% / 513.9\%}
\defmmnum{ws4c.wiki.comp.moonshine.qwentogemmatwo.layernorm.short_clean_cell}{460.4\%}
\defmmnum{ws4c.wiki.comp.whisper.gemmatwotogemmathree.layernorm.clean}{475.17}
\defmmnum{ws4c.wiki.comp.whisper.gemmatwotogemmathree.layernorm.clean_cell}{475.17\%}
\defmmnum{ws4c.wiki.comp.whisper.gemmatwotogemmathree.layernorm.empty_clean}{0}
\defmmnum{ws4c.wiki.comp.whisper.gemmatwotogemmathree.layernorm.empty_other}{0}
\defmmnum{ws4c.wiki.comp.whisper.gemmatwotogemmathree.layernorm.other}{534.63}
\defmmnum{ws4c.wiki.comp.whisper.gemmatwotogemmathree.layernorm.other_cell}{534.63\%}
\defmmnum{ws4c.wiki.comp.whisper.gemmatwotogemmathree.layernorm.pair_cell}{475.2\% / 534.6\%}
\defmmnum{ws4c.wiki.comp.whisper.gemmatwotogemmathree.layernorm.short_clean_cell}{475.2\%}
\defmmnum{ws4c.wiki.comp.whisper.gemmatwotoqwen.layernorm.clean}{108.25}
\defmmnum{ws4c.wiki.comp.whisper.gemmatwotoqwen.layernorm.clean_cell}{108.25\%}
\defmmnum{ws4c.wiki.comp.whisper.gemmatwotoqwen.layernorm.empty_clean}{0}
\defmmnum{ws4c.wiki.comp.whisper.gemmatwotoqwen.layernorm.empty_other}{0}
\defmmnum{ws4c.wiki.comp.whisper.gemmatwotoqwen.layernorm.other}{113.56}
\defmmnum{ws4c.wiki.comp.whisper.gemmatwotoqwen.layernorm.other_cell}{113.56\%}
\defmmnum{ws4c.wiki.comp.whisper.gemmatwotoqwen.layernorm.pair_cell}{108.2\% / 113.6\%}
\defmmnum{ws4c.wiki.comp.whisper.gemmatwotoqwen.layernorm.short_clean_cell}{108.2\%}
\defmmnum{ws4c.wiki.comp.whisper.qwentogemmathree.layernorm.clean}{383.34}
\defmmnum{ws4c.wiki.comp.whisper.qwentogemmathree.layernorm.clean_cell}{383.34\%}
\defmmnum{ws4c.wiki.comp.whisper.qwentogemmathree.layernorm.empty_clean}{0}
\defmmnum{ws4c.wiki.comp.whisper.qwentogemmathree.layernorm.empty_other}{0}
\defmmnum{ws4c.wiki.comp.whisper.qwentogemmathree.layernorm.other}{440.11}
\defmmnum{ws4c.wiki.comp.whisper.qwentogemmathree.layernorm.other_cell}{440.11\%}
\defmmnum{ws4c.wiki.comp.whisper.qwentogemmathree.layernorm.pair_cell}{383.3\% / 440.1\%}
\defmmnum{ws4c.wiki.comp.whisper.qwentogemmathree.layernorm.short_clean_cell}{383.3\%}
\defmmnum{ws4c.wiki.comp.whisper.qwentogemmatwo.layernorm.clean}{436.11}
\defmmnum{ws4c.wiki.comp.whisper.qwentogemmatwo.layernorm.clean_cell}{436.11\%}
\defmmnum{ws4c.wiki.comp.whisper.qwentogemmatwo.layernorm.empty_clean}{0}
\defmmnum{ws4c.wiki.comp.whisper.qwentogemmatwo.layernorm.empty_other}{0}
\defmmnum{ws4c.wiki.comp.whisper.qwentogemmatwo.layernorm.other}{486.19}
\defmmnum{ws4c.wiki.comp.whisper.qwentogemmatwo.layernorm.other_cell}{486.19\%}
\defmmnum{ws4c.wiki.comp.whisper.qwentogemmatwo.layernorm.pair_cell}{436.1\% / 486.2\%}
\defmmnum{ws4c.wiki.comp.whisper.qwentogemmatwo.layernorm.short_clean_cell}{436.1\%}
\defmmnum{ws4c.wiki.comp_done}{8}
\defmmnum{ws4c.wiki.comp_total}{8}
\defmmnum{ws4c.wiki.fit.gemmatwotogemmathree.layernorm.anchors}{163{,}983}
\defmmnum{ws4c.wiki.fit.gemmatwotogemmathree.layernorm.cos}{0.1297}
\defmmnum{ws4c.wiki.fit.gemmatwotogemmathree.layernorm.ls_text_val}{0.1800}
\defmmnum{ws4c.wiki.fit.gemmatwotogemmathree.layernorm.n_test}{32{,}796}
\defmmnum{ws4c.wiki.fit.gemmatwotogemmathree.layernorm.p1}{52.91}
\defmmnum{ws4c.wiki.fit.gemmatwotogemmathree.layernorm.params}{5{,}910{,}528}
\defmmnum{ws4c.wiki.fit.gemmatwotogemmathree.layernorm.params_m}{5.91M}
\defmmnum{ws4c.wiki.fit.gemmatwotogemmathree.layernorm.rr}{1.030}
\defmmnum{ws4c.wiki.fit.gemmatwotogemmathree.layernorm.wiki_val}{0.0093}
\defmmnum{ws4c.wiki.fit.gemmatwotoqwen.layernorm.anchors}{83{,}378}
\defmmnum{ws4c.wiki.fit.gemmatwotoqwen.layernorm.cos}{0.1621}
\defmmnum{ws4c.wiki.fit.gemmatwotoqwen.layernorm.ls_text_val}{0.2760}
\defmmnum{ws4c.wiki.fit.gemmatwotoqwen.layernorm.n_test}{16{,}675}
\defmmnum{ws4c.wiki.fit.gemmatwotoqwen.layernorm.p1}{33.21}
\defmmnum{ws4c.wiki.fit.gemmatwotoqwen.layernorm.params}{3{,}548{,}160}
\defmmnum{ws4c.wiki.fit.gemmatwotoqwen.layernorm.params_m}{3.55M}
\defmmnum{ws4c.wiki.fit.gemmatwotoqwen.layernorm.rr}{4.419}
\defmmnum{ws4c.wiki.fit.gemmatwotoqwen.layernorm.wiki_val}{0.0182}
\defmmnum{ws4c.wiki.fit.qwentogemmathree.layernorm.anchors}{82{,}663}
\defmmnum{ws4c.wiki.fit.qwentogemmathree.layernorm.cos}{0.1358}
\defmmnum{ws4c.wiki.fit.qwentogemmathree.layernorm.ls_text_val}{0.2533}
\defmmnum{ws4c.wiki.fit.qwentogemmathree.layernorm.n_test}{16{,}532}
\defmmnum{ws4c.wiki.fit.qwentogemmathree.layernorm.p1}{68.19}
\defmmnum{ws4c.wiki.fit.qwentogemmathree.layernorm.params}{3{,}942{,}912}
\defmmnum{ws4c.wiki.fit.qwentogemmathree.layernorm.params_m}{3.94M}
\defmmnum{ws4c.wiki.fit.qwentogemmathree.layernorm.rr}{1.017}
\defmmnum{ws4c.wiki.fit.qwentogemmathree.layernorm.wiki_val}{0.0105}
\defmmnum{ws4c.wiki.fit.qwentogemmatwo.layernorm.anchors}{83{,}378}
\defmmnum{ws4c.wiki.fit.qwentogemmatwo.layernorm.cos}{0.1383}
\defmmnum{ws4c.wiki.fit.qwentogemmatwo.layernorm.ls_text_val}{0.1716}
\defmmnum{ws4c.wiki.fit.qwentogemmatwo.layernorm.n_test}{16{,}675}
\defmmnum{ws4c.wiki.fit.qwentogemmatwo.layernorm.p1}{64.03}
\defmmnum{ws4c.wiki.fit.qwentogemmatwo.layernorm.params}{3{,}548{,}928}
\defmmnum{ws4c.wiki.fit.qwentogemmatwo.layernorm.params_m}{3.55M}
\defmmnum{ws4c.wiki.fit.qwentogemmatwo.layernorm.rr}{1.031}
\defmmnum{ws4c.wiki.fit.qwentogemmatwo.layernorm.wiki_val}{0.0145}
\defmmnum{ws4c.wiki.fits_done}{4}
\defmmnum{ws4c.wiki.fits_total}{4}
\defmmnum{ws4c.wiki.max_clean}{490.78}
\defmmnum{ws4c.wiki.max_cos}{0.1621}
\defmmnum{ws4c.wiki.max_ls_val}{0.2760}
\defmmnum{ws4c.wiki.max_p1}{68.19}
\defmmnum{ws4c.wiki.max_val}{0.0182}
\defmmnum{ws4c.wiki.min_clean}{99.65}
\defmmnum{ws4c.wiki.min_cos}{0.1297}
\defmmnum{ws4c.wiki.min_ls_val}{0.1716}
\defmmnum{ws4c.wiki.min_p1}{33.21}
\defmmnum{ws4c.wiki.min_val}{0.0093}
\defmmnum{ws4d.b0.above_chance_hs}{+32.50}
\defmmnum{ws4d.b0.above_chance_piqa}{+28.56}
\defmmnum{ws4d.b0.delta_b0_hs}{+0.00}
\defmmnum{ws4d.b0.delta_b0_piqa}{+0.00}
\defmmnum{ws4d.b0.hs}{57.50}
\defmmnum{ws4d.b0.hs_cell}{57.50\%}
\defmmnum{ws4d.b0.hs_correct}{1{,}150}
\defmmnum{ws4d.b0.pair_cell}{78.56\% / 57.50\%}
\defmmnum{ws4d.b0.piqa}{78.56}
\defmmnum{ws4d.b0.piqa_cell}{78.56\%}
\defmmnum{ws4d.b0.piqa_correct}{1{,}444}
\defmmnum{ws4d.b0.ret_hs}{100.0}
\defmmnum{ws4d.b0.ret_piqa}{100.0}
\defmmnum{ws4d.b0.rr}{1.000}
\defmmnum{ws4d.best_trained.above_chance_hs}{+9.45}
\defmmnum{ws4d.best_trained.above_chance_piqa}{+9.47}
\defmmnum{ws4d.best_trained.hs}{34.45}
\defmmnum{ws4d.best_trained.piqa}{59.47}
\defmmnum{ws4d.c1.above_chance_hs}{+32.10}
\defmmnum{ws4d.c1.above_chance_piqa}{+27.15}
\defmmnum{ws4d.c1.anchors}{163{,}983}
\defmmnum{ws4d.c1.cos}{0.5768}
\defmmnum{ws4d.c1.delta_b0_hs}{-0.40}
\defmmnum{ws4d.c1.delta_b0_piqa}{-1.41}
\defmmnum{ws4d.c1.hs}{57.10}
\defmmnum{ws4d.c1.hs_cell}{57.10\%}
\defmmnum{ws4d.c1.hs_correct}{1{,}142}
\defmmnum{ws4d.c1.p1}{98.56}
\defmmnum{ws4d.c1.pair_cell}{77.15\% / 57.10\%}
\defmmnum{ws4d.c1.piqa}{77.15}
\defmmnum{ws4d.c1.piqa_cell}{77.15\%}
\defmmnum{ws4d.c1.piqa_correct}{1{,}418}
\defmmnum{ws4d.c1.ret_hs}{99.3}
\defmmnum{ws4d.c1.ret_piqa}{98.2}
\defmmnum{ws4d.c1.rows_overwritten}{163{,}977}
\defmmnum{ws4d.c1.rr}{1.000}
\defmmnum{ws4d.c2.above_chance_hs}{+29.40}
\defmmnum{ws4d.c2.above_chance_piqa}{+26.44}
\defmmnum{ws4d.c2.anchors}{163{,}983}
\defmmnum{ws4d.c2.cos}{0.5768}
\defmmnum{ws4d.c2.delta_b0_hs}{-3.10}
\defmmnum{ws4d.c2.delta_b0_piqa}{-2.12}
\defmmnum{ws4d.c2.hs}{54.40}
\defmmnum{ws4d.c2.hs_cell}{54.40\%}
\defmmnum{ws4d.c2.hs_correct}{1{,}088}
\defmmnum{ws4d.c2.p1}{98.56}
\defmmnum{ws4d.c2.pair_cell}{76.44\% / 54.40\%}
\defmmnum{ws4d.c2.piqa}{76.44}
\defmmnum{ws4d.c2.piqa_cell}{76.44\%}
\defmmnum{ws4d.c2.piqa_correct}{1{,}405}
\defmmnum{ws4d.c2.ret_hs}{94.6}
\defmmnum{ws4d.c2.ret_piqa}{97.3}
\defmmnum{ws4d.c2.rows_overwritten}{163{,}977}
\defmmnum{ws4d.c2.rr}{0.603}
\defmmnum{ws4d.c3.above_chance_hs}{+3.35}
\defmmnum{ws4d.c3.above_chance_piqa}{+4.08}
\defmmnum{ws4d.c3.anchors}{163{,}983}
\defmmnum{ws4d.c3.cos}{0.5080}
\defmmnum{ws4d.c3.delta_b0_hs}{-29.15}
\defmmnum{ws4d.c3.delta_b0_piqa}{-24.48}
\defmmnum{ws4d.c3.hs}{28.35}
\defmmnum{ws4d.c3.hs_cell}{28.35\%}
\defmmnum{ws4d.c3.hs_correct}{567}
\defmmnum{ws4d.c3.p1}{97.47}
\defmmnum{ws4d.c3.pair_cell}{54.08\% / 28.35\%}
\defmmnum{ws4d.c3.piqa}{54.08}
\defmmnum{ws4d.c3.piqa_cell}{54.08\%}
\defmmnum{ws4d.c3.piqa_correct}{994}
\defmmnum{ws4d.c3.ret_hs}{49.3}
\defmmnum{ws4d.c3.ret_piqa}{68.8}
\defmmnum{ws4d.c3.rows_overwritten}{163{,}977}
\defmmnum{ws4d.c3.rr}{1.000}
\defmmnum{ws4d.c4.above_chance_hs}{+0.05}
\defmmnum{ws4d.c4.above_chance_piqa}{-0.11}
\defmmnum{ws4d.c4.anchors}{163{,}983}
\defmmnum{ws4d.c4.cos}{0.0005}
\defmmnum{ws4d.c4.delta_b0_hs}{-32.45}
\defmmnum{ws4d.c4.delta_b0_piqa}{-28.67}
\defmmnum{ws4d.c4.hs}{25.05}
\defmmnum{ws4d.c4.hs_cell}{25.05\%}
\defmmnum{ws4d.c4.hs_correct}{501}
\defmmnum{ws4d.c4.p1}{0.00}
\defmmnum{ws4d.c4.pair_cell}{49.89\% / 25.05\%}
\defmmnum{ws4d.c4.piqa}{49.89}
\defmmnum{ws4d.c4.piqa_cell}{49.89\%}
\defmmnum{ws4d.c4.piqa_correct}{917}
\defmmnum{ws4d.c4.ret_hs}{43.6}
\defmmnum{ws4d.c4.ret_piqa}{63.5}
\defmmnum{ws4d.c4.rows_overwritten}{163{,}977}
\defmmnum{ws4d.c4.rr}{1.000}
\defmmnum{ws4d.closed_done_cells}{4}
\defmmnum{ws4d.closed_total_cells}{4}
\defmmnum{ws4d.delta.c1_vs_b0.hs}{-0.40}
\defmmnum{ws4d.delta.c1_vs_b0.piqa}{-1.41}
\defmmnum{ws4d.delta.t2_vs_c1.hs}{-22.90}
\defmmnum{ws4d.delta.t2_vs_c1.piqa}{-17.68}
\defmmnum{ws4d.delta.t3_vs_t2.hs}{+0.25}
\defmmnum{ws4d.delta.t3_vs_t2.piqa}{-1.74}
\defmmnum{ws4d.done_cells}{12}
\defmmnum{ws4d.e7_b0.hs}{57.50}
\defmmnum{ws4d.e7_b0.hs_cell}{57.50\%}
\defmmnum{ws4d.e7_b0.hs_correct}{1{,}150}
\defmmnum{ws4d.e7_b0.piqa}{78.56}
\defmmnum{ws4d.e7_b0.piqa_cell}{78.56\%}
\defmmnum{ws4d.e7_b0.piqa_correct}{1{,}444}
\defmmnum{ws4d.t1.above_chance_hs}{+5.40}
\defmmnum{ws4d.t1.above_chance_piqa}{+4.24}
\defmmnum{ws4d.t1.anchors}{163{,}983}
\defmmnum{ws4d.t1.cos}{0.1136}
\defmmnum{ws4d.t1.delta_b0_hs}{-27.10}
\defmmnum{ws4d.t1.delta_b0_piqa}{-24.32}
\defmmnum{ws4d.t1.hs}{30.40}
\defmmnum{ws4d.t1.hs_cell}{30.40\%}
\defmmnum{ws4d.t1.hs_correct}{608}
\defmmnum{ws4d.t1.p1}{36.11}
\defmmnum{ws4d.t1.pair_cell}{54.24\% / 30.40\%}
\defmmnum{ws4d.t1.piqa}{54.24}
\defmmnum{ws4d.t1.piqa_cell}{54.24\%}
\defmmnum{ws4d.t1.piqa_correct}{997}
\defmmnum{ws4d.t1.ret_hs}{52.9}
\defmmnum{ws4d.t1.ret_piqa}{69.0}
\defmmnum{ws4d.t1.rows_overwritten}{163{,}977}
\defmmnum{ws4d.t1.rr}{1.028}
\defmmnum{ws4d.t2.above_chance_hs}{+9.20}
\defmmnum{ws4d.t2.above_chance_piqa}{+9.47}
\defmmnum{ws4d.t2.anchors}{163{,}983}
\defmmnum{ws4d.t2.cos}{0.1297}
\defmmnum{ws4d.t2.delta_b0_hs}{-23.30}
\defmmnum{ws4d.t2.delta_b0_piqa}{-19.10}
\defmmnum{ws4d.t2.hs}{34.20}
\defmmnum{ws4d.t2.hs_cell}{34.20\%}
\defmmnum{ws4d.t2.hs_correct}{684}
\defmmnum{ws4d.t2.p1}{52.91}
\defmmnum{ws4d.t2.pair_cell}{59.47\% / 34.20\%}
\defmmnum{ws4d.t2.piqa}{59.47}
\defmmnum{ws4d.t2.piqa_cell}{59.47\%}
\defmmnum{ws4d.t2.piqa_correct}{1{,}093}
\defmmnum{ws4d.t2.ret_hs}{59.5}
\defmmnum{ws4d.t2.ret_piqa}{75.7}
\defmmnum{ws4d.t2.rows_overwritten}{163{,}977}
\defmmnum{ws4d.t2.rr}{1.030}
\defmmnum{ws4d.t3.above_chance_hs}{+9.45}
\defmmnum{ws4d.t3.above_chance_piqa}{+7.73}
\defmmnum{ws4d.t3.anchors}{163{,}983}
\defmmnum{ws4d.t3.cos}{0.1119}
\defmmnum{ws4d.t3.delta_b0_hs}{-23.05}
\defmmnum{ws4d.t3.delta_b0_piqa}{-20.84}
\defmmnum{ws4d.t3.hs}{34.45}
\defmmnum{ws4d.t3.hs_cell}{34.45\%}
\defmmnum{ws4d.t3.hs_correct}{689}
\defmmnum{ws4d.t3.p1}{44.61}
\defmmnum{ws4d.t3.pair_cell}{57.73\% / 34.45\%}
\defmmnum{ws4d.t3.piqa}{57.73}
\defmmnum{ws4d.t3.piqa_cell}{57.73\%}
\defmmnum{ws4d.t3.piqa_correct}{1{,}061}
\defmmnum{ws4d.t3.ret_hs}{59.9}
\defmmnum{ws4d.t3.ret_piqa}{73.5}
\defmmnum{ws4d.t3.rows_overwritten}{163{,}977}
\defmmnum{ws4d.t3.rr}{1.080}
\defmmnum{ws4d.t4.above_chance_hs}{+2.45}
\defmmnum{ws4d.t4.above_chance_piqa}{+1.90}
\defmmnum{ws4d.t4.anchors}{163{,}983}
\defmmnum{ws4d.t4.cos}{0.0339}
\defmmnum{ws4d.t4.delta_b0_hs}{-30.05}
\defmmnum{ws4d.t4.delta_b0_piqa}{-26.66}
\defmmnum{ws4d.t4.hs}{27.45}
\defmmnum{ws4d.t4.hs_cell}{27.45\%}
\defmmnum{ws4d.t4.hs_correct}{549}
\defmmnum{ws4d.t4.p1}{0.83}
\defmmnum{ws4d.t4.pair_cell}{51.90\% / 27.45\%}
\defmmnum{ws4d.t4.piqa}{51.90}
\defmmnum{ws4d.t4.piqa_cell}{51.90\%}
\defmmnum{ws4d.t4.piqa_correct}{954}
\defmmnum{ws4d.t4.ret_hs}{47.7}
\defmmnum{ws4d.t4.ret_piqa}{66.1}
\defmmnum{ws4d.t4.rows_overwritten}{163{,}977}
\defmmnum{ws4d.t4.rr}{1.058}
\defmmnum{ws4d.t5.above_chance_hs}{+0.50}
\defmmnum{ws4d.t5.above_chance_piqa}{+2.50}
\defmmnum{ws4d.t5.anchors}{163{,}983}
\defmmnum{ws4d.t5.cos}{0.0387}
\defmmnum{ws4d.t5.delta_b0_hs}{-32.00}
\defmmnum{ws4d.t5.delta_b0_piqa}{-26.22}
\defmmnum{ws4d.t5.hs}{25.50}
\defmmnum{ws4d.t5.hs_cell}{25.50\%}
\defmmnum{ws4d.t5.hs_correct}{510}
\defmmnum{ws4d.t5.p1}{1.27}
\defmmnum{ws4d.t5.pair_cell}{52.50\% / 25.50\%}
\defmmnum{ws4d.t5.piqa}{52.50}
\defmmnum{ws4d.t5.piqa_cell}{52.50\%}
\defmmnum{ws4d.t5.piqa_correct}{965}
\defmmnum{ws4d.t5.ret_hs}{44.3}
\defmmnum{ws4d.t5.ret_piqa}{66.7}
\defmmnum{ws4d.t5.rows_overwritten}{163{,}977}
\defmmnum{ws4d.t5.rr}{1.053}
\defmmnum{ws4d.t6.above_chance_hs}{+8.85}
\defmmnum{ws4d.t6.above_chance_piqa}{+7.56}
\defmmnum{ws4d.t6.anchors}{163{,}983}
\defmmnum{ws4d.t6.cos}{0.1082}
\defmmnum{ws4d.t6.delta_b0_hs}{-23.65}
\defmmnum{ws4d.t6.delta_b0_piqa}{-21.16}
\defmmnum{ws4d.t6.hs}{33.85}
\defmmnum{ws4d.t6.hs_cell}{33.85\%}
\defmmnum{ws4d.t6.hs_correct}{677}
\defmmnum{ws4d.t6.p1}{44.66}
\defmmnum{ws4d.t6.pair_cell}{57.56\% / 33.85\%}
\defmmnum{ws4d.t6.piqa}{57.56}
\defmmnum{ws4d.t6.piqa_cell}{57.56\%}
\defmmnum{ws4d.t6.piqa_correct}{1{,}058}
\defmmnum{ws4d.t6.ret_hs}{58.9}
\defmmnum{ws4d.t6.ret_piqa}{73.1}
\defmmnum{ws4d.t6.rows_overwritten}{163{,}977}
\defmmnum{ws4d.t6.rr}{1.090}
\defmmnum{ws4d.t7.above_chance_hs}{+1.45}
\defmmnum{ws4d.t7.above_chance_piqa}{+1.96}
\defmmnum{ws4d.t7.anchors}{163{,}983}
\defmmnum{ws4d.t7.cos}{0.0391}
\defmmnum{ws4d.t7.delta_b0_hs}{-31.05}
\defmmnum{ws4d.t7.delta_b0_piqa}{-26.61}
\defmmnum{ws4d.t7.hs}{26.45}
\defmmnum{ws4d.t7.hs_cell}{26.45\%}
\defmmnum{ws4d.t7.hs_correct}{529}
\defmmnum{ws4d.t7.p1}{1.21}
\defmmnum{ws4d.t7.pair_cell}{51.96\% / 26.45\%}
\defmmnum{ws4d.t7.piqa}{51.96}
\defmmnum{ws4d.t7.piqa_cell}{51.96\%}
\defmmnum{ws4d.t7.piqa_correct}{955}
\defmmnum{ws4d.t7.ret_hs}{46.0}
\defmmnum{ws4d.t7.ret_piqa}{66.1}
\defmmnum{ws4d.t7.rows_overwritten}{163{,}977}
\defmmnum{ws4d.t7.rr}{1.090}
\defmmnum{ws4d.text_bridges.max_cos}{0.1297}
\defmmnum{ws4d.text_bridges.min_cos}{0.1136}
\defmmnum{ws4d.total_cells}{12}
\defmmnum{ws4d.trained_done_cells}{7}
\defmmnum{ws4d.trained_total_cells}{7}
\defmmnum{ws4d.w1_b0.hs}{57.50}
\defmmnum{ws4d.w1_b0.hs_cell}{57.50\%}
\defmmnum{ws4d.w1_b0.hs_correct}{1{,}150}
\defmmnum{ws4d.w1_b0.piqa}{78.73}
\defmmnum{ws4d.w1_b0.piqa_cell}{78.73\%}
\defmmnum{ws4d.w1_b0.piqa_correct}{1{,}447}
\defmmnum{ws4e.direct.moonshine.adamw.med_clean}{0.00}
\defmmnum{ws4e.direct.moonshine.adamw.med_other}{8.33}
\defmmnum{ws4e.direct.moonshine.muon.med_clean}{0.00}
\defmmnum{ws4e.direct.moonshine.muon.med_other}{8.33}
\defmmnum{ws4e.direct.whisper.adamw.med_clean}{0.00}
\defmmnum{ws4e.direct.whisper.adamw.med_other}{4.35}
\defmmnum{ws4e.direct.whisper.muon.med_clean}{0.00}
\defmmnum{ws4e.direct.whisper.muon.med_other}{4.35}
\defmmnum{ws4e.fit.joint_best_step}{32{,}500}
\defmmnum{ws4e.fit.joint_best_val}{0.1052}
\defmmnum{ws4e.fit.joint_half_step}{17{,}578}
\defmmnum{ws4e.fit.joint_half_text_val}{6.79}
\defmmnum{ws4e.fit.joint_half_val}{0.1138}
\defmmnum{ws4e.fit.joint_moonshine_val}{0.1220}
\defmmnum{ws4e.fit.joint_samples_per_adapter_step}{8}
\defmmnum{ws4e.fit.joint_samples_per_step}{16}
\defmmnum{ws4e.fit.joint_text_val}{6.91}
\defmmnum{ws4e.fit.joint_total_steps}{35{,}156}
\defmmnum{ws4e.fit.joint_whisper_val}{0.0883}
\defmmnum{ws4e.fit.single_best_step}{17{,}500}
\defmmnum{ws4e.fit.single_steps}{17{,}578}
\defmmnum{ws4e.joint.moonshine.muon.clean}{4.27}
\defmmnum{ws4e.joint.moonshine.muon.clean_cell}{4.27\%}
\defmmnum{ws4e.joint.moonshine.muon.empty_clean}{0}
\defmmnum{ws4e.joint.moonshine.muon.empty_other}{0}
\defmmnum{ws4e.joint.moonshine.muon.med_clean}{0.00}
\defmmnum{ws4e.joint.moonshine.muon.med_other}{7.69}
\defmmnum{ws4e.joint.moonshine.muon.other}{10.66}
\defmmnum{ws4e.joint.moonshine.muon.other_cell}{10.66\%}
\defmmnum{ws4e.joint.whisper.muon.clean}{3.09}
\defmmnum{ws4e.joint.whisper.muon.clean_cell}{3.09\%}
\defmmnum{ws4e.joint.whisper.muon.empty_clean}{0}
\defmmnum{ws4e.joint.whisper.muon.empty_other}{0}
\defmmnum{ws4e.joint.whisper.muon.med_clean}{0.00}
\defmmnum{ws4e.joint.whisper.muon.med_other}{3.85}
\defmmnum{ws4e.joint.whisper.muon.other}{7.51}
\defmmnum{ws4e.joint.whisper.muon.other_cell}{7.51\%}
\defmmnum{ws4e.joint_ctrl.moonshine.adamw.clean}{12.78}
\defmmnum{ws4e.joint_ctrl.moonshine.adamw.clean_cell}{12.78\%}
\defmmnum{ws4e.joint_ctrl.moonshine.adamw.empty_clean}{4}
\defmmnum{ws4e.joint_ctrl.moonshine.adamw.empty_other}{3}
\defmmnum{ws4e.joint_ctrl.moonshine.adamw.med_clean}{5.13}
\defmmnum{ws4e.joint_ctrl.moonshine.adamw.med_other}{14.89}
\defmmnum{ws4e.joint_ctrl.moonshine.adamw.other}{23.26}
\defmmnum{ws4e.joint_ctrl.moonshine.adamw.other_cell}{23.26\%}
\defmmnum{ws4e.joint_ctrl.whisper.adamw.clean}{41.20}
\defmmnum{ws4e.joint_ctrl.whisper.adamw.clean_cell}{41.20\%}
\defmmnum{ws4e.joint_ctrl.whisper.adamw.empty_clean}{552}
\defmmnum{ws4e.joint_ctrl.whisper.adamw.empty_other}{584}
\defmmnum{ws4e.joint_ctrl.whisper.adamw.med_clean}{4.76}
\defmmnum{ws4e.joint_ctrl.whisper.adamw.med_other}{12.50}
\defmmnum{ws4e.joint_ctrl.whisper.adamw.other}{44.23}
\defmmnum{ws4e.joint_ctrl.whisper.adamw.other_cell}{44.23\%}
\defmmnum{ws4e.joint_half.moonshine.muon.clean}{4.52}
\defmmnum{ws4e.joint_half.moonshine.muon.clean_cell}{4.52\%}
\defmmnum{ws4e.joint_half.moonshine.muon.empty_clean}{0}
\defmmnum{ws4e.joint_half.moonshine.muon.empty_other}{0}
\defmmnum{ws4e.joint_half.moonshine.muon.med_clean}{0.00}
\defmmnum{ws4e.joint_half.moonshine.muon.med_other}{8.33}
\defmmnum{ws4e.joint_half.moonshine.muon.other}{11.27}
\defmmnum{ws4e.joint_half.moonshine.muon.other_cell}{11.27\%}
\defmmnum{ws4e.joint_half.trail_vs_single.moonshine.clean}{0.23}
\defmmnum{ws4e.joint_half.trail_vs_single.whisper.clean}{0.22}
\defmmnum{ws4e.joint_half.whisper.muon.clean}{3.44}
\defmmnum{ws4e.joint_half.whisper.muon.clean_cell}{3.44\%}
\defmmnum{ws4e.joint_half.whisper.muon.empty_clean}{1}
\defmmnum{ws4e.joint_half.whisper.muon.empty_other}{0}
\defmmnum{ws4e.joint_half.whisper.muon.med_clean}{0.00}
\defmmnum{ws4e.joint_half.whisper.muon.med_other}{4.55}
\defmmnum{ws4e.joint_half.whisper.muon.other}{7.75}
\defmmnum{ws4e.joint_half.whisper.muon.other_cell}{7.75\%}
\defmmnum{ws4e.single_ctrl.moonshine.adamw.clean}{6.54}
\defmmnum{ws4e.single_ctrl.moonshine.adamw.clean_cell}{6.54\%}
\defmmnum{ws4e.single_ctrl.moonshine.adamw.empty_clean}{0}
\defmmnum{ws4e.single_ctrl.moonshine.adamw.empty_other}{0}
\defmmnum{ws4e.single_ctrl.moonshine.adamw.med_clean}{3.57}
\defmmnum{ws4e.single_ctrl.moonshine.adamw.med_other}{11.11}
\defmmnum{ws4e.single_ctrl.moonshine.adamw.other}{13.74}
\defmmnum{ws4e.single_ctrl.moonshine.adamw.other_cell}{13.74\%}
\defmmnum{ws4e.single_ctrl.whisper.adamw.clean}{40.49}
\defmmnum{ws4e.single_ctrl.whisper.adamw.clean_cell}{40.49\%}
\defmmnum{ws4e.single_ctrl.whisper.adamw.empty_clean}{286}
\defmmnum{ws4e.single_ctrl.whisper.adamw.empty_other}{263}
\defmmnum{ws4e.single_ctrl.whisper.adamw.med_clean}{5.26}
\defmmnum{ws4e.single_ctrl.whisper.adamw.med_other}{12.50}
\defmmnum{ws4e.single_ctrl.whisper.adamw.other}{52.23}
\defmmnum{ws4e.single_ctrl.whisper.adamw.other_cell}{52.23\%}
\defmmnum{ws4e.single_same.moonshine.muon.med_clean}{0.00}
\defmmnum{ws4e.single_same.moonshine.muon.med_other}{7.69}
\defmmnum{ws4e.single_same.whisper.muon.med_clean}{0.00}
\defmmnum{ws4e.single_same.whisper.muon.med_other}{3.85}
\defmmnum{xid.cells_in_window}{2}
\defmmnum{xid.earliest_log_date}{2026-10-11}
\defmmnum{xid.files_in_window}{5}
\defmmnum{xid.files_quarantined}{5}
\defmmnum{xid.gpu03.boot1_xids}{214}
\defmmnum{xid.gpu03.boot2_xids}{73}
\defmmnum{xid.gpu03.clean_reboot_utc}{2026-11-14T06:20:17Z}
\defmmnum{xid.gpu03.fault_boots}{2}
\defmmnum{xid.gpu03.first_xid_utc}{2026-11-13T03:41:59Z}
\defmmnum{xid.gpu03.last_xid_utc}{2026-11-13T20:50:26Z}
\defmmnum{xid.gpu03.post_reboot_xids}{0}
\defmmnum{xid.gpu03.total_xids}{287}
\defmmnum{xid.hosts_audited}{15}
\defmmnum{xid.hosts_reachable}{15}
\defmmnum{xid.hosts_with_xids}{1}
\defmmnum{xid.hosts_zero_xids}{14}
\defmmnum{xid.log_rotations_per_host}{5}
\defmmnum{xid.unquarantined_affected}{0}
\defmmnum{xprofile.a1.clean.cap159_n}{0}
\defmmnum{xprofile.a1.clean.cap159_pct}{0.00}
\defmmnum{xprofile.a1.clean.cap160_n}{0}
\defmmnum{xprofile.a1.clean.cap160_pct}{0.00}
\defmmnum{xprofile.a1.clean.del_pct}{1.15}
\defmmnum{xprofile.a1.clean.del_words}{607}
\defmmnum{xprofile.a1.clean.empty_pct}{0.00}
\defmmnum{xprofile.a1.clean.ins_pct}{1.02}
\defmmnum{xprofile.a1.clean.ins_words}{537}
\defmmnum{xprofile.a1.clean.n_empty}{0}
\defmmnum{xprofile.a1.clean.n_utt}{2{,}620}
\defmmnum{xprofile.a1.clean.nonempty_n}{2{,}620}
\defmmnum{xprofile.a1.clean.nonempty_utt_median_wer}{6.82}
\defmmnum{xprofile.a1.clean.nonempty_wer}{9.04}
\defmmnum{xprofile.a1.clean.sub_pct}{6.87}
\defmmnum{xprofile.a1.clean.sub_words}{3{,}610}
\defmmnum{xprofile.a1.clean.utt_eq0_n}{867}
\defmmnum{xprofile.a1.clean.utt_eq0_pct}{33.09}
\defmmnum{xprofile.a1.clean.utt_ge100_n}{6}
\defmmnum{xprofile.a1.clean.utt_ge100_pct}{0.23}
\defmmnum{xprofile.a1.clean.utt_lt10_n}{1{,}599}
\defmmnum{xprofile.a1.clean.utt_lt10_pct}{61.03}
\defmmnum{xprofile.a1.clean.utt_lt50_n}{2{,}565}
\defmmnum{xprofile.a1.clean.utt_lt50_pct}{97.90}
\defmmnum{xprofile.a1.clean.utt_median_wer}{6.82}
\defmmnum{xprofile.a1.clean.wer}{9.04}
\defmmnum{xprofile.a1.clean.word_ratio_corpus}{1.00}
\defmmnum{xprofile.a1.clean.word_ratio_median}{1.00}
\defmmnum{xprofile.a1.other.cap159_n}{1}
\defmmnum{xprofile.a1.other.cap159_pct}{0.03}
\defmmnum{xprofile.a1.other.cap160_n}{1}
\defmmnum{xprofile.a1.other.cap160_pct}{0.03}
\defmmnum{xprofile.a1.other.del_pct}{2.84}
\defmmnum{xprofile.a1.other.del_words}{1{,}485}
\defmmnum{xprofile.a1.other.empty_pct}{0.00}
\defmmnum{xprofile.a1.other.ins_pct}{2.16}
\defmmnum{xprofile.a1.other.ins_words}{1{,}128}
\defmmnum{xprofile.a1.other.n_empty}{0}
\defmmnum{xprofile.a1.other.n_utt}{2{,}939}
\defmmnum{xprofile.a1.other.nonempty_n}{2{,}939}
\defmmnum{xprofile.a1.other.nonempty_utt_median_wer}{16.67}
\defmmnum{xprofile.a1.other.nonempty_wer}{18.91}
\defmmnum{xprofile.a1.other.sub_pct}{13.92}
\defmmnum{xprofile.a1.other.sub_words}{7{,}286}
\defmmnum{xprofile.a1.other.utt_eq0_n}{515}
\defmmnum{xprofile.a1.other.utt_eq0_pct}{17.52}
\defmmnum{xprofile.a1.other.utt_ge100_n}{30}
\defmmnum{xprofile.a1.other.utt_ge100_pct}{1.02}
\defmmnum{xprofile.a1.other.utt_lt10_n}{922}
\defmmnum{xprofile.a1.other.utt_lt10_pct}{31.37}
\defmmnum{xprofile.a1.other.utt_lt50_n}{2{,}657}
\defmmnum{xprofile.a1.other.utt_lt50_pct}{90.40}
\defmmnum{xprofile.a1.other.utt_median_wer}{16.67}
\defmmnum{xprofile.a1.other.wer}{18.91}
\defmmnum{xprofile.a1.other.word_ratio_corpus}{0.99}
\defmmnum{xprofile.a1.other.word_ratio_median}{1.00}
\defmmnum{xprofile.a1_cross.clean.cap159_n}{3}
\defmmnum{xprofile.a1_cross.clean.cap159_pct}{0.11}
\defmmnum{xprofile.a1_cross.clean.cap160_n}{3}
\defmmnum{xprofile.a1_cross.clean.cap160_pct}{0.11}
\defmmnum{xprofile.a1_cross.clean.del_pct}{38.82}
\defmmnum{xprofile.a1_cross.clean.del_words}{20{,}409}
\defmmnum{xprofile.a1_cross.clean.empty_pct}{0.00}
\defmmnum{xprofile.a1_cross.clean.ins_pct}{17.22}
\defmmnum{xprofile.a1_cross.clean.ins_words}{9{,}051}
\defmmnum{xprofile.a1_cross.clean.n_empty}{0}
\defmmnum{xprofile.a1_cross.clean.n_utt}{2{,}620}
\defmmnum{xprofile.a1_cross.clean.nonempty_n}{2{,}620}
\defmmnum{xprofile.a1_cross.clean.nonempty_utt_median_wer}{98.00}
\defmmnum{xprofile.a1_cross.clean.nonempty_wer}{98.63}
\defmmnum{xprofile.a1_cross.clean.sub_pct}{42.60}
\defmmnum{xprofile.a1_cross.clean.sub_words}{22{,}398}
\defmmnum{xprofile.a1_cross.clean.utt_eq0_n}{15}
\defmmnum{xprofile.a1_cross.clean.utt_eq0_pct}{0.57}
\defmmnum{xprofile.a1_cross.clean.utt_ge100_n}{1{,}273}
\defmmnum{xprofile.a1_cross.clean.utt_ge100_pct}{48.59}
\defmmnum{xprofile.a1_cross.clean.utt_lt10_n}{20}
\defmmnum{xprofile.a1_cross.clean.utt_lt10_pct}{0.76}
\defmmnum{xprofile.a1_cross.clean.utt_lt50_n}{362}
\defmmnum{xprofile.a1_cross.clean.utt_lt50_pct}{13.82}
\defmmnum{xprofile.a1_cross.clean.utt_median_wer}{98.00}
\defmmnum{xprofile.a1_cross.clean.wer}{98.63}
\defmmnum{xprofile.a1_cross.clean.word_ratio_corpus}{0.78}
\defmmnum{xprofile.a1_cross.clean.word_ratio_median}{0.83}
\defmmnum{xprofile.a1_cross.other.cap159_n}{1}
\defmmnum{xprofile.a1_cross.other.cap159_pct}{0.03}
\defmmnum{xprofile.a1_cross.other.cap160_n}{1}
\defmmnum{xprofile.a1_cross.other.cap160_pct}{0.03}
\defmmnum{xprofile.a1_cross.other.del_pct}{39.24}
\defmmnum{xprofile.a1_cross.other.del_words}{20{,}537}
\defmmnum{xprofile.a1_cross.other.empty_pct}{0.00}
\defmmnum{xprofile.a1_cross.other.ins_pct}{22.31}
\defmmnum{xprofile.a1_cross.other.ins_words}{11{,}679}
\defmmnum{xprofile.a1_cross.other.n_empty}{0}
\defmmnum{xprofile.a1_cross.other.n_utt}{2{,}939}
\defmmnum{xprofile.a1_cross.other.nonempty_n}{2{,}939}
\defmmnum{xprofile.a1_cross.other.nonempty_utt_median_wer}{100.00}
\defmmnum{xprofile.a1_cross.other.nonempty_wer}{107.15}
\defmmnum{xprofile.a1_cross.other.sub_pct}{45.61}
\defmmnum{xprofile.a1_cross.other.sub_words}{23{,}872}
\defmmnum{xprofile.a1_cross.other.utt_eq0_n}{12}
\defmmnum{xprofile.a1_cross.other.utt_eq0_pct}{0.41}
\defmmnum{xprofile.a1_cross.other.utt_ge100_n}{1{,}616}
\defmmnum{xprofile.a1_cross.other.utt_ge100_pct}{54.98}
\defmmnum{xprofile.a1_cross.other.utt_lt10_n}{12}
\defmmnum{xprofile.a1_cross.other.utt_lt10_pct}{0.41}
\defmmnum{xprofile.a1_cross.other.utt_lt50_n}{244}
\defmmnum{xprofile.a1_cross.other.utt_lt50_pct}{8.30}
\defmmnum{xprofile.a1_cross.other.utt_median_wer}{100.00}
\defmmnum{xprofile.a1_cross.other.wer}{107.15}
\defmmnum{xprofile.a1_cross.other.word_ratio_corpus}{0.83}
\defmmnum{xprofile.a1_cross.other.word_ratio_median}{0.83}
\defmmnum{xprofile.a2.clean.cap159_n}{0}
\defmmnum{xprofile.a2.clean.cap159_pct}{0.00}
\defmmnum{xprofile.a2.clean.cap160_n}{0}
\defmmnum{xprofile.a2.clean.cap160_pct}{0.00}
\defmmnum{xprofile.a2.clean.del_pct}{1.24}
\defmmnum{xprofile.a2.clean.del_words}{654}
\defmmnum{xprofile.a2.clean.empty_pct}{0.00}
\defmmnum{xprofile.a2.clean.ins_pct}{0.98}
\defmmnum{xprofile.a2.clean.ins_words}{513}
\defmmnum{xprofile.a2.clean.n_empty}{0}
\defmmnum{xprofile.a2.clean.n_utt}{2{,}620}
\defmmnum{xprofile.a2.clean.nonempty_n}{2{,}620}
\defmmnum{xprofile.a2.clean.nonempty_utt_median_wer}{6.90}
\defmmnum{xprofile.a2.clean.nonempty_wer}{9.03}
\defmmnum{xprofile.a2.clean.sub_pct}{6.81}
\defmmnum{xprofile.a2.clean.sub_words}{3{,}582}
\defmmnum{xprofile.a2.clean.utt_eq0_n}{859}
\defmmnum{xprofile.a2.clean.utt_eq0_pct}{32.79}
\defmmnum{xprofile.a2.clean.utt_ge100_n}{8}
\defmmnum{xprofile.a2.clean.utt_ge100_pct}{0.31}
\defmmnum{xprofile.a2.clean.utt_lt10_n}{1{,}579}
\defmmnum{xprofile.a2.clean.utt_lt10_pct}{60.27}
\defmmnum{xprofile.a2.clean.utt_lt50_n}{2{,}568}
\defmmnum{xprofile.a2.clean.utt_lt50_pct}{98.02}
\defmmnum{xprofile.a2.clean.utt_median_wer}{6.90}
\defmmnum{xprofile.a2.clean.wer}{9.03}
\defmmnum{xprofile.a2.clean.word_ratio_corpus}{1.00}
\defmmnum{xprofile.a2.clean.word_ratio_median}{1.00}
\defmmnum{xprofile.a2.other.cap159_n}{1}
\defmmnum{xprofile.a2.other.cap159_pct}{0.03}
\defmmnum{xprofile.a2.other.cap160_n}{1}
\defmmnum{xprofile.a2.other.cap160_pct}{0.03}
\defmmnum{xprofile.a2.other.del_pct}{2.55}
\defmmnum{xprofile.a2.other.del_words}{1{,}335}
\defmmnum{xprofile.a2.other.empty_pct}{0.00}
\defmmnum{xprofile.a2.other.ins_pct}{2.27}
\defmmnum{xprofile.a2.other.ins_words}{1{,}186}
\defmmnum{xprofile.a2.other.n_empty}{0}
\defmmnum{xprofile.a2.other.n_utt}{2{,}939}
\defmmnum{xprofile.a2.other.nonempty_n}{2{,}939}
\defmmnum{xprofile.a2.other.nonempty_utt_median_wer}{16.67}
\defmmnum{xprofile.a2.other.nonempty_wer}{18.79}
\defmmnum{xprofile.a2.other.sub_pct}{13.97}
\defmmnum{xprofile.a2.other.sub_words}{7{,}313}
\defmmnum{xprofile.a2.other.utt_eq0_n}{486}
\defmmnum{xprofile.a2.other.utt_eq0_pct}{16.54}
\defmmnum{xprofile.a2.other.utt_ge100_n}{32}
\defmmnum{xprofile.a2.other.utt_ge100_pct}{1.09}
\defmmnum{xprofile.a2.other.utt_lt10_n}{919}
\defmmnum{xprofile.a2.other.utt_lt10_pct}{31.27}
\defmmnum{xprofile.a2.other.utt_lt50_n}{2{,}643}
\defmmnum{xprofile.a2.other.utt_lt50_pct}{89.93}
\defmmnum{xprofile.a2.other.utt_median_wer}{16.67}
\defmmnum{xprofile.a2.other.wer}{18.79}
\defmmnum{xprofile.a2.other.word_ratio_corpus}{1.00}
\defmmnum{xprofile.a2.other.word_ratio_median}{1.00}
\defmmnum{xprofile.a3.clean.cap159_n}{0}
\defmmnum{xprofile.a3.clean.cap159_pct}{0.00}
\defmmnum{xprofile.a3.clean.cap160_n}{0}
\defmmnum{xprofile.a3.clean.cap160_pct}{0.00}
\defmmnum{xprofile.a3.clean.del_pct}{0.56}
\defmmnum{xprofile.a3.clean.del_words}{297}
\defmmnum{xprofile.a3.clean.empty_pct}{0.00}
\defmmnum{xprofile.a3.clean.ins_pct}{0.54}
\defmmnum{xprofile.a3.clean.ins_words}{282}
\defmmnum{xprofile.a3.clean.n_empty}{0}
\defmmnum{xprofile.a3.clean.n_utt}{2{,}620}
\defmmnum{xprofile.a3.clean.nonempty_n}{2{,}620}
\defmmnum{xprofile.a3.clean.nonempty_utt_median_wer}{1.63}
\defmmnum{xprofile.a3.clean.nonempty_wer}{5.12}
\defmmnum{xprofile.a3.clean.sub_pct}{4.02}
\defmmnum{xprofile.a3.clean.sub_words}{2{,}115}
\defmmnum{xprofile.a3.clean.utt_eq0_n}{1{,}304}
\defmmnum{xprofile.a3.clean.utt_eq0_pct}{49.77}
\defmmnum{xprofile.a3.clean.utt_ge100_n}{1}
\defmmnum{xprofile.a3.clean.utt_ge100_pct}{0.04}
\defmmnum{xprofile.a3.clean.utt_lt10_n}{2{,}072}
\defmmnum{xprofile.a3.clean.utt_lt10_pct}{79.08}
\defmmnum{xprofile.a3.clean.utt_lt50_n}{2{,}596}
\defmmnum{xprofile.a3.clean.utt_lt50_pct}{99.08}
\defmmnum{xprofile.a3.clean.utt_median_wer}{1.63}
\defmmnum{xprofile.a3.clean.wer}{5.12}
\defmmnum{xprofile.a3.clean.word_ratio_corpus}{1.00}
\defmmnum{xprofile.a3.clean.word_ratio_median}{1.00}
\defmmnum{xprofile.a3.other.cap159_n}{0}
\defmmnum{xprofile.a3.other.cap159_pct}{0.00}
\defmmnum{xprofile.a3.other.cap160_n}{0}
\defmmnum{xprofile.a3.other.cap160_pct}{0.00}
\defmmnum{xprofile.a3.other.del_pct}{1.40}
\defmmnum{xprofile.a3.other.del_words}{733}
\defmmnum{xprofile.a3.other.empty_pct}{0.00}
\defmmnum{xprofile.a3.other.ins_pct}{1.28}
\defmmnum{xprofile.a3.other.ins_words}{669}
\defmmnum{xprofile.a3.other.n_empty}{0}
\defmmnum{xprofile.a3.other.n_utt}{2{,}939}
\defmmnum{xprofile.a3.other.nonempty_n}{2{,}939}
\defmmnum{xprofile.a3.other.nonempty_utt_median_wer}{9.09}
\defmmnum{xprofile.a3.other.nonempty_wer}{11.28}
\defmmnum{xprofile.a3.other.sub_pct}{8.60}
\defmmnum{xprofile.a3.other.sub_words}{4{,}501}
\defmmnum{xprofile.a3.other.utt_eq0_n}{899}
\defmmnum{xprofile.a3.other.utt_eq0_pct}{30.59}
\defmmnum{xprofile.a3.other.utt_ge100_n}{18}
\defmmnum{xprofile.a3.other.utt_ge100_pct}{0.61}
\defmmnum{xprofile.a3.other.utt_lt10_n}{1{,}542}
\defmmnum{xprofile.a3.other.utt_lt10_pct}{52.47}
\defmmnum{xprofile.a3.other.utt_lt50_n}{2{,}793}
\defmmnum{xprofile.a3.other.utt_lt50_pct}{95.03}
\defmmnum{xprofile.a3.other.utt_median_wer}{9.09}
\defmmnum{xprofile.a3.other.wer}{11.28}
\defmmnum{xprofile.a3.other.word_ratio_corpus}{1.00}
\defmmnum{xprofile.a3.other.word_ratio_median}{1.00}
\defmmnum{xprofile.a3_cross.clean.cap159_n}{2}
\defmmnum{xprofile.a3_cross.clean.cap159_pct}{0.08}
\defmmnum{xprofile.a3_cross.clean.cap160_n}{2}
\defmmnum{xprofile.a3_cross.clean.cap160_pct}{0.08}
\defmmnum{xprofile.a3_cross.clean.del_pct}{55.89}
\defmmnum{xprofile.a3_cross.clean.del_words}{29{,}385}
\defmmnum{xprofile.a3_cross.clean.empty_pct}{0.00}
\defmmnum{xprofile.a3_cross.clean.ins_pct}{28.56}
\defmmnum{xprofile.a3_cross.clean.ins_words}{15{,}017}
\defmmnum{xprofile.a3_cross.clean.n_empty}{0}
\defmmnum{xprofile.a3_cross.clean.n_utt}{2{,}620}
\defmmnum{xprofile.a3_cross.clean.nonempty_n}{2{,}620}
\defmmnum{xprofile.a3_cross.clean.nonempty_utt_median_wer}{100.00}
\defmmnum{xprofile.a3_cross.clean.nonempty_wer}{123.83}
\defmmnum{xprofile.a3_cross.clean.sub_pct}{39.38}
\defmmnum{xprofile.a3_cross.clean.sub_words}{20{,}702}
\defmmnum{xprofile.a3_cross.clean.utt_eq0_n}{1}
\defmmnum{xprofile.a3_cross.clean.utt_eq0_pct}{0.04}
\defmmnum{xprofile.a3_cross.clean.utt_ge100_n}{1{,}776}
\defmmnum{xprofile.a3_cross.clean.utt_ge100_pct}{67.79}
\defmmnum{xprofile.a3_cross.clean.utt_lt10_n}{1}
\defmmnum{xprofile.a3_cross.clean.utt_lt10_pct}{0.04}
\defmmnum{xprofile.a3_cross.clean.utt_lt50_n}{15}
\defmmnum{xprofile.a3_cross.clean.utt_lt50_pct}{0.57}
\defmmnum{xprofile.a3_cross.clean.utt_median_wer}{100.00}
\defmmnum{xprofile.a3_cross.clean.wer}{123.83}
\defmmnum{xprofile.a3_cross.clean.word_ratio_corpus}{0.73}
\defmmnum{xprofile.a3_cross.clean.word_ratio_median}{0.57}
\defmmnum{xprofile.a3_cross.other.cap159_n}{2}
\defmmnum{xprofile.a3_cross.other.cap159_pct}{0.07}
\defmmnum{xprofile.a3_cross.other.cap160_n}{2}
\defmmnum{xprofile.a3_cross.other.cap160_pct}{0.07}
\defmmnum{xprofile.a3_cross.other.del_pct}{52.22}
\defmmnum{xprofile.a3_cross.other.del_words}{27{,}332}
\defmmnum{xprofile.a3_cross.other.empty_pct}{0.00}
\defmmnum{xprofile.a3_cross.other.ins_pct}{34.48}
\defmmnum{xprofile.a3_cross.other.ins_words}{18{,}049}
\defmmnum{xprofile.a3_cross.other.n_empty}{0}
\defmmnum{xprofile.a3_cross.other.n_utt}{2{,}939}
\defmmnum{xprofile.a3_cross.other.nonempty_n}{2{,}939}
\defmmnum{xprofile.a3_cross.other.nonempty_utt_median_wer}{100.00}
\defmmnum{xprofile.a3_cross.other.nonempty_wer}{130.33}
\defmmnum{xprofile.a3_cross.other.sub_pct}{43.63}
\defmmnum{xprofile.a3_cross.other.sub_words}{22{,}836}
\defmmnum{xprofile.a3_cross.other.utt_eq0_n}{0}
\defmmnum{xprofile.a3_cross.other.utt_eq0_pct}{0.00}
\defmmnum{xprofile.a3_cross.other.utt_ge100_n}{2{,}142}
\defmmnum{xprofile.a3_cross.other.utt_ge100_pct}{72.88}
\defmmnum{xprofile.a3_cross.other.utt_lt10_n}{0}
\defmmnum{xprofile.a3_cross.other.utt_lt10_pct}{0.00}
\defmmnum{xprofile.a3_cross.other.utt_lt50_n}{3}
\defmmnum{xprofile.a3_cross.other.utt_lt50_pct}{0.10}
\defmmnum{xprofile.a3_cross.other.utt_median_wer}{100.00}
\defmmnum{xprofile.a3_cross.other.wer}{130.33}
\defmmnum{xprofile.a3_cross.other.word_ratio_corpus}{0.82}
\defmmnum{xprofile.a3_cross.other.word_ratio_median}{0.67}
\defmmnum{xprofile.a4.clean.cap159_n}{0}
\defmmnum{xprofile.a4.clean.cap159_pct}{0.00}
\defmmnum{xprofile.a4.clean.cap160_n}{0}
\defmmnum{xprofile.a4.clean.cap160_pct}{0.00}
\defmmnum{xprofile.a4.clean.del_pct}{0.38}
\defmmnum{xprofile.a4.clean.del_words}{200}
\defmmnum{xprofile.a4.clean.empty_pct}{0.00}
\defmmnum{xprofile.a4.clean.ins_pct}{0.51}
\defmmnum{xprofile.a4.clean.ins_words}{266}
\defmmnum{xprofile.a4.clean.n_empty}{0}
\defmmnum{xprofile.a4.clean.n_utt}{2{,}620}
\defmmnum{xprofile.a4.clean.nonempty_n}{2{,}620}
\defmmnum{xprofile.a4.clean.nonempty_utt_median_wer}{0.00}
\defmmnum{xprofile.a4.clean.nonempty_wer}{4.29}
\defmmnum{xprofile.a4.clean.sub_pct}{3.40}
\defmmnum{xprofile.a4.clean.sub_words}{1{,}790}
\defmmnum{xprofile.a4.clean.utt_eq0_n}{1{,}454}
\defmmnum{xprofile.a4.clean.utt_eq0_pct}{55.50}
\defmmnum{xprofile.a4.clean.utt_ge100_n}{6}
\defmmnum{xprofile.a4.clean.utt_ge100_pct}{0.23}
\defmmnum{xprofile.a4.clean.utt_lt10_n}{2{,}121}
\defmmnum{xprofile.a4.clean.utt_lt10_pct}{80.95}
\defmmnum{xprofile.a4.clean.utt_lt50_n}{2{,}589}
\defmmnum{xprofile.a4.clean.utt_lt50_pct}{98.82}
\defmmnum{xprofile.a4.clean.utt_median_wer}{0.00}
\defmmnum{xprofile.a4.clean.wer}{4.29}
\defmmnum{xprofile.a4.clean.word_ratio_corpus}{1.00}
\defmmnum{xprofile.a4.clean.word_ratio_median}{1.00}
\defmmnum{xprofile.a4.other.cap159_n}{0}
\defmmnum{xprofile.a4.other.cap159_pct}{0.00}
\defmmnum{xprofile.a4.other.cap160_n}{0}
\defmmnum{xprofile.a4.other.cap160_pct}{0.00}
\defmmnum{xprofile.a4.other.del_pct}{1.00}
\defmmnum{xprofile.a4.other.del_words}{521}
\defmmnum{xprofile.a4.other.empty_pct}{0.00}
\defmmnum{xprofile.a4.other.ins_pct}{1.44}
\defmmnum{xprofile.a4.other.ins_words}{753}
\defmmnum{xprofile.a4.other.n_empty}{0}
\defmmnum{xprofile.a4.other.n_utt}{2{,}939}
\defmmnum{xprofile.a4.other.nonempty_n}{2{,}939}
\defmmnum{xprofile.a4.other.nonempty_utt_median_wer}{7.69}
\defmmnum{xprofile.a4.other.nonempty_wer}{10.85}
\defmmnum{xprofile.a4.other.sub_pct}{8.42}
\defmmnum{xprofile.a4.other.sub_words}{4{,}407}
\defmmnum{xprofile.a4.other.utt_eq0_n}{1{,}000}
\defmmnum{xprofile.a4.other.utt_eq0_pct}{34.03}
\defmmnum{xprofile.a4.other.utt_ge100_n}{19}
\defmmnum{xprofile.a4.other.utt_ge100_pct}{0.65}
\defmmnum{xprofile.a4.other.utt_lt10_n}{1{,}639}
\defmmnum{xprofile.a4.other.utt_lt10_pct}{55.77}
\defmmnum{xprofile.a4.other.utt_lt50_n}{2{,}781}
\defmmnum{xprofile.a4.other.utt_lt50_pct}{94.62}
\defmmnum{xprofile.a4.other.utt_median_wer}{7.69}
\defmmnum{xprofile.a4.other.wer}{10.85}
\defmmnum{xprofile.a4.other.word_ratio_corpus}{1.00}
\defmmnum{xprofile.a4.other.word_ratio_median}{1.00}
\defmmnum{xprofile.a5.clean.cap159_n}{0}
\defmmnum{xprofile.a5.clean.cap159_pct}{0.00}
\defmmnum{xprofile.a5.clean.cap160_n}{0}
\defmmnum{xprofile.a5.clean.cap160_pct}{0.00}
\defmmnum{xprofile.a5.clean.del_pct}{0.38}
\defmmnum{xprofile.a5.clean.del_words}{200}
\defmmnum{xprofile.a5.clean.empty_pct}{0.00}
\defmmnum{xprofile.a5.clean.ins_pct}{0.71}
\defmmnum{xprofile.a5.clean.ins_words}{372}
\defmmnum{xprofile.a5.clean.n_empty}{0}
\defmmnum{xprofile.a5.clean.n_utt}{2{,}620}
\defmmnum{xprofile.a5.clean.nonempty_n}{2{,}620}
\defmmnum{xprofile.a5.clean.nonempty_utt_median_wer}{0.00}
\defmmnum{xprofile.a5.clean.nonempty_wer}{5.10}
\defmmnum{xprofile.a5.clean.sub_pct}{4.01}
\defmmnum{xprofile.a5.clean.sub_words}{2{,}107}
\defmmnum{xprofile.a5.clean.utt_eq0_n}{1{,}331}
\defmmnum{xprofile.a5.clean.utt_eq0_pct}{50.80}
\defmmnum{xprofile.a5.clean.utt_ge100_n}{5}
\defmmnum{xprofile.a5.clean.utt_ge100_pct}{0.19}
\defmmnum{xprofile.a5.clean.utt_lt10_n}{2{,}035}
\defmmnum{xprofile.a5.clean.utt_lt10_pct}{77.67}
\defmmnum{xprofile.a5.clean.utt_lt50_n}{2{,}592}
\defmmnum{xprofile.a5.clean.utt_lt50_pct}{98.93}
\defmmnum{xprofile.a5.clean.utt_median_wer}{0.00}
\defmmnum{xprofile.a5.clean.wer}{5.10}
\defmmnum{xprofile.a5.clean.word_ratio_corpus}{1.00}
\defmmnum{xprofile.a5.clean.word_ratio_median}{1.00}
\defmmnum{xprofile.a5.other.cap159_n}{1}
\defmmnum{xprofile.a5.other.cap159_pct}{0.03}
\defmmnum{xprofile.a5.other.cap160_n}{1}
\defmmnum{xprofile.a5.other.cap160_pct}{0.03}
\defmmnum{xprofile.a5.other.del_pct}{1.18}
\defmmnum{xprofile.a5.other.del_words}{617}
\defmmnum{xprofile.a5.other.empty_pct}{0.00}
\defmmnum{xprofile.a5.other.ins_pct}{2.13}
\defmmnum{xprofile.a5.other.ins_words}{1{,}116}
\defmmnum{xprofile.a5.other.n_empty}{0}
\defmmnum{xprofile.a5.other.n_utt}{2{,}939}
\defmmnum{xprofile.a5.other.nonempty_n}{2{,}939}
\defmmnum{xprofile.a5.other.nonempty_utt_median_wer}{9.09}
\defmmnum{xprofile.a5.other.nonempty_wer}{12.45}
\defmmnum{xprofile.a5.other.sub_pct}{9.14}
\defmmnum{xprofile.a5.other.sub_words}{4{,}785}
\defmmnum{xprofile.a5.other.utt_eq0_n}{893}
\defmmnum{xprofile.a5.other.utt_eq0_pct}{30.38}
\defmmnum{xprofile.a5.other.utt_ge100_n}{27}
\defmmnum{xprofile.a5.other.utt_ge100_pct}{0.92}
\defmmnum{xprofile.a5.other.utt_lt10_n}{1{,}519}
\defmmnum{xprofile.a5.other.utt_lt10_pct}{51.68}
\defmmnum{xprofile.a5.other.utt_lt50_n}{2{,}743}
\defmmnum{xprofile.a5.other.utt_lt50_pct}{93.33}
\defmmnum{xprofile.a5.other.utt_median_wer}{9.09}
\defmmnum{xprofile.a5.other.wer}{12.45}
\defmmnum{xprofile.a5.other.word_ratio_corpus}{1.01}
\defmmnum{xprofile.a5.other.word_ratio_median}{1.00}
\defmmnum{xprofile.a6_k050.clean.cap159_n}{0}
\defmmnum{xprofile.a6_k050.clean.cap159_pct}{0.00}
\defmmnum{xprofile.a6_k050.clean.cap160_n}{0}
\defmmnum{xprofile.a6_k050.clean.cap160_pct}{0.00}
\defmmnum{xprofile.a6_k050.clean.del_pct}{57.02}
\defmmnum{xprofile.a6_k050.clean.del_words}{29{,}980}
\defmmnum{xprofile.a6_k050.clean.empty_pct}{0.00}
\defmmnum{xprofile.a6_k050.clean.ins_pct}{12.70}
\defmmnum{xprofile.a6_k050.clean.ins_words}{6{,}678}
\defmmnum{xprofile.a6_k050.clean.n_empty}{0}
\defmmnum{xprofile.a6_k050.clean.n_utt}{2{,}620}
\defmmnum{xprofile.a6_k050.clean.nonempty_n}{2{,}620}
\defmmnum{xprofile.a6_k050.clean.nonempty_utt_median_wer}{100.00}
\defmmnum{xprofile.a6_k050.clean.nonempty_wer}{110.06}
\defmmnum{xprofile.a6_k050.clean.sub_pct}{40.34}
\defmmnum{xprofile.a6_k050.clean.sub_words}{21{,}209}
\defmmnum{xprofile.a6_k050.clean.utt_eq0_n}{0}
\defmmnum{xprofile.a6_k050.clean.utt_eq0_pct}{0.00}
\defmmnum{xprofile.a6_k050.clean.utt_ge100_n}{1{,}814}
\defmmnum{xprofile.a6_k050.clean.utt_ge100_pct}{69.24}
\defmmnum{xprofile.a6_k050.clean.utt_lt10_n}{0}
\defmmnum{xprofile.a6_k050.clean.utt_lt10_pct}{0.00}
\defmmnum{xprofile.a6_k050.clean.utt_lt50_n}{0}
\defmmnum{xprofile.a6_k050.clean.utt_lt50_pct}{0.00}
\defmmnum{xprofile.a6_k050.clean.utt_median_wer}{100.00}
\defmmnum{xprofile.a6_k050.clean.wer}{110.06}
\defmmnum{xprofile.a6_k050.clean.word_ratio_corpus}{0.56}
\defmmnum{xprofile.a6_k050.clean.word_ratio_median}{0.53}
\defmmnum{xprofile.a6_k050.other.cap159_n}{0}
\defmmnum{xprofile.a6_k050.other.cap159_pct}{0.00}
\defmmnum{xprofile.a6_k050.other.cap160_n}{0}
\defmmnum{xprofile.a6_k050.other.cap160_pct}{0.00}
\defmmnum{xprofile.a6_k050.other.del_pct}{54.01}
\defmmnum{xprofile.a6_k050.other.del_words}{28{,}268}
\defmmnum{xprofile.a6_k050.other.empty_pct}{0.00}
\defmmnum{xprofile.a6_k050.other.ins_pct}{14.95}
\defmmnum{xprofile.a6_k050.other.ins_words}{7{,}823}
\defmmnum{xprofile.a6_k050.other.n_empty}{0}
\defmmnum{xprofile.a6_k050.other.n_utt}{2{,}939}
\defmmnum{xprofile.a6_k050.other.nonempty_n}{2{,}939}
\defmmnum{xprofile.a6_k050.other.nonempty_utt_median_wer}{100.00}
\defmmnum{xprofile.a6_k050.other.nonempty_wer}{112.18}
\defmmnum{xprofile.a6_k050.other.sub_pct}{43.23}
\defmmnum{xprofile.a6_k050.other.sub_words}{22{,}629}
\defmmnum{xprofile.a6_k050.other.utt_eq0_n}{0}
\defmmnum{xprofile.a6_k050.other.utt_eq0_pct}{0.00}
\defmmnum{xprofile.a6_k050.other.utt_ge100_n}{2{,}117}
\defmmnum{xprofile.a6_k050.other.utt_ge100_pct}{72.03}
\defmmnum{xprofile.a6_k050.other.utt_lt10_n}{0}
\defmmnum{xprofile.a6_k050.other.utt_lt10_pct}{0.00}
\defmmnum{xprofile.a6_k050.other.utt_lt50_n}{0}
\defmmnum{xprofile.a6_k050.other.utt_lt50_pct}{0.00}
\defmmnum{xprofile.a6_k050.other.utt_median_wer}{100.00}
\defmmnum{xprofile.a6_k050.other.wer}{112.18}
\defmmnum{xprofile.a6_k050.other.word_ratio_corpus}{0.61}
\defmmnum{xprofile.a6_k050.other.word_ratio_median}{0.62}
\defmmnum{xprofile.a6_k100.clean.cap159_n}{1}
\defmmnum{xprofile.a6_k100.clean.cap159_pct}{0.04}
\defmmnum{xprofile.a6_k100.clean.cap160_n}{1}
\defmmnum{xprofile.a6_k100.clean.cap160_pct}{0.04}
\defmmnum{xprofile.a6_k100.clean.del_pct}{61.81}
\defmmnum{xprofile.a6_k100.clean.del_words}{32{,}496}
\defmmnum{xprofile.a6_k100.clean.empty_pct}{0.00}
\defmmnum{xprofile.a6_k100.clean.ins_pct}{2.14}
\defmmnum{xprofile.a6_k100.clean.ins_words}{1{,}125}
\defmmnum{xprofile.a6_k100.clean.n_empty}{0}
\defmmnum{xprofile.a6_k100.clean.n_utt}{2{,}620}
\defmmnum{xprofile.a6_k100.clean.nonempty_n}{2{,}620}
\defmmnum{xprofile.a6_k100.clean.nonempty_utt_median_wer}{100.00}
\defmmnum{xprofile.a6_k100.clean.nonempty_wer}{99.82}
\defmmnum{xprofile.a6_k100.clean.sub_pct}{35.87}
\defmmnum{xprofile.a6_k100.clean.sub_words}{18{,}860}
\defmmnum{xprofile.a6_k100.clean.utt_eq0_n}{0}
\defmmnum{xprofile.a6_k100.clean.utt_eq0_pct}{0.00}
\defmmnum{xprofile.a6_k100.clean.utt_ge100_n}{1{,}747}
\defmmnum{xprofile.a6_k100.clean.utt_ge100_pct}{66.68}
\defmmnum{xprofile.a6_k100.clean.utt_lt10_n}{0}
\defmmnum{xprofile.a6_k100.clean.utt_lt10_pct}{0.00}
\defmmnum{xprofile.a6_k100.clean.utt_lt50_n}{0}
\defmmnum{xprofile.a6_k100.clean.utt_lt50_pct}{0.00}
\defmmnum{xprofile.a6_k100.clean.utt_median_wer}{100.00}
\defmmnum{xprofile.a6_k100.clean.wer}{99.82}
\defmmnum{xprofile.a6_k100.clean.word_ratio_corpus}{0.40}
\defmmnum{xprofile.a6_k100.clean.word_ratio_median}{0.50}
\defmmnum{xprofile.a6_k100.other.cap159_n}{0}
\defmmnum{xprofile.a6_k100.other.cap159_pct}{0.00}
\defmmnum{xprofile.a6_k100.other.cap160_n}{0}
\defmmnum{xprofile.a6_k100.other.cap160_pct}{0.00}
\defmmnum{xprofile.a6_k100.other.del_pct}{57.81}
\defmmnum{xprofile.a6_k100.other.del_words}{30{,}262}
\defmmnum{xprofile.a6_k100.other.empty_pct}{0.00}
\defmmnum{xprofile.a6_k100.other.ins_pct}{3.27}
\defmmnum{xprofile.a6_k100.other.ins_words}{1{,}710}
\defmmnum{xprofile.a6_k100.other.n_empty}{0}
\defmmnum{xprofile.a6_k100.other.n_utt}{2{,}939}
\defmmnum{xprofile.a6_k100.other.nonempty_n}{2{,}939}
\defmmnum{xprofile.a6_k100.other.nonempty_utt_median_wer}{100.00}
\defmmnum{xprofile.a6_k100.other.nonempty_wer}{100.86}
\defmmnum{xprofile.a6_k100.other.sub_pct}{39.78}
\defmmnum{xprofile.a6_k100.other.sub_words}{20{,}821}
\defmmnum{xprofile.a6_k100.other.utt_eq0_n}{0}
\defmmnum{xprofile.a6_k100.other.utt_eq0_pct}{0.00}
\defmmnum{xprofile.a6_k100.other.utt_ge100_n}{2{,}062}
\defmmnum{xprofile.a6_k100.other.utt_ge100_pct}{70.16}
\defmmnum{xprofile.a6_k100.other.utt_lt10_n}{0}
\defmmnum{xprofile.a6_k100.other.utt_lt10_pct}{0.00}
\defmmnum{xprofile.a6_k100.other.utt_lt50_n}{0}
\defmmnum{xprofile.a6_k100.other.utt_lt50_pct}{0.00}
\defmmnum{xprofile.a6_k100.other.utt_median_wer}{100.00}
\defmmnum{xprofile.a6_k100.other.wer}{100.86}
\defmmnum{xprofile.a6_k100.other.word_ratio_corpus}{0.45}
\defmmnum{xprofile.a6_k100.other.word_ratio_median}{0.57}
\defmmnum{xprofile.a6_k773.clean.cap159_n}{61}
\defmmnum{xprofile.a6_k773.clean.cap159_pct}{2.33}
\defmmnum{xprofile.a6_k773.clean.cap160_n}{49}
\defmmnum{xprofile.a6_k773.clean.cap160_pct}{1.87}
\defmmnum{xprofile.a6_k773.clean.del_pct}{28.64}
\defmmnum{xprofile.a6_k773.clean.del_words}{15{,}056}
\defmmnum{xprofile.a6_k773.clean.empty_pct}{0.69}
\defmmnum{xprofile.a6_k773.clean.ins_pct}{49.59}
\defmmnum{xprofile.a6_k773.clean.ins_words}{26{,}071}
\defmmnum{xprofile.a6_k773.clean.n_empty}{18}
\defmmnum{xprofile.a6_k773.clean.n_utt}{2{,}620}
\defmmnum{xprofile.a6_k773.clean.nonempty_n}{2{,}602}
\defmmnum{xprofile.a6_k773.clean.nonempty_utt_median_wer}{100.00}
\defmmnum{xprofile.a6_k773.clean.nonempty_wer}{145.03}
\defmmnum{xprofile.a6_k773.clean.sub_pct}{65.76}
\defmmnum{xprofile.a6_k773.clean.sub_words}{34{,}572}
\defmmnum{xprofile.a6_k773.clean.utt_eq0_n}{0}
\defmmnum{xprofile.a6_k773.clean.utt_eq0_pct}{0.00}
\defmmnum{xprofile.a6_k773.clean.utt_ge100_n}{2{,}032}
\defmmnum{xprofile.a6_k773.clean.utt_ge100_pct}{77.56}
\defmmnum{xprofile.a6_k773.clean.utt_lt10_n}{0}
\defmmnum{xprofile.a6_k773.clean.utt_lt10_pct}{0.00}
\defmmnum{xprofile.a6_k773.clean.utt_lt50_n}{0}
\defmmnum{xprofile.a6_k773.clean.utt_lt50_pct}{0.00}
\defmmnum{xprofile.a6_k773.clean.utt_median_wer}{100.00}
\defmmnum{xprofile.a6_k773.clean.wer}{143.98}
\defmmnum{xprofile.a6_k773.clean.word_ratio_corpus}{1.21}
\defmmnum{xprofile.a6_k773.clean.word_ratio_median}{0.89}
\defmmnum{xprofile.a6_k773.other.cap159_n}{55}
\defmmnum{xprofile.a6_k773.other.cap159_pct}{1.87}
\defmmnum{xprofile.a6_k773.other.cap160_n}{39}
\defmmnum{xprofile.a6_k773.other.cap160_pct}{1.33}
\defmmnum{xprofile.a6_k773.other.del_pct}{25.02}
\defmmnum{xprofile.a6_k773.other.del_words}{13{,}098}
\defmmnum{xprofile.a6_k773.other.empty_pct}{0.14}
\defmmnum{xprofile.a6_k773.other.ins_pct}{55.74}
\defmmnum{xprofile.a6_k773.other.ins_words}{29{,}174}
\defmmnum{xprofile.a6_k773.other.n_empty}{4}
\defmmnum{xprofile.a6_k773.other.n_utt}{2{,}939}
\defmmnum{xprofile.a6_k773.other.nonempty_n}{2{,}935}
\defmmnum{xprofile.a6_k773.other.nonempty_utt_median_wer}{100.00}
\defmmnum{xprofile.a6_k773.other.nonempty_wer}{150.42}
\defmmnum{xprofile.a6_k773.other.sub_pct}{69.37}
\defmmnum{xprofile.a6_k773.other.sub_words}{36{,}308}
\defmmnum{xprofile.a6_k773.other.utt_eq0_n}{0}
\defmmnum{xprofile.a6_k773.other.utt_eq0_pct}{0.00}
\defmmnum{xprofile.a6_k773.other.utt_ge100_n}{2{,}387}
\defmmnum{xprofile.a6_k773.other.utt_ge100_pct}{81.22}
\defmmnum{xprofile.a6_k773.other.utt_lt10_n}{0}
\defmmnum{xprofile.a6_k773.other.utt_lt10_pct}{0.00}
\defmmnum{xprofile.a6_k773.other.utt_lt50_n}{0}
\defmmnum{xprofile.a6_k773.other.utt_lt50_pct}{0.00}
\defmmnum{xprofile.a6_k773.other.utt_median_wer}{100.00}
\defmmnum{xprofile.a6_k773.other.wer}{150.13}
\defmmnum{xprofile.a6_k773.other.word_ratio_corpus}{1.31}
\defmmnum{xprofile.a6_k773.other.word_ratio_median}{0.90}
\defmmnum{xprofile.a7.caponesixtypct}{0.00}
\defmmnum{xprofile.a7.clean.cap159_n}{0}
\defmmnum{xprofile.a7.clean.cap159_pct}{0.00}
\defmmnum{xprofile.a7.clean.cap160_n}{0}
\defmmnum{xprofile.a7.clean.cap160_pct}{0.00}
\defmmnum{xprofile.a7.clean.del_pct}{0.30}
\defmmnum{xprofile.a7.clean.del_words}{159}
\defmmnum{xprofile.a7.clean.empty_pct}{0.00}
\defmmnum{xprofile.a7.clean.ins_pct}{0.39}
\defmmnum{xprofile.a7.clean.ins_words}{206}
\defmmnum{xprofile.a7.clean.n_empty}{0}
\defmmnum{xprofile.a7.clean.n_utt}{2{,}620}
\defmmnum{xprofile.a7.clean.nonempty_n}{2{,}620}
\defmmnum{xprofile.a7.clean.nonempty_utt_median_wer}{0.00}
\defmmnum{xprofile.a7.clean.nonempty_wer}{3.22}
\defmmnum{xprofile.a7.clean.sub_pct}{2.53}
\defmmnum{xprofile.a7.clean.sub_words}{1{,}328}
\defmmnum{xprofile.a7.clean.utt_eq0_n}{1{,}682}
\defmmnum{xprofile.a7.clean.utt_eq0_pct}{64.20}
\defmmnum{xprofile.a7.clean.utt_ge100_n}{1}
\defmmnum{xprofile.a7.clean.utt_ge100_pct}{0.04}
\defmmnum{xprofile.a7.clean.utt_lt10_n}{2{,}281}
\defmmnum{xprofile.a7.clean.utt_lt10_pct}{87.06}
\defmmnum{xprofile.a7.clean.utt_lt50_n}{2{,}602}
\defmmnum{xprofile.a7.clean.utt_lt50_pct}{99.31}
\defmmnum{xprofile.a7.clean.utt_median_wer}{0.00}
\defmmnum{xprofile.a7.clean.wer}{3.22}
\defmmnum{xprofile.a7.clean.word_ratio_corpus}{1.00}
\defmmnum{xprofile.a7.clean.word_ratio_median}{1.00}
\defmmnum{xprofile.a7.delpct}{0.30}
\defmmnum{xprofile.a7.inspct}{0.39}
\defmmnum{xprofile.a7.nonemptymedwer}{0.00}
\defmmnum{xprofile.a7.nonemptywer}{3.22}
\defmmnum{xprofile.a7.other.cap159_n}{1}
\defmmnum{xprofile.a7.other.cap159_pct}{0.03}
\defmmnum{xprofile.a7.other.cap160_n}{1}
\defmmnum{xprofile.a7.other.cap160_pct}{0.03}
\defmmnum{xprofile.a7.other.del_pct}{0.72}
\defmmnum{xprofile.a7.other.del_words}{378}
\defmmnum{xprofile.a7.other.empty_pct}{0.00}
\defmmnum{xprofile.a7.other.ins_pct}{0.93}
\defmmnum{xprofile.a7.other.ins_words}{487}
\defmmnum{xprofile.a7.other.n_empty}{0}
\defmmnum{xprofile.a7.other.n_utt}{2{,}939}
\defmmnum{xprofile.a7.other.nonempty_n}{2{,}939}
\defmmnum{xprofile.a7.other.nonempty_utt_median_wer}{3.85}
\defmmnum{xprofile.a7.other.nonempty_wer}{7.53}
\defmmnum{xprofile.a7.other.sub_pct}{5.88}
\defmmnum{xprofile.a7.other.sub_words}{3{,}077}
\defmmnum{xprofile.a7.other.utt_eq0_n}{1{,}307}
\defmmnum{xprofile.a7.other.utt_eq0_pct}{44.47}
\defmmnum{xprofile.a7.other.utt_ge100_n}{16}
\defmmnum{xprofile.a7.other.utt_ge100_pct}{0.54}
\defmmnum{xprofile.a7.other.utt_lt10_n}{1{,}995}
\defmmnum{xprofile.a7.other.utt_lt10_pct}{67.88}
\defmmnum{xprofile.a7.other.utt_lt50_n}{2{,}844}
\defmmnum{xprofile.a7.other.utt_lt50_pct}{96.77}
\defmmnum{xprofile.a7.other.utt_median_wer}{3.85}
\defmmnum{xprofile.a7.other.wer}{7.53}
\defmmnum{xprofile.a7.other.word_ratio_corpus}{1.00}
\defmmnum{xprofile.a7.other.word_ratio_median}{1.00}
\defmmnum{xprofile.a7.subpct}{2.53}
\defmmnum{xprofile.a7.uttltfiftypct}{99.31}
\defmmnum{xprofile.a7.uttmedwer}{0.00}
\defmmnum{xprofile.a7.wer}{3.22}
\defmmnum{xprofile.best_closed.moonshine-base__Qwen2.5-1.5B-Instruct__to__gemma-2-2b-it__l2norm.clean.cap159_n}{5}
\defmmnum{xprofile.best_closed.moonshine-base__Qwen2.5-1.5B-Instruct__to__gemma-2-2b-it__l2norm.clean.cap159_pct}{0.19}
\defmmnum{xprofile.best_closed.moonshine-base__Qwen2.5-1.5B-Instruct__to__gemma-2-2b-it__l2norm.clean.cap160_n}{4}
\defmmnum{xprofile.best_closed.moonshine-base__Qwen2.5-1.5B-Instruct__to__gemma-2-2b-it__l2norm.clean.cap160_pct}{0.15}
\defmmnum{xprofile.best_closed.moonshine-base__Qwen2.5-1.5B-Instruct__to__gemma-2-2b-it__l2norm.clean.del_pct}{10.13}
\defmmnum{xprofile.best_closed.moonshine-base__Qwen2.5-1.5B-Instruct__to__gemma-2-2b-it__l2norm.clean.del_words}{5{,}324}
\defmmnum{xprofile.best_closed.moonshine-base__Qwen2.5-1.5B-Instruct__to__gemma-2-2b-it__l2norm.clean.empty_pct}{0.00}
\defmmnum{xprofile.best_closed.moonshine-base__Qwen2.5-1.5B-Instruct__to__gemma-2-2b-it__l2norm.clean.ins_pct}{110.74}
\defmmnum{xprofile.best_closed.moonshine-base__Qwen2.5-1.5B-Instruct__to__gemma-2-2b-it__l2norm.clean.ins_words}{58{,}222}
\defmmnum{xprofile.best_closed.moonshine-base__Qwen2.5-1.5B-Instruct__to__gemma-2-2b-it__l2norm.clean.n_empty}{0}
\defmmnum{xprofile.best_closed.moonshine-base__Qwen2.5-1.5B-Instruct__to__gemma-2-2b-it__l2norm.clean.n_utt}{2{,}620}
\defmmnum{xprofile.best_closed.moonshine-base__Qwen2.5-1.5B-Instruct__to__gemma-2-2b-it__l2norm.clean.nonempty_n}{2{,}620}
\defmmnum{xprofile.best_closed.moonshine-base__Qwen2.5-1.5B-Instruct__to__gemma-2-2b-it__l2norm.clean.nonempty_utt_median_wer}{206.25}
\defmmnum{xprofile.best_closed.moonshine-base__Qwen2.5-1.5B-Instruct__to__gemma-2-2b-it__l2norm.clean.nonempty_wer}{201.54}
\defmmnum{xprofile.best_closed.moonshine-base__Qwen2.5-1.5B-Instruct__to__gemma-2-2b-it__l2norm.clean.sub_pct}{80.68}
\defmmnum{xprofile.best_closed.moonshine-base__Qwen2.5-1.5B-Instruct__to__gemma-2-2b-it__l2norm.clean.sub_words}{42{,}417}
\defmmnum{xprofile.best_closed.moonshine-base__Qwen2.5-1.5B-Instruct__to__gemma-2-2b-it__l2norm.clean.utt_eq0_n}{0}
\defmmnum{xprofile.best_closed.moonshine-base__Qwen2.5-1.5B-Instruct__to__gemma-2-2b-it__l2norm.clean.utt_eq0_pct}{0.00}
\defmmnum{xprofile.best_closed.moonshine-base__Qwen2.5-1.5B-Instruct__to__gemma-2-2b-it__l2norm.clean.utt_ge100_n}{2{,}248}
\defmmnum{xprofile.best_closed.moonshine-base__Qwen2.5-1.5B-Instruct__to__gemma-2-2b-it__l2norm.clean.utt_ge100_pct}{85.80}
\defmmnum{xprofile.best_closed.moonshine-base__Qwen2.5-1.5B-Instruct__to__gemma-2-2b-it__l2norm.clean.utt_lt10_n}{0}
\defmmnum{xprofile.best_closed.moonshine-base__Qwen2.5-1.5B-Instruct__to__gemma-2-2b-it__l2norm.clean.utt_lt10_pct}{0.00}
\defmmnum{xprofile.best_closed.moonshine-base__Qwen2.5-1.5B-Instruct__to__gemma-2-2b-it__l2norm.clean.utt_lt50_n}{0}
\defmmnum{xprofile.best_closed.moonshine-base__Qwen2.5-1.5B-Instruct__to__gemma-2-2b-it__l2norm.clean.utt_lt50_pct}{0.00}
\defmmnum{xprofile.best_closed.moonshine-base__Qwen2.5-1.5B-Instruct__to__gemma-2-2b-it__l2norm.clean.utt_median_wer}{206.25}
\defmmnum{xprofile.best_closed.moonshine-base__Qwen2.5-1.5B-Instruct__to__gemma-2-2b-it__l2norm.clean.wer}{201.54}
\defmmnum{xprofile.best_closed.moonshine-base__Qwen2.5-1.5B-Instruct__to__gemma-2-2b-it__l2norm.clean.word_ratio_corpus}{2.01}
\defmmnum{xprofile.best_closed.moonshine-base__Qwen2.5-1.5B-Instruct__to__gemma-2-2b-it__l2norm.clean.word_ratio_median}{2.17}
\defmmnum{xprofile.best_closed.moonshine-base__Qwen2.5-1.5B-Instruct__to__gemma-2-2b-it__l2norm.other.cap159_n}{6}
\defmmnum{xprofile.best_closed.moonshine-base__Qwen2.5-1.5B-Instruct__to__gemma-2-2b-it__l2norm.other.cap159_pct}{0.20}
\defmmnum{xprofile.best_closed.moonshine-base__Qwen2.5-1.5B-Instruct__to__gemma-2-2b-it__l2norm.other.cap160_n}{5}
\defmmnum{xprofile.best_closed.moonshine-base__Qwen2.5-1.5B-Instruct__to__gemma-2-2b-it__l2norm.other.cap160_pct}{0.17}
\defmmnum{xprofile.best_closed.moonshine-base__Qwen2.5-1.5B-Instruct__to__gemma-2-2b-it__l2norm.other.del_pct}{8.41}
\defmmnum{xprofile.best_closed.moonshine-base__Qwen2.5-1.5B-Instruct__to__gemma-2-2b-it__l2norm.other.del_words}{4{,}404}
\defmmnum{xprofile.best_closed.moonshine-base__Qwen2.5-1.5B-Instruct__to__gemma-2-2b-it__l2norm.other.empty_pct}{0.00}
\defmmnum{xprofile.best_closed.moonshine-base__Qwen2.5-1.5B-Instruct__to__gemma-2-2b-it__l2norm.other.ins_pct}{137.33}
\defmmnum{xprofile.best_closed.moonshine-base__Qwen2.5-1.5B-Instruct__to__gemma-2-2b-it__l2norm.other.ins_words}{71{,}882}
\defmmnum{xprofile.best_closed.moonshine-base__Qwen2.5-1.5B-Instruct__to__gemma-2-2b-it__l2norm.other.n_empty}{0}
\defmmnum{xprofile.best_closed.moonshine-base__Qwen2.5-1.5B-Instruct__to__gemma-2-2b-it__l2norm.other.n_utt}{2{,}939}
\defmmnum{xprofile.best_closed.moonshine-base__Qwen2.5-1.5B-Instruct__to__gemma-2-2b-it__l2norm.other.nonempty_n}{2{,}939}
\defmmnum{xprofile.best_closed.moonshine-base__Qwen2.5-1.5B-Instruct__to__gemma-2-2b-it__l2norm.other.nonempty_utt_median_wer}{244.44}
\defmmnum{xprofile.best_closed.moonshine-base__Qwen2.5-1.5B-Instruct__to__gemma-2-2b-it__l2norm.other.nonempty_wer}{227.24}
\defmmnum{xprofile.best_closed.moonshine-base__Qwen2.5-1.5B-Instruct__to__gemma-2-2b-it__l2norm.other.sub_pct}{81.50}
\defmmnum{xprofile.best_closed.moonshine-base__Qwen2.5-1.5B-Instruct__to__gemma-2-2b-it__l2norm.other.sub_words}{42{,}660}
\defmmnum{xprofile.best_closed.moonshine-base__Qwen2.5-1.5B-Instruct__to__gemma-2-2b-it__l2norm.other.utt_eq0_n}{0}
\defmmnum{xprofile.best_closed.moonshine-base__Qwen2.5-1.5B-Instruct__to__gemma-2-2b-it__l2norm.other.utt_eq0_pct}{0.00}
\defmmnum{xprofile.best_closed.moonshine-base__Qwen2.5-1.5B-Instruct__to__gemma-2-2b-it__l2norm.other.utt_ge100_n}{2{,}634}
\defmmnum{xprofile.best_closed.moonshine-base__Qwen2.5-1.5B-Instruct__to__gemma-2-2b-it__l2norm.other.utt_ge100_pct}{89.62}
\defmmnum{xprofile.best_closed.moonshine-base__Qwen2.5-1.5B-Instruct__to__gemma-2-2b-it__l2norm.other.utt_lt10_n}{0}
\defmmnum{xprofile.best_closed.moonshine-base__Qwen2.5-1.5B-Instruct__to__gemma-2-2b-it__l2norm.other.utt_lt10_pct}{0.00}
\defmmnum{xprofile.best_closed.moonshine-base__Qwen2.5-1.5B-Instruct__to__gemma-2-2b-it__l2norm.other.utt_lt50_n}{0}
\defmmnum{xprofile.best_closed.moonshine-base__Qwen2.5-1.5B-Instruct__to__gemma-2-2b-it__l2norm.other.utt_lt50_pct}{0.00}
\defmmnum{xprofile.best_closed.moonshine-base__Qwen2.5-1.5B-Instruct__to__gemma-2-2b-it__l2norm.other.utt_median_wer}{244.44}
\defmmnum{xprofile.best_closed.moonshine-base__Qwen2.5-1.5B-Instruct__to__gemma-2-2b-it__l2norm.other.wer}{227.24}
\defmmnum{xprofile.best_closed.moonshine-base__Qwen2.5-1.5B-Instruct__to__gemma-2-2b-it__l2norm.other.word_ratio_corpus}{2.29}
\defmmnum{xprofile.best_closed.moonshine-base__Qwen2.5-1.5B-Instruct__to__gemma-2-2b-it__l2norm.other.word_ratio_median}{2.58}
\defmmnum{xprofile.best_closed.moonshine-base__Qwen2.5-1.5B-Instruct__to__gemma-2-2b-it__rmsnorm.clean.cap159_n}{423}
\defmmnum{xprofile.best_closed.moonshine-base__Qwen2.5-1.5B-Instruct__to__gemma-2-2b-it__rmsnorm.clean.cap159_pct}{16.15}
\defmmnum{xprofile.best_closed.moonshine-base__Qwen2.5-1.5B-Instruct__to__gemma-2-2b-it__rmsnorm.clean.cap160_n}{286}
\defmmnum{xprofile.best_closed.moonshine-base__Qwen2.5-1.5B-Instruct__to__gemma-2-2b-it__rmsnorm.clean.cap160_pct}{10.92}
\defmmnum{xprofile.best_closed.moonshine-base__Qwen2.5-1.5B-Instruct__to__gemma-2-2b-it__rmsnorm.clean.del_pct}{93.02}
\defmmnum{xprofile.best_closed.moonshine-base__Qwen2.5-1.5B-Instruct__to__gemma-2-2b-it__rmsnorm.clean.del_words}{48{,}906}
\defmmnum{xprofile.best_closed.moonshine-base__Qwen2.5-1.5B-Instruct__to__gemma-2-2b-it__rmsnorm.clean.empty_pct}{83.70}
\defmmnum{xprofile.best_closed.moonshine-base__Qwen2.5-1.5B-Instruct__to__gemma-2-2b-it__rmsnorm.clean.ins_pct}{61.72}
\defmmnum{xprofile.best_closed.moonshine-base__Qwen2.5-1.5B-Instruct__to__gemma-2-2b-it__rmsnorm.clean.ins_words}{32{,}449}
\defmmnum{xprofile.best_closed.moonshine-base__Qwen2.5-1.5B-Instruct__to__gemma-2-2b-it__rmsnorm.clean.n_empty}{2{,}193}
\defmmnum{xprofile.best_closed.moonshine-base__Qwen2.5-1.5B-Instruct__to__gemma-2-2b-it__rmsnorm.clean.n_utt}{2{,}620}
\defmmnum{xprofile.best_closed.moonshine-base__Qwen2.5-1.5B-Instruct__to__gemma-2-2b-it__rmsnorm.clean.nonempty_n}{427}
\defmmnum{xprofile.best_closed.moonshine-base__Qwen2.5-1.5B-Instruct__to__gemma-2-2b-it__rmsnorm.clean.nonempty_utt_median_wer}{1030.00}
\defmmnum{xprofile.best_closed.moonshine-base__Qwen2.5-1.5B-Instruct__to__gemma-2-2b-it__rmsnorm.clean.nonempty_wer}{946.03}
\defmmnum{xprofile.best_closed.moonshine-base__Qwen2.5-1.5B-Instruct__to__gemma-2-2b-it__rmsnorm.clean.sub_pct}{6.02}
\defmmnum{xprofile.best_closed.moonshine-base__Qwen2.5-1.5B-Instruct__to__gemma-2-2b-it__rmsnorm.clean.sub_words}{3{,}167}
\defmmnum{xprofile.best_closed.moonshine-base__Qwen2.5-1.5B-Instruct__to__gemma-2-2b-it__rmsnorm.clean.utt_eq0_n}{0}
\defmmnum{xprofile.best_closed.moonshine-base__Qwen2.5-1.5B-Instruct__to__gemma-2-2b-it__rmsnorm.clean.utt_eq0_pct}{0.00}
\defmmnum{xprofile.best_closed.moonshine-base__Qwen2.5-1.5B-Instruct__to__gemma-2-2b-it__rmsnorm.clean.utt_ge100_n}{2{,}614}
\defmmnum{xprofile.best_closed.moonshine-base__Qwen2.5-1.5B-Instruct__to__gemma-2-2b-it__rmsnorm.clean.utt_ge100_pct}{99.77}
\defmmnum{xprofile.best_closed.moonshine-base__Qwen2.5-1.5B-Instruct__to__gemma-2-2b-it__rmsnorm.clean.utt_lt10_n}{0}
\defmmnum{xprofile.best_closed.moonshine-base__Qwen2.5-1.5B-Instruct__to__gemma-2-2b-it__rmsnorm.clean.utt_lt10_pct}{0.00}
\defmmnum{xprofile.best_closed.moonshine-base__Qwen2.5-1.5B-Instruct__to__gemma-2-2b-it__rmsnorm.clean.utt_lt50_n}{0}
\defmmnum{xprofile.best_closed.moonshine-base__Qwen2.5-1.5B-Instruct__to__gemma-2-2b-it__rmsnorm.clean.utt_lt50_pct}{0.00}
\defmmnum{xprofile.best_closed.moonshine-base__Qwen2.5-1.5B-Instruct__to__gemma-2-2b-it__rmsnorm.clean.utt_median_wer}{100.00}
\defmmnum{xprofile.best_closed.moonshine-base__Qwen2.5-1.5B-Instruct__to__gemma-2-2b-it__rmsnorm.clean.wer}{160.76}
\defmmnum{xprofile.best_closed.moonshine-base__Qwen2.5-1.5B-Instruct__to__gemma-2-2b-it__rmsnorm.clean.word_ratio_corpus}{0.69}
\defmmnum{xprofile.best_closed.moonshine-base__Qwen2.5-1.5B-Instruct__to__gemma-2-2b-it__rmsnorm.clean.word_ratio_median}{0.00}
\defmmnum{xprofile.best_closed.moonshine-base__Qwen2.5-1.5B-Instruct__to__gemma-2-2b-it__rmsnorm.other.cap159_n}{570}
\defmmnum{xprofile.best_closed.moonshine-base__Qwen2.5-1.5B-Instruct__to__gemma-2-2b-it__rmsnorm.other.cap159_pct}{19.39}
\defmmnum{xprofile.best_closed.moonshine-base__Qwen2.5-1.5B-Instruct__to__gemma-2-2b-it__rmsnorm.other.cap160_n}{445}
\defmmnum{xprofile.best_closed.moonshine-base__Qwen2.5-1.5B-Instruct__to__gemma-2-2b-it__rmsnorm.other.cap160_pct}{15.14}
\defmmnum{xprofile.best_closed.moonshine-base__Qwen2.5-1.5B-Instruct__to__gemma-2-2b-it__rmsnorm.other.del_pct}{89.33}
\defmmnum{xprofile.best_closed.moonshine-base__Qwen2.5-1.5B-Instruct__to__gemma-2-2b-it__rmsnorm.other.del_words}{46{,}756}
\defmmnum{xprofile.best_closed.moonshine-base__Qwen2.5-1.5B-Instruct__to__gemma-2-2b-it__rmsnorm.other.empty_pct}{77.75}
\defmmnum{xprofile.best_closed.moonshine-base__Qwen2.5-1.5B-Instruct__to__gemma-2-2b-it__rmsnorm.other.ins_pct}{95.44}
\defmmnum{xprofile.best_closed.moonshine-base__Qwen2.5-1.5B-Instruct__to__gemma-2-2b-it__rmsnorm.other.ins_words}{49{,}954}
\defmmnum{xprofile.best_closed.moonshine-base__Qwen2.5-1.5B-Instruct__to__gemma-2-2b-it__rmsnorm.other.n_empty}{2{,}285}
\defmmnum{xprofile.best_closed.moonshine-base__Qwen2.5-1.5B-Instruct__to__gemma-2-2b-it__rmsnorm.other.n_utt}{2{,}939}
\defmmnum{xprofile.best_closed.moonshine-base__Qwen2.5-1.5B-Instruct__to__gemma-2-2b-it__rmsnorm.other.nonempty_n}{654}
\defmmnum{xprofile.best_closed.moonshine-base__Qwen2.5-1.5B-Instruct__to__gemma-2-2b-it__rmsnorm.other.nonempty_utt_median_wer}{1078.89}
\defmmnum{xprofile.best_closed.moonshine-base__Qwen2.5-1.5B-Instruct__to__gemma-2-2b-it__rmsnorm.other.nonempty_wer}{962.04}
\defmmnum{xprofile.best_closed.moonshine-base__Qwen2.5-1.5B-Instruct__to__gemma-2-2b-it__rmsnorm.other.sub_pct}{9.08}
\defmmnum{xprofile.best_closed.moonshine-base__Qwen2.5-1.5B-Instruct__to__gemma-2-2b-it__rmsnorm.other.sub_words}{4{,}752}
\defmmnum{xprofile.best_closed.moonshine-base__Qwen2.5-1.5B-Instruct__to__gemma-2-2b-it__rmsnorm.other.utt_eq0_n}{0}
\defmmnum{xprofile.best_closed.moonshine-base__Qwen2.5-1.5B-Instruct__to__gemma-2-2b-it__rmsnorm.other.utt_eq0_pct}{0.00}
\defmmnum{xprofile.best_closed.moonshine-base__Qwen2.5-1.5B-Instruct__to__gemma-2-2b-it__rmsnorm.other.utt_ge100_n}{2{,}933}
\defmmnum{xprofile.best_closed.moonshine-base__Qwen2.5-1.5B-Instruct__to__gemma-2-2b-it__rmsnorm.other.utt_ge100_pct}{99.80}
\defmmnum{xprofile.best_closed.moonshine-base__Qwen2.5-1.5B-Instruct__to__gemma-2-2b-it__rmsnorm.other.utt_lt10_n}{0}
\defmmnum{xprofile.best_closed.moonshine-base__Qwen2.5-1.5B-Instruct__to__gemma-2-2b-it__rmsnorm.other.utt_lt10_pct}{0.00}
\defmmnum{xprofile.best_closed.moonshine-base__Qwen2.5-1.5B-Instruct__to__gemma-2-2b-it__rmsnorm.other.utt_lt50_n}{0}
\defmmnum{xprofile.best_closed.moonshine-base__Qwen2.5-1.5B-Instruct__to__gemma-2-2b-it__rmsnorm.other.utt_lt50_pct}{0.00}
\defmmnum{xprofile.best_closed.moonshine-base__Qwen2.5-1.5B-Instruct__to__gemma-2-2b-it__rmsnorm.other.utt_median_wer}{100.00}
\defmmnum{xprofile.best_closed.moonshine-base__Qwen2.5-1.5B-Instruct__to__gemma-2-2b-it__rmsnorm.other.wer}{193.84}
\defmmnum{xprofile.best_closed.moonshine-base__Qwen2.5-1.5B-Instruct__to__gemma-2-2b-it__rmsnorm.other.word_ratio_corpus}{1.06}
\defmmnum{xprofile.best_closed.moonshine-base__Qwen2.5-1.5B-Instruct__to__gemma-2-2b-it__rmsnorm.other.word_ratio_median}{0.00}
\defmmnum{xprofile.best_closed.moonshine-base__Qwen2.5-1.5B-Instruct__to__gemma-3-4b-it__none.clean.cap159_n}{1{,}116}
\defmmnum{xprofile.best_closed.moonshine-base__Qwen2.5-1.5B-Instruct__to__gemma-3-4b-it__none.clean.cap159_pct}{42.60}
\defmmnum{xprofile.best_closed.moonshine-base__Qwen2.5-1.5B-Instruct__to__gemma-3-4b-it__none.clean.cap160_n}{953}
\defmmnum{xprofile.best_closed.moonshine-base__Qwen2.5-1.5B-Instruct__to__gemma-3-4b-it__none.clean.cap160_pct}{36.37}
\defmmnum{xprofile.best_closed.moonshine-base__Qwen2.5-1.5B-Instruct__to__gemma-3-4b-it__none.clean.del_pct}{39.45}
\defmmnum{xprofile.best_closed.moonshine-base__Qwen2.5-1.5B-Instruct__to__gemma-3-4b-it__none.clean.del_words}{20{,}741}
\defmmnum{xprofile.best_closed.moonshine-base__Qwen2.5-1.5B-Instruct__to__gemma-3-4b-it__none.clean.empty_pct}{11.76}
\defmmnum{xprofile.best_closed.moonshine-base__Qwen2.5-1.5B-Instruct__to__gemma-3-4b-it__none.clean.ins_pct}{211.46}
\defmmnum{xprofile.best_closed.moonshine-base__Qwen2.5-1.5B-Instruct__to__gemma-3-4b-it__none.clean.ins_words}{111{,}179}
\defmmnum{xprofile.best_closed.moonshine-base__Qwen2.5-1.5B-Instruct__to__gemma-3-4b-it__none.clean.n_empty}{308}
\defmmnum{xprofile.best_closed.moonshine-base__Qwen2.5-1.5B-Instruct__to__gemma-3-4b-it__none.clean.n_utt}{2{,}620}
\defmmnum{xprofile.best_closed.moonshine-base__Qwen2.5-1.5B-Instruct__to__gemma-3-4b-it__none.clean.nonempty_n}{2{,}312}
\defmmnum{xprofile.best_closed.moonshine-base__Qwen2.5-1.5B-Instruct__to__gemma-3-4b-it__none.clean.nonempty_utt_median_wer}{390.45}
\defmmnum{xprofile.best_closed.moonshine-base__Qwen2.5-1.5B-Instruct__to__gemma-3-4b-it__none.clean.nonempty_wer}{363.32}
\defmmnum{xprofile.best_closed.moonshine-base__Qwen2.5-1.5B-Instruct__to__gemma-3-4b-it__none.clean.sub_pct}{53.73}
\defmmnum{xprofile.best_closed.moonshine-base__Qwen2.5-1.5B-Instruct__to__gemma-3-4b-it__none.clean.sub_words}{28{,}250}
\defmmnum{xprofile.best_closed.moonshine-base__Qwen2.5-1.5B-Instruct__to__gemma-3-4b-it__none.clean.utt_eq0_n}{0}
\defmmnum{xprofile.best_closed.moonshine-base__Qwen2.5-1.5B-Instruct__to__gemma-3-4b-it__none.clean.utt_eq0_pct}{0.00}
\defmmnum{xprofile.best_closed.moonshine-base__Qwen2.5-1.5B-Instruct__to__gemma-3-4b-it__none.clean.utt_ge100_n}{2{,}395}
\defmmnum{xprofile.best_closed.moonshine-base__Qwen2.5-1.5B-Instruct__to__gemma-3-4b-it__none.clean.utt_ge100_pct}{91.41}
\defmmnum{xprofile.best_closed.moonshine-base__Qwen2.5-1.5B-Instruct__to__gemma-3-4b-it__none.clean.utt_lt10_n}{0}
\defmmnum{xprofile.best_closed.moonshine-base__Qwen2.5-1.5B-Instruct__to__gemma-3-4b-it__none.clean.utt_lt10_pct}{0.00}
\defmmnum{xprofile.best_closed.moonshine-base__Qwen2.5-1.5B-Instruct__to__gemma-3-4b-it__none.clean.utt_lt50_n}{0}
\defmmnum{xprofile.best_closed.moonshine-base__Qwen2.5-1.5B-Instruct__to__gemma-3-4b-it__none.clean.utt_lt50_pct}{0.00}
\defmmnum{xprofile.best_closed.moonshine-base__Qwen2.5-1.5B-Instruct__to__gemma-3-4b-it__none.clean.utt_median_wer}{320.14}
\defmmnum{xprofile.best_closed.moonshine-base__Qwen2.5-1.5B-Instruct__to__gemma-3-4b-it__none.clean.wer}{304.64}
\defmmnum{xprofile.best_closed.moonshine-base__Qwen2.5-1.5B-Instruct__to__gemma-3-4b-it__none.clean.word_ratio_corpus}{2.72}
\defmmnum{xprofile.best_closed.moonshine-base__Qwen2.5-1.5B-Instruct__to__gemma-3-4b-it__none.clean.word_ratio_median}{3.31}
\defmmnum{xprofile.best_closed.moonshine-base__Qwen2.5-1.5B-Instruct__to__gemma-3-4b-it__none.other.cap159_n}{1{,}214}
\defmmnum{xprofile.best_closed.moonshine-base__Qwen2.5-1.5B-Instruct__to__gemma-3-4b-it__none.other.cap159_pct}{41.31}
\defmmnum{xprofile.best_closed.moonshine-base__Qwen2.5-1.5B-Instruct__to__gemma-3-4b-it__none.other.cap160_n}{1{,}093}
\defmmnum{xprofile.best_closed.moonshine-base__Qwen2.5-1.5B-Instruct__to__gemma-3-4b-it__none.other.cap160_pct}{37.19}
\defmmnum{xprofile.best_closed.moonshine-base__Qwen2.5-1.5B-Instruct__to__gemma-3-4b-it__none.other.del_pct}{29.63}
\defmmnum{xprofile.best_closed.moonshine-base__Qwen2.5-1.5B-Instruct__to__gemma-3-4b-it__none.other.del_words}{15{,}508}
\defmmnum{xprofile.best_closed.moonshine-base__Qwen2.5-1.5B-Instruct__to__gemma-3-4b-it__none.other.empty_pct}{7.83}
\defmmnum{xprofile.best_closed.moonshine-base__Qwen2.5-1.5B-Instruct__to__gemma-3-4b-it__none.other.ins_pct}{245.89}
\defmmnum{xprofile.best_closed.moonshine-base__Qwen2.5-1.5B-Instruct__to__gemma-3-4b-it__none.other.ins_words}{128{,}705}
\defmmnum{xprofile.best_closed.moonshine-base__Qwen2.5-1.5B-Instruct__to__gemma-3-4b-it__none.other.n_empty}{230}
\defmmnum{xprofile.best_closed.moonshine-base__Qwen2.5-1.5B-Instruct__to__gemma-3-4b-it__none.other.n_utt}{2{,}939}
\defmmnum{xprofile.best_closed.moonshine-base__Qwen2.5-1.5B-Instruct__to__gemma-3-4b-it__none.other.nonempty_n}{2{,}709}
\defmmnum{xprofile.best_closed.moonshine-base__Qwen2.5-1.5B-Instruct__to__gemma-3-4b-it__none.other.nonempty_utt_median_wer}{407.69}
\defmmnum{xprofile.best_closed.moonshine-base__Qwen2.5-1.5B-Instruct__to__gemma-3-4b-it__none.other.nonempty_wer}{381.48}
\defmmnum{xprofile.best_closed.moonshine-base__Qwen2.5-1.5B-Instruct__to__gemma-3-4b-it__none.other.sub_pct}{62.39}
\defmmnum{xprofile.best_closed.moonshine-base__Qwen2.5-1.5B-Instruct__to__gemma-3-4b-it__none.other.sub_words}{32{,}659}
\defmmnum{xprofile.best_closed.moonshine-base__Qwen2.5-1.5B-Instruct__to__gemma-3-4b-it__none.other.utt_eq0_n}{0}
\defmmnum{xprofile.best_closed.moonshine-base__Qwen2.5-1.5B-Instruct__to__gemma-3-4b-it__none.other.utt_eq0_pct}{0.00}
\defmmnum{xprofile.best_closed.moonshine-base__Qwen2.5-1.5B-Instruct__to__gemma-3-4b-it__none.other.utt_ge100_n}{2{,}732}
\defmmnum{xprofile.best_closed.moonshine-base__Qwen2.5-1.5B-Instruct__to__gemma-3-4b-it__none.other.utt_ge100_pct}{92.96}
\defmmnum{xprofile.best_closed.moonshine-base__Qwen2.5-1.5B-Instruct__to__gemma-3-4b-it__none.other.utt_lt10_n}{0}
\defmmnum{xprofile.best_closed.moonshine-base__Qwen2.5-1.5B-Instruct__to__gemma-3-4b-it__none.other.utt_lt10_pct}{0.00}
\defmmnum{xprofile.best_closed.moonshine-base__Qwen2.5-1.5B-Instruct__to__gemma-3-4b-it__none.other.utt_lt50_n}{0}
\defmmnum{xprofile.best_closed.moonshine-base__Qwen2.5-1.5B-Instruct__to__gemma-3-4b-it__none.other.utt_lt50_pct}{0.00}
\defmmnum{xprofile.best_closed.moonshine-base__Qwen2.5-1.5B-Instruct__to__gemma-3-4b-it__none.other.utt_median_wer}{362.50}
\defmmnum{xprofile.best_closed.moonshine-base__Qwen2.5-1.5B-Instruct__to__gemma-3-4b-it__none.other.wer}{337.91}
\defmmnum{xprofile.best_closed.moonshine-base__Qwen2.5-1.5B-Instruct__to__gemma-3-4b-it__none.other.word_ratio_corpus}{3.16}
\defmmnum{xprofile.best_closed.moonshine-base__Qwen2.5-1.5B-Instruct__to__gemma-3-4b-it__none.other.word_ratio_median}{3.78}
\defmmnum{xprofile.best_closed.moonshine-base__Qwen2.5-1.5B-Instruct__to__gemma-3-4b-it__rmsnorm.clean.cap159_n}{488}
\defmmnum{xprofile.best_closed.moonshine-base__Qwen2.5-1.5B-Instruct__to__gemma-3-4b-it__rmsnorm.clean.cap159_pct}{18.63}
\defmmnum{xprofile.best_closed.moonshine-base__Qwen2.5-1.5B-Instruct__to__gemma-3-4b-it__rmsnorm.clean.cap160_n}{168}
\defmmnum{xprofile.best_closed.moonshine-base__Qwen2.5-1.5B-Instruct__to__gemma-3-4b-it__rmsnorm.clean.cap160_pct}{6.41}
\defmmnum{xprofile.best_closed.moonshine-base__Qwen2.5-1.5B-Instruct__to__gemma-3-4b-it__rmsnorm.clean.del_pct}{87.54}
\defmmnum{xprofile.best_closed.moonshine-base__Qwen2.5-1.5B-Instruct__to__gemma-3-4b-it__rmsnorm.clean.del_words}{46{,}025}
\defmmnum{xprofile.best_closed.moonshine-base__Qwen2.5-1.5B-Instruct__to__gemma-3-4b-it__rmsnorm.clean.empty_pct}{63.97}
\defmmnum{xprofile.best_closed.moonshine-base__Qwen2.5-1.5B-Instruct__to__gemma-3-4b-it__rmsnorm.clean.ins_pct}{30.53}
\defmmnum{xprofile.best_closed.moonshine-base__Qwen2.5-1.5B-Instruct__to__gemma-3-4b-it__rmsnorm.clean.ins_words}{16{,}051}
\defmmnum{xprofile.best_closed.moonshine-base__Qwen2.5-1.5B-Instruct__to__gemma-3-4b-it__rmsnorm.clean.n_empty}{1{,}676}
\defmmnum{xprofile.best_closed.moonshine-base__Qwen2.5-1.5B-Instruct__to__gemma-3-4b-it__rmsnorm.clean.n_utt}{2{,}620}
\defmmnum{xprofile.best_closed.moonshine-base__Qwen2.5-1.5B-Instruct__to__gemma-3-4b-it__rmsnorm.clean.nonempty_n}{944}
\defmmnum{xprofile.best_closed.moonshine-base__Qwen2.5-1.5B-Instruct__to__gemma-3-4b-it__rmsnorm.clean.nonempty_utt_median_wer}{100.00}
\defmmnum{xprofile.best_closed.moonshine-base__Qwen2.5-1.5B-Instruct__to__gemma-3-4b-it__rmsnorm.clean.nonempty_wer}{193.68}
\defmmnum{xprofile.best_closed.moonshine-base__Qwen2.5-1.5B-Instruct__to__gemma-3-4b-it__rmsnorm.clean.sub_pct}{12.04}
\defmmnum{xprofile.best_closed.moonshine-base__Qwen2.5-1.5B-Instruct__to__gemma-3-4b-it__rmsnorm.clean.sub_words}{6{,}330}
\defmmnum{xprofile.best_closed.moonshine-base__Qwen2.5-1.5B-Instruct__to__gemma-3-4b-it__rmsnorm.clean.utt_eq0_n}{0}
\defmmnum{xprofile.best_closed.moonshine-base__Qwen2.5-1.5B-Instruct__to__gemma-3-4b-it__rmsnorm.clean.utt_eq0_pct}{0.00}
\defmmnum{xprofile.best_closed.moonshine-base__Qwen2.5-1.5B-Instruct__to__gemma-3-4b-it__rmsnorm.clean.utt_ge100_n}{2{,}510}
\defmmnum{xprofile.best_closed.moonshine-base__Qwen2.5-1.5B-Instruct__to__gemma-3-4b-it__rmsnorm.clean.utt_ge100_pct}{95.80}
\defmmnum{xprofile.best_closed.moonshine-base__Qwen2.5-1.5B-Instruct__to__gemma-3-4b-it__rmsnorm.clean.utt_lt10_n}{0}
\defmmnum{xprofile.best_closed.moonshine-base__Qwen2.5-1.5B-Instruct__to__gemma-3-4b-it__rmsnorm.clean.utt_lt10_pct}{0.00}
\defmmnum{xprofile.best_closed.moonshine-base__Qwen2.5-1.5B-Instruct__to__gemma-3-4b-it__rmsnorm.clean.utt_lt50_n}{0}
\defmmnum{xprofile.best_closed.moonshine-base__Qwen2.5-1.5B-Instruct__to__gemma-3-4b-it__rmsnorm.clean.utt_lt50_pct}{0.00}
\defmmnum{xprofile.best_closed.moonshine-base__Qwen2.5-1.5B-Instruct__to__gemma-3-4b-it__rmsnorm.clean.utt_median_wer}{100.00}
\defmmnum{xprofile.best_closed.moonshine-base__Qwen2.5-1.5B-Instruct__to__gemma-3-4b-it__rmsnorm.clean.wer}{130.11}
\defmmnum{xprofile.best_closed.moonshine-base__Qwen2.5-1.5B-Instruct__to__gemma-3-4b-it__rmsnorm.clean.word_ratio_corpus}{0.43}
\defmmnum{xprofile.best_closed.moonshine-base__Qwen2.5-1.5B-Instruct__to__gemma-3-4b-it__rmsnorm.clean.word_ratio_median}{0.00}
\defmmnum{xprofile.best_closed.moonshine-base__Qwen2.5-1.5B-Instruct__to__gemma-3-4b-it__rmsnorm.other.cap159_n}{576}
\defmmnum{xprofile.best_closed.moonshine-base__Qwen2.5-1.5B-Instruct__to__gemma-3-4b-it__rmsnorm.other.cap159_pct}{19.60}
\defmmnum{xprofile.best_closed.moonshine-base__Qwen2.5-1.5B-Instruct__to__gemma-3-4b-it__rmsnorm.other.cap160_n}{216}
\defmmnum{xprofile.best_closed.moonshine-base__Qwen2.5-1.5B-Instruct__to__gemma-3-4b-it__rmsnorm.other.cap160_pct}{7.35}
\defmmnum{xprofile.best_closed.moonshine-base__Qwen2.5-1.5B-Instruct__to__gemma-3-4b-it__rmsnorm.other.del_pct}{85.57}
\defmmnum{xprofile.best_closed.moonshine-base__Qwen2.5-1.5B-Instruct__to__gemma-3-4b-it__rmsnorm.other.del_words}{44{,}789}
\defmmnum{xprofile.best_closed.moonshine-base__Qwen2.5-1.5B-Instruct__to__gemma-3-4b-it__rmsnorm.other.empty_pct}{62.30}
\defmmnum{xprofile.best_closed.moonshine-base__Qwen2.5-1.5B-Instruct__to__gemma-3-4b-it__rmsnorm.other.ins_pct}{39.97}
\defmmnum{xprofile.best_closed.moonshine-base__Qwen2.5-1.5B-Instruct__to__gemma-3-4b-it__rmsnorm.other.ins_words}{20{,}920}
\defmmnum{xprofile.best_closed.moonshine-base__Qwen2.5-1.5B-Instruct__to__gemma-3-4b-it__rmsnorm.other.n_empty}{1{,}831}
\defmmnum{xprofile.best_closed.moonshine-base__Qwen2.5-1.5B-Instruct__to__gemma-3-4b-it__rmsnorm.other.n_utt}{2{,}939}
\defmmnum{xprofile.best_closed.moonshine-base__Qwen2.5-1.5B-Instruct__to__gemma-3-4b-it__rmsnorm.other.nonempty_n}{1{,}108}
\defmmnum{xprofile.best_closed.moonshine-base__Qwen2.5-1.5B-Instruct__to__gemma-3-4b-it__rmsnorm.other.nonempty_utt_median_wer}{100.00}
\defmmnum{xprofile.best_closed.moonshine-base__Qwen2.5-1.5B-Instruct__to__gemma-3-4b-it__rmsnorm.other.nonempty_wer}{215.02}
\defmmnum{xprofile.best_closed.moonshine-base__Qwen2.5-1.5B-Instruct__to__gemma-3-4b-it__rmsnorm.other.sub_pct}{13.98}
\defmmnum{xprofile.best_closed.moonshine-base__Qwen2.5-1.5B-Instruct__to__gemma-3-4b-it__rmsnorm.other.sub_words}{7{,}320}
\defmmnum{xprofile.best_closed.moonshine-base__Qwen2.5-1.5B-Instruct__to__gemma-3-4b-it__rmsnorm.other.utt_eq0_n}{0}
\defmmnum{xprofile.best_closed.moonshine-base__Qwen2.5-1.5B-Instruct__to__gemma-3-4b-it__rmsnorm.other.utt_eq0_pct}{0.00}
\defmmnum{xprofile.best_closed.moonshine-base__Qwen2.5-1.5B-Instruct__to__gemma-3-4b-it__rmsnorm.other.utt_ge100_n}{2{,}821}
\defmmnum{xprofile.best_closed.moonshine-base__Qwen2.5-1.5B-Instruct__to__gemma-3-4b-it__rmsnorm.other.utt_ge100_pct}{95.99}
\defmmnum{xprofile.best_closed.moonshine-base__Qwen2.5-1.5B-Instruct__to__gemma-3-4b-it__rmsnorm.other.utt_lt10_n}{0}
\defmmnum{xprofile.best_closed.moonshine-base__Qwen2.5-1.5B-Instruct__to__gemma-3-4b-it__rmsnorm.other.utt_lt10_pct}{0.00}
\defmmnum{xprofile.best_closed.moonshine-base__Qwen2.5-1.5B-Instruct__to__gemma-3-4b-it__rmsnorm.other.utt_lt50_n}{0}
\defmmnum{xprofile.best_closed.moonshine-base__Qwen2.5-1.5B-Instruct__to__gemma-3-4b-it__rmsnorm.other.utt_lt50_pct}{0.00}
\defmmnum{xprofile.best_closed.moonshine-base__Qwen2.5-1.5B-Instruct__to__gemma-3-4b-it__rmsnorm.other.utt_median_wer}{100.00}
\defmmnum{xprofile.best_closed.moonshine-base__Qwen2.5-1.5B-Instruct__to__gemma-3-4b-it__rmsnorm.other.wer}{139.52}
\defmmnum{xprofile.best_closed.moonshine-base__Qwen2.5-1.5B-Instruct__to__gemma-3-4b-it__rmsnorm.other.word_ratio_corpus}{0.54}
\defmmnum{xprofile.best_closed.moonshine-base__Qwen2.5-1.5B-Instruct__to__gemma-3-4b-it__rmsnorm.other.word_ratio_median}{0.00}
\defmmnum{xprofile.best_closed.moonshine-base__gemma-2-2b-it__to__Qwen2.5-1.5B-Instruct__l2norm.clean.cap159_n}{1}
\defmmnum{xprofile.best_closed.moonshine-base__gemma-2-2b-it__to__Qwen2.5-1.5B-Instruct__l2norm.clean.cap159_pct}{0.04}
\defmmnum{xprofile.best_closed.moonshine-base__gemma-2-2b-it__to__Qwen2.5-1.5B-Instruct__l2norm.clean.cap160_n}{0}
\defmmnum{xprofile.best_closed.moonshine-base__gemma-2-2b-it__to__Qwen2.5-1.5B-Instruct__l2norm.clean.cap160_pct}{0.00}
\defmmnum{xprofile.best_closed.moonshine-base__gemma-2-2b-it__to__Qwen2.5-1.5B-Instruct__l2norm.clean.del_pct}{15.10}
\defmmnum{xprofile.best_closed.moonshine-base__gemma-2-2b-it__to__Qwen2.5-1.5B-Instruct__l2norm.clean.del_words}{7{,}938}
\defmmnum{xprofile.best_closed.moonshine-base__gemma-2-2b-it__to__Qwen2.5-1.5B-Instruct__l2norm.clean.empty_pct}{0.00}
\defmmnum{xprofile.best_closed.moonshine-base__gemma-2-2b-it__to__Qwen2.5-1.5B-Instruct__l2norm.clean.ins_pct}{65.35}
\defmmnum{xprofile.best_closed.moonshine-base__gemma-2-2b-it__to__Qwen2.5-1.5B-Instruct__l2norm.clean.ins_words}{34{,}358}
\defmmnum{xprofile.best_closed.moonshine-base__gemma-2-2b-it__to__Qwen2.5-1.5B-Instruct__l2norm.clean.n_empty}{0}
\defmmnum{xprofile.best_closed.moonshine-base__gemma-2-2b-it__to__Qwen2.5-1.5B-Instruct__l2norm.clean.n_utt}{2{,}620}
\defmmnum{xprofile.best_closed.moonshine-base__gemma-2-2b-it__to__Qwen2.5-1.5B-Instruct__l2norm.clean.nonempty_n}{2{,}620}
\defmmnum{xprofile.best_closed.moonshine-base__gemma-2-2b-it__to__Qwen2.5-1.5B-Instruct__l2norm.clean.nonempty_utt_median_wer}{169.78}
\defmmnum{xprofile.best_closed.moonshine-base__gemma-2-2b-it__to__Qwen2.5-1.5B-Instruct__l2norm.clean.nonempty_wer}{159.00}
\defmmnum{xprofile.best_closed.moonshine-base__gemma-2-2b-it__to__Qwen2.5-1.5B-Instruct__l2norm.clean.sub_pct}{78.55}
\defmmnum{xprofile.best_closed.moonshine-base__gemma-2-2b-it__to__Qwen2.5-1.5B-Instruct__l2norm.clean.sub_words}{41{,}298}
\defmmnum{xprofile.best_closed.moonshine-base__gemma-2-2b-it__to__Qwen2.5-1.5B-Instruct__l2norm.clean.utt_eq0_n}{0}
\defmmnum{xprofile.best_closed.moonshine-base__gemma-2-2b-it__to__Qwen2.5-1.5B-Instruct__l2norm.clean.utt_eq0_pct}{0.00}
\defmmnum{xprofile.best_closed.moonshine-base__gemma-2-2b-it__to__Qwen2.5-1.5B-Instruct__l2norm.clean.utt_ge100_n}{2{,}198}
\defmmnum{xprofile.best_closed.moonshine-base__gemma-2-2b-it__to__Qwen2.5-1.5B-Instruct__l2norm.clean.utt_ge100_pct}{83.89}
\defmmnum{xprofile.best_closed.moonshine-base__gemma-2-2b-it__to__Qwen2.5-1.5B-Instruct__l2norm.clean.utt_lt10_n}{0}
\defmmnum{xprofile.best_closed.moonshine-base__gemma-2-2b-it__to__Qwen2.5-1.5B-Instruct__l2norm.clean.utt_lt10_pct}{0.00}
\defmmnum{xprofile.best_closed.moonshine-base__gemma-2-2b-it__to__Qwen2.5-1.5B-Instruct__l2norm.clean.utt_lt50_n}{0}
\defmmnum{xprofile.best_closed.moonshine-base__gemma-2-2b-it__to__Qwen2.5-1.5B-Instruct__l2norm.clean.utt_lt50_pct}{0.00}
\defmmnum{xprofile.best_closed.moonshine-base__gemma-2-2b-it__to__Qwen2.5-1.5B-Instruct__l2norm.clean.utt_median_wer}{169.78}
\defmmnum{xprofile.best_closed.moonshine-base__gemma-2-2b-it__to__Qwen2.5-1.5B-Instruct__l2norm.clean.wer}{159.00}
\defmmnum{xprofile.best_closed.moonshine-base__gemma-2-2b-it__to__Qwen2.5-1.5B-Instruct__l2norm.clean.word_ratio_corpus}{1.50}
\defmmnum{xprofile.best_closed.moonshine-base__gemma-2-2b-it__to__Qwen2.5-1.5B-Instruct__l2norm.clean.word_ratio_median}{1.78}
\defmmnum{xprofile.best_closed.moonshine-base__gemma-2-2b-it__to__Qwen2.5-1.5B-Instruct__l2norm.other.cap159_n}{1}
\defmmnum{xprofile.best_closed.moonshine-base__gemma-2-2b-it__to__Qwen2.5-1.5B-Instruct__l2norm.other.cap159_pct}{0.03}
\defmmnum{xprofile.best_closed.moonshine-base__gemma-2-2b-it__to__Qwen2.5-1.5B-Instruct__l2norm.other.cap160_n}{1}
\defmmnum{xprofile.best_closed.moonshine-base__gemma-2-2b-it__to__Qwen2.5-1.5B-Instruct__l2norm.other.cap160_pct}{0.03}
\defmmnum{xprofile.best_closed.moonshine-base__gemma-2-2b-it__to__Qwen2.5-1.5B-Instruct__l2norm.other.del_pct}{13.25}
\defmmnum{xprofile.best_closed.moonshine-base__gemma-2-2b-it__to__Qwen2.5-1.5B-Instruct__l2norm.other.del_words}{6{,}934}
\defmmnum{xprofile.best_closed.moonshine-base__gemma-2-2b-it__to__Qwen2.5-1.5B-Instruct__l2norm.other.empty_pct}{0.00}
\defmmnum{xprofile.best_closed.moonshine-base__gemma-2-2b-it__to__Qwen2.5-1.5B-Instruct__l2norm.other.ins_pct}{82.19}
\defmmnum{xprofile.best_closed.moonshine-base__gemma-2-2b-it__to__Qwen2.5-1.5B-Instruct__l2norm.other.ins_words}{43{,}020}
\defmmnum{xprofile.best_closed.moonshine-base__gemma-2-2b-it__to__Qwen2.5-1.5B-Instruct__l2norm.other.n_empty}{0}
\defmmnum{xprofile.best_closed.moonshine-base__gemma-2-2b-it__to__Qwen2.5-1.5B-Instruct__l2norm.other.n_utt}{2{,}939}
\defmmnum{xprofile.best_closed.moonshine-base__gemma-2-2b-it__to__Qwen2.5-1.5B-Instruct__l2norm.other.nonempty_n}{2{,}939}
\defmmnum{xprofile.best_closed.moonshine-base__gemma-2-2b-it__to__Qwen2.5-1.5B-Instruct__l2norm.other.nonempty_utt_median_wer}{193.75}
\defmmnum{xprofile.best_closed.moonshine-base__gemma-2-2b-it__to__Qwen2.5-1.5B-Instruct__l2norm.other.nonempty_wer}{175.40}
\defmmnum{xprofile.best_closed.moonshine-base__gemma-2-2b-it__to__Qwen2.5-1.5B-Instruct__l2norm.other.sub_pct}{79.96}
\defmmnum{xprofile.best_closed.moonshine-base__gemma-2-2b-it__to__Qwen2.5-1.5B-Instruct__l2norm.other.sub_words}{41{,}854}
\defmmnum{xprofile.best_closed.moonshine-base__gemma-2-2b-it__to__Qwen2.5-1.5B-Instruct__l2norm.other.utt_eq0_n}{0}
\defmmnum{xprofile.best_closed.moonshine-base__gemma-2-2b-it__to__Qwen2.5-1.5B-Instruct__l2norm.other.utt_eq0_pct}{0.00}
\defmmnum{xprofile.best_closed.moonshine-base__gemma-2-2b-it__to__Qwen2.5-1.5B-Instruct__l2norm.other.utt_ge100_n}{2{,}599}
\defmmnum{xprofile.best_closed.moonshine-base__gemma-2-2b-it__to__Qwen2.5-1.5B-Instruct__l2norm.other.utt_ge100_pct}{88.43}
\defmmnum{xprofile.best_closed.moonshine-base__gemma-2-2b-it__to__Qwen2.5-1.5B-Instruct__l2norm.other.utt_lt10_n}{0}
\defmmnum{xprofile.best_closed.moonshine-base__gemma-2-2b-it__to__Qwen2.5-1.5B-Instruct__l2norm.other.utt_lt10_pct}{0.00}
\defmmnum{xprofile.best_closed.moonshine-base__gemma-2-2b-it__to__Qwen2.5-1.5B-Instruct__l2norm.other.utt_lt50_n}{0}
\defmmnum{xprofile.best_closed.moonshine-base__gemma-2-2b-it__to__Qwen2.5-1.5B-Instruct__l2norm.other.utt_lt50_pct}{0.00}
\defmmnum{xprofile.best_closed.moonshine-base__gemma-2-2b-it__to__Qwen2.5-1.5B-Instruct__l2norm.other.utt_median_wer}{193.75}
\defmmnum{xprofile.best_closed.moonshine-base__gemma-2-2b-it__to__Qwen2.5-1.5B-Instruct__l2norm.other.wer}{175.40}
\defmmnum{xprofile.best_closed.moonshine-base__gemma-2-2b-it__to__Qwen2.5-1.5B-Instruct__l2norm.other.word_ratio_corpus}{1.69}
\defmmnum{xprofile.best_closed.moonshine-base__gemma-2-2b-it__to__Qwen2.5-1.5B-Instruct__l2norm.other.word_ratio_median}{2.06}
\defmmnum{xprofile.best_closed.moonshine-base__gemma-2-2b-it__to__Qwen2.5-1.5B-Instruct__layernorm.clean.cap159_n}{1}
\defmmnum{xprofile.best_closed.moonshine-base__gemma-2-2b-it__to__Qwen2.5-1.5B-Instruct__layernorm.clean.cap159_pct}{0.04}
\defmmnum{xprofile.best_closed.moonshine-base__gemma-2-2b-it__to__Qwen2.5-1.5B-Instruct__layernorm.clean.cap160_n}{1}
\defmmnum{xprofile.best_closed.moonshine-base__gemma-2-2b-it__to__Qwen2.5-1.5B-Instruct__layernorm.clean.cap160_pct}{0.04}
\defmmnum{xprofile.best_closed.moonshine-base__gemma-2-2b-it__to__Qwen2.5-1.5B-Instruct__layernorm.clean.del_pct}{59.25}
\defmmnum{xprofile.best_closed.moonshine-base__gemma-2-2b-it__to__Qwen2.5-1.5B-Instruct__layernorm.clean.del_words}{31{,}150}
\defmmnum{xprofile.best_closed.moonshine-base__gemma-2-2b-it__to__Qwen2.5-1.5B-Instruct__layernorm.clean.empty_pct}{0.00}
\defmmnum{xprofile.best_closed.moonshine-base__gemma-2-2b-it__to__Qwen2.5-1.5B-Instruct__layernorm.clean.ins_pct}{10.85}
\defmmnum{xprofile.best_closed.moonshine-base__gemma-2-2b-it__to__Qwen2.5-1.5B-Instruct__layernorm.clean.ins_words}{5{,}706}
\defmmnum{xprofile.best_closed.moonshine-base__gemma-2-2b-it__to__Qwen2.5-1.5B-Instruct__layernorm.clean.n_empty}{0}
\defmmnum{xprofile.best_closed.moonshine-base__gemma-2-2b-it__to__Qwen2.5-1.5B-Instruct__layernorm.clean.n_utt}{2{,}620}
\defmmnum{xprofile.best_closed.moonshine-base__gemma-2-2b-it__to__Qwen2.5-1.5B-Instruct__layernorm.clean.nonempty_n}{2{,}620}
\defmmnum{xprofile.best_closed.moonshine-base__gemma-2-2b-it__to__Qwen2.5-1.5B-Instruct__layernorm.clean.nonempty_utt_median_wer}{100.00}
\defmmnum{xprofile.best_closed.moonshine-base__gemma-2-2b-it__to__Qwen2.5-1.5B-Instruct__layernorm.clean.nonempty_wer}{108.32}
\defmmnum{xprofile.best_closed.moonshine-base__gemma-2-2b-it__to__Qwen2.5-1.5B-Instruct__layernorm.clean.sub_pct}{38.22}
\defmmnum{xprofile.best_closed.moonshine-base__gemma-2-2b-it__to__Qwen2.5-1.5B-Instruct__layernorm.clean.sub_words}{20{,}092}
\defmmnum{xprofile.best_closed.moonshine-base__gemma-2-2b-it__to__Qwen2.5-1.5B-Instruct__layernorm.clean.utt_eq0_n}{0}
\defmmnum{xprofile.best_closed.moonshine-base__gemma-2-2b-it__to__Qwen2.5-1.5B-Instruct__layernorm.clean.utt_eq0_pct}{0.00}
\defmmnum{xprofile.best_closed.moonshine-base__gemma-2-2b-it__to__Qwen2.5-1.5B-Instruct__layernorm.clean.utt_ge100_n}{1{,}788}
\defmmnum{xprofile.best_closed.moonshine-base__gemma-2-2b-it__to__Qwen2.5-1.5B-Instruct__layernorm.clean.utt_ge100_pct}{68.24}
\defmmnum{xprofile.best_closed.moonshine-base__gemma-2-2b-it__to__Qwen2.5-1.5B-Instruct__layernorm.clean.utt_lt10_n}{0}
\defmmnum{xprofile.best_closed.moonshine-base__gemma-2-2b-it__to__Qwen2.5-1.5B-Instruct__layernorm.clean.utt_lt10_pct}{0.00}
\defmmnum{xprofile.best_closed.moonshine-base__gemma-2-2b-it__to__Qwen2.5-1.5B-Instruct__layernorm.clean.utt_lt50_n}{0}
\defmmnum{xprofile.best_closed.moonshine-base__gemma-2-2b-it__to__Qwen2.5-1.5B-Instruct__layernorm.clean.utt_lt50_pct}{0.00}
\defmmnum{xprofile.best_closed.moonshine-base__gemma-2-2b-it__to__Qwen2.5-1.5B-Instruct__layernorm.clean.utt_median_wer}{100.00}
\defmmnum{xprofile.best_closed.moonshine-base__gemma-2-2b-it__to__Qwen2.5-1.5B-Instruct__layernorm.clean.wer}{108.32}
\defmmnum{xprofile.best_closed.moonshine-base__gemma-2-2b-it__to__Qwen2.5-1.5B-Instruct__layernorm.clean.word_ratio_corpus}{0.52}
\defmmnum{xprofile.best_closed.moonshine-base__gemma-2-2b-it__to__Qwen2.5-1.5B-Instruct__layernorm.clean.word_ratio_median}{0.50}
\defmmnum{xprofile.best_closed.moonshine-base__gemma-2-2b-it__to__Qwen2.5-1.5B-Instruct__layernorm.other.cap159_n}{0}
\defmmnum{xprofile.best_closed.moonshine-base__gemma-2-2b-it__to__Qwen2.5-1.5B-Instruct__layernorm.other.cap159_pct}{0.00}
\defmmnum{xprofile.best_closed.moonshine-base__gemma-2-2b-it__to__Qwen2.5-1.5B-Instruct__layernorm.other.cap160_n}{0}
\defmmnum{xprofile.best_closed.moonshine-base__gemma-2-2b-it__to__Qwen2.5-1.5B-Instruct__layernorm.other.cap160_pct}{0.00}
\defmmnum{xprofile.best_closed.moonshine-base__gemma-2-2b-it__to__Qwen2.5-1.5B-Instruct__layernorm.other.del_pct}{54.42}
\defmmnum{xprofile.best_closed.moonshine-base__gemma-2-2b-it__to__Qwen2.5-1.5B-Instruct__layernorm.other.del_words}{28{,}487}
\defmmnum{xprofile.best_closed.moonshine-base__gemma-2-2b-it__to__Qwen2.5-1.5B-Instruct__layernorm.other.empty_pct}{0.00}
\defmmnum{xprofile.best_closed.moonshine-base__gemma-2-2b-it__to__Qwen2.5-1.5B-Instruct__layernorm.other.ins_pct}{16.28}
\defmmnum{xprofile.best_closed.moonshine-base__gemma-2-2b-it__to__Qwen2.5-1.5B-Instruct__layernorm.other.ins_words}{8{,}522}
\defmmnum{xprofile.best_closed.moonshine-base__gemma-2-2b-it__to__Qwen2.5-1.5B-Instruct__layernorm.other.n_empty}{0}
\defmmnum{xprofile.best_closed.moonshine-base__gemma-2-2b-it__to__Qwen2.5-1.5B-Instruct__layernorm.other.n_utt}{2{,}939}
\defmmnum{xprofile.best_closed.moonshine-base__gemma-2-2b-it__to__Qwen2.5-1.5B-Instruct__layernorm.other.nonempty_n}{2{,}939}
\defmmnum{xprofile.best_closed.moonshine-base__gemma-2-2b-it__to__Qwen2.5-1.5B-Instruct__layernorm.other.nonempty_utt_median_wer}{100.00}
\defmmnum{xprofile.best_closed.moonshine-base__gemma-2-2b-it__to__Qwen2.5-1.5B-Instruct__layernorm.other.nonempty_wer}{113.60}
\defmmnum{xprofile.best_closed.moonshine-base__gemma-2-2b-it__to__Qwen2.5-1.5B-Instruct__layernorm.other.sub_pct}{42.90}
\defmmnum{xprofile.best_closed.moonshine-base__gemma-2-2b-it__to__Qwen2.5-1.5B-Instruct__layernorm.other.sub_words}{22{,}454}
\defmmnum{xprofile.best_closed.moonshine-base__gemma-2-2b-it__to__Qwen2.5-1.5B-Instruct__layernorm.other.utt_eq0_n}{0}
\defmmnum{xprofile.best_closed.moonshine-base__gemma-2-2b-it__to__Qwen2.5-1.5B-Instruct__layernorm.other.utt_eq0_pct}{0.00}
\defmmnum{xprofile.best_closed.moonshine-base__gemma-2-2b-it__to__Qwen2.5-1.5B-Instruct__layernorm.other.utt_ge100_n}{2{,}121}
\defmmnum{xprofile.best_closed.moonshine-base__gemma-2-2b-it__to__Qwen2.5-1.5B-Instruct__layernorm.other.utt_ge100_pct}{72.17}
\defmmnum{xprofile.best_closed.moonshine-base__gemma-2-2b-it__to__Qwen2.5-1.5B-Instruct__layernorm.other.utt_lt10_n}{0}
\defmmnum{xprofile.best_closed.moonshine-base__gemma-2-2b-it__to__Qwen2.5-1.5B-Instruct__layernorm.other.utt_lt10_pct}{0.00}
\defmmnum{xprofile.best_closed.moonshine-base__gemma-2-2b-it__to__Qwen2.5-1.5B-Instruct__layernorm.other.utt_lt50_n}{0}
\defmmnum{xprofile.best_closed.moonshine-base__gemma-2-2b-it__to__Qwen2.5-1.5B-Instruct__layernorm.other.utt_lt50_pct}{0.00}
\defmmnum{xprofile.best_closed.moonshine-base__gemma-2-2b-it__to__Qwen2.5-1.5B-Instruct__layernorm.other.utt_median_wer}{100.00}
\defmmnum{xprofile.best_closed.moonshine-base__gemma-2-2b-it__to__Qwen2.5-1.5B-Instruct__layernorm.other.wer}{113.60}
\defmmnum{xprofile.best_closed.moonshine-base__gemma-2-2b-it__to__Qwen2.5-1.5B-Instruct__layernorm.other.word_ratio_corpus}{0.62}
\defmmnum{xprofile.best_closed.moonshine-base__gemma-2-2b-it__to__Qwen2.5-1.5B-Instruct__layernorm.other.word_ratio_median}{0.62}
\defmmnum{xprofile.best_closed.moonshine-base__gemma-2-2b-it__to__gemma-3-4b-it__l2norm.clean.cap159_n}{849}
\defmmnum{xprofile.best_closed.moonshine-base__gemma-2-2b-it__to__gemma-3-4b-it__l2norm.clean.cap159_pct}{32.40}
\defmmnum{xprofile.best_closed.moonshine-base__gemma-2-2b-it__to__gemma-3-4b-it__l2norm.clean.cap160_n}{795}
\defmmnum{xprofile.best_closed.moonshine-base__gemma-2-2b-it__to__gemma-3-4b-it__l2norm.clean.cap160_pct}{30.34}
\defmmnum{xprofile.best_closed.moonshine-base__gemma-2-2b-it__to__gemma-3-4b-it__l2norm.clean.del_pct}{23.67}
\defmmnum{xprofile.best_closed.moonshine-base__gemma-2-2b-it__to__gemma-3-4b-it__l2norm.clean.del_words}{12{,}445}
\defmmnum{xprofile.best_closed.moonshine-base__gemma-2-2b-it__to__gemma-3-4b-it__l2norm.clean.empty_pct}{0.19}
\defmmnum{xprofile.best_closed.moonshine-base__gemma-2-2b-it__to__gemma-3-4b-it__l2norm.clean.ins_pct}{173.54}
\defmmnum{xprofile.best_closed.moonshine-base__gemma-2-2b-it__to__gemma-3-4b-it__l2norm.clean.ins_words}{91{,}238}
\defmmnum{xprofile.best_closed.moonshine-base__gemma-2-2b-it__to__gemma-3-4b-it__l2norm.clean.n_empty}{5}
\defmmnum{xprofile.best_closed.moonshine-base__gemma-2-2b-it__to__gemma-3-4b-it__l2norm.clean.n_utt}{2{,}620}
\defmmnum{xprofile.best_closed.moonshine-base__gemma-2-2b-it__to__gemma-3-4b-it__l2norm.clean.nonempty_n}{2{,}615}
\defmmnum{xprofile.best_closed.moonshine-base__gemma-2-2b-it__to__gemma-3-4b-it__l2norm.clean.nonempty_utt_median_wer}{266.67}
\defmmnum{xprofile.best_closed.moonshine-base__gemma-2-2b-it__to__gemma-3-4b-it__l2norm.clean.nonempty_wer}{265.74}
\defmmnum{xprofile.best_closed.moonshine-base__gemma-2-2b-it__to__gemma-3-4b-it__l2norm.clean.sub_pct}{67.74}
\defmmnum{xprofile.best_closed.moonshine-base__gemma-2-2b-it__to__gemma-3-4b-it__l2norm.clean.sub_words}{35{,}613}
\defmmnum{xprofile.best_closed.moonshine-base__gemma-2-2b-it__to__gemma-3-4b-it__l2norm.clean.utt_eq0_n}{0}
\defmmnum{xprofile.best_closed.moonshine-base__gemma-2-2b-it__to__gemma-3-4b-it__l2norm.clean.utt_eq0_pct}{0.00}
\defmmnum{xprofile.best_closed.moonshine-base__gemma-2-2b-it__to__gemma-3-4b-it__l2norm.clean.utt_ge100_n}{2{,}227}
\defmmnum{xprofile.best_closed.moonshine-base__gemma-2-2b-it__to__gemma-3-4b-it__l2norm.clean.utt_ge100_pct}{85.00}
\defmmnum{xprofile.best_closed.moonshine-base__gemma-2-2b-it__to__gemma-3-4b-it__l2norm.clean.utt_lt10_n}{0}
\defmmnum{xprofile.best_closed.moonshine-base__gemma-2-2b-it__to__gemma-3-4b-it__l2norm.clean.utt_lt10_pct}{0.00}
\defmmnum{xprofile.best_closed.moonshine-base__gemma-2-2b-it__to__gemma-3-4b-it__l2norm.clean.utt_lt50_n}{0}
\defmmnum{xprofile.best_closed.moonshine-base__gemma-2-2b-it__to__gemma-3-4b-it__l2norm.clean.utt_lt50_pct}{0.00}
\defmmnum{xprofile.best_closed.moonshine-base__gemma-2-2b-it__to__gemma-3-4b-it__l2norm.clean.utt_median_wer}{264.61}
\defmmnum{xprofile.best_closed.moonshine-base__gemma-2-2b-it__to__gemma-3-4b-it__l2norm.clean.wer}{264.94}
\defmmnum{xprofile.best_closed.moonshine-base__gemma-2-2b-it__to__gemma-3-4b-it__l2norm.clean.word_ratio_corpus}{2.50}
\defmmnum{xprofile.best_closed.moonshine-base__gemma-2-2b-it__to__gemma-3-4b-it__l2norm.clean.word_ratio_median}{2.81}
\defmmnum{xprofile.best_closed.moonshine-base__gemma-2-2b-it__to__gemma-3-4b-it__l2norm.other.cap159_n}{1{,}107}
\defmmnum{xprofile.best_closed.moonshine-base__gemma-2-2b-it__to__gemma-3-4b-it__l2norm.other.cap159_pct}{37.67}
\defmmnum{xprofile.best_closed.moonshine-base__gemma-2-2b-it__to__gemma-3-4b-it__l2norm.other.cap160_n}{1{,}008}
\defmmnum{xprofile.best_closed.moonshine-base__gemma-2-2b-it__to__gemma-3-4b-it__l2norm.other.cap160_pct}{34.30}
\defmmnum{xprofile.best_closed.moonshine-base__gemma-2-2b-it__to__gemma-3-4b-it__l2norm.other.del_pct}{18.45}
\defmmnum{xprofile.best_closed.moonshine-base__gemma-2-2b-it__to__gemma-3-4b-it__l2norm.other.del_words}{9{,}658}
\defmmnum{xprofile.best_closed.moonshine-base__gemma-2-2b-it__to__gemma-3-4b-it__l2norm.other.empty_pct}{0.03}
\defmmnum{xprofile.best_closed.moonshine-base__gemma-2-2b-it__to__gemma-3-4b-it__l2norm.other.ins_pct}{224.68}
\defmmnum{xprofile.best_closed.moonshine-base__gemma-2-2b-it__to__gemma-3-4b-it__l2norm.other.ins_words}{117{,}604}
\defmmnum{xprofile.best_closed.moonshine-base__gemma-2-2b-it__to__gemma-3-4b-it__l2norm.other.n_empty}{1}
\defmmnum{xprofile.best_closed.moonshine-base__gemma-2-2b-it__to__gemma-3-4b-it__l2norm.other.n_utt}{2{,}939}
\defmmnum{xprofile.best_closed.moonshine-base__gemma-2-2b-it__to__gemma-3-4b-it__l2norm.other.nonempty_n}{2{,}938}
\defmmnum{xprofile.best_closed.moonshine-base__gemma-2-2b-it__to__gemma-3-4b-it__l2norm.other.nonempty_utt_median_wer}{336.36}
\defmmnum{xprofile.best_closed.moonshine-base__gemma-2-2b-it__to__gemma-3-4b-it__l2norm.other.nonempty_wer}{315.31}
\defmmnum{xprofile.best_closed.moonshine-base__gemma-2-2b-it__to__gemma-3-4b-it__l2norm.other.sub_pct}{72.11}
\defmmnum{xprofile.best_closed.moonshine-base__gemma-2-2b-it__to__gemma-3-4b-it__l2norm.other.sub_words}{37{,}743}
\defmmnum{xprofile.best_closed.moonshine-base__gemma-2-2b-it__to__gemma-3-4b-it__l2norm.other.utt_eq0_n}{0}
\defmmnum{xprofile.best_closed.moonshine-base__gemma-2-2b-it__to__gemma-3-4b-it__l2norm.other.utt_eq0_pct}{0.00}
\defmmnum{xprofile.best_closed.moonshine-base__gemma-2-2b-it__to__gemma-3-4b-it__l2norm.other.utt_ge100_n}{2{,}593}
\defmmnum{xprofile.best_closed.moonshine-base__gemma-2-2b-it__to__gemma-3-4b-it__l2norm.other.utt_ge100_pct}{88.23}
\defmmnum{xprofile.best_closed.moonshine-base__gemma-2-2b-it__to__gemma-3-4b-it__l2norm.other.utt_lt10_n}{0}
\defmmnum{xprofile.best_closed.moonshine-base__gemma-2-2b-it__to__gemma-3-4b-it__l2norm.other.utt_lt10_pct}{0.00}
\defmmnum{xprofile.best_closed.moonshine-base__gemma-2-2b-it__to__gemma-3-4b-it__l2norm.other.utt_lt50_n}{0}
\defmmnum{xprofile.best_closed.moonshine-base__gemma-2-2b-it__to__gemma-3-4b-it__l2norm.other.utt_lt50_pct}{0.00}
\defmmnum{xprofile.best_closed.moonshine-base__gemma-2-2b-it__to__gemma-3-4b-it__l2norm.other.utt_median_wer}{336.36}
\defmmnum{xprofile.best_closed.moonshine-base__gemma-2-2b-it__to__gemma-3-4b-it__l2norm.other.wer}{315.24}
\defmmnum{xprofile.best_closed.moonshine-base__gemma-2-2b-it__to__gemma-3-4b-it__l2norm.other.word_ratio_corpus}{3.06}
\defmmnum{xprofile.best_closed.moonshine-base__gemma-2-2b-it__to__gemma-3-4b-it__l2norm.other.word_ratio_median}{3.50}
\defmmnum{xprofile.best_closed.whisper-small__Qwen2.5-1.5B-Instruct__to__gemma-2-2b-it__none.clean.cap159_n}{818}
\defmmnum{xprofile.best_closed.whisper-small__Qwen2.5-1.5B-Instruct__to__gemma-2-2b-it__none.clean.cap159_pct}{31.22}
\defmmnum{xprofile.best_closed.whisper-small__Qwen2.5-1.5B-Instruct__to__gemma-2-2b-it__none.clean.cap160_n}{208}
\defmmnum{xprofile.best_closed.whisper-small__Qwen2.5-1.5B-Instruct__to__gemma-2-2b-it__none.clean.cap160_pct}{7.94}
\defmmnum{xprofile.best_closed.whisper-small__Qwen2.5-1.5B-Instruct__to__gemma-2-2b-it__none.clean.del_pct}{91.49}
\defmmnum{xprofile.best_closed.whisper-small__Qwen2.5-1.5B-Instruct__to__gemma-2-2b-it__none.clean.del_words}{48{,}101}
\defmmnum{xprofile.best_closed.whisper-small__Qwen2.5-1.5B-Instruct__to__gemma-2-2b-it__none.clean.empty_pct}{78.05}
\defmmnum{xprofile.best_closed.whisper-small__Qwen2.5-1.5B-Instruct__to__gemma-2-2b-it__none.clean.ins_pct}{12.38}
\defmmnum{xprofile.best_closed.whisper-small__Qwen2.5-1.5B-Instruct__to__gemma-2-2b-it__none.clean.ins_words}{6{,}508}
\defmmnum{xprofile.best_closed.whisper-small__Qwen2.5-1.5B-Instruct__to__gemma-2-2b-it__none.clean.n_empty}{2{,}045}
\defmmnum{xprofile.best_closed.whisper-small__Qwen2.5-1.5B-Instruct__to__gemma-2-2b-it__none.clean.n_utt}{2{,}620}
\defmmnum{xprofile.best_closed.whisper-small__Qwen2.5-1.5B-Instruct__to__gemma-2-2b-it__none.clean.nonempty_n}{575}
\defmmnum{xprofile.best_closed.whisper-small__Qwen2.5-1.5B-Instruct__to__gemma-2-2b-it__none.clean.nonempty_utt_median_wer}{100.00}
\defmmnum{xprofile.best_closed.whisper-small__Qwen2.5-1.5B-Instruct__to__gemma-2-2b-it__none.clean.nonempty_wer}{152.34}
\defmmnum{xprofile.best_closed.whisper-small__Qwen2.5-1.5B-Instruct__to__gemma-2-2b-it__none.clean.sub_pct}{7.94}
\defmmnum{xprofile.best_closed.whisper-small__Qwen2.5-1.5B-Instruct__to__gemma-2-2b-it__none.clean.sub_words}{4{,}175}
\defmmnum{xprofile.best_closed.whisper-small__Qwen2.5-1.5B-Instruct__to__gemma-2-2b-it__none.clean.utt_eq0_n}{0}
\defmmnum{xprofile.best_closed.whisper-small__Qwen2.5-1.5B-Instruct__to__gemma-2-2b-it__none.clean.utt_eq0_pct}{0.00}
\defmmnum{xprofile.best_closed.whisper-small__Qwen2.5-1.5B-Instruct__to__gemma-2-2b-it__none.clean.utt_ge100_n}{2{,}484}
\defmmnum{xprofile.best_closed.whisper-small__Qwen2.5-1.5B-Instruct__to__gemma-2-2b-it__none.clean.utt_ge100_pct}{94.81}
\defmmnum{xprofile.best_closed.whisper-small__Qwen2.5-1.5B-Instruct__to__gemma-2-2b-it__none.clean.utt_lt10_n}{0}
\defmmnum{xprofile.best_closed.whisper-small__Qwen2.5-1.5B-Instruct__to__gemma-2-2b-it__none.clean.utt_lt10_pct}{0.00}
\defmmnum{xprofile.best_closed.whisper-small__Qwen2.5-1.5B-Instruct__to__gemma-2-2b-it__none.clean.utt_lt50_n}{0}
\defmmnum{xprofile.best_closed.whisper-small__Qwen2.5-1.5B-Instruct__to__gemma-2-2b-it__none.clean.utt_lt50_pct}{0.00}
\defmmnum{xprofile.best_closed.whisper-small__Qwen2.5-1.5B-Instruct__to__gemma-2-2b-it__none.clean.utt_median_wer}{100.00}
\defmmnum{xprofile.best_closed.whisper-small__Qwen2.5-1.5B-Instruct__to__gemma-2-2b-it__none.clean.wer}{111.81}
\defmmnum{xprofile.best_closed.whisper-small__Qwen2.5-1.5B-Instruct__to__gemma-2-2b-it__none.clean.word_ratio_corpus}{0.21}
\defmmnum{xprofile.best_closed.whisper-small__Qwen2.5-1.5B-Instruct__to__gemma-2-2b-it__none.clean.word_ratio_median}{0.00}
\defmmnum{xprofile.best_closed.whisper-small__Qwen2.5-1.5B-Instruct__to__gemma-2-2b-it__none.other.cap159_n}{588}
\defmmnum{xprofile.best_closed.whisper-small__Qwen2.5-1.5B-Instruct__to__gemma-2-2b-it__none.other.cap159_pct}{20.01}
\defmmnum{xprofile.best_closed.whisper-small__Qwen2.5-1.5B-Instruct__to__gemma-2-2b-it__none.other.cap160_n}{176}
\defmmnum{xprofile.best_closed.whisper-small__Qwen2.5-1.5B-Instruct__to__gemma-2-2b-it__none.other.cap160_pct}{5.99}
\defmmnum{xprofile.best_closed.whisper-small__Qwen2.5-1.5B-Instruct__to__gemma-2-2b-it__none.other.del_pct}{89.33}
\defmmnum{xprofile.best_closed.whisper-small__Qwen2.5-1.5B-Instruct__to__gemma-2-2b-it__none.other.del_words}{46{,}757}
\defmmnum{xprofile.best_closed.whisper-small__Qwen2.5-1.5B-Instruct__to__gemma-2-2b-it__none.other.empty_pct}{74.75}
\defmmnum{xprofile.best_closed.whisper-small__Qwen2.5-1.5B-Instruct__to__gemma-2-2b-it__none.other.ins_pct}{19.92}
\defmmnum{xprofile.best_closed.whisper-small__Qwen2.5-1.5B-Instruct__to__gemma-2-2b-it__none.other.ins_words}{10{,}428}
\defmmnum{xprofile.best_closed.whisper-small__Qwen2.5-1.5B-Instruct__to__gemma-2-2b-it__none.other.n_empty}{2{,}197}
\defmmnum{xprofile.best_closed.whisper-small__Qwen2.5-1.5B-Instruct__to__gemma-2-2b-it__none.other.n_utt}{2{,}939}
\defmmnum{xprofile.best_closed.whisper-small__Qwen2.5-1.5B-Instruct__to__gemma-2-2b-it__none.other.nonempty_n}{742}
\defmmnum{xprofile.best_closed.whisper-small__Qwen2.5-1.5B-Instruct__to__gemma-2-2b-it__none.other.nonempty_utt_median_wer}{100.00}
\defmmnum{xprofile.best_closed.whisper-small__Qwen2.5-1.5B-Instruct__to__gemma-2-2b-it__none.other.nonempty_wer}{187.03}
\defmmnum{xprofile.best_closed.whisper-small__Qwen2.5-1.5B-Instruct__to__gemma-2-2b-it__none.other.sub_pct}{10.06}
\defmmnum{xprofile.best_closed.whisper-small__Qwen2.5-1.5B-Instruct__to__gemma-2-2b-it__none.other.sub_words}{5{,}264}
\defmmnum{xprofile.best_closed.whisper-small__Qwen2.5-1.5B-Instruct__to__gemma-2-2b-it__none.other.utt_eq0_n}{0}
\defmmnum{xprofile.best_closed.whisper-small__Qwen2.5-1.5B-Instruct__to__gemma-2-2b-it__none.other.utt_eq0_pct}{0.00}
\defmmnum{xprofile.best_closed.whisper-small__Qwen2.5-1.5B-Instruct__to__gemma-2-2b-it__none.other.utt_ge100_n}{2{,}807}
\defmmnum{xprofile.best_closed.whisper-small__Qwen2.5-1.5B-Instruct__to__gemma-2-2b-it__none.other.utt_ge100_pct}{95.51}
\defmmnum{xprofile.best_closed.whisper-small__Qwen2.5-1.5B-Instruct__to__gemma-2-2b-it__none.other.utt_lt10_n}{0}
\defmmnum{xprofile.best_closed.whisper-small__Qwen2.5-1.5B-Instruct__to__gemma-2-2b-it__none.other.utt_lt10_pct}{0.00}
\defmmnum{xprofile.best_closed.whisper-small__Qwen2.5-1.5B-Instruct__to__gemma-2-2b-it__none.other.utt_lt50_n}{0}
\defmmnum{xprofile.best_closed.whisper-small__Qwen2.5-1.5B-Instruct__to__gemma-2-2b-it__none.other.utt_lt50_pct}{0.00}
\defmmnum{xprofile.best_closed.whisper-small__Qwen2.5-1.5B-Instruct__to__gemma-2-2b-it__none.other.utt_median_wer}{100.00}
\defmmnum{xprofile.best_closed.whisper-small__Qwen2.5-1.5B-Instruct__to__gemma-2-2b-it__none.other.wer}{119.31}
\defmmnum{xprofile.best_closed.whisper-small__Qwen2.5-1.5B-Instruct__to__gemma-2-2b-it__none.other.word_ratio_corpus}{0.31}
\defmmnum{xprofile.best_closed.whisper-small__Qwen2.5-1.5B-Instruct__to__gemma-2-2b-it__none.other.word_ratio_median}{0.00}
\defmmnum{xprofile.best_closed.whisper-small__Qwen2.5-1.5B-Instruct__to__gemma-2-2b-it__rmsnorm.clean.cap159_n}{1{,}126}
\defmmnum{xprofile.best_closed.whisper-small__Qwen2.5-1.5B-Instruct__to__gemma-2-2b-it__rmsnorm.clean.cap159_pct}{42.98}
\defmmnum{xprofile.best_closed.whisper-small__Qwen2.5-1.5B-Instruct__to__gemma-2-2b-it__rmsnorm.clean.cap160_n}{256}
\defmmnum{xprofile.best_closed.whisper-small__Qwen2.5-1.5B-Instruct__to__gemma-2-2b-it__rmsnorm.clean.cap160_pct}{9.77}
\defmmnum{xprofile.best_closed.whisper-small__Qwen2.5-1.5B-Instruct__to__gemma-2-2b-it__rmsnorm.clean.del_pct}{91.31}
\defmmnum{xprofile.best_closed.whisper-small__Qwen2.5-1.5B-Instruct__to__gemma-2-2b-it__rmsnorm.clean.del_words}{48{,}007}
\defmmnum{xprofile.best_closed.whisper-small__Qwen2.5-1.5B-Instruct__to__gemma-2-2b-it__rmsnorm.clean.empty_pct}{85.65}
\defmmnum{xprofile.best_closed.whisper-small__Qwen2.5-1.5B-Instruct__to__gemma-2-2b-it__rmsnorm.clean.ins_pct}{54.21}
\defmmnum{xprofile.best_closed.whisper-small__Qwen2.5-1.5B-Instruct__to__gemma-2-2b-it__rmsnorm.clean.ins_words}{28{,}504}
\defmmnum{xprofile.best_closed.whisper-small__Qwen2.5-1.5B-Instruct__to__gemma-2-2b-it__rmsnorm.clean.n_empty}{2{,}244}
\defmmnum{xprofile.best_closed.whisper-small__Qwen2.5-1.5B-Instruct__to__gemma-2-2b-it__rmsnorm.clean.n_utt}{2{,}620}
\defmmnum{xprofile.best_closed.whisper-small__Qwen2.5-1.5B-Instruct__to__gemma-2-2b-it__rmsnorm.clean.nonempty_n}{376}
\defmmnum{xprofile.best_closed.whisper-small__Qwen2.5-1.5B-Instruct__to__gemma-2-2b-it__rmsnorm.clean.nonempty_utt_median_wer}{730.71}
\defmmnum{xprofile.best_closed.whisper-small__Qwen2.5-1.5B-Instruct__to__gemma-2-2b-it__rmsnorm.clean.nonempty_wer}{611.75}
\defmmnum{xprofile.best_closed.whisper-small__Qwen2.5-1.5B-Instruct__to__gemma-2-2b-it__rmsnorm.clean.sub_pct}{7.91}
\defmmnum{xprofile.best_closed.whisper-small__Qwen2.5-1.5B-Instruct__to__gemma-2-2b-it__rmsnorm.clean.sub_words}{4{,}160}
\defmmnum{xprofile.best_closed.whisper-small__Qwen2.5-1.5B-Instruct__to__gemma-2-2b-it__rmsnorm.clean.utt_eq0_n}{0}
\defmmnum{xprofile.best_closed.whisper-small__Qwen2.5-1.5B-Instruct__to__gemma-2-2b-it__rmsnorm.clean.utt_eq0_pct}{0.00}
\defmmnum{xprofile.best_closed.whisper-small__Qwen2.5-1.5B-Instruct__to__gemma-2-2b-it__rmsnorm.clean.utt_ge100_n}{2{,}589}
\defmmnum{xprofile.best_closed.whisper-small__Qwen2.5-1.5B-Instruct__to__gemma-2-2b-it__rmsnorm.clean.utt_ge100_pct}{98.82}
\defmmnum{xprofile.best_closed.whisper-small__Qwen2.5-1.5B-Instruct__to__gemma-2-2b-it__rmsnorm.clean.utt_lt10_n}{0}
\defmmnum{xprofile.best_closed.whisper-small__Qwen2.5-1.5B-Instruct__to__gemma-2-2b-it__rmsnorm.clean.utt_lt10_pct}{0.00}
\defmmnum{xprofile.best_closed.whisper-small__Qwen2.5-1.5B-Instruct__to__gemma-2-2b-it__rmsnorm.clean.utt_lt50_n}{0}
\defmmnum{xprofile.best_closed.whisper-small__Qwen2.5-1.5B-Instruct__to__gemma-2-2b-it__rmsnorm.clean.utt_lt50_pct}{0.00}
\defmmnum{xprofile.best_closed.whisper-small__Qwen2.5-1.5B-Instruct__to__gemma-2-2b-it__rmsnorm.clean.utt_median_wer}{100.00}
\defmmnum{xprofile.best_closed.whisper-small__Qwen2.5-1.5B-Instruct__to__gemma-2-2b-it__rmsnorm.clean.wer}{153.44}
\defmmnum{xprofile.best_closed.whisper-small__Qwen2.5-1.5B-Instruct__to__gemma-2-2b-it__rmsnorm.clean.word_ratio_corpus}{0.63}
\defmmnum{xprofile.best_closed.whisper-small__Qwen2.5-1.5B-Instruct__to__gemma-2-2b-it__rmsnorm.clean.word_ratio_median}{0.00}
\defmmnum{xprofile.best_closed.whisper-small__Qwen2.5-1.5B-Instruct__to__gemma-2-2b-it__rmsnorm.other.cap159_n}{1{,}020}
\defmmnum{xprofile.best_closed.whisper-small__Qwen2.5-1.5B-Instruct__to__gemma-2-2b-it__rmsnorm.other.cap159_pct}{34.71}
\defmmnum{xprofile.best_closed.whisper-small__Qwen2.5-1.5B-Instruct__to__gemma-2-2b-it__rmsnorm.other.cap160_n}{274}
\defmmnum{xprofile.best_closed.whisper-small__Qwen2.5-1.5B-Instruct__to__gemma-2-2b-it__rmsnorm.other.cap160_pct}{9.32}
\defmmnum{xprofile.best_closed.whisper-small__Qwen2.5-1.5B-Instruct__to__gemma-2-2b-it__rmsnorm.other.del_pct}{92.21}
\defmmnum{xprofile.best_closed.whisper-small__Qwen2.5-1.5B-Instruct__to__gemma-2-2b-it__rmsnorm.other.del_words}{48{,}268}
\defmmnum{xprofile.best_closed.whisper-small__Qwen2.5-1.5B-Instruct__to__gemma-2-2b-it__rmsnorm.other.empty_pct}{87.31}
\defmmnum{xprofile.best_closed.whisper-small__Qwen2.5-1.5B-Instruct__to__gemma-2-2b-it__rmsnorm.other.ins_pct}{55.67}
\defmmnum{xprofile.best_closed.whisper-small__Qwen2.5-1.5B-Instruct__to__gemma-2-2b-it__rmsnorm.other.ins_words}{29{,}139}
\defmmnum{xprofile.best_closed.whisper-small__Qwen2.5-1.5B-Instruct__to__gemma-2-2b-it__rmsnorm.other.n_empty}{2{,}566}
\defmmnum{xprofile.best_closed.whisper-small__Qwen2.5-1.5B-Instruct__to__gemma-2-2b-it__rmsnorm.other.n_utt}{2{,}939}
\defmmnum{xprofile.best_closed.whisper-small__Qwen2.5-1.5B-Instruct__to__gemma-2-2b-it__rmsnorm.other.nonempty_n}{373}
\defmmnum{xprofile.best_closed.whisper-small__Qwen2.5-1.5B-Instruct__to__gemma-2-2b-it__rmsnorm.other.nonempty_utt_median_wer}{833.33}
\defmmnum{xprofile.best_closed.whisper-small__Qwen2.5-1.5B-Instruct__to__gemma-2-2b-it__rmsnorm.other.nonempty_wer}{730.08}
\defmmnum{xprofile.best_closed.whisper-small__Qwen2.5-1.5B-Instruct__to__gemma-2-2b-it__rmsnorm.other.sub_pct}{7.10}
\defmmnum{xprofile.best_closed.whisper-small__Qwen2.5-1.5B-Instruct__to__gemma-2-2b-it__rmsnorm.other.sub_words}{3{,}718}
\defmmnum{xprofile.best_closed.whisper-small__Qwen2.5-1.5B-Instruct__to__gemma-2-2b-it__rmsnorm.other.utt_eq0_n}{0}
\defmmnum{xprofile.best_closed.whisper-small__Qwen2.5-1.5B-Instruct__to__gemma-2-2b-it__rmsnorm.other.utt_eq0_pct}{0.00}
\defmmnum{xprofile.best_closed.whisper-small__Qwen2.5-1.5B-Instruct__to__gemma-2-2b-it__rmsnorm.other.utt_ge100_n}{2{,}920}
\defmmnum{xprofile.best_closed.whisper-small__Qwen2.5-1.5B-Instruct__to__gemma-2-2b-it__rmsnorm.other.utt_ge100_pct}{99.35}
\defmmnum{xprofile.best_closed.whisper-small__Qwen2.5-1.5B-Instruct__to__gemma-2-2b-it__rmsnorm.other.utt_lt10_n}{0}
\defmmnum{xprofile.best_closed.whisper-small__Qwen2.5-1.5B-Instruct__to__gemma-2-2b-it__rmsnorm.other.utt_lt10_pct}{0.00}
\defmmnum{xprofile.best_closed.whisper-small__Qwen2.5-1.5B-Instruct__to__gemma-2-2b-it__rmsnorm.other.utt_lt50_n}{0}
\defmmnum{xprofile.best_closed.whisper-small__Qwen2.5-1.5B-Instruct__to__gemma-2-2b-it__rmsnorm.other.utt_lt50_pct}{0.00}
\defmmnum{xprofile.best_closed.whisper-small__Qwen2.5-1.5B-Instruct__to__gemma-2-2b-it__rmsnorm.other.utt_median_wer}{100.00}
\defmmnum{xprofile.best_closed.whisper-small__Qwen2.5-1.5B-Instruct__to__gemma-2-2b-it__rmsnorm.other.wer}{154.99}
\defmmnum{xprofile.best_closed.whisper-small__Qwen2.5-1.5B-Instruct__to__gemma-2-2b-it__rmsnorm.other.word_ratio_corpus}{0.63}
\defmmnum{xprofile.best_closed.whisper-small__Qwen2.5-1.5B-Instruct__to__gemma-2-2b-it__rmsnorm.other.word_ratio_median}{0.00}
\defmmnum{xprofile.best_closed.whisper-small__Qwen2.5-1.5B-Instruct__to__gemma-3-4b-it__l2norm.clean.cap159_n}{1{,}357}
\defmmnum{xprofile.best_closed.whisper-small__Qwen2.5-1.5B-Instruct__to__gemma-3-4b-it__l2norm.clean.cap159_pct}{51.79}
\defmmnum{xprofile.best_closed.whisper-small__Qwen2.5-1.5B-Instruct__to__gemma-3-4b-it__l2norm.clean.cap160_n}{1{,}212}
\defmmnum{xprofile.best_closed.whisper-small__Qwen2.5-1.5B-Instruct__to__gemma-3-4b-it__l2norm.clean.cap160_pct}{46.26}
\defmmnum{xprofile.best_closed.whisper-small__Qwen2.5-1.5B-Instruct__to__gemma-3-4b-it__l2norm.clean.del_pct}{31.32}
\defmmnum{xprofile.best_closed.whisper-small__Qwen2.5-1.5B-Instruct__to__gemma-3-4b-it__l2norm.clean.del_words}{16{,}467}
\defmmnum{xprofile.best_closed.whisper-small__Qwen2.5-1.5B-Instruct__to__gemma-3-4b-it__l2norm.clean.empty_pct}{11.68}
\defmmnum{xprofile.best_closed.whisper-small__Qwen2.5-1.5B-Instruct__to__gemma-3-4b-it__l2norm.clean.ins_pct}{218.80}
\defmmnum{xprofile.best_closed.whisper-small__Qwen2.5-1.5B-Instruct__to__gemma-3-4b-it__l2norm.clean.ins_words}{115{,}037}
\defmmnum{xprofile.best_closed.whisper-small__Qwen2.5-1.5B-Instruct__to__gemma-3-4b-it__l2norm.clean.n_empty}{306}
\defmmnum{xprofile.best_closed.whisper-small__Qwen2.5-1.5B-Instruct__to__gemma-3-4b-it__l2norm.clean.n_utt}{2{,}620}
\defmmnum{xprofile.best_closed.whisper-small__Qwen2.5-1.5B-Instruct__to__gemma-3-4b-it__l2norm.clean.nonempty_n}{2{,}314}
\defmmnum{xprofile.best_closed.whisper-small__Qwen2.5-1.5B-Instruct__to__gemma-3-4b-it__l2norm.clean.nonempty_utt_median_wer}{381.65}
\defmmnum{xprofile.best_closed.whisper-small__Qwen2.5-1.5B-Instruct__to__gemma-3-4b-it__l2norm.clean.nonempty_wer}{354.89}
\defmmnum{xprofile.best_closed.whisper-small__Qwen2.5-1.5B-Instruct__to__gemma-3-4b-it__l2norm.clean.sub_pct}{60.97}
\defmmnum{xprofile.best_closed.whisper-small__Qwen2.5-1.5B-Instruct__to__gemma-3-4b-it__l2norm.clean.sub_words}{32{,}055}
\defmmnum{xprofile.best_closed.whisper-small__Qwen2.5-1.5B-Instruct__to__gemma-3-4b-it__l2norm.clean.utt_eq0_n}{0}
\defmmnum{xprofile.best_closed.whisper-small__Qwen2.5-1.5B-Instruct__to__gemma-3-4b-it__l2norm.clean.utt_eq0_pct}{0.00}
\defmmnum{xprofile.best_closed.whisper-small__Qwen2.5-1.5B-Instruct__to__gemma-3-4b-it__l2norm.clean.utt_ge100_n}{2{,}392}
\defmmnum{xprofile.best_closed.whisper-small__Qwen2.5-1.5B-Instruct__to__gemma-3-4b-it__l2norm.clean.utt_ge100_pct}{91.30}
\defmmnum{xprofile.best_closed.whisper-small__Qwen2.5-1.5B-Instruct__to__gemma-3-4b-it__l2norm.clean.utt_lt10_n}{0}
\defmmnum{xprofile.best_closed.whisper-small__Qwen2.5-1.5B-Instruct__to__gemma-3-4b-it__l2norm.clean.utt_lt10_pct}{0.00}
\defmmnum{xprofile.best_closed.whisper-small__Qwen2.5-1.5B-Instruct__to__gemma-3-4b-it__l2norm.clean.utt_lt50_n}{0}
\defmmnum{xprofile.best_closed.whisper-small__Qwen2.5-1.5B-Instruct__to__gemma-3-4b-it__l2norm.clean.utt_lt50_pct}{0.00}
\defmmnum{xprofile.best_closed.whisper-small__Qwen2.5-1.5B-Instruct__to__gemma-3-4b-it__l2norm.clean.utt_median_wer}{310.88}
\defmmnum{xprofile.best_closed.whisper-small__Qwen2.5-1.5B-Instruct__to__gemma-3-4b-it__l2norm.clean.wer}{311.09}
\defmmnum{xprofile.best_closed.whisper-small__Qwen2.5-1.5B-Instruct__to__gemma-3-4b-it__l2norm.clean.word_ratio_corpus}{2.87}
\defmmnum{xprofile.best_closed.whisper-small__Qwen2.5-1.5B-Instruct__to__gemma-3-4b-it__l2norm.clean.word_ratio_median}{3.21}
\defmmnum{xprofile.best_closed.whisper-small__Qwen2.5-1.5B-Instruct__to__gemma-3-4b-it__l2norm.other.cap159_n}{1{,}559}
\defmmnum{xprofile.best_closed.whisper-small__Qwen2.5-1.5B-Instruct__to__gemma-3-4b-it__l2norm.other.cap159_pct}{53.05}
\defmmnum{xprofile.best_closed.whisper-small__Qwen2.5-1.5B-Instruct__to__gemma-3-4b-it__l2norm.other.cap160_n}{1{,}394}
\defmmnum{xprofile.best_closed.whisper-small__Qwen2.5-1.5B-Instruct__to__gemma-3-4b-it__l2norm.other.cap160_pct}{47.43}
\defmmnum{xprofile.best_closed.whisper-small__Qwen2.5-1.5B-Instruct__to__gemma-3-4b-it__l2norm.other.del_pct}{25.74}
\defmmnum{xprofile.best_closed.whisper-small__Qwen2.5-1.5B-Instruct__to__gemma-3-4b-it__l2norm.other.del_words}{13{,}472}
\defmmnum{xprofile.best_closed.whisper-small__Qwen2.5-1.5B-Instruct__to__gemma-3-4b-it__l2norm.other.empty_pct}{8.47}
\defmmnum{xprofile.best_closed.whisper-small__Qwen2.5-1.5B-Instruct__to__gemma-3-4b-it__l2norm.other.ins_pct}{256.13}
\defmmnum{xprofile.best_closed.whisper-small__Qwen2.5-1.5B-Instruct__to__gemma-3-4b-it__l2norm.other.ins_words}{134{,}067}
\defmmnum{xprofile.best_closed.whisper-small__Qwen2.5-1.5B-Instruct__to__gemma-3-4b-it__l2norm.other.n_empty}{249}
\defmmnum{xprofile.best_closed.whisper-small__Qwen2.5-1.5B-Instruct__to__gemma-3-4b-it__l2norm.other.n_utt}{2{,}939}
\defmmnum{xprofile.best_closed.whisper-small__Qwen2.5-1.5B-Instruct__to__gemma-3-4b-it__l2norm.other.nonempty_n}{2{,}690}
\defmmnum{xprofile.best_closed.whisper-small__Qwen2.5-1.5B-Instruct__to__gemma-3-4b-it__l2norm.other.nonempty_utt_median_wer}{411.32}
\defmmnum{xprofile.best_closed.whisper-small__Qwen2.5-1.5B-Instruct__to__gemma-3-4b-it__l2norm.other.nonempty_wer}{380.59}
\defmmnum{xprofile.best_closed.whisper-small__Qwen2.5-1.5B-Instruct__to__gemma-3-4b-it__l2norm.other.sub_pct}{66.25}
\defmmnum{xprofile.best_closed.whisper-small__Qwen2.5-1.5B-Instruct__to__gemma-3-4b-it__l2norm.other.sub_words}{34{,}676}
\defmmnum{xprofile.best_closed.whisper-small__Qwen2.5-1.5B-Instruct__to__gemma-3-4b-it__l2norm.other.utt_eq0_n}{0}
\defmmnum{xprofile.best_closed.whisper-small__Qwen2.5-1.5B-Instruct__to__gemma-3-4b-it__l2norm.other.utt_eq0_pct}{0.00}
\defmmnum{xprofile.best_closed.whisper-small__Qwen2.5-1.5B-Instruct__to__gemma-3-4b-it__l2norm.other.utt_ge100_n}{2{,}692}
\defmmnum{xprofile.best_closed.whisper-small__Qwen2.5-1.5B-Instruct__to__gemma-3-4b-it__l2norm.other.utt_ge100_pct}{91.60}
\defmmnum{xprofile.best_closed.whisper-small__Qwen2.5-1.5B-Instruct__to__gemma-3-4b-it__l2norm.other.utt_lt10_n}{0}
\defmmnum{xprofile.best_closed.whisper-small__Qwen2.5-1.5B-Instruct__to__gemma-3-4b-it__l2norm.other.utt_lt10_pct}{0.00}
\defmmnum{xprofile.best_closed.whisper-small__Qwen2.5-1.5B-Instruct__to__gemma-3-4b-it__l2norm.other.utt_lt50_n}{0}
\defmmnum{xprofile.best_closed.whisper-small__Qwen2.5-1.5B-Instruct__to__gemma-3-4b-it__l2norm.other.utt_lt50_pct}{0.00}
\defmmnum{xprofile.best_closed.whisper-small__Qwen2.5-1.5B-Instruct__to__gemma-3-4b-it__l2norm.other.utt_median_wer}{355.56}
\defmmnum{xprofile.best_closed.whisper-small__Qwen2.5-1.5B-Instruct__to__gemma-3-4b-it__l2norm.other.wer}{348.12}
\defmmnum{xprofile.best_closed.whisper-small__Qwen2.5-1.5B-Instruct__to__gemma-3-4b-it__l2norm.other.word_ratio_corpus}{3.30}
\defmmnum{xprofile.best_closed.whisper-small__Qwen2.5-1.5B-Instruct__to__gemma-3-4b-it__l2norm.other.word_ratio_median}{3.67}
\defmmnum{xprofile.best_closed.whisper-small__gemma-2-2b-it__to__Qwen2.5-1.5B-Instruct__layernorm.clean.cap159_n}{0}
\defmmnum{xprofile.best_closed.whisper-small__gemma-2-2b-it__to__Qwen2.5-1.5B-Instruct__layernorm.clean.cap159_pct}{0.00}
\defmmnum{xprofile.best_closed.whisper-small__gemma-2-2b-it__to__Qwen2.5-1.5B-Instruct__layernorm.clean.cap160_n}{0}
\defmmnum{xprofile.best_closed.whisper-small__gemma-2-2b-it__to__Qwen2.5-1.5B-Instruct__layernorm.clean.cap160_pct}{0.00}
\defmmnum{xprofile.best_closed.whisper-small__gemma-2-2b-it__to__Qwen2.5-1.5B-Instruct__layernorm.clean.del_pct}{25.73}
\defmmnum{xprofile.best_closed.whisper-small__gemma-2-2b-it__to__Qwen2.5-1.5B-Instruct__layernorm.clean.del_words}{13{,}528}
\defmmnum{xprofile.best_closed.whisper-small__gemma-2-2b-it__to__Qwen2.5-1.5B-Instruct__layernorm.clean.empty_pct}{0.00}
\defmmnum{xprofile.best_closed.whisper-small__gemma-2-2b-it__to__Qwen2.5-1.5B-Instruct__layernorm.clean.ins_pct}{38.36}
\defmmnum{xprofile.best_closed.whisper-small__gemma-2-2b-it__to__Qwen2.5-1.5B-Instruct__layernorm.clean.ins_words}{20{,}169}
\defmmnum{xprofile.best_closed.whisper-small__gemma-2-2b-it__to__Qwen2.5-1.5B-Instruct__layernorm.clean.n_empty}{0}
\defmmnum{xprofile.best_closed.whisper-small__gemma-2-2b-it__to__Qwen2.5-1.5B-Instruct__layernorm.clean.n_utt}{2{,}620}
\defmmnum{xprofile.best_closed.whisper-small__gemma-2-2b-it__to__Qwen2.5-1.5B-Instruct__layernorm.clean.nonempty_n}{2{,}620}
\defmmnum{xprofile.best_closed.whisper-small__gemma-2-2b-it__to__Qwen2.5-1.5B-Instruct__layernorm.clean.nonempty_utt_median_wer}{120.00}
\defmmnum{xprofile.best_closed.whisper-small__gemma-2-2b-it__to__Qwen2.5-1.5B-Instruct__layernorm.clean.nonempty_wer}{134.84}
\defmmnum{xprofile.best_closed.whisper-small__gemma-2-2b-it__to__Qwen2.5-1.5B-Instruct__layernorm.clean.sub_pct}{70.75}
\defmmnum{xprofile.best_closed.whisper-small__gemma-2-2b-it__to__Qwen2.5-1.5B-Instruct__layernorm.clean.sub_words}{37{,}195}
\defmmnum{xprofile.best_closed.whisper-small__gemma-2-2b-it__to__Qwen2.5-1.5B-Instruct__layernorm.clean.utt_eq0_n}{0}
\defmmnum{xprofile.best_closed.whisper-small__gemma-2-2b-it__to__Qwen2.5-1.5B-Instruct__layernorm.clean.utt_eq0_pct}{0.00}
\defmmnum{xprofile.best_closed.whisper-small__gemma-2-2b-it__to__Qwen2.5-1.5B-Instruct__layernorm.clean.utt_ge100_n}{2{,}082}
\defmmnum{xprofile.best_closed.whisper-small__gemma-2-2b-it__to__Qwen2.5-1.5B-Instruct__layernorm.clean.utt_ge100_pct}{79.47}
\defmmnum{xprofile.best_closed.whisper-small__gemma-2-2b-it__to__Qwen2.5-1.5B-Instruct__layernorm.clean.utt_lt10_n}{0}
\defmmnum{xprofile.best_closed.whisper-small__gemma-2-2b-it__to__Qwen2.5-1.5B-Instruct__layernorm.clean.utt_lt10_pct}{0.00}
\defmmnum{xprofile.best_closed.whisper-small__gemma-2-2b-it__to__Qwen2.5-1.5B-Instruct__layernorm.clean.utt_lt50_n}{0}
\defmmnum{xprofile.best_closed.whisper-small__gemma-2-2b-it__to__Qwen2.5-1.5B-Instruct__layernorm.clean.utt_lt50_pct}{0.00}
\defmmnum{xprofile.best_closed.whisper-small__gemma-2-2b-it__to__Qwen2.5-1.5B-Instruct__layernorm.clean.utt_median_wer}{120.00}
\defmmnum{xprofile.best_closed.whisper-small__gemma-2-2b-it__to__Qwen2.5-1.5B-Instruct__layernorm.clean.wer}{134.84}
\defmmnum{xprofile.best_closed.whisper-small__gemma-2-2b-it__to__Qwen2.5-1.5B-Instruct__layernorm.clean.word_ratio_corpus}{1.13}
\defmmnum{xprofile.best_closed.whisper-small__gemma-2-2b-it__to__Qwen2.5-1.5B-Instruct__layernorm.clean.word_ratio_median}{1.22}
\defmmnum{xprofile.best_closed.whisper-small__gemma-2-2b-it__to__Qwen2.5-1.5B-Instruct__layernorm.other.cap159_n}{0}
\defmmnum{xprofile.best_closed.whisper-small__gemma-2-2b-it__to__Qwen2.5-1.5B-Instruct__layernorm.other.cap159_pct}{0.00}
\defmmnum{xprofile.best_closed.whisper-small__gemma-2-2b-it__to__Qwen2.5-1.5B-Instruct__layernorm.other.cap160_n}{0}
\defmmnum{xprofile.best_closed.whisper-small__gemma-2-2b-it__to__Qwen2.5-1.5B-Instruct__layernorm.other.cap160_pct}{0.00}
\defmmnum{xprofile.best_closed.whisper-small__gemma-2-2b-it__to__Qwen2.5-1.5B-Instruct__layernorm.other.del_pct}{28.88}
\defmmnum{xprofile.best_closed.whisper-small__gemma-2-2b-it__to__Qwen2.5-1.5B-Instruct__layernorm.other.del_words}{15{,}118}
\defmmnum{xprofile.best_closed.whisper-small__gemma-2-2b-it__to__Qwen2.5-1.5B-Instruct__layernorm.other.empty_pct}{0.00}
\defmmnum{xprofile.best_closed.whisper-small__gemma-2-2b-it__to__Qwen2.5-1.5B-Instruct__layernorm.other.ins_pct}{44.72}
\defmmnum{xprofile.best_closed.whisper-small__gemma-2-2b-it__to__Qwen2.5-1.5B-Instruct__layernorm.other.ins_words}{23{,}409}
\defmmnum{xprofile.best_closed.whisper-small__gemma-2-2b-it__to__Qwen2.5-1.5B-Instruct__layernorm.other.n_empty}{0}
\defmmnum{xprofile.best_closed.whisper-small__gemma-2-2b-it__to__Qwen2.5-1.5B-Instruct__layernorm.other.n_utt}{2{,}939}
\defmmnum{xprofile.best_closed.whisper-small__gemma-2-2b-it__to__Qwen2.5-1.5B-Instruct__layernorm.other.nonempty_n}{2{,}939}
\defmmnum{xprofile.best_closed.whisper-small__gemma-2-2b-it__to__Qwen2.5-1.5B-Instruct__layernorm.other.nonempty_utt_median_wer}{126.32}
\defmmnum{xprofile.best_closed.whisper-small__gemma-2-2b-it__to__Qwen2.5-1.5B-Instruct__layernorm.other.nonempty_wer}{141.08}
\defmmnum{xprofile.best_closed.whisper-small__gemma-2-2b-it__to__Qwen2.5-1.5B-Instruct__layernorm.other.sub_pct}{67.47}
\defmmnum{xprofile.best_closed.whisper-small__gemma-2-2b-it__to__Qwen2.5-1.5B-Instruct__layernorm.other.sub_words}{35{,}318}
\defmmnum{xprofile.best_closed.whisper-small__gemma-2-2b-it__to__Qwen2.5-1.5B-Instruct__layernorm.other.utt_eq0_n}{0}
\defmmnum{xprofile.best_closed.whisper-small__gemma-2-2b-it__to__Qwen2.5-1.5B-Instruct__layernorm.other.utt_eq0_pct}{0.00}
\defmmnum{xprofile.best_closed.whisper-small__gemma-2-2b-it__to__Qwen2.5-1.5B-Instruct__layernorm.other.utt_ge100_n}{2{,}385}
\defmmnum{xprofile.best_closed.whisper-small__gemma-2-2b-it__to__Qwen2.5-1.5B-Instruct__layernorm.other.utt_ge100_pct}{81.15}
\defmmnum{xprofile.best_closed.whisper-small__gemma-2-2b-it__to__Qwen2.5-1.5B-Instruct__layernorm.other.utt_lt10_n}{0}
\defmmnum{xprofile.best_closed.whisper-small__gemma-2-2b-it__to__Qwen2.5-1.5B-Instruct__layernorm.other.utt_lt10_pct}{0.00}
\defmmnum{xprofile.best_closed.whisper-small__gemma-2-2b-it__to__Qwen2.5-1.5B-Instruct__layernorm.other.utt_lt50_n}{0}
\defmmnum{xprofile.best_closed.whisper-small__gemma-2-2b-it__to__Qwen2.5-1.5B-Instruct__layernorm.other.utt_lt50_pct}{0.00}
\defmmnum{xprofile.best_closed.whisper-small__gemma-2-2b-it__to__Qwen2.5-1.5B-Instruct__layernorm.other.utt_median_wer}{126.32}
\defmmnum{xprofile.best_closed.whisper-small__gemma-2-2b-it__to__Qwen2.5-1.5B-Instruct__layernorm.other.wer}{141.08}
\defmmnum{xprofile.best_closed.whisper-small__gemma-2-2b-it__to__Qwen2.5-1.5B-Instruct__layernorm.other.word_ratio_corpus}{1.16}
\defmmnum{xprofile.best_closed.whisper-small__gemma-2-2b-it__to__Qwen2.5-1.5B-Instruct__layernorm.other.word_ratio_median}{1.32}
\defmmnum{xprofile.best_closed.whisper-small__gemma-2-2b-it__to__gemma-3-4b-it__l2norm.clean.cap159_n}{585}
\defmmnum{xprofile.best_closed.whisper-small__gemma-2-2b-it__to__gemma-3-4b-it__l2norm.clean.cap159_pct}{22.33}
\defmmnum{xprofile.best_closed.whisper-small__gemma-2-2b-it__to__gemma-3-4b-it__l2norm.clean.cap160_n}{451}
\defmmnum{xprofile.best_closed.whisper-small__gemma-2-2b-it__to__gemma-3-4b-it__l2norm.clean.cap160_pct}{17.21}
\defmmnum{xprofile.best_closed.whisper-small__gemma-2-2b-it__to__gemma-3-4b-it__l2norm.clean.del_pct}{56.61}
\defmmnum{xprofile.best_closed.whisper-small__gemma-2-2b-it__to__gemma-3-4b-it__l2norm.clean.del_words}{29{,}764}
\defmmnum{xprofile.best_closed.whisper-small__gemma-2-2b-it__to__gemma-3-4b-it__l2norm.clean.empty_pct}{11.72}
\defmmnum{xprofile.best_closed.whisper-small__gemma-2-2b-it__to__gemma-3-4b-it__l2norm.clean.ins_pct}{101.32}
\defmmnum{xprofile.best_closed.whisper-small__gemma-2-2b-it__to__gemma-3-4b-it__l2norm.clean.ins_words}{53{,}268}
\defmmnum{xprofile.best_closed.whisper-small__gemma-2-2b-it__to__gemma-3-4b-it__l2norm.clean.n_empty}{307}
\defmmnum{xprofile.best_closed.whisper-small__gemma-2-2b-it__to__gemma-3-4b-it__l2norm.clean.n_utt}{2{,}620}
\defmmnum{xprofile.best_closed.whisper-small__gemma-2-2b-it__to__gemma-3-4b-it__l2norm.clean.nonempty_n}{2{,}313}
\defmmnum{xprofile.best_closed.whisper-small__gemma-2-2b-it__to__gemma-3-4b-it__l2norm.clean.nonempty_utt_median_wer}{115.00}
\defmmnum{xprofile.best_closed.whisper-small__gemma-2-2b-it__to__gemma-3-4b-it__l2norm.clean.nonempty_wer}{223.19}
\defmmnum{xprofile.best_closed.whisper-small__gemma-2-2b-it__to__gemma-3-4b-it__l2norm.clean.sub_pct}{39.56}
\defmmnum{xprofile.best_closed.whisper-small__gemma-2-2b-it__to__gemma-3-4b-it__l2norm.clean.sub_words}{20{,}801}
\defmmnum{xprofile.best_closed.whisper-small__gemma-2-2b-it__to__gemma-3-4b-it__l2norm.clean.utt_eq0_n}{0}
\defmmnum{xprofile.best_closed.whisper-small__gemma-2-2b-it__to__gemma-3-4b-it__l2norm.clean.utt_eq0_pct}{0.00}
\defmmnum{xprofile.best_closed.whisper-small__gemma-2-2b-it__to__gemma-3-4b-it__l2norm.clean.utt_ge100_n}{2{,}160}
\defmmnum{xprofile.best_closed.whisper-small__gemma-2-2b-it__to__gemma-3-4b-it__l2norm.clean.utt_ge100_pct}{82.44}
\defmmnum{xprofile.best_closed.whisper-small__gemma-2-2b-it__to__gemma-3-4b-it__l2norm.clean.utt_lt10_n}{0}
\defmmnum{xprofile.best_closed.whisper-small__gemma-2-2b-it__to__gemma-3-4b-it__l2norm.clean.utt_lt10_pct}{0.00}
\defmmnum{xprofile.best_closed.whisper-small__gemma-2-2b-it__to__gemma-3-4b-it__l2norm.clean.utt_lt50_n}{0}
\defmmnum{xprofile.best_closed.whisper-small__gemma-2-2b-it__to__gemma-3-4b-it__l2norm.clean.utt_lt50_pct}{0.00}
\defmmnum{xprofile.best_closed.whisper-small__gemma-2-2b-it__to__gemma-3-4b-it__l2norm.clean.utt_median_wer}{100.00}
\defmmnum{xprofile.best_closed.whisper-small__gemma-2-2b-it__to__gemma-3-4b-it__l2norm.clean.wer}{197.49}
\defmmnum{xprofile.best_closed.whisper-small__gemma-2-2b-it__to__gemma-3-4b-it__l2norm.clean.word_ratio_corpus}{1.45}
\defmmnum{xprofile.best_closed.whisper-small__gemma-2-2b-it__to__gemma-3-4b-it__l2norm.clean.word_ratio_median}{0.83}
\defmmnum{xprofile.best_closed.whisper-small__gemma-2-2b-it__to__gemma-3-4b-it__l2norm.other.cap159_n}{710}
\defmmnum{xprofile.best_closed.whisper-small__gemma-2-2b-it__to__gemma-3-4b-it__l2norm.other.cap159_pct}{24.16}
\defmmnum{xprofile.best_closed.whisper-small__gemma-2-2b-it__to__gemma-3-4b-it__l2norm.other.cap160_n}{585}
\defmmnum{xprofile.best_closed.whisper-small__gemma-2-2b-it__to__gemma-3-4b-it__l2norm.other.cap160_pct}{19.90}
\defmmnum{xprofile.best_closed.whisper-small__gemma-2-2b-it__to__gemma-3-4b-it__l2norm.other.del_pct}{51.15}
\defmmnum{xprofile.best_closed.whisper-small__gemma-2-2b-it__to__gemma-3-4b-it__l2norm.other.del_words}{26{,}775}
\defmmnum{xprofile.best_closed.whisper-small__gemma-2-2b-it__to__gemma-3-4b-it__l2norm.other.empty_pct}{10.41}
\defmmnum{xprofile.best_closed.whisper-small__gemma-2-2b-it__to__gemma-3-4b-it__l2norm.other.ins_pct}{133.17}
\defmmnum{xprofile.best_closed.whisper-small__gemma-2-2b-it__to__gemma-3-4b-it__l2norm.other.ins_words}{69{,}704}
\defmmnum{xprofile.best_closed.whisper-small__gemma-2-2b-it__to__gemma-3-4b-it__l2norm.other.n_empty}{306}
\defmmnum{xprofile.best_closed.whisper-small__gemma-2-2b-it__to__gemma-3-4b-it__l2norm.other.n_utt}{2{,}939}
\defmmnum{xprofile.best_closed.whisper-small__gemma-2-2b-it__to__gemma-3-4b-it__l2norm.other.nonempty_n}{2{,}633}
\defmmnum{xprofile.best_closed.whisper-small__gemma-2-2b-it__to__gemma-3-4b-it__l2norm.other.nonempty_utt_median_wer}{161.54}
\defmmnum{xprofile.best_closed.whisper-small__gemma-2-2b-it__to__gemma-3-4b-it__l2norm.other.nonempty_wer}{259.71}
\defmmnum{xprofile.best_closed.whisper-small__gemma-2-2b-it__to__gemma-3-4b-it__l2norm.other.sub_pct}{44.80}
\defmmnum{xprofile.best_closed.whisper-small__gemma-2-2b-it__to__gemma-3-4b-it__l2norm.other.sub_words}{23{,}450}
\defmmnum{xprofile.best_closed.whisper-small__gemma-2-2b-it__to__gemma-3-4b-it__l2norm.other.utt_eq0_n}{0}
\defmmnum{xprofile.best_closed.whisper-small__gemma-2-2b-it__to__gemma-3-4b-it__l2norm.other.utt_eq0_pct}{0.00}
\defmmnum{xprofile.best_closed.whisper-small__gemma-2-2b-it__to__gemma-3-4b-it__l2norm.other.utt_ge100_n}{2{,}533}
\defmmnum{xprofile.best_closed.whisper-small__gemma-2-2b-it__to__gemma-3-4b-it__l2norm.other.utt_ge100_pct}{86.19}
\defmmnum{xprofile.best_closed.whisper-small__gemma-2-2b-it__to__gemma-3-4b-it__l2norm.other.utt_lt10_n}{0}
\defmmnum{xprofile.best_closed.whisper-small__gemma-2-2b-it__to__gemma-3-4b-it__l2norm.other.utt_lt10_pct}{0.00}
\defmmnum{xprofile.best_closed.whisper-small__gemma-2-2b-it__to__gemma-3-4b-it__l2norm.other.utt_lt50_n}{0}
\defmmnum{xprofile.best_closed.whisper-small__gemma-2-2b-it__to__gemma-3-4b-it__l2norm.other.utt_lt50_pct}{0.00}
\defmmnum{xprofile.best_closed.whisper-small__gemma-2-2b-it__to__gemma-3-4b-it__l2norm.other.utt_median_wer}{116.67}
\defmmnum{xprofile.best_closed.whisper-small__gemma-2-2b-it__to__gemma-3-4b-it__l2norm.other.wer}{229.12}
\defmmnum{xprofile.best_closed.whisper-small__gemma-2-2b-it__to__gemma-3-4b-it__l2norm.other.word_ratio_corpus}{1.82}
\defmmnum{xprofile.best_closed.whisper-small__gemma-2-2b-it__to__gemma-3-4b-it__l2norm.other.word_ratio_median}{1.21}
\defmmnum{xprofile.best_closed.whisper-small__gemma-2-2b-it__to__gemma-3-4b-it__none.clean.cap159_n}{2{,}062}
\defmmnum{xprofile.best_closed.whisper-small__gemma-2-2b-it__to__gemma-3-4b-it__none.clean.cap159_pct}{78.70}
\defmmnum{xprofile.best_closed.whisper-small__gemma-2-2b-it__to__gemma-3-4b-it__none.clean.cap160_n}{1{,}823}
\defmmnum{xprofile.best_closed.whisper-small__gemma-2-2b-it__to__gemma-3-4b-it__none.clean.cap160_pct}{69.58}
\defmmnum{xprofile.best_closed.whisper-small__gemma-2-2b-it__to__gemma-3-4b-it__none.clean.del_pct}{11.44}
\defmmnum{xprofile.best_closed.whisper-small__gemma-2-2b-it__to__gemma-3-4b-it__none.clean.del_words}{6{,}017}
\defmmnum{xprofile.best_closed.whisper-small__gemma-2-2b-it__to__gemma-3-4b-it__none.clean.empty_pct}{2.82}
\defmmnum{xprofile.best_closed.whisper-small__gemma-2-2b-it__to__gemma-3-4b-it__none.clean.ins_pct}{271.74}
\defmmnum{xprofile.best_closed.whisper-small__gemma-2-2b-it__to__gemma-3-4b-it__none.clean.ins_words}{142{,}871}
\defmmnum{xprofile.best_closed.whisper-small__gemma-2-2b-it__to__gemma-3-4b-it__none.clean.n_empty}{74}
\defmmnum{xprofile.best_closed.whisper-small__gemma-2-2b-it__to__gemma-3-4b-it__none.clean.n_utt}{2{,}620}
\defmmnum{xprofile.best_closed.whisper-small__gemma-2-2b-it__to__gemma-3-4b-it__none.clean.nonempty_n}{2{,}546}
\defmmnum{xprofile.best_closed.whisper-small__gemma-2-2b-it__to__gemma-3-4b-it__none.clean.nonempty_utt_median_wer}{437.50}
\defmmnum{xprofile.best_closed.whisper-small__gemma-2-2b-it__to__gemma-3-4b-it__none.clean.nonempty_wer}{381.97}
\defmmnum{xprofile.best_closed.whisper-small__gemma-2-2b-it__to__gemma-3-4b-it__none.clean.sub_pct}{77.69}
\defmmnum{xprofile.best_closed.whisper-small__gemma-2-2b-it__to__gemma-3-4b-it__none.clean.sub_words}{40{,}846}
\defmmnum{xprofile.best_closed.whisper-small__gemma-2-2b-it__to__gemma-3-4b-it__none.clean.utt_eq0_n}{0}
\defmmnum{xprofile.best_closed.whisper-small__gemma-2-2b-it__to__gemma-3-4b-it__none.clean.utt_eq0_pct}{0.00}
\defmmnum{xprofile.best_closed.whisper-small__gemma-2-2b-it__to__gemma-3-4b-it__none.clean.utt_ge100_n}{2{,}518}
\defmmnum{xprofile.best_closed.whisper-small__gemma-2-2b-it__to__gemma-3-4b-it__none.clean.utt_ge100_pct}{96.11}
\defmmnum{xprofile.best_closed.whisper-small__gemma-2-2b-it__to__gemma-3-4b-it__none.clean.utt_lt10_n}{0}
\defmmnum{xprofile.best_closed.whisper-small__gemma-2-2b-it__to__gemma-3-4b-it__none.clean.utt_lt10_pct}{0.00}
\defmmnum{xprofile.best_closed.whisper-small__gemma-2-2b-it__to__gemma-3-4b-it__none.clean.utt_lt50_n}{0}
\defmmnum{xprofile.best_closed.whisper-small__gemma-2-2b-it__to__gemma-3-4b-it__none.clean.utt_lt50_pct}{0.00}
\defmmnum{xprofile.best_closed.whisper-small__gemma-2-2b-it__to__gemma-3-4b-it__none.clean.utt_median_wer}{426.79}
\defmmnum{xprofile.best_closed.whisper-small__gemma-2-2b-it__to__gemma-3-4b-it__none.clean.wer}{360.88}
\defmmnum{xprofile.best_closed.whisper-small__gemma-2-2b-it__to__gemma-3-4b-it__none.clean.word_ratio_corpus}{3.60}
\defmmnum{xprofile.best_closed.whisper-small__gemma-2-2b-it__to__gemma-3-4b-it__none.clean.word_ratio_median}{4.41}
\defmmnum{xprofile.best_closed.whisper-small__gemma-2-2b-it__to__gemma-3-4b-it__none.other.cap159_n}{2{,}270}
\defmmnum{xprofile.best_closed.whisper-small__gemma-2-2b-it__to__gemma-3-4b-it__none.other.cap159_pct}{77.24}
\defmmnum{xprofile.best_closed.whisper-small__gemma-2-2b-it__to__gemma-3-4b-it__none.other.cap160_n}{2{,}042}
\defmmnum{xprofile.best_closed.whisper-small__gemma-2-2b-it__to__gemma-3-4b-it__none.other.cap160_pct}{69.48}
\defmmnum{xprofile.best_closed.whisper-small__gemma-2-2b-it__to__gemma-3-4b-it__none.other.del_pct}{10.36}
\defmmnum{xprofile.best_closed.whisper-small__gemma-2-2b-it__to__gemma-3-4b-it__none.other.del_words}{5{,}424}
\defmmnum{xprofile.best_closed.whisper-small__gemma-2-2b-it__to__gemma-3-4b-it__none.other.empty_pct}{2.08}
\defmmnum{xprofile.best_closed.whisper-small__gemma-2-2b-it__to__gemma-3-4b-it__none.other.ins_pct}{310.68}
\defmmnum{xprofile.best_closed.whisper-small__gemma-2-2b-it__to__gemma-3-4b-it__none.other.ins_words}{162{,}618}
\defmmnum{xprofile.best_closed.whisper-small__gemma-2-2b-it__to__gemma-3-4b-it__none.other.n_empty}{61}
\defmmnum{xprofile.best_closed.whisper-small__gemma-2-2b-it__to__gemma-3-4b-it__none.other.n_utt}{2{,}939}
\defmmnum{xprofile.best_closed.whisper-small__gemma-2-2b-it__to__gemma-3-4b-it__none.other.nonempty_n}{2{,}878}
\defmmnum{xprofile.best_closed.whisper-small__gemma-2-2b-it__to__gemma-3-4b-it__none.other.nonempty_utt_median_wer}{478.76}
\defmmnum{xprofile.best_closed.whisper-small__gemma-2-2b-it__to__gemma-3-4b-it__none.other.nonempty_wer}{418.61}
\defmmnum{xprofile.best_closed.whisper-small__gemma-2-2b-it__to__gemma-3-4b-it__none.other.sub_pct}{79.12}
\defmmnum{xprofile.best_closed.whisper-small__gemma-2-2b-it__to__gemma-3-4b-it__none.other.sub_words}{41{,}416}
\defmmnum{xprofile.best_closed.whisper-small__gemma-2-2b-it__to__gemma-3-4b-it__none.other.utt_eq0_n}{0}
\defmmnum{xprofile.best_closed.whisper-small__gemma-2-2b-it__to__gemma-3-4b-it__none.other.utt_eq0_pct}{0.00}
\defmmnum{xprofile.best_closed.whisper-small__gemma-2-2b-it__to__gemma-3-4b-it__none.other.utt_ge100_n}{2{,}833}
\defmmnum{xprofile.best_closed.whisper-small__gemma-2-2b-it__to__gemma-3-4b-it__none.other.utt_ge100_pct}{96.39}
\defmmnum{xprofile.best_closed.whisper-small__gemma-2-2b-it__to__gemma-3-4b-it__none.other.utt_lt10_n}{0}
\defmmnum{xprofile.best_closed.whisper-small__gemma-2-2b-it__to__gemma-3-4b-it__none.other.utt_lt10_pct}{0.00}
\defmmnum{xprofile.best_closed.whisper-small__gemma-2-2b-it__to__gemma-3-4b-it__none.other.utt_lt50_n}{0}
\defmmnum{xprofile.best_closed.whisper-small__gemma-2-2b-it__to__gemma-3-4b-it__none.other.utt_lt50_pct}{0.00}
\defmmnum{xprofile.best_closed.whisper-small__gemma-2-2b-it__to__gemma-3-4b-it__none.other.utt_median_wer}{470.00}
\defmmnum{xprofile.best_closed.whisper-small__gemma-2-2b-it__to__gemma-3-4b-it__none.other.wer}{400.16}
\defmmnum{xprofile.best_closed.whisper-small__gemma-2-2b-it__to__gemma-3-4b-it__none.other.word_ratio_corpus}{4.00}
\defmmnum{xprofile.best_closed.whisper-small__gemma-2-2b-it__to__gemma-3-4b-it__none.other.word_ratio_median}{4.84}
\defmmnum{xprofile.cells_with_predictions}{154}
\defmmnum{xprofile.cross_all.audio_only_cells}{8}
\defmmnum{xprofile.cross_all.fail_cells}{14}
\defmmnum{xprofile.cross_all.fail_max_med_clean}{583.33}
\defmmnum{xprofile.cross_all.fail_max_wer_clean}{480.30}
\defmmnum{xprofile.cross_all.fail_min_med_clean}{98.00}
\defmmnum{xprofile.cross_all.fail_min_wer_clean}{98.63}
\defmmnum{xprofile.cross_all.max_clean}{480.30}
\defmmnum{xprofile.cross_all.max_other}{536.30}
\defmmnum{xprofile.cross_all.min_clean}{67.35}
\defmmnum{xprofile.cross_all.min_clean_with_partial}{67.35}
\defmmnum{xprofile.cross_all.min_other}{77.47}
\defmmnum{xprofile.cross_all.mixed_cells}{8}
\defmmnum{xprofile.cross_all.partial_cells}{2}
\defmmnum{xprofile.cross_all.partial_max_med_clean}{23.81}
\defmmnum{xprofile.cross_all.partial_max_wer_clean}{75.35}
\defmmnum{xprofile.cross_all.partial_min_med_clean}{20.83}
\defmmnum{xprofile.cross_all.partial_min_wer_clean}{67.35}
\defmmnum{xprofile.cross_all.total_cells}{16}
\defmmnum{xprofile.g2_to_g3_same.max_med_clean}{0.00}
\defmmnum{xprofile.group.ws4_libri.max_cap160_clean}{91.95}
\defmmnum{xprofile.group.ws4_libri.max_cap160_other}{90.95}
\defmmnum{xprofile.group.ws4_libri.max_lt50_clean}{0.00}
\defmmnum{xprofile.group.ws4_libri.max_lt50_other}{0.00}
\defmmnum{xprofile.group.ws4_libri.max_med_clean}{617.65}
\defmmnum{xprofile.group.ws4_libri.max_med_other}{700.00}
\defmmnum{xprofile.group.ws4_libri.min_lt50_clean}{0.00}
\defmmnum{xprofile.group.ws4_libri.min_lt50_other}{0.00}
\defmmnum{xprofile.group.ws4_libri.min_med_clean}{100.00}
\defmmnum{xprofile.group.ws4_libri.min_med_other}{100.00}
\defmmnum{xprofile.group.ws4_libri.n_cells}{16}
\defmmnum{xprofile.group.ws4b_cross_g3.max_cap160_clean}{82.21}
\defmmnum{xprofile.group.ws4b_cross_g3.max_cap160_other}{79.35}
\defmmnum{xprofile.group.ws4b_cross_g3.max_lt50_clean}{75.15}
\defmmnum{xprofile.group.ws4b_cross_g3.max_lt50_other}{60.33}
\defmmnum{xprofile.group.ws4b_cross_g3.max_med_clean}{555.78}
\defmmnum{xprofile.group.ws4b_cross_g3.max_med_other}{611.11}
\defmmnum{xprofile.group.ws4b_cross_g3.min_lt50_clean}{0.00}
\defmmnum{xprofile.group.ws4b_cross_g3.min_lt50_other}{0.00}
\defmmnum{xprofile.group.ws4b_cross_g3.min_med_clean}{20.83}
\defmmnum{xprofile.group.ws4b_cross_g3.min_med_other}{36.36}
\defmmnum{xprofile.group.ws4b_cross_g3.n_cells}{4}
\defmmnum{xprofile.group.ws4b_cross_qwen.max_cap160_clean}{0.11}
\defmmnum{xprofile.group.ws4b_cross_qwen.max_cap160_other}{0.07}
\defmmnum{xprofile.group.ws4b_cross_qwen.max_lt50_clean}{13.82}
\defmmnum{xprofile.group.ws4b_cross_qwen.max_lt50_other}{8.30}
\defmmnum{xprofile.group.ws4b_cross_qwen.max_med_clean}{100.00}
\defmmnum{xprofile.group.ws4b_cross_qwen.max_med_other}{100.00}
\defmmnum{xprofile.group.ws4b_cross_qwen.min_lt50_clean}{0.57}
\defmmnum{xprofile.group.ws4b_cross_qwen.min_lt50_other}{0.10}
\defmmnum{xprofile.group.ws4b_cross_qwen.min_med_clean}{98.00}
\defmmnum{xprofile.group.ws4b_cross_qwen.min_med_other}{100.00}
\defmmnum{xprofile.group.ws4b_cross_qwen.n_cells}{2}
\defmmnum{xprofile.group.ws4b_noise.max_cap160_clean}{1.87}
\defmmnum{xprofile.group.ws4b_noise.max_cap160_other}{1.33}
\defmmnum{xprofile.group.ws4b_noise.max_lt50_clean}{0.00}
\defmmnum{xprofile.group.ws4b_noise.max_lt50_other}{0.00}
\defmmnum{xprofile.group.ws4b_noise.max_med_clean}{100.00}
\defmmnum{xprofile.group.ws4b_noise.max_med_other}{100.00}
\defmmnum{xprofile.group.ws4b_noise.min_lt50_clean}{0.00}
\defmmnum{xprofile.group.ws4b_noise.min_lt50_other}{0.00}
\defmmnum{xprofile.group.ws4b_noise.min_med_clean}{100.00}
\defmmnum{xprofile.group.ws4b_noise.min_med_other}{100.00}
\defmmnum{xprofile.group.ws4b_noise.n_cells}{3}
\defmmnum{xprofile.group.ws4b_same.max_cap160_clean}{0.00}
\defmmnum{xprofile.group.ws4b_same.max_cap160_other}{0.03}
\defmmnum{xprofile.group.ws4b_same.max_lt50_clean}{99.31}
\defmmnum{xprofile.group.ws4b_same.max_lt50_other}{96.90}
\defmmnum{xprofile.group.ws4b_same.max_med_clean}{6.90}
\defmmnum{xprofile.group.ws4b_same.max_med_other}{16.67}
\defmmnum{xprofile.group.ws4b_same.min_lt50_clean}{97.90}
\defmmnum{xprofile.group.ws4b_same.min_lt50_other}{89.93}
\defmmnum{xprofile.group.ws4b_same.min_med_clean}{0.00}
\defmmnum{xprofile.group.ws4b_same.min_med_other}{3.85}
\defmmnum{xprofile.group.ws4b_same.n_cells}{9}
\defmmnum{xprofile.group.ws4c_mix_cross.max_cap160_clean}{85.61}
\defmmnum{xprofile.group.ws4c_mix_cross.max_cap160_other}{83.43}
\defmmnum{xprofile.group.ws4c_mix_cross.max_lt50_clean}{10.08}
\defmmnum{xprofile.group.ws4c_mix_cross.max_lt50_other}{3.44}
\defmmnum{xprofile.group.ws4c_mix_cross.max_med_clean}{583.33}
\defmmnum{xprofile.group.ws4c_mix_cross.max_med_other}{666.67}
\defmmnum{xprofile.group.ws4c_mix_cross.min_lt50_clean}{0.00}
\defmmnum{xprofile.group.ws4c_mix_cross.min_lt50_other}{0.00}
\defmmnum{xprofile.group.ws4c_mix_cross.min_med_clean}{100.00}
\defmmnum{xprofile.group.ws4c_mix_cross.min_med_other}{100.00}
\defmmnum{xprofile.group.ws4c_mix_cross.n_cells}{8}
\defmmnum{xprofile.group.ws4c_mix_same.max_cap160_clean}{0.04}
\defmmnum{xprofile.group.ws4c_mix_same.max_cap160_other}{0.07}
\defmmnum{xprofile.group.ws4c_mix_same.max_lt50_clean}{99.27}
\defmmnum{xprofile.group.ws4c_mix_same.max_lt50_other}{97.04}
\defmmnum{xprofile.group.ws4c_mix_same.max_med_clean}{5.56}
\defmmnum{xprofile.group.ws4c_mix_same.max_med_other}{15.38}
\defmmnum{xprofile.group.ws4c_mix_same.min_lt50_clean}{98.32}
\defmmnum{xprofile.group.ws4c_mix_same.min_lt50_other}{91.80}
\defmmnum{xprofile.group.ws4c_mix_same.min_med_clean}{0.00}
\defmmnum{xprofile.group.ws4c_mix_same.min_med_other}{4.00}
\defmmnum{xprofile.group.ws4c_mix_same.n_cells}{8}
\defmmnum{xprofile.group.ws4c_wiki.max_cap160_clean}{87.90}
\defmmnum{xprofile.group.ws4c_wiki.max_cap160_other}{89.38}
\defmmnum{xprofile.group.ws4c_wiki.max_lt50_clean}{0.00}
\defmmnum{xprofile.group.ws4c_wiki.max_lt50_other}{0.00}
\defmmnum{xprofile.group.ws4c_wiki.max_med_clean}{625.00}
\defmmnum{xprofile.group.ws4c_wiki.max_med_other}{725.00}
\defmmnum{xprofile.group.ws4c_wiki.min_lt50_clean}{0.00}
\defmmnum{xprofile.group.ws4c_wiki.min_lt50_other}{0.00}
\defmmnum{xprofile.group.ws4c_wiki.min_med_clean}{100.00}
\defmmnum{xprofile.group.ws4c_wiki.min_med_other}{100.00}
\defmmnum{xprofile.group.ws4c_wiki.n_cells}{8}
\defmmnum{xprofile.m1_cross.clean.cap159_n}{2}
\defmmnum{xprofile.m1_cross.clean.cap159_pct}{0.08}
\defmmnum{xprofile.m1_cross.clean.cap160_n}{2}
\defmmnum{xprofile.m1_cross.clean.cap160_pct}{0.08}
\defmmnum{xprofile.m1_cross.clean.del_pct}{60.08}
\defmmnum{xprofile.m1_cross.clean.del_words}{31{,}588}
\defmmnum{xprofile.m1_cross.clean.empty_pct}{0.00}
\defmmnum{xprofile.m1_cross.clean.ins_pct}{8.68}
\defmmnum{xprofile.m1_cross.clean.ins_words}{4{,}565}
\defmmnum{xprofile.m1_cross.clean.n_empty}{0}
\defmmnum{xprofile.m1_cross.clean.n_utt}{2{,}620}
\defmmnum{xprofile.m1_cross.clean.nonempty_n}{2{,}620}
\defmmnum{xprofile.m1_cross.clean.nonempty_utt_median_wer}{100.00}
\defmmnum{xprofile.m1_cross.clean.nonempty_wer}{105.96}
\defmmnum{xprofile.m1_cross.clean.sub_pct}{37.19}
\defmmnum{xprofile.m1_cross.clean.sub_words}{19{,}554}
\defmmnum{xprofile.m1_cross.clean.utt_eq0_n}{0}
\defmmnum{xprofile.m1_cross.clean.utt_eq0_pct}{0.00}
\defmmnum{xprofile.m1_cross.clean.utt_ge100_n}{1{,}773}
\defmmnum{xprofile.m1_cross.clean.utt_ge100_pct}{67.67}
\defmmnum{xprofile.m1_cross.clean.utt_lt10_n}{0}
\defmmnum{xprofile.m1_cross.clean.utt_lt10_pct}{0.00}
\defmmnum{xprofile.m1_cross.clean.utt_lt50_n}{9}
\defmmnum{xprofile.m1_cross.clean.utt_lt50_pct}{0.34}
\defmmnum{xprofile.m1_cross.clean.utt_median_wer}{100.00}
\defmmnum{xprofile.m1_cross.clean.wer}{105.96}
\defmmnum{xprofile.m1_cross.clean.word_ratio_corpus}{0.49}
\defmmnum{xprofile.m1_cross.clean.word_ratio_median}{0.50}
\defmmnum{xprofile.m1_cross.other.cap159_n}{1}
\defmmnum{xprofile.m1_cross.other.cap159_pct}{0.03}
\defmmnum{xprofile.m1_cross.other.cap160_n}{1}
\defmmnum{xprofile.m1_cross.other.cap160_pct}{0.03}
\defmmnum{xprofile.m1_cross.other.del_pct}{54.40}
\defmmnum{xprofile.m1_cross.other.del_words}{28{,}473}
\defmmnum{xprofile.m1_cross.other.empty_pct}{0.00}
\defmmnum{xprofile.m1_cross.other.ins_pct}{12.05}
\defmmnum{xprofile.m1_cross.other.ins_words}{6{,}306}
\defmmnum{xprofile.m1_cross.other.n_empty}{0}
\defmmnum{xprofile.m1_cross.other.n_utt}{2{,}939}
\defmmnum{xprofile.m1_cross.other.nonempty_n}{2{,}939}
\defmmnum{xprofile.m1_cross.other.nonempty_utt_median_wer}{100.00}
\defmmnum{xprofile.m1_cross.other.nonempty_wer}{109.21}
\defmmnum{xprofile.m1_cross.other.sub_pct}{42.77}
\defmmnum{xprofile.m1_cross.other.sub_words}{22{,}387}
\defmmnum{xprofile.m1_cross.other.utt_eq0_n}{1}
\defmmnum{xprofile.m1_cross.other.utt_eq0_pct}{0.03}
\defmmnum{xprofile.m1_cross.other.utt_ge100_n}{2{,}112}
\defmmnum{xprofile.m1_cross.other.utt_ge100_pct}{71.86}
\defmmnum{xprofile.m1_cross.other.utt_lt10_n}{1}
\defmmnum{xprofile.m1_cross.other.utt_lt10_pct}{0.03}
\defmmnum{xprofile.m1_cross.other.utt_lt50_n}{8}
\defmmnum{xprofile.m1_cross.other.utt_lt50_pct}{0.27}
\defmmnum{xprofile.m1_cross.other.utt_median_wer}{100.00}
\defmmnum{xprofile.m1_cross.other.wer}{109.21}
\defmmnum{xprofile.m1_cross.other.word_ratio_corpus}{0.58}
\defmmnum{xprofile.m1_cross.other.word_ratio_median}{0.62}
\defmmnum{xprofile.m1_same.clean.cap159_n}{0}
\defmmnum{xprofile.m1_same.clean.cap159_pct}{0.00}
\defmmnum{xprofile.m1_same.clean.cap160_n}{0}
\defmmnum{xprofile.m1_same.clean.cap160_pct}{0.00}
\defmmnum{xprofile.m1_same.clean.del_pct}{1.02}
\defmmnum{xprofile.m1_same.clean.del_words}{537}
\defmmnum{xprofile.m1_same.clean.empty_pct}{0.00}
\defmmnum{xprofile.m1_same.clean.ins_pct}{0.83}
\defmmnum{xprofile.m1_same.clean.ins_words}{437}
\defmmnum{xprofile.m1_same.clean.n_empty}{0}
\defmmnum{xprofile.m1_same.clean.n_utt}{2{,}620}
\defmmnum{xprofile.m1_same.clean.nonempty_n}{2{,}620}
\defmmnum{xprofile.m1_same.clean.nonempty_utt_median_wer}{5.56}
\defmmnum{xprofile.m1_same.clean.nonempty_wer}{8.03}
\defmmnum{xprofile.m1_same.clean.sub_pct}{6.18}
\defmmnum{xprofile.m1_same.clean.sub_words}{3{,}247}
\defmmnum{xprofile.m1_same.clean.utt_eq0_n}{940}
\defmmnum{xprofile.m1_same.clean.utt_eq0_pct}{35.88}
\defmmnum{xprofile.m1_same.clean.utt_ge100_n}{4}
\defmmnum{xprofile.m1_same.clean.utt_ge100_pct}{0.15}
\defmmnum{xprofile.m1_same.clean.utt_lt10_n}{1{,}693}
\defmmnum{xprofile.m1_same.clean.utt_lt10_pct}{64.62}
\defmmnum{xprofile.m1_same.clean.utt_lt50_n}{2{,}576}
\defmmnum{xprofile.m1_same.clean.utt_lt50_pct}{98.32}
\defmmnum{xprofile.m1_same.clean.utt_median_wer}{5.56}
\defmmnum{xprofile.m1_same.clean.wer}{8.03}
\defmmnum{xprofile.m1_same.clean.word_ratio_corpus}{1.00}
\defmmnum{xprofile.m1_same.clean.word_ratio_median}{1.00}
\defmmnum{xprofile.m1_same.other.cap159_n}{0}
\defmmnum{xprofile.m1_same.other.cap159_pct}{0.00}
\defmmnum{xprofile.m1_same.other.cap160_n}{0}
\defmmnum{xprofile.m1_same.other.cap160_pct}{0.00}
\defmmnum{xprofile.m1_same.other.del_pct}{2.39}
\defmmnum{xprofile.m1_same.other.del_words}{1{,}249}
\defmmnum{xprofile.m1_same.other.empty_pct}{0.00}
\defmmnum{xprofile.m1_same.other.ins_pct}{2.01}
\defmmnum{xprofile.m1_same.other.ins_words}{1{,}052}
\defmmnum{xprofile.m1_same.other.n_empty}{0}
\defmmnum{xprofile.m1_same.other.n_utt}{2{,}939}
\defmmnum{xprofile.m1_same.other.nonempty_n}{2{,}939}
\defmmnum{xprofile.m1_same.other.nonempty_utt_median_wer}{15.38}
\defmmnum{xprofile.m1_same.other.nonempty_wer}{17.45}
\defmmnum{xprofile.m1_same.other.sub_pct}{13.06}
\defmmnum{xprofile.m1_same.other.sub_words}{6{,}835}
\defmmnum{xprofile.m1_same.other.utt_eq0_n}{529}
\defmmnum{xprofile.m1_same.other.utt_eq0_pct}{18.00}
\defmmnum{xprofile.m1_same.other.utt_ge100_n}{23}
\defmmnum{xprofile.m1_same.other.utt_ge100_pct}{0.78}
\defmmnum{xprofile.m1_same.other.utt_lt10_n}{1{,}012}
\defmmnum{xprofile.m1_same.other.utt_lt10_pct}{34.43}
\defmmnum{xprofile.m1_same.other.utt_lt50_n}{2{,}698}
\defmmnum{xprofile.m1_same.other.utt_lt50_pct}{91.80}
\defmmnum{xprofile.m1_same.other.utt_median_wer}{15.38}
\defmmnum{xprofile.m1_same.other.wer}{17.45}
\defmmnum{xprofile.m1_same.other.word_ratio_corpus}{1.00}
\defmmnum{xprofile.m1_same.other.word_ratio_median}{1.00}
\defmmnum{xprofile.m2_cross.clean.cap159_n}{15}
\defmmnum{xprofile.m2_cross.clean.cap159_pct}{0.57}
\defmmnum{xprofile.m2_cross.clean.cap160_n}{15}
\defmmnum{xprofile.m2_cross.clean.cap160_pct}{0.57}
\defmmnum{xprofile.m2_cross.clean.del_pct}{51.47}
\defmmnum{xprofile.m2_cross.clean.del_words}{27{,}062}
\defmmnum{xprofile.m2_cross.clean.empty_pct}{0.00}
\defmmnum{xprofile.m2_cross.clean.ins_pct}{30.75}
\defmmnum{xprofile.m2_cross.clean.ins_words}{16{,}168}
\defmmnum{xprofile.m2_cross.clean.n_empty}{0}
\defmmnum{xprofile.m2_cross.clean.n_utt}{2{,}620}
\defmmnum{xprofile.m2_cross.clean.nonempty_n}{2{,}620}
\defmmnum{xprofile.m2_cross.clean.nonempty_utt_median_wer}{100.00}
\defmmnum{xprofile.m2_cross.clean.nonempty_wer}{123.29}
\defmmnum{xprofile.m2_cross.clean.sub_pct}{41.06}
\defmmnum{xprofile.m2_cross.clean.sub_words}{21{,}589}
\defmmnum{xprofile.m2_cross.clean.utt_eq0_n}{2}
\defmmnum{xprofile.m2_cross.clean.utt_eq0_pct}{0.08}
\defmmnum{xprofile.m2_cross.clean.utt_ge100_n}{1{,}592}
\defmmnum{xprofile.m2_cross.clean.utt_ge100_pct}{60.76}
\defmmnum{xprofile.m2_cross.clean.utt_lt10_n}{2}
\defmmnum{xprofile.m2_cross.clean.utt_lt10_pct}{0.08}
\defmmnum{xprofile.m2_cross.clean.utt_lt50_n}{57}
\defmmnum{xprofile.m2_cross.clean.utt_lt50_pct}{2.18}
\defmmnum{xprofile.m2_cross.clean.utt_median_wer}{100.00}
\defmmnum{xprofile.m2_cross.clean.wer}{123.29}
\defmmnum{xprofile.m2_cross.clean.word_ratio_corpus}{0.79}
\defmmnum{xprofile.m2_cross.clean.word_ratio_median}{0.67}
\defmmnum{xprofile.m2_cross.other.cap159_n}{20}
\defmmnum{xprofile.m2_cross.other.cap159_pct}{0.68}
\defmmnum{xprofile.m2_cross.other.cap160_n}{18}
\defmmnum{xprofile.m2_cross.other.cap160_pct}{0.61}
\defmmnum{xprofile.m2_cross.other.del_pct}{45.95}
\defmmnum{xprofile.m2_cross.other.del_words}{24{,}054}
\defmmnum{xprofile.m2_cross.other.empty_pct}{0.00}
\defmmnum{xprofile.m2_cross.other.ins_pct}{41.93}
\defmmnum{xprofile.m2_cross.other.ins_words}{21{,}945}
\defmmnum{xprofile.m2_cross.other.n_empty}{0}
\defmmnum{xprofile.m2_cross.other.n_utt}{2{,}939}
\defmmnum{xprofile.m2_cross.other.nonempty_n}{2{,}939}
\defmmnum{xprofile.m2_cross.other.nonempty_utt_median_wer}{100.00}
\defmmnum{xprofile.m2_cross.other.nonempty_wer}{134.18}
\defmmnum{xprofile.m2_cross.other.sub_pct}{46.30}
\defmmnum{xprofile.m2_cross.other.sub_words}{24{,}236}
\defmmnum{xprofile.m2_cross.other.utt_eq0_n}{2}
\defmmnum{xprofile.m2_cross.other.utt_eq0_pct}{0.07}
\defmmnum{xprofile.m2_cross.other.utt_ge100_n}{1{,}920}
\defmmnum{xprofile.m2_cross.other.utt_ge100_pct}{65.33}
\defmmnum{xprofile.m2_cross.other.utt_lt10_n}{2}
\defmmnum{xprofile.m2_cross.other.utt_lt10_pct}{0.07}
\defmmnum{xprofile.m2_cross.other.utt_lt50_n}{44}
\defmmnum{xprofile.m2_cross.other.utt_lt50_pct}{1.50}
\defmmnum{xprofile.m2_cross.other.utt_median_wer}{100.00}
\defmmnum{xprofile.m2_cross.other.wer}{134.18}
\defmmnum{xprofile.m2_cross.other.word_ratio_corpus}{0.96}
\defmmnum{xprofile.m2_cross.other.word_ratio_median}{0.80}
\defmmnum{xprofile.m2_same.clean.cap159_n}{0}
\defmmnum{xprofile.m2_same.clean.cap159_pct}{0.00}
\defmmnum{xprofile.m2_same.clean.cap160_n}{0}
\defmmnum{xprofile.m2_same.clean.cap160_pct}{0.00}
\defmmnum{xprofile.m2_same.clean.del_pct}{0.56}
\defmmnum{xprofile.m2_same.clean.del_words}{293}
\defmmnum{xprofile.m2_same.clean.empty_pct}{0.00}
\defmmnum{xprofile.m2_same.clean.ins_pct}{0.59}
\defmmnum{xprofile.m2_same.clean.ins_words}{310}
\defmmnum{xprofile.m2_same.clean.n_empty}{0}
\defmmnum{xprofile.m2_same.clean.n_utt}{2{,}620}
\defmmnum{xprofile.m2_same.clean.nonempty_n}{2{,}620}
\defmmnum{xprofile.m2_same.clean.nonempty_utt_median_wer}{1.90}
\defmmnum{xprofile.m2_same.clean.nonempty_wer}{5.13}
\defmmnum{xprofile.m2_same.clean.sub_pct}{3.98}
\defmmnum{xprofile.m2_same.clean.sub_words}{2{,}094}
\defmmnum{xprofile.m2_same.clean.utt_eq0_n}{1{,}299}
\defmmnum{xprofile.m2_same.clean.utt_eq0_pct}{49.58}
\defmmnum{xprofile.m2_same.clean.utt_ge100_n}{3}
\defmmnum{xprofile.m2_same.clean.utt_ge100_pct}{0.11}
\defmmnum{xprofile.m2_same.clean.utt_lt10_n}{2{,}054}
\defmmnum{xprofile.m2_same.clean.utt_lt10_pct}{78.40}
\defmmnum{xprofile.m2_same.clean.utt_lt50_n}{2{,}597}
\defmmnum{xprofile.m2_same.clean.utt_lt50_pct}{99.12}
\defmmnum{xprofile.m2_same.clean.utt_median_wer}{1.90}
\defmmnum{xprofile.m2_same.clean.wer}{5.13}
\defmmnum{xprofile.m2_same.clean.word_ratio_corpus}{1.00}
\defmmnum{xprofile.m2_same.clean.word_ratio_median}{1.00}
\defmmnum{xprofile.m2_same.other.cap159_n}{0}
\defmmnum{xprofile.m2_same.other.cap159_pct}{0.00}
\defmmnum{xprofile.m2_same.other.cap160_n}{0}
\defmmnum{xprofile.m2_same.other.cap160_pct}{0.00}
\defmmnum{xprofile.m2_same.other.del_pct}{1.28}
\defmmnum{xprofile.m2_same.other.del_words}{670}
\defmmnum{xprofile.m2_same.other.empty_pct}{0.00}
\defmmnum{xprofile.m2_same.other.ins_pct}{1.43}
\defmmnum{xprofile.m2_same.other.ins_words}{749}
\defmmnum{xprofile.m2_same.other.n_empty}{0}
\defmmnum{xprofile.m2_same.other.n_utt}{2{,}939}
\defmmnum{xprofile.m2_same.other.nonempty_n}{2{,}939}
\defmmnum{xprofile.m2_same.other.nonempty_utt_median_wer}{9.09}
\defmmnum{xprofile.m2_same.other.nonempty_wer}{11.15}
\defmmnum{xprofile.m2_same.other.sub_pct}{8.43}
\defmmnum{xprofile.m2_same.other.sub_words}{4{,}415}
\defmmnum{xprofile.m2_same.other.utt_eq0_n}{951}
\defmmnum{xprofile.m2_same.other.utt_eq0_pct}{32.36}
\defmmnum{xprofile.m2_same.other.utt_ge100_n}{17}
\defmmnum{xprofile.m2_same.other.utt_ge100_pct}{0.58}
\defmmnum{xprofile.m2_same.other.utt_lt10_n}{1{,}553}
\defmmnum{xprofile.m2_same.other.utt_lt10_pct}{52.84}
\defmmnum{xprofile.m2_same.other.utt_lt50_n}{2{,}812}
\defmmnum{xprofile.m2_same.other.utt_lt50_pct}{95.68}
\defmmnum{xprofile.m2_same.other.utt_median_wer}{9.09}
\defmmnum{xprofile.m2_same.other.wer}{11.15}
\defmmnum{xprofile.m2_same.other.word_ratio_corpus}{1.00}
\defmmnum{xprofile.m2_same.other.word_ratio_median}{1.00}
\defmmnum{xprofile.m3_cross.caponesixtypct}{75.95}
\defmmnum{xprofile.m3_cross.clean.cap159_n}{2{,}240}
\defmmnum{xprofile.m3_cross.clean.cap159_pct}{85.50}
\defmmnum{xprofile.m3_cross.clean.cap160_n}{1{,}990}
\defmmnum{xprofile.m3_cross.clean.cap160_pct}{75.95}
\defmmnum{xprofile.m3_cross.clean.del_pct}{0.95}
\defmmnum{xprofile.m3_cross.clean.del_words}{501}
\defmmnum{xprofile.m3_cross.clean.empty_pct}{0.00}
\defmmnum{xprofile.m3_cross.clean.ins_pct}{382.74}
\defmmnum{xprofile.m3_cross.clean.ins_words}{201{,}229}
\defmmnum{xprofile.m3_cross.clean.n_empty}{0}
\defmmnum{xprofile.m3_cross.clean.n_utt}{2{,}620}
\defmmnum{xprofile.m3_cross.clean.nonempty_n}{2{,}620}
\defmmnum{xprofile.m3_cross.clean.nonempty_utt_median_wer}{550.00}
\defmmnum{xprofile.m3_cross.clean.nonempty_wer}{452.43}
\defmmnum{xprofile.m3_cross.clean.sub_pct}{68.74}
\defmmnum{xprofile.m3_cross.clean.sub_words}{36{,}141}
\defmmnum{xprofile.m3_cross.clean.utt_eq0_n}{1}
\defmmnum{xprofile.m3_cross.clean.utt_eq0_pct}{0.04}
\defmmnum{xprofile.m3_cross.clean.utt_ge100_n}{2{,}594}
\defmmnum{xprofile.m3_cross.clean.utt_ge100_pct}{99.01}
\defmmnum{xprofile.m3_cross.clean.utt_lt10_n}{3}
\defmmnum{xprofile.m3_cross.clean.utt_lt10_pct}{0.11}
\defmmnum{xprofile.m3_cross.clean.utt_lt50_n}{17}
\defmmnum{xprofile.m3_cross.clean.utt_lt50_pct}{0.65}
\defmmnum{xprofile.m3_cross.clean.utt_median_wer}{550.00}
\defmmnum{xprofile.m3_cross.clean.wer}{452.43}
\defmmnum{xprofile.m3_cross.clean.word_ratio_corpus}{4.82}
\defmmnum{xprofile.m3_cross.clean.word_ratio_median}{5.83}
\defmmnum{xprofile.m3_cross.delpct}{0.95}
\defmmnum{xprofile.m3_cross.inspct}{382.74}
\defmmnum{xprofile.m3_cross.nonemptymedwer}{550.00}
\defmmnum{xprofile.m3_cross.nonemptywer}{452.43}
\defmmnum{xprofile.m3_cross.other.cap159_n}{2{,}500}
\defmmnum{xprofile.m3_cross.other.cap159_pct}{85.06}
\defmmnum{xprofile.m3_cross.other.cap160_n}{2{,}245}
\defmmnum{xprofile.m3_cross.other.cap160_pct}{76.39}
\defmmnum{xprofile.m3_cross.other.del_pct}{0.83}
\defmmnum{xprofile.m3_cross.other.del_words}{433}
\defmmnum{xprofile.m3_cross.other.empty_pct}{0.00}
\defmmnum{xprofile.m3_cross.other.ins_pct}{432.81}
\defmmnum{xprofile.m3_cross.other.ins_words}{226{,}545}
\defmmnum{xprofile.m3_cross.other.n_empty}{0}
\defmmnum{xprofile.m3_cross.other.n_utt}{2{,}939}
\defmmnum{xprofile.m3_cross.other.nonempty_n}{2{,}939}
\defmmnum{xprofile.m3_cross.other.nonempty_utt_median_wer}{617.65}
\defmmnum{xprofile.m3_cross.other.nonempty_wer}{505.36}
\defmmnum{xprofile.m3_cross.other.sub_pct}{71.72}
\defmmnum{xprofile.m3_cross.other.sub_words}{37{,}543}
\defmmnum{xprofile.m3_cross.other.utt_eq0_n}{0}
\defmmnum{xprofile.m3_cross.other.utt_eq0_pct}{0.00}
\defmmnum{xprofile.m3_cross.other.utt_ge100_n}{2{,}916}
\defmmnum{xprofile.m3_cross.other.utt_ge100_pct}{99.22}
\defmmnum{xprofile.m3_cross.other.utt_lt10_n}{0}
\defmmnum{xprofile.m3_cross.other.utt_lt10_pct}{0.00}
\defmmnum{xprofile.m3_cross.other.utt_lt50_n}{3}
\defmmnum{xprofile.m3_cross.other.utt_lt50_pct}{0.10}
\defmmnum{xprofile.m3_cross.other.utt_median_wer}{617.65}
\defmmnum{xprofile.m3_cross.other.wer}{505.36}
\defmmnum{xprofile.m3_cross.other.word_ratio_corpus}{5.32}
\defmmnum{xprofile.m3_cross.other.word_ratio_median}{6.53}
\defmmnum{xprofile.m3_cross.subpct}{68.74}
\defmmnum{xprofile.m3_cross.uttltfiftypct}{0.65}
\defmmnum{xprofile.m3_cross.uttmedwer}{550.00}
\defmmnum{xprofile.m3_cross.wer}{452.43}
\defmmnum{xprofile.m3_same.clean.cap159_n}{0}
\defmmnum{xprofile.m3_same.clean.cap159_pct}{0.00}
\defmmnum{xprofile.m3_same.clean.cap160_n}{0}
\defmmnum{xprofile.m3_same.clean.cap160_pct}{0.00}
\defmmnum{xprofile.m3_same.clean.del_pct}{0.36}
\defmmnum{xprofile.m3_same.clean.del_words}{191}
\defmmnum{xprofile.m3_same.clean.empty_pct}{0.00}
\defmmnum{xprofile.m3_same.clean.ins_pct}{0.54}
\defmmnum{xprofile.m3_same.clean.ins_words}{282}
\defmmnum{xprofile.m3_same.clean.n_empty}{0}
\defmmnum{xprofile.m3_same.clean.n_utt}{2{,}620}
\defmmnum{xprofile.m3_same.clean.nonempty_n}{2{,}620}
\defmmnum{xprofile.m3_same.clean.nonempty_utt_median_wer}{0.00}
\defmmnum{xprofile.m3_same.clean.nonempty_wer}{4.21}
\defmmnum{xprofile.m3_same.clean.sub_pct}{3.31}
\defmmnum{xprofile.m3_same.clean.sub_words}{1{,}740}
\defmmnum{xprofile.m3_same.clean.utt_eq0_n}{1{,}485}
\defmmnum{xprofile.m3_same.clean.utt_eq0_pct}{56.68}
\defmmnum{xprofile.m3_same.clean.utt_ge100_n}{4}
\defmmnum{xprofile.m3_same.clean.utt_ge100_pct}{0.15}
\defmmnum{xprofile.m3_same.clean.utt_lt10_n}{2{,}136}
\defmmnum{xprofile.m3_same.clean.utt_lt10_pct}{81.53}
\defmmnum{xprofile.m3_same.clean.utt_lt50_n}{2{,}587}
\defmmnum{xprofile.m3_same.clean.utt_lt50_pct}{98.74}
\defmmnum{xprofile.m3_same.clean.utt_median_wer}{0.00}
\defmmnum{xprofile.m3_same.clean.wer}{4.21}
\defmmnum{xprofile.m3_same.clean.word_ratio_corpus}{1.00}
\defmmnum{xprofile.m3_same.clean.word_ratio_median}{1.00}
\defmmnum{xprofile.m3_same.other.cap159_n}{0}
\defmmnum{xprofile.m3_same.other.cap159_pct}{0.00}
\defmmnum{xprofile.m3_same.other.cap160_n}{0}
\defmmnum{xprofile.m3_same.other.cap160_pct}{0.00}
\defmmnum{xprofile.m3_same.other.del_pct}{0.98}
\defmmnum{xprofile.m3_same.other.del_words}{514}
\defmmnum{xprofile.m3_same.other.empty_pct}{0.00}
\defmmnum{xprofile.m3_same.other.ins_pct}{1.54}
\defmmnum{xprofile.m3_same.other.ins_words}{806}
\defmmnum{xprofile.m3_same.other.n_empty}{0}
\defmmnum{xprofile.m3_same.other.n_utt}{2{,}939}
\defmmnum{xprofile.m3_same.other.nonempty_n}{2{,}939}
\defmmnum{xprofile.m3_same.other.nonempty_utt_median_wer}{7.69}
\defmmnum{xprofile.m3_same.other.nonempty_wer}{10.81}
\defmmnum{xprofile.m3_same.other.sub_pct}{8.29}
\defmmnum{xprofile.m3_same.other.sub_words}{4{,}339}
\defmmnum{xprofile.m3_same.other.utt_eq0_n}{995}
\defmmnum{xprofile.m3_same.other.utt_eq0_pct}{33.86}
\defmmnum{xprofile.m3_same.other.utt_ge100_n}{16}
\defmmnum{xprofile.m3_same.other.utt_ge100_pct}{0.54}
\defmmnum{xprofile.m3_same.other.utt_lt10_n}{1{,}650}
\defmmnum{xprofile.m3_same.other.utt_lt10_pct}{56.14}
\defmmnum{xprofile.m3_same.other.utt_lt50_n}{2{,}790}
\defmmnum{xprofile.m3_same.other.utt_lt50_pct}{94.93}
\defmmnum{xprofile.m3_same.other.utt_median_wer}{7.69}
\defmmnum{xprofile.m3_same.other.wer}{10.81}
\defmmnum{xprofile.m3_same.other.word_ratio_corpus}{1.01}
\defmmnum{xprofile.m3_same.other.word_ratio_median}{1.00}
\defmmnum{xprofile.m4_cross.clean.cap159_n}{1{,}723}
\defmmnum{xprofile.m4_cross.clean.cap159_pct}{65.76}
\defmmnum{xprofile.m4_cross.clean.cap160_n}{1{,}599}
\defmmnum{xprofile.m4_cross.clean.cap160_pct}{61.03}
\defmmnum{xprofile.m4_cross.clean.del_pct}{0.26}
\defmmnum{xprofile.m4_cross.clean.del_words}{137}
\defmmnum{xprofile.m4_cross.clean.empty_pct}{0.00}
\defmmnum{xprofile.m4_cross.clean.ins_pct}{382.21}
\defmmnum{xprofile.m4_cross.clean.ins_words}{200{,}953}
\defmmnum{xprofile.m4_cross.clean.n_empty}{0}
\defmmnum{xprofile.m4_cross.clean.n_utt}{2{,}620}
\defmmnum{xprofile.m4_cross.clean.nonempty_n}{2{,}620}
\defmmnum{xprofile.m4_cross.clean.nonempty_utt_median_wer}{566.67}
\defmmnum{xprofile.m4_cross.clean.nonempty_wer}{465.22}
\defmmnum{xprofile.m4_cross.clean.sub_pct}{82.75}
\defmmnum{xprofile.m4_cross.clean.sub_words}{43{,}506}
\defmmnum{xprofile.m4_cross.clean.utt_eq0_n}{0}
\defmmnum{xprofile.m4_cross.clean.utt_eq0_pct}{0.00}
\defmmnum{xprofile.m4_cross.clean.utt_ge100_n}{2{,}610}
\defmmnum{xprofile.m4_cross.clean.utt_ge100_pct}{99.62}
\defmmnum{xprofile.m4_cross.clean.utt_lt10_n}{0}
\defmmnum{xprofile.m4_cross.clean.utt_lt10_pct}{0.00}
\defmmnum{xprofile.m4_cross.clean.utt_lt50_n}{0}
\defmmnum{xprofile.m4_cross.clean.utt_lt50_pct}{0.00}
\defmmnum{xprofile.m4_cross.clean.utt_median_wer}{566.67}
\defmmnum{xprofile.m4_cross.clean.wer}{465.22}
\defmmnum{xprofile.m4_cross.clean.word_ratio_corpus}{4.82}
\defmmnum{xprofile.m4_cross.clean.word_ratio_median}{5.88}
\defmmnum{xprofile.m4_cross.other.cap159_n}{1{,}825}
\defmmnum{xprofile.m4_cross.other.cap159_pct}{62.10}
\defmmnum{xprofile.m4_cross.other.cap160_n}{1{,}683}
\defmmnum{xprofile.m4_cross.other.cap160_pct}{57.26}
\defmmnum{xprofile.m4_cross.other.del_pct}{0.28}
\defmmnum{xprofile.m4_cross.other.del_words}{149}
\defmmnum{xprofile.m4_cross.other.empty_pct}{0.00}
\defmmnum{xprofile.m4_cross.other.ins_pct}{439.95}
\defmmnum{xprofile.m4_cross.other.ins_words}{230{,}281}
\defmmnum{xprofile.m4_cross.other.n_empty}{0}
\defmmnum{xprofile.m4_cross.other.n_utt}{2{,}939}
\defmmnum{xprofile.m4_cross.other.nonempty_n}{2{,}939}
\defmmnum{xprofile.m4_cross.other.nonempty_utt_median_wer}{652.94}
\defmmnum{xprofile.m4_cross.other.nonempty_wer}{522.81}
\defmmnum{xprofile.m4_cross.other.sub_pct}{82.58}
\defmmnum{xprofile.m4_cross.other.sub_words}{43{,}223}
\defmmnum{xprofile.m4_cross.other.utt_eq0_n}{0}
\defmmnum{xprofile.m4_cross.other.utt_eq0_pct}{0.00}
\defmmnum{xprofile.m4_cross.other.utt_ge100_n}{2{,}935}
\defmmnum{xprofile.m4_cross.other.utt_ge100_pct}{99.86}
\defmmnum{xprofile.m4_cross.other.utt_lt10_n}{0}
\defmmnum{xprofile.m4_cross.other.utt_lt10_pct}{0.00}
\defmmnum{xprofile.m4_cross.other.utt_lt50_n}{0}
\defmmnum{xprofile.m4_cross.other.utt_lt50_pct}{0.00}
\defmmnum{xprofile.m4_cross.other.utt_median_wer}{652.94}
\defmmnum{xprofile.m4_cross.other.wer}{522.81}
\defmmnum{xprofile.m4_cross.other.word_ratio_corpus}{5.40}
\defmmnum{xprofile.m4_cross.other.word_ratio_median}{6.71}
\defmmnum{xprofile.m4_same.clean.cap159_n}{0}
\defmmnum{xprofile.m4_same.clean.cap159_pct}{0.00}
\defmmnum{xprofile.m4_same.clean.cap160_n}{0}
\defmmnum{xprofile.m4_same.clean.cap160_pct}{0.00}
\defmmnum{xprofile.m4_same.clean.del_pct}{0.38}
\defmmnum{xprofile.m4_same.clean.del_words}{199}
\defmmnum{xprofile.m4_same.clean.empty_pct}{0.00}
\defmmnum{xprofile.m4_same.clean.ins_pct}{0.73}
\defmmnum{xprofile.m4_same.clean.ins_words}{383}
\defmmnum{xprofile.m4_same.clean.n_empty}{0}
\defmmnum{xprofile.m4_same.clean.n_utt}{2{,}620}
\defmmnum{xprofile.m4_same.clean.nonempty_n}{2{,}620}
\defmmnum{xprofile.m4_same.clean.nonempty_utt_median_wer}{0.00}
\defmmnum{xprofile.m4_same.clean.nonempty_wer}{5.05}
\defmmnum{xprofile.m4_same.clean.sub_pct}{3.94}
\defmmnum{xprofile.m4_same.clean.sub_words}{2{,}074}
\defmmnum{xprofile.m4_same.clean.utt_eq0_n}{1{,}341}
\defmmnum{xprofile.m4_same.clean.utt_eq0_pct}{51.18}
\defmmnum{xprofile.m4_same.clean.utt_ge100_n}{4}
\defmmnum{xprofile.m4_same.clean.utt_ge100_pct}{0.15}
\defmmnum{xprofile.m4_same.clean.utt_lt10_n}{2{,}055}
\defmmnum{xprofile.m4_same.clean.utt_lt10_pct}{78.44}
\defmmnum{xprofile.m4_same.clean.utt_lt50_n}{2{,}586}
\defmmnum{xprofile.m4_same.clean.utt_lt50_pct}{98.70}
\defmmnum{xprofile.m4_same.clean.utt_median_wer}{0.00}
\defmmnum{xprofile.m4_same.clean.wer}{5.05}
\defmmnum{xprofile.m4_same.clean.word_ratio_corpus}{1.00}
\defmmnum{xprofile.m4_same.clean.word_ratio_median}{1.00}
\defmmnum{xprofile.m4_same.other.cap159_n}{2}
\defmmnum{xprofile.m4_same.other.cap159_pct}{0.07}
\defmmnum{xprofile.m4_same.other.cap160_n}{2}
\defmmnum{xprofile.m4_same.other.cap160_pct}{0.07}
\defmmnum{xprofile.m4_same.other.del_pct}{1.13}
\defmmnum{xprofile.m4_same.other.del_words}{592}
\defmmnum{xprofile.m4_same.other.empty_pct}{0.00}
\defmmnum{xprofile.m4_same.other.ins_pct}{2.38}
\defmmnum{xprofile.m4_same.other.ins_words}{1{,}244}
\defmmnum{xprofile.m4_same.other.n_empty}{0}
\defmmnum{xprofile.m4_same.other.n_utt}{2{,}939}
\defmmnum{xprofile.m4_same.other.nonempty_n}{2{,}939}
\defmmnum{xprofile.m4_same.other.nonempty_utt_median_wer}{9.09}
\defmmnum{xprofile.m4_same.other.nonempty_wer}{12.71}
\defmmnum{xprofile.m4_same.other.sub_pct}{9.20}
\defmmnum{xprofile.m4_same.other.sub_words}{4{,}818}
\defmmnum{xprofile.m4_same.other.utt_eq0_n}{900}
\defmmnum{xprofile.m4_same.other.utt_eq0_pct}{30.62}
\defmmnum{xprofile.m4_same.other.utt_ge100_n}{22}
\defmmnum{xprofile.m4_same.other.utt_ge100_pct}{0.75}
\defmmnum{xprofile.m4_same.other.utt_lt10_n}{1{,}514}
\defmmnum{xprofile.m4_same.other.utt_lt10_pct}{51.51}
\defmmnum{xprofile.m4_same.other.utt_lt50_n}{2{,}751}
\defmmnum{xprofile.m4_same.other.utt_lt50_pct}{93.60}
\defmmnum{xprofile.m4_same.other.utt_median_wer}{9.09}
\defmmnum{xprofile.m4_same.other.wer}{12.71}
\defmmnum{xprofile.m4_same.other.word_ratio_corpus}{1.01}
\defmmnum{xprofile.m4_same.other.word_ratio_median}{1.00}
\defmmnum{xprofile.m5_cross.caponesixtypct}{13.02}
\defmmnum{xprofile.m5_cross.clean.cap159_n}{357}
\defmmnum{xprofile.m5_cross.clean.cap159_pct}{13.63}
\defmmnum{xprofile.m5_cross.clean.cap160_n}{341}
\defmmnum{xprofile.m5_cross.clean.cap160_pct}{13.02}
\defmmnum{xprofile.m5_cross.clean.del_pct}{2.37}
\defmmnum{xprofile.m5_cross.clean.del_words}{1{,}245}
\defmmnum{xprofile.m5_cross.clean.empty_pct}{0.00}
\defmmnum{xprofile.m5_cross.clean.ins_pct}{123.44}
\defmmnum{xprofile.m5_cross.clean.ins_words}{64{,}898}
\defmmnum{xprofile.m5_cross.clean.n_empty}{0}
\defmmnum{xprofile.m5_cross.clean.n_utt}{2{,}620}
\defmmnum{xprofile.m5_cross.clean.nonempty_n}{2{,}620}
\defmmnum{xprofile.m5_cross.clean.nonempty_utt_median_wer}{108.33}
\defmmnum{xprofile.m5_cross.clean.nonempty_wer}{154.00}
\defmmnum{xprofile.m5_cross.clean.sub_pct}{28.19}
\defmmnum{xprofile.m5_cross.clean.sub_words}{14{,}823}
\defmmnum{xprofile.m5_cross.clean.utt_eq0_n}{7}
\defmmnum{xprofile.m5_cross.clean.utt_eq0_pct}{0.27}
\defmmnum{xprofile.m5_cross.clean.utt_ge100_n}{1{,}473}
\defmmnum{xprofile.m5_cross.clean.utt_ge100_pct}{56.22}
\defmmnum{xprofile.m5_cross.clean.utt_lt10_n}{21}
\defmmnum{xprofile.m5_cross.clean.utt_lt10_pct}{0.80}
\defmmnum{xprofile.m5_cross.clean.utt_lt50_n}{264}
\defmmnum{xprofile.m5_cross.clean.utt_lt50_pct}{10.08}
\defmmnum{xprofile.m5_cross.clean.utt_median_wer}{108.33}
\defmmnum{xprofile.m5_cross.clean.wer}{154.00}
\defmmnum{xprofile.m5_cross.clean.word_ratio_corpus}{2.21}
\defmmnum{xprofile.m5_cross.clean.word_ratio_median}{1.82}
\defmmnum{xprofile.m5_cross.delpct}{2.37}
\defmmnum{xprofile.m5_cross.inspct}{123.44}
\defmmnum{xprofile.m5_cross.nonemptymedwer}{108.33}
\defmmnum{xprofile.m5_cross.nonemptywer}{154.00}
\defmmnum{xprofile.m5_cross.other.cap159_n}{524}
\defmmnum{xprofile.m5_cross.other.cap159_pct}{17.83}
\defmmnum{xprofile.m5_cross.other.cap160_n}{507}
\defmmnum{xprofile.m5_cross.other.cap160_pct}{17.25}
\defmmnum{xprofile.m5_cross.other.del_pct}{2.07}
\defmmnum{xprofile.m5_cross.other.del_words}{1{,}081}
\defmmnum{xprofile.m5_cross.other.empty_pct}{0.00}
\defmmnum{xprofile.m5_cross.other.ins_pct}{162.39}
\defmmnum{xprofile.m5_cross.other.ins_words}{85{,}001}
\defmmnum{xprofile.m5_cross.other.n_empty}{0}
\defmmnum{xprofile.m5_cross.other.n_utt}{2{,}939}
\defmmnum{xprofile.m5_cross.other.nonempty_n}{2{,}939}
\defmmnum{xprofile.m5_cross.other.nonempty_utt_median_wer}{150.00}
\defmmnum{xprofile.m5_cross.other.nonempty_wer}{205.91}
\defmmnum{xprofile.m5_cross.other.sub_pct}{41.45}
\defmmnum{xprofile.m5_cross.other.sub_words}{21{,}698}
\defmmnum{xprofile.m5_cross.other.utt_eq0_n}{4}
\defmmnum{xprofile.m5_cross.other.utt_eq0_pct}{0.14}
\defmmnum{xprofile.m5_cross.other.utt_ge100_n}{2{,}101}
\defmmnum{xprofile.m5_cross.other.utt_ge100_pct}{71.49}
\defmmnum{xprofile.m5_cross.other.utt_lt10_n}{6}
\defmmnum{xprofile.m5_cross.other.utt_lt10_pct}{0.20}
\defmmnum{xprofile.m5_cross.other.utt_lt50_n}{101}
\defmmnum{xprofile.m5_cross.other.utt_lt50_pct}{3.44}
\defmmnum{xprofile.m5_cross.other.utt_median_wer}{150.00}
\defmmnum{xprofile.m5_cross.other.wer}{205.91}
\defmmnum{xprofile.m5_cross.other.word_ratio_corpus}{2.60}
\defmmnum{xprofile.m5_cross.other.word_ratio_median}{2.06}
\defmmnum{xprofile.m5_cross.subpct}{28.19}
\defmmnum{xprofile.m5_cross.uttltfiftypct}{10.08}
\defmmnum{xprofile.m5_cross.uttmedwer}{108.33}
\defmmnum{xprofile.m5_cross.wer}{154.00}
\defmmnum{xprofile.m5_same.clean.cap159_n}{0}
\defmmnum{xprofile.m5_same.clean.cap159_pct}{0.00}
\defmmnum{xprofile.m5_same.clean.cap160_n}{0}
\defmmnum{xprofile.m5_same.clean.cap160_pct}{0.00}
\defmmnum{xprofile.m5_same.clean.del_pct}{0.26}
\defmmnum{xprofile.m5_same.clean.del_words}{138}
\defmmnum{xprofile.m5_same.clean.empty_pct}{0.00}
\defmmnum{xprofile.m5_same.clean.ins_pct}{0.46}
\defmmnum{xprofile.m5_same.clean.ins_words}{243}
\defmmnum{xprofile.m5_same.clean.n_empty}{0}
\defmmnum{xprofile.m5_same.clean.n_utt}{2{,}620}
\defmmnum{xprofile.m5_same.clean.nonempty_n}{2{,}620}
\defmmnum{xprofile.m5_same.clean.nonempty_utt_median_wer}{0.00}
\defmmnum{xprofile.m5_same.clean.nonempty_wer}{3.26}
\defmmnum{xprofile.m5_same.clean.sub_pct}{2.53}
\defmmnum{xprofile.m5_same.clean.sub_words}{1{,}332}
\defmmnum{xprofile.m5_same.clean.utt_eq0_n}{1{,}697}
\defmmnum{xprofile.m5_same.clean.utt_eq0_pct}{64.77}
\defmmnum{xprofile.m5_same.clean.utt_ge100_n}{2}
\defmmnum{xprofile.m5_same.clean.utt_ge100_pct}{0.08}
\defmmnum{xprofile.m5_same.clean.utt_lt10_n}{2{,}270}
\defmmnum{xprofile.m5_same.clean.utt_lt10_pct}{86.64}
\defmmnum{xprofile.m5_same.clean.utt_lt50_n}{2{,}596}
\defmmnum{xprofile.m5_same.clean.utt_lt50_pct}{99.08}
\defmmnum{xprofile.m5_same.clean.utt_median_wer}{0.00}
\defmmnum{xprofile.m5_same.clean.wer}{3.26}
\defmmnum{xprofile.m5_same.clean.word_ratio_corpus}{1.00}
\defmmnum{xprofile.m5_same.clean.word_ratio_median}{1.00}
\defmmnum{xprofile.m5_same.other.cap159_n}{0}
\defmmnum{xprofile.m5_same.other.cap159_pct}{0.00}
\defmmnum{xprofile.m5_same.other.cap160_n}{0}
\defmmnum{xprofile.m5_same.other.cap160_pct}{0.00}
\defmmnum{xprofile.m5_same.other.del_pct}{0.69}
\defmmnum{xprofile.m5_same.other.del_words}{361}
\defmmnum{xprofile.m5_same.other.empty_pct}{0.00}
\defmmnum{xprofile.m5_same.other.ins_pct}{1.18}
\defmmnum{xprofile.m5_same.other.ins_words}{616}
\defmmnum{xprofile.m5_same.other.n_empty}{0}
\defmmnum{xprofile.m5_same.other.n_utt}{2{,}939}
\defmmnum{xprofile.m5_same.other.nonempty_n}{2{,}939}
\defmmnum{xprofile.m5_same.other.nonempty_utt_median_wer}{4.08}
\defmmnum{xprofile.m5_same.other.nonempty_wer}{7.71}
\defmmnum{xprofile.m5_same.other.sub_pct}{5.84}
\defmmnum{xprofile.m5_same.other.sub_words}{3{,}059}
\defmmnum{xprofile.m5_same.other.utt_eq0_n}{1{,}295}
\defmmnum{xprofile.m5_same.other.utt_eq0_pct}{44.06}
\defmmnum{xprofile.m5_same.other.utt_ge100_n}{18}
\defmmnum{xprofile.m5_same.other.utt_ge100_pct}{0.61}
\defmmnum{xprofile.m5_same.other.utt_lt10_n}{1{,}979}
\defmmnum{xprofile.m5_same.other.utt_lt10_pct}{67.34}
\defmmnum{xprofile.m5_same.other.utt_lt50_n}{2{,}848}
\defmmnum{xprofile.m5_same.other.utt_lt50_pct}{96.90}
\defmmnum{xprofile.m5_same.other.utt_median_wer}{4.08}
\defmmnum{xprofile.m5_same.other.wer}{7.71}
\defmmnum{xprofile.m5_same.other.word_ratio_corpus}{1.00}
\defmmnum{xprofile.m5_same.other.word_ratio_median}{1.00}
\defmmnum{xprofile.missing_cells}{0}
\defmmnum{xprofile.ref_words.clean}{52{,}576}
\defmmnum{xprofile.ref_words.other}{52{,}343}
\defmmnum{xprofile.regime1.max_lt50_clean}{99.31}
\defmmnum{xprofile.regime1.max_lt50_other}{97.04}
\defmmnum{xprofile.regime1.max_med_clean}{6.90}
\defmmnum{xprofile.regime1.max_med_other}{16.67}
\defmmnum{xprofile.regime1.max_wer_clean}{9.04}
\defmmnum{xprofile.regime1.max_wer_other}{18.91}
\defmmnum{xprofile.regime1.min_lt50_clean}{97.90}
\defmmnum{xprofile.regime1.min_lt50_other}{89.93}
\defmmnum{xprofile.regime1.min_med_clean}{0.00}
\defmmnum{xprofile.regime1.min_med_other}{3.85}
\defmmnum{xprofile.regime1.min_wer_clean}{3.08}
\defmmnum{xprofile.regime1.min_wer_other}{7.38}
\defmmnum{xprofile.regime1.n_cells}{17}
\defmmnum{xprofile.regime2.max_lt50_clean}{75.15}
\defmmnum{xprofile.regime2.max_lt50_other}{60.33}
\defmmnum{xprofile.regime2.max_med_clean}{23.81}
\defmmnum{xprofile.regime2.max_med_other}{40.00}
\defmmnum{xprofile.regime2.max_wer_clean}{75.35}
\defmmnum{xprofile.regime2.max_wer_other}{89.82}
\defmmnum{xprofile.regime2.min_lt50_clean}{66.30}
\defmmnum{xprofile.regime2.min_lt50_other}{57.06}
\defmmnum{xprofile.regime2.min_med_clean}{20.83}
\defmmnum{xprofile.regime2.min_med_other}{36.36}
\defmmnum{xprofile.regime2.min_wer_clean}{67.35}
\defmmnum{xprofile.regime2.min_wer_other}{77.47}
\defmmnum{xprofile.regime2.n_cells}{2}
\defmmnum{xprofile.regime3.max_lt50_clean}{13.82}
\defmmnum{xprofile.regime3.max_lt50_other}{8.30}
\defmmnum{xprofile.regime3.max_med_clean}{625.00}
\defmmnum{xprofile.regime3.max_med_other}{725.00}
\defmmnum{xprofile.regime3.max_wer_clean}{490.78}
\defmmnum{xprofile.regime3.max_wer_other}{562.71}
\defmmnum{xprofile.regime3.min_lt50_clean}{0.00}
\defmmnum{xprofile.regime3.min_lt50_other}{0.00}
\defmmnum{xprofile.regime3.min_med_clean}{98.00}
\defmmnum{xprofile.regime3.min_med_other}{100.00}
\defmmnum{xprofile.regime3.min_wer_clean}{98.63}
\defmmnum{xprofile.regime3.min_wer_other}{100.43}
\defmmnum{xprofile.regime3.n_cells}{127}
\defmmnum{xprofile.total_cells}{154}
\defmmnum{xprofile.w1_ms.clean.cap159_n}{0}
\defmmnum{xprofile.w1_ms.clean.cap159_pct}{0.00}
\defmmnum{xprofile.w1_ms.clean.cap160_n}{0}
\defmmnum{xprofile.w1_ms.clean.cap160_pct}{0.00}
\defmmnum{xprofile.w1_ms.clean.del_pct}{62.08}
\defmmnum{xprofile.w1_ms.clean.del_words}{32{,}640}
\defmmnum{xprofile.w1_ms.clean.empty_pct}{0.00}
\defmmnum{xprofile.w1_ms.clean.ins_pct}{1.95}
\defmmnum{xprofile.w1_ms.clean.ins_words}{1{,}025}
\defmmnum{xprofile.w1_ms.clean.n_empty}{0}
\defmmnum{xprofile.w1_ms.clean.n_utt}{2{,}620}
\defmmnum{xprofile.w1_ms.clean.nonempty_n}{2{,}620}
\defmmnum{xprofile.w1_ms.clean.nonempty_utt_median_wer}{100.00}
\defmmnum{xprofile.w1_ms.clean.nonempty_wer}{99.65}
\defmmnum{xprofile.w1_ms.clean.sub_pct}{35.62}
\defmmnum{xprofile.w1_ms.clean.sub_words}{18{,}725}
\defmmnum{xprofile.w1_ms.clean.utt_eq0_n}{0}
\defmmnum{xprofile.w1_ms.clean.utt_eq0_pct}{0.00}
\defmmnum{xprofile.w1_ms.clean.utt_ge100_n}{1{,}748}
\defmmnum{xprofile.w1_ms.clean.utt_ge100_pct}{66.72}
\defmmnum{xprofile.w1_ms.clean.utt_lt10_n}{0}
\defmmnum{xprofile.w1_ms.clean.utt_lt10_pct}{0.00}
\defmmnum{xprofile.w1_ms.clean.utt_lt50_n}{0}
\defmmnum{xprofile.w1_ms.clean.utt_lt50_pct}{0.00}
\defmmnum{xprofile.w1_ms.clean.utt_median_wer}{100.00}
\defmmnum{xprofile.w1_ms.clean.wer}{99.65}
\defmmnum{xprofile.w1_ms.clean.word_ratio_corpus}{0.40}
\defmmnum{xprofile.w1_ms.clean.word_ratio_median}{0.49}
\defmmnum{xprofile.w1_ms.other.cap159_n}{0}
\defmmnum{xprofile.w1_ms.other.cap159_pct}{0.00}
\defmmnum{xprofile.w1_ms.other.cap160_n}{0}
\defmmnum{xprofile.w1_ms.other.cap160_pct}{0.00}
\defmmnum{xprofile.w1_ms.other.del_pct}{57.90}
\defmmnum{xprofile.w1_ms.other.del_words}{30{,}307}
\defmmnum{xprofile.w1_ms.other.empty_pct}{0.00}
\defmmnum{xprofile.w1_ms.other.ins_pct}{2.82}
\defmmnum{xprofile.w1_ms.other.ins_words}{1{,}477}
\defmmnum{xprofile.w1_ms.other.n_empty}{0}
\defmmnum{xprofile.w1_ms.other.n_utt}{2{,}939}
\defmmnum{xprofile.w1_ms.other.nonempty_n}{2{,}939}
\defmmnum{xprofile.w1_ms.other.nonempty_utt_median_wer}{100.00}
\defmmnum{xprofile.w1_ms.other.nonempty_wer}{100.43}
\defmmnum{xprofile.w1_ms.other.sub_pct}{39.71}
\defmmnum{xprofile.w1_ms.other.sub_words}{20{,}785}
\defmmnum{xprofile.w1_ms.other.utt_eq0_n}{0}
\defmmnum{xprofile.w1_ms.other.utt_eq0_pct}{0.00}
\defmmnum{xprofile.w1_ms.other.utt_ge100_n}{2{,}063}
\defmmnum{xprofile.w1_ms.other.utt_ge100_pct}{70.19}
\defmmnum{xprofile.w1_ms.other.utt_lt10_n}{0}
\defmmnum{xprofile.w1_ms.other.utt_lt10_pct}{0.00}
\defmmnum{xprofile.w1_ms.other.utt_lt50_n}{0}
\defmmnum{xprofile.w1_ms.other.utt_lt50_pct}{0.00}
\defmmnum{xprofile.w1_ms.other.utt_median_wer}{100.00}
\defmmnum{xprofile.w1_ms.other.wer}{100.43}
\defmmnum{xprofile.w1_ms.other.word_ratio_corpus}{0.45}
\defmmnum{xprofile.w1_ms.other.word_ratio_median}{0.57}
\defmmnum{xprofile.w1_wh.clean.cap159_n}{0}
\defmmnum{xprofile.w1_wh.clean.cap159_pct}{0.00}
\defmmnum{xprofile.w1_wh.clean.cap160_n}{0}
\defmmnum{xprofile.w1_wh.clean.cap160_pct}{0.00}
\defmmnum{xprofile.w1_wh.clean.del_pct}{56.59}
\defmmnum{xprofile.w1_wh.clean.del_words}{29{,}753}
\defmmnum{xprofile.w1_wh.clean.empty_pct}{0.00}
\defmmnum{xprofile.w1_wh.clean.ins_pct}{10.80}
\defmmnum{xprofile.w1_wh.clean.ins_words}{5{,}679}
\defmmnum{xprofile.w1_wh.clean.n_empty}{0}
\defmmnum{xprofile.w1_wh.clean.n_utt}{2{,}620}
\defmmnum{xprofile.w1_wh.clean.nonempty_n}{2{,}620}
\defmmnum{xprofile.w1_wh.clean.nonempty_utt_median_wer}{100.00}
\defmmnum{xprofile.w1_wh.clean.nonempty_wer}{108.25}
\defmmnum{xprofile.w1_wh.clean.sub_pct}{40.86}
\defmmnum{xprofile.w1_wh.clean.sub_words}{21{,}480}
\defmmnum{xprofile.w1_wh.clean.utt_eq0_n}{0}
\defmmnum{xprofile.w1_wh.clean.utt_eq0_pct}{0.00}
\defmmnum{xprofile.w1_wh.clean.utt_ge100_n}{1{,}812}
\defmmnum{xprofile.w1_wh.clean.utt_ge100_pct}{69.16}
\defmmnum{xprofile.w1_wh.clean.utt_lt10_n}{0}
\defmmnum{xprofile.w1_wh.clean.utt_lt10_pct}{0.00}
\defmmnum{xprofile.w1_wh.clean.utt_lt50_n}{0}
\defmmnum{xprofile.w1_wh.clean.utt_lt50_pct}{0.00}
\defmmnum{xprofile.w1_wh.clean.utt_median_wer}{100.00}
\defmmnum{xprofile.w1_wh.clean.wer}{108.25}
\defmmnum{xprofile.w1_wh.clean.word_ratio_corpus}{0.54}
\defmmnum{xprofile.w1_wh.clean.word_ratio_median}{0.53}
\defmmnum{xprofile.w1_wh.other.cap159_n}{0}
\defmmnum{xprofile.w1_wh.other.cap159_pct}{0.00}
\defmmnum{xprofile.w1_wh.other.cap160_n}{0}
\defmmnum{xprofile.w1_wh.other.cap160_pct}{0.00}
\defmmnum{xprofile.w1_wh.other.del_pct}{51.73}
\defmmnum{xprofile.w1_wh.other.del_words}{27{,}076}
\defmmnum{xprofile.w1_wh.other.empty_pct}{0.00}
\defmmnum{xprofile.w1_wh.other.ins_pct}{16.28}
\defmmnum{xprofile.w1_wh.other.ins_words}{8{,}522}
\defmmnum{xprofile.w1_wh.other.n_empty}{0}
\defmmnum{xprofile.w1_wh.other.n_utt}{2{,}939}
\defmmnum{xprofile.w1_wh.other.nonempty_n}{2{,}939}
\defmmnum{xprofile.w1_wh.other.nonempty_utt_median_wer}{100.00}
\defmmnum{xprofile.w1_wh.other.nonempty_wer}{113.56}
\defmmnum{xprofile.w1_wh.other.sub_pct}{45.56}
\defmmnum{xprofile.w1_wh.other.sub_words}{23{,}845}
\defmmnum{xprofile.w1_wh.other.utt_eq0_n}{0}
\defmmnum{xprofile.w1_wh.other.utt_eq0_pct}{0.00}
\defmmnum{xprofile.w1_wh.other.utt_ge100_n}{2{,}150}
\defmmnum{xprofile.w1_wh.other.utt_ge100_pct}{73.15}
\defmmnum{xprofile.w1_wh.other.utt_lt10_n}{0}
\defmmnum{xprofile.w1_wh.other.utt_lt10_pct}{0.00}
\defmmnum{xprofile.w1_wh.other.utt_lt50_n}{0}
\defmmnum{xprofile.w1_wh.other.utt_lt50_pct}{0.00}
\defmmnum{xprofile.w1_wh.other.utt_median_wer}{100.00}
\defmmnum{xprofile.w1_wh.other.wer}{113.56}
\defmmnum{xprofile.w1_wh.other.word_ratio_corpus}{0.65}
\defmmnum{xprofile.w1_wh.other.word_ratio_median}{0.67}
\defmmnum{xprofile.w2_ms.clean.cap159_n}{2{,}405}
\defmmnum{xprofile.w2_ms.clean.cap159_pct}{91.79}
\defmmnum{xprofile.w2_ms.clean.cap160_n}{2{,}303}
\defmmnum{xprofile.w2_ms.clean.cap160_pct}{87.90}
\defmmnum{xprofile.w2_ms.clean.del_pct}{3.67}
\defmmnum{xprofile.w2_ms.clean.del_words}{1{,}931}
\defmmnum{xprofile.w2_ms.clean.empty_pct}{0.00}
\defmmnum{xprofile.w2_ms.clean.ins_pct}{406.25}
\defmmnum{xprofile.w2_ms.clean.ins_words}{213{,}592}
\defmmnum{xprofile.w2_ms.clean.n_empty}{0}
\defmmnum{xprofile.w2_ms.clean.n_utt}{2{,}620}
\defmmnum{xprofile.w2_ms.clean.nonempty_n}{2{,}620}
\defmmnum{xprofile.w2_ms.clean.nonempty_utt_median_wer}{625.00}
\defmmnum{xprofile.w2_ms.clean.nonempty_wer}{490.78}
\defmmnum{xprofile.w2_ms.clean.sub_pct}{80.85}
\defmmnum{xprofile.w2_ms.clean.sub_words}{42{,}508}
\defmmnum{xprofile.w2_ms.clean.utt_eq0_n}{0}
\defmmnum{xprofile.w2_ms.clean.utt_eq0_pct}{0.00}
\defmmnum{xprofile.w2_ms.clean.utt_ge100_n}{2{,}500}
\defmmnum{xprofile.w2_ms.clean.utt_ge100_pct}{95.42}
\defmmnum{xprofile.w2_ms.clean.utt_lt10_n}{0}
\defmmnum{xprofile.w2_ms.clean.utt_lt10_pct}{0.00}
\defmmnum{xprofile.w2_ms.clean.utt_lt50_n}{0}
\defmmnum{xprofile.w2_ms.clean.utt_lt50_pct}{0.00}
\defmmnum{xprofile.w2_ms.clean.utt_median_wer}{625.00}
\defmmnum{xprofile.w2_ms.clean.wer}{490.78}
\defmmnum{xprofile.w2_ms.clean.word_ratio_corpus}{5.03}
\defmmnum{xprofile.w2_ms.clean.word_ratio_median}{6.50}
\defmmnum{xprofile.w2_ms.other.cap159_n}{2{,}752}
\defmmnum{xprofile.w2_ms.other.cap159_pct}{93.64}
\defmmnum{xprofile.w2_ms.other.cap160_n}{2{,}627}
\defmmnum{xprofile.w2_ms.other.cap160_pct}{89.38}
\defmmnum{xprofile.w2_ms.other.del_pct}{3.07}
\defmmnum{xprofile.w2_ms.other.del_words}{1{,}605}
\defmmnum{xprofile.w2_ms.other.empty_pct}{0.00}
\defmmnum{xprofile.w2_ms.other.ins_pct}{478.38}
\defmmnum{xprofile.w2_ms.other.ins_words}{250{,}398}
\defmmnum{xprofile.w2_ms.other.n_empty}{0}
\defmmnum{xprofile.w2_ms.other.n_utt}{2{,}939}
\defmmnum{xprofile.w2_ms.other.nonempty_n}{2{,}939}
\defmmnum{xprofile.w2_ms.other.nonempty_utt_median_wer}{725.00}
\defmmnum{xprofile.w2_ms.other.nonempty_wer}{562.71}
\defmmnum{xprofile.w2_ms.other.sub_pct}{81.27}
\defmmnum{xprofile.w2_ms.other.sub_words}{42{,}538}
\defmmnum{xprofile.w2_ms.other.utt_eq0_n}{0}
\defmmnum{xprofile.w2_ms.other.utt_eq0_pct}{0.00}
\defmmnum{xprofile.w2_ms.other.utt_ge100_n}{2{,}848}
\defmmnum{xprofile.w2_ms.other.utt_ge100_pct}{96.90}
\defmmnum{xprofile.w2_ms.other.utt_lt10_n}{0}
\defmmnum{xprofile.w2_ms.other.utt_lt10_pct}{0.00}
\defmmnum{xprofile.w2_ms.other.utt_lt50_n}{0}
\defmmnum{xprofile.w2_ms.other.utt_lt50_pct}{0.00}
\defmmnum{xprofile.w2_ms.other.utt_median_wer}{725.00}
\defmmnum{xprofile.w2_ms.other.wer}{562.71}
\defmmnum{xprofile.w2_ms.other.word_ratio_corpus}{5.75}
\defmmnum{xprofile.w2_ms.other.word_ratio_median}{7.43}
\defmmnum{xprofile.w2_wh.clean.cap159_n}{2{,}544}
\defmmnum{xprofile.w2_wh.clean.cap159_pct}{97.10}
\defmmnum{xprofile.w2_wh.clean.cap160_n}{2{,}275}
\defmmnum{xprofile.w2_wh.clean.cap160_pct}{86.83}
\defmmnum{xprofile.w2_wh.clean.del_pct}{1.76}
\defmmnum{xprofile.w2_wh.clean.del_words}{924}
\defmmnum{xprofile.w2_wh.clean.empty_pct}{0.00}
\defmmnum{xprofile.w2_wh.clean.ins_pct}{390.15}
\defmmnum{xprofile.w2_wh.clean.ins_words}{205{,}127}
\defmmnum{xprofile.w2_wh.clean.n_empty}{0}
\defmmnum{xprofile.w2_wh.clean.n_utt}{2{,}620}
\defmmnum{xprofile.w2_wh.clean.nonempty_n}{2{,}620}
\defmmnum{xprofile.w2_wh.clean.nonempty_utt_median_wer}{587.50}
\defmmnum{xprofile.w2_wh.clean.nonempty_wer}{475.17}
\defmmnum{xprofile.w2_wh.clean.sub_pct}{83.25}
\defmmnum{xprofile.w2_wh.clean.sub_words}{43{,}772}
\defmmnum{xprofile.w2_wh.clean.utt_eq0_n}{0}
\defmmnum{xprofile.w2_wh.clean.utt_eq0_pct}{0.00}
\defmmnum{xprofile.w2_wh.clean.utt_ge100_n}{2{,}581}
\defmmnum{xprofile.w2_wh.clean.utt_ge100_pct}{98.51}
\defmmnum{xprofile.w2_wh.clean.utt_lt10_n}{0}
\defmmnum{xprofile.w2_wh.clean.utt_lt10_pct}{0.00}
\defmmnum{xprofile.w2_wh.clean.utt_lt50_n}{0}
\defmmnum{xprofile.w2_wh.clean.utt_lt50_pct}{0.00}
\defmmnum{xprofile.w2_wh.clean.utt_median_wer}{587.50}
\defmmnum{xprofile.w2_wh.clean.wer}{475.17}
\defmmnum{xprofile.w2_wh.clean.word_ratio_corpus}{4.88}
\defmmnum{xprofile.w2_wh.clean.word_ratio_median}{6.06}
\defmmnum{xprofile.w2_wh.other.cap159_n}{2{,}855}
\defmmnum{xprofile.w2_wh.other.cap159_pct}{97.14}
\defmmnum{xprofile.w2_wh.other.cap160_n}{2{,}567}
\defmmnum{xprofile.w2_wh.other.cap160_pct}{87.34}
\defmmnum{xprofile.w2_wh.other.del_pct}{1.74}
\defmmnum{xprofile.w2_wh.other.del_words}{910}
\defmmnum{xprofile.w2_wh.other.empty_pct}{0.00}
\defmmnum{xprofile.w2_wh.other.ins_pct}{449.75}
\defmmnum{xprofile.w2_wh.other.ins_words}{235{,}413}
\defmmnum{xprofile.w2_wh.other.n_empty}{0}
\defmmnum{xprofile.w2_wh.other.n_utt}{2{,}939}
\defmmnum{xprofile.w2_wh.other.nonempty_n}{2{,}939}
\defmmnum{xprofile.w2_wh.other.nonempty_utt_median_wer}{676.47}
\defmmnum{xprofile.w2_wh.other.nonempty_wer}{534.63}
\defmmnum{xprofile.w2_wh.other.sub_pct}{83.14}
\defmmnum{xprofile.w2_wh.other.sub_words}{43{,}516}
\defmmnum{xprofile.w2_wh.other.utt_eq0_n}{0}
\defmmnum{xprofile.w2_wh.other.utt_eq0_pct}{0.00}
\defmmnum{xprofile.w2_wh.other.utt_ge100_n}{2{,}909}
\defmmnum{xprofile.w2_wh.other.utt_ge100_pct}{98.98}
\defmmnum{xprofile.w2_wh.other.utt_lt10_n}{0}
\defmmnum{xprofile.w2_wh.other.utt_lt10_pct}{0.00}
\defmmnum{xprofile.w2_wh.other.utt_lt50_n}{0}
\defmmnum{xprofile.w2_wh.other.utt_lt50_pct}{0.00}
\defmmnum{xprofile.w2_wh.other.utt_median_wer}{676.47}
\defmmnum{xprofile.w2_wh.other.wer}{534.63}
\defmmnum{xprofile.w2_wh.other.word_ratio_corpus}{5.48}
\defmmnum{xprofile.w2_wh.other.word_ratio_median}{6.93}
\defmmnum{xprofile.w3_ms.clean.cap159_n}{1{,}294}
\defmmnum{xprofile.w3_ms.clean.cap159_pct}{49.39}
\defmmnum{xprofile.w3_ms.clean.cap160_n}{1{,}200}
\defmmnum{xprofile.w3_ms.clean.cap160_pct}{45.80}
\defmmnum{xprofile.w3_ms.clean.del_pct}{0.30}
\defmmnum{xprofile.w3_ms.clean.del_words}{156}
\defmmnum{xprofile.w3_ms.clean.empty_pct}{0.00}
\defmmnum{xprofile.w3_ms.clean.ins_pct}{376.21}
\defmmnum{xprofile.w3_ms.clean.ins_words}{197{,}796}
\defmmnum{xprofile.w3_ms.clean.n_empty}{0}
\defmmnum{xprofile.w3_ms.clean.n_utt}{2{,}620}
\defmmnum{xprofile.w3_ms.clean.nonempty_n}{2{,}620}
\defmmnum{xprofile.w3_ms.clean.nonempty_utt_median_wer}{561.11}
\defmmnum{xprofile.w3_ms.clean.nonempty_wer}{460.44}
\defmmnum{xprofile.w3_ms.clean.sub_pct}{83.94}
\defmmnum{xprofile.w3_ms.clean.sub_words}{44{,}131}
\defmmnum{xprofile.w3_ms.clean.utt_eq0_n}{0}
\defmmnum{xprofile.w3_ms.clean.utt_eq0_pct}{0.00}
\defmmnum{xprofile.w3_ms.clean.utt_ge100_n}{2{,}613}
\defmmnum{xprofile.w3_ms.clean.utt_ge100_pct}{99.73}
\defmmnum{xprofile.w3_ms.clean.utt_lt10_n}{0}
\defmmnum{xprofile.w3_ms.clean.utt_lt10_pct}{0.00}
\defmmnum{xprofile.w3_ms.clean.utt_lt50_n}{0}
\defmmnum{xprofile.w3_ms.clean.utt_lt50_pct}{0.00}
\defmmnum{xprofile.w3_ms.clean.utt_median_wer}{561.11}
\defmmnum{xprofile.w3_ms.clean.wer}{460.44}
\defmmnum{xprofile.w3_ms.clean.word_ratio_corpus}{4.76}
\defmmnum{xprofile.w3_ms.clean.word_ratio_median}{5.81}
\defmmnum{xprofile.w3_ms.other.cap159_n}{1{,}300}
\defmmnum{xprofile.w3_ms.other.cap159_pct}{44.23}
\defmmnum{xprofile.w3_ms.other.cap160_n}{1{,}216}
\defmmnum{xprofile.w3_ms.other.cap160_pct}{41.37}
\defmmnum{xprofile.w3_ms.other.del_pct}{0.28}
\defmmnum{xprofile.w3_ms.other.del_words}{145}
\defmmnum{xprofile.w3_ms.other.empty_pct}{0.00}
\defmmnum{xprofile.w3_ms.other.ins_pct}{430.54}
\defmmnum{xprofile.w3_ms.other.ins_words}{225{,}357}
\defmmnum{xprofile.w3_ms.other.n_empty}{0}
\defmmnum{xprofile.w3_ms.other.n_utt}{2{,}939}
\defmmnum{xprofile.w3_ms.other.nonempty_n}{2{,}939}
\defmmnum{xprofile.w3_ms.other.nonempty_utt_median_wer}{635.29}
\defmmnum{xprofile.w3_ms.other.nonempty_wer}{513.95}
\defmmnum{xprofile.w3_ms.other.sub_pct}{83.13}
\defmmnum{xprofile.w3_ms.other.sub_words}{43{,}513}
\defmmnum{xprofile.w3_ms.other.utt_eq0_n}{0}
\defmmnum{xprofile.w3_ms.other.utt_eq0_pct}{0.00}
\defmmnum{xprofile.w3_ms.other.utt_ge100_n}{2{,}931}
\defmmnum{xprofile.w3_ms.other.utt_ge100_pct}{99.73}
\defmmnum{xprofile.w3_ms.other.utt_lt10_n}{0}
\defmmnum{xprofile.w3_ms.other.utt_lt10_pct}{0.00}
\defmmnum{xprofile.w3_ms.other.utt_lt50_n}{0}
\defmmnum{xprofile.w3_ms.other.utt_lt50_pct}{0.00}
\defmmnum{xprofile.w3_ms.other.utt_median_wer}{635.29}
\defmmnum{xprofile.w3_ms.other.wer}{513.95}
\defmmnum{xprofile.w3_ms.other.word_ratio_corpus}{5.30}
\defmmnum{xprofile.w3_ms.other.word_ratio_median}{6.53}
\defmmnum{xprofile.w3_wh.clean.cap159_n}{970}
\defmmnum{xprofile.w3_wh.clean.cap159_pct}{37.02}
\defmmnum{xprofile.w3_wh.clean.cap160_n}{881}
\defmmnum{xprofile.w3_wh.clean.cap160_pct}{33.63}
\defmmnum{xprofile.w3_wh.clean.del_pct}{1.03}
\defmmnum{xprofile.w3_wh.clean.del_words}{544}
\defmmnum{xprofile.w3_wh.clean.empty_pct}{0.00}
\defmmnum{xprofile.w3_wh.clean.ins_pct}{351.71}
\defmmnum{xprofile.w3_wh.clean.ins_words}{184{,}916}
\defmmnum{xprofile.w3_wh.clean.n_empty}{0}
\defmmnum{xprofile.w3_wh.clean.n_utt}{2{,}620}
\defmmnum{xprofile.w3_wh.clean.nonempty_n}{2{,}620}
\defmmnum{xprofile.w3_wh.clean.nonempty_utt_median_wer}{534.17}
\defmmnum{xprofile.w3_wh.clean.nonempty_wer}{436.11}
\defmmnum{xprofile.w3_wh.clean.sub_pct}{83.36}
\defmmnum{xprofile.w3_wh.clean.sub_words}{43{,}828}
\defmmnum{xprofile.w3_wh.clean.utt_eq0_n}{0}
\defmmnum{xprofile.w3_wh.clean.utt_eq0_pct}{0.00}
\defmmnum{xprofile.w3_wh.clean.utt_ge100_n}{2{,}597}
\defmmnum{xprofile.w3_wh.clean.utt_ge100_pct}{99.12}
\defmmnum{xprofile.w3_wh.clean.utt_lt10_n}{0}
\defmmnum{xprofile.w3_wh.clean.utt_lt10_pct}{0.00}
\defmmnum{xprofile.w3_wh.clean.utt_lt50_n}{0}
\defmmnum{xprofile.w3_wh.clean.utt_lt50_pct}{0.00}
\defmmnum{xprofile.w3_wh.clean.utt_median_wer}{534.17}
\defmmnum{xprofile.w3_wh.clean.wer}{436.11}
\defmmnum{xprofile.w3_wh.clean.word_ratio_corpus}{4.51}
\defmmnum{xprofile.w3_wh.clean.word_ratio_median}{5.53}
\defmmnum{xprofile.w3_wh.other.cap159_n}{965}
\defmmnum{xprofile.w3_wh.other.cap159_pct}{32.83}
\defmmnum{xprofile.w3_wh.other.cap160_n}{888}
\defmmnum{xprofile.w3_wh.other.cap160_pct}{30.21}
\defmmnum{xprofile.w3_wh.other.del_pct}{0.97}
\defmmnum{xprofile.w3_wh.other.del_words}{508}
\defmmnum{xprofile.w3_wh.other.empty_pct}{0.00}
\defmmnum{xprofile.w3_wh.other.ins_pct}{402.14}
\defmmnum{xprofile.w3_wh.other.ins_words}{210{,}493}
\defmmnum{xprofile.w3_wh.other.n_empty}{0}
\defmmnum{xprofile.w3_wh.other.n_utt}{2{,}939}
\defmmnum{xprofile.w3_wh.other.nonempty_n}{2{,}939}
\defmmnum{xprofile.w3_wh.other.nonempty_utt_median_wer}{590.91}
\defmmnum{xprofile.w3_wh.other.nonempty_wer}{486.19}
\defmmnum{xprofile.w3_wh.other.sub_pct}{83.08}
\defmmnum{xprofile.w3_wh.other.sub_words}{43{,}487}
\defmmnum{xprofile.w3_wh.other.utt_eq0_n}{0}
\defmmnum{xprofile.w3_wh.other.utt_eq0_pct}{0.00}
\defmmnum{xprofile.w3_wh.other.utt_ge100_n}{2{,}920}
\defmmnum{xprofile.w3_wh.other.utt_ge100_pct}{99.35}
\defmmnum{xprofile.w3_wh.other.utt_lt10_n}{0}
\defmmnum{xprofile.w3_wh.other.utt_lt10_pct}{0.00}
\defmmnum{xprofile.w3_wh.other.utt_lt50_n}{0}
\defmmnum{xprofile.w3_wh.other.utt_lt50_pct}{0.00}
\defmmnum{xprofile.w3_wh.other.utt_median_wer}{590.91}
\defmmnum{xprofile.w3_wh.other.wer}{486.19}
\defmmnum{xprofile.w3_wh.other.word_ratio_corpus}{5.01}
\defmmnum{xprofile.w3_wh.other.word_ratio_median}{6.08}
\defmmnum{xprofile.ws4_ms_g2_g3.clean.cap159_n}{2{,}438}
\defmmnum{xprofile.ws4_ms_g2_g3.clean.cap159_pct}{93.05}
\defmmnum{xprofile.ws4_ms_g2_g3.clean.cap160_n}{2{,}409}
\defmmnum{xprofile.ws4_ms_g2_g3.clean.cap160_pct}{91.95}
\defmmnum{xprofile.ws4_ms_g2_g3.clean.del_pct}{0.99}
\defmmnum{xprofile.ws4_ms_g2_g3.clean.del_words}{520}
\defmmnum{xprofile.ws4_ms_g2_g3.clean.empty_pct}{0.00}
\defmmnum{xprofile.ws4_ms_g2_g3.clean.ins_pct}{302.06}
\defmmnum{xprofile.ws4_ms_g2_g3.clean.ins_words}{158{,}810}
\defmmnum{xprofile.ws4_ms_g2_g3.clean.n_empty}{0}
\defmmnum{xprofile.ws4_ms_g2_g3.clean.n_utt}{2{,}620}
\defmmnum{xprofile.ws4_ms_g2_g3.clean.nonempty_n}{2{,}620}
\defmmnum{xprofile.ws4_ms_g2_g3.clean.nonempty_utt_median_wer}{466.67}
\defmmnum{xprofile.ws4_ms_g2_g3.clean.nonempty_wer}{394.50}
\defmmnum{xprofile.ws4_ms_g2_g3.clean.sub_pct}{91.46}
\defmmnum{xprofile.ws4_ms_g2_g3.clean.sub_words}{48{,}084}
\defmmnum{xprofile.ws4_ms_g2_g3.clean.utt_eq0_n}{0}
\defmmnum{xprofile.ws4_ms_g2_g3.clean.utt_eq0_pct}{0.00}
\defmmnum{xprofile.ws4_ms_g2_g3.clean.utt_ge100_n}{2{,}594}
\defmmnum{xprofile.ws4_ms_g2_g3.clean.utt_ge100_pct}{99.01}
\defmmnum{xprofile.ws4_ms_g2_g3.clean.utt_lt10_n}{0}
\defmmnum{xprofile.ws4_ms_g2_g3.clean.utt_lt10_pct}{0.00}
\defmmnum{xprofile.ws4_ms_g2_g3.clean.utt_lt50_n}{0}
\defmmnum{xprofile.ws4_ms_g2_g3.clean.utt_lt50_pct}{0.00}
\defmmnum{xprofile.ws4_ms_g2_g3.clean.utt_median_wer}{466.67}
\defmmnum{xprofile.ws4_ms_g2_g3.clean.wer}{394.50}
\defmmnum{xprofile.ws4_ms_g2_g3.clean.word_ratio_corpus}{4.01}
\defmmnum{xprofile.ws4_ms_g2_g3.clean.word_ratio_median}{4.76}
\defmmnum{xprofile.ws4_ms_g2_g3.other.cap159_n}{2{,}719}
\defmmnum{xprofile.ws4_ms_g2_g3.other.cap159_pct}{92.51}
\defmmnum{xprofile.ws4_ms_g2_g3.other.cap160_n}{2{,}673}
\defmmnum{xprofile.ws4_ms_g2_g3.other.cap160_pct}{90.95}
\defmmnum{xprofile.ws4_ms_g2_g3.other.del_pct}{0.91}
\defmmnum{xprofile.ws4_ms_g2_g3.other.del_words}{477}
\defmmnum{xprofile.ws4_ms_g2_g3.other.empty_pct}{0.00}
\defmmnum{xprofile.ws4_ms_g2_g3.other.ins_pct}{351.18}
\defmmnum{xprofile.ws4_ms_g2_g3.other.ins_words}{183{,}816}
\defmmnum{xprofile.ws4_ms_g2_g3.other.n_empty}{0}
\defmmnum{xprofile.ws4_ms_g2_g3.other.n_utt}{2{,}939}
\defmmnum{xprofile.ws4_ms_g2_g3.other.nonempty_n}{2{,}939}
\defmmnum{xprofile.ws4_ms_g2_g3.other.nonempty_utt_median_wer}{527.59}
\defmmnum{xprofile.ws4_ms_g2_g3.other.nonempty_wer}{443.33}
\defmmnum{xprofile.ws4_ms_g2_g3.other.sub_pct}{91.24}
\defmmnum{xprofile.ws4_ms_g2_g3.other.sub_words}{47{,}758}
\defmmnum{xprofile.ws4_ms_g2_g3.other.utt_eq0_n}{0}
\defmmnum{xprofile.ws4_ms_g2_g3.other.utt_eq0_pct}{0.00}
\defmmnum{xprofile.ws4_ms_g2_g3.other.utt_ge100_n}{2{,}919}
\defmmnum{xprofile.ws4_ms_g2_g3.other.utt_ge100_pct}{99.32}
\defmmnum{xprofile.ws4_ms_g2_g3.other.utt_lt10_n}{0}
\defmmnum{xprofile.ws4_ms_g2_g3.other.utt_lt10_pct}{0.00}
\defmmnum{xprofile.ws4_ms_g2_g3.other.utt_lt50_n}{0}
\defmmnum{xprofile.ws4_ms_g2_g3.other.utt_lt50_pct}{0.00}
\defmmnum{xprofile.ws4_ms_g2_g3.other.utt_median_wer}{527.59}
\defmmnum{xprofile.ws4_ms_g2_g3.other.wer}{443.33}
\defmmnum{xprofile.ws4_ms_g2_g3.other.word_ratio_corpus}{4.50}
\defmmnum{xprofile.ws4_ms_g2_g3.other.word_ratio_median}{5.38}
\defmmnum{xprofile.ws4_ms_g2_qwen_ln.clean.cap159_n}{0}
\defmmnum{xprofile.ws4_ms_g2_qwen_ln.clean.cap159_pct}{0.00}
\defmmnum{xprofile.ws4_ms_g2_qwen_ln.clean.cap160_n}{0}
\defmmnum{xprofile.ws4_ms_g2_qwen_ln.clean.cap160_pct}{0.00}
\defmmnum{xprofile.ws4_ms_g2_qwen_ln.clean.del_pct}{60.88}
\defmmnum{xprofile.ws4_ms_g2_qwen_ln.clean.del_words}{32{,}010}
\defmmnum{xprofile.ws4_ms_g2_qwen_ln.clean.empty_pct}{0.00}
\defmmnum{xprofile.ws4_ms_g2_qwen_ln.clean.ins_pct}{3.09}
\defmmnum{xprofile.ws4_ms_g2_qwen_ln.clean.ins_words}{1{,}624}
\defmmnum{xprofile.ws4_ms_g2_qwen_ln.clean.n_empty}{0}
\defmmnum{xprofile.ws4_ms_g2_qwen_ln.clean.n_utt}{2{,}620}
\defmmnum{xprofile.ws4_ms_g2_qwen_ln.clean.nonempty_n}{2{,}620}
\defmmnum{xprofile.ws4_ms_g2_qwen_ln.clean.nonempty_utt_median_wer}{100.00}
\defmmnum{xprofile.ws4_ms_g2_qwen_ln.clean.nonempty_wer}{100.70}
\defmmnum{xprofile.ws4_ms_g2_qwen_ln.clean.sub_pct}{36.73}
\defmmnum{xprofile.ws4_ms_g2_qwen_ln.clean.sub_words}{19{,}311}
\defmmnum{xprofile.ws4_ms_g2_qwen_ln.clean.utt_eq0_n}{0}
\defmmnum{xprofile.ws4_ms_g2_qwen_ln.clean.utt_eq0_pct}{0.00}
\defmmnum{xprofile.ws4_ms_g2_qwen_ln.clean.utt_ge100_n}{1{,}757}
\defmmnum{xprofile.ws4_ms_g2_qwen_ln.clean.utt_ge100_pct}{67.06}
\defmmnum{xprofile.ws4_ms_g2_qwen_ln.clean.utt_lt10_n}{0}
\defmmnum{xprofile.ws4_ms_g2_qwen_ln.clean.utt_lt10_pct}{0.00}
\defmmnum{xprofile.ws4_ms_g2_qwen_ln.clean.utt_lt50_n}{0}
\defmmnum{xprofile.ws4_ms_g2_qwen_ln.clean.utt_lt50_pct}{0.00}
\defmmnum{xprofile.ws4_ms_g2_qwen_ln.clean.utt_median_wer}{100.00}
\defmmnum{xprofile.ws4_ms_g2_qwen_ln.clean.wer}{100.70}
\defmmnum{xprofile.ws4_ms_g2_qwen_ln.clean.word_ratio_corpus}{0.42}
\defmmnum{xprofile.ws4_ms_g2_qwen_ln.clean.word_ratio_median}{0.50}
\defmmnum{xprofile.ws4_ms_g2_qwen_ln.other.cap159_n}{0}
\defmmnum{xprofile.ws4_ms_g2_qwen_ln.other.cap159_pct}{0.00}
\defmmnum{xprofile.ws4_ms_g2_qwen_ln.other.cap160_n}{0}
\defmmnum{xprofile.ws4_ms_g2_qwen_ln.other.cap160_pct}{0.00}
\defmmnum{xprofile.ws4_ms_g2_qwen_ln.other.del_pct}{56.65}
\defmmnum{xprofile.ws4_ms_g2_qwen_ln.other.del_words}{29{,}650}
\defmmnum{xprofile.ws4_ms_g2_qwen_ln.other.empty_pct}{0.00}
\defmmnum{xprofile.ws4_ms_g2_qwen_ln.other.ins_pct}{4.71}
\defmmnum{xprofile.ws4_ms_g2_qwen_ln.other.ins_words}{2{,}464}
\defmmnum{xprofile.ws4_ms_g2_qwen_ln.other.n_empty}{0}
\defmmnum{xprofile.ws4_ms_g2_qwen_ln.other.n_utt}{2{,}939}
\defmmnum{xprofile.ws4_ms_g2_qwen_ln.other.nonempty_n}{2{,}939}
\defmmnum{xprofile.ws4_ms_g2_qwen_ln.other.nonempty_utt_median_wer}{100.00}
\defmmnum{xprofile.ws4_ms_g2_qwen_ln.other.nonempty_wer}{102.22}
\defmmnum{xprofile.ws4_ms_g2_qwen_ln.other.sub_pct}{40.87}
\defmmnum{xprofile.ws4_ms_g2_qwen_ln.other.sub_words}{21{,}390}
\defmmnum{xprofile.ws4_ms_g2_qwen_ln.other.utt_eq0_n}{0}
\defmmnum{xprofile.ws4_ms_g2_qwen_ln.other.utt_eq0_pct}{0.00}
\defmmnum{xprofile.ws4_ms_g2_qwen_ln.other.utt_ge100_n}{2{,}069}
\defmmnum{xprofile.ws4_ms_g2_qwen_ln.other.utt_ge100_pct}{70.40}
\defmmnum{xprofile.ws4_ms_g2_qwen_ln.other.utt_lt10_n}{0}
\defmmnum{xprofile.ws4_ms_g2_qwen_ln.other.utt_lt10_pct}{0.00}
\defmmnum{xprofile.ws4_ms_g2_qwen_ln.other.utt_lt50_n}{0}
\defmmnum{xprofile.ws4_ms_g2_qwen_ln.other.utt_lt50_pct}{0.00}
\defmmnum{xprofile.ws4_ms_g2_qwen_ln.other.utt_median_wer}{100.00}
\defmmnum{xprofile.ws4_ms_g2_qwen_ln.other.wer}{102.22}
\defmmnum{xprofile.ws4_ms_g2_qwen_ln.other.word_ratio_corpus}{0.48}
\defmmnum{xprofile.ws4_ms_g2_qwen_ln.other.word_ratio_median}{0.57}
\defmmnum{xprofile.ws4_ms_g2_qwen_sc.clean.cap159_n}{24}
\defmmnum{xprofile.ws4_ms_g2_qwen_sc.clean.cap159_pct}{0.92}
\defmmnum{xprofile.ws4_ms_g2_qwen_sc.clean.cap160_n}{24}
\defmmnum{xprofile.ws4_ms_g2_qwen_sc.clean.cap160_pct}{0.92}
\defmmnum{xprofile.ws4_ms_g2_qwen_sc.clean.del_pct}{53.20}
\defmmnum{xprofile.ws4_ms_g2_qwen_sc.clean.del_words}{27{,}972}
\defmmnum{xprofile.ws4_ms_g2_qwen_sc.clean.empty_pct}{0.00}
\defmmnum{xprofile.ws4_ms_g2_qwen_sc.clean.ins_pct}{19.29}
\defmmnum{xprofile.ws4_ms_g2_qwen_sc.clean.ins_words}{10{,}140}
\defmmnum{xprofile.ws4_ms_g2_qwen_sc.clean.n_empty}{0}
\defmmnum{xprofile.ws4_ms_g2_qwen_sc.clean.n_utt}{2{,}620}
\defmmnum{xprofile.ws4_ms_g2_qwen_sc.clean.nonempty_n}{2{,}620}
\defmmnum{xprofile.ws4_ms_g2_qwen_sc.clean.nonempty_utt_median_wer}{100.00}
\defmmnum{xprofile.ws4_ms_g2_qwen_sc.clean.nonempty_wer}{116.28}
\defmmnum{xprofile.ws4_ms_g2_qwen_sc.clean.sub_pct}{43.80}
\defmmnum{xprofile.ws4_ms_g2_qwen_sc.clean.sub_words}{23{,}026}
\defmmnum{xprofile.ws4_ms_g2_qwen_sc.clean.utt_eq0_n}{0}
\defmmnum{xprofile.ws4_ms_g2_qwen_sc.clean.utt_eq0_pct}{0.00}
\defmmnum{xprofile.ws4_ms_g2_qwen_sc.clean.utt_ge100_n}{1{,}861}
\defmmnum{xprofile.ws4_ms_g2_qwen_sc.clean.utt_ge100_pct}{71.03}
\defmmnum{xprofile.ws4_ms_g2_qwen_sc.clean.utt_lt10_n}{0}
\defmmnum{xprofile.ws4_ms_g2_qwen_sc.clean.utt_lt10_pct}{0.00}
\defmmnum{xprofile.ws4_ms_g2_qwen_sc.clean.utt_lt50_n}{0}
\defmmnum{xprofile.ws4_ms_g2_qwen_sc.clean.utt_lt50_pct}{0.00}
\defmmnum{xprofile.ws4_ms_g2_qwen_sc.clean.utt_median_wer}{100.00}
\defmmnum{xprofile.ws4_ms_g2_qwen_sc.clean.wer}{116.28}
\defmmnum{xprofile.ws4_ms_g2_qwen_sc.clean.word_ratio_corpus}{0.66}
\defmmnum{xprofile.ws4_ms_g2_qwen_sc.clean.word_ratio_median}{0.57}
\defmmnum{xprofile.ws4_ms_g2_qwen_sc.other.cap159_n}{8}
\defmmnum{xprofile.ws4_ms_g2_qwen_sc.other.cap159_pct}{0.27}
\defmmnum{xprofile.ws4_ms_g2_qwen_sc.other.cap160_n}{8}
\defmmnum{xprofile.ws4_ms_g2_qwen_sc.other.cap160_pct}{0.27}
\defmmnum{xprofile.ws4_ms_g2_qwen_sc.other.del_pct}{50.33}
\defmmnum{xprofile.ws4_ms_g2_qwen_sc.other.del_words}{26{,}344}
\defmmnum{xprofile.ws4_ms_g2_qwen_sc.other.empty_pct}{0.00}
\defmmnum{xprofile.ws4_ms_g2_qwen_sc.other.ins_pct}{17.47}
\defmmnum{xprofile.ws4_ms_g2_qwen_sc.other.ins_words}{9{,}143}
\defmmnum{xprofile.ws4_ms_g2_qwen_sc.other.n_empty}{0}
\defmmnum{xprofile.ws4_ms_g2_qwen_sc.other.n_utt}{2{,}939}
\defmmnum{xprofile.ws4_ms_g2_qwen_sc.other.nonempty_n}{2{,}939}
\defmmnum{xprofile.ws4_ms_g2_qwen_sc.other.nonempty_utt_median_wer}{100.00}
\defmmnum{xprofile.ws4_ms_g2_qwen_sc.other.nonempty_wer}{114.48}
\defmmnum{xprofile.ws4_ms_g2_qwen_sc.other.sub_pct}{46.68}
\defmmnum{xprofile.ws4_ms_g2_qwen_sc.other.sub_words}{24{,}435}
\defmmnum{xprofile.ws4_ms_g2_qwen_sc.other.utt_eq0_n}{0}
\defmmnum{xprofile.ws4_ms_g2_qwen_sc.other.utt_eq0_pct}{0.00}
\defmmnum{xprofile.ws4_ms_g2_qwen_sc.other.utt_ge100_n}{2{,}149}
\defmmnum{xprofile.ws4_ms_g2_qwen_sc.other.utt_ge100_pct}{73.12}
\defmmnum{xprofile.ws4_ms_g2_qwen_sc.other.utt_lt10_n}{0}
\defmmnum{xprofile.ws4_ms_g2_qwen_sc.other.utt_lt10_pct}{0.00}
\defmmnum{xprofile.ws4_ms_g2_qwen_sc.other.utt_lt50_n}{0}
\defmmnum{xprofile.ws4_ms_g2_qwen_sc.other.utt_lt50_pct}{0.00}
\defmmnum{xprofile.ws4_ms_g2_qwen_sc.other.utt_median_wer}{100.00}
\defmmnum{xprofile.ws4_ms_g2_qwen_sc.other.wer}{114.48}
\defmmnum{xprofile.ws4_ms_g2_qwen_sc.other.word_ratio_corpus}{0.67}
\defmmnum{xprofile.ws4_ms_g2_qwen_sc.other.word_ratio_median}{0.63}
\defmmnum{xprofile.ws4_ms_qwen_g2.clean.cap159_n}{999}
\defmmnum{xprofile.ws4_ms_qwen_g2.clean.cap159_pct}{38.13}
\defmmnum{xprofile.ws4_ms_qwen_g2.clean.cap160_n}{930}
\defmmnum{xprofile.ws4_ms_qwen_g2.clean.cap160_pct}{35.50}
\defmmnum{xprofile.ws4_ms_qwen_g2.clean.del_pct}{4.02}
\defmmnum{xprofile.ws4_ms_qwen_g2.clean.del_words}{2{,}114}
\defmmnum{xprofile.ws4_ms_qwen_g2.clean.empty_pct}{0.38}
\defmmnum{xprofile.ws4_ms_qwen_g2.clean.ins_pct}{255.33}
\defmmnum{xprofile.ws4_ms_qwen_g2.clean.ins_words}{134{,}244}
\defmmnum{xprofile.ws4_ms_qwen_g2.clean.n_empty}{10}
\defmmnum{xprofile.ws4_ms_qwen_g2.clean.n_utt}{2{,}620}
\defmmnum{xprofile.ws4_ms_qwen_g2.clean.nonempty_n}{2{,}610}
\defmmnum{xprofile.ws4_ms_qwen_g2.clean.nonempty_utt_median_wer}{366.67}
\defmmnum{xprofile.ws4_ms_qwen_g2.clean.nonempty_wer}{342.43}
\defmmnum{xprofile.ws4_ms_qwen_g2.clean.sub_pct}{82.50}
\defmmnum{xprofile.ws4_ms_qwen_g2.clean.sub_words}{43{,}374}
\defmmnum{xprofile.ws4_ms_qwen_g2.clean.utt_eq0_n}{0}
\defmmnum{xprofile.ws4_ms_qwen_g2.clean.utt_eq0_pct}{0.00}
\defmmnum{xprofile.ws4_ms_qwen_g2.clean.utt_ge100_n}{2{,}487}
\defmmnum{xprofile.ws4_ms_qwen_g2.clean.utt_ge100_pct}{94.92}
\defmmnum{xprofile.ws4_ms_qwen_g2.clean.utt_lt10_n}{0}
\defmmnum{xprofile.ws4_ms_qwen_g2.clean.utt_lt10_pct}{0.00}
\defmmnum{xprofile.ws4_ms_qwen_g2.clean.utt_lt50_n}{0}
\defmmnum{xprofile.ws4_ms_qwen_g2.clean.utt_lt50_pct}{0.00}
\defmmnum{xprofile.ws4_ms_qwen_g2.clean.utt_median_wer}{366.67}
\defmmnum{xprofile.ws4_ms_qwen_g2.clean.wer}{341.85}
\defmmnum{xprofile.ws4_ms_qwen_g2.clean.word_ratio_corpus}{3.51}
\defmmnum{xprofile.ws4_ms_qwen_g2.clean.word_ratio_median}{3.81}
\defmmnum{xprofile.ws4_ms_qwen_g2.other.cap159_n}{1{,}083}
\defmmnum{xprofile.ws4_ms_qwen_g2.other.cap159_pct}{36.85}
\defmmnum{xprofile.ws4_ms_qwen_g2.other.cap160_n}{1{,}025}
\defmmnum{xprofile.ws4_ms_qwen_g2.other.cap160_pct}{34.88}
\defmmnum{xprofile.ws4_ms_qwen_g2.other.del_pct}{3.41}
\defmmnum{xprofile.ws4_ms_qwen_g2.other.del_words}{1{,}783}
\defmmnum{xprofile.ws4_ms_qwen_g2.other.empty_pct}{0.14}
\defmmnum{xprofile.ws4_ms_qwen_g2.other.ins_pct}{304.17}
\defmmnum{xprofile.ws4_ms_qwen_g2.other.ins_words}{159{,}212}
\defmmnum{xprofile.ws4_ms_qwen_g2.other.n_empty}{4}
\defmmnum{xprofile.ws4_ms_qwen_g2.other.n_utt}{2{,}939}
\defmmnum{xprofile.ws4_ms_qwen_g2.other.nonempty_n}{2{,}935}
\defmmnum{xprofile.ws4_ms_qwen_g2.other.nonempty_utt_median_wer}{420.83}
\defmmnum{xprofile.ws4_ms_qwen_g2.other.nonempty_wer}{390.13}
\defmmnum{xprofile.ws4_ms_qwen_g2.other.sub_pct}{82.44}
\defmmnum{xprofile.ws4_ms_qwen_g2.other.sub_words}{43{,}149}
\defmmnum{xprofile.ws4_ms_qwen_g2.other.utt_eq0_n}{0}
\defmmnum{xprofile.ws4_ms_qwen_g2.other.utt_eq0_pct}{0.00}
\defmmnum{xprofile.ws4_ms_qwen_g2.other.utt_ge100_n}{2{,}824}
\defmmnum{xprofile.ws4_ms_qwen_g2.other.utt_ge100_pct}{96.09}
\defmmnum{xprofile.ws4_ms_qwen_g2.other.utt_lt10_n}{0}
\defmmnum{xprofile.ws4_ms_qwen_g2.other.utt_lt10_pct}{0.00}
\defmmnum{xprofile.ws4_ms_qwen_g2.other.utt_lt50_n}{0}
\defmmnum{xprofile.ws4_ms_qwen_g2.other.utt_lt50_pct}{0.00}
\defmmnum{xprofile.ws4_ms_qwen_g2.other.utt_median_wer}{420.00}
\defmmnum{xprofile.ws4_ms_qwen_g2.other.wer}{390.01}
\defmmnum{xprofile.ws4_ms_qwen_g2.other.word_ratio_corpus}{4.01}
\defmmnum{xprofile.ws4_ms_qwen_g2.other.word_ratio_median}{4.38}
\defmmnum{xprofile.ws4_wh_qwen_g2.clean.cap159_n}{1{,}949}
\defmmnum{xprofile.ws4_wh_qwen_g2.clean.cap159_pct}{74.39}
\defmmnum{xprofile.ws4_wh_qwen_g2.clean.cap160_n}{1{,}817}
\defmmnum{xprofile.ws4_wh_qwen_g2.clean.cap160_pct}{69.35}
\defmmnum{xprofile.ws4_wh_qwen_g2.clean.del_pct}{0.97}
\defmmnum{xprofile.ws4_wh_qwen_g2.clean.del_words}{511}
\defmmnum{xprofile.ws4_wh_qwen_g2.clean.empty_pct}{0.00}
\defmmnum{xprofile.ws4_wh_qwen_g2.clean.ins_pct}{392.91}
\defmmnum{xprofile.ws4_wh_qwen_g2.clean.ins_words}{206{,}576}
\defmmnum{xprofile.ws4_wh_qwen_g2.clean.n_empty}{0}
\defmmnum{xprofile.ws4_wh_qwen_g2.clean.n_utt}{2{,}620}
\defmmnum{xprofile.ws4_wh_qwen_g2.clean.nonempty_n}{2{,}620}
\defmmnum{xprofile.ws4_wh_qwen_g2.clean.nonempty_utt_median_wer}{600.00}
\defmmnum{xprofile.ws4_wh_qwen_g2.clean.nonempty_wer}{477.13}
\defmmnum{xprofile.ws4_wh_qwen_g2.clean.sub_pct}{83.25}
\defmmnum{xprofile.ws4_wh_qwen_g2.clean.sub_words}{43{,}768}
\defmmnum{xprofile.ws4_wh_qwen_g2.clean.utt_eq0_n}{0}
\defmmnum{xprofile.ws4_wh_qwen_g2.clean.utt_eq0_pct}{0.00}
\defmmnum{xprofile.ws4_wh_qwen_g2.clean.utt_ge100_n}{2{,}588}
\defmmnum{xprofile.ws4_wh_qwen_g2.clean.utt_ge100_pct}{98.78}
\defmmnum{xprofile.ws4_wh_qwen_g2.clean.utt_lt10_n}{0}
\defmmnum{xprofile.ws4_wh_qwen_g2.clean.utt_lt10_pct}{0.00}
\defmmnum{xprofile.ws4_wh_qwen_g2.clean.utt_lt50_n}{0}
\defmmnum{xprofile.ws4_wh_qwen_g2.clean.utt_lt50_pct}{0.00}
\defmmnum{xprofile.ws4_wh_qwen_g2.clean.utt_median_wer}{600.00}
\defmmnum{xprofile.ws4_wh_qwen_g2.clean.wer}{477.13}
\defmmnum{xprofile.ws4_wh_qwen_g2.clean.word_ratio_corpus}{4.92}
\defmmnum{xprofile.ws4_wh_qwen_g2.clean.word_ratio_median}{6.16}
\defmmnum{xprofile.ws4_wh_qwen_g2.other.cap159_n}{2{,}053}
\defmmnum{xprofile.ws4_wh_qwen_g2.other.cap159_pct}{69.85}
\defmmnum{xprofile.ws4_wh_qwen_g2.other.cap160_n}{1{,}897}
\defmmnum{xprofile.ws4_wh_qwen_g2.other.cap160_pct}{64.55}
\defmmnum{xprofile.ws4_wh_qwen_g2.other.del_pct}{0.92}
\defmmnum{xprofile.ws4_wh_qwen_g2.other.del_words}{482}
\defmmnum{xprofile.ws4_wh_qwen_g2.other.empty_pct}{0.00}
\defmmnum{xprofile.ws4_wh_qwen_g2.other.ins_pct}{448.45}
\defmmnum{xprofile.ws4_wh_qwen_g2.other.ins_words}{234{,}730}
\defmmnum{xprofile.ws4_wh_qwen_g2.other.n_empty}{0}
\defmmnum{xprofile.ws4_wh_qwen_g2.other.n_utt}{2{,}939}
\defmmnum{xprofile.ws4_wh_qwen_g2.other.nonempty_n}{2{,}939}
\defmmnum{xprofile.ws4_wh_qwen_g2.other.nonempty_utt_median_wer}{680.00}
\defmmnum{xprofile.ws4_wh_qwen_g2.other.nonempty_wer}{532.68}
\defmmnum{xprofile.ws4_wh_qwen_g2.other.sub_pct}{83.31}
\defmmnum{xprofile.ws4_wh_qwen_g2.other.sub_words}{43{,}608}
\defmmnum{xprofile.ws4_wh_qwen_g2.other.utt_eq0_n}{0}
\defmmnum{xprofile.ws4_wh_qwen_g2.other.utt_eq0_pct}{0.00}
\defmmnum{xprofile.ws4_wh_qwen_g2.other.utt_ge100_n}{2{,}915}
\defmmnum{xprofile.ws4_wh_qwen_g2.other.utt_ge100_pct}{99.18}
\defmmnum{xprofile.ws4_wh_qwen_g2.other.utt_lt10_n}{0}
\defmmnum{xprofile.ws4_wh_qwen_g2.other.utt_lt10_pct}{0.00}
\defmmnum{xprofile.ws4_wh_qwen_g2.other.utt_lt50_n}{0}
\defmmnum{xprofile.ws4_wh_qwen_g2.other.utt_lt50_pct}{0.00}
\defmmnum{xprofile.ws4_wh_qwen_g2.other.utt_median_wer}{680.00}
\defmmnum{xprofile.ws4_wh_qwen_g2.other.wer}{532.68}
\defmmnum{xprofile.ws4_wh_qwen_g2.other.word_ratio_corpus}{5.48}
\defmmnum{xprofile.ws4_wh_qwen_g2.other.word_ratio_median}{7.00}
\defmmnum{xprofile.ws4_wh_qwen_g3.clean.cap159_n}{2{,}272}
\defmmnum{xprofile.ws4_wh_qwen_g3.clean.cap159_pct}{86.72}
\defmmnum{xprofile.ws4_wh_qwen_g3.clean.cap160_n}{2{,}161}
\defmmnum{xprofile.ws4_wh_qwen_g3.clean.cap160_pct}{82.48}
\defmmnum{xprofile.ws4_wh_qwen_g3.clean.del_pct}{1.34}
\defmmnum{xprofile.ws4_wh_qwen_g3.clean.del_words}{704}
\defmmnum{xprofile.ws4_wh_qwen_g3.clean.empty_pct}{0.00}
\defmmnum{xprofile.ws4_wh_qwen_g3.clean.ins_pct}{353.75}
\defmmnum{xprofile.ws4_wh_qwen_g3.clean.ins_words}{185{,}988}
\defmmnum{xprofile.ws4_wh_qwen_g3.clean.n_empty}{0}
\defmmnum{xprofile.ws4_wh_qwen_g3.clean.n_utt}{2{,}620}
\defmmnum{xprofile.ws4_wh_qwen_g3.clean.nonempty_n}{2{,}620}
\defmmnum{xprofile.ws4_wh_qwen_g3.clean.nonempty_utt_median_wer}{511.44}
\defmmnum{xprofile.ws4_wh_qwen_g3.clean.nonempty_wer}{442.77}
\defmmnum{xprofile.ws4_wh_qwen_g3.clean.sub_pct}{87.68}
\defmmnum{xprofile.ws4_wh_qwen_g3.clean.sub_words}{46{,}097}
\defmmnum{xprofile.ws4_wh_qwen_g3.clean.utt_eq0_n}{0}
\defmmnum{xprofile.ws4_wh_qwen_g3.clean.utt_eq0_pct}{0.00}
\defmmnum{xprofile.ws4_wh_qwen_g3.clean.utt_ge100_n}{2{,}587}
\defmmnum{xprofile.ws4_wh_qwen_g3.clean.utt_ge100_pct}{98.74}
\defmmnum{xprofile.ws4_wh_qwen_g3.clean.utt_lt10_n}{0}
\defmmnum{xprofile.ws4_wh_qwen_g3.clean.utt_lt10_pct}{0.00}
\defmmnum{xprofile.ws4_wh_qwen_g3.clean.utt_lt50_n}{0}
\defmmnum{xprofile.ws4_wh_qwen_g3.clean.utt_lt50_pct}{0.00}
\defmmnum{xprofile.ws4_wh_qwen_g3.clean.utt_median_wer}{511.44}
\defmmnum{xprofile.ws4_wh_qwen_g3.clean.wer}{442.77}
\defmmnum{xprofile.ws4_wh_qwen_g3.clean.word_ratio_corpus}{4.52}
\defmmnum{xprofile.ws4_wh_qwen_g3.clean.word_ratio_median}{5.25}
\defmmnum{xprofile.ws4_wh_qwen_g3.other.cap159_n}{2{,}480}
\defmmnum{xprofile.ws4_wh_qwen_g3.other.cap159_pct}{84.38}
\defmmnum{xprofile.ws4_wh_qwen_g3.other.cap160_n}{2{,}370}
\defmmnum{xprofile.ws4_wh_qwen_g3.other.cap160_pct}{80.64}
\defmmnum{xprofile.ws4_wh_qwen_g3.other.del_pct}{0.91}
\defmmnum{xprofile.ws4_wh_qwen_g3.other.del_words}{475}
\defmmnum{xprofile.ws4_wh_qwen_g3.other.empty_pct}{0.00}
\defmmnum{xprofile.ws4_wh_qwen_g3.other.ins_pct}{416.88}
\defmmnum{xprofile.ws4_wh_qwen_g3.other.ins_words}{218{,}210}
\defmmnum{xprofile.ws4_wh_qwen_g3.other.n_empty}{0}
\defmmnum{xprofile.ws4_wh_qwen_g3.other.n_utt}{2{,}939}
\defmmnum{xprofile.ws4_wh_qwen_g3.other.nonempty_n}{2{,}939}
\defmmnum{xprofile.ws4_wh_qwen_g3.other.nonempty_utt_median_wer}{600.00}
\defmmnum{xprofile.ws4_wh_qwen_g3.other.nonempty_wer}{506.34}
\defmmnum{xprofile.ws4_wh_qwen_g3.other.sub_pct}{88.55}
\defmmnum{xprofile.ws4_wh_qwen_g3.other.sub_words}{46{,}349}
\defmmnum{xprofile.ws4_wh_qwen_g3.other.utt_eq0_n}{0}
\defmmnum{xprofile.ws4_wh_qwen_g3.other.utt_eq0_pct}{0.00}
\defmmnum{xprofile.ws4_wh_qwen_g3.other.utt_ge100_n}{2{,}917}
\defmmnum{xprofile.ws4_wh_qwen_g3.other.utt_ge100_pct}{99.25}
\defmmnum{xprofile.ws4_wh_qwen_g3.other.utt_lt10_n}{0}
\defmmnum{xprofile.ws4_wh_qwen_g3.other.utt_lt10_pct}{0.00}
\defmmnum{xprofile.ws4_wh_qwen_g3.other.utt_lt50_n}{0}
\defmmnum{xprofile.ws4_wh_qwen_g3.other.utt_lt50_pct}{0.00}
\defmmnum{xprofile.ws4_wh_qwen_g3.other.utt_median_wer}{600.00}
\defmmnum{xprofile.ws4_wh_qwen_g3.other.wer}{506.34}
\defmmnum{xprofile.ws4_wh_qwen_g3.other.word_ratio_corpus}{5.16}
\defmmnum{xprofile.ws4_wh_qwen_g3.other.word_ratio_median}{6.14}
\defmmnum{xprofile.x1.caponesixtypct}{9.89}
\defmmnum{xprofile.x1.clean.cap159_n}{283}
\defmmnum{xprofile.x1.clean.cap159_pct}{10.80}
\defmmnum{xprofile.x1.clean.cap160_n}{259}
\defmmnum{xprofile.x1.clean.cap160_pct}{9.89}
\defmmnum{xprofile.x1.clean.del_pct}{2.77}
\defmmnum{xprofile.x1.clean.del_words}{1{,}455}
\defmmnum{xprofile.x1.clean.empty_pct}{0.00}
\defmmnum{xprofile.x1.clean.ins_pct}{53.15}
\defmmnum{xprofile.x1.clean.ins_words}{27{,}944}
\defmmnum{xprofile.x1.clean.n_empty}{0}
\defmmnum{xprofile.x1.clean.n_utt}{2{,}620}
\defmmnum{xprofile.x1.clean.nonempty_n}{2{,}620}
\defmmnum{xprofile.x1.clean.nonempty_utt_median_wer}{23.81}
\defmmnum{xprofile.x1.clean.nonempty_wer}{75.35}
\defmmnum{xprofile.x1.clean.sub_pct}{19.44}
\defmmnum{xprofile.x1.clean.sub_words}{10{,}219}
\defmmnum{xprofile.x1.clean.utt_eq0_n}{251}
\defmmnum{xprofile.x1.clean.utt_eq0_pct}{9.58}
\defmmnum{xprofile.x1.clean.utt_ge100_n}{440}
\defmmnum{xprofile.x1.clean.utt_ge100_pct}{16.79}
\defmmnum{xprofile.x1.clean.utt_lt10_n}{552}
\defmmnum{xprofile.x1.clean.utt_lt10_pct}{21.07}
\defmmnum{xprofile.x1.clean.utt_lt50_n}{1{,}969}
\defmmnum{xprofile.x1.clean.utt_lt50_pct}{75.15}
\defmmnum{xprofile.x1.clean.utt_median_wer}{23.81}
\defmmnum{xprofile.x1.clean.wer}{75.35}
\defmmnum{xprofile.x1.clean.word_ratio_corpus}{1.50}
\defmmnum{xprofile.x1.clean.word_ratio_median}{1.00}
\defmmnum{xprofile.x1.delpct}{2.77}
\defmmnum{xprofile.x1.inspct}{53.15}
\defmmnum{xprofile.x1.nonemptymedwer}{23.81}
\defmmnum{xprofile.x1.nonemptywer}{75.35}
\defmmnum{xprofile.x1.other.cap159_n}{275}
\defmmnum{xprofile.x1.other.cap159_pct}{9.36}
\defmmnum{xprofile.x1.other.cap160_n}{252}
\defmmnum{xprofile.x1.other.cap160_pct}{8.57}
\defmmnum{xprofile.x1.other.del_pct}{4.64}
\defmmnum{xprofile.x1.other.del_words}{2{,}428}
\defmmnum{xprofile.x1.other.empty_pct}{0.00}
\defmmnum{xprofile.x1.other.ins_pct}{57.28}
\defmmnum{xprofile.x1.other.ins_words}{29{,}984}
\defmmnum{xprofile.x1.other.n_empty}{0}
\defmmnum{xprofile.x1.other.n_utt}{2{,}939}
\defmmnum{xprofile.x1.other.nonempty_n}{2{,}939}
\defmmnum{xprofile.x1.other.nonempty_utt_median_wer}{40.00}
\defmmnum{xprofile.x1.other.nonempty_wer}{89.82}
\defmmnum{xprofile.x1.other.sub_pct}{27.90}
\defmmnum{xprofile.x1.other.sub_words}{14{,}605}
\defmmnum{xprofile.x1.other.utt_eq0_n}{149}
\defmmnum{xprofile.x1.other.utt_eq0_pct}{5.07}
\defmmnum{xprofile.x1.other.utt_ge100_n}{539}
\defmmnum{xprofile.x1.other.utt_ge100_pct}{18.34}
\defmmnum{xprofile.x1.other.utt_lt10_n}{261}
\defmmnum{xprofile.x1.other.utt_lt10_pct}{8.88}
\defmmnum{xprofile.x1.other.utt_lt50_n}{1{,}773}
\defmmnum{xprofile.x1.other.utt_lt50_pct}{60.33}
\defmmnum{xprofile.x1.other.utt_median_wer}{40.00}
\defmmnum{xprofile.x1.other.wer}{89.82}
\defmmnum{xprofile.x1.other.word_ratio_corpus}{1.53}
\defmmnum{xprofile.x1.other.word_ratio_median}{1.00}
\defmmnum{xprofile.x1.subpct}{19.44}
\defmmnum{xprofile.x1.uttltfiftypct}{75.15}
\defmmnum{xprofile.x1.uttmedwer}{23.81}
\defmmnum{xprofile.x1.wer}{75.35}
\defmmnum{xprofile.x2.caponesixtypct}{3.89}
\defmmnum{xprofile.x2.clean.cap159_n}{280}
\defmmnum{xprofile.x2.clean.cap159_pct}{10.69}
\defmmnum{xprofile.x2.clean.cap160_n}{102}
\defmmnum{xprofile.x2.clean.cap160_pct}{3.89}
\defmmnum{xprofile.x2.clean.del_pct}{31.71}
\defmmnum{xprofile.x2.clean.del_words}{16{,}673}
\defmmnum{xprofile.x2.clean.empty_pct}{13.78}
\defmmnum{xprofile.x2.clean.ins_pct}{18.92}
\defmmnum{xprofile.x2.clean.ins_words}{9{,}945}
\defmmnum{xprofile.x2.clean.n_empty}{361}
\defmmnum{xprofile.x2.clean.n_utt}{2{,}620}
\defmmnum{xprofile.x2.clean.nonempty_n}{2{,}259}
\defmmnum{xprofile.x2.clean.nonempty_utt_median_wer}{16.67}
\defmmnum{xprofile.x2.clean.nonempty_wer}{57.62}
\defmmnum{xprofile.x2.clean.sub_pct}{16.72}
\defmmnum{xprofile.x2.clean.sub_words}{8{,}793}
\defmmnum{xprofile.x2.clean.utt_eq0_n}{409}
\defmmnum{xprofile.x2.clean.utt_eq0_pct}{15.61}
\defmmnum{xprofile.x2.clean.utt_ge100_n}{676}
\defmmnum{xprofile.x2.clean.utt_ge100_pct}{25.80}
\defmmnum{xprofile.x2.clean.utt_lt10_n}{768}
\defmmnum{xprofile.x2.clean.utt_lt10_pct}{29.31}
\defmmnum{xprofile.x2.clean.utt_lt50_n}{1{,}737}
\defmmnum{xprofile.x2.clean.utt_lt50_pct}{66.30}
\defmmnum{xprofile.x2.clean.utt_median_wer}{20.83}
\defmmnum{xprofile.x2.clean.wer}{67.35}
\defmmnum{xprofile.x2.clean.word_ratio_corpus}{0.87}
\defmmnum{xprofile.x2.clean.word_ratio_median}{1.00}
\defmmnum{xprofile.x2.delpct}{31.71}
\defmmnum{xprofile.x2.inspct}{18.92}
\defmmnum{xprofile.x2.nonemptymedwer}{16.67}
\defmmnum{xprofile.x2.nonemptywer}{57.62}
\defmmnum{xprofile.x2.other.cap159_n}{334}
\defmmnum{xprofile.x2.other.cap159_pct}{11.36}
\defmmnum{xprofile.x2.other.cap160_n}{131}
\defmmnum{xprofile.x2.other.cap160_pct}{4.46}
\defmmnum{xprofile.x2.other.del_pct}{32.49}
\defmmnum{xprofile.x2.other.del_words}{17{,}006}
\defmmnum{xprofile.x2.other.empty_pct}{13.17}
\defmmnum{xprofile.x2.other.ins_pct}{24.30}
\defmmnum{xprofile.x2.other.ins_words}{12{,}720}
\defmmnum{xprofile.x2.other.n_empty}{387}
\defmmnum{xprofile.x2.other.n_utt}{2{,}939}
\defmmnum{xprofile.x2.other.nonempty_n}{2{,}552}
\defmmnum{xprofile.x2.other.nonempty_utt_median_wer}{30.00}
\defmmnum{xprofile.x2.other.nonempty_wer}{70.97}
\defmmnum{xprofile.x2.other.sub_pct}{20.68}
\defmmnum{xprofile.x2.other.sub_words}{10{,}822}
\defmmnum{xprofile.x2.other.utt_eq0_n}{265}
\defmmnum{xprofile.x2.other.utt_eq0_pct}{9.02}
\defmmnum{xprofile.x2.other.utt_ge100_n}{847}
\defmmnum{xprofile.x2.other.utt_ge100_pct}{28.82}
\defmmnum{xprofile.x2.other.utt_lt10_n}{440}
\defmmnum{xprofile.x2.other.utt_lt10_pct}{14.97}
\defmmnum{xprofile.x2.other.utt_lt50_n}{1{,}677}
\defmmnum{xprofile.x2.other.utt_lt50_pct}{57.06}
\defmmnum{xprofile.x2.other.utt_median_wer}{36.36}
\defmmnum{xprofile.x2.other.wer}{77.47}
\defmmnum{xprofile.x2.other.word_ratio_corpus}{0.92}
\defmmnum{xprofile.x2.other.word_ratio_median}{1.00}
\defmmnum{xprofile.x2.subpct}{16.72}
\defmmnum{xprofile.x2.uttltfiftypct}{66.30}
\defmmnum{xprofile.x2.uttmedwer}{20.83}
\defmmnum{xprofile.x2.wer}{67.35}
\defmmnum{xprofile.zero_audio.closed_form_cells}{84}
\defmmnum{xprofile.zero_audio.lt50_pct}{0.00}
\defmmnum{xprofile.zero_audio.lt50_utts}{0}
\defmmnum{xprofile.zero_audio.min_med_clean}{100.00}
\defmmnum{xprofile.zero_audio.min_med_other}{100.00}
\defmmnum{xprofile.zero_audio.min_wer_clean}{99.65}
\defmmnum{xprofile.zero_audio.n_cells}{113}
\defmmnum{xprofile.zero_audio.noise_cells}{3}
\defmmnum{xprofile.zero_audio.rand_orth_cells}{2}
\defmmnum{xprofile.zero_audio.total_utts}{628{,}167}
\defmmnum{xprofile.zero_audio.transcript_cells}{16}
\defmmnum{xprofile.zero_audio.wiki_cells}{8}

%% file: sections/00_abstract.tex
\begin{abstract}
Connecting a new input modality or a new language model normally
requires training a connector for each pair.
If the input-embedding spaces of independently trained models are
linearly related, connectors could be computed in closed form or
reused.
We test this hypothesis across three experiments with all encoders and language models frozen:
(1)~transplanting vocabulary embeddings across model families, tokenizers, and widths into a
frozen target via closed-form least squares plus one radius scalar
retains up to $\WsOneBestPiqa\%$ PIQA (vs.\ $\WsOneNativePiqa\%$ unmodified)
with zero gradient steps;
(2)~a single normalized linear layer from a frozen speech encoder into a
frozen language model, with no frame stacking and no hidden
nonlinearity, trained on LibriSpeech, reaches
\WsTwoBestMuonRunWerPairCell\ word error rate (clean / other)
from Whisper-Small into Gemma-3-4B-IT, below Whisper-Small's own
decoder used zero-shot ($\NativeWhisperWerClean\%$ / $\NativeWhisperWerOther\%$);
(3)~chaining a speech adapter with a linear bridge into a second
language model succeeds when the bridge is fit on that adapter's
soft-prompt outputs ($\WsFourBCompBestClean\%$ /
$\WsFourBCompBestOther\%$ composed WER)
but fails when fit on vocabulary tokens or text
($\XprofileZeroAudioLtFiftyPct\%$ of
$\XprofileZeroAudioTotalUtts$ decoded utterances below
$\UttWerHalfThresholdPct\%$ WER, over all
$\XprofileZeroAudioCells$ configurations that never saw audio)
and does not generalize to another speech encoder's adapter
($\CrossAdapterFailCells$ of $\CrossAdapterTotalCells$ cells
fail completely) unless the bridge is fit on both adapters' outputs.
Each linear map works on the inputs and the task it was fit on,
and not on the others.
\end{abstract}

%% file: sections/01_intro.tex
\section{Introduction}
\label{sec:intro}

How far does a linear map between two language models' input-embedding
spaces go?
If independently trained models occupy linearly related coordinate
systems, then a trained adapter---a speech recognizer, say---could be
moved to a new language model with a single matrix multiplication,
no paired data, and no gradient step through the target model.

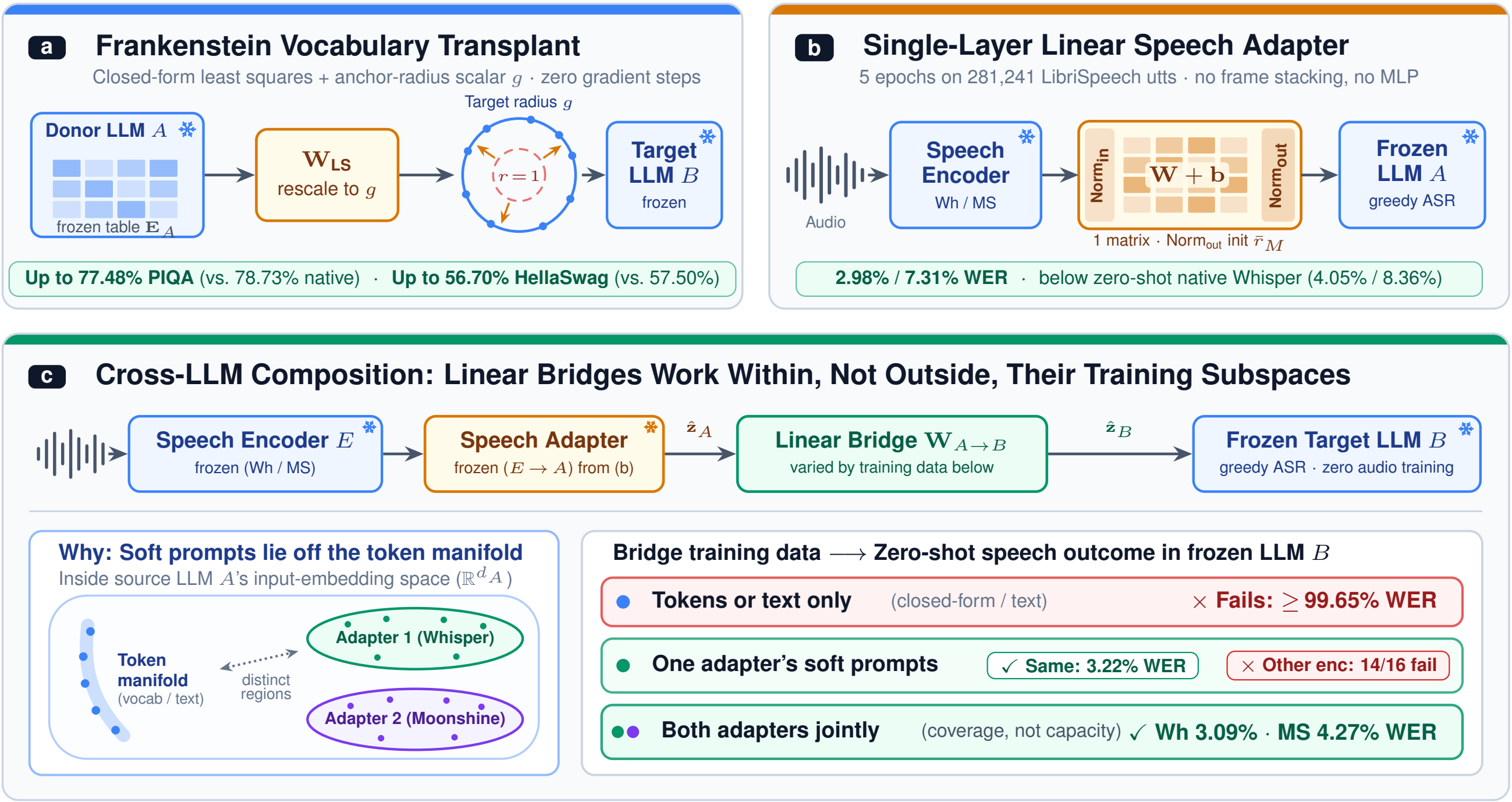
\begin{figure*}[!t]
\centering
\resizebox{\textwidth}{!}{%
\begin{tikzpicture}[
  x=1cm, y=1cm,
  font=\sffamily,
  badge/.style={fill=ink, text=white, rounded corners=2.2pt, font=\sffamily\scriptsize\bfseries, inner xsep=4pt, inner ysep=1.8pt},
  ctitle/.style={font=\sffamily\footnotesize\bfseries, text=ink, anchor=west},
  csub/.style={font=\sffamily\tiny, text=muted, anchor=west},
  fbox/.style={draw=icebd, fill=icefill, line width=0.8pt, rounded corners=3.5pt, align=center, font=\sffamily\scriptsize\bfseries, text=icedark},
  abox/.style={draw=amberbd, fill=amberfill, line width=0.85pt, rounded corners=3.5pt, align=center, font=\sffamily\scriptsize\bfseries, text=amberdark},
  ebox/.style={draw=emebd, fill=emefill, line width=0.85pt, rounded corners=3.5pt, align=center, font=\sffamily\scriptsize\bfseries, text=emedark},
  arr/.style={-{Stealth[length=2.1mm,width=1.7mm]}, line width=0.85pt, draw=arrowc},
  kpigreen/.style={draw=emebd!60, fill=emefill, rounded corners=3pt, inner xsep=5pt, inner ysep=2.6pt, font=\sffamily\tiny, text=emedark}
]

\begin{scope}
  \clip[rounded corners=5pt] (0.00, 5.48) rectangle (8.25, 8.88);
  \fill[cardbg] (0.00, 5.48) rectangle (8.25, 8.88);
  \fill[icebd] (0.00, 8.76) rectangle (8.25, 8.88);
\end{scope}
\draw[cardbd, line width=0.65pt, rounded corners=5pt] (0.00, 5.48) rectangle (8.25, 8.88);

\node[badge, anchor=west] (ba) at (0.28, 8.39) {a};
\node[ctitle, right=2mm of ba] {Frankenstein Vocabulary Transplant};
\node[csub] at (0.88, 8.05) {Closed-form least squares + anchor-radius scalar $g$ $\cdot$ zero gradient steps};

\node[fbox, minimum width=1.92cm, minimum height=1.38cm] (don) at (1.28, 6.97) {};
\node[font=\sffamily\tiny\bfseries, text=icedark] at (1.16, 7.47) {Donor LLM $A$};
\pic[scale=0.09] at (2.06, 7.48) {snowflake};
\foreach \r in {0,1,2} {
  \foreach \c in {0,1,2,3} {
    \pgfmathtruncatemacro{\op}{25 + 15*mod(\r+\c,3)}
    \fill[icebd!\op, rounded corners=0.4pt] (0.56+\c*0.36, 6.49+\r*0.23) rectangle (0.87+\c*0.36, 6.68+\r*0.23);
  }
}
\node[font=\sffamily\fontsize{5.1}{5.8}\selectfont, text=slate] at (1.28, 6.37) {frozen table $\mathbf{E}_A$};

\node[abox, minimum width=1.58cm, minimum height=1.02cm] (wls) at (3.62, 6.97) {$\mathbf{W}_{\text{LS}}$\\[1pt]{\mdseries\tiny rescale to $g$}};

\begin{scope}[shift={(5.76, 6.97)}]
  \draw[rosebd!65, densely dashed, line width=0.65pt] (0,0) circle (0.29);
  \draw[icebd, line width=0.85pt] (0,0) circle (0.63);
  \foreach \ang in {35, 145, 250} {
    \draw[-{Stealth[length=1.4mm,width=1.1mm]}, amberbd, line width=0.65pt] (\ang:0.33) -- (\ang:0.58);
  }
  \foreach \ang in {20, 45, 78, 125, 155, 200, 240, 295, 335} {
    \fill[icebd] (\ang:0.63) circle (0.045);
  }
  \node[font=\sffamily\fontsize{5.1}{5.8}\selectfont, text=icedark] at (0, 0.80) {Target radius $g$};
  \node[font=\sffamily\fontsize{4.8}{5.5}\selectfont, text=rosedark] at (0, -0.01) {$r\!=\!1$};
\end{scope}

\node[fbox, minimum width=1.28cm, minimum height=1.18cm] (tgt) at (7.38, 6.97) {Target\\LLM $B$\\{\mdseries\fontsize{5.1}{5.8}\selectfont frozen}};
\pic[scale=0.085] at (7.86, 7.40) {snowflake};

\draw[arr] (don.east) -- (wls.west);
\draw[arr] (wls.east) -- (5.04, 6.97);
\draw[arr] (6.46, 6.97) -- (tgt.west);

\node[kpigreen, minimum width=7.65cm, align=center] at (4.125, 5.81) {\textbf{Up to \WsOneBestPiqa\% PIQA} (vs.\ \WsOneNativePiqa\% native) $\;\cdot\;$ \textbf{Up to \WsOneBestHs\% HellaSwag} (vs.\ \WsOneNativeHs\%)};

\begin{scope}
  \clip[rounded corners=5pt] (8.55, 5.48) rectangle (16.80, 8.88);
  \fill[cardbg] (8.55, 5.48) rectangle (16.80, 8.88);
  \fill[amberbd] (8.55, 8.76) rectangle (16.80, 8.88);
\end{scope}
\draw[cardbd, line width=0.65pt, rounded corners=5pt] (8.55, 5.48) rectangle (16.80, 8.88);

\node[badge, anchor=west] (bb) at (8.83, 8.39) {b};
\node[ctitle, right=2mm of bb] {Single-Layer Linear Speech Adapter};
\node[csub] at (9.43, 8.05) {\WsTwoUniformEpochs\ epochs on \LibriTrainUtts\ LibriSpeech utts $\cdot$ no frame stacking, no MLP};

\begin{scope}[shift={(9.18, 6.97)}]
  \foreach \i/\h in {0/0.14, 1/0.34, 2/0.52, 3/0.24, 4/0.60, 5/0.40, 6/0.18, 7/0.46, 8/0.28, 9/0.12} {
    \draw[slate, line width=1.1pt, line cap=round] (-0.42+\i*0.092, -\h/2) -- (-0.42+\i*0.092, \h/2);
  }
  \node[font=\sffamily\fontsize{5.1}{5.8}\selectfont, text=muted] at (0.0, -0.52) {Audio};
\end{scope}

\node[fbox, minimum width=1.68cm, minimum height=1.18cm] (encb) at (10.74, 6.97) {Speech\\Encoder\\{\mdseries\fontsize{5.1}{5.8}\selectfont Wh / MS}};
\pic[scale=0.085] at (11.42, 7.40) {snowflake};

\draw[amberbd, fill=amberfill, line width=0.85pt, rounded corners=3.5pt] (12.00, 6.37) rectangle (14.48, 7.57);
\fill[amberbd!22, rounded corners=1.5pt] (12.07, 6.45) rectangle (12.40, 7.49);
\node[rotate=90, font=\sffamily\fontsize{5}{5.5}\selectfont\bfseries, text=amberdark] at (12.235, 6.97) {Norm$_{\text{in}}$};
\foreach \r in {0,1,2,3} {
  \foreach \c in {0,1,2,3} {
    \pgfmathtruncatemacro{\op}{22 + 16*mod(2*\r+\c,3)}
    \fill[amberbd!\op, rounded corners=0.4pt] (12.50+\c*0.36, 6.55+\r*0.22) rectangle (12.80+\c*0.36, 6.73+\r*0.22);
  }
}
\node[font=\sffamily\scriptsize\bfseries, text=amberdark, fill=amberfill, fill opacity=0.88, text opacity=1, inner sep=1.2pt, rounded corners=1.5pt] at (13.21, 6.97) {$\mathbf{W}+\mathbf{b}$};
\fill[amberbd!28, rounded corners=1.5pt] (14.04, 6.45) rectangle (14.41, 7.49);
\node[rotate=90, font=\sffamily\fontsize{5}{5.5}\selectfont\bfseries, text=amberdark] at (14.225, 6.97) {Norm$_{\text{out}}$};
\node[font=\sffamily\fontsize{5.1}{5.8}\selectfont, text=amberdark] at (13.24, 6.22) {1 matrix $\cdot$ Norm$_{\text{out}}$ init $\bar r_M$};

\node[fbox, minimum width=1.62cm, minimum height=1.18cm] (llma) at (15.72, 6.97) {Frozen\\LLM $A$\\{\mdseries\fontsize{5.1}{5.8}\selectfont greedy ASR}};
\pic[scale=0.085] at (16.37, 7.40) {snowflake};

\draw[arr] (9.65, 6.97) -- (encb.west);
\draw[arr] (encb.east) -- (12.00, 6.97);
\draw[arr] (14.48, 6.97) -- (llma.west);

\node[kpigreen, minimum width=7.65cm, align=center] at (12.675, 5.81) {\textbf{\WsTwoBestMuonRunWerPairCell\ WER} $\;\cdot\;$ below zero-shot native Whisper (\NativeWhisperWerClean\% / \NativeWhisperWerOther\%)};

\begin{scope}
  \clip[rounded corners=5pt] (0.00, 0.00) rectangle (16.80, 5.20);
  \fill[cardbg] (0.00, 0.00) rectangle (16.80, 5.20);
  \fill[emebd] (0.00, 5.08) rectangle (16.80, 5.20);
\end{scope}
\draw[cardbd, line width=0.65pt, rounded corners=5pt] (0.00, 0.00) rectangle (16.80, 5.20);

\node[badge, anchor=west] (bc) at (0.28, 4.72) {c};
\node[ctitle, right=2mm of bc] {Cross-LLM Composition: Linear Bridges Work Within, Not Outside, Their Training Subspaces};

\begin{scope}[shift={(0.76, 3.86)}]
  \foreach \i/\h in {0/0.12, 1/0.30, 2/0.46, 3/0.22, 4/0.52, 5/0.34, 6/0.18, 7/0.40, 8/0.22} {
    \draw[slate, line width=1.0pt, line cap=round] (-0.36+\i*0.09, -\h/2) -- (-0.36+\i*0.09, \h/2);
  }
\end{scope}
\node[fbox, minimum width=2.82cm, minimum height=0.84cm] (p1) at (2.82, 3.86) {Speech Encoder $E$\\{\mdseries\fontsize{5.1}{5.8}\selectfont frozen (Wh / MS)}};
\pic[scale=0.068] at (4.09, 4.16) {snowflake};

\node[abox, minimum width=2.66cm, minimum height=0.84cm] (p2) at (6.04, 3.86) {Speech Adapter\\{\mdseries\fontsize{5.1}{5.8}\selectfont frozen ($E\!\to\!A$) from (b)}};
\pic[scale=0.068] at (7.23, 4.16) {snowamber};

\node[ebox, minimum width=3.45cm, minimum height=0.84cm] (p3) at (9.92, 3.86) {Linear Bridge $\mathbf{W}_{A\to B}$\\{\mdseries\fontsize{5.1}{5.8}\selectfont varied by training data below}};

\node[fbox, minimum width=3.20cm, minimum height=0.84cm] (p4) at (14.88, 3.86) {Frozen Target LLM $B$\\{\mdseries\fontsize{5.1}{5.8}\selectfont greedy ASR $\cdot$ zero audio training}};
\pic[scale=0.075] at (16.32, 4.14) {snowflake};

\draw[arr] (1.18, 3.86) -- (p1.west);
\draw[arr] (p1.east) -- (p2.west);
\draw[arr] (p2.east) -- (p3.west) node[midway, above=1.5pt, font=\sffamily\fontsize{5.4}{6}\selectfont\bfseries, text=amberdark] {$\hat{\mathbf{z}}_A$};
\draw[arr] (p3.east) -- (p4.west) node[midway, above=1.5pt, font=\sffamily\fontsize{5.4}{6}\selectfont\bfseries, text=emedark] {$\hat{\mathbf{z}}_B$};

\draw[cardbd, line width=0.55pt] (0.30, 3.22) -- (16.50, 3.22);

\draw[draw=icebd!55, fill=white, line width=0.7pt, rounded corners=4pt] (0.30, 0.28) rectangle (6.20, 3.00);
\node[font=\sffamily\scriptsize\bfseries, text=icedark, anchor=west] at (0.50, 2.74) {Why: Soft prompts lie off the token manifold};
\node[font=\sffamily\tiny, text=muted, anchor=west] at (0.50, 2.48) {Inside source LLM $A$'s input-embedding space ($\mathbb{R}^{d_A}$)};

\draw[draw=icebd!40, fill=icefill!25, line width=0.65pt, rounded corners=12pt] (0.52, 0.46) rectangle (5.98, 2.28);

\draw[icebd!45, line width=4.2pt, line cap=round, opacity=0.55] (0.95, 1.95) to[bend right=25] (1.35, 0.72);
\foreach \x/\y in {0.98/1.88, 0.90/1.60, 0.92/1.30, 1.04/1.00, 1.26/0.78} {\fill[icebd] (\x, \y) circle (0.048);}
\node[font=\sffamily\fontsize{5.4}{6}\selectfont\bfseries, text=icedark, anchor=west] at (1.16, 1.56) {Token};
\node[font=\sffamily\fontsize{5.4}{6}\selectfont\bfseries, text=icedark, anchor=west] at (1.16, 1.34) {manifold};
\node[font=\sffamily\fontsize{4.8}{5.4}\selectfont, text=muted, anchor=west] at (1.16, 1.12) {(vocab / text)};

\draw[{Stealth[length=1.4mm,width=1.1mm]}-{Stealth[length=1.4mm,width=1.1mm]}, muted, densely dotted, line width=0.75pt] (2.42, 1.46) -- (3.30, 1.66);
\node[font=\sffamily\fontsize{4.9}{5.4}\selectfont, text=muted, align=center] at (2.94, 1.25) {distinct\\[-1pt]regions};

\fill[emefill, draw=emebd, line width=0.8pt] (4.60, 1.80) ellipse (1.20 and 0.34);
\foreach \x/\y in {3.85/1.96, 4.28/2.02, 4.92/2.01, 5.36/1.94, 4.18/1.59, 5.06/1.60} {\fill[emebd] (\x, \y) circle (0.036);}
\node[font=\sffamily\fontsize{5.3}{5.8}\selectfont\bfseries, text=emedark] at (4.60, 1.80) {Adapter 1 (Whisper)};

\fill[viofill, draw=viobd, line width=0.8pt] (4.60, 0.90) ellipse (1.20 and 0.34);
\foreach \x/\y in {3.88/1.05, 4.32/1.12, 4.95/1.11, 5.34/1.04, 4.22/0.69, 5.04/0.70} {\fill[viobd] (\x, \y) circle (0.036);}
\node[font=\sffamily\fontsize{5.3}{5.8}\selectfont\bfseries, text=viodark] at (4.60, 0.90) {Adapter 2 (Moonshine)};

\draw[draw=cardbd, fill=white, line width=0.7pt, rounded corners=4pt] (6.46, 0.28) rectangle (16.50, 3.00);
\node[font=\sffamily\scriptsize\bfseries, text=ink, anchor=west] at (6.68, 2.74) {Bridge training data $\longrightarrow$ Zero-shot speech outcome in frozen LLM $B$};

\draw[draw=rosebd!75, fill=rosefill, line width=0.75pt, rounded corners=3.5pt] (6.68, 1.94) rectangle (16.28, 2.48);
\fill[icebd] (6.92, 2.21) circle (0.075);
\node[anchor=west, font=\sffamily\scriptsize\bfseries, text=ink] at (7.12, 2.21) {Tokens or text only};
\node[anchor=west, font=\sffamily\tiny, text=muted] at (9.78, 2.21) {(closed-form / text)};
\node[anchor=east, font=\sffamily\scriptsize\bfseries, text=rosedark] at (16.12, 2.21) {$\times$ Fails: $\ge\,$\ZeroAudioCompBestClean\% WER};

\draw[draw=emebd!75, fill=emefill!65, line width=0.75pt, rounded corners=3.5pt] (6.68, 1.20) rectangle (16.28, 1.80);
\fill[emebd] (6.92, 1.50) circle (0.075);
\node[anchor=west, font=\sffamily\scriptsize\bfseries, text=ink] at (7.12, 1.50) {One adapter's soft prompts};
\node[draw=emebd, fill=white, rounded corners=2.5pt, inner xsep=3.8pt, inner ysep=2.0pt, anchor=east, font=\sffamily\fontsize{5.8}{6.3}\selectfont\bfseries, text=emedark] at (13.34, 1.50) {$\checkmark$ Same: \WsFourBCompBestClean\% WER};
\node[draw=rosebd, fill=rosefill, rounded corners=2.5pt, inner xsep=3.8pt, inner ysep=2.0pt, anchor=east, font=\sffamily\fontsize{5.8}{6.3}\selectfont\bfseries, text=rosedark] at (16.14, 1.50) {$\times$ Other enc: \CrossAdapterFailCells/\CrossAdapterTotalCells\ fail};

\draw[draw=emebd!85, fill=emefill, line width=0.8pt, rounded corners=3.5pt] (6.68, 0.46) rectangle (16.28, 1.06);
\fill[emebd] (6.86, 0.76) circle (0.068); \fill[viobd] (7.03, 0.76) circle (0.068);
\node[anchor=west, font=\sffamily\scriptsize\bfseries, text=ink] at (7.22, 0.76) {Both adapters jointly};
\node[anchor=west, font=\sffamily\tiny, text=slate] at (10.12, 0.76) {(coverage, not capacity)};
\node[anchor=east, font=\sffamily\scriptsize\bfseries, text=emedark] at (16.12, 0.76) {$\checkmark$ Wh \WsFourEJointWhisperMuonClean\% $\cdot$ MS \WsFourEJointMoonshineMuonClean\% WER};
\end{tikzpicture}%
}
\vspace{-1pt}
\caption{\textbf{Overview of our three tests in frozen language model input-embedding space.} \textbf{(a)}~Overwriting shared vocabulary rows of a frozen target LLM via closed-form least squares plus target-radius calibration retains near-native PIQA and HellaSwag accuracy ($\uparrow$). \textbf{(b)}~Training a single normalized linear layer, its output normalization initialized at the target mean token radius $\bar r_M$, reaches, from Whisper-Small into Gemma-3-4B-IT, lower LibriSpeech word error rate ($\downarrow$) than Whisper-Small's own pretrained decoder used zero-shot. \textbf{(c)}~Chaining a trained speech adapter with a linear bridge $A \to B$ succeeds when the bridge is fit on that adapter's soft prompts, fails when fit only on vocabulary tokens or text, and does not reach usable accuracy on another speech encoder's adapter unless it is fit on both (WER in (c) is \texttt{test-clean}; Wh: Whisper-Small, MS: Moonshine-Base; the 16 cross-adapter cells include bridges fit on one adapter's soft prompts mixed with text).}
\label{fig:teaser}
\end{figure*}

We test this hypothesis through three experiments with every model
frozen (Figure~\ref{fig:teaser}).
First, we transplant vocabulary embeddings from a donor model into
a frozen target model using a closed-form least-squares map.
Second, we train a single linear layer from a frozen speech encoder
into a frozen language model for automatic speech recognition.
Third, we chain the two: a speech adapter trained with one language
model, composed with a linear bridge into a second language model
that has never heard audio.
The scope of this paper is speech recognition; other audio tasks and vision
are left for later work.

Our findings:
\begin{enumerate}[leftmargin=1.6em,itemsep=1.5pt,topsep=2pt]
    \item \textbf{Tokens transfer with a closed-form map plus one
    scalar.}
    Overwriting shared vocabulary rows via least squares and
    calibrating output radius to the target's mean token radius keeps a frozen language model at up to
    $\WsOneBestPiqa\%$ PIQA ($\WsOneNativePiqa\%$ native) and up to
    $\WsOneBestHs\%$ HellaSwag ($\WsOneNativeHs\%$ native),
    across donor families and mismatched embedding widths,
    with zero gradient steps.
    Projecting onto the unit sphere without radius restoration
    collapses every configuration to chance.

    \item \textbf{Speech enters a frozen language model through
    one trained linear layer.}
    With encoder and language model frozen, a single matrix,
    its output normalization initialized at the target's mean token radius, reaches
    \WsTwoBestMuonRunWerPairCell\ WER on LibriSpeech from Whisper-Small into Gemma-3-4B-IT,
    below Whisper-Small's own pretrained decoder used zero-shot
    ($\NativeWhisperWerClean\%$ / $\NativeWhisperWerOther\%$).
    Unlike the SLAM-ASR projector \citep{ma2024embarrassingly},
    our adapter uses no frame stacking and no hidden nonlinearity.

    \item \textbf{Chaining the two has a sharp boundary.}
    A bridge fit on one adapter's soft-prompt outputs carries that
    adapter to a second language model, on the best pair at the adapter's own accuracy
    ($\WsFourBCompBestClean\%$ / $\WsFourBCompBestOther\%$ WER),
    with zero empty decodes.
    Same-adapter composition works; transfer to another encoder's
    adapter does not:
    $\CrossAdapterFailCells$ of $\CrossAdapterTotalCells$
    cross-adapter cells (bridges fit on one adapter's soft prompts, alone or mixed with text; Table~\ref{tab:ws3_comp}) fail completely and two decode partially.
    The limit is coverage, not capacity: one bridge fit on both
    adapters' soft prompts serves both.
    A bridge fit on vocabulary tokens or text---even one with
    near-zero text loss---never produces usable speech recognition
    ($\XprofileZeroAudioLtFiftyPct\%$ of $\XprofileZeroAudioTotalUtts$
    decoded utterances below $\UttWerHalfThresholdPct\%$ WER, over all
    $\XprofileZeroAudioCells$ configurations that never saw audio).
    Conversely, bridges trained through the target model make poor static vocabularies.
    Each linear map works on the inputs and the task it was fit on,
    and not on the others.
\end{enumerate}

%% file: sections/02_related_work.tex
\section{Related Work}
\label{sec:related}

\paragraph{Representation convergence and model stitching.}
Independently trained networks learn geometrically related
representations \citep{li2015convergent, huh2024platonic, nelson2026statistical}.
Model stitching aligns frozen internal representations via task-trained linear layers~\citep{lenc2015understanding,bansal2021revisiting} or unsupervised activation-matching maps~\citep{chen2025transferring}. Related work fits closed-form ridge regression on late-stage hidden states~\citep{gorbett2026characterizing}, whereas task-level model fusion fine-tunes nonlinear connectors end to end~\citep{jiang2024stitching}. We map static input-embedding tables directly in closed form, with zero forward passes or gradient steps.

\paragraph{Embedding alignment.}
Linear and orthogonal Procrustes maps~\citep{schonemann1966generalized} between word-embedding spaces~\citep{mikolov2013exploiting,smith2017offline,conneau2018word} underpin cross-lingual transfer, while relative representations align spaces through shared anchors without fitting a map~\citep{moschella2023relative}.
Our vocabulary transplants extend this idea from retrieval to
downstream reasoning: a map that retrieves tokens accurately
may still fail to preserve function (Section~\ref{sec:tokens}).

\paragraph{Vocabulary and tokenizer transfer.}
Embedding tables can be warm-started for a new tokenizer
\citep{minixhofer2022wechsel, dobler2023focus} or predicted by a
trained hypernetwork \citep{minixhofer2024zeroshot}.
We evaluate closed-form linear transplants with no further training
and show that the map must keep the target's embedding magnitude:
the same directions at unit length leave the model at chance.
Unlike tokenizer transfer, a transplant keeps the target's tokenizer and
output head and replaces only the input rows of the tokens that the two
vocabularies share.

\paragraph{Connectors for multimodal language models.}
Standard multimodal language models pair linear or MLP projectors with language-model fine-tuning, use cross-attention resamplers, or fine-tune the backbone~\citep{liu2023llava,alayrac2022flamingo,chen2023pali}.
Training a vision encoder into a frozen text-only language model produces meaningful captions
\citep{tsimpoukelli2021multimodal}, even when the vision encoder is also frozen and only a single linear map is trained \citep{merullo2023linearly};
our speech adapter is an audio counterpart of these linear vision-to-language maps, with input and output normalizations added.
Projected visual tokens can enter a language model at norms far above those of text tokens; \citet{li2026unseen} align them with a layer norm after the projector, initialized so that its output has the mean text-embedding norm, which is also how we initialize the adapter's output normalization (Section~\ref{sec:setup}).

\paragraph{Projector-only speech recognition into frozen language models.}
Frozen speech encoders and frozen language models have been connected via stacked-frame MLP, cross-attention, and Q-Former modules~\citep{yu2024connecting}, with SLAM-ASR~\citep{ma2024embarrassingly} showing that a two-layer projector alone gives strong ASR ($k=5$ frame concatenation followed by $\mathrm{Linear}(\mathrm{ReLU}(\mathrm{Linear}(\cdot)))$ with a 2048-unit hidden layer). Subsequent work explored mixture-of-expert projectors~\citep{pandey2026dynamic} and zero-shot spoken language understanding with single-matrix projectors~\citep{rios2026zeroshot}, whereas multi-component speech- and audio-language models train parameters beyond the projector~\citep{fathullah2024prompting,tang2024salmonn,chu2023qwenaudio}. Unlike stacked, nonlinear, or multi-component designs, our adapter maps each frame through one matrix with no frame stacking and no hidden nonlinearity, its output normalization initialized at the target language model's mean token radius.

\paragraph{Soft-prompt transfer.}
Soft prompts trained for one frozen language model can be recycled to
another via linear projections fit on shared embeddings
\citep{lester2022reducing} or via a supervised cross-model projector
\citep{su2022transferability};
within a single model, prompts transfer across tasks
\citep{vu2022spot}.
Our composition test
(Section~\ref{sec:composition}) asks a different question:
whether a bridge between two language models, trained on one
adapter's continuous outputs, generalizes to a different adapter's
outputs.
A continuous prompt can solve a task while its nearest tokens spell out
an unrelated text \citep{khashabi2022prompt}; our adapters' soft prompts
are far from every token row (Section~\ref{sec:why}).
Connectors that write each output as a weighted combination of vocabulary
embeddings have been proposed for images \citep{liao2025langbridge} and,
with a codebook initialized from the embedding table, for speech
\citep{yang2025bridging}; \citet{liao2025langbridge} reuse such a connector
with a different language model without repeating its alignment
pretraining (a fine-tuning stage with the new model follows).
Our adapters are not constrained to the vocabulary, and our negative
result for token-fit bridges applies to such unconstrained adapters.
For vision, \citet{zhang2023vpgtrans} move a visual prompt generator
to another language model, start its projector from a word-embedding
converter fit on shared tokens, and still train on paired data.
For speech, we test each step separately and then their composition,
and locate where it stops working.

%% file: sections/03_setup.tex
\section{Setup}
\label{sec:setup}

\paragraph{Input-embedding space and token radius.}
Every autoregressive language model $M$ maps each token index $i$ to a
continuous vector $\mathbf{e}^{(M)}_i \in \mathbb{R}^{d_M}$ in its
\textbf{input-embedding space} of width $d_M$: the token's row of the
embedding table multiplied by an architecture-specific embedding scale $s_M$
($s_M = \sqrt{d_M}$ in Gemma~2 and Gemma~3, as stored in \texttt{bfloat16}:
$\GemmaThreeEmbedScaleBf$ for Gemma-3-4B-IT;
$s_M = 1$ in Qwen, SmolLM2, and Phi).
We define the \textbf{mean token radius}
\begin{equation}
    \bar r_M \;=\; \frac{1}{|V_M|}\sum_{i \in V_M}
    \bigl\lVert \mathbf{e}^{(M)}_i \bigr\rVert_2
    \label{eq:token-radius}
\end{equation}
as the average Euclidean norm of the $s_M$-scaled vocabulary rows,
where $V_M$ is the set of token indices we average over
(the non-zero rows among the first $\TokenRadiusMaxRows$ of the
embedding table, or among all rows of a smaller table).
Measured on loaded \texttt{bfloat16} checkpoints, $\bar r_M$ spans
a $\RadiusSpreadFactor\times$ range across our six language models,
from $\RadiusQwenOneFiveB$ (Qwen2.5-1.5B-IT)
to $\RadiusGemmaTwo$ (Gemma-2-2B-IT).
This scale disparity motivates the normalization hierarchy below.

\paragraph{Closed-form maps.}
Given a donor model $A$ and a target model $B$, we extract $N$ shared
non-special tokens (\textbf{anchors}) by rendering each vocabulary piece
through its native detokenizer and matching the resulting UTF-8 strings.
We partition the anchors into $\AnchorTrainPct\%$ train and
$\AnchorTestPct\%$ held-out test.
Row $n$ of $\mathbf{X} \in \mathbb{R}^{N_{\text{tr}} \times d_A}$ is the
donor's embedding $\mathbf{e}^{(A)}$ of the $n$-th training anchor, and
row $n$ of $\mathbf{Y} \in \mathbb{R}^{N_{\text{tr}} \times d_B}$ is the
target's embedding $\mathbf{e}^{(B)}$ of the same token.
A map is a matrix in $\mathbb{R}^{d_A \times d_B}$ that acts on row
vectors from the right.
We solve for two maps in closed form in \texttt{float64}:
\begin{align}
    \mathbf{W}_{\text{LS}} &= \arg\min_{\mathbf{W}}
    \lVert \mathbf{X}\mathbf{W} - \mathbf{Y} \rVert_F^2,
    \label{eq:lstsq} \\
    \mathbf{Q}_{\text{Proc}} &= \arg\max_{\mathbf{Q} \in \mathcal{S}}
    \operatorname{tr}\bigl(\mathbf{Q}^\top \mathbf{X}^\top \mathbf{Y}\bigr)
    = \mathbf{V}\mathbf{U}^\top,
    \label{eq:maps}
\end{align}
where $\lVert\cdot\rVert_F$ is the Frobenius norm,
$\mathbf{Y}^\top \mathbf{X} = \mathbf{U}\bm{\Sigma}\mathbf{V}^\top$
is the thin singular value decomposition, and $\mathcal{S}$ is the set of
semi-orthogonal $d_A \times d_B$ matrices
($\mathbf{Q}\mathbf{Q}^\top = \mathbf{I}_{d_A}$ if $d_A \le d_B$,
$\mathbf{Q}^\top\mathbf{Q} = \mathbf{I}_{d_B}$ otherwise).
For $d_A \le d_B$, $\mathbf{Q}_{\text{Proc}}$ is the orthogonal Procrustes
solution \citep{schonemann1966generalized}: it also minimizes
$\lVert \mathbf{X}\mathbf{Q} - \mathbf{Y} \rVert_F^2$ over $\mathcal{S}$.
A matrix drawn Haar-uniformly from $\mathcal{S}$
(\textbf{random orthogonal}) serves as a negative control.
Unlike orthogonal Procrustes, unconstrained least squares rescales principal
axes anisotropically, shrinking noisy directions.

\paragraph{Output normalizations.}
To disentangle angular alignment from radial scale, we apply one of five
normalizations $\mathcal{N}$ to each mapped vector
$\mathbf{v} \in \mathbb{R}^{d}$ ($d$ is the width of the space the map
writes into), in order of increasing capacity:
\textbf{none} ($\mathbf{v}$; $0$ parameters);
\textbf{unit $\ell_2$ norm}
($\mathbf{v}/\lVert\mathbf{v}\rVert_2$; $0$ parameters, projects onto the
unit sphere);
\textbf{target-radius norm}
($g\,\mathbf{v}/\lVert\mathbf{v}\rVert_2$; $1$ scalar $g$, rescales unit
vectors to the target's mean token radius);
\textbf{RMS norm}
($\bm{\gamma} \odot \mathbf{v}/\mathrm{rms}(\mathbf{v})$; $d$ per-channel
gains $\bm{\gamma}$); and
\textbf{layer norm}
($\bm{\gamma} \odot (\mathbf{v} - \mu(\mathbf{v}))/\sigma(\mathbf{v}) + \bm{\beta}$;
$2d$ parameters, gains $\bm{\gamma}$ and offsets $\bm{\beta}$).
Here $\mathrm{rms}$, $\mu$, and $\sigma$ are the root mean square, mean,
and standard deviation over the $d$ coordinates of $\mathbf{v}$, and
$\odot$ is the coordinate-wise product.
In closed-form vocabulary transplants, the parameters are calibrated on
the target's training anchors $\mathbf{Y}$ with zero gradient steps:
$g$ is the mean row norm of $\mathbf{Y}$ (the mean token radius of
Eq.~\ref{eq:token-radius} restricted to the anchors),
$\bm{\gamma}$ is the per-channel RMS (RMS norm) or standard deviation
(layer norm) of $\mathbf{Y}$, and $\bm{\beta}$ is its per-channel mean.
In trained speech adapters and trained bridges, all normalization
parameters are learned.

\paragraph{Speech adapter architecture.}
A speech adapter maps each output frame
$\mathbf{h}_E \in \mathbb{R}^{d_E}$ of a frozen speech encoder $E$
to one soft token $\hat{\mathbf{z}} \in \mathbb{R}^{d_M}$ in a frozen
language model $M$'s input-embedding space:
\begin{equation}
    \hat{\mathbf{z}} \;=\; \mathrm{Norm}_{\text{out}}\bigl(
    \mathrm{Norm}_{\text{in}}(\mathbf{h}_E)\,\mathbf{W} + \mathbf{b}\bigr),
    \label{eq:adapter}
\end{equation}
a single matrix $\mathbf{W} \in \mathbb{R}^{d_E \times d_M}$ with a bias
$\mathbf{b} \in \mathbb{R}^{d_M}$ and input and output normalizations.
$\mathrm{Norm}_{\text{in}}$ and $\mathrm{Norm}_{\text{out}}$ are the same
one of the five normalizations above (layer norm unless stated otherwise),
over $d_E$ and $d_M$ coordinates.
Throughout, we call an adapter or a bridge \emph{linear} when it has this
form: one weight matrix and no hidden layer, with no nonlinearity other
than its input and output normalizations.
We initialize $\mathrm{Norm}_{\text{out}}$ at the target's mean token
radius $\bar r_M$ (Eq.~\ref{eq:token-radius}):
$g = \bar r_M$ for the target-radius norm, and
$\bm{\gamma} = (\bar r_M / \sqrt{d_M})\,\mathbf{1}$ for RMS norm and layer norm
(\citet{li2026unseen} initialize a layer norm after a visual projector in the same way).
Adapters therefore start at unit ratio to the target language model's
native token embeddings (the variants with no normalization or with unit
$\ell_2$ norm have no scale to set).
Unlike the SLAM-ASR projector \citep{ma2024embarrassingly},
which concatenates $k = \SlamAsrFrameStack$ consecutive frames and applies
$\mathrm{Linear}(\mathrm{ReLU}(\mathrm{Linear}(\cdot)))$
with a $\SlamAsrHiddenDim$-unit hidden layer,
our adapter uses no frame stacking and no hidden nonlinearity.

\paragraph{Models and data.}
We use six frozen language models from four families (Gemma-3-4B-IT~\citep{gemma2025gemma3}, Gemma-2-2B-IT~\citep{gemma2024gemma2}, Qwen2.5-1.5B-IT~\citep{yang2024qwen25}, Qwen3-4B~\citep{qwen2025qwen3}, Phi-2~\citep{li2023phi2}, and SmolLM2-1.7B-IT~\citep{allal2025smollm2}) and two frozen speech encoders (Whisper-Small~\citep{radford2023whisper} and Moonshine-Base~\citep{jeffries2024moonshine}).
Gemma-3-4B-IT is a vision-language model: its training included images, encoded by a SigLIP encoder and supplied in place of token embeddings \citep{gemma2025gemma3}; we use only its text interface.
Gemma-2-2B-IT and Qwen2.5-1.5B-IT, the other two language models that receive speech, are text-only.
Speech training and evaluation use the $\LibriTrainHours$-hour
LibriSpeech corpus \citep{panayotov2015librispeech} (test-clean and test-other).
Text reasoning is evaluated zero-shot on
PIQA \citep{bisk2020piqa} ($\PiqaEvalItems$ items, $\PiqaChancePct\%$ chance)
and a $\HsEvalItems$-item subset of HellaSwag  \citep{zellers2019hellaswag}
($\HsTotalEvalItems$ items total, $\HsChancePct\%$ chance)
by mean continuation log-probability.
All speech decoding is greedy;
all soft-prompt inputs are slotted into each language model's native
chat template.
Protocol details are in Appendix~\ref{app:protocol}.

\paragraph{What is frozen and what is trained.}
In vocabulary transplants (Section~\ref{sec:tokens}),
nothing is trained: maps are closed-form and normalizations
are calibrated from anchor statistics.
In speech adapters (Section~\ref{sec:speech}),
only the matrix, its bias, and the two normalization layers are trained;
the speech encoder and the language model are both frozen.
In composition (Section~\ref{sec:composition}),
the bridge $A \to B$ is trained on inputs from model $A$'s
embedding space with model $B$ frozen;
the speech adapter into $A$ is also frozen.

%% file: sections/04_tokens.tex
\section{Tokens: Frankenstein Vocabulary Transplants}
\label{sec:tokens}

\input{tables/tab_ws1_s2}

We overwrite the shared-vocabulary rows of a frozen language model's
input embedding table with rows mapped from a donor model, with no
gradient updates.
We call this procedure a \textbf{Frankenstein vocabulary transplant}:
for every shared token, the donor's embedding is mapped, normalized,
and written into the target's table.
Gemma-3-4B-IT shares one matrix between its input embedding table and
its output classifier (\texttt{lm\_head}).
We untie the two and overwrite the input table only; the output
classifier keeps its native rows
(tying it to the transplanted table lowers PIQA by
$\WsOneUntiedVsTiedMinGainPiqa$--$\WsOneUntiedVsTiedMaxGainPiqa$\,pp;
Table~\ref{tab:ws1_5norm}).

\paragraph{Protocol.}
We transplant into frozen Gemma-3-4B-IT from $\WsOneDonorsCount$
independently pretrained donors spanning three families and both
matching and mismatched widths:
Qwen3-4B ($\WsOneQwenLstsqScaledAnchors$ shared anchors),
SmolLM2-1.7B-IT ($d = \SmolDim \to \GemmaThreeDim$;
$\WsOneSmolLstsqScaledAnchors$ anchors),
and Phi-2 ($\WsOnePhiLstsqScaledAnchors$ anchors).
An identity gate ($\mathbf{W} = \mathbf{I}$, donor $=$ target)
verifies zero degradation from the surgery write path;
Haar-random orthogonal maps serve as negative controls.
On the evaluation text, $\TokenHitsTabOneExactMinOverwrittenPct\%$--$\TokenHitsTabOneExactMaxOverwrittenPct\%$ of input token occurrences
(by donor and task, for these three donors) fall on overwritten rows, $\TokenHitsTabOneExactMinTrainAnchorPct\%$--$\TokenHitsTabOneExactMaxTrainAnchorPct\%$ on rows
the map was fit on and $\TokenHitsTabOneExactMinHeldoutAnchorPct\%$--$\TokenHitsTabOneExactMaxHeldoutAnchorPct\%$ on held-out rows; the other
occurrences use the target's native rows (Table~\ref{tab:token_hits}).
Overwriting the same rows with an unaligned or a random map drops the
model to chance (below), so the overwritten rows carry the task.

\paragraph{Least squares plus one scalar preserves function.}
The identity gate reproduces native Gemma-3-4B-IT to
$\WsOneGateDeltaPiqa$\,pp on both benchmarks
($\WsOneGatePiqa\%$ PIQA vs.\ $\WsOneNativePiqa\%$ native;
$\WsOneGateHs\%$ vs.\ $\WsOneNativeHs\%$ HellaSwag).
Across real donors, closed-form least squares with target-radius
normalization retains
$\WsOneQwenLstsqScaledPiqa\%$ PIQA and
$\WsOneQwenLstsqScaledHs\%$ HellaSwag on Qwen3-4B,
$\WsOnePhiLstsqScaledPiqa\%$ and
$\WsOnePhiLstsqScaledHs\%$ on Phi-2,
and---across a mismatched width ($\SmolDim \to \GemmaThreeDim$)---$\WsOneSmolLstsqScaledPiqa\%$ PIQA and
$\WsOneSmolLstsqScaledHs\%$ HellaSwag
on SmolLM2-1.7B-IT, all with zero gradient steps.
Random orthogonal controls collapse to chance
($\WsOneSmolRandNonePiqa\%$--$\WsOnePhiRandNonePiqa\%$ PIQA, $\WsOneSmolRandScaledHs\%$--$\WsOnePhiRandNoneHs\%$ HellaSwag).

\begin{figure*}[!t]
\centering
\includegraphics[width=\textwidth]{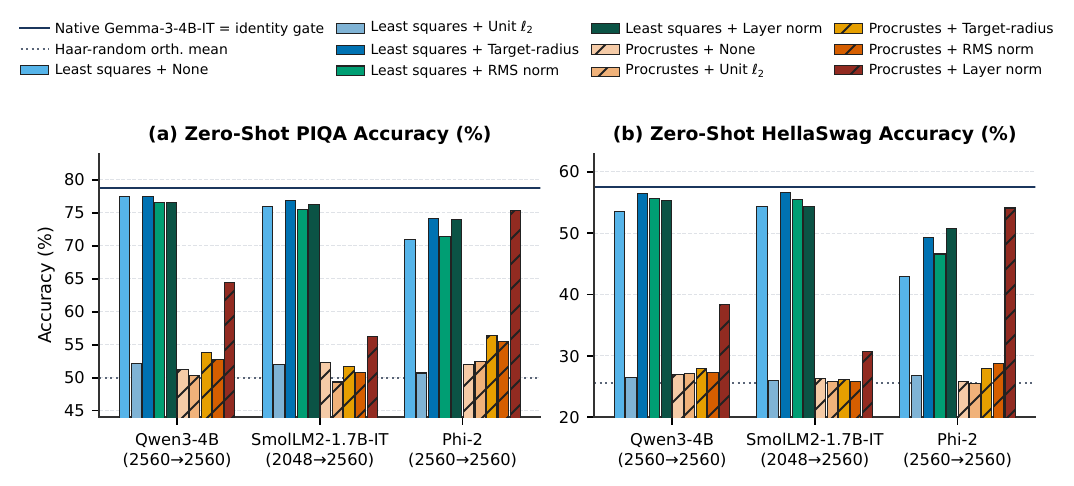}
\caption{\textbf{Zero-shot PIQA (\PiqaEvalItems\ items, accuracy $\uparrow$) and HellaSwag (\HsEvalItems\ items, accuracy $\uparrow$) across the five output normalizations for closed-form vocabulary transplants into frozen Gemma-3-4B-IT (Table~\ref{tab:ws1_5norm}).} Solid dark lines mark unmodified Gemma-3-4B-IT and the identity transplant gate (\WsOneNativePiqa\% PIQA, \WsOneNativeHs\% HellaSwag); dotted lines mark mean Haar-random orthogonal controls. Unit $\ell_2$ normalization without target radius scaling (median radius ratio \WsOneQwenLstsqLtwoRr) drops every donor and solver to chance (\WsOneLtwoMinPiqa\%--\WsOneLtwoMaxPiqa\% PIQA), whereas rescaling by the target mean token radius restores \WsOneSmolLstsqScaledHs\% HellaSwag on SmolLM2-1.7B-IT and \WsOneQwenLstsqScaledPiqa\% PIQA on Qwen3-4B.}
\label{fig:ws1_norms}
\end{figure*}

\paragraph{Direction is not enough.}
Projecting mapped rows onto the unit sphere
(unit $\ell_2$ norm, median radius ratio $\WsOneQwenLstsqLtwoRr$)
drops every donor--map pair to chance
($\WsOneLtwoMinPiqa\%$--$\WsOneLtwoMaxPiqa\%$ PIQA,
$\WsOneLtwoMinHs\%$--$\WsOneLtwoMaxHs\%$ HellaSwag),
although the directions are those of target-radius normalization.
Rescaling the same unit vectors by one scalar, the mean radius $g$ of
the target's anchors, brings least-squares maps back:
$\WsOneLstsqLtwoRestoreMinGainHs$--$\WsOneLstsqLtwoRestoreMaxGainHs$\,pp HellaSwag over unit $\ell_2$ norm.
Least squares with no normalization already carries most of the scale
(radius ratio $\WsOneLstsqNoneMinRr$--$\WsOneLstsqNoneMaxRr$); target-radius normalization adds
$\WsOneQwenLstsqNormGainHs$, $\WsOneSmolLstsqNormGainHs$, and
$\WsOnePhiLstsqNormGainHs$\,pp HellaSwag over it on Qwen3-4B,
SmolLM2-1.7B-IT, and Phi-2
(Table~\ref{tab:ws1_s2}; all five normalizations in
Table~\ref{tab:ws1_5norm} and Figure~\ref{fig:ws1_norms}).
Among least-squares maps, target-radius normalization scores highest in
$\WsOneLstsqScaledTopCells$ of $\WsOneLstsqDonorTaskCells$ donor--task cells (in one of them, PIQA on Qwen3-4B, level with no normalization); in the other, layer norm is
$\WsOneLstsqPhiLnOverScaledHs$\,pp higher (HellaSwag on Phi-2).

\paragraph{Retrieval does not predict function.}
On held-out vocabulary anchors, orthogonal Procrustes achieves
$\WsOneQwenProcScaledPatOne\%$--$\WsOneSmolProcScaledPatOne\%$ top-1
nearest-neighbor retrieval (searched among the held-out anchors)---virtually tied with least squares
($\WsOnePhiLstsqScaledPatOne\%$--$\WsOneSmolLstsqScaledPatOne\%$).
Yet even after target-radius normalization restores exact unit
radius ratio, Procrustes remains near chance on downstream reasoning
($\WsOneSmolProcScaledPiqa\%$--$\WsOnePhiProcScaledPiqa\%$ PIQA,
$\WsOneSmolProcScaledHs\%$--$\WsOneQwenProcScaledHs\%$ HellaSwag).
Layer norm changes this picture for Procrustes:
$\WsOneQwenProcLnPiqa\%$, $\WsOneSmolProcLnPiqa\%$, and
$\WsOnePhiProcLnPiqa\%$ PIQA on Qwen3-4B, SmolLM2-1.7B-IT, and Phi-2,
against $\WsOneQwenProcScaledPiqa\%$, $\WsOneSmolProcScaledPiqa\%$, and
$\WsOnePhiProcScaledPiqa\%$ with target-radius norm
(Table~\ref{tab:ws1_5norm}, Figure~\ref{fig:ws1_norms}).
RMS norm has per-channel gains but no centering of each vector over its
coordinates and no per-channel offset, and gives no such gain
($\WsOneQwenProcRmsPiqa\%$, $\WsOneSmolProcRmsPiqa\%$,
$\WsOnePhiProcRmsPiqa\%$).
The rescue therefore comes with that centering and the offset, which is
the target's mean embedding; a rotation with no intercept cannot supply
the offset. We did not test the two separately, and we did not test a
rotation fit on centered anchors.
On Phi-2, Procrustes with layer norm
($\WsOnePhiProcLnPiqa\%$ PIQA, $\WsOnePhiProcLnHs\%$ HellaSwag)
is above every least-squares configuration.

\paragraph{Same-family confirmation and comparison to trained maps.}
To rule out a cross-family confound, we repeat the protocol on a
same-family pair: Gemma-2-2B-IT $\to$ frozen Gemma-3-4B-IT
($\WsFourDConeLstsqScaledAnchors$ shared anchors).
Closed-form least squares retains
$\WsFourDConeLstsqScaledPiqa\%$ PIQA
($\WsFourDDeltaConeVsBzeroPiqa$\,pp of same-host native,
$\WsFourDConeLstsqScaledAboveChancePiqa$\,pp above chance)
and $\WsFourDConeLstsqScaledHs\%$ HellaSwag
($\WsFourDDeltaConeVsBzeroHs$\,pp of native,
$\WsFourDConeLstsqScaledAboveChanceHs$\,pp above chance).
We then evaluate the seven gradient-trained bridges of Section~\ref{sec:composition}---fit on LibriSpeech
transcripts, WikiText-103 \citep{merity2017pointer},
one adapter's soft prompts (Whisper or Moonshine),
a half-text half-soft-prompt mixture (with either adapter), and
both adapters' soft prompts---as static vocabulary replacements.
Every trained map lands between chance
($\PiqaChancePct\%$ PIQA, $\HsChancePct\%$ HellaSwag) and
$\WsFourDBestTrainedPiqa\%$ PIQA
($\WsFourDBestTrainedAboveChancePiqa$\,pp above chance) /
$\WsFourDBestTrainedHs\%$ HellaSwag
($\WsFourDBestTrainedAboveChanceHs$\,pp above chance;
Table~\ref{tab:ws4d}).
These bridges are trained through the frozen target model on a
sequence loss (for the two text bridges, copying a text supplied in the
soft-prompt slot; Section~\ref{sec:composition}), not on token pairs; as static vocabularies they recover
far less than the closed-form map, including the two fit on text.

%% file: tables/tab_ws1_s2.tex
\begin{table*}[t]
\caption{\textbf{Closed-form Frankenstein vocabulary transplants into frozen Gemma-3-4B-IT (\WsOneDoneCells\ configurations; full five-normalization grid in Table~\ref{tab:ws1_5norm}).} Linear maps are fit without gradient updates on $\AnchorTrainPct\%$ of exact-string shared vocabulary tokens and evaluated on the held-out $\AnchorTestPct\%$ split (mean cosine similarity $\uparrow$, nearest-neighbor retrieval P@1 among the held-out anchors $\uparrow$, median output-to-target radius ratio; closer to $\WsOneNativeRr$ matches native scale) and zero-shot multiple-choice accuracy ($\uparrow$) on PIQA (\PiqaEvalItems\ items, \PiqaChancePct\% chance) and HellaSwag (\HsEvalItems\ items, \HsChancePct\% chance) with the output head kept native. $^\dagger$Uses candidate surface-form matching (Appendix~\ref{app:protocol}).}
\label{tab:ws1_s2}
\centering
\resizebox{\textwidth}{!}{%
\setlength{\tabcolsep}{4.0pt}
\begin{tabular}{@{}lllcrrrrrr@{}}
\toprule
\textbf{Donor LLM} & \textbf{Linear Map} & \textbf{Output Norm} & \textbf{Width} & \textbf{Shared Tokens} & \textbf{Held-out $\cos\uparrow$} & \textbf{P@1 $\uparrow$} & \textbf{Radius Ratio} & \textbf{PIQA $\uparrow$} & \textbf{HellaSwag $\uparrow$} \\
\midrule
\multicolumn{10}{@{}l}{\textit{Reference model and transplant invariance gate (target: Gemma-3-4B-IT)}} \\
Native Gemma-3-4B-IT (unmodified) & --- & --- & $\GemmaThreeDim$ & --- & --- & --- & \WsOneNativeRr & \textbf{\WsOneNativePiqa\%} & \textbf{\WsOneNativeHs\%} \\
Identity gate (target $\to$ target) & Identity & None & $\GemmaThreeDim\!\to\!\GemmaThreeDim$ & \WsOneGateAnchors & \WsOneGateCos & \WsOneGatePatOne\% & \WsOneGateRr & \textbf{\WsOneGatePiqa\%} & \textbf{\WsOneGateHs\%} \\
\midrule
\multicolumn{10}{@{}l}{\textit{Closed-form cross-model vocabulary transplants (donor $\to$ frozen Gemma-3-4B-IT, untied output head)}} \\
Qwen3-4B & Least squares & Target-radius norm & $\QwenThreeDim\!\to\!\GemmaThreeDim$ & \WsOneQwenLstsqScaledAnchors & \WsOneQwenLstsqScaledCos & \WsOneQwenLstsqScaledPatOne\% & \WsOneQwenLstsqScaledRr & \textbf{\WsOneQwenLstsqScaledPiqa\%} & \WsOneQwenLstsqScaledHs\% \\
Qwen3-4B & Least squares & None & $\QwenThreeDim\!\to\!\GemmaThreeDim$ & \WsOneQwenLstsqNoneAnchors & \WsOneQwenLstsqNoneCos & \WsOneQwenLstsqNonePatOne\% & \WsOneQwenLstsqNoneRr & \textbf{\WsOneQwenLstsqNonePiqa\%} & \WsOneQwenLstsqNoneHs\% \\
Qwen3-4B (surface matcher$^\dagger$) & Least squares & None & $\QwenThreeDim\!\to\!\GemmaThreeDim$ & \WsOneQwenClReplicaAnchors & \WsOneQwenClReplicaCos & \WsOneQwenClReplicaPatOne\% & \WsOneQwenClReplicaRr & \WsOneQwenClReplicaPiqa\% & \WsOneQwenClReplicaHs\% \\
Qwen3-4B & Orthogonal Procrustes & Target-radius norm & $\QwenThreeDim\!\to\!\GemmaThreeDim$ & \WsOneQwenProcScaledAnchors & \WsOneQwenProcScaledCos & \WsOneQwenProcScaledPatOne\% & \WsOneQwenProcScaledRr & \WsOneQwenProcScaledPiqa\% & \WsOneQwenProcScaledHs\% \\
Qwen3-4B & Orthogonal Procrustes & None & $\QwenThreeDim\!\to\!\GemmaThreeDim$ & \WsOneQwenProcNoneAnchors & \WsOneQwenProcNoneCos & \WsOneQwenProcNonePatOne\% & \WsOneQwenProcNoneRr & \WsOneQwenProcNonePiqa\% & \WsOneQwenProcNoneHs\% \\
\cmidrule(lr){1-10}
SmolLM2-1.7B-IT & Least squares & Target-radius norm & $\SmolDim\!\to\!\GemmaThreeDim$ & \WsOneSmolLstsqScaledAnchors & \WsOneSmolLstsqScaledCos & \WsOneSmolLstsqScaledPatOne\% & \WsOneSmolLstsqScaledRr & \WsOneSmolLstsqScaledPiqa\% & \textbf{\WsOneSmolLstsqScaledHs\%} \\
SmolLM2-1.7B-IT & Least squares & None & $\SmolDim\!\to\!\GemmaThreeDim$ & \WsOneSmolLstsqNoneAnchors & \WsOneSmolLstsqNoneCos & \WsOneSmolLstsqNonePatOne\% & \WsOneSmolLstsqNoneRr & \WsOneSmolLstsqNonePiqa\% & \WsOneSmolLstsqNoneHs\% \\
SmolLM2-1.7B-IT & Orthogonal Procrustes & Target-radius norm & $\SmolDim\!\to\!\GemmaThreeDim$ & \WsOneSmolProcScaledAnchors & \WsOneSmolProcScaledCos & \WsOneSmolProcScaledPatOne\% & \WsOneSmolProcScaledRr & \WsOneSmolProcScaledPiqa\% & \WsOneSmolProcScaledHs\% \\
SmolLM2-1.7B-IT & Orthogonal Procrustes & None & $\SmolDim\!\to\!\GemmaThreeDim$ & \WsOneSmolProcNoneAnchors & \WsOneSmolProcNoneCos & \WsOneSmolProcNonePatOne\% & \WsOneSmolProcNoneRr & \WsOneSmolProcNonePiqa\% & \WsOneSmolProcNoneHs\% \\
\cmidrule(lr){1-10}
Phi-2 & Least squares & Target-radius norm & $\PhiDim\!\to\!\GemmaThreeDim$ & \WsOnePhiLstsqScaledAnchors & \WsOnePhiLstsqScaledCos & \WsOnePhiLstsqScaledPatOne\% & \WsOnePhiLstsqScaledRr & \WsOnePhiLstsqScaledPiqa\% & \WsOnePhiLstsqScaledHs\% \\
Phi-2 & Least squares & None & $\PhiDim\!\to\!\GemmaThreeDim$ & \WsOnePhiLstsqNoneAnchors & \WsOnePhiLstsqNoneCos & \WsOnePhiLstsqNonePatOne\% & \WsOnePhiLstsqNoneRr & \WsOnePhiLstsqNonePiqa\% & \WsOnePhiLstsqNoneHs\% \\
Phi-2 & Orthogonal Procrustes & Target-radius norm & $\PhiDim\!\to\!\GemmaThreeDim$ & \WsOnePhiProcScaledAnchors & \WsOnePhiProcScaledCos & \WsOnePhiProcScaledPatOne\% & \WsOnePhiProcScaledRr & \WsOnePhiProcScaledPiqa\% & \WsOnePhiProcScaledHs\% \\
Phi-2 & Orthogonal Procrustes & None & $\PhiDim\!\to\!\GemmaThreeDim$ & \WsOnePhiProcNoneAnchors & \WsOnePhiProcNoneCos & \WsOnePhiProcNonePatOne\% & \WsOnePhiProcNoneRr & \WsOnePhiProcNonePiqa\% & \WsOnePhiProcNoneHs\% \\
\midrule
\multicolumn{10}{@{}l}{\textit{Negative controls: unaligned direct swap and Haar-random orthogonal map}} \\
Qwen3-4B & Direct swap & Target-radius norm & $\QwenThreeDim\!\to\!\GemmaThreeDim$ & \WsOneQwenSwapScaledAnchors & \WsOneQwenSwapScaledCos & \WsOneQwenSwapScaledPatOne\% & \WsOneQwenSwapScaledRr & \WsOneQwenSwapScaledPiqa\% & \WsOneQwenSwapScaledHs\% \\
Qwen3-4B & Direct swap & None & $\QwenThreeDim\!\to\!\GemmaThreeDim$ & \WsOneQwenSwapNoneAnchors & \WsOneQwenSwapNoneCos & \WsOneQwenSwapNonePatOne\% & \WsOneQwenSwapNoneRr & \WsOneQwenSwapNonePiqa\% & \WsOneQwenSwapNoneHs\% \\
Phi-2 & Direct swap & Target-radius norm & $\PhiDim\!\to\!\GemmaThreeDim$ & \WsOnePhiSwapScaledAnchors & \WsOnePhiSwapScaledCos & \WsOnePhiSwapScaledPatOne\% & \WsOnePhiSwapScaledRr & \WsOnePhiSwapScaledPiqa\% & \WsOnePhiSwapScaledHs\% \\
Phi-2 & Direct swap & None & $\PhiDim\!\to\!\GemmaThreeDim$ & \WsOnePhiSwapNoneAnchors & \WsOnePhiSwapNoneCos & \WsOnePhiSwapNonePatOne\% & \WsOnePhiSwapNoneRr & \WsOnePhiSwapNonePiqa\% & \WsOnePhiSwapNoneHs\% \\
Qwen3-4B & Random orthogonal & Target-radius norm & $\QwenThreeDim\!\to\!\GemmaThreeDim$ & \WsOneQwenRandScaledAnchors & \WsOneQwenRandScaledCos & \WsOneQwenRandScaledPatOne\% & \WsOneQwenRandScaledRr & \WsOneQwenRandScaledPiqa\% & \WsOneQwenRandScaledHs\% \\
Qwen3-4B & Random orthogonal & None & $\QwenThreeDim\!\to\!\GemmaThreeDim$ & \WsOneQwenRandNoneAnchors & \WsOneQwenRandNoneCos & \WsOneQwenRandNonePatOne\% & \WsOneQwenRandNoneRr & \WsOneQwenRandNonePiqa\% & \WsOneQwenRandNoneHs\% \\
SmolLM2-1.7B-IT & Random orthogonal & Target-radius norm & $\SmolDim\!\to\!\GemmaThreeDim$ & \WsOneSmolRandScaledAnchors & \WsOneSmolRandScaledCos & \WsOneSmolRandScaledPatOne\% & \WsOneSmolRandScaledRr & \WsOneSmolRandScaledPiqa\% & \WsOneSmolRandScaledHs\% \\
SmolLM2-1.7B-IT & Random orthogonal & None & $\SmolDim\!\to\!\GemmaThreeDim$ & \WsOneSmolRandNoneAnchors & \WsOneSmolRandNoneCos & \WsOneSmolRandNonePatOne\% & \WsOneSmolRandNoneRr & \WsOneSmolRandNonePiqa\% & \WsOneSmolRandNoneHs\% \\
Phi-2 & Random orthogonal & Target-radius norm & $\PhiDim\!\to\!\GemmaThreeDim$ & \WsOnePhiRandScaledAnchors & \WsOnePhiRandScaledCos & \WsOnePhiRandScaledPatOne\% & \WsOnePhiRandScaledRr & \WsOnePhiRandScaledPiqa\% & \WsOnePhiRandScaledHs\% \\
Phi-2 & Random orthogonal & None & $\PhiDim\!\to\!\GemmaThreeDim$ & \WsOnePhiRandNoneAnchors & \WsOnePhiRandNoneCos & \WsOnePhiRandNonePatOne\% & \WsOnePhiRandNoneRr & \WsOnePhiRandNonePiqa\% & \WsOnePhiRandNoneHs\% \\
\bottomrule
\end{tabular}%
}
\end{table*}

%% file: sections/05_speech.tex
\section{Speech: One Linear Layer}
\label{sec:speech}

\input{tables/tab_ws2_asr}

We train a single linear matrix between a frozen speech encoder and a
frozen language model on $\LibriTrainHours$ hours of LibriSpeech
\citep{panayotov2015librispeech}, with no frame stacking, no hidden
nonlinearity, and no fine-tuning of either model.
Unlike the SLAM-ASR projector
\citep{ma2024embarrassingly}, which concatenates
$k = \SlamAsrFrameStack$ frames and applies a
$\SlamAsrHiddenDim$-unit hidden layer with ReLU,
our adapter contains only
$\mathrm{Norm}_{\text{in}} \to \mathrm{Linear} \to \mathrm{Norm}_{\text{out}}$ (Eq.~\ref{eq:adapter}), with $\mathrm{Norm}_{\text{out}}$ initialized at the target's mean token
radius $\bar r_M$.

\paragraph{A single matrix suffices.}
On Whisper-Small $\to$ Gemma-3-4B-IT, the best adapter reaches
\WsTwoBestMuonRunWerPairCell\ WER (\texttt{test-clean} /
\texttt{test-other}), surpassing Whisper-Small's own pretrained
decoder ($\NativeWhisperWerClean\%$ / $\NativeWhisperWerOther\%$).
The adapter is trained on LibriSpeech and feeds a language model of
1.5B--4B parameters, whereas Whisper-Small was not trained on
LibriSpeech, so the comparison shows that a linear interface is
enough, not that it is the better recognizer.
Our scoring applies the same minimal text normalization to every system
(Appendix~\ref{app:protocol}); it counts the numerals and abbreviations
that Whisper's decoder writes as errors.
With Whisper's own English text normalizer \citep{radford2023whisper}
applied to the same transcripts, the native decoder scores
$\WnormNativeWhisperClean\%$ / $\WnormNativeWhisperOther\%$ and the adapter
$\WnormWhisperGemmaThreeMuonClean\%$ / $\WnormWhisperGemmaThreeMuonOther\%$:
the adapter stays below on both splits, with a smaller margin.
Under our normalization, Whisper-Small adapters are below the native decoder into both Gemma
models and above it into Qwen2.5-1.5B-IT
($\WsTwoWhisperQwenLnMuonWerClean\%$ /
$\WsTwoWhisperQwenLnMuonWerOther\%$); under Whisper's normalizer, the
adapter into Gemma-2-2B-IT ($\WnormWhisperGemmaTwoMuonClean\%$ /
$\WnormWhisperGemmaTwoMuonOther\%$) is below it on \texttt{test-clean} only.
No Moonshine-Base adapter
(best $\WsTwoMoonshineGemmaThreeLnMuonWerClean\%$ /
$\WsTwoMoonshineGemmaThreeLnMuonWerOther\%$) reaches Moonshine's own
decoder ($\NativeMoonshineWerClean\%$ / $\NativeMoonshineWerOther\%$)
under either normalization; Moonshine's training data include LibriSpeech.
The same architecture trains on all six encoder--LLM pairs
(Table~\ref{tab:ws2_asr}).

\paragraph{Optimizer: Muon vs.\ AdamW.}
We compare Muon \citep{jordan2024muon}, which orthogonalizes
weight updates via a five-step quintic Newton--Schulz iteration,
against AdamW \citep{loshchilov2018decoupled} with the same schedule
and one fixed learning-rate setting per optimizer
(Appendix~\ref{app:protocol}); neither was tuned.
The two settings also differ in weight decay on the matrix
($0.01$ for AdamW, none for Muon; Table~\ref{tab:protocol_optim}).
Muon has lower \texttt{test-clean} WER on
$\WsTwoMuonWinsClean / \WsTwoPairsDone$ pairs
(by $\MuonCleanMarginMin$--$\MuonCleanMarginMax$\,pp; mean difference $\WsTwoMeanMuonDeltaClean$\,pp; single training seeds, with the two runs of a pair on different hosts, Appendix~\ref{app:reproducibility}).
On \texttt{test-other} it is lower on $\WsTwoMuonLowerOther$, tied on $\WsTwoMuonTiedOther$
(within our $\MixVsAudioTieCriterionPp$\,pp criterion), and higher on
$\WsTwoMuonHigherOther$ of $\WsTwoPairsDone$ pairs.
The learned matrices differ in spectrum
(Figure~\ref{fig:adapter_geometry}).
Muon has higher Shannon effective rank on every pair
($\WsTwoMuonMinErankPct\%$--$\WsTwoMuonMaxErankPct\%$ of the maximum, against
$\WsTwoAdamwMinErankPct\%$--$\WsTwoAdamwMaxErankPct\%$ for AdamW).
Its condition number is lower on the $\WsTwoWhisperMuonLowerKappa$ Whisper-Small pairs and
$\WsTwoMoonshineMinKappaRatio$--$\WsTwoMoonshineMaxKappaRatio$ times higher on the three Moonshine-Base pairs, where one direction sets it.
In every Muon adapter, the smallest singular value belongs to the one input direction in which the layer-normalized frames do not vary (cosine $\MuonLastDirCosMin$--$\MuonLastDirCosMax$ in absolute value; at most $\AdamwLastDirCosMax$ for AdamW).
Over the other directions, the condition number is $\MuonActiveKappaMin$--$\MuonActiveKappaMax$ for Muon and $\AdamwActiveKappaMin$--$\AdamwActiveKappaMax$ for AdamW, lower for Muon on $\MuonActiveKappaLowerPairs$ of $\WsTwoPairsDone$ pairs.
Neither optimizer collapses the spectrum.
These are spectra of $\mathbf{W}$ alone, without the normalization gains,
and the two settings differ in weight decay as well as in update rule, so
we do not attribute the difference to orthogonalized updates alone.

\paragraph{A length fixed at one hurts; a learned length is enough.}
Table~\ref{tab:ws2_asr} varies the normalization on both sides of the
matrix ($\mathrm{Norm}_{\text{in}}$ and $\mathrm{Norm}_{\text{out}}$
change together, so the two sides are not separated).
Unit $\ell_2$ normalization, which fixes the length at one on both sides,
is the worst variant on both targets:
$\WsTwoWhisperGemmaTwoLtwoMuonWerClean\%$ /
$\WsTwoWhisperGemmaTwoLtwoMuonWerOther\%$ WER on Gemma-2
($\bar r_M = \RadiusGemmaTwo$),
$\WsTwoNormDeltaWhisperGemmaTwoLtwoVsLnMuonClean$\,pp above layer norm,
and $\WsTwoWhisperQwenLtwoMuonWerClean\%$ on Qwen2.5
against $\WsTwoWhisperQwenLnMuonWerClean\%$.
On Qwen2.5 a length of one is close to the token radius
($\bar r_M = \RadiusQwenOneFiveB$), and the variant still loses to
no normalization ($\WsTwoWhisperQwenNoneMuonWerClean\%$):
matching the token radius is not what matters there.
A learned length is enough: target-radius normalization, which gives
every frame the same length and learns that length (one scalar on each
side), ties layer norm on both targets
($\WsTwoWhisperQwenScaledMuonWerClean\%$ on Qwen2.5,
$\WsTwoWhisperGemmaTwoScaledMuonWerClean\%$ on Gemma-2; within our
$\MixVsAudioTieCriterionPp$\,pp criterion).
Training does not keep the outputs at the token radius on every target:
adapter outputs have $\ManifoldSrcGemmaMinRrMed$--$\ManifoldSrcGemmaMaxRrMed$ times the mean token radius on
the Gemma models and $\ManifoldSrcQwenMinRrMed$--$\ManifoldSrcQwenMaxRrMed$ times on Qwen2.5
(Figure~\ref{fig:manifold_distance}h).

%% file: tables/tab_ws2_asr.tex
\begin{table*}[t]
\caption{\textbf{Single-linear-layer speech adapters on LibriSpeech (\LibriTrainUtts\ training utterances, \WsTwoUniformEpochs\ epochs, \WsTwoTotalSteps\ optimizer steps).} Both the speech encoder and the language model remain frozen; only the normalized linear projection ($\mathrm{Norm}_{\text{in}} \to \mathrm{Linear} \to \mathrm{Norm}_{\text{out}}$, where input and output normalization change together) is trained. Validation negative log-likelihood ($\downarrow$) is comparable within each target language model's tokenizer. Word error rate (WER $\downarrow$, greedy decoding) is reported on LibriSpeech \texttt{test-clean} (\LibriTestCleanUtts\ utterances) and \texttt{test-other} (\LibriTestOtherUtts\ utterances); within each block, the lowest WER and every value within $\MixVsAudioTieCriterionPp$\,pp of it are shown in bold. $^\ddagger$Same run as in the paired optimizer block above.}
\label{tab:ws2_asr}
\centering
\resizebox{\textwidth}{!}{%
\setlength{\tabcolsep}{4.2pt}
\begin{tabular}{@{}llcllrrr@{}}
\toprule
\textbf{Speech Encoder} & \textbf{Frozen LLM / Decoder} & \textbf{Width} & \textbf{Normalization} & \textbf{Optimizer} & \textbf{Params} & \textbf{Best Val NLL $\downarrow$} & \textbf{WER (\texttt{clean} / \texttt{other}) $\downarrow$} \\
\midrule
\multicolumn{8}{@{}l}{\textit{Reference: speech encoders decoded by their own pretrained sequence-to-sequence decoders (same text normalization)}} \\
Whisper-Small & Native Whisper decoder & --- & --- & Pretrained & --- & --- & \NativeWhisperWerClean\% / \NativeWhisperWerOther\% \\
Moonshine-Base & Native Moonshine decoder & --- & --- & Pretrained & --- & --- & \NativeMoonshineWerClean\% / \NativeMoonshineWerOther\% \\
\midrule
\multicolumn{8}{@{}l}{\textit{Paired optimizer comparison (layer norm initialized at target mean token radius, \WsTwoUniformEpochs\ epochs on LibriSpeech)}} \\
Whisper-Small & Qwen2.5-1.5B-IT & $\WhisperDim\!\to\!\QwenDim$ & Layer norm & Muon & \WsTwoWhisperQwenLnMuonParamsM & \WsTwoWhisperQwenLnMuonBestValCell & \textbf{\WsTwoWhisperQwenLnMuonWerCleanCell} / \textbf{\WsTwoWhisperQwenLnMuonWerOtherCell} \\
Whisper-Small & Qwen2.5-1.5B-IT & $\WhisperDim\!\to\!\QwenDim$ & Layer norm & AdamW & \WsTwoWhisperQwenLnAdamwParamsM & \WsTwoWhisperQwenLnAdamwBestValCell & \WsTwoWhisperQwenLnAdamwWerCleanCell{} / \WsTwoWhisperQwenLnAdamwWerOtherCell \\
\cmidrule(lr){1-8}
Moonshine-Base & Qwen2.5-1.5B-IT & $\MoonshineDim\!\to\!\QwenDim$ & Layer norm & Muon & \WsTwoMoonshineQwenLnMuonParamsM & \WsTwoMoonshineQwenLnMuonBestValCell & \textbf{\WsTwoMoonshineQwenLnMuonWerCleanCell} / \textbf{\WsTwoMoonshineQwenLnMuonWerOtherCell} \\
Moonshine-Base & Qwen2.5-1.5B-IT & $\MoonshineDim\!\to\!\QwenDim$ & Layer norm & AdamW & \WsTwoMoonshineQwenLnAdamwParamsM & \WsTwoMoonshineQwenLnAdamwBestValCell & \WsTwoMoonshineQwenLnAdamwWerCleanCell{} / \textbf{\WsTwoMoonshineQwenLnAdamwWerOtherCell} \\
\cmidrule(lr){1-8}
Moonshine-Base & Gemma-2-2B-IT & $\MoonshineDim\!\to\!\GemmaTwoDim$ & Layer norm & Muon & \WsTwoMoonshineGemmaTwoLnMuonParamsM & \WsTwoMoonshineGemmaTwoLnMuonBestValCell & \textbf{\WsTwoMoonshineGemmaTwoLnMuonWerCleanCell} / \textbf{\WsTwoMoonshineGemmaTwoLnMuonWerOtherCell} \\
Moonshine-Base & Gemma-2-2B-IT & $\MoonshineDim\!\to\!\GemmaTwoDim$ & Layer norm & AdamW & \WsTwoMoonshineGemmaTwoLnAdamwParamsM & \WsTwoMoonshineGemmaTwoLnAdamwBestValCell & \WsTwoMoonshineGemmaTwoLnAdamwWerCleanCell{} / \WsTwoMoonshineGemmaTwoLnAdamwWerOtherCell \\
\cmidrule(lr){1-8}
Whisper-Small & Gemma-2-2B-IT & $\WhisperDim\!\to\!\GemmaTwoDim$ & Layer norm & Muon & \WsTwoWhisperGemmaTwoLnMuonParamsM & \WsTwoWhisperGemmaTwoLnMuonBestValCell & \textbf{\WsTwoWhisperGemmaTwoLnMuonWerCleanCell} / \WsTwoWhisperGemmaTwoLnMuonWerOtherCell \\
Whisper-Small & Gemma-2-2B-IT & $\WhisperDim\!\to\!\GemmaTwoDim$ & Layer norm & AdamW & \WsTwoWhisperGemmaTwoLnAdamwParamsM & \WsTwoWhisperGemmaTwoLnAdamwBestValCell & \WsTwoWhisperGemmaTwoLnAdamwWerCleanCell{} / \textbf{\WsTwoWhisperGemmaTwoLnAdamwWerOtherCell} \\
\cmidrule(lr){1-8}
Moonshine-Base & Gemma-3-4B-IT & $\MoonshineDim\!\to\!\GemmaThreeDim$ & Layer norm & Muon & \WsTwoMoonshineGemmaThreeLnMuonParamsM & \WsTwoMoonshineGemmaThreeLnMuonBestValCell & \textbf{\WsTwoMoonshineGemmaThreeLnMuonWerCleanCell} / \textbf{\WsTwoMoonshineGemmaThreeLnMuonWerOtherCell} \\
Moonshine-Base & Gemma-3-4B-IT & $\MoonshineDim\!\to\!\GemmaThreeDim$ & Layer norm & AdamW & \WsTwoMoonshineGemmaThreeLnAdamwParamsM & \WsTwoMoonshineGemmaThreeLnAdamwBestValCell & \WsTwoMoonshineGemmaThreeLnAdamwWerCleanCell{} / \WsTwoMoonshineGemmaThreeLnAdamwWerOtherCell \\
\cmidrule(lr){1-8}
Whisper-Small & Gemma-3-4B-IT & $\WhisperDim\!\to\!\GemmaThreeDim$ & Layer norm & Muon & \WsTwoWhisperGemmaThreeLnMuonParamsM & \WsTwoWhisperGemmaThreeLnMuonBestValCell & \textbf{\WsTwoWhisperGemmaThreeLnMuonWerCleanCell} / \WsTwoWhisperGemmaThreeLnMuonWerOtherCell \\
Whisper-Small & Gemma-3-4B-IT & $\WhisperDim\!\to\!\GemmaThreeDim$ & Layer norm & AdamW & \WsTwoWhisperGemmaThreeLnAdamwParamsM & \WsTwoWhisperGemmaThreeLnAdamwBestValCell & \WsTwoWhisperGemmaThreeLnAdamwWerCleanCell{} / \textbf{\WsTwoWhisperGemmaThreeLnAdamwWerOtherCell} \\
\midrule
\multicolumn{8}{@{}l}{\textit{Normalization ladder on large-radius target (Whisper-Small $\to$ Gemma-2-2B-IT, mean token radius \RadiusGemmaTwo, Muon)}} \\
Whisper-Small & Gemma-2-2B-IT & $\WhisperDim\!\to\!\GemmaTwoDim$ & None & Muon & \WsTwoWhisperGemmaTwoNoneMuonParamsM & \WsTwoWhisperGemmaTwoNoneMuonBestValCell & \WsTwoWhisperGemmaTwoNoneMuonWerCleanCell{} / \WsTwoWhisperGemmaTwoNoneMuonWerOtherCell \\
Whisper-Small & Gemma-2-2B-IT & $\WhisperDim\!\to\!\GemmaTwoDim$ & Unit $\ell_2$ norm & Muon & \WsTwoWhisperGemmaTwoLtwoMuonParamsM & \WsTwoWhisperGemmaTwoLtwoMuonBestValCell & \WsTwoWhisperGemmaTwoLtwoMuonWerCleanCell{} / \WsTwoWhisperGemmaTwoLtwoMuonWerOtherCell \\
Whisper-Small & Gemma-2-2B-IT & $\WhisperDim\!\to\!\GemmaTwoDim$ & Target-radius norm & Muon & \WsTwoWhisperGemmaTwoScaledMuonParamsM & \WsTwoWhisperGemmaTwoScaledMuonBestValCell & \textbf{\WsTwoWhisperGemmaTwoScaledMuonWerCleanCell} / \WsTwoWhisperGemmaTwoScaledMuonWerOtherCell \\
Whisper-Small & Gemma-2-2B-IT & $\WhisperDim\!\to\!\GemmaTwoDim$ & RMS norm & Muon & \WsTwoWhisperGemmaTwoRmsMuonParamsM & \WsTwoWhisperGemmaTwoRmsMuonBestValCell & \textbf{\WsTwoWhisperGemmaTwoRmsMuonWerCleanCell} / \textbf{\WsTwoWhisperGemmaTwoRmsMuonWerOtherCell} \\
Whisper-Small & Gemma-2-2B-IT & $\WhisperDim\!\to\!\GemmaTwoDim$ & Layer norm$^\ddagger$ & Muon & \WsTwoWhisperGemmaTwoLnMuonParamsM & \WsTwoWhisperGemmaTwoLnMuonBestValCell & \textbf{\WsTwoWhisperGemmaTwoLnMuonWerCleanCell} / \textbf{\WsTwoWhisperGemmaTwoLnMuonWerOtherCell} \\
\midrule
\multicolumn{8}{@{}l}{\textit{Normalization ladder on unit-radius target (Whisper-Small $\to$ Qwen2.5-1.5B-IT, mean token radius \RadiusQwenOneFiveB, Muon)}} \\
Whisper-Small & Qwen2.5-1.5B-IT & $\WhisperDim\!\to\!\QwenDim$ & None & Muon & \WsTwoWhisperQwenNoneMuonParamsM & \WsTwoWhisperQwenNoneMuonBestValCell & \WsTwoWhisperQwenNoneMuonWerCleanCell{} / \WsTwoWhisperQwenNoneMuonWerOtherCell \\
Whisper-Small & Qwen2.5-1.5B-IT & $\WhisperDim\!\to\!\QwenDim$ & Unit $\ell_2$ norm & Muon & \WsTwoWhisperQwenLtwoMuonParamsM & \WsTwoWhisperQwenLtwoMuonBestValCell & \WsTwoWhisperQwenLtwoMuonWerCleanCell{} / \WsTwoWhisperQwenLtwoMuonWerOtherCell \\
Whisper-Small & Qwen2.5-1.5B-IT & $\WhisperDim\!\to\!\QwenDim$ & Target-radius norm & Muon & \WsTwoWhisperQwenScaledMuonParamsM & \WsTwoWhisperQwenScaledMuonBestValCell & \textbf{\WsTwoWhisperQwenScaledMuonWerCleanCell} / \textbf{\WsTwoWhisperQwenScaledMuonWerOtherCell} \\
Whisper-Small & Qwen2.5-1.5B-IT & $\WhisperDim\!\to\!\QwenDim$ & RMS norm & Muon & \WsTwoWhisperQwenRmsMuonParamsM & \WsTwoWhisperQwenRmsMuonBestValCell & \textbf{\WsTwoWhisperQwenRmsMuonWerCleanCell} / \textbf{\WsTwoWhisperQwenRmsMuonWerOtherCell} \\
Whisper-Small & Qwen2.5-1.5B-IT & $\WhisperDim\!\to\!\QwenDim$ & Layer norm$^\ddagger$ & Muon & \WsTwoWhisperQwenLnMuonParamsM & \WsTwoWhisperQwenLnMuonBestValCell & \textbf{\WsTwoWhisperQwenLnMuonWerCleanCell} / \textbf{\WsTwoWhisperQwenLnMuonWerOtherCell} \\
\bottomrule
\end{tabular}%
}
\end{table*}

%% file: sections/06_composition.tex
\section{Composition: Moving an Adapter to a Second Language Model}
\label{sec:composition}

\input{tables/tab_centerpiece}

Sections~\ref{sec:tokens} and~\ref{sec:speech} establish two linear
bridges into a language model's input-embedding space: closed-form
vocabulary maps and trained speech adapters.
We now test whether these bridges compose:
given a frozen speech encoder $E$, a speech adapter trained with
frozen source LLM $A$, and a linear bridge $A \to B$ fitted between
$A$ and a frozen target LLM $B$, we inject the composed soft tokens
\begin{equation}
    \hat{\mathbf{z}}_B \;=\; \mathcal{N}_B\!\Bigl(
    \mathcal{N}_A(\hat{\mathbf{z}}_A)\,\mathbf{W}_{A\to B}
    + \mathbf{b}_{A\to B}\Bigr)
    \label{eq:composition}
\end{equation}
into $B$'s native chat template.
Here $\hat{\mathbf{z}}_A = \mathrm{Adapter}_{EA}(\mathbf{h}_E) \in \mathbb{R}^{d_A}$
is the output of the speech adapter from $E$ into $A$
(Eq.~\ref{eq:adapter}),
$\mathbf{W}_{A\to B} \in \mathbb{R}^{d_A \times d_B}$ and
$\mathbf{b}_{A\to B} \in \mathbb{R}^{d_B}$ are the bridge's matrix and bias,
and $\mathcal{N}_A$ and $\mathcal{N}_B$ are normalizations over $d_A$ and
$d_B$ coordinates (Section~\ref{sec:setup}).
A closed-form bridge has no input normalization and no bias
($\mathcal{N}_A$ is the identity and $\mathbf{b}_{A\to B} = \mathbf{0}$);
its matrix is $\mathbf{W}_{\text{LS}}$ or $\mathbf{Q}_{\text{Proc}}$
(Eqs.~\ref{eq:lstsq}--\ref{eq:maps}) and $\mathcal{N}_B$ is calibrated on
the target's training anchors.
A gradient-trained bridge has the form of the speech adapter
(Eq.~\ref{eq:adapter}): its matrix, bias, and both normalizations are
trained by gradient descent through frozen $B$, on the negative
log-likelihood that $B$ assigns to the reference text.
The composed chain then has a normalization between its two matrices
and does not reduce to one matrix.
A bridge fit on text receives the text's token embeddings in $A$,
placed through the bridge where the soft prompt goes in $B$'s chat
template, and $B$ must reproduce the same text: a copy task, which is
why its loss is close to zero (Table~\ref{tab:centerpiece}).
The weights of $B$ are never updated.

\paragraph{Maps that never saw audio fail.}
Closed-form vocabulary maps (orthogonal Procrustes and least squares,
each combined with five normalizations) and gradient-trained bridges
fit on text (LibriSpeech transcripts or WikiText-103
\citep{merity2017pointer}) all produce unusable speech recognition.
Across $\XprofileZeroAudioCells$ configurations (itemized in Appendix~\ref{app:composition-details}),
$\XprofileZeroAudioTotalUtts$ decoded utterances,
and all four ordered LLM pairs,
$\XprofileZeroAudioLtFiftyPct\%$ of utterances fall below
$\UttWerHalfThresholdPct\%$ WER
(Tables~\ref{tab:ws3_comp},~\ref{tab:ws4c}).
Least squares improves held-out anchor cosine over Procrustes on
every pair,
but even the best closed-form composition reaches only
$\ClosedFormCompBestClean\%$ \texttt{test-clean} WER.
The primary obstruction is the input-distribution shift from
continuous acoustic soft prompts to discrete token embeddings,
not map quality.

\begin{figure*}[!t]
\centering
\includegraphics[width=\textwidth]{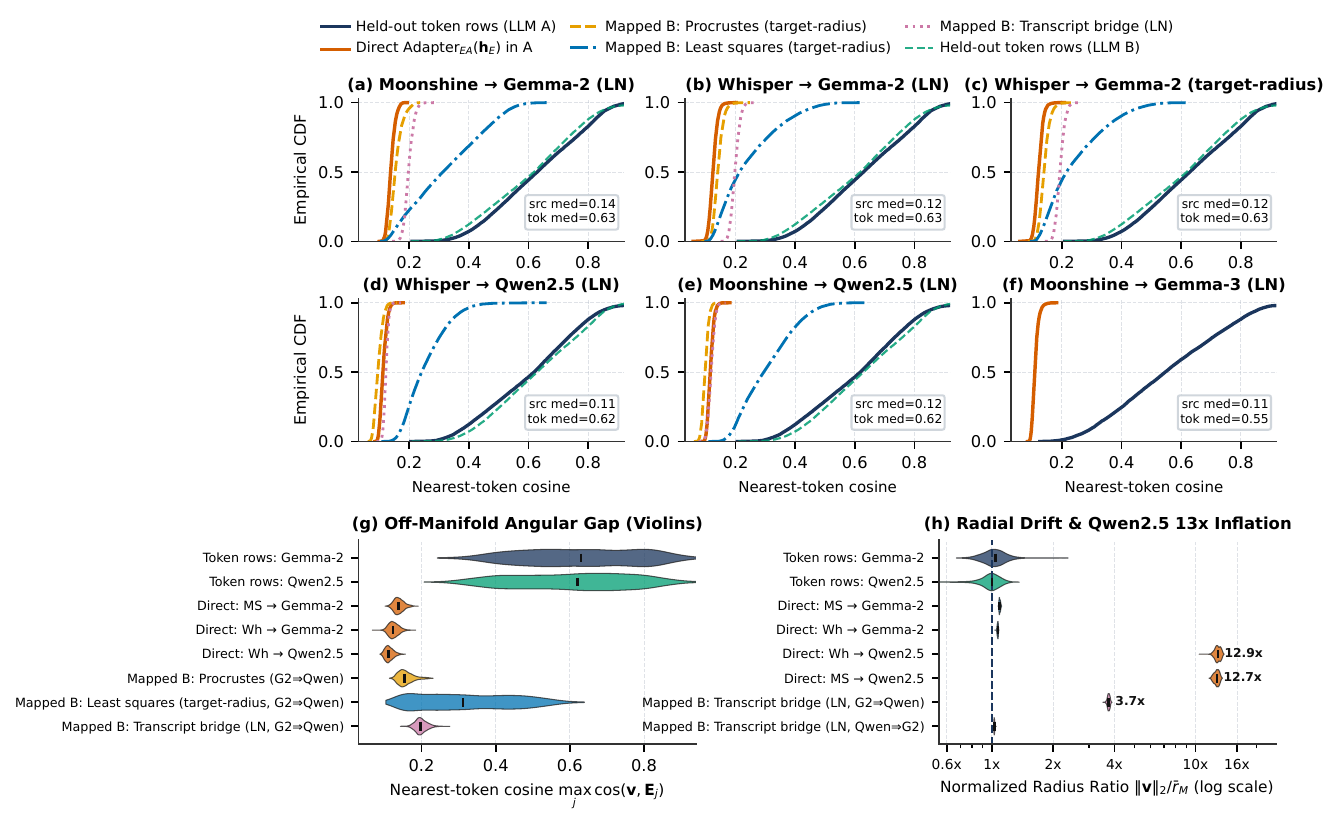}
\caption{\textbf{Angular and radial separation between discrete token embeddings and continuous speech-adapter soft prompts (\ManifoldNUtterances\ LibriSpeech utterances vs.\ \ManifoldNTokenEval\ held-out vocabulary rows).} \textbf{(a--f)}~Empirical cumulative distributions of nearest-token cosine similarity ($\uparrow$ means closer to a discrete token row; the nearest row is searched over the model's whole embedding table, without the query's own row, and every frame of the utterances is used): held-out vocabulary tokens cluster at median cosine \ManifoldTokMinNnCosMed--\ManifoldTokMaxNnCosMed, whereas speech adapter outputs in the source LLM sit far off the token set at median cosine \ManifoldSrcMinNnCosMed--\ManifoldSrcMaxNnCosMed; after bridging into target LLM $B$ (Qwen2.5-1.5B-IT in a--c, Gemma-2-2B-IT in d--e; no bridge in f) the value depends on the map (Section~\ref{sec:why}). \textbf{(g--h)}~Summary violins of nearest-token cosine similarity and output-to-token radius ratio; in (g), the mapped outputs are those of panel (a).}
\label{fig:manifold_distance}
\end{figure*}

\paragraph{A map trained on the adapter's own outputs works.}
Training $\mathbf{W}_{A \to B}$ on the speech adapter's own
frozen outputs---so the map sees the actual continuous soft-prompt
distribution that $A$ decoded---closes the gap.
On the best pair the composed system carries the adapter to target
LLM $B$ at the adapter's own source-LLM accuracy:
$\WsFourBCompBestClean\%$ / $\WsFourBCompBestOther\%$ WER, against
$\WsTwoWhisperGemmaTwoLnMuonWerClean\%$ /
$\WsTwoWhisperGemmaTwoLnMuonWerOther\%$ for the same adapter
evaluated directly on $A$.
All $\WsFourBInDistTotal$ in-distribution cells decode with zero empty
outputs ($\WsFourBInDistMinClean\%$--$\WsFourBInDistMaxClean\%$
\texttt{test-clean} WER).
The loss is largest into the narrower Qwen2.5-1.5B-IT: the Moonshine
adapter has $\WsTwoMoonshineGemmaTwoLnMuonWerClean\%$ in Gemma-2 and
$\WsFourBCompInDistMoonshineGemmaTwoToQwenLnClean\%$ after the bridge,
against $\WsTwoMoonshineQwenLnMuonWerClean\%$ for an adapter trained
directly into Qwen2.5.
A working bridge is not a free route into $B$: each training step needs
paired audio and a backward pass through $B$, as for a direct
adapter, and the bridge has more parameters.
Our bridges ran for one epoch ($\WsFourEFitSingleSteps$ steps) and the direct
adapters for $\WsTwoUniformEpochs$ epochs ($\WsTwoTotalSteps$ steps); at these unequal
budgets, the bridge ends above the direct adapter's WER
on $\WsThreeSameAboveDirectCleanRows$ of $\WsThreeCompTotalRows$ rows (Table~\ref{tab:ws3_comp}).
We use it to measure what a linear map between two models can carry.
Such a bridge is trained with the same data and loss as a direct adapter into $B$, on inputs computed from the same encoder frames, so its success does not by itself show that the input spaces of $A$ and $B$ are related.
Gaussian input-noise controls---transcript bridges trained with isotropic
Gaussian noise of expected norm $k\,\bar r_A$ added to their inputs,
$k \in \{\WsFourBNoiseKHalf, \WsFourBNoiseKOne, \WsFourBNoiseKSevenSevenThree\}$,
where the largest $k$ matches the soft prompts' median nearest-token
cosine---produce
$\WsFourBNoiseMinClean\%$--$\WsFourBNoiseMaxClean\%$ WER
(Table~\ref{tab:ws4b}),
confirming that structured soft-prompt inputs are required.

\paragraph{Half text, half soft prompts.}
Training $\mathbf{W}$ on a $\MixtureTextPct\%$ WikiText-103 /
$\MixtureAudioPct\%$ soft-prompt mixture keeps the text-copy loss low
(WikiText NLL $\WsFourCMixMinWikiVal$--$\WsFourCMixMaxWikiVal$)
at no consistent cost in in-distribution speech accuracy:
across $\MixVsAudioSameTriplesTotal$ matched triples, mixed minus
soft-prompt-only \texttt{test-clean} WER ranges from $\MixVsAudioMinDeltaClean$ to
$\MixVsAudioMaxDeltaClean$\,pp ($\MixVsAudioSameTriplesTied$ of them within
$\MixVsAudioTieCriterionPp$\,pp; single training seeds).
A mixed bridge sees the same audio once but takes twice as many steps (\WsFourEFitJointTotalSteps\ against \WsFourEFitSingleSteps).
One linear map, with its input and output normalizations, can read both
discrete text and continuous acoustic prefixes.

\begin{figure*}[!t]
\centering
\includegraphics[width=\textwidth]{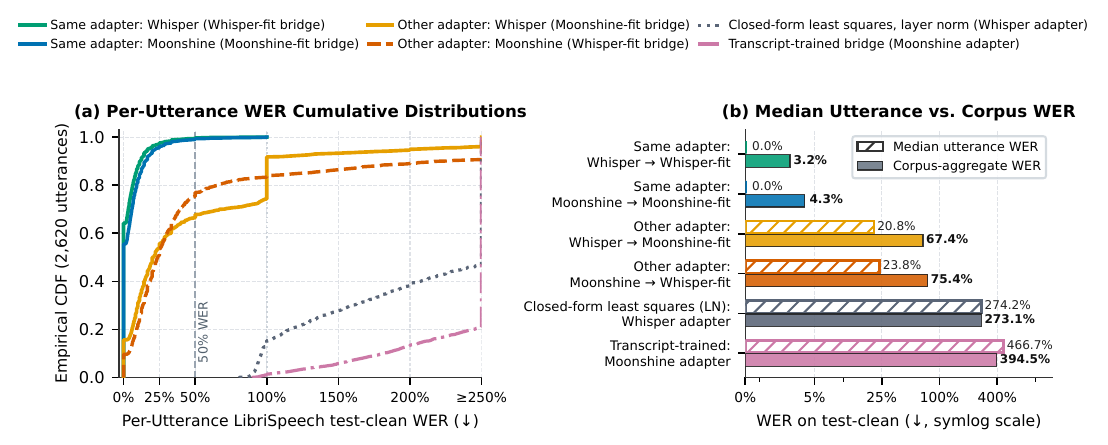}
\caption{\textbf{Per-utterance LibriSpeech \texttt{test-clean} word error rate ($\downarrow$) distributions across representative Gemma-2-2B-IT $\Rightarrow$ Gemma-3-4B-IT linear bridges (\LibriTestCleanUtts\ utterances).} \textbf{(a)}~Empirical cumulative distributions separate three regimes: (i)~bridges evaluated with the same speech adapter seen during fitting have a median utterance WER of \WsFourESingleSameMaxMedClean\% (\WsFourBCompInDistWhisperGemmaTwoToGemmaThreeLnClean\% Whisper and \WsFourBCompInDistMoonshineGemmaTwoToGemmaThreeLnClean\% Moonshine corpus WER); (ii)~evaluating those soft-prompt bridges on the other speech encoder's adapter decodes the median utterance partially (\XprofileXtwoCleanUttMedWer\% and \XprofileXoneCleanUttMedWer\% median WER) while run-on outputs, and for Whisper through the Moonshine-fit bridge \WsFourBCompCrossWhisperOnMoonshineGemmaTwoToGemmaThreeLnEmptyClean\ empty outputs, raise corpus WER to \WsFourBCompCrossWhisperOnMoonshineGemmaTwoToGemmaThreeLnClean\% and \WsFourBCompCrossMoonshineOnWhisperGemmaTwoToGemmaThreeLnClean\%; and (iii)~closed-form least squares (with layer norm) and transcript-trained bridges place \XprofileZeroAudioLtFiftyPct\% of utterances below \UttWerHalfThresholdPct\% WER (\WsThreeCompWhisperGemmaTwoToGemmaThreeLstsqLnClean\% and \WsFourCompMoonshineGemmaTwoToGemmaThreeLnClean\% corpus WER). \textbf{(b)}~Paired comparison of per-utterance median WER versus corpus-aggregate WER.}
\label{fig:utt_wer_dist}
\end{figure*}

\paragraph{A different adapter.}
Evaluating the same bridge on a different speech encoder's adapter
is the critical zero-shot porting test.
Across $\CrossAdapterTotalCells$ cross-adapter cells
($\CrossAdapterAudioOnlyCells$ from soft-prompt bridges and
$\CrossAdapterMixedCells$ from half-text bridges; Table~\ref{tab:ws3_comp}),
$\CrossAdapterFailCells$ fail completely
(per-utterance median \texttt{test-clean} WER
$\ge \CrossAdapterFailMinMedClean\%$;
aggregate $\CrossAdapterFailMinWerClean\%$--$\CrossAdapterFailMaxWerClean\%$).
Two cells, both from soft-prompt bridges on the Gemma-2 $\to$ Gemma-3
pair, decode partially
(median $\CrossAdapterPartialMinMedClean\%$ and
$\CrossAdapterPartialMaxMedClean\%$;
aggregate $\CrossAdapterPartialMinWerClean\%$ and
$\CrossAdapterPartialMaxWerClean\%$).
Run-on outputs raise the aggregate in both; in one,
$\WsFourBCompCrossWhisperOnMoonshineGemmaTwoToGemmaThreeLnEmptyClean$
empty outputs add to it (Figure~\ref{fig:utt_wer_dist}).
A second adapter of the \emph{same} encoder, trained with AdamW in place
of Muon from the same seed (same initialization and data order), sent
through the bridge fit on its Muon
counterpart, falls between the same-adapter and cross-encoder cases:
$\WsFourESingleCtrlMoonshineAdamwMedClean\%$ (Moonshine) and $\WsFourESingleCtrlWhisperAdamwMedClean\%$ (Whisper) median utterance WER,
with aggregate WER of $\WsFourESingleCtrlMoonshineAdamwClean\%$ and $\WsFourESingleCtrlWhisperAdamwClean\%$.
Of the $\WsFourESingleCtrlWhisperAdamwClean\%$, $\HeldoutWhisperSingleEmptyPpClean$\,pp come from $\WsFourESingleCtrlWhisperAdamwEmptyClean$ empty decodes, which fall on long utterances, and $\HeldoutWhisperSingleRunOnPpClean$\,pp from $\HeldoutWhisperSingleRunOnUttsClean$ run-on outputs (more errors than reference words; Table~\ref{tab:ws4e}).

\paragraph{One map for two adapters.}
On the Gemma-2 $\to$ Gemma-3 pair, cross-adapter failure is a matter
of coverage, not of linear capacity: one map trained on both adapters'
soft prompts serves both at single-adapter accuracy,
$\WsFourEJointWhisperMuonClean\%$ (Whisper) and $\WsFourEJointMoonshineMuonClean\%$ (Moonshine) \texttt{test-clean} WER
against $\WsFourBCompInDistWhisperGemmaTwoToGemmaThreeLnClean\%$ and $\WsFourBCompInDistMoonshineGemmaTwoToGemmaThreeLnClean\%$ for the two single-adapter bridges,
with zero empty decodes.
It runs twice as many steps as a single-adapter bridge (one pass over
each adapter's data); its midpoint checkpoint, at the single-adapter step
count, trails them by $\WsFourEJointHalfTrailWhisperClean$ and $\WsFourEJointHalfTrailMoonshineClean$\,pp.
The same map is a poor vocabulary ($\WsFourDTsevenWteJointPiqa\%$ PIQA) and reconstructs
text poorly (validation NLL $\WsFourEFitJointTextVal$; Table~\ref{tab:centerpiece}):
it serves the two inputs it was fit on, not tokens or text.
The map does not reach beyond the two adapters it was fit on.
Through it, the held-out AdamW adapters do not do reliably better than
through the single-adapter bridge of their own encoder.
Moonshine does worse: $\WsFourEJointCtrlMoonshineAdamwClean\%$ against
$\WsFourESingleCtrlMoonshineAdamwClean\%$ aggregate \texttt{test-clean} WER.
Whisper is mixed: $\WsFourEJointCtrlWhisperAdamwEmptyClean$ against
$\WsFourESingleCtrlWhisperAdamwEmptyClean$ empty decodes and
$\WsFourEJointCtrlWhisperAdamwClean\%$ against
$\WsFourESingleCtrlWhisperAdamwClean\%$ aggregate WER on \texttt{test-clean},
but $\WsFourEJointCtrlWhisperAdamwOther\%$ against
$\WsFourESingleCtrlWhisperAdamwOther\%$ on \texttt{test-other}
(Table~\ref{tab:ws4e}).
A bridge serves the adapters it was fit on; an unseen adapter of the
same encoder is carried only in part, and not reliably better when the
bridge has seen two adapters.

\paragraph{The reverse test: bridges as vocabularies.}
Bridges trained through the target model make poor static vocabularies, including the two fit on text.
Evaluated as Frankenstein vocabulary transplants on the same
Gemma-2 $\to$ Gemma-3 pair (Table~\ref{tab:ws4d}),
every gradient-trained map lands between chance
($\PiqaChancePct\%$ / $\HsChancePct\%$) and
$\WsFourDBestTrainedPiqa\%$ PIQA
($\WsFourDBestTrainedAboveChancePiqa$\,pp above chance) /
$\WsFourDBestTrainedHs\%$ HellaSwag
($\WsFourDBestTrainedAboveChanceHs$\,pp above chance),
against $\WsFourDConeLstsqScaledPiqa\%$
($\WsFourDConeLstsqScaledAboveChancePiqa$\,pp above chance) /
$\WsFourDConeLstsqScaledHs\%$
($\WsFourDConeLstsqScaledAboveChanceHs$\,pp above chance)
for the closed-form least-squares map
($\WsFourDDeltaConeVsBzeroPiqa$\,pp of native on PIQA).
The two text bridges show why.
They are trained so that $B$ can copy a text from a mapped prefix;
nothing in that objective asks a mapped row to stand in for $B$'s own
row of the same token, and it does not
(held-out anchor cosine $\WsFourDToneWteLibriCos$--$\WsFourDTtwoWteWikiCos$, against $\WsFourDConeLstsqScaledCos$ for the
least-squares map; the nearest held-out row is still the right one for $\WsFourDToneWteLibriPatOne\%$ and $\WsFourDTtwoWteWikiPatOne\%$ of anchors, against $\WsFourDConeLstsqScaledPatOne\%$; Table~\ref{tab:ws4d}).

%% file: tables/tab_centerpiece.tex
\begin{table*}[t]
\caption{\textbf{Each linear map from Gemma-2-2B-IT to frozen Gemma-3-4B-IT works on the inputs and the task it was fit on and fails outside them.} Rows vary what the $\GemmaTwoDim\!\to\!\GemmaThreeDim$ linear map was fit on (closed-form least squares uses target-radius normalization in every cell; gradient-trained bridges use layer norm); columns test the resulting map as a replacement input vocabulary on PIQA (\PiqaEvalItems\ items, accuracy $\uparrow$, \PiqaChancePct\% chance), on discrete text reconstruction (validation negative log-likelihood $\downarrow$: $^{\text{a}}$LibriSpeech \texttt{validation.clean} transcripts, including on the soft-prompt-only rows; $^{\text{b}}$WikiText-103 validation chunks), and on zero-shot speech recognition (LibriSpeech \texttt{test-clean} WER $\downarrow$, \LibriTestCleanUtts\ utterances) when chained behind the Gemma-2-2B-IT speech adapter seen during fitting versus the other speech encoder's Gemma-2-2B-IT adapter (Wh = Whisper-Small, MS = Moonshine-Base). $^\dagger$Aggregate WER is inflated by a minority of run-on or empty decodes; per-utterance median WER is shown in parentheses; --- indicates not applicable or not measured. The half-text and two-adapter bridges run \WsFourEFitJointTotalSteps\ steps (one pass over each of their two data streams); the other gradient-trained bridges run \WsFourEFitSingleSteps. The PIQA values of the two rows fit on the Whisper adapter (alone or with text) were measured on a second host, where unmodified Gemma-3-4B-IT scores \WsFourDWoneBzeroNativePiqa\% (Table~\ref{tab:ws4d}).}
\label{tab:centerpiece}
\centering
\resizebox{\textwidth}{!}{%
\setlength{\tabcolsep}{4.5pt}
\begin{tabular}{@{}lrrrr@{}}
\toprule
\textbf{What the Gemma-2 $\to$ Gemma-3 Map Was Fit On} & \textbf{As Vocabulary: PIQA $\uparrow$} & \textbf{Text Val NLL $\downarrow$} & \textbf{Speech WER, Same Adapter $\downarrow$} & \textbf{Speech WER, Other Adapter $\downarrow$} \\
\midrule
\multicolumn{5}{@{}l}{\textit{Reference: unmodified Gemma-3-4B-IT (PIQA) and direct speech adapters trained into Gemma-3-4B-IT (WER)}} \\
No bridge (native Gemma-3-4B-IT / direct adapter) & \textbf{\WsFourDBzeroNativePiqa\%} & --- & \textbf{\WsTwoWhisperGemmaThreeLnMuonWerClean\%} (Wh) / \textbf{\WsTwoMoonshineGemmaThreeLnMuonWerClean\%} (MS) & --- \\
\midrule
\multicolumn{5}{@{}l}{\textit{Linear bridges from Gemma-2-2B-IT ($\GemmaTwoDim$) to frozen Gemma-3-4B-IT ($\GemmaThreeDim$), grouped by what the map was fit on}} \\
Shared vocabulary tokens, closed-form least squares (target-radius norm) & \textbf{\WsFourDConeLstsqScaledPiqa\%} & --- & --- & \WsThreeCompWhisperGemmaTwoToGemmaThreeLstsqScaledClean\% (Wh) / \WsThreeCompMoonshineGemmaTwoToGemmaThreeLstsqScaledClean\% (MS) \\
Speech transcripts (LibriSpeech text), gradient & \WsFourDToneWteLibriPiqa\% & \WsFourFitGemmaTwoToGemmaThreeLnVal$^{\text{a}}$ & --- & \WsFourCompWhisperGemmaTwoToGemmaThreeLnClean\% (Wh) / \WsFourCompMoonshineGemmaTwoToGemmaThreeLnClean\% (MS) \\
General text (WikiText-103), gradient & \WsFourDTtwoWteWikiPiqa\% & \WsFourCWikiFitGemmaTwoToGemmaThreeLnWikiVal$^{\text{b}}$ & --- & \WsFourCWikiCompWhisperGemmaTwoToGemmaThreeLnClean\% (Wh) / \WsFourCWikiCompMoonshineGemmaTwoToGemmaThreeLnClean\% (MS) \\
\cmidrule(lr){1-5}
One adapter's soft prompts (Moonshine), gradient & \WsFourDTfourWteAudioMoonshinePiqa\% & \WsFourBFitMoonshineGemmaTwoToGemmaThreeLnTextVal$^{\text{a}}$ & \textbf{\WsFourBCompInDistMoonshineGemmaTwoToGemmaThreeLnClean\%} (MS) & \WsFourBCompCrossWhisperOnMoonshineGemmaTwoToGemmaThreeLnClean\% (Wh; med.\ \XprofileXtwoCleanUttMedWer\%)$^\dagger$ \\
One adapter's soft prompts (Whisper), gradient & \WsFourDTfiveWteAudioWhisperPiqa\% & \WsFourBFitWhisperGemmaTwoToGemmaThreeLnTextVal$^{\text{a}}$ & \textbf{\WsFourBCompInDistWhisperGemmaTwoToGemmaThreeLnClean\%} (Wh) & \WsFourBCompCrossMoonshineOnWhisperGemmaTwoToGemmaThreeLnClean\% (MS; med.\ \XprofileXoneCleanUttMedWer\%)$^\dagger$ \\
\cmidrule(lr){1-5}
Half text, half soft prompts (WikiText + Moonshine) & \WsFourDTthreeWteMixMoonshinePiqa\% & \WsFourCMixFitMoonshineGemmaTwoToGemmaThreeLnWikiVal$^{\text{b}}$ & \textbf{\WsFourCMixCompInDistMoonshineGemmaTwoToGemmaThreeLnClean\%} (MS) & \WsFourCMixCompCrossWhisperOnMoonshineGemmaTwoToGemmaThreeLnClean\% (Wh) \\
Half text, half soft prompts (WikiText + Whisper) & \WsFourDTsixWteMixWhisperPiqa\% & \WsFourCMixFitWhisperGemmaTwoToGemmaThreeLnWikiVal$^{\text{b}}$ & \textbf{\WsFourCMixCompInDistWhisperGemmaTwoToGemmaThreeLnClean\%} (Wh) & \WsFourCMixCompCrossMoonshineOnWhisperGemmaTwoToGemmaThreeLnClean\% (MS) \\
\cmidrule(lr){1-5}
Two adapters' soft prompts (Whisper + Moonshine), gradient & \WsFourDTsevenWteJointPiqa\% & \WsFourEFitJointTextVal$^{\text{a}}$ & \textbf{\WsFourEJointWhisperMuonClean\%} (Wh) / \textbf{\WsFourEJointMoonshineMuonClean\%} (MS) & --- \\
\bottomrule
\end{tabular}%
}
\end{table*}

%% file: sections/07_why.tex
\section{Why: Where Soft Prompts Sit}
\label{sec:why}

Why does a map fit on vocabulary tokens fail on soft prompts, and why does a bridge fit on one speech adapter fail on another?
We measure where soft prompts sit relative to token embeddings: by direction, by length, and in maps of the source and target input spaces.

\paragraph{Soft prompts lie far from the token set.}
Figure~\ref{fig:manifold_distance} compares the nearest-token
cosine similarity of held-out vocabulary rows against the
nearest-token cosine similarity of speech-adapter soft-prompt
outputs across $\ManifoldNUtterances$ LibriSpeech utterances
and $\ManifoldNTokenEval$ held-out tokens.
Held-out tokens have a median nearest-token cosine of
$\ManifoldTokMinNnCosMed$--$\ManifoldTokMaxNnCosMed$; adapter outputs
in the source LLM have
$\ManifoldSrcMinNnCosMed$--$\ManifoldSrcMaxNnCosMed$: no soft token is
close to any token row.
After bridging, the value depends on the map:
$\ManifoldMapProcMinNnCosMed$--$\ManifoldMapProcMaxNnCosMed$ for Procrustes,
$\ManifoldMapTranscriptMinNnCosMed$--$\ManifoldMapTranscriptMaxNnCosMed$ for transcript-trained bridges,
$\ManifoldMapLstsqScaledMinNnCosMed$--$\ManifoldMapLstsqScaledMaxNnCosMed$ for least squares with target-radius norm, and
$\ManifoldMapLstsqLnMinNnCosMed$--$\ManifoldMapLstsqLnMaxNnCosMed$ for least squares with layer norm.
The last range reaches that of real tokens, and those bridges still
fail (Table~\ref{tab:ws3_comp}): landing near token rows in the target
does not make a soft prompt decodable
(Section~\ref{sec:why-target} shows where these outputs land).
The token-fit maps are fit on token rows and applied to vectors that
are not near any of them.

\paragraph{Length depends on the target model.}
On the Gemma models, adapter outputs have $\ManifoldSrcGemmaMinRrMed$--$\ManifoldSrcGemmaMaxRrMed$ times the
mean token radius; on Qwen2.5-1.5B-IT they have $\ManifoldSrcQwenMinRrMed$--$\ManifoldSrcQwenMaxRrMed$
times (Figure~\ref{fig:manifold_distance}h).
Length does not separate soft prompts from tokens in general;
direction does.

\subsection{A Map of the Source Input Space}
\label{sec:why-source}

\begin{figure*}[p]
\centering
\includegraphics[width=0.95\textwidth]{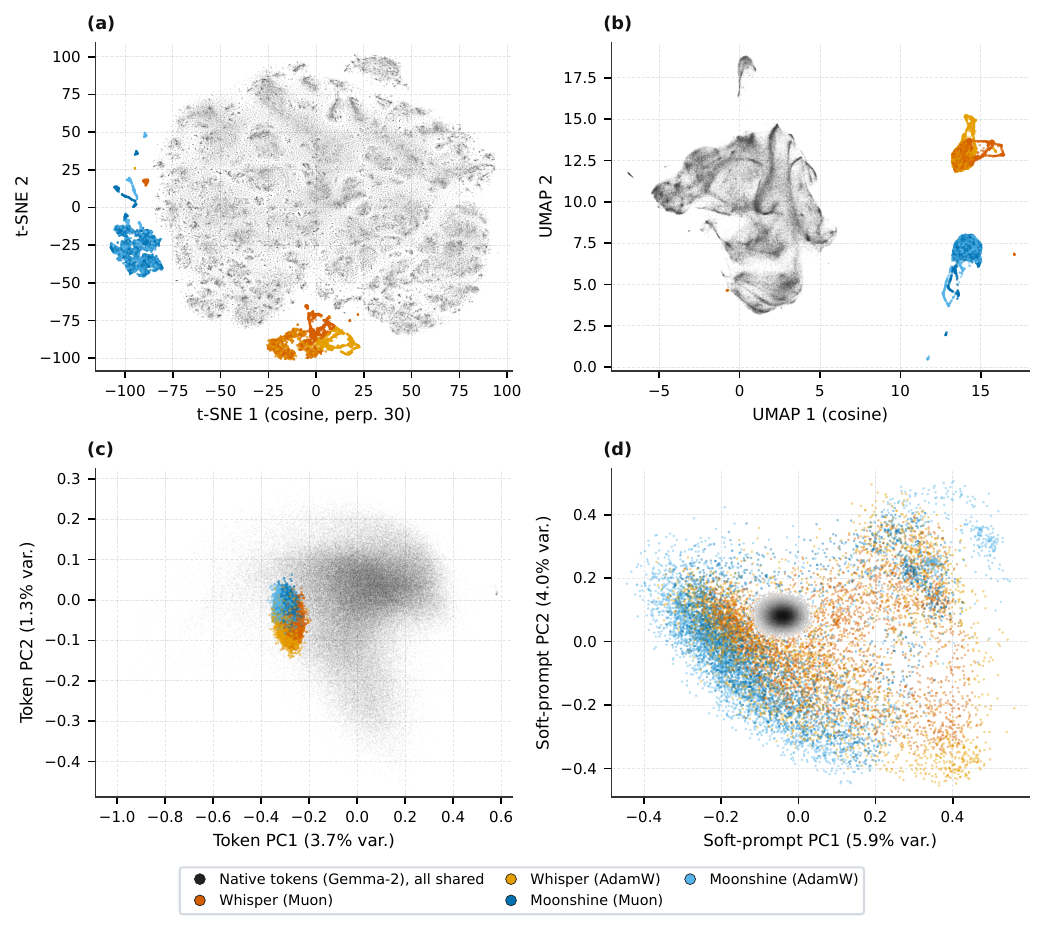}
\caption{\textbf{Token rows and speech soft tokens in Gemma-2-2B-IT's input space: four views of \GeoSrcTotalPoints\ vectors.}
\textbf{Points.} Unit-length input vectors. Black: all \GeoSrcSharedTokCount\ token rows shared with Gemma-3-4B-IT. Colors: soft tokens (a speech adapter's per-frame vectors, supplied in place of token rows): Whisper-Small (orange), Moonshine-Base (blue); Muon (darker), AdamW (lighter); \GeoSrcFramesPerAdapter\ per adapter (\GeoSrcFramesPerUtt\ frames from each of Figure~\ref{fig:manifold_distance}'s \ManifoldNUtterances\ utterances).
\textbf{(a)~t-SNE.} \emph{Method:} t-SNE with cosine distance (perplexity \GeoSrcPerp); it keeps similar vectors together. \emph{Axes:} unitless; only which points are neighbors has meaning, not distances between groups. \emph{Result:} three groups. Of the ten nearest map neighbors, \GeoSrcTsneTokSelfPct\% of a token row's are token rows; \GeoSrcTsneCrossEncMaxPct\% of a soft token's come from the other encoder and \GeoSrcTsneWhMixMinPct--\GeoSrcTsneWhMixMaxPct\% (Whisper) or \GeoSrcTsneMsMixPct\% (Moonshine) from the other optimizer's adapter. \emph{Interpretation:} modality and encoder set the neighborhood; the optimizer matters less (\GeoSrcOtherOptFullMinIntPct--\GeoSrcOtherOptFullMaxIntPct\% in all dimensions).
\textbf{(b)~UMAP.} \emph{Method:} a different objective on the same input (cosine distance, \GeoSrcUmapNbrs\ neighbors). \emph{Axes:} as in (a). \emph{Result:} the same three groups (\GeoSrcUmapTokSelfPct\%; $\le$\,\GeoSrcUmapCrossEncMaxPct\%; \GeoSrcUmapWhMixPct\% and \GeoSrcUmapMsMixPct\%). \emph{Interpretation:} the groups do not depend on the method (Appendix~\ref{app:geometry-details}).
\textbf{(c)~Token directions.} \emph{Method:} linear projection onto the token rows' first two principal components (PCs). \emph{Axes:} PC1, PC2 (\GeoSrcPanelCPcOnePct\%, \GeoSrcPanelCPcTwoPct\% of token variance), from the mean token row. \emph{Result:} token rows spread; soft tokens fall on one patch. The plane holds \GeoSrcPanelCTokOwnPct\% of token variance and \GeoSrcPanelCSpOwnMinPct--\GeoSrcPanelCSpOwnMaxPct\% of each adapter's (random plane: \GeoSrcRandomPlanePct\%). \emph{Interpretation:} directions that separate tokens barely separate audio frames.
\textbf{(d)~Soft-prompt directions.} \emph{Method:} as (c), with the soft tokens' PCs (token rows as a density). \emph{Axes:} PC1, PC2 (\GeoSrcPanelDPcOnePct\%, \GeoSrcPanelDPcTwoPct\% of soft-token variance), from the mean soft token. \emph{Result:} the reverse: the plane holds \GeoSrcPanelDSpOwnMinPct--\GeoSrcPanelDSpOwnMaxPct\% of each adapter's variance and \GeoSrcPanelDTokOwnPct\% of token variance; the encoders overlap. \emph{Interpretation:} both encoders vary most along shared directions, where tokens vary as in a random plane.
\textbf{Together.} Three groups (a, b) that vary along different directions (c, d). In all \GeoSrcDim\ dimensions, \GeoSrcTokWithSpCount\ of \GeoSrcSharedTokCount\ token rows have a soft token among their ten nearest neighbors and \GeoSrcSpWithTokCount\ of \GeoSrcSoftPromptTotal\ soft tokens have a token row (Section~\ref{sec:why-source}).}
\label{fig:geo_source_alltok}
\end{figure*}

Section~\ref{sec:composition} showed that a bridge works on the kind of input it was fit on and fails on the others.
Figure~\ref{fig:geo_source_alltok} shows those inputs themselves, in the input space of Gemma-2-2B-IT, where the bridges into Gemma-3-4B-IT take their input: all $\GeoSrcSharedTokCount$ token rows that the two models share, and $\GeoSrcFramesPerAdapter$ soft tokens from each of four speech adapters for the same $\ManifoldNUtterances$ utterances.
A soft token is the vector that an adapter supplies for one audio frame.
A two-dimensional map can mislead, so each statement below is a count or a cosine in all $\GeoSrcDim$ dimensions.
All vectors are scaled to unit length, so length plays no part.

\paragraph{Tokens and soft tokens form separate groups.}
For each of the $\GeoSrcTotalPoints$ vectors we find the ten vectors most similar to it by cosine.
Of the $\GeoSrcSharedTokCount$ token rows, $\GeoSrcTokWithSpCount$ have a soft token among these ten; of the $\GeoSrcSoftPromptTotal$ soft tokens, $\GeoSrcSpWithTokCount$ have a token row.
The largest cosine between a soft token and a token row is $\GeoSrcAlltokNearTokCosMax$.
The median cosine of a soft token with its nearest token row is $\GeoSrcAlltokNearTokCosMinMed$--$\GeoSrcAlltokNearTokCosMaxMed$ across the four adapters, against $\GeoSrcTokSelfNearTokCosMed$ for a token row with its nearest other token row.
The two kinds also point in different directions on average: the mean cosine between a token row and a soft token is $\GeoSrcMeanPairCosTokSpMin$--$\GeoSrcMeanPairCosTokSpMax$, against $\GeoSrcMeanPairCosTokTok$ between two token rows and $\GeoSrcMeanPairCosSpSpMin$--$\GeoSrcMeanPairCosSpSpMax$ between two soft tokens.

\paragraph{Each encoder has its own group; its two adapters lie side by side.}
Of the ten nearest neighbors of a Whisper soft token, at most $\GeoSrcWhToMsFullPct\%$ are Moonshine soft tokens; of those of a Moonshine soft token, $\GeoSrcMsToWhFullPct\%$ are Whisper soft tokens.
The two adapters of one encoder, trained with different optimizers from the same seed, lie side by side.
The same audio frame gives two vectors with a median cosine of $\GeoSrcPairedCosWh$ (Whisper) and $\GeoSrcPairedCosMs$ (Moonshine); frames of different utterances give $\GeoSrcXuttCosWh$ and $\GeoSrcXuttCosMs$.
The two adapters are still not the same: of a soft token's ten nearest neighbors, $\GeoSrcOtherOptFullMinPct\%$--$\GeoSrcOtherOptFullMaxPct\%$ come from the other adapter of its encoder, and $\GeoSrcXuttOtherOptMinPct\%$--$\GeoSrcXuttOtherOptMaxPct\%$ when frames of the query's own utterance are left out (Appendix~\ref{app:geometry-details}).

\paragraph{Tokens and soft tokens vary along different directions.}
Panels (c) and (d) of Figure~\ref{fig:geo_source_alltok} look along the two directions in which one kind of vector varies most.
The plane of the token directions holds $\GeoSrcPanelCTokOwnPct\%$ of the token variance and $\GeoSrcPanelCSpOwnMinPct\%$--$\GeoSrcPanelCSpOwnMaxPct\%$ of each adapter's; a plane chosen at random would hold $\GeoSrcRandomPlanePct\%$.
The plane of the soft-token directions holds $\GeoSrcPanelDSpOwnMinPct\%$--$\GeoSrcPanelDSpOwnMaxPct\%$ of each adapter's variance and $\GeoSrcPanelDTokOwnPct\%$ of the token variance.
The same holds over all directions.
We order the $\GeoSrcDim$ directions of the space by the energy of the training token rows along them (the eigenvectors of $\mathbf{X}^{\top}\mathbf{X}$ by decreasing eigenvalue, with $\mathbf{X}$ as in Section~\ref{sec:setup}; energy is measured from the origin, not from the mean as in panels (c) and (d), and on the vectors as they are), and measure what share of a population's summed squared length lies along them.
Held-out token rows have $\GeoLinSrcTokTopSixtyFourPct\%$ of it along the first $64$ directions and $\GeoLinSrcTokBotHalfPct\%$ along the last half; the four adapters have $\GeoLinSrcSpTopSixtyFourMinPct\%$--$\GeoLinSrcSpTopSixtyFourMaxPct\%$ and $\GeoLinSrcSpBotHalfMinPct\%$--$\GeoLinSrcSpBotHalfMaxPct\%$ (Figure~\ref{fig:geo_linear}).
Soft tokens are spread almost evenly over directions that token rows use very unevenly.
They do not fill the space: the soft tokens of one adapter span at most $d_E + 1$ dimensions ($\GeoSubDimMs$ for Moonshine-Base, $\GeoSubDimWh$ for Whisper-Small), a subspace that is not aligned with the token directions (Appendix~\ref{app:geometry-details}).
What tells one token from another tells almost nothing about audio frames, and the reverse.

\paragraph{The bridge results follow these groups.}
A bridge is a linear map fit on vectors of one group.
\begin{itemize}
\item Fit on token rows, or on text as a sequence of token rows, and applied to soft tokens, it fails: $\XprofileZeroAudioLtFiftyPct\%$ of $\XprofileZeroAudioTotalUtts$ decoded utterances fall below $\UttWerHalfThresholdPct\%$ WER. Fit on soft tokens and used as a vocabulary, it is poor (Table~\ref{tab:ws4d}).
\item Fit on one encoder's soft tokens, alone or mixed with text, and applied to the other's, it fails in $\CrossAdapterFailCells$ of $\CrossAdapterTotalCells$ cells ($\CrossAdapterFailMinWerClean\%$--$\CrossAdapterFailMaxWerClean\%$ WER). Fit on both, it works for both (Section~\ref{sec:composition}).
\item Fit on one adapter and applied to the other adapter of the same encoder, it works in part: WER rises from $\WsFourBCompInDistMoonshineGemmaTwoToGemmaThreeLnClean\%$ to $\WsFourESingleCtrlMoonshineAdamwClean\%$ for Moonshine, and from $\WsFourBCompInDistWhisperGemmaTwoToGemmaThreeLnClean\%$ to $\WsFourESingleCtrlWhisperAdamwClean\%$, with $\WsFourESingleCtrlWhisperAdamwEmptyClean$ empty outputs, for Whisper (Table~\ref{tab:ws4e}). Moonshine's pair of adapters is also the closer one in Figure~\ref{fig:geo_source_alltok} (same-frame cosine $\GeoSrcPairedCosMs$ against $\GeoSrcPairedCosWh$). With two pairs, this agreement is an observation and not a rule.
\end{itemize}

\subsection{Where Transplants and Bridges Land in the Target Space}
\label{sec:why-target}

\begin{figure*}[p]
\centering
\includegraphics[width=0.95\textwidth]{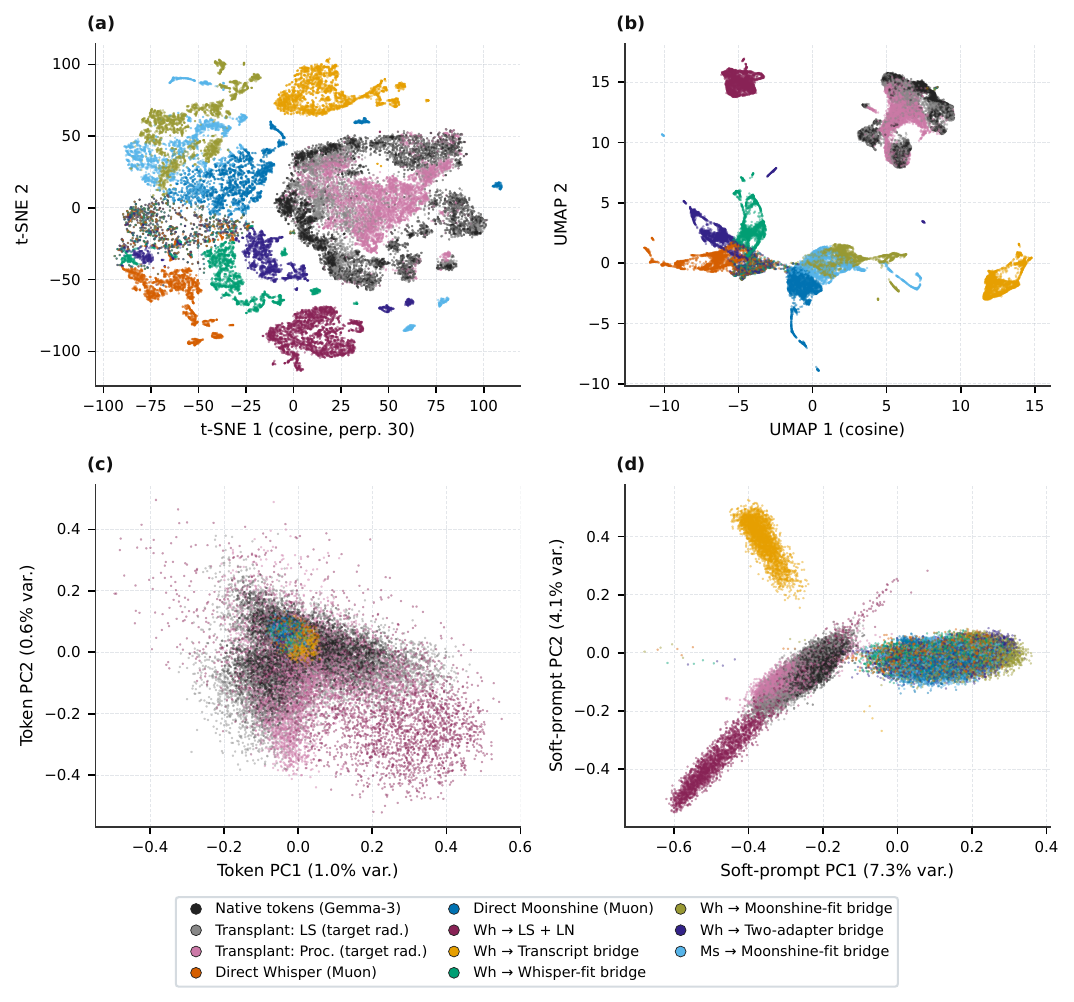}
\caption{\textbf{Transplanted token rows and bridged soft tokens in Gemma-3-4B-IT's input space: four views of \GeoTgtTotalPoints\ vectors.}
\textbf{Points.} Unit-length input vectors. Token rows of \GeoTgtTokPerPop\ held-out shared tokens: native (charcoal), transplanted from Gemma-2-2B-IT by least squares (gray) or Procrustes (rose). Soft tokens of Figure~\ref{fig:geo_source_alltok}'s frames: from adapters trained directly into Gemma-3-4B-IT, and from Gemma-2-2B-IT adapters through a bridge (Wh: Whisper; Ms: Moonshine; LS\,+\,LN: least squares with layer norm).
\textbf{(a)~t-SNE.} \emph{Method:} as in Figure~\ref{fig:geo_source_alltok}a. \emph{Axes:} unitless. \emph{Result:} five groups, each keeping nearly all ten nearest map neighbors: the three kinds of token rows ($\ge$\,\GeoTgtTsneTokGrpMinPct\%); direct Whisper with the Whisper-fit and two-adapter bridges (\GeoTgtTsneWhGrpMinPct\%); direct Moonshine with both outputs of the Moonshine-fit bridge (\GeoTgtTsneMsGrpMinPct\%); the LS\,+\,LN output (\GeoTgtTsneLsLnSelfPct\%); the transcript-bridge output (\GeoTgtTsneTransSelfPct\%). \emph{Interpretation:} bridges fit on soft prompts deliver vectors beside a direct adapter's; bridges fit on tokens or text do not.
\textbf{(b)~UMAP.} \emph{Method:} as in Figure~\ref{fig:geo_source_alltok}b. \emph{Axes:} unitless. \emph{Result:} the same five groups (\GeoTgtUmapTokGrpMinPct\%, \GeoTgtUmapWhGrpMinPct\%, \GeoTgtUmapMsGrpMinPct\%, \GeoTgtUmapLsLnSelfPct\%, \GeoTgtUmapTransSelfPct\%). \emph{Interpretation:} the groups do not depend on the method.
\textbf{(c)~Token directions.} \emph{Method:} linear projection onto the native rows' first two principal components (PCs). \emph{Axes:} PC1, PC2 (\GeoTgtPanelCPcOnePct\%, \GeoTgtPanelCPcTwoPct\% of native-row variance), from the mean native row. \emph{Result:} token rows spread (\GeoTgtPanelCTokOwnPct--\GeoTgtPanelCFlsOwnPct\% of each kind's variance in the plane); soft tokens of direct adapters and soft-prompt bridges fall on one patch (\GeoTgtPanelCSpOwnMinPct--\GeoTgtPanelCSpOwnMaxPct\%); the LS\,+\,LN output spreads widest (\GeoTgtPanelCLsLnOwnPct\%). \emph{Interpretation:} the LS\,+\,LN output varies along token directions; soft tokens that decode do not.
\textbf{(d)~Soft-prompt directions.} \emph{Method:} as (c), with the eight speech populations' pooled PCs. \emph{Axes:} PC1, PC2 (\GeoTgtPanelDPcOnePct\%, \GeoTgtPanelDPcTwoPct\% of their variance), from their mean. \emph{Result:} direct adapters and soft-prompt bridges overlap; the transcript-bridge output lies apart; the LS\,+\,LN output lines up with the token rows. \emph{Interpretation:} the two failing outputs miss the decodable soft tokens in different directions.
\textbf{Together.} Transplanted rows land on native rows. For the same audio frame, a bridge fit on an adapter delivers a vector close to a direct adapter's (median cosine \GeoTgtPairCosWhBridge\ for Whisper, \GeoTgtPairCosMsBridge\ for Moonshine); bridges fit on tokens or text do not (\GeoTgtPairCosLsLn\ and \GeoTgtPairCosTrans; Section~\ref{sec:why-target}).}
\label{fig:geo_target}
\end{figure*}

Figure~\ref{fig:geo_target} shows the other end of the bridges: the input space of Gemma-3-4B-IT, with $\GeoTgtTotalPoints$ vectors of $\GeoTgtNumPops$ kinds.
Three kinds are token rows of the same $\GeoTgtTokPerPop$ held-out shared tokens: native rows, and rows transplanted from Gemma-2-2B-IT by least squares and by Procrustes, both with target-radius norm.
Two kinds are the soft tokens of adapters trained directly into Gemma-3-4B-IT, one per encoder.
Six kinds are soft tokens of the Muon-trained Gemma-2-2B-IT adapters of Figure~\ref{fig:geo_source_alltok}, sent through a bridge: Whisper soft tokens through the least-squares map with layer norm, the transcript-trained bridge, the bridge fit on the Whisper adapter, the bridge fit on the Moonshine adapter, and the bridge fit on both adapters; and Moonshine soft tokens through the bridge fit on the Moonshine adapter.
All speech vectors come from the same $\GeoSrcFramesPerAdapter$ frames per encoder, so one audio frame can be followed along every route.
As in Section~\ref{sec:why-source}, the statements below are counts and cosines in all $\GeoTgtDim$ dimensions, on vectors of unit length.

\paragraph{Transplanted rows land on the native rows.}
In the maps, the three kinds of token rows form one group.
Among the held-out tokens, the native row nearest to a transplanted row is the row of the same token for $\WsFourDConeLstsqScaledPatOne\%$ of the least-squares rows and $\WsFourDCthreeProcScaledPatOne\%$ of the Procrustes rows (Table~\ref{tab:ws4d}).
The median cosine of a plotted transplanted row with its nearest plotted native row is $\GeoTgtFlsNearTokMed$ (least squares) and $\GeoTgtFprocNearTokMed$ (Procrustes), against $\GeoTgtTokSelfNearTokMed$ for a native row with its nearest other native row.
Landing there is not sufficient for a working vocabulary: both transplants land on the native rows, and the least-squares rows keep $\WsFourDConeLstsqScaledPiqa\%$ PIQA where the Procrustes rows keep $\WsFourDCthreeProcScaledPiqa\%$ (Section~\ref{sec:tokens}).

\paragraph{A bridge fit on an adapter delivers vectors close to those of a direct adapter.}
We compare, for the same audio frame, the vector that arrives through a Gemma-2-2B-IT adapter and a bridge with the vector of the adapter trained directly into Gemma-3-4B-IT.
Through the bridge fit on that adapter, the two have a median cosine of $\GeoTgtPairCosWhBridge$ for Whisper and $\GeoTgtPairCosMsBridge$ for Moonshine, against $\GeoTgtXuttCosWhBridge$ and $\GeoTgtXuttCosMsBridge$ for frames of different utterances.
Through the two-adapter bridge, Whisper gives $\GeoTgtPairCosWhJoint$.
For comparison, two adapters into the same model that differ in the optimizer gave $\GeoSrcPairedCosWh$ and $\GeoSrcPairedCosMs$ (Section~\ref{sec:why-source}).
These vectors stay away from the token rows.
Of the $\GeoTgtSpFitTotal$ soft tokens of direct adapters and soft-prompt bridges (the bridges fit on soft prompts), $\GeoTgtSpFitWithTokCount$ have a token row among their ten nearest neighbors; of the $\GeoTgtTokRowsTotal$ token rows, $\GeoTgtTokRowsWithSpFitCount$ have one of these soft tokens; and the largest cosine between one of these soft tokens and a native row is $\GeoTgtWorkSpNearTokMax$.

\paragraph{A bridge fit on the other encoder delivers vectors in between.}
Whisper soft tokens sent through the bridge fit on the Moonshine adapter have a median cosine of $\GeoTgtPairCosWhOnMs$ with the direct Whisper adapter's vectors for the same frames ($\GeoTgtXuttCosWhOnMs$ for different utterances).
Of their ten nearest neighbors, $\GeoTgtFullWhOnMsSelfPct\%$ are of their own kind, $\GeoTgtFullWhOnMsToMsOnMsPct\%$ are Moonshine soft tokens sent through the same bridge, and $\GeoTgtFullWhOnMsToWhGrpPct\%$ belong to the three Whisper populations of Figure~\ref{fig:geo_target}a.
Figure~\ref{fig:geo_target} counts them with the Moonshine group; they join it through the other output of the same bridge, not through the direct Moonshine adapter.
This is one of the two cross-adapter cells that decode in part: $\XprofileXtwoCleanUttMedWer\%$ median utterance WER and $\WsFourBCompCrossWhisperOnMoonshineGemmaTwoToGemmaThreeLnClean\%$ in aggregate (Section~\ref{sec:composition}).

\paragraph{Bridges fit on tokens or text deliver vectors elsewhere.}
Through the least-squares map with layer norm, the same Whisper frames have a median cosine of $\GeoTgtPairCosLsLn$ with the direct adapter's vectors.
They land at the token rows instead.
Their median cosine with the nearest plotted native row is $\GeoTgtLsLnNearTokMed$, about that of a native row with its nearest other native row ($\GeoTgtTokSelfNearTokMed$); $\GeoTgtNativeWithLsLnCount$ of the $\GeoTgtTokPerPop$ native rows have one of them among their ten nearest neighbors; and $\GeoLinTgtLsLnTopSixtyFourPct\%$ of their summed squared length lies along the first $64$ token directions of Gemma-3-4B-IT, against $\GeoLinTgtTokTopSixtyFourPct\%$ for held-out native rows (Figure~\ref{fig:geo_linear}).
This accounts for the token-like nearest-token cosine noted at the start of this section: the map carries these soft tokens to the token rows, and the soft tokens that Gemma-3-4B-IT decodes do not lie there.
Through the transcript-trained bridge, the cosine with the direct adapter's vectors is $\GeoTgtPairCosTrans$.
These vectors lie near neither group: $\GeoTgtFullTransSelfPct\%$ of their ten nearest neighbors are of their own kind, and their median cosine with the nearest plotted native row is $\GeoTgtTransNearTokMed$.
Both routes fail ($\WsThreeCompWhisperGemmaTwoToGemmaThreeLstsqLnClean\%$ and $\WsFourCompWhisperGemmaTwoToGemmaThreeLnClean\%$ WER; Table~\ref{tab:geo_pairs} lists all six routes).

\paragraph{Limits.}
The two figures cover one language-model pair, $\ManifoldNUtterances$ utterances of read speech, and one trained instance of each adapter and bridge.
They show where inputs lie.
They do not show by themselves why a map fit on one group fails on another; that the maps fail is the result of Section~\ref{sec:composition}.
Position does not settle function for token rows either, as the two transplants show.
The settings of the maps, a control without the principal-component step, repeated runs, and further counts are in Appendix~\ref{app:geometry-details}.

\subsection{Per-Utterance Error Profiles}
\label{sec:why-utt}

Figure~\ref{fig:utt_wer_dist} decomposes corpus-aggregate WER
into per-utterance distributions for representative bridges
on the Gemma-2 $\Rightarrow$ Gemma-3 pair.
Three regimes emerge:
(i)~same-adapter bridges have a median utterance WER of
$\WsFourESingleSameMaxMedClean\%$;
(ii)~cross-adapter partial cells show
$\XprofileXtwoCleanUttMedWer\%$ and
$\XprofileXoneCleanUttMedWer\%$ median utterance WER,
with aggregates raised by run-on outputs in both cells and by empty outputs in one, not by every utterance failing;
and (iii)~token-fit and text-fit bridges place
$\XprofileZeroAudioLtFiftyPct\%$ of utterances below
$\UttWerHalfThresholdPct\%$ WER.
The aggregate WER of a token-fit bridge can exceed twice the
reference length---more erroneous words inserted than exist in the
reference---while the median utterance produces no recognizable
speech at all.

%% file: sections/08_conclusion.tex
\section{Limitations and Conclusion}
\label{sec:conclusion}

\paragraph{Limitations.}
Each configuration uses one training seed, and seed-to-seed variance was
not measured; where two results differ
by less than $\MixVsAudioTieCriterionPp$\,pp, we report them as
tied rather than claiming a difference.
Runs were spread over several hosts and software stacks; a change of host moved $\NativeGemmaThreeHostPiqaDiffItems$ of \PiqaEvalItems\ PIQA items of the unmodified model ($\WsFourDWoneBzeroNativePiqa\%$ against $\WsFourDBzeroNativePiqa\%$), more than this threshold, so PIQA differences are taken against the native value of the same host, and the effect of host on training was not measured (Appendix~\ref{app:reproducibility}).
All speech experiments use one task (read English speech on
LibriSpeech \citep{panayotov2015librispeech}),
language models up to $4$B parameters, and two speech encoders
(Whisper-Small and Moonshine-Base), so ``another adapter'' means
one other.
All decoding is greedy; WER uses our minimal text normalization, which
costs the native decoders more than the adapters (Section~\ref{sec:speech}).
The two-adapter bridge, the same-encoder control, and the input-space maps of Section~\ref{sec:why} were each run on one language-model pair (Gemma-2 $\to$ Gemma-3).
Vocabulary transplants were tested into one target model, from donors
no wider than the target.
They are scored by multiple-choice log-probability; generation with a
transplanted table was not tested, and most evaluation tokens fall on rows
the map was fit on (Table~\ref{tab:token_hits}).
The text bridges were trained on a copy objective
(Section~\ref{sec:composition}); a bridge trained on next-token prediction
over mapped inputs was not tested.
Every adapter and every trained bridge includes input and output
normalizations; we trained a variant with none for the adapter only
(Table~\ref{tab:ws2_asr}).
Our adapters are not constrained to the vocabulary; adapters that build
their outputs from token embeddings
\citep{liao2025langbridge,yang2025bridging} were not tested and may
transfer differently.
Gemma-3-4B-IT, the target of our best adapter and of the two-adapter
bridge, is a vision-language model; whether its training on image inputs
helps it accept speech soft prompts was not tested.
Without frame stacking, the adapter passes every encoder frame to the
language model, $\SlamAsrFrameStack$ times as many soft tokens as the
stacked projector of \citet{ma2024embarrassingly}; we did not train
that projector on our models and do not compare accuracy with it.
A working bridge needs paired audio and backward passes through the
target model, like a direct adapter, and is specific to the adapters it
was fit on; our bridges ran for one epoch and the direct adapters for
$\WsTwoUniformEpochs$ epochs, so their WERs are not compared at equal training
budgets.

\paragraph{Conclusion.}
A single linear map between two frozen language models' input-embedding
spaces can carry vocabulary tokens, one speech adapter's soft prompts,
or two adapters' soft prompts at once; however, it generalizes only to the inputs it was fit on.
A closed-form least-squares map with one radius scalar keeps a frozen
model at up to $\WsOneBestPiqa\%$ PIQA ($\WsOneNativePiqa\%$ native).
A single trained matrix with input and output normalization, no frame
stacking, and no hidden nonlinearity reaches \WsTwoBestMuonRunWerPairCell\ WER from Whisper-Small into Gemma-3-4B-IT, below Whisper-Small's own decoder used zero-shot.
Chaining the two succeeds when the bridge is fit on the adapter's
own outputs ($\WsFourBCompBestClean\%$ /
$\WsFourBCompBestOther\%$ composed WER)
and fails on another encoder's adapter
($\CrossAdapterFailCells$ of $\CrossAdapterTotalCells$ cells)
unless the bridge is fit on both.
Bridges fit on tokens or text never produce usable speech
recognition; bridges trained through the target model make poor vocabularies.
In the input spaces themselves, token rows and the soft tokens of each
speech encoder form separate groups, and a bridge fit on an adapter
delivers vectors close to those of an adapter trained directly into the
target model (Section~\ref{sec:why}).
Each linear map works on the inputs and the task it was fit on and
not on the others: linear bridges connect frozen models within,
not outside, their training subspaces.

%% file: sections/90_appendix.tex
\section{Appendix}
\label{sec:appendix}

\subsection{Full Result Grids}
\label{app:full-grids}

\begin{figure}[ht]
\centering
\includegraphics[width=\textwidth]{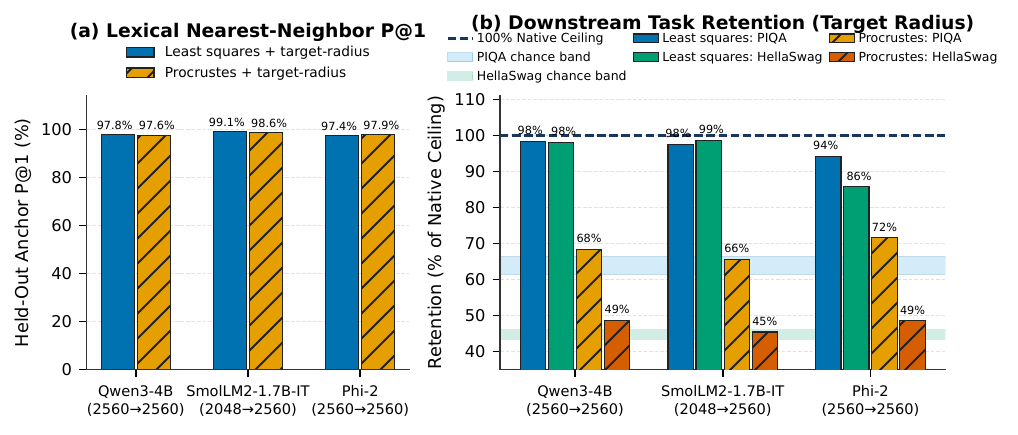}
\caption{\textbf{Lexical retrieval accuracy does not predict downstream reasoning retention in closed-form vocabulary transplants into frozen Gemma-3-4B-IT.} \textbf{(a)}~On held-out shared vocabulary tokens, orthogonal Procrustes matches unconstrained least squares in nearest-neighbor retrieval P@1 among the held-out anchors ($\uparrow$, \WsOneQwenProcScaledPatOne\%--\WsOneSmolProcScaledPatOne\% vs.\ \WsOnePhiLstsqScaledPatOne\%--\WsOneSmolLstsqScaledPatOne\%). \textbf{(b)}~At matched target radius, least squares retains \WsOnePhiLstsqScaledRetPiqa\%--\WsOneQwenLstsqScaledRetPiqa\% of native PIQA and \WsOnePhiLstsqScaledRetHs\%--\WsOneSmolLstsqScaledRetHs\% of native HellaSwag ($\uparrow$), whereas orthogonal Procrustes remains near chance.}
\label{fig:ws1_retention}
\end{figure}

Table~\ref{tab:app_models} lists the models and Table~\ref{tab:app_licenses} their licenses.
Table~\ref{tab:ws1_5norm} reports the complete five-normalization
vocabulary transplant hierarchy across all donor--map--normalization
combinations into frozen Gemma-3-4B-IT.
Figure~\ref{fig:ws1_retention} contrasts held-out lexical retrieval
accuracy against downstream reasoning retention.

\input{tables/tab_app_models}
\input{tables/tab_app_licenses}

\input{tables/tab_ws1_5norm}
\input{tables/tab_token_hits}

\subsection{Speech Adapter Training Curves and Geometry}
\label{app:ws2-details}

\begin{figure}[ht]
\centering
\includegraphics[width=\textwidth]{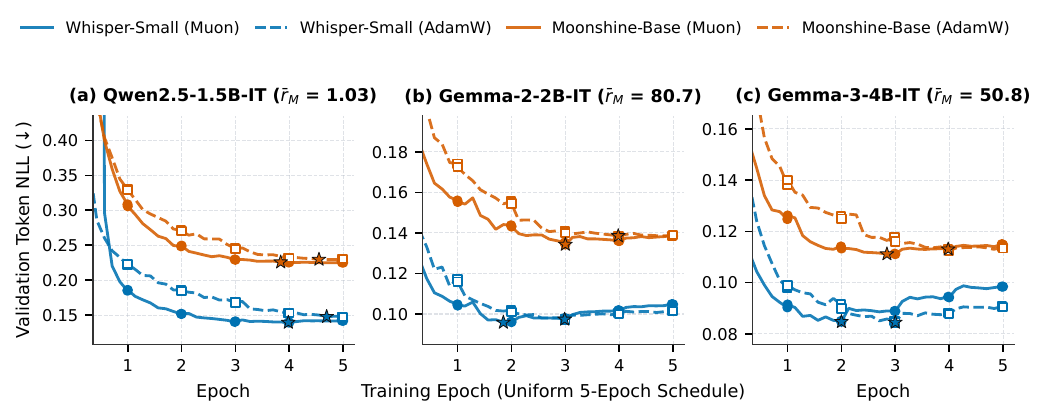}
\caption{\textbf{Validation negative log-likelihood ($\downarrow$) across the uniform \WsTwoUniformEpochs-epoch LibriSpeech training schedule for single-linear speech adapters.} Solid curves show orthogonalized momentum updates (Muon); dashed curves show AdamW; circles (Muon) and squares (AdamW) mark epoch ends; stars mark the best validation checkpoint selected for greedy WER evaluation on Qwen2.5-1.5B-IT, Gemma-2-2B-IT, and Gemma-3-4B-IT. Panel titles give each target's mean token radius.}
\label{fig:ws2_curves}
\end{figure}

\begin{figure}[ht]
\centering
\includegraphics[width=\textwidth]{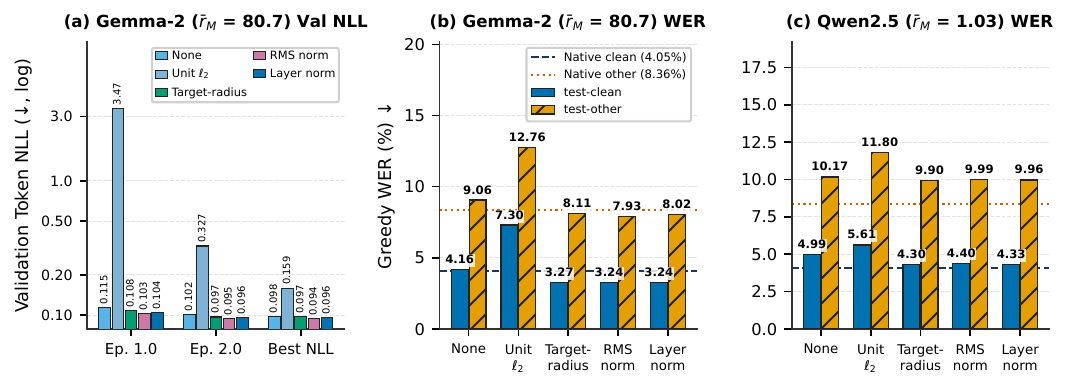}
\caption{\textbf{Normalization ladders (input and output normalization change together) under gradient training for Whisper-Small speech adapters (Muon, \WsTwoUniformEpochs\ epochs on LibriSpeech).} \textbf{(a)}~Validation negative log-likelihood ($\downarrow$) on large-radius Gemma-2-2B-IT (mean token radius \RadiusGemmaTwo). \textbf{(b--c)}~Greedy LibriSpeech \texttt{test-clean} and \texttt{test-other} WER ($\downarrow$) on Gemma-2-2B-IT and unit-radius Qwen2.5-1.5B-IT (mean token radius \RadiusQwenOneFiveB) relative to Whisper-Small's pretrained decoder (\NativeWhisperWerClean\% / \NativeWhisperWerOther\%).}
\label{fig:ws2_norm_ladder}
\end{figure}

\begin{figure}[ht]
\centering
\includegraphics[width=\textwidth]{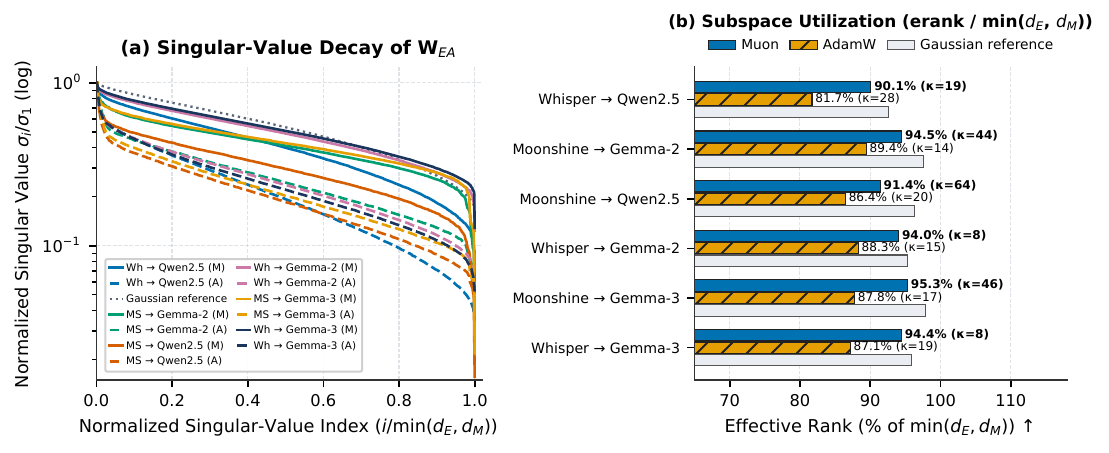}
\caption{\textbf{Spectral geometry of learned single-linear speech adapter matrices across all \WsTwoPairsDone\ encoder--LLM pairs.} \textbf{(a)}~Normalized singular-value spectra ($\sigma_i / \sigma_1$, log scale) of the matrix $\mathbf{W}$ alone, without the normalization gains, for Muon (solid) versus AdamW (dashed), and of one i.i.d.\ Gaussian random matrix of the Whisper-Small $\to$ Qwen2.5-1.5B-IT shape as a reference (dotted). \textbf{(b)}~Shannon effective rank (the exponential of the Shannon entropy of the singular values normalized to sum to one) as a percentage of $\min(d_E, d_M)$ ($\uparrow$) and condition number $\kappa$ (largest over smallest singular value) ($\downarrow$), with a Gaussian random matrix of each pair's shape as a reference. The adapters themselves start from the uniform initialization of Table~\ref{tab:protocol_optim}.}
\label{fig:adapter_geometry}
\end{figure}

Figure~\ref{fig:ws2_curves} shows validation negative log-likelihood
trajectories across the uniform training schedule.
Figure~\ref{fig:ws2_norm_ladder} shows the normalization ladder (input and
output normalization change together) on the large-radius Gemma-2-2B-IT
target and, for WER, on Qwen2.5-1.5B-IT.
Figure~\ref{fig:adapter_geometry} shows the spectral geometry of
learned adapter matrices.

\paragraph{WER under Whisper's text normalizer.}
With Whisper's \texttt{EnglishTextNormalizer} \citep{radford2023whisper} applied to both strings in place of our normalization (Appendix~\ref{app:protocol}), the native decoders score $\WnormNativeWhisperClean\%$ / $\WnormNativeWhisperOther\%$ (Whisper-Small; \citet{radford2023whisper} report $3.4\%$ / $7.6\%$ for this model) and $\WnormNativeMoonshineClean\%$ / $\WnormNativeMoonshineOther\%$ (Moonshine-Base), and the best adapter of each encoder $\WnormWhisperGemmaThreeMuonClean\%$ / $\WnormWhisperGemmaThreeMuonOther\%$ and $\WnormMoonshineGemmaThreeMuonClean\%$ / $\WnormMoonshineGemmaThreeMuonOther\%$.
The WER of each of the $\WnormAdapterCount$ adapters of the optimizer block of Table~\ref{tab:ws2_asr} falls by $\WnormAdapterMinDropPp$--$\WnormAdapterMaxDropPp$\,pp; that of the native decoders falls more, because they write numerals and abbreviations that the LibriSpeech references spell out.

\subsection{Composition Fit Diagnostics and Full WER Grid}
\label{app:composition-details}

\begin{figure}[ht]
\centering
\includegraphics[width=\textwidth]{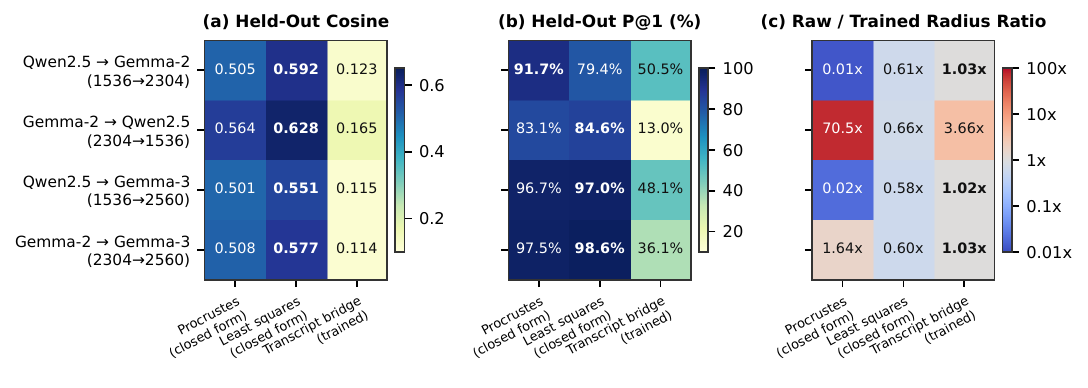}
\caption{\textbf{Lexical alignment diagnostics across the \WsThreeTotalPairs\ ordered LLM pairs for orthogonal Procrustes, unconstrained least squares, and transcript-trained bridges.} \textbf{(a)}~Mean held-out token cosine similarity ($\uparrow$). \textbf{(b)}~Held-out nearest-neighbor retrieval P@1 ($\uparrow$). \textbf{(c)}~Median output-to-target token radius ratio (closer to $1\times$ matches native scale): shown before target-radius rescaling for Procrustes and least squares, and after the learned output normalization for the transcript-trained bridge.}
\label{fig:ws3_fit_heatmaps}
\end{figure}

\begin{figure}[ht]
\centering
\includegraphics[width=\textwidth]{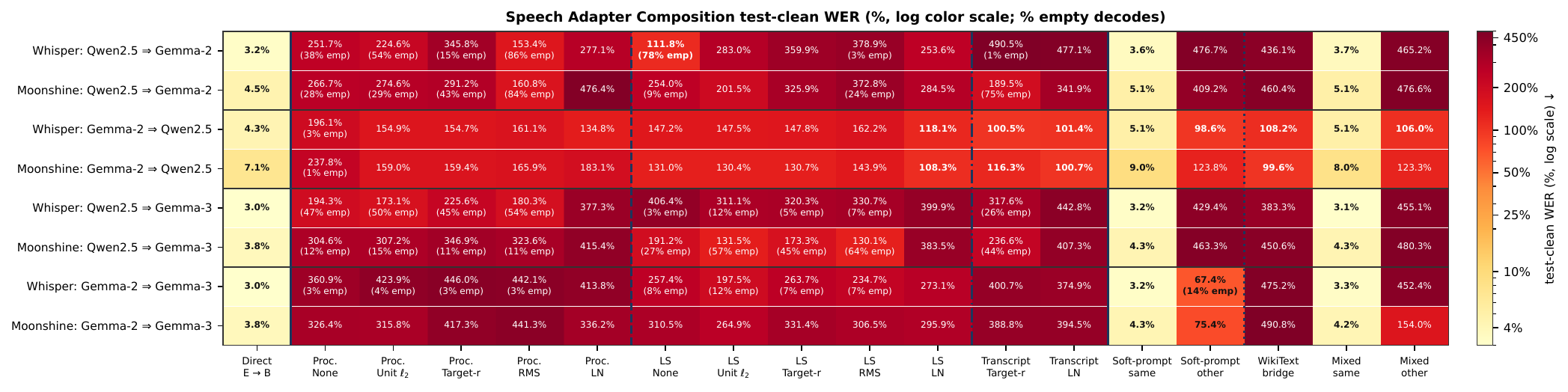}
\caption{\textbf{LibriSpeech \texttt{test-clean} word error rate ($\downarrow$, log color scale) across direct speech adapters and all composed bridge families.} Each row pairs a frozen speech encoder (Whisper-Small or Moonshine-Base) with an ordered source-to-target LLM port ($A \Rightarrow B$); cells with $\ge 0.5\%$ empty decodes display the empty-hypothesis percentage in parentheses. In the two columns labeled ``other'', the row's encoder is evaluated through the bridge fit on the other encoder's adapter. Not shown: the $4$ least-squares cells with the AdamW-trained Moonshine adapter that Appendix~\ref{app:composition-details} counts (Gemma-2 $\Rightarrow$ Qwen2.5 and Gemma-2 $\Rightarrow$ Gemma-3, without normalization and with target-radius norm; $\ZeroAudioAdamwLstsqMinClean\%$--$\ZeroAudioAdamwLstsqMaxClean\%$ \texttt{test-clean} WER).}
\label{fig:ws3_comp_heatmap}
\end{figure}

Table~\ref{tab:ws3_fits} reports lexical alignment diagnostics for
the closed-form, text-trained, soft-prompt-trained, and mixed bridges (Figure~\ref{fig:ws3_fit_heatmaps});
those of the two-adapter bridge are in Table~\ref{tab:ws4d} and those of the noise-control bridges in Table~\ref{tab:ws4b}.
Table~\ref{tab:ws3_comp} reports the speech adapter composition
WER grid; Figure~\ref{fig:ws3_comp_heatmap} adds unit $\ell_2$ and RMS norm for closed-form maps.
The $\XprofileZeroAudioCells$ configurations that never saw audio are cells, each one map evaluated with one speech adapter: $\XprofileZeroAudioClosedFormCells$ closed-form cells (those of Figure~\ref{fig:ws3_comp_heatmap}, and $4$ least-squares cells with the AdamW-trained Moonshine adapter, which no table shows), $\XprofileZeroAudioRandOrthCells$ random-orthogonal controls (Moonshine adapter, target-radius norm), $\XprofileZeroAudioTranscriptCells$ transcript-bridge cells ($8$ bridges, two encoders each), $\XprofileZeroAudioNoiseCells$ cells of transcript bridges trained with input noise, and $\XprofileZeroAudioWikiCells$ WikiText-103 cells ($4$ bridges, two encoders each).
\par\penalty0\relax 

\input{tables/tab_ws3_fits}

\input{tables/tab_ws3_comp}

\subsection{Soft-Prompt-Trained Bridges}
\label{app:ws4b}

Table~\ref{tab:ws4b} reports all soft-prompt-trained bridge results,
including in-distribution and cross-adapter cells
and Gaussian input-noise controls.

\input{tables/tab_ws4b}

\subsection{Two-Adapter Bridge and Same-Encoder Controls}
\label{app:ws4e}

Table~\ref{tab:ws4e} reports the two-adapter bridge and held-out
same-encoder adapter controls from Gemma-2-2B-IT to frozen Gemma-3-4B-IT.
Single-adapter bridges are trained on one frozen Muon speech adapter's
LibriSpeech soft prompts for $\WsFourEFitSingleSteps$ Muon steps
($\WsTwoBatchSize$ utterances per step; best validation checkpoint at
step $\WsFourEFitSingleBestStep$).
The two-adapter bridge is trained on frozen Muon Whisper-Small and
Moonshine-Base soft prompts together for $\WsFourEFitJointTotalSteps$
Muon steps ($\WsFourEFitJointSamplesPerAdapterStep$ utterances from
each adapter per step; best validation checkpoint at step
$\WsFourEFitJointBestStep$, midpoint checkpoint saved at step
$\WsFourEFitJointHalfStep$).
The two AdamW speech adapters on Gemma-2-2B-IT (one per encoder,
trained for $\WsTwoTotalSteps$ steps under identical data order and
initialization) are held out from all bridge fitting and evaluated
zero-shot through both single-adapter and two-adapter bridges.

\input{tables/tab_ws4e}

\subsection{Mixed and Text-Only Bridges}
\label{app:ws4c}

Table~\ref{tab:ws4c} reports WikiText-103-only and half-text
half-soft-prompt bridges.

\input{tables/tab_ws4c}

\subsection{Bridges as Vocabularies}
\label{app:ws4d}

Table~\ref{tab:ws4d} reports closed-form and gradient-trained maps
evaluated as static vocabulary replacements on the Gemma-2 $\to$
Gemma-3 pair against chance ($\PiqaChancePct\%$ PIQA,
$\HsChancePct\%$ HellaSwag).
The seven gradient-trained bridges are fit on:
(i)~LibriSpeech transcripts (text; $\WsFourDToneWteLibriPiqa\%$ PIQA,
$\WsFourDToneWteLibriAboveChancePiqa$\,pp above chance;
$\WsFourDToneWteLibriHs\%$ HellaSwag,
$\WsFourDToneWteLibriAboveChanceHs$\,pp above chance);
(ii)~WikiText-103 text ($\WsFourDTtwoWteWikiPiqa\%$ PIQA,
$\WsFourDTtwoWteWikiAboveChancePiqa$\,pp;
$\WsFourDTtwoWteWikiHs\%$ HellaSwag,
$\WsFourDTtwoWteWikiAboveChanceHs$\,pp);
(iii)~half WikiText-103 text and half Moonshine soft prompts
($\WsFourDTthreeWteMixMoonshinePiqa\%$ PIQA,
$\WsFourDTthreeWteMixMoonshineAboveChancePiqa$\,pp;
$\WsFourDTthreeWteMixMoonshineHs\%$ HellaSwag,
$\WsFourDTthreeWteMixMoonshineAboveChanceHs$\,pp);
(iv)~half WikiText-103 text and half Whisper soft prompts
($\WsFourDTsixWteMixWhisperPiqa\%$ PIQA,
$\WsFourDTsixWteMixWhisperAboveChancePiqa$\,pp;
$\WsFourDTsixWteMixWhisperHs\%$ HellaSwag,
$\WsFourDTsixWteMixWhisperAboveChanceHs$\,pp);
(v)~Moonshine soft prompts only
($\WsFourDTfourWteAudioMoonshinePiqa\%$ PIQA,
$\WsFourDTfourWteAudioMoonshineAboveChancePiqa$\,pp;
$\WsFourDTfourWteAudioMoonshineHs\%$ HellaSwag,
$\WsFourDTfourWteAudioMoonshineAboveChanceHs$\,pp);
(vi)~Whisper soft prompts only
($\WsFourDTfiveWteAudioWhisperPiqa\%$ PIQA,
$\WsFourDTfiveWteAudioWhisperAboveChancePiqa$\,pp;
$\WsFourDTfiveWteAudioWhisperHs\%$ HellaSwag,
$\WsFourDTfiveWteAudioWhisperAboveChanceHs$\,pp); and
(vii)~Whisper and Moonshine soft prompts
($\WsFourDTsevenWteJointPiqa\%$ PIQA,
$\WsFourDTsevenWteJointAboveChancePiqa$\,pp;
$\WsFourDTsevenWteJointHs\%$ HellaSwag,
$\WsFourDTsevenWteJointAboveChanceHs$\,pp).

\input{tables/tab_ws4d}

\clearpage
\subsection{Details of the Input-Space Maps}
\label{app:geometry-details}

This subsection gives the settings of the maps in Figures~\ref{fig:geo_source_alltok} and~\ref{fig:geo_target}, the checks behind them, and the measurements that Section~\ref{sec:why} cites.

\paragraph{Vectors and settings.}
Every vector is scaled to unit length before the maps, the neighbor counts, and the cosines; the energy shares and the leverage below use the vectors as they are.
The $\ManifoldNUtterances$ utterances are a seeded random sample of LibriSpeech \texttt{test-clean}, the same in every figure.
Figure~\ref{fig:manifold_distance} uses every frame of these utterances; the maps and the measurements of this subsection use $\GeoSrcFramesPerUtt$ frames per utterance, drawn at random without replacement after Whisper's padding frames are removed.
The held-out token rows are seeded random subsets of the held-out split.
The Gemma-2-2B-IT map uses all $\GeoSrcSharedTokCount$ shared token rows ($\GeoSrcZeroNormRows$ have zero norm and $\GeoSrcDupRows$ duplicate another row); they make up $\GeoSrcTokSharePct\%$ of its $\GeoSrcTotalPoints$ vectors.
The Gemma-3-4B-IT map uses $\GeoTgtTokPerPop$ of the held-out shared tokens, the same tokens for the native and the two transplanted kinds, and the audio frames of the Gemma-2-2B-IT map.
t-SNE uses cosine distance, perplexity $\GeoSrcPerp$, a random start, and seed $42$; UMAP uses cosine distance, $\GeoSrcUmapNbrs$ neighbors, and a minimum distance of $0.1$.
Both run on the first $\GeoSrcPcaDims$ principal components of the pooled vectors.
Before scaling, the median length of each population is $\GeoSrcRadMin$--$\GeoSrcRadMax$ times the mean token radius in the Gemma-2-2B-IT map and $\GeoTgtRadMin$--$\GeoTgtRadMax$ times in the Gemma-3-4B-IT map.
A neighbor share \emph{in the map} is counted over the ten nearest points in the plane; a share \emph{in all dimensions} is counted over the ten highest cosines among all vectors of the figure.

\begin{figure}[htbp]
\centering
\includegraphics[width=0.90\textwidth]{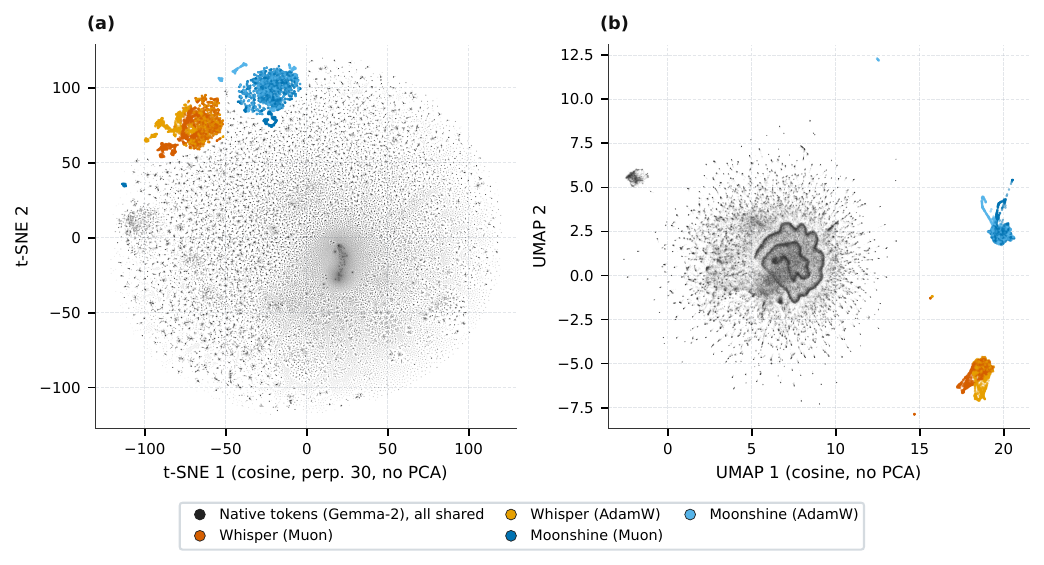}
\caption{\textbf{The maps of Figure~\ref{fig:geo_source_alltok}a--b without the principal-component step.} \textbf{(a)}~t-SNE and \textbf{(b)}~UMAP with the settings of Figure~\ref{fig:geo_source_alltok}, computed from the exact cosine neighbors of the \GeoSrcTotalPoints\ unit vectors in all \GeoSrcDim\ dimensions. Both show the same three groups; the neighbor shares in the map differ from those of Figure~\ref{fig:geo_source_alltok} by at most \GeoSrcNoPcaTsneMaxDiffPp\,pp (t-SNE) and \GeoSrcNoPcaUmapMaxDiffPp\,pp (UMAP). The token group has a different shape, which carries no meaning.}
\label{fig:geo_source_nopca}
\end{figure}

\paragraph{The principal-component step does not create the groups.}
The first $\GeoSrcPcaDims$ principal components keep $\GeoSrcAlltokPcaEvrTwoDecPct\%$ of the pooled variance of the Gemma-2-2B-IT vectors ($\GeoSrcAlltokPcaTokOwnPct\%$ of the token rows' variance about their own mean and $\GeoSrcAlltokPcaSpOwnMinPct\%$--$\GeoSrcAlltokPcaSpOwnMaxPct\%$ of each adapter's) and $\GeoTgtPcaEvrPct\%$ of that of the Gemma-3-4B-IT vectors.
Figure~\ref{fig:geo_source_nopca} repeats both maps on the full $\GeoSrcDim$-dimensional vectors without this step.
With and without it, the neighbor shares in the map differ by at most $\GeoSrcNoPcaTsneMaxDiffPp$\,pp (t-SNE) and $\GeoSrcNoPcaUmapMaxDiffPp$\,pp (UMAP).
In all four maps, at most $\GeoSrcAllFourMapsTokSpPct\%$ of a soft token's neighbors are token rows and at most $\GeoSrcUmapCrossEncMaxPct\%$ come from the other encoder.
The maps differ from the count in all dimensions in one respect: they mix the two adapters of an encoder more than the full space does.
In those entries the largest difference is $\GeoSrcRbTsneMaxDiffStepOnePp$--$\GeoSrcRaUmapMaxDiffStepOnePp$\,pp across the four maps; in all other entries it is at most $\GeoSrcMapsVsFullOtherMaxDiffPp$\,pp.

\begin{figure}[!t]
\centering
\includegraphics[width=0.90\textwidth]{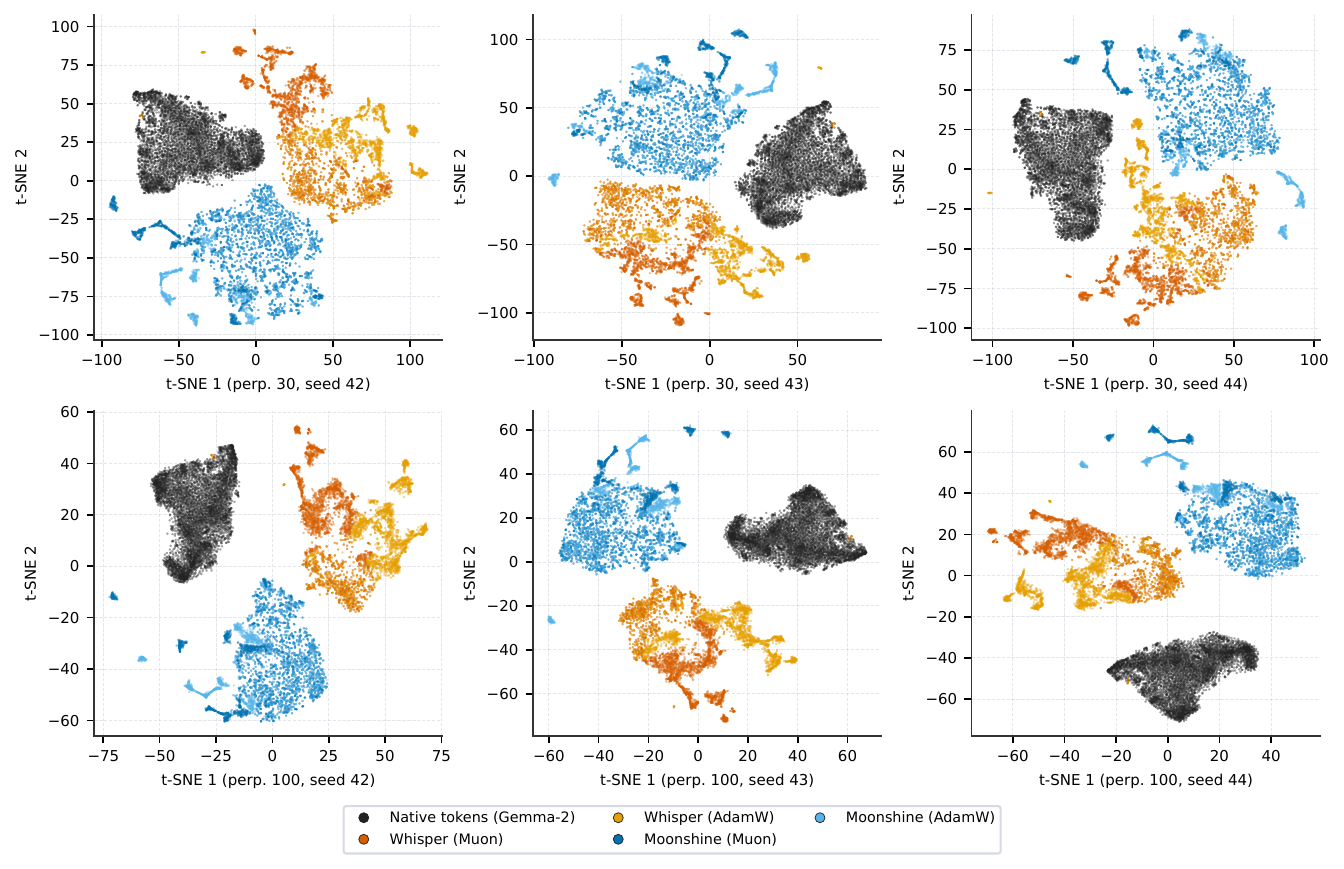}
\caption{\textbf{Six t-SNE runs on \GeoSrcFiveKTotalPoints\ Gemma-2-2B-IT vectors (\GeoSrcFiveKTokCount\ held-out token rows and \GeoSrcSoftPromptTotal\ soft tokens).} Top row: perplexity \GeoSrcPerp; bottom row: perplexity $100$; columns: seeds $42$, $43$, and $44$. Every run gives the same three groups (the neighbor shares in the map differ by at most \GeoSrcSixRunMaxDiffPp\,pp between runs); their arrangement changes from run to run.}
\label{fig:geo_tsne_6runs}
\end{figure}

\paragraph{Repeated t-SNE runs.}
Figure~\ref{fig:geo_tsne_6runs} shows six t-SNE runs (perplexities $\GeoSrcPerp$ and $100$; seeds $42$, $43$, and $44$) on $\GeoSrcFiveKTotalPoints$ Gemma-2-2B-IT vectors: $\GeoSrcFiveKTokCount$ held-out token rows and the $\GeoSrcSoftPromptTotal$ soft tokens (the first $\GeoSrcPcaDims$ principal components keep $\GeoSrcFiveKPcaEvrPct\%$ of their variance).
Between runs, the neighbor shares in the map differ by at most $\GeoSrcSixRunMaxDiffPp$\,pp; for six such runs on the Gemma-3-4B-IT vectors of Figure~\ref{fig:geo_target}, by at most $\GeoTgtSixRunMaxDiffPp$\,pp.
The arrangement of the groups is not stable: the rank correlation, between two runs, of the distances between group centers has a minimum of $\GeoSrcSixRunCentroidRhoMin$ and a mean of $\GeoSrcSixRunCentroidRhoMean$ over the pairs of runs.
Which points are neighbors is reproducible; where a group lies in the plane is not.

\begin{figure}[!t]
\centering
\includegraphics[width=0.85\textwidth]{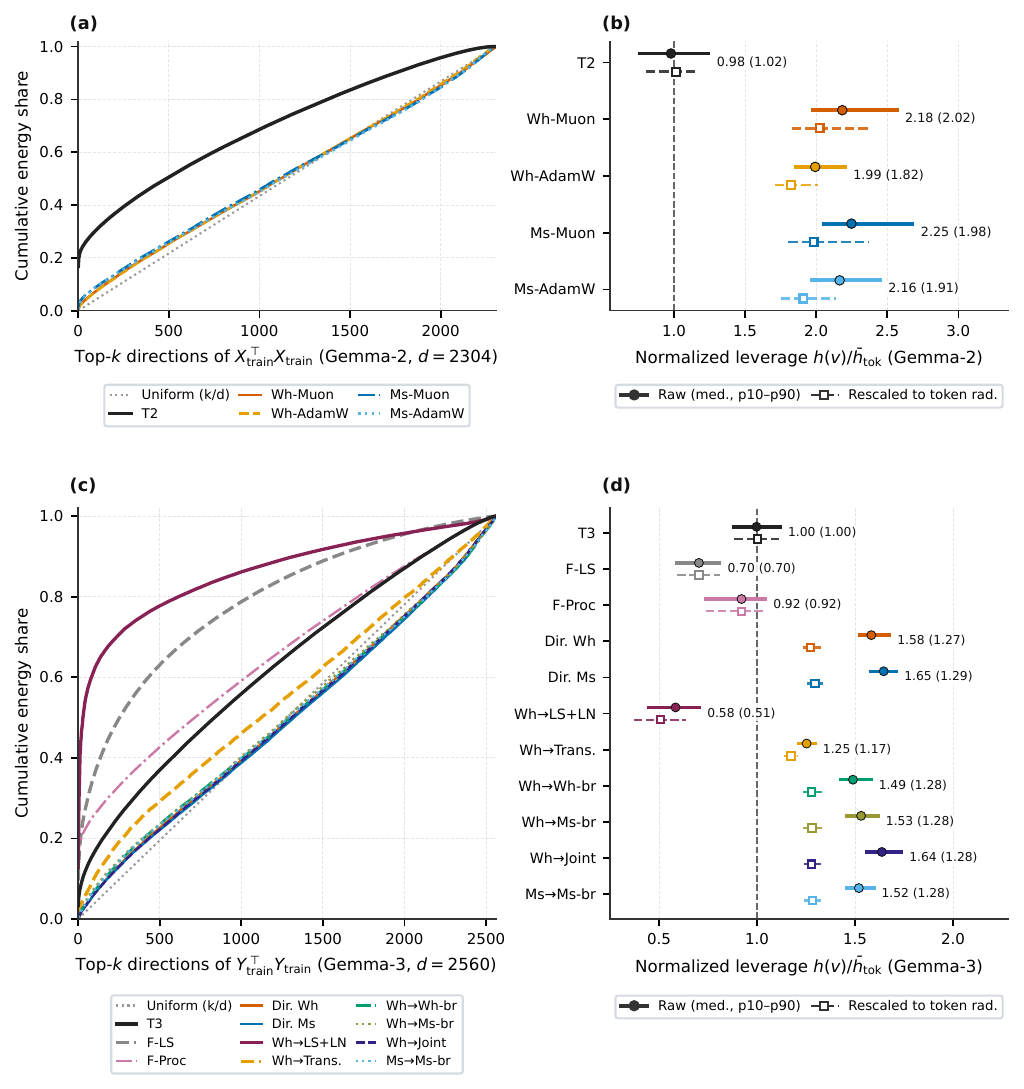}
\caption{\textbf{Energy along token directions (left) and leverage (right) in Gemma-2-2B-IT (top) and Gemma-3-4B-IT (bottom).} \textbf{(a, c)}~Share of a population's summed squared length along the first $k$ eigenvectors of $\mathbf{X}^{\top}\mathbf{X}$ ($d=\GeoSrcDim$) and of $\mathbf{Y}^{\top}\mathbf{Y}$ ($d=\GeoTgtDim$), with $\mathbf{X}$ and $\mathbf{Y}$ the training-anchor rows; the dotted diagonal $k/d$ is a population whose energy is spread evenly over these directions. \textbf{(b, d)}~Leverage relative to the mean leverage of held-out token rows: median and 10th--90th percentiles of the vectors as they are (filled circles; first number) and rescaled to the mean token radius (open squares; number in parentheses). \textbf{Labels.} T2 and T3: held-out token rows of Gemma-2-2B-IT and Gemma-3-4B-IT. F-LS and F-Proc: rows transplanted by least squares and by Procrustes. Wh and Ms: Whisper and Moonshine. Dir.: adapter trained directly into Gemma-3-4B-IT. Wh$\to$LS+LN, Wh$\to$Trans., Wh$\to$Wh-br, Wh$\to$Ms-br, and Wh$\to$Joint: Whisper soft tokens through the least-squares map with layer norm, the transcript-trained bridge, the Whisper-fit bridge, the Moonshine-fit bridge, and the two-adapter bridge. Ms$\to$Ms-br: Moonshine soft tokens through the Moonshine-fit bridge.}
\label{fig:geo_linear}
\end{figure}

\paragraph{Energy along token directions, and leverage.}
Let $\mathbf{X}$ hold the Gemma-2-2B-IT rows of the training anchors and $\mathbf{Y}$ their Gemma-3-4B-IT rows (Section~\ref{sec:setup}).
The eigenvectors of $\mathbf{X}^{\top}\mathbf{X}$, by decreasing eigenvalue, are the directions of the Gemma-2-2B-IT input space ordered by the energy (summed squared projection) of the training rows along them; those of $\mathbf{Y}^{\top}\mathbf{Y}$ do the same for Gemma-3-4B-IT.
Figure~\ref{fig:geo_linear}a,c shows, for each population, the share of its summed squared length that lies along the first $k$ directions.
A population whose energy is spread evenly over these directions follows the diagonal $k/d$. A population confined to a subspace stays close to the diagonal too when the subspace is not aligned with the token directions.
In Gemma-2-2B-IT the four adapters stay close to it: $\GeoLinSrcSpBotHalfMinPct\%$--$\GeoLinSrcSpBotHalfMaxPct\%$ of their energy lies in the last half of the directions, against $\GeoLinSrcTokBotHalfPct\%$ for held-out token rows.
In Gemma-3-4B-IT so do the soft tokens of direct adapters and soft-prompt bridges ($\GeoLinTgtWorkSpBotHalfMinPct\%$--$\GeoLinTgtWorkSpBotHalfMaxPct\%$, against $\GeoLinTgtTokBotHalfPct\%$ for held-out native rows).
The output of the least-squares map with layer norm does the opposite: $\GeoLinTgtLsLnTopSixtyFourPct\%$ of its energy lies in the first $64$ directions, $\GeoLinTgtLsLnTopTwoFiveSixPct\%$ in the first $256$, and $\GeoLinTgtLsLnBotHalfPct\%$ in the last half.
The leverage $h(\mathbf{v}) = \mathbf{v}\,(\mathbf{X}^{\top}\mathbf{X})^{-1}\mathbf{v}^{\top}$ of a row vector $\mathbf{v}$ (with $\mathbf{Y}$ in place of $\mathbf{X}$ in Gemma-3-4B-IT) puts the same fact in one number: it is large when $\mathbf{v}$ lies along directions in which the training rows have little energy.
Relative to the mean leverage of held-out token rows, the median leverage is $\GeoLinSrcSpLevRawMin$--$\GeoLinSrcSpLevRawMax$ for the four Gemma-2-2B-IT adapters ($\GeoLinSrcSpLevRescMin$--$\GeoLinSrcSpLevRescMax$ when each vector is first rescaled to the mean token radius), $\GeoLinTgtWorkSpLevRawMin$--$\GeoLinTgtWorkSpLevRawMax$ ($\GeoLinTgtWorkSpLevRescMin$--$\GeoLinTgtWorkSpLevRescMax$) for the Gemma-3-4B-IT soft tokens of direct adapters and soft-prompt bridges, and $\GeoLinTgtLsLnLevRaw$ ($\GeoLinTgtLsLnLevResc$) for the output of the least-squares map with layer norm (Figure~\ref{fig:geo_linear}b,d).

\paragraph{The subspace of an adapter's soft tokens.}
The soft tokens of one adapter do not fill the input space.
By Eq.~\ref{eq:adapter} with layer norm on both sides, they lie in a subspace of at most $d_E + 1$ dimensions: the input normalization leaves $d_E - 1$ free directions and an offset, the matrix and bias map these into a subspace of at most $d_E$ dimensions, the output normalization acts on it linearly up to a scalar factor, and its offset adds one dimension.
The singular values of the sampled soft tokens agree: for each adapter, $\GeoSubDimWh$ (Whisper-Small) or $\GeoSubDimMs$ (Moonshine-Base) of them exceed $10^{-6}$ times the largest, in Gemma-2-2B-IT and in Gemma-3-4B-IT.
The diagonal of Figure~\ref{fig:geo_linear}a therefore shows that this subspace is not aligned with the token directions, not that soft tokens point everywhere.
In Gemma-2-2B-IT, the subspace of one adapter holds $\GeoSubSameEncMinPct\%$--$\GeoSubSameEncMaxPct\%$ of the summed squared length of the soft tokens of the other adapter of the same encoder.
A Moonshine adapter's subspace holds $\GeoSubWhInMsMinPct\%$--$\GeoSubWhInMsMaxPct\%$ of that of a Whisper adapter's soft tokens, and a Whisper adapter's subspace $\GeoSubMsInWhMinPct\%$--$\GeoSubMsInWhMaxPct\%$ of that of a Moonshine adapter's; a random subspace of the same dimension would hold $\GeoSubRandMsPct\%$ and $\GeoSubRandWhPct\%$.
A bridge fit on one adapter is thus fit on a subspace that holds only part of any other adapter's soft tokens.

\paragraph{Neighbor shares in all dimensions.}
Counted in all dimensions on the $\GeoTgtTotalPoints$ vectors of Figure~\ref{fig:geo_target}, at least $\GeoTgtFullTokGrpMinPct\%$ of the ten nearest neighbors of a token row of any of the three kinds are token rows; nearly all of the others, up to $\GeoTgtFullTokToLsLnMaxPct\%$, are outputs of the least-squares map with layer norm.
At least $\GeoTgtFullWhGrpMinPct\%$ of the neighbors of each of the three Whisper populations of Figure~\ref{fig:geo_target}a lie within the three, and at least $\GeoTgtFullMsGrpMinPct\%$ of those of direct Moonshine and of the two outputs of the Moonshine-fit bridge lie within those three.
The outputs of the least-squares map with layer norm and of the transcript-trained bridge keep $\GeoTgtFullLsLnSelfPct\%$ and $\GeoTgtFullTransSelfPct\%$ of their neighbors to themselves.

\begin{figure}[!t]
\centering
\includegraphics[width=0.90\textwidth]{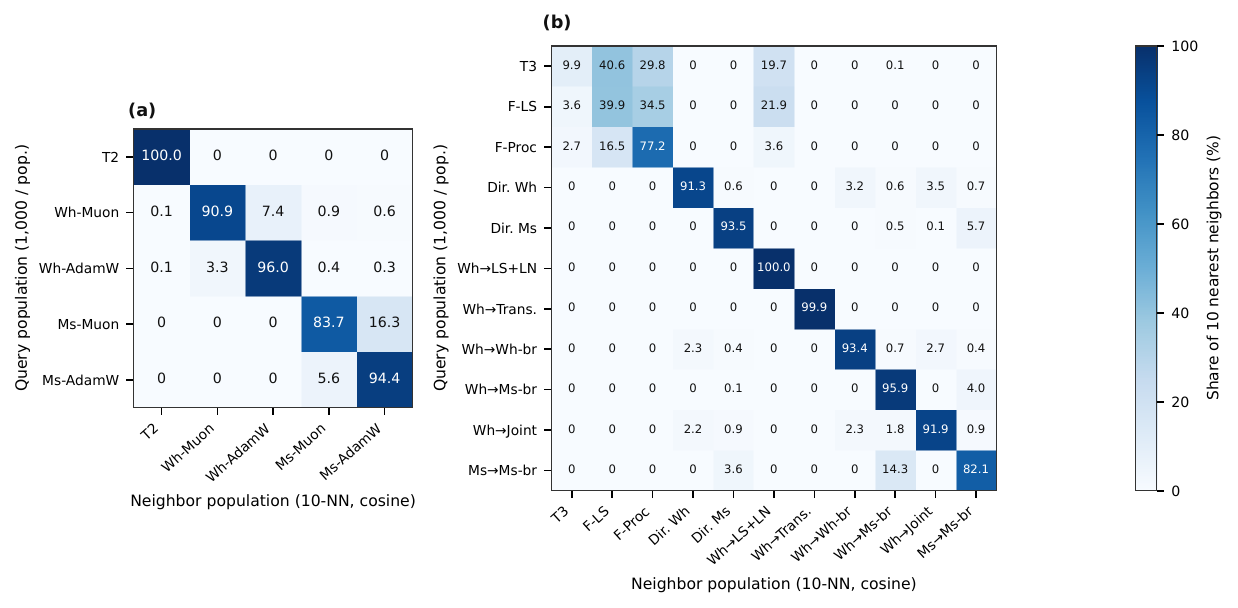}
\caption{\textbf{Shares of the ten nearest neighbors in all dimensions (\%), on \GeoBalancedPerPop\ vectors per population drawn separately for each population.} Rows: population of the query; columns: population of the neighbor. \textbf{(a)}~Gemma-2-2B-IT. \textbf{(b)}~Gemma-3-4B-IT. Labels as in Figure~\ref{fig:geo_linear}.}
\label{fig:geo_neighbours}
\end{figure}

In Figures~\ref{fig:geo_source_alltok} and~\ref{fig:geo_target}, some populations share their tokens or their audio frames.
Figure~\ref{fig:geo_neighbours} repeats the count on $\GeoBalancedPerPop$ vectors per population, drawn separately for each population: different tokens for the three kinds of token rows in Gemma-3-4B-IT, and separately drawn frames for every speech population.
The groups remain.
In Gemma-2-2B-IT, $\GeoSrcDisjTokSelfPct\%$ of a token row's neighbors are token rows.
In Gemma-3-4B-IT, the outputs of the least-squares map with layer norm and of the transcript-trained bridge keep $\GeoTgtDisjLsLnSelfPct\%$ and $\GeoTgtDisjTransSelfPct\%$ of their neighbors to themselves, and $\GeoTgtDisjMsToWhOnMsPct\%$ of the neighbors of Moonshine soft tokens sent through the Moonshine-fit bridge are Whisper soft tokens sent through the same bridge ($\GeoTgtDisjWhOnMsToMsPct\%$ in the other direction).
The three kinds of token rows mix with one another, and $\GeoTgtDisjTokToLsLnPct\%$ of a native row's neighbors are outputs of the least-squares map with layer norm: these outputs lie close to many token rows at once, while their own neighbors are of their own kind.

Three further counts use the $\GeoSrcFiveKTotalPoints$ Gemma-2-2B-IT vectors of Figure~\ref{fig:geo_tsne_6runs}.
(i)~Without neighbors from the query's own utterance, $\GeoSrcXuttOwnMinPct\%$--$\GeoSrcXuttOwnMaxPct\%$ of a soft token's neighbors come from its own adapter and $\GeoSrcXuttOtherOptMinPct\%$--$\GeoSrcXuttOtherOptMaxPct\%$ from the other adapter of its encoder: the two adapters of an encoder are neighbors mainly through frames of the same utterance.
(ii)~For $\GeoSrcTwinTopTenWhPct\%$ of Whisper frames and $\GeoSrcTwinTopTenMsPct\%$ of Moonshine frames, the vector that the other adapter gives for the same frame is among the ten nearest of all $\GeoSrcFiveKTotalPoints$ vectors.
(iii)~The mean directions of the four adapters have cosines of $\GeoSrcMeanDirCosAdaptersMin$--$\GeoSrcMeanDirCosAdaptersMax$ with one another and of $\GeoSrcMeanDirCosTokAdaptersMin$--$\GeoSrcMeanDirCosTokAdaptersMax$ with the mean direction of the token rows.

\paragraph{The same frame along different routes.}
Table~\ref{tab:geo_pairs} follows each audio frame along six routes into Gemma-3-4B-IT and compares the vector that arrives with the vector that the adapter trained directly into Gemma-3-4B-IT gives for the same frame.
As a control, it compares the arriving vector with that adapter's vectors for all frames of other utterances.
The three routes that decode have same-frame cosines of $\GeoTgtPairCosWhBridge$, $\GeoTgtPairCosWhJoint$, and $\GeoTgtPairCosMsBridge$; the route that decodes in part, $\GeoTgtPairCosWhOnMs$; the two routes that fail, $\GeoTgtPairCosLsLn$ and $\GeoTgtPairCosTrans$.
With six routes, this agreement is an observation and not a rule.

\begin{table}[!h]
\caption{\textbf{The same audio frame along different routes into Gemma-3-4B-IT.} Median cosine between the vector that a Gemma-2-2B-IT adapter and a bridge deliver and the vector of the adapter of the same encoder trained directly into Gemma-3-4B-IT (\GeoSrcFramesPerAdapter\ frames per encoder): for the same frame, and over all pairs of frames from different utterances. WER is the LibriSpeech \texttt{test-clean} word error rate of the route; the two direct adapters reach \WsTwoWhisperGemmaThreeLnMuonWerClean\% (Whisper) and \WsTwoMoonshineGemmaThreeLnMuonWerClean\% (Moonshine).}
\label{tab:geo_pairs}
\vspace{4pt}
\centering
\small
\begin{tabular}{@{}llrrr@{}}
\toprule
\textbf{Soft Tokens} & \textbf{Bridge} & \shortstack[r]{\strut\textbf{Same}\\\textbf{Frame}} & \shortstack[r]{\strut\textbf{Different}\\\textbf{Utterances}} & \textbf{WER $\downarrow$} \\
\midrule
Whisper & Fit on the Whisper adapter & \GeoTgtPairCosWhBridge & \GeoTgtXuttCosWhBridge & \WsFourBCompInDistWhisperGemmaTwoToGemmaThreeLnClean\% \\
Whisper & Fit on both adapters & \GeoTgtPairCosWhJoint & \GeoTgtXuttCosWhJoint & \WsFourEJointWhisperMuonClean\% \\
Moonshine & Fit on the Moonshine adapter & \GeoTgtPairCosMsBridge & \GeoTgtXuttCosMsBridge & \WsFourBCompInDistMoonshineGemmaTwoToGemmaThreeLnClean\% \\
Whisper & Fit on the Moonshine adapter & \GeoTgtPairCosWhOnMs & \GeoTgtXuttCosWhOnMs & \WsFourBCompCrossWhisperOnMoonshineGemmaTwoToGemmaThreeLnClean\% \\
Whisper & Least squares with layer norm (fit on tokens) & \GeoTgtPairCosLsLn & \GeoTgtXuttCosLsLn & \WsThreeCompWhisperGemmaTwoToGemmaThreeLstsqLnClean\% \\
Whisper & Transcript-trained (fit on text) & \GeoTgtPairCosTrans & \GeoTgtXuttCosTrans & \WsFourCompWhisperGemmaTwoToGemmaThreeLnClean\% \\
\bottomrule
\end{tabular}
\end{table}

\subsection{Reproducibility}
\label{app:reproducibility}

Runs used $\VersionsDistinctStacks$ distinct PyTorch/Transformers stacks across the
compute fleet (PyTorch \VersionsTorchList; Transformers \VersionsTransformersList).
Two checks bound their effect.
Rerunning one composed evaluation on a second stack moved
\texttt{test-clean} WER by $\StackcheckAsevenDeltaClean$\,pp and
\texttt{test-other} WER by $\StackcheckAsevenDeltaOther$\,pp
($\StackcheckAsevenDiffTotalUtts$ of $\StackcheckAsevenTotalUtts$
transcripts differ).
Unmodified Gemma-3-4B-IT scores $\WsOneNativePiqa\%$ PIQA on one host
and $\WsFourDBzeroNativePiqa\%$ on another ($\NativeGemmaThreeHostPiqaDiffItems$ of $\PiqaEvalItems$
items); within Tables~\ref{tab:ws1_s2} and~\ref{tab:ws1_5norm} every
row was run on the same host as its native reference.
In Table~\ref{tab:ws4d}, the two rows marked $\dagger$ were run on a host where the unmodified model scores $\WsFourDWoneBzeroNativePiqa\%$ and the other rows on one where it scores $\WsFourDBzeroNativePiqa\%$; each row's difference is taken against the native value of its own host, and Table~\ref{tab:centerpiece} repeats the PIQA values.
The two runs of each optimizer pair in Table~\ref{tab:ws2_asr} were made on different hosts, and for the two pairs into Gemma-3-4B-IT on different stacks; we did not measure the effect of host or stack on training.
One host produced GPU faults on one day; the two affected cells were
rerun.

\input{sections/91_protocol}

%% file: tables/tab_app_models.tex
\afterpage{\clearpage\begin{landscape}\thispagestyle{plain}
\begin{table}[p]
\caption{\textbf{Pretrained frozen language models and speech encoders evaluated in this study.} For language models, $s_M$ is the architecture's input embedding multiplier measured in \texttt{bfloat16} and $\bar r_M$ is the mean token radius of Eq.~\ref{eq:token-radius} in \texttt{inputs\_embeds} space (spanning a $\RadiusSpreadFactor\times$ range across families). Gemma-3-4B-IT is a vision-language model; we use its text interface only.}
\label{tab:app_models}
\vspace{4pt}
\centering
\begin{adjustbox}{max width=\linewidth, max totalheight=395pt}%
\setlength{\tabcolsep}{4.2pt}
\begin{tabular}{@{}llrrrr>{\raggedright\hyphenpenalty=10000\exhyphenpenalty=10000\arraybackslash}p{4.5cm}@{}}
\toprule
\textbf{Model / Encoder} & \textbf{HuggingFace Checkpoint ID} & \textbf{Width $d$} & \textbf{Vocab $|V|$} & \textbf{Scale $s_M$} & \textbf{Radius $\bar r_M$} & \textbf{Experimental Role} \\
\midrule
Gemma-3-4B-IT \citep{gemma2025gemma3} & \texttt{google/gemma-3-4b-it} & $\GemmaThreeDim$ & $\GemmaThreeVocab$ & $\GemmaThreeEmbedScaleBf$ & $\RadiusGemmaThree$ & Vocabulary transplant target; speech adapter \& bridge target \\
Gemma-2-2B-IT \citep{gemma2024gemma2} & \texttt{google/gemma-2-2b-it} & $\GemmaTwoDim$ & $\GemmaTwoVocab$ & $\GemmaTwoEmbedScaleBf$ & $\RadiusGemmaTwo$ & Speech adapter target; bridge source \& target; transplant donor \\
Qwen2.5-1.5B-IT \citep{yang2024qwen25} & \texttt{Qwen/Qwen2.5-1.5B-Instruct} & $\QwenDim$ & $\QwenVocab$ & $\UnitEmbedScale$ & $\RadiusQwenOneFiveB$ & Speech adapter target; bridge source \& target \\
Qwen3-4B \citep{qwen2025qwen3} & \texttt{Qwen/Qwen3-4B} & $\QwenThreeDim$ & $\QwenThreeVocab$ & $\UnitEmbedScale$ & $\RadiusQwenThreeFourB$ & Same-width vocabulary transplant donor \\
SmolLM2-1.7B-IT \citep{allal2025smollm2} & \texttt{HuggingFaceTB/SmolLM2-1.7B-Instruct} & $\SmolDim$ & $\SmolVocab$ & $\UnitEmbedScale$ & $\RadiusSmolLMTwo$ & Cross-width vocabulary transplant donor \\
Phi-2 \citep{li2023phi2} & \texttt{microsoft/phi-2} & $\PhiDim$ & $\PhiVocab$ & $\UnitEmbedScale$ & $\RadiusPhiTwo$ & Base-model vocabulary transplant donor \\
\midrule
Whisper-Small \citep{radford2023whisper} & \texttt{openai/whisper-small} & $\WhisperDim$ & --- & --- & --- & Frozen speech encoder ($\WhisperFrameHz$\,Hz active-frame trimmed) \\
Moonshine-Base \citep{jeffries2024moonshine} & \texttt{UsefulSensors/moonshine-base} & $\MoonshineDim$ & --- & --- & --- & Frozen speech encoder ($\MoonshineFrameHz$\,Hz rotary variable-length) \\
\bottomrule
\end{tabular}%
\end{adjustbox}
\end{table}
\end{landscape}}

%% file: tables/tab_app_licenses.tex
\afterpage{\clearpage\begin{landscape}\thispagestyle{plain}
\begin{table}[p]
\caption{\textbf{Licenses of the pretrained models, speech encoders, training corpora, and evaluation benchmarks used in this study,} as stated on each artifact's upstream repository or model card.}
\label{tab:app_licenses}
\vspace{4pt}
\centering
\begin{adjustbox}{max width=\linewidth, max totalheight=395pt}%
\setlength{\tabcolsep}{5.0pt}
\begin{tabular}{@{}ll>{\raggedright\hyphenpenalty=10000\exhyphenpenalty=10000\arraybackslash}p{6.2cm}l@{}}
\toprule
\textbf{Artifact} & \textbf{Identifier / Split} & \textbf{Usage in Study} & \textbf{License} \\
\midrule
\multicolumn{4}{@{}l}{\textit{Pretrained frozen language model backbones and tokenizers}} \\
Gemma-3-4B-IT \citep{gemma2025gemma3} & \texttt{google/gemma-3-4b-it} & Frozen LLM \& tokenizer & Gemma Terms of Use \\
Gemma-2-2B-IT \citep{gemma2024gemma2} & \texttt{google/gemma-2-2b-it} & Frozen LLM \& tokenizer & Gemma Terms of Use \\
Qwen2.5-1.5B-IT \citep{yang2024qwen25} & \texttt{Qwen/Qwen2.5-1.5B-Instruct} & Frozen LLM \& tokenizer & Apache-2.0 \\
Qwen3-4B \citep{qwen2025qwen3} & \texttt{Qwen/Qwen3-4B} & Frozen donor LLM \& tokenizer & Apache-2.0 \\
SmolLM2-1.7B-IT \citep{allal2025smollm2} & \texttt{HuggingFaceTB/SmolLM2-1.7B-Instruct} & Frozen donor LLM \& tokenizer & Apache-2.0 \\
Phi-2 \citep{li2023phi2} & \texttt{microsoft/phi-2} & Frozen donor LLM \& tokenizer & MIT \\
\midrule
\multicolumn{4}{@{}l}{\textit{Pretrained frozen speech encoders}} \\
Whisper-Small \citep{radford2023whisper} & \texttt{openai/whisper-small} & Frozen speech encoder; native decoder as reference (Table~\ref{tab:ws2_asr}) & Apache-2.0 / MIT \\
Moonshine-Base \citep{jeffries2024moonshine} & \texttt{UsefulSensors/moonshine-base} & Frozen speech encoder; native decoder as reference (Table~\ref{tab:ws2_asr}) & MIT \\
\midrule
\multicolumn{4}{@{}l}{\textit{Training corpora and zero-shot evaluation benchmarks}} \\
LibriSpeech ASR \citep{panayotov2015librispeech} & \texttt{openslr/librispeech\_asr} & Speech adapter \& bridge fit (\LibriTrainHours\,h, \LibriTrainUtts\ utts); \texttt{test-clean}/\texttt{other} WER & CC-BY-4.0 \\
WikiText-103 \citep{merity2017pointer} & \texttt{Salesforce/wikitext (wikitext-103-raw-v1)} & Text \& mixed bridge fit (\LibriTrainUtts\ chunks) & CC-BY-SA-3.0 \\
PIQA \citep{bisk2020piqa} & \texttt{ybisk/piqa} (\texttt{validation}) & Zero-shot commonsense evaluation (\PiqaEvalItems\ items) & AFL-3.0 \\
HellaSwag \citep{zellers2019hellaswag} & \texttt{Rowan/hellaswag} (\texttt{validation}) & Zero-shot NLI completion evaluation (\HsEvalItems\ items) & MIT \\
\bottomrule
\end{tabular}%
\end{adjustbox}
\end{table}
\end{landscape}}

%% file: tables/tab_ws1_5norm.tex
\begin{table}[ht]
\caption{\textbf{Full five-normalization ablation for closed-form vocabulary transplants into frozen Gemma-3-4B-IT.} Accuracy ($\uparrow$) is measured zero-shot on PIQA (\PiqaEvalItems\ items, \PiqaChancePct\% chance) and HellaSwag (\HsEvalItems\ items, \HsChancePct\% chance) with the output head kept native; tying the output head to the transplanted table lowers least-squares target-radius accuracy by \WsOneUntiedVsTiedMinGainPiqa--\WsOneUntiedVsTiedMaxGainPiqa\,pp on PIQA and \WsOneUntiedVsTiedMinGainHs--\WsOneUntiedVsTiedMaxGainHs\,pp on HellaSwag. Unit $\ell_2$ normalization without target radius scaling (radius ratio \WsOneQwenLstsqLtwoRr) drops every donor and solver to chance (\WsOneLtwoMinPiqa\%--\WsOneLtwoMaxPiqa\% PIQA), whereas RMS norm and layer norm calibrated from anchor statistics give radius ratios of \WsOneLstsqRmsLnMinRr--\WsOneLstsqRmsLnMaxRr\ on least-squares maps.}
\label{tab:ws1_5norm}
\centering
\resizebox{\textwidth}{!}{%
\setlength{\tabcolsep}{4.5pt}
\begin{tabular}{@{}llrrrrrr@{}}
\toprule
& & \multicolumn{3}{c}{\textbf{Unconstrained Least Squares}} & \multicolumn{3}{c}{\textbf{Orthogonal Procrustes}} \\
\cmidrule(lr){3-5} \cmidrule(lr){6-8}
\textbf{Donor LLM ($\to$ Gemma-3-4B-IT)} & \textbf{Output Norm} & \textbf{Radius Ratio} & \textbf{PIQA $\uparrow$} & \textbf{HellaSwag $\uparrow$} & \textbf{Radius Ratio} & \textbf{PIQA $\uparrow$} & \textbf{HellaSwag $\uparrow$} \\
\midrule
Native / identity gate (Gemma-3-4B-IT) & --- & \WsOneNativeRr & \textbf{\WsOneNativePiqa\%} & \textbf{\WsOneNativeHs\%} & \WsOneGateRr & \textbf{\WsOneGatePiqa\%} & \textbf{\WsOneGateHs\%} \\
\midrule
Qwen3-4B & None & \WsOneQwenLstsqNoneRr & \textbf{\WsOneQwenLstsqNonePiqa\%} & \WsOneQwenLstsqNoneHs\% & \WsOneQwenProcNoneRr & \WsOneQwenProcNonePiqa\% & \WsOneQwenProcNoneHs\% \\
Qwen3-4B & Unit $\ell_2$ norm & \WsOneQwenLstsqLtwoRr & \WsOneQwenLstsqLtwoPiqa\% & \WsOneQwenLstsqLtwoHs\% & \WsOneQwenProcLtwoRr & \WsOneQwenProcLtwoPiqa\% & \WsOneQwenProcLtwoHs\% \\
Qwen3-4B & Target-radius norm & \WsOneQwenLstsqScaledRr & \textbf{\WsOneQwenLstsqScaledPiqa\%} & \textbf{\WsOneQwenLstsqScaledHs\%} & \WsOneQwenProcScaledRr & \WsOneQwenProcScaledPiqa\% & \WsOneQwenProcScaledHs\% \\
Qwen3-4B & RMS norm & \WsOneQwenLstsqRmsRr & \WsOneQwenLstsqRmsPiqa\% & \WsOneQwenLstsqRmsHs\% & \WsOneQwenProcRmsRr & \WsOneQwenProcRmsPiqa\% & \WsOneQwenProcRmsHs\% \\
Qwen3-4B & Layer norm & \WsOneQwenLstsqLnRr & \WsOneQwenLstsqLnPiqa\% & \WsOneQwenLstsqLnHs\% & \WsOneQwenProcLnRr & \textbf{\WsOneQwenProcLnPiqa\%} & \textbf{\WsOneQwenProcLnHs\%} \\
\cmidrule(lr){1-8}
SmolLM2-1.7B-IT & None & \WsOneSmolLstsqNoneRr & \WsOneSmolLstsqNonePiqa\% & \WsOneSmolLstsqNoneHs\% & \WsOneSmolProcNoneRr & \WsOneSmolProcNonePiqa\% & \WsOneSmolProcNoneHs\% \\
SmolLM2-1.7B-IT & Unit $\ell_2$ norm & \WsOneSmolLstsqLtwoRr & \WsOneSmolLstsqLtwoPiqa\% & \WsOneSmolLstsqLtwoHs\% & \WsOneSmolProcLtwoRr & \WsOneSmolProcLtwoPiqa\% & \WsOneSmolProcLtwoHs\% \\
SmolLM2-1.7B-IT & Target-radius norm & \WsOneSmolLstsqScaledRr & \textbf{\WsOneSmolLstsqScaledPiqa\%} & \textbf{\WsOneSmolLstsqScaledHs\%} & \WsOneSmolProcScaledRr & \WsOneSmolProcScaledPiqa\% & \WsOneSmolProcScaledHs\% \\
SmolLM2-1.7B-IT & RMS norm & \WsOneSmolLstsqRmsRr & \WsOneSmolLstsqRmsPiqa\% & \WsOneSmolLstsqRmsHs\% & \WsOneSmolProcRmsRr & \WsOneSmolProcRmsPiqa\% & \WsOneSmolProcRmsHs\% \\
SmolLM2-1.7B-IT & Layer norm & \WsOneSmolLstsqLnRr & \WsOneSmolLstsqLnPiqa\% & \WsOneSmolLstsqLnHs\% & \WsOneSmolProcLnRr & \textbf{\WsOneSmolProcLnPiqa\%} & \textbf{\WsOneSmolProcLnHs\%} \\
\cmidrule(lr){1-8}
Phi-2 & None & \WsOnePhiLstsqNoneRr & \WsOnePhiLstsqNonePiqa\% & \WsOnePhiLstsqNoneHs\% & \WsOnePhiProcNoneRr & \WsOnePhiProcNonePiqa\% & \WsOnePhiProcNoneHs\% \\
Phi-2 & Unit $\ell_2$ norm & \WsOnePhiLstsqLtwoRr & \WsOnePhiLstsqLtwoPiqa\% & \WsOnePhiLstsqLtwoHs\% & \WsOnePhiProcLtwoRr & \WsOnePhiProcLtwoPiqa\% & \WsOnePhiProcLtwoHs\% \\
Phi-2 & Target-radius norm & \WsOnePhiLstsqScaledRr & \textbf{\WsOnePhiLstsqScaledPiqa\%} & \WsOnePhiLstsqScaledHs\% & \WsOnePhiProcScaledRr & \WsOnePhiProcScaledPiqa\% & \WsOnePhiProcScaledHs\% \\
Phi-2 & RMS norm & \WsOnePhiLstsqRmsRr & \WsOnePhiLstsqRmsPiqa\% & \WsOnePhiLstsqRmsHs\% & \WsOnePhiProcRmsRr & \WsOnePhiProcRmsPiqa\% & \WsOnePhiProcRmsHs\% \\
Phi-2 & Layer norm & \WsOnePhiLstsqLnRr & \WsOnePhiLstsqLnPiqa\% & \textbf{\WsOnePhiLstsqLnHs\%} & \WsOnePhiProcLnRr & \textbf{\WsOnePhiProcLnPiqa\%} & \textbf{\WsOnePhiProcLnHs\%} \\
\bottomrule
\end{tabular}%
}
\end{table}

%% file: tables/tab_token_hits.tex
\afterpage{\clearpage\begin{landscape}\thispagestyle{plain}
\begin{table}[p]
\caption{\textbf{Evaluation token-hit breakdown on frozen Gemma-3-4B-IT across vocabulary-transplant donors and zero-shot benchmarks (Table~\ref{tab:ws1_s2} and Table~\ref{tab:ws4d}).} Each cell reports percentage shares ($\text{training anchor} / \text{held-out anchor} / \text{native row}$, summing to $\FullSignalPct\%$) of token occurrences and unique token types under Gemma-3-4B-IT tokenization across all scored $(\text{context}, \text{continuation})$ inputs, context prefixes alone, and scored continuations alone on PIQA (\PiqaEvalItems\ items) and HellaSwag (\HsEvalItems\ items). $^\dagger$Candidate surface-form matching ablation (\TokenHitsQwenThreeSurfaceOverwrittenRows\ anchor rows overwritten with chat-control exemptions disabled; Appendix~\ref{app:protocol}).}
\label{tab:token_hits}
\vspace{4pt}
\centering
\begin{adjustbox}{max width=\linewidth, max totalheight=395pt}%
\setlength{\tabcolsep}{3.0pt}
\begin{tabular}{@{}llrrrrrr@{}}
\toprule
& & \multicolumn{2}{c}{\textbf{All Input Tokens (\%)}} & \multicolumn{2}{c}{\textbf{Context Prefix (\%)}} & \multicolumn{2}{c}{\textbf{Scored Continuation (\%)}} \\
\cmidrule(lr){3-4} \cmidrule(lr){5-6} \cmidrule(lr){7-8}
\textbf{Donor LLM} & \textbf{Benchmark} & \shortstack[r]{\strut\textbf{Occurrences}\\\textbf{Train / Test / Nat.}} & \shortstack[r]{\strut\textbf{Types}\\\textbf{Train / Test / Nat.}} & \shortstack[r]{\strut\textbf{Occurrences}\\\textbf{Train / Test / Nat.}} & \shortstack[r]{\strut\textbf{Types}\\\textbf{Train / Test / Nat.}} & \shortstack[r]{\strut\textbf{Occurrences}\\\textbf{Train / Test / Nat.}} & \shortstack[r]{\strut\textbf{Types}\\\textbf{Train / Test / Nat.}} \\
\midrule
Qwen3-4B & PIQA & \TokenHitsQwenThreePiqaCombOccTr / \TokenHitsQwenThreePiqaCombOccTe / \TokenHitsQwenThreePiqaCombOccNat & \TokenHitsQwenThreePiqaCombTypTr / \TokenHitsQwenThreePiqaCombTypTe / \TokenHitsQwenThreePiqaCombTypNat & \TokenHitsQwenThreePiqaCtxOccTr / \TokenHitsQwenThreePiqaCtxOccTe / \TokenHitsQwenThreePiqaCtxOccNat & \TokenHitsQwenThreePiqaCtxTypTr / \TokenHitsQwenThreePiqaCtxTypTe / \TokenHitsQwenThreePiqaCtxTypNat & \TokenHitsQwenThreePiqaContOccTr / \TokenHitsQwenThreePiqaContOccTe / \TokenHitsQwenThreePiqaContOccNat & \TokenHitsQwenThreePiqaContTypTr / \TokenHitsQwenThreePiqaContTypTe / \TokenHitsQwenThreePiqaContTypNat \\
Qwen3-4B & HellaSwag & \TokenHitsQwenThreeHsCombOccTr / \TokenHitsQwenThreeHsCombOccTe / \TokenHitsQwenThreeHsCombOccNat & \TokenHitsQwenThreeHsCombTypTr / \TokenHitsQwenThreeHsCombTypTe / \TokenHitsQwenThreeHsCombTypNat & \TokenHitsQwenThreeHsCtxOccTr / \TokenHitsQwenThreeHsCtxOccTe / \TokenHitsQwenThreeHsCtxOccNat & \TokenHitsQwenThreeHsCtxTypTr / \TokenHitsQwenThreeHsCtxTypTe / \TokenHitsQwenThreeHsCtxTypNat & \TokenHitsQwenThreeHsContOccTr / \TokenHitsQwenThreeHsContOccTe / \TokenHitsQwenThreeHsContOccNat & \TokenHitsQwenThreeHsContTypTr / \TokenHitsQwenThreeHsContTypTe / \TokenHitsQwenThreeHsContTypNat \\
\cmidrule(lr){1-8}
SmolLM2-1.7B-IT & PIQA & \TokenHitsSmolPiqaCombOccTr / \TokenHitsSmolPiqaCombOccTe / \TokenHitsSmolPiqaCombOccNat & \TokenHitsSmolPiqaCombTypTr / \TokenHitsSmolPiqaCombTypTe / \TokenHitsSmolPiqaCombTypNat & \TokenHitsSmolPiqaCtxOccTr / \TokenHitsSmolPiqaCtxOccTe / \TokenHitsSmolPiqaCtxOccNat & \TokenHitsSmolPiqaCtxTypTr / \TokenHitsSmolPiqaCtxTypTe / \TokenHitsSmolPiqaCtxTypNat & \TokenHitsSmolPiqaContOccTr / \TokenHitsSmolPiqaContOccTe / \TokenHitsSmolPiqaContOccNat & \TokenHitsSmolPiqaContTypTr / \TokenHitsSmolPiqaContTypTe / \TokenHitsSmolPiqaContTypNat \\
SmolLM2-1.7B-IT & HellaSwag & \TokenHitsSmolHsCombOccTr / \TokenHitsSmolHsCombOccTe / \TokenHitsSmolHsCombOccNat & \TokenHitsSmolHsCombTypTr / \TokenHitsSmolHsCombTypTe / \TokenHitsSmolHsCombTypNat & \TokenHitsSmolHsCtxOccTr / \TokenHitsSmolHsCtxOccTe / \TokenHitsSmolHsCtxOccNat & \TokenHitsSmolHsCtxTypTr / \TokenHitsSmolHsCtxTypTe / \TokenHitsSmolHsCtxTypNat & \TokenHitsSmolHsContOccTr / \TokenHitsSmolHsContOccTe / \TokenHitsSmolHsContOccNat & \TokenHitsSmolHsContTypTr / \TokenHitsSmolHsContTypTe / \TokenHitsSmolHsContTypNat \\
\cmidrule(lr){1-8}
Phi-2 & PIQA & \TokenHitsPhiPiqaCombOccTr / \TokenHitsPhiPiqaCombOccTe / \TokenHitsPhiPiqaCombOccNat & \TokenHitsPhiPiqaCombTypTr / \TokenHitsPhiPiqaCombTypTe / \TokenHitsPhiPiqaCombTypNat & \TokenHitsPhiPiqaCtxOccTr / \TokenHitsPhiPiqaCtxOccTe / \TokenHitsPhiPiqaCtxOccNat & \TokenHitsPhiPiqaCtxTypTr / \TokenHitsPhiPiqaCtxTypTe / \TokenHitsPhiPiqaCtxTypNat & \TokenHitsPhiPiqaContOccTr / \TokenHitsPhiPiqaContOccTe / \TokenHitsPhiPiqaContOccNat & \TokenHitsPhiPiqaContTypTr / \TokenHitsPhiPiqaContTypTe / \TokenHitsPhiPiqaContTypNat \\
Phi-2 & HellaSwag & \TokenHitsPhiHsCombOccTr / \TokenHitsPhiHsCombOccTe / \TokenHitsPhiHsCombOccNat & \TokenHitsPhiHsCombTypTr / \TokenHitsPhiHsCombTypTe / \TokenHitsPhiHsCombTypNat & \TokenHitsPhiHsCtxOccTr / \TokenHitsPhiHsCtxOccTe / \TokenHitsPhiHsCtxOccNat & \TokenHitsPhiHsCtxTypTr / \TokenHitsPhiHsCtxTypTe / \TokenHitsPhiHsCtxTypNat & \TokenHitsPhiHsContOccTr / \TokenHitsPhiHsContOccTe / \TokenHitsPhiHsContOccNat & \TokenHitsPhiHsContTypTr / \TokenHitsPhiHsContTypTe / \TokenHitsPhiHsContTypNat \\
\cmidrule(lr){1-8}
Gemma-2-2B-IT & PIQA & \TokenHitsGemmaTwoPiqaCombOccTr / \TokenHitsGemmaTwoPiqaCombOccTe / \TokenHitsGemmaTwoPiqaCombOccNat & \TokenHitsGemmaTwoPiqaCombTypTr / \TokenHitsGemmaTwoPiqaCombTypTe / \TokenHitsGemmaTwoPiqaCombTypNat & \TokenHitsGemmaTwoPiqaCtxOccTr / \TokenHitsGemmaTwoPiqaCtxOccTe / \TokenHitsGemmaTwoPiqaCtxOccNat & \TokenHitsGemmaTwoPiqaCtxTypTr / \TokenHitsGemmaTwoPiqaCtxTypTe / \TokenHitsGemmaTwoPiqaCtxTypNat & \TokenHitsGemmaTwoPiqaContOccTr / \TokenHitsGemmaTwoPiqaContOccTe / \TokenHitsGemmaTwoPiqaContOccNat & \TokenHitsGemmaTwoPiqaContTypTr / \TokenHitsGemmaTwoPiqaContTypTe / \TokenHitsGemmaTwoPiqaContTypNat \\
Gemma-2-2B-IT & HellaSwag & \TokenHitsGemmaTwoHsCombOccTr / \TokenHitsGemmaTwoHsCombOccTe / \TokenHitsGemmaTwoHsCombOccNat & \TokenHitsGemmaTwoHsCombTypTr / \TokenHitsGemmaTwoHsCombTypTe / \TokenHitsGemmaTwoHsCombTypNat & \TokenHitsGemmaTwoHsCtxOccTr / \TokenHitsGemmaTwoHsCtxOccTe / \TokenHitsGemmaTwoHsCtxOccNat & \TokenHitsGemmaTwoHsCtxTypTr / \TokenHitsGemmaTwoHsCtxTypTe / \TokenHitsGemmaTwoHsCtxTypNat & \TokenHitsGemmaTwoHsContOccTr / \TokenHitsGemmaTwoHsContOccTe / \TokenHitsGemmaTwoHsContOccNat & \TokenHitsGemmaTwoHsContTypTr / \TokenHitsGemmaTwoHsContTypTe / \TokenHitsGemmaTwoHsContTypNat \\
\midrule
Qwen3-4B (surface matcher$^\dagger$) & PIQA & \TokenHitsQwenThreeSurfacePiqaCombOccTr / \TokenHitsQwenThreeSurfacePiqaCombOccTe / \TokenHitsQwenThreeSurfacePiqaCombOccNat & \TokenHitsQwenThreeSurfacePiqaCombTypTr / \TokenHitsQwenThreeSurfacePiqaCombTypTe / \TokenHitsQwenThreeSurfacePiqaCombTypNat & \TokenHitsQwenThreeSurfacePiqaCtxOccTr / \TokenHitsQwenThreeSurfacePiqaCtxOccTe / \TokenHitsQwenThreeSurfacePiqaCtxOccNat & \TokenHitsQwenThreeSurfacePiqaCtxTypTr / \TokenHitsQwenThreeSurfacePiqaCtxTypTe / \TokenHitsQwenThreeSurfacePiqaCtxTypNat & \TokenHitsQwenThreeSurfacePiqaContOccTr / \TokenHitsQwenThreeSurfacePiqaContOccTe / \TokenHitsQwenThreeSurfacePiqaContOccNat & \TokenHitsQwenThreeSurfacePiqaContTypTr / \TokenHitsQwenThreeSurfacePiqaContTypTe / \TokenHitsQwenThreeSurfacePiqaContTypNat \\
Qwen3-4B (surface matcher$^\dagger$) & HellaSwag & \TokenHitsQwenThreeSurfaceHsCombOccTr / \TokenHitsQwenThreeSurfaceHsCombOccTe / \TokenHitsQwenThreeSurfaceHsCombOccNat & \TokenHitsQwenThreeSurfaceHsCombTypTr / \TokenHitsQwenThreeSurfaceHsCombTypTe / \TokenHitsQwenThreeSurfaceHsCombTypNat & \TokenHitsQwenThreeSurfaceHsCtxOccTr / \TokenHitsQwenThreeSurfaceHsCtxOccTe / \TokenHitsQwenThreeSurfaceHsCtxOccNat & \TokenHitsQwenThreeSurfaceHsCtxTypTr / \TokenHitsQwenThreeSurfaceHsCtxTypTe / \TokenHitsQwenThreeSurfaceHsCtxTypNat & \TokenHitsQwenThreeSurfaceHsContOccTr / \TokenHitsQwenThreeSurfaceHsContOccTe / \TokenHitsQwenThreeSurfaceHsContOccNat & \TokenHitsQwenThreeSurfaceHsContTypTr / \TokenHitsQwenThreeSurfaceHsContTypTe / \TokenHitsQwenThreeSurfaceHsContTypNat \\
\bottomrule
\end{tabular}%
\end{adjustbox}
\end{table}
\end{landscape}}

%% file: tables/tab_ws3_fits.tex
\afterpage{\clearpage\begin{landscape}\thispagestyle{plain}
\begin{table}[p]
\caption{\textbf{Lexical alignment diagnostics across all \WsThreeTotalPairs\ ordered LLM pairs for closed-form, text-trained, soft-prompt-trained, and mixed bridges.} Held-out cosine similarity ($\uparrow$), nearest-neighbor retrieval P@1 ($\uparrow$), and median output-to-target token radius ratio are evaluated on the unseen \AnchorTestPct\% shared-vocabulary split ($N_{\text{test}}$ tokens). Validation negative log-likelihoods ($\downarrow$) are reported on each gradient-trained bridge's held-out split (transcripts \WsFourMinVal--\WsFourMaxVal; soft-prompt audio \WsFourBMinAudioVal--\WsFourBMaxAudioVal\ vs.\ text \WsFourBMinTextVal--\WsFourBMaxTextVal; WikiText \WsFourCWikiMinVal--\WsFourCWikiMaxVal).}
\label{tab:ws3_fits}
\vspace{2pt}
\centering
\renewcommand{\arraystretch}{0.86}
\begin{adjustbox}{max width=\linewidth, max totalheight=422pt}%
\setlength{\tabcolsep}{4.2pt}
\begin{tabular}{@{}llcrrrrr@{}}
\toprule
\textbf{Source LLM $\to$ Target LLM} & \textbf{Bridge Fit \& Val NLL} & \textbf{Width} & \shortstack[r]{\strut\textbf{Shared}\\\textbf{($N_{\text{test}}$)}} & \shortstack[r]{\strut\textbf{Held-out}\\\textbf{$\cos\uparrow$}} & \shortstack[r]{\strut\textbf{Held-out}\\\textbf{P@1 $\uparrow$}} & \shortstack[r]{\strut\textbf{Raw}\\\textbf{Ratio}} & \shortstack[r]{\strut\textbf{Calib.}\\\textbf{Ratio}} \\
\midrule
\multicolumn{8}{@{}l}{\textit{Closed-form orthogonal Procrustes bridges (fit on \AnchorTrainPct\% of shared vocabulary tokens)}} \\
Qwen2.5-1.5B-IT $\to$ Gemma-2-2B-IT & Procrustes & $\QwenDim\!\to\!\GemmaTwoDim$ & \WsThreeFitQwenToGemmaTwoProcAnchors (\WsThreeFitQwenToGemmaTwoProcTestN) & \WsThreeFitQwenToGemmaTwoProcCos & \textbf{\WsThreeFitQwenToGemmaTwoProcPatOne\%} & \WsThreeFitQwenToGemmaTwoProcRawRr & \WsThreeFitQwenToGemmaTwoProcNormRr \\
Gemma-2-2B-IT $\to$ Qwen2.5-1.5B-IT & Procrustes & $\GemmaTwoDim\!\to\!\QwenDim$ & \WsThreeFitGemmaTwoToQwenProcAnchors (\WsThreeFitGemmaTwoToQwenProcTestN) & \WsThreeFitGemmaTwoToQwenProcCos & \textbf{\WsThreeFitGemmaTwoToQwenProcPatOne\%} & \WsThreeFitGemmaTwoToQwenProcRawRr & \WsThreeFitGemmaTwoToQwenProcNormRr \\
Qwen2.5-1.5B-IT $\to$ Gemma-3-4B-IT & Procrustes & $\QwenDim\!\to\!\GemmaThreeDim$ & \WsThreeFitQwenToGemmaThreeProcAnchors (\WsThreeFitQwenToGemmaThreeProcTestN) & \WsThreeFitQwenToGemmaThreeProcCos & \textbf{\WsThreeFitQwenToGemmaThreeProcPatOne\%} & \WsThreeFitQwenToGemmaThreeProcRawRr & \WsThreeFitQwenToGemmaThreeProcNormRr \\
Gemma-2-2B-IT $\to$ Gemma-3-4B-IT & Procrustes & $\GemmaTwoDim\!\to\!\GemmaThreeDim$ & \WsThreeFitGemmaTwoToGemmaThreeProcAnchors (\WsThreeFitGemmaTwoToGemmaThreeProcTestN) & \WsThreeFitGemmaTwoToGemmaThreeProcCos & \textbf{\WsThreeFitGemmaTwoToGemmaThreeProcPatOne\%} & \WsThreeFitGemmaTwoToGemmaThreeProcRawRr & \WsThreeFitGemmaTwoToGemmaThreeProcNormRr \\
\midrule
\multicolumn{8}{@{}l}{\textit{Closed-form unconstrained least-squares bridges (fit on \AnchorTrainPct\% of shared vocabulary tokens)}} \\
Qwen2.5-1.5B-IT $\to$ Gemma-2-2B-IT & Least squares & $\QwenDim\!\to\!\GemmaTwoDim$ & \WsThreeFitQwenToGemmaTwoLstsqAnchors (\WsThreeFitQwenToGemmaTwoLstsqTestN) & \WsThreeFitQwenToGemmaTwoLstsqCos & \textbf{\WsThreeFitQwenToGemmaTwoLstsqPatOne\%} & \WsThreeFitQwenToGemmaTwoLstsqRawRr & \WsThreeFitQwenToGemmaTwoLstsqNormRr \\
Gemma-2-2B-IT $\to$ Qwen2.5-1.5B-IT & Least squares & $\GemmaTwoDim\!\to\!\QwenDim$ & \WsThreeFitGemmaTwoToQwenLstsqAnchors (\WsThreeFitGemmaTwoToQwenLstsqTestN) & \WsThreeFitGemmaTwoToQwenLstsqCos & \textbf{\WsThreeFitGemmaTwoToQwenLstsqPatOne\%} & \WsThreeFitGemmaTwoToQwenLstsqRawRr & \WsThreeFitGemmaTwoToQwenLstsqNormRr \\
Qwen2.5-1.5B-IT $\to$ Gemma-3-4B-IT & Least squares & $\QwenDim\!\to\!\GemmaThreeDim$ & \WsThreeFitQwenToGemmaThreeLstsqAnchors (\WsThreeFitQwenToGemmaThreeLstsqTestN) & \WsThreeFitQwenToGemmaThreeLstsqCos & \textbf{\WsThreeFitQwenToGemmaThreeLstsqPatOne\%} & \WsThreeFitQwenToGemmaThreeLstsqRawRr & \WsThreeFitQwenToGemmaThreeLstsqNormRr \\
Gemma-2-2B-IT $\to$ Gemma-3-4B-IT & Least squares & $\GemmaTwoDim\!\to\!\GemmaThreeDim$ & \WsThreeFitGemmaTwoToGemmaThreeLstsqAnchors (\WsThreeFitGemmaTwoToGemmaThreeLstsqTestN) & \WsThreeFitGemmaTwoToGemmaThreeLstsqCos & \textbf{\WsThreeFitGemmaTwoToGemmaThreeLstsqPatOne\%} & \WsThreeFitGemmaTwoToGemmaThreeLstsqRawRr & \WsThreeFitGemmaTwoToGemmaThreeLstsqNormRr \\
\midrule
\multicolumn{8}{@{}l}{\textit{Text-trained bridges on LibriSpeech transcripts (Muon, \LibriTrainUtts\ utterances; transcript val NLL)}} \\
Qwen2.5-1.5B-IT $\to$ Gemma-2-2B-IT & Transcript (Layer norm; NLL \WsFourFitQwenToGemmaTwoLnVal) & $\QwenDim\!\to\!\GemmaTwoDim$ & \WsFourFitQwenToGemmaTwoLnAnchors (\WsFourFitQwenToGemmaTwoLnTestN) & \WsFourFitQwenToGemmaTwoLnCos & \WsFourFitQwenToGemmaTwoLnPatOne\% & --- & \WsFourFitQwenToGemmaTwoLnRr \\
Qwen2.5-1.5B-IT $\to$ Gemma-2-2B-IT & Transcript (Target-radius norm; NLL \WsFourFitQwenToGemmaTwoScaledVal) & $\QwenDim\!\to\!\GemmaTwoDim$ & \WsFourFitQwenToGemmaTwoScaledAnchors (\WsFourFitQwenToGemmaTwoScaledTestN) & \WsFourFitQwenToGemmaTwoScaledCos & \WsFourFitQwenToGemmaTwoScaledPatOne\% & --- & \WsFourFitQwenToGemmaTwoScaledRr \\
Gemma-2-2B-IT $\to$ Qwen2.5-1.5B-IT & Transcript (Layer norm; NLL \WsFourFitGemmaTwoToQwenLnVal) & $\GemmaTwoDim\!\to\!\QwenDim$ & \WsFourFitGemmaTwoToQwenLnAnchors (\WsFourFitGemmaTwoToQwenLnTestN) & \WsFourFitGemmaTwoToQwenLnCos & \WsFourFitGemmaTwoToQwenLnPatOne\% & --- & \WsFourFitGemmaTwoToQwenLnRr \\
Gemma-2-2B-IT $\to$ Qwen2.5-1.5B-IT & Transcript (Target-radius norm; NLL \WsFourFitGemmaTwoToQwenScaledVal) & $\GemmaTwoDim\!\to\!\QwenDim$ & \WsFourFitGemmaTwoToQwenScaledAnchors (\WsFourFitGemmaTwoToQwenScaledTestN) & \WsFourFitGemmaTwoToQwenScaledCos & \WsFourFitGemmaTwoToQwenScaledPatOne\% & --- & \WsFourFitGemmaTwoToQwenScaledRr \\
Qwen2.5-1.5B-IT $\to$ Gemma-3-4B-IT & Transcript (Layer norm; NLL \WsFourFitQwenToGemmaThreeLnVal) & $\QwenDim\!\to\!\GemmaThreeDim$ & \WsFourFitQwenToGemmaThreeLnAnchors (\WsFourFitQwenToGemmaThreeLnTestN) & \WsFourFitQwenToGemmaThreeLnCos & \WsFourFitQwenToGemmaThreeLnPatOne\% & --- & \WsFourFitQwenToGemmaThreeLnRr \\
Qwen2.5-1.5B-IT $\to$ Gemma-3-4B-IT & Transcript (Target-radius norm; NLL \WsFourFitQwenToGemmaThreeScaledVal) & $\QwenDim\!\to\!\GemmaThreeDim$ & \WsFourFitQwenToGemmaThreeScaledAnchors (\WsFourFitQwenToGemmaThreeScaledTestN) & \WsFourFitQwenToGemmaThreeScaledCos & \WsFourFitQwenToGemmaThreeScaledPatOne\% & --- & \WsFourFitQwenToGemmaThreeScaledRr \\
Gemma-2-2B-IT $\to$ Gemma-3-4B-IT & Transcript (Layer norm; NLL \WsFourFitGemmaTwoToGemmaThreeLnVal) & $\GemmaTwoDim\!\to\!\GemmaThreeDim$ & \WsFourFitGemmaTwoToGemmaThreeLnAnchors (\WsFourFitGemmaTwoToGemmaThreeLnTestN) & \WsFourFitGemmaTwoToGemmaThreeLnCos & \WsFourFitGemmaTwoToGemmaThreeLnPatOne\% & --- & \WsFourFitGemmaTwoToGemmaThreeLnRr \\
Gemma-2-2B-IT $\to$ Gemma-3-4B-IT & Transcript (Target-radius norm; NLL \WsFourFitGemmaTwoToGemmaThreeScaledVal) & $\GemmaTwoDim\!\to\!\GemmaThreeDim$ & \WsFourFitGemmaTwoToGemmaThreeScaledAnchors (\WsFourFitGemmaTwoToGemmaThreeScaledTestN) & \WsFourFitGemmaTwoToGemmaThreeScaledCos & \WsFourFitGemmaTwoToGemmaThreeScaledPatOne\% & --- & \WsFourFitGemmaTwoToGemmaThreeScaledRr \\
\midrule
\multicolumn{8}{@{}l}{\textit{Soft-prompt-trained bridges on frozen speech adapter outputs (Muon; audio / text val NLL)}} \\
Qwen2.5-1.5B-IT $\to$ Gemma-2-2B-IT & Soft-prompt (MS, LN; \WsFourBFitMoonshineQwenToGemmaTwoLnAudioVal / \WsFourBFitMoonshineQwenToGemmaTwoLnTextVal) & $\QwenDim\!\to\!\GemmaTwoDim$ & \WsFourBFitMoonshineQwenToGemmaTwoLnAnchors (\WsFourBFitMoonshineQwenToGemmaTwoLnTestN) & \WsFourBFitMoonshineQwenToGemmaTwoLnCos & \WsFourBFitMoonshineQwenToGemmaTwoLnPatOne\% & --- & \WsFourBFitMoonshineQwenToGemmaTwoLnRr \\
Qwen2.5-1.5B-IT $\to$ Gemma-2-2B-IT & Soft-prompt (Wh, LN; \WsFourBFitWhisperQwenToGemmaTwoLnAudioVal / \WsFourBFitWhisperQwenToGemmaTwoLnTextVal) & $\QwenDim\!\to\!\GemmaTwoDim$ & \WsFourBFitWhisperQwenToGemmaTwoLnAnchors (\WsFourBFitWhisperQwenToGemmaTwoLnTestN) & \WsFourBFitWhisperQwenToGemmaTwoLnCos & \WsFourBFitWhisperQwenToGemmaTwoLnPatOne\% & --- & \WsFourBFitWhisperQwenToGemmaTwoLnRr \\
Gemma-2-2B-IT $\to$ Qwen2.5-1.5B-IT & Soft-prompt (MS, LN; \WsFourBFitMoonshineGemmaTwoToQwenLnAudioVal / \WsFourBFitMoonshineGemmaTwoToQwenLnTextVal) & $\GemmaTwoDim\!\to\!\QwenDim$ & \WsFourBFitMoonshineGemmaTwoToQwenLnAnchors (\WsFourBFitMoonshineGemmaTwoToQwenLnTestN) & \WsFourBFitMoonshineGemmaTwoToQwenLnCos & \WsFourBFitMoonshineGemmaTwoToQwenLnPatOne\% & --- & \WsFourBFitMoonshineGemmaTwoToQwenLnRr \\
Gemma-2-2B-IT $\to$ Qwen2.5-1.5B-IT & Soft-prompt (MS, target-radius; \WsFourBFitMoonshineGemmaTwoToQwenScaledAudioVal / \WsFourBFitMoonshineGemmaTwoToQwenScaledTextVal) & $\GemmaTwoDim\!\to\!\QwenDim$ & \WsFourBFitMoonshineGemmaTwoToQwenScaledAnchors (\WsFourBFitMoonshineGemmaTwoToQwenScaledTestN) & \WsFourBFitMoonshineGemmaTwoToQwenScaledCos & \WsFourBFitMoonshineGemmaTwoToQwenScaledPatOne\% & --- & \WsFourBFitMoonshineGemmaTwoToQwenScaledRr \\
Gemma-2-2B-IT $\to$ Qwen2.5-1.5B-IT & Soft-prompt (Wh, LN; \WsFourBFitWhisperGemmaTwoToQwenLnAudioVal / \WsFourBFitWhisperGemmaTwoToQwenLnTextVal) & $\GemmaTwoDim\!\to\!\QwenDim$ & \WsFourBFitWhisperGemmaTwoToQwenLnAnchors (\WsFourBFitWhisperGemmaTwoToQwenLnTestN) & \WsFourBFitWhisperGemmaTwoToQwenLnCos & \WsFourBFitWhisperGemmaTwoToQwenLnPatOne\% & --- & \WsFourBFitWhisperGemmaTwoToQwenLnRr \\
Qwen2.5-1.5B-IT $\to$ Gemma-3-4B-IT & Soft-prompt (MS, LN; \WsFourBFitMoonshineQwenToGemmaThreeLnAudioVal / \WsFourBFitMoonshineQwenToGemmaThreeLnTextVal) & $\QwenDim\!\to\!\GemmaThreeDim$ & \WsFourBFitMoonshineQwenToGemmaThreeLnAnchors (\WsFourBFitMoonshineQwenToGemmaThreeLnTestN) & \WsFourBFitMoonshineQwenToGemmaThreeLnCos & \WsFourBFitMoonshineQwenToGemmaThreeLnPatOne\% & --- & \WsFourBFitMoonshineQwenToGemmaThreeLnRr \\
Qwen2.5-1.5B-IT $\to$ Gemma-3-4B-IT & Soft-prompt (Wh, LN; \WsFourBFitWhisperQwenToGemmaThreeLnAudioVal / \WsFourBFitWhisperQwenToGemmaThreeLnTextVal) & $\QwenDim\!\to\!\GemmaThreeDim$ & \WsFourBFitWhisperQwenToGemmaThreeLnAnchors (\WsFourBFitWhisperQwenToGemmaThreeLnTestN) & \WsFourBFitWhisperQwenToGemmaThreeLnCos & \WsFourBFitWhisperQwenToGemmaThreeLnPatOne\% & --- & \WsFourBFitWhisperQwenToGemmaThreeLnRr \\
Gemma-2-2B-IT $\to$ Gemma-3-4B-IT & Soft-prompt (MS, LN; \WsFourBFitMoonshineGemmaTwoToGemmaThreeLnAudioVal / \WsFourBFitMoonshineGemmaTwoToGemmaThreeLnTextVal) & $\GemmaTwoDim\!\to\!\GemmaThreeDim$ & \WsFourBFitMoonshineGemmaTwoToGemmaThreeLnAnchors (\WsFourBFitMoonshineGemmaTwoToGemmaThreeLnTestN) & \WsFourBFitMoonshineGemmaTwoToGemmaThreeLnCos & \WsFourBFitMoonshineGemmaTwoToGemmaThreeLnPatOne\% & --- & \WsFourBFitMoonshineGemmaTwoToGemmaThreeLnRr \\
Gemma-2-2B-IT $\to$ Gemma-3-4B-IT & Soft-prompt (Wh, LN; \WsFourBFitWhisperGemmaTwoToGemmaThreeLnAudioVal / \WsFourBFitWhisperGemmaTwoToGemmaThreeLnTextVal) & $\GemmaTwoDim\!\to\!\GemmaThreeDim$ & \WsFourBFitWhisperGemmaTwoToGemmaThreeLnAnchors (\WsFourBFitWhisperGemmaTwoToGemmaThreeLnTestN) & \WsFourBFitWhisperGemmaTwoToGemmaThreeLnCos & \WsFourBFitWhisperGemmaTwoToGemmaThreeLnPatOne\% & --- & \WsFourBFitWhisperGemmaTwoToGemmaThreeLnRr \\
\midrule
\multicolumn{8}{@{}l}{\textit{Text-trained bridges on WikiText-103 (Muon; WikiText / transcript val NLL)}} \\
Qwen2.5-1.5B-IT $\to$ Gemma-2-2B-IT & WikiText (LN; \WsFourCWikiFitQwenToGemmaTwoLnWikiVal / \WsFourCWikiFitQwenToGemmaTwoLnLsVal) & $\QwenDim\!\to\!\GemmaTwoDim$ & \WsFourCWikiFitQwenToGemmaTwoLnAnchors (\WsFourCWikiFitQwenToGemmaTwoLnTestN) & \WsFourCWikiFitQwenToGemmaTwoLnCos & \WsFourCWikiFitQwenToGemmaTwoLnPatOne\% & --- & \WsFourCWikiFitQwenToGemmaTwoLnRr \\
Gemma-2-2B-IT $\to$ Qwen2.5-1.5B-IT & WikiText (LN; \WsFourCWikiFitGemmaTwoToQwenLnWikiVal / \WsFourCWikiFitGemmaTwoToQwenLnLsVal) & $\GemmaTwoDim\!\to\!\QwenDim$ & \WsFourCWikiFitGemmaTwoToQwenLnAnchors (\WsFourCWikiFitGemmaTwoToQwenLnTestN) & \WsFourCWikiFitGemmaTwoToQwenLnCos & \WsFourCWikiFitGemmaTwoToQwenLnPatOne\% & --- & \WsFourCWikiFitGemmaTwoToQwenLnRr \\
Qwen2.5-1.5B-IT $\to$ Gemma-3-4B-IT & WikiText (LN; \WsFourCWikiFitQwenToGemmaThreeLnWikiVal / \WsFourCWikiFitQwenToGemmaThreeLnLsVal) & $\QwenDim\!\to\!\GemmaThreeDim$ & \WsFourCWikiFitQwenToGemmaThreeLnAnchors (\WsFourCWikiFitQwenToGemmaThreeLnTestN) & \WsFourCWikiFitQwenToGemmaThreeLnCos & \WsFourCWikiFitQwenToGemmaThreeLnPatOne\% & --- & \WsFourCWikiFitQwenToGemmaThreeLnRr \\
Gemma-2-2B-IT $\to$ Gemma-3-4B-IT & WikiText (LN; \WsFourCWikiFitGemmaTwoToGemmaThreeLnWikiVal / \WsFourCWikiFitGemmaTwoToGemmaThreeLnLsVal) & $\GemmaTwoDim\!\to\!\GemmaThreeDim$ & \WsFourCWikiFitGemmaTwoToGemmaThreeLnAnchors (\WsFourCWikiFitGemmaTwoToGemmaThreeLnTestN) & \WsFourCWikiFitGemmaTwoToGemmaThreeLnCos & \WsFourCWikiFitGemmaTwoToGemmaThreeLnPatOne\% & --- & \WsFourCWikiFitGemmaTwoToGemmaThreeLnRr \\
\midrule
\multicolumn{8}{@{}l}{\textit{Mixed bridges (\MixtureTextPct\% WikiText-103 + \MixtureAudioPct\% frozen speech soft prompts; WikiText / audio val NLL)}} \\
Qwen2.5-1.5B-IT $\to$ Gemma-2-2B-IT & Mixed (MS, LN; \WsFourCMixFitMoonshineQwenToGemmaTwoLnWikiVal / \WsFourCMixFitMoonshineQwenToGemmaTwoLnAudioVal) & $\QwenDim\!\to\!\GemmaTwoDim$ & \WsFourCMixFitMoonshineQwenToGemmaTwoLnAnchors (\WsFourCMixFitMoonshineQwenToGemmaTwoLnTestN) & \WsFourCMixFitMoonshineQwenToGemmaTwoLnCos & \WsFourCMixFitMoonshineQwenToGemmaTwoLnPatOne\% & --- & \WsFourCMixFitMoonshineQwenToGemmaTwoLnRr \\
Qwen2.5-1.5B-IT $\to$ Gemma-2-2B-IT & Mixed (Wh, LN; \WsFourCMixFitWhisperQwenToGemmaTwoLnWikiVal / \WsFourCMixFitWhisperQwenToGemmaTwoLnAudioVal) & $\QwenDim\!\to\!\GemmaTwoDim$ & \WsFourCMixFitWhisperQwenToGemmaTwoLnAnchors (\WsFourCMixFitWhisperQwenToGemmaTwoLnTestN) & \WsFourCMixFitWhisperQwenToGemmaTwoLnCos & \WsFourCMixFitWhisperQwenToGemmaTwoLnPatOne\% & --- & \WsFourCMixFitWhisperQwenToGemmaTwoLnRr \\
Gemma-2-2B-IT $\to$ Qwen2.5-1.5B-IT & Mixed (MS, LN; \WsFourCMixFitMoonshineGemmaTwoToQwenLnWikiVal / \WsFourCMixFitMoonshineGemmaTwoToQwenLnAudioVal) & $\GemmaTwoDim\!\to\!\QwenDim$ & \WsFourCMixFitMoonshineGemmaTwoToQwenLnAnchors (\WsFourCMixFitMoonshineGemmaTwoToQwenLnTestN) & \WsFourCMixFitMoonshineGemmaTwoToQwenLnCos & \WsFourCMixFitMoonshineGemmaTwoToQwenLnPatOne\% & --- & \WsFourCMixFitMoonshineGemmaTwoToQwenLnRr \\
Gemma-2-2B-IT $\to$ Qwen2.5-1.5B-IT & Mixed (Wh, LN; \WsFourCMixFitWhisperGemmaTwoToQwenLnWikiVal / \WsFourCMixFitWhisperGemmaTwoToQwenLnAudioVal) & $\GemmaTwoDim\!\to\!\QwenDim$ & \WsFourCMixFitWhisperGemmaTwoToQwenLnAnchors (\WsFourCMixFitWhisperGemmaTwoToQwenLnTestN) & \WsFourCMixFitWhisperGemmaTwoToQwenLnCos & \WsFourCMixFitWhisperGemmaTwoToQwenLnPatOne\% & --- & \WsFourCMixFitWhisperGemmaTwoToQwenLnRr \\
Qwen2.5-1.5B-IT $\to$ Gemma-3-4B-IT & Mixed (MS, LN; \WsFourCMixFitMoonshineQwenToGemmaThreeLnWikiVal / \WsFourCMixFitMoonshineQwenToGemmaThreeLnAudioVal) & $\QwenDim\!\to\!\GemmaThreeDim$ & \WsFourCMixFitMoonshineQwenToGemmaThreeLnAnchors (\WsFourCMixFitMoonshineQwenToGemmaThreeLnTestN) & \WsFourCMixFitMoonshineQwenToGemmaThreeLnCos & \WsFourCMixFitMoonshineQwenToGemmaThreeLnPatOne\% & --- & \WsFourCMixFitMoonshineQwenToGemmaThreeLnRr \\
Qwen2.5-1.5B-IT $\to$ Gemma-3-4B-IT & Mixed (Wh, LN; \WsFourCMixFitWhisperQwenToGemmaThreeLnWikiVal / \WsFourCMixFitWhisperQwenToGemmaThreeLnAudioVal) & $\QwenDim\!\to\!\GemmaThreeDim$ & \WsFourCMixFitWhisperQwenToGemmaThreeLnAnchors (\WsFourCMixFitWhisperQwenToGemmaThreeLnTestN) & \WsFourCMixFitWhisperQwenToGemmaThreeLnCos & \WsFourCMixFitWhisperQwenToGemmaThreeLnPatOne\% & --- & \WsFourCMixFitWhisperQwenToGemmaThreeLnRr \\
Gemma-2-2B-IT $\to$ Gemma-3-4B-IT & Mixed (MS, LN; \WsFourCMixFitMoonshineGemmaTwoToGemmaThreeLnWikiVal / \WsFourCMixFitMoonshineGemmaTwoToGemmaThreeLnAudioVal) & $\GemmaTwoDim\!\to\!\GemmaThreeDim$ & \WsFourCMixFitMoonshineGemmaTwoToGemmaThreeLnAnchors (\WsFourCMixFitMoonshineGemmaTwoToGemmaThreeLnTestN) & \WsFourCMixFitMoonshineGemmaTwoToGemmaThreeLnCos & \WsFourCMixFitMoonshineGemmaTwoToGemmaThreeLnPatOne\% & --- & \WsFourCMixFitMoonshineGemmaTwoToGemmaThreeLnRr \\
Gemma-2-2B-IT $\to$ Gemma-3-4B-IT & Mixed (Wh, LN; \WsFourCMixFitWhisperGemmaTwoToGemmaThreeLnWikiVal / \WsFourCMixFitWhisperGemmaTwoToGemmaThreeLnAudioVal) & $\GemmaTwoDim\!\to\!\GemmaThreeDim$ & \WsFourCMixFitWhisperGemmaTwoToGemmaThreeLnAnchors (\WsFourCMixFitWhisperGemmaTwoToGemmaThreeLnTestN) & \WsFourCMixFitWhisperGemmaTwoToGemmaThreeLnCos & \WsFourCMixFitWhisperGemmaTwoToGemmaThreeLnPatOne\% & --- & \WsFourCMixFitWhisperGemmaTwoToGemmaThreeLnRr \\
\bottomrule
\end{tabular}%
\end{adjustbox}
\end{table}
\end{landscape}}

%% file: tables/tab_ws3_comp.tex
\afterpage{\clearpage\begin{landscape}\thispagestyle{plain}
\begin{table}[p]
\caption{\textbf{Full speech adapter composition grid across ordered frozen LLM pairs (LibriSpeech WER $\downarrow$).} Columns 5--14 report \texttt{test-clean} / \texttt{test-other} WER; columns 3--4 and 15--17 report \texttt{test-clean} WER (both splits in Table~\ref{tab:ws2_asr} for columns 3--4 and in Table~\ref{tab:ws4c} for columns 15--17). All maps that never observe speech soft prompts (closed-form Procrustes, closed-form least squares, transcript-trained bridges, and WikiText-trained bridges) remain at $\ge \WsFourCWikiMinClean\%$ \texttt{test-clean} WER; random orthogonal controls on Moonshine Gemma-2 $\Rightarrow$ Qwen2.5 / Gemma-3 score \WsThreeCompMoonshineGemmaTwoToQwenRandScaledClean\% / \WsThreeCompMoonshineGemmaTwoToGemmaThreeRandScaledClean\%. Soft-prompt and mixed bridges reach \WsFourBCompBestClean\%--\WsFourBInDistMaxClean\% and \WsFourCMixCompBestClean\%--\WsFourCMixInDistMaxClean\% on the same adapter, whereas evaluating on the other encoder's adapter yields \CrossAdapterMinClean\%--\CrossAdapterMaxClean\%. In the columns ``Other'' and ``Mixed other'', the row's encoder is evaluated through the bridge fit on the other encoder's adapter.}
\label{tab:ws3_comp}
\vspace{4pt}
\centering
\begin{adjustbox}{max width=\linewidth, max totalheight=395pt}%
\setlength{\tabcolsep}{1.7pt}
\begin{tabular}{@{}llrrrrrrrrrrrrrrr@{}}
\toprule
& & \multicolumn{2}{c}{\shortstack[c]{\strut\textbf{Direct}\\\textbf{\texttt{clean}}}} & \multicolumn{3}{c}{\textbf{Closed-Form Procrustes}} & \multicolumn{3}{c}{\textbf{Closed-Form Least Squares}} & \multicolumn{2}{c}{\textbf{Transcript Bridge}} & \multicolumn{2}{c}{\textbf{Soft-Prompt Bridge}} & \multicolumn{3}{c}{\shortstack[c]{\strut\textbf{WikiText \&}\\\textbf{Mixed (\texttt{clean})}}} \\
\cmidrule(lr){3-4} \cmidrule(lr){5-7} \cmidrule(lr){8-10} \cmidrule(lr){11-12} \cmidrule(lr){13-14} \cmidrule(lr){15-17}
\textbf{Encoder} & \textbf{Port ($A \Rightarrow B$)} & $E\!\to\!A$ & $E\!\to\!B$ & None & Target-$r$ & LN & None & Target-$r$ & LN & Target-$r$ & LN & Same & Other & WikiText & \shortstack[r]{\strut Mixed\\same} & \shortstack[r]{\strut Mixed\\other} \\
\midrule
Whisper-Small & Qwen2.5-1.5B-IT $\Rightarrow$ Gemma-2-2B-IT & \WsTwoWhisperQwenLnMuonWerCleanCell & \WsTwoWhisperGemmaTwoLnMuonWerCleanCell & \WsThreeCompWhisperQwenToGemmaTwoProcNonePairCell & \WsThreeCompWhisperQwenToGemmaTwoProcScaledPairCell & \WsThreeCompWhisperQwenToGemmaTwoProcLnPairCell & \WsThreeCompWhisperQwenToGemmaTwoLstsqNonePairCell & \WsThreeCompWhisperQwenToGemmaTwoLstsqScaledPairCell & \WsThreeCompWhisperQwenToGemmaTwoLstsqLnPairCell & \WsFourCompWhisperQwenToGemmaTwoScaledPairCell & \WsFourCompWhisperQwenToGemmaTwoLnPairCell & \textbf{\WsFourBCompInDistWhisperQwenToGemmaTwoLnPairCell} & \WsFourBCompCrossWhisperOnMoonshineQwenToGemmaTwoLnPairCell & \WsFourCWikiCompWhisperQwenToGemmaTwoLnShortCleanCell & \textbf{\WsFourCMixCompInDistWhisperQwenToGemmaTwoLnShortCleanCell} & \WsFourCMixCompCrossWhisperOnMoonshineQwenToGemmaTwoLnShortCleanCell \\
Moonshine-Base & Qwen2.5-1.5B-IT $\Rightarrow$ Gemma-2-2B-IT & \WsTwoMoonshineQwenLnMuonWerCleanCell & \WsTwoMoonshineGemmaTwoLnMuonWerCleanCell & \WsThreeCompMoonshineQwenToGemmaTwoProcNonePairCell & \WsThreeCompMoonshineQwenToGemmaTwoProcScaledPairCell & \WsThreeCompMoonshineQwenToGemmaTwoProcLnPairCell & \WsThreeCompMoonshineQwenToGemmaTwoLstsqNonePairCell & \WsThreeCompMoonshineQwenToGemmaTwoLstsqScaledPairCell & \WsThreeCompMoonshineQwenToGemmaTwoLstsqLnPairCell & \WsFourCompMoonshineQwenToGemmaTwoScaledPairCell & \WsFourCompMoonshineQwenToGemmaTwoLnPairCell & \textbf{\WsFourBCompInDistMoonshineQwenToGemmaTwoLnPairCell} & \WsFourBCompCrossMoonshineOnWhisperQwenToGemmaTwoLnPairCell & \WsFourCWikiCompMoonshineQwenToGemmaTwoLnShortCleanCell & \textbf{\WsFourCMixCompInDistMoonshineQwenToGemmaTwoLnShortCleanCell} & \WsFourCMixCompCrossMoonshineOnWhisperQwenToGemmaTwoLnShortCleanCell \\
\cmidrule(lr){1-17}
Whisper-Small & Gemma-2-2B-IT $\Rightarrow$ Qwen2.5-1.5B-IT & \WsTwoWhisperGemmaTwoLnMuonWerCleanCell & \WsTwoWhisperQwenLnMuonWerCleanCell & \WsThreeCompWhisperGemmaTwoToQwenProcNonePairCell & \WsThreeCompWhisperGemmaTwoToQwenProcScaledPairCell & \WsThreeCompWhisperGemmaTwoToQwenProcLnPairCell & \WsThreeCompWhisperGemmaTwoToQwenLstsqNonePairCell & \WsThreeCompWhisperGemmaTwoToQwenLstsqScaledPairCell & \WsThreeCompWhisperGemmaTwoToQwenLstsqLnPairCell & \WsFourCompWhisperGemmaTwoToQwenScaledPairCell & \WsFourCompWhisperGemmaTwoToQwenLnPairCell & \textbf{\WsFourBCompInDistWhisperGemmaTwoToQwenLnPairCell} & \WsFourBCompCrossWhisperOnMoonshineGemmaTwoToQwenLnPairCell & \WsFourCWikiCompWhisperGemmaTwoToQwenLnShortCleanCell & \textbf{\WsFourCMixCompInDistWhisperGemmaTwoToQwenLnShortCleanCell} & \WsFourCMixCompCrossWhisperOnMoonshineGemmaTwoToQwenLnShortCleanCell \\
Moonshine-Base & Gemma-2-2B-IT $\Rightarrow$ Qwen2.5-1.5B-IT & \WsTwoMoonshineGemmaTwoLnMuonWerCleanCell & \WsTwoMoonshineQwenLnMuonWerCleanCell & \WsThreeCompMoonshineGemmaTwoToQwenProcNonePairCell & \WsThreeCompMoonshineGemmaTwoToQwenProcScaledPairCell & \WsThreeCompMoonshineGemmaTwoToQwenProcLnPairCell & \WsThreeCompMoonshineGemmaTwoToQwenLstsqNonePairCell & \WsThreeCompMoonshineGemmaTwoToQwenLstsqScaledPairCell & \WsThreeCompMoonshineGemmaTwoToQwenLstsqLnPairCell & \WsFourCompMoonshineGemmaTwoToQwenScaledPairCell & \WsFourCompMoonshineGemmaTwoToQwenLnPairCell & \textbf{\WsFourBCompInDistMoonshineGemmaTwoToQwenLnPairCell} & \WsFourBCompCrossMoonshineOnWhisperGemmaTwoToQwenLnPairCell & \WsFourCWikiCompMoonshineGemmaTwoToQwenLnShortCleanCell & \textbf{\WsFourCMixCompInDistMoonshineGemmaTwoToQwenLnShortCleanCell} & \WsFourCMixCompCrossMoonshineOnWhisperGemmaTwoToQwenLnShortCleanCell \\
\cmidrule(lr){1-17}
Whisper-Small & Qwen2.5-1.5B-IT $\Rightarrow$ Gemma-3-4B-IT & \WsTwoWhisperQwenLnMuonWerCleanCell & \WsTwoWhisperGemmaThreeLnMuonWerCleanCell & \WsThreeCompWhisperQwenToGemmaThreeProcNonePairCell & \WsThreeCompWhisperQwenToGemmaThreeProcScaledPairCell & \WsThreeCompWhisperQwenToGemmaThreeProcLnPairCell & \WsThreeCompWhisperQwenToGemmaThreeLstsqNonePairCell & \WsThreeCompWhisperQwenToGemmaThreeLstsqScaledPairCell & \WsThreeCompWhisperQwenToGemmaThreeLstsqLnPairCell & \WsFourCompWhisperQwenToGemmaThreeScaledPairCell & \WsFourCompWhisperQwenToGemmaThreeLnPairCell & \textbf{\WsFourBCompInDistWhisperQwenToGemmaThreeLnPairCell} & \WsFourBCompCrossWhisperOnMoonshineQwenToGemmaThreeLnPairCell & \WsFourCWikiCompWhisperQwenToGemmaThreeLnShortCleanCell & \textbf{\WsFourCMixCompInDistWhisperQwenToGemmaThreeLnShortCleanCell} & \WsFourCMixCompCrossWhisperOnMoonshineQwenToGemmaThreeLnShortCleanCell \\
Moonshine-Base & Qwen2.5-1.5B-IT $\Rightarrow$ Gemma-3-4B-IT & \WsTwoMoonshineQwenLnMuonWerCleanCell & \WsTwoMoonshineGemmaThreeLnMuonWerCleanCell & \WsThreeCompMoonshineQwenToGemmaThreeProcNonePairCell & \WsThreeCompMoonshineQwenToGemmaThreeProcScaledPairCell & \WsThreeCompMoonshineQwenToGemmaThreeProcLnPairCell & \WsThreeCompMoonshineQwenToGemmaThreeLstsqNonePairCell & \WsThreeCompMoonshineQwenToGemmaThreeLstsqScaledPairCell & \WsThreeCompMoonshineQwenToGemmaThreeLstsqLnPairCell & \WsFourCompMoonshineQwenToGemmaThreeScaledPairCell & \WsFourCompMoonshineQwenToGemmaThreeLnPairCell & \textbf{\WsFourBCompInDistMoonshineQwenToGemmaThreeLnPairCell} & \WsFourBCompCrossMoonshineOnWhisperQwenToGemmaThreeLnPairCell & \WsFourCWikiCompMoonshineQwenToGemmaThreeLnShortCleanCell & \textbf{\WsFourCMixCompInDistMoonshineQwenToGemmaThreeLnShortCleanCell} & \WsFourCMixCompCrossMoonshineOnWhisperQwenToGemmaThreeLnShortCleanCell \\
\cmidrule(lr){1-17}
Whisper-Small & Gemma-2-2B-IT $\Rightarrow$ Gemma-3-4B-IT & \WsTwoWhisperGemmaTwoLnMuonWerCleanCell & \WsTwoWhisperGemmaThreeLnMuonWerCleanCell & \WsThreeCompWhisperGemmaTwoToGemmaThreeProcNonePairCell & \WsThreeCompWhisperGemmaTwoToGemmaThreeProcScaledPairCell & \WsThreeCompWhisperGemmaTwoToGemmaThreeProcLnPairCell & \WsThreeCompWhisperGemmaTwoToGemmaThreeLstsqNonePairCell & \WsThreeCompWhisperGemmaTwoToGemmaThreeLstsqScaledPairCell & \WsThreeCompWhisperGemmaTwoToGemmaThreeLstsqLnPairCell & \WsFourCompWhisperGemmaTwoToGemmaThreeScaledPairCell & \WsFourCompWhisperGemmaTwoToGemmaThreeLnPairCell & \textbf{\WsFourBCompInDistWhisperGemmaTwoToGemmaThreeLnPairCell} & \WsFourBCompCrossWhisperOnMoonshineGemmaTwoToGemmaThreeLnPairCell & \WsFourCWikiCompWhisperGemmaTwoToGemmaThreeLnShortCleanCell & \textbf{\WsFourCMixCompInDistWhisperGemmaTwoToGemmaThreeLnShortCleanCell} & \WsFourCMixCompCrossWhisperOnMoonshineGemmaTwoToGemmaThreeLnShortCleanCell \\
Moonshine-Base & Gemma-2-2B-IT $\Rightarrow$ Gemma-3-4B-IT & \WsTwoMoonshineGemmaTwoLnMuonWerCleanCell & \WsTwoMoonshineGemmaThreeLnMuonWerCleanCell & \WsThreeCompMoonshineGemmaTwoToGemmaThreeProcNonePairCell & \WsThreeCompMoonshineGemmaTwoToGemmaThreeProcScaledPairCell & \WsThreeCompMoonshineGemmaTwoToGemmaThreeProcLnPairCell & \WsThreeCompMoonshineGemmaTwoToGemmaThreeLstsqNonePairCell & \WsThreeCompMoonshineGemmaTwoToGemmaThreeLstsqScaledPairCell & \WsThreeCompMoonshineGemmaTwoToGemmaThreeLstsqLnPairCell & \WsFourCompMoonshineGemmaTwoToGemmaThreeScaledPairCell & \WsFourCompMoonshineGemmaTwoToGemmaThreeLnPairCell & \textbf{\WsFourBCompInDistMoonshineGemmaTwoToGemmaThreeLnPairCell} & \WsFourBCompCrossMoonshineOnWhisperGemmaTwoToGemmaThreeLnPairCell & \WsFourCWikiCompMoonshineGemmaTwoToGemmaThreeLnShortCleanCell & \textbf{\WsFourCMixCompInDistMoonshineGemmaTwoToGemmaThreeLnShortCleanCell} & \WsFourCMixCompCrossMoonshineOnWhisperGemmaTwoToGemmaThreeLnShortCleanCell \\
\bottomrule
\end{tabular}%
\end{adjustbox}
\end{table}
\end{landscape}}

%% file: tables/tab_ws4b.tex
\afterpage{\clearpage\begin{landscape}\thispagestyle{plain}
\begin{table}[p]
\caption{\textbf{Soft-prompt-trained linear bridges and isotropic Gaussian input-noise controls (LibriSpeech WER $\downarrow$, validation NLL $\downarrow$, held-out lexical $\cos$/P@1 $\uparrow$).} Top: bridges trained for one pass (\LibriTrainUtts\ utterances, Muon) on frozen speech adapter outputs reach \WsFourBCompBestClean\% / \WsFourBCompBestOther\% to \WsFourBInDistMaxClean\% / \WsFourBInDistMaxOther\% WER on the same adapter while diverging on discrete text (text NLL \WsFourBMinTextVal--\WsFourBMaxTextVal). Middle: evaluating the same soft-prompt bridges on the other speech encoder's adapter yields \WsFourBCrossMinClean\%--\WsFourBCrossMaxClean\% \texttt{test-clean} WER ($^\dagger$\WsFourBCompCrossWhisperOnMoonshineGemmaTwoToGemmaThreeLnEmptyClean\ empty decodes, median utterance WER \XprofileXtwoCleanUttMedWer\%; $^\ddagger$\WsFourBCompCrossMoonshineOnWhisperGemmaTwoToGemmaThreeLnEmptyClean\ empty decodes, median utterance WER \XprofileXoneCleanUttMedWer\%). In the middle block, the audio NLL, text NLL and anchor $\cos$/P@1 columns repeat the values of the top block: they describe the bridge, with audio NLL measured on the adapter it was fit on, not on the evaluated adapter. Bottom: isotropic Gaussian noise of expected norm $k\,\bar r_A$ added to the bridge's inputs during training ($k \in \{\WsFourBNoiseKHalf, \WsFourBNoiseKOne, \WsFourBNoiseKSevenSevenThree\}$) yields \WsFourBNoiseMinClean\%--\WsFourBNoiseMaxClean\% WER.}
\label{tab:ws4b}
\vspace{4pt}
\centering
\begin{adjustbox}{max width=\linewidth, max totalheight=395pt}%
\setlength{\tabcolsep}{3.0pt}
\begin{tabular}{@{}lllrrrrr@{}}
\toprule
\textbf{Eval Adapter (Bridge Fit)} & \textbf{Port ($A \Rightarrow B$)} & \textbf{Norm} & \shortstack[r]{\strut\textbf{Audio}\\\textbf{NLL $\downarrow$}} & \shortstack[r]{\strut\textbf{Text}\\\textbf{NLL $\downarrow$}} & \shortstack[r]{\strut\textbf{Anchor}\\\textbf{$\cos$ / P@1 $\uparrow$}} & \shortstack[r]{\strut\textbf{Direct}\\\textbf{$E\!\to\!A$ / $B$}} & \shortstack[r]{\strut\textbf{Composed}\\\textbf{(\texttt{clean} / \texttt{other}) $\downarrow$}} \\
\midrule
\multicolumn{8}{@{}l}{\textit{Same-adapter evaluation (bridge trained and evaluated with the same frozen speech adapter into LLM $A$)}} \\
Moonshine-Base (Moonshine-fit) & Qwen2.5-1.5B-IT $\Rightarrow$ Gemma-2-2B-IT & Layer norm & \WsFourBFitMoonshineQwenToGemmaTwoLnAudioVal & \WsFourBFitMoonshineQwenToGemmaTwoLnTextVal & \WsFourBFitMoonshineQwenToGemmaTwoLnCos / \WsFourBFitMoonshineQwenToGemmaTwoLnPatOne\% & \WsTwoMoonshineQwenLnMuonWerCleanCell\ / \WsTwoMoonshineGemmaTwoLnMuonWerCleanCell & \textbf{\WsFourBCompInDistMoonshineQwenToGemmaTwoLnCleanCell / \WsFourBCompInDistMoonshineQwenToGemmaTwoLnOtherCell} \\
Whisper-Small (Whisper-fit) & Qwen2.5-1.5B-IT $\Rightarrow$ Gemma-2-2B-IT & Layer norm & \WsFourBFitWhisperQwenToGemmaTwoLnAudioVal & \WsFourBFitWhisperQwenToGemmaTwoLnTextVal & \WsFourBFitWhisperQwenToGemmaTwoLnCos / \WsFourBFitWhisperQwenToGemmaTwoLnPatOne\% & \WsTwoWhisperQwenLnMuonWerCleanCell\ / \WsTwoWhisperGemmaTwoLnMuonWerCleanCell & \textbf{\WsFourBCompInDistWhisperQwenToGemmaTwoLnCleanCell / \WsFourBCompInDistWhisperQwenToGemmaTwoLnOtherCell} \\
Moonshine-Base (Moonshine-fit) & Gemma-2-2B-IT $\Rightarrow$ Qwen2.5-1.5B-IT & Layer norm & \WsFourBFitMoonshineGemmaTwoToQwenLnAudioVal & \WsFourBFitMoonshineGemmaTwoToQwenLnTextVal & \WsFourBFitMoonshineGemmaTwoToQwenLnCos / \WsFourBFitMoonshineGemmaTwoToQwenLnPatOne\% & \WsTwoMoonshineGemmaTwoLnMuonWerCleanCell\ / \WsTwoMoonshineQwenLnMuonWerCleanCell & \textbf{\WsFourBCompInDistMoonshineGemmaTwoToQwenLnCleanCell / \WsFourBCompInDistMoonshineGemmaTwoToQwenLnOtherCell} \\
Moonshine-Base (Moonshine-fit) & Gemma-2-2B-IT $\Rightarrow$ Qwen2.5-1.5B-IT & Target-radius & \WsFourBFitMoonshineGemmaTwoToQwenScaledAudioVal & \WsFourBFitMoonshineGemmaTwoToQwenScaledTextVal & \WsFourBFitMoonshineGemmaTwoToQwenScaledCos / \WsFourBFitMoonshineGemmaTwoToQwenScaledPatOne\% & \WsTwoMoonshineGemmaTwoLnMuonWerCleanCell\ / \WsTwoMoonshineQwenLnMuonWerCleanCell & \textbf{\WsFourBCompInDistMoonshineGemmaTwoToQwenScaledCleanCell / \WsFourBCompInDistMoonshineGemmaTwoToQwenScaledOtherCell} \\
Whisper-Small (Whisper-fit) & Gemma-2-2B-IT $\Rightarrow$ Qwen2.5-1.5B-IT & Layer norm & \WsFourBFitWhisperGemmaTwoToQwenLnAudioVal & \WsFourBFitWhisperGemmaTwoToQwenLnTextVal & \WsFourBFitWhisperGemmaTwoToQwenLnCos / \WsFourBFitWhisperGemmaTwoToQwenLnPatOne\% & \WsTwoWhisperGemmaTwoLnMuonWerCleanCell\ / \WsTwoWhisperQwenLnMuonWerCleanCell & \textbf{\WsFourBCompInDistWhisperGemmaTwoToQwenLnCleanCell / \WsFourBCompInDistWhisperGemmaTwoToQwenLnOtherCell} \\
Moonshine-Base (Moonshine-fit) & Qwen2.5-1.5B-IT $\Rightarrow$ Gemma-3-4B-IT & Layer norm & \WsFourBFitMoonshineQwenToGemmaThreeLnAudioVal & \WsFourBFitMoonshineQwenToGemmaThreeLnTextVal & \WsFourBFitMoonshineQwenToGemmaThreeLnCos / \WsFourBFitMoonshineQwenToGemmaThreeLnPatOne\% & \WsTwoMoonshineQwenLnMuonWerCleanCell\ / \WsTwoMoonshineGemmaThreeLnMuonWerCleanCell & \textbf{\WsFourBCompInDistMoonshineQwenToGemmaThreeLnCleanCell / \WsFourBCompInDistMoonshineQwenToGemmaThreeLnOtherCell} \\
Whisper-Small (Whisper-fit) & Qwen2.5-1.5B-IT $\Rightarrow$ Gemma-3-4B-IT & Layer norm & \WsFourBFitWhisperQwenToGemmaThreeLnAudioVal & \WsFourBFitWhisperQwenToGemmaThreeLnTextVal & \WsFourBFitWhisperQwenToGemmaThreeLnCos / \WsFourBFitWhisperQwenToGemmaThreeLnPatOne\% & \WsTwoWhisperQwenLnMuonWerCleanCell\ / \WsTwoWhisperGemmaThreeLnMuonWerCleanCell & \textbf{\WsFourBCompInDistWhisperQwenToGemmaThreeLnCleanCell / \WsFourBCompInDistWhisperQwenToGemmaThreeLnOtherCell} \\
Moonshine-Base (Moonshine-fit) & Gemma-2-2B-IT $\Rightarrow$ Gemma-3-4B-IT & Layer norm & \WsFourBFitMoonshineGemmaTwoToGemmaThreeLnAudioVal & \WsFourBFitMoonshineGemmaTwoToGemmaThreeLnTextVal & \WsFourBFitMoonshineGemmaTwoToGemmaThreeLnCos / \WsFourBFitMoonshineGemmaTwoToGemmaThreeLnPatOne\% & \WsTwoMoonshineGemmaTwoLnMuonWerCleanCell\ / \WsTwoMoonshineGemmaThreeLnMuonWerCleanCell & \textbf{\WsFourBCompInDistMoonshineGemmaTwoToGemmaThreeLnCleanCell / \WsFourBCompInDistMoonshineGemmaTwoToGemmaThreeLnOtherCell} \\
Whisper-Small (Whisper-fit) & Gemma-2-2B-IT $\Rightarrow$ Gemma-3-4B-IT & Layer norm & \WsFourBFitWhisperGemmaTwoToGemmaThreeLnAudioVal & \WsFourBFitWhisperGemmaTwoToGemmaThreeLnTextVal & \WsFourBFitWhisperGemmaTwoToGemmaThreeLnCos / \WsFourBFitWhisperGemmaTwoToGemmaThreeLnPatOne\% & \WsTwoWhisperGemmaTwoLnMuonWerCleanCell\ / \WsTwoWhisperGemmaThreeLnMuonWerCleanCell & \textbf{\WsFourBCompInDistWhisperGemmaTwoToGemmaThreeLnCleanCell / \WsFourBCompInDistWhisperGemmaTwoToGemmaThreeLnOtherCell} \\
\midrule
\multicolumn{8}{@{}l}{\textit{Other-adapter evaluation (identical soft-prompt bridges evaluated on the unseen speech encoder's adapter)}} \\
Whisper-Small (Moonshine-fit) & Qwen2.5-1.5B-IT $\Rightarrow$ Gemma-2-2B-IT & Layer norm & \WsFourBFitMoonshineQwenToGemmaTwoLnAudioVal & \WsFourBFitMoonshineQwenToGemmaTwoLnTextVal & \WsFourBFitMoonshineQwenToGemmaTwoLnCos / \WsFourBFitMoonshineQwenToGemmaTwoLnPatOne\% & \WsTwoWhisperQwenLnMuonWerCleanCell\ / \WsTwoWhisperGemmaTwoLnMuonWerCleanCell & \WsFourBCompCrossWhisperOnMoonshineQwenToGemmaTwoLnCleanCell / \WsFourBCompCrossWhisperOnMoonshineQwenToGemmaTwoLnOtherCell \\
Moonshine-Base (Whisper-fit) & Qwen2.5-1.5B-IT $\Rightarrow$ Gemma-2-2B-IT & Layer norm & \WsFourBFitWhisperQwenToGemmaTwoLnAudioVal & \WsFourBFitWhisperQwenToGemmaTwoLnTextVal & \WsFourBFitWhisperQwenToGemmaTwoLnCos / \WsFourBFitWhisperQwenToGemmaTwoLnPatOne\% & \WsTwoMoonshineQwenLnMuonWerCleanCell\ / \WsTwoMoonshineGemmaTwoLnMuonWerCleanCell & \WsFourBCompCrossMoonshineOnWhisperQwenToGemmaTwoLnCleanCell / \WsFourBCompCrossMoonshineOnWhisperQwenToGemmaTwoLnOtherCell \\
Whisper-Small (Moonshine-fit) & Gemma-2-2B-IT $\Rightarrow$ Qwen2.5-1.5B-IT & Layer norm & \WsFourBFitMoonshineGemmaTwoToQwenLnAudioVal & \WsFourBFitMoonshineGemmaTwoToQwenLnTextVal & \WsFourBFitMoonshineGemmaTwoToQwenLnCos / \WsFourBFitMoonshineGemmaTwoToQwenLnPatOne\% & \WsTwoWhisperGemmaTwoLnMuonWerCleanCell\ / \WsTwoWhisperQwenLnMuonWerCleanCell & \WsFourBCompCrossWhisperOnMoonshineGemmaTwoToQwenLnCleanCell / \WsFourBCompCrossWhisperOnMoonshineGemmaTwoToQwenLnOtherCell \\
Moonshine-Base (Whisper-fit) & Gemma-2-2B-IT $\Rightarrow$ Qwen2.5-1.5B-IT & Layer norm & \WsFourBFitWhisperGemmaTwoToQwenLnAudioVal & \WsFourBFitWhisperGemmaTwoToQwenLnTextVal & \WsFourBFitWhisperGemmaTwoToQwenLnCos / \WsFourBFitWhisperGemmaTwoToQwenLnPatOne\% & \WsTwoMoonshineGemmaTwoLnMuonWerCleanCell\ / \WsTwoMoonshineQwenLnMuonWerCleanCell & \WsFourBCompCrossMoonshineOnWhisperGemmaTwoToQwenLnCleanCell / \WsFourBCompCrossMoonshineOnWhisperGemmaTwoToQwenLnOtherCell \\
Whisper-Small (Moonshine-fit) & Qwen2.5-1.5B-IT $\Rightarrow$ Gemma-3-4B-IT & Layer norm & \WsFourBFitMoonshineQwenToGemmaThreeLnAudioVal & \WsFourBFitMoonshineQwenToGemmaThreeLnTextVal & \WsFourBFitMoonshineQwenToGemmaThreeLnCos / \WsFourBFitMoonshineQwenToGemmaThreeLnPatOne\% & \WsTwoWhisperQwenLnMuonWerCleanCell\ / \WsTwoWhisperGemmaThreeLnMuonWerCleanCell & \WsFourBCompCrossWhisperOnMoonshineQwenToGemmaThreeLnCleanCell / \WsFourBCompCrossWhisperOnMoonshineQwenToGemmaThreeLnOtherCell \\
Moonshine-Base (Whisper-fit) & Qwen2.5-1.5B-IT $\Rightarrow$ Gemma-3-4B-IT & Layer norm & \WsFourBFitWhisperQwenToGemmaThreeLnAudioVal & \WsFourBFitWhisperQwenToGemmaThreeLnTextVal & \WsFourBFitWhisperQwenToGemmaThreeLnCos / \WsFourBFitWhisperQwenToGemmaThreeLnPatOne\% & \WsTwoMoonshineQwenLnMuonWerCleanCell\ / \WsTwoMoonshineGemmaThreeLnMuonWerCleanCell & \WsFourBCompCrossMoonshineOnWhisperQwenToGemmaThreeLnCleanCell / \WsFourBCompCrossMoonshineOnWhisperQwenToGemmaThreeLnOtherCell \\
Whisper-Small (Moonshine-fit)$^\dagger$ & Gemma-2-2B-IT $\Rightarrow$ Gemma-3-4B-IT & Layer norm & \WsFourBFitMoonshineGemmaTwoToGemmaThreeLnAudioVal & \WsFourBFitMoonshineGemmaTwoToGemmaThreeLnTextVal & \WsFourBFitMoonshineGemmaTwoToGemmaThreeLnCos / \WsFourBFitMoonshineGemmaTwoToGemmaThreeLnPatOne\% & \WsTwoWhisperGemmaTwoLnMuonWerCleanCell\ / \WsTwoWhisperGemmaThreeLnMuonWerCleanCell & \WsFourBCompCrossWhisperOnMoonshineGemmaTwoToGemmaThreeLnCleanCell / \WsFourBCompCrossWhisperOnMoonshineGemmaTwoToGemmaThreeLnOtherCell \\
Moonshine-Base (Whisper-fit)$^\ddagger$ & Gemma-2-2B-IT $\Rightarrow$ Gemma-3-4B-IT & Layer norm & \WsFourBFitWhisperGemmaTwoToGemmaThreeLnAudioVal & \WsFourBFitWhisperGemmaTwoToGemmaThreeLnTextVal & \WsFourBFitWhisperGemmaTwoToGemmaThreeLnCos / \WsFourBFitWhisperGemmaTwoToGemmaThreeLnPatOne\% & \WsTwoMoonshineGemmaTwoLnMuonWerCleanCell\ / \WsTwoMoonshineGemmaThreeLnMuonWerCleanCell & \WsFourBCompCrossMoonshineOnWhisperGemmaTwoToGemmaThreeLnCleanCell / \WsFourBCompCrossMoonshineOnWhisperGemmaTwoToGemmaThreeLnOtherCell \\
\midrule
\multicolumn{8}{@{}l}{\textit{Gaussian input-noise controls on discrete transcript bridge training (Moonshine-Base, Gemma-2-2B-IT $\Rightarrow$ Qwen2.5-1.5B-IT)}} \\
Transcript bridge (noise $k\!=\!\WsFourBNoiseKZero$) & Gemma-2-2B-IT $\Rightarrow$ Qwen2.5-1.5B-IT & Layer norm & --- & \WsFourFitGemmaTwoToQwenLnVal & \WsFourFitGemmaTwoToQwenLnCos / \WsFourFitGemmaTwoToQwenLnPatOne\% & \WsTwoMoonshineGemmaTwoLnMuonWerCleanCell\ / \WsTwoMoonshineQwenLnMuonWerCleanCell & \WsFourCompMoonshineGemmaTwoToQwenLnCleanCell / \WsFourCompMoonshineGemmaTwoToQwenLnOtherCell \\
Transcript + noise ($k\!=\!\WsFourBNoiseKHalf$) & Gemma-2-2B-IT $\Rightarrow$ Qwen2.5-1.5B-IT & Layer norm & --- & \WsFourBNoiseHalfVal & \WsFourBNoiseHalfCos / \WsFourBNoiseHalfPatOne\% & \WsTwoMoonshineGemmaTwoLnMuonWerCleanCell\ / \WsTwoMoonshineQwenLnMuonWerCleanCell & \WsFourBNoiseHalfCleanCell / \WsFourBNoiseHalfOtherCell \\
Transcript + noise ($k\!=\!\WsFourBNoiseKOne$) & Gemma-2-2B-IT $\Rightarrow$ Qwen2.5-1.5B-IT & Layer norm & --- & \WsFourBNoiseOneVal & \WsFourBNoiseOneCos / \WsFourBNoiseOnePatOne\% & \WsTwoMoonshineGemmaTwoLnMuonWerCleanCell\ / \WsTwoMoonshineQwenLnMuonWerCleanCell & \WsFourBNoiseOneCleanCell / \WsFourBNoiseOneOtherCell \\
Transcript + noise ($k\!=\!\WsFourBNoiseKSevenSevenThree$) & Gemma-2-2B-IT $\Rightarrow$ Qwen2.5-1.5B-IT & Layer norm & --- & \WsFourBNoiseSevenSevenThreeVal & \WsFourBNoiseSevenSevenThreeCos / \WsFourBNoiseSevenSevenThreePatOne\% & \WsTwoMoonshineGemmaTwoLnMuonWerCleanCell\ / \WsTwoMoonshineQwenLnMuonWerCleanCell & \WsFourBNoiseSevenSevenThreeCleanCell / \WsFourBNoiseSevenSevenThreeOtherCell \\
\bottomrule
\end{tabular}%
\end{adjustbox}
\end{table}
\end{landscape}}

%% file: tables/tab_ws4e.tex
\afterpage{\clearpage\begin{landscape}\thispagestyle{plain}
\begin{table}[p]
\caption{\textbf{Two-adapter linear bridge and held-out same-encoder adapter controls from Gemma-2-2B-IT ($\GemmaTwoDim$) to frozen Gemma-3-4B-IT ($\GemmaThreeDim$) on LibriSpeech.} Columns report corpus-aggregate word error rate (WER $\downarrow$), empty-hypothesis count ($\downarrow$), and per-utterance median WER ($\downarrow$) on \texttt{test-clean} (\LibriTestCleanUtts\ utterances) and \texttt{test-other} (\LibriTestOtherUtts\ utterances). Each single-adapter bridge is trained for \WsFourEFitSingleSteps\ Muon steps (\WsTwoBatchSize\ utterances per step, best validation checkpoint at step \WsFourEFitSingleBestStep); the two-adapter bridge is trained for \WsFourEFitJointTotalSteps\ Muon steps (\WsFourEFitJointSamplesPerAdapterStep\ Whisper + \WsFourEFitJointSamplesPerAdapterStep\ Moonshine = \WsFourEFitJointSamplesPerStep\ utterances per step; best validation checkpoint at step \WsFourEFitJointBestStep, midpoint checkpoint at step \WsFourEFitJointHalfStep).}
\label{tab:ws4e}
\vspace{4pt}
\centering
\begin{adjustbox}{max width=\linewidth, max totalheight=395pt}%
\setlength{\tabcolsep}{4.8pt}
\begin{tabular}{@{}llrrr@{}}
\toprule
\textbf{Gemma-2-2B-IT Speech Adapter} & \textbf{Gemma-2 $\to$ Gemma-3 Bridge} & \shortstack[r]{\strut\textbf{WER}\\\textbf{(\texttt{clean} / \texttt{other}) $\downarrow$}} & \shortstack[r]{\strut\textbf{Empty Decodes}\\\textbf{(\texttt{clean} / \texttt{other}) $\downarrow$}} & \shortstack[r]{\strut\textbf{Median Utt.\ WER}\\\textbf{(\texttt{clean} / \texttt{other}) $\downarrow$}} \\
\midrule
\multicolumn{5}{@{}l}{\textit{Direct speech adapters on frozen Gemma-2-2B-IT (no bridge; layer norm)}} \\
Whisper-Small (Muon, fitted) & --- & \WsTwoWhisperGemmaTwoLnMuonWerClean\% / \WsTwoWhisperGemmaTwoLnMuonWerOther\% & \WsTwoWhisperGemmaTwoLnMuonEmptyClean\ / \WsTwoWhisperGemmaTwoLnMuonEmptyOther & \WsFourEDirectWhisperMuonMedClean\% / \WsFourEDirectWhisperMuonMedOther\% \\
Whisper-Small (AdamW, held-out) & --- & \WsTwoWhisperGemmaTwoLnAdamwWerClean\% / \WsTwoWhisperGemmaTwoLnAdamwWerOther\% & \WsTwoWhisperGemmaTwoLnAdamwEmptyClean\ / \WsTwoWhisperGemmaTwoLnAdamwEmptyOther & \WsFourEDirectWhisperAdamwMedClean\% / \WsFourEDirectWhisperAdamwMedOther\% \\
Moonshine-Base (Muon, fitted) & --- & \WsTwoMoonshineGemmaTwoLnMuonWerClean\% / \WsTwoMoonshineGemmaTwoLnMuonWerOther\% & \WsTwoMoonshineGemmaTwoLnMuonEmptyClean\ / \WsTwoMoonshineGemmaTwoLnMuonEmptyOther & \WsFourEDirectMoonshineMuonMedClean\% / \WsFourEDirectMoonshineMuonMedOther\% \\
Moonshine-Base (AdamW, held-out) & --- & \WsTwoMoonshineGemmaTwoLnAdamwWerClean\% / \WsTwoMoonshineGemmaTwoLnAdamwWerOther\% & \WsTwoMoonshineGemmaTwoLnAdamwEmptyClean\ / \WsTwoMoonshineGemmaTwoLnAdamwEmptyOther & \WsFourEDirectMoonshineAdamwMedClean\% / \WsFourEDirectMoonshineAdamwMedOther\% \\
\midrule
\multicolumn{5}{@{}l}{\textit{Single-adapter bridges ($\GemmaTwoDim \to \GemmaThreeDim$, layer norm) with the fitted Muon adapter}} \\
Whisper-Small (Muon, fitted) & Single-adapter (Whisper-fit) & \textbf{\WsFourBCompInDistWhisperGemmaTwoToGemmaThreeLnClean\% / \WsFourBCompInDistWhisperGemmaTwoToGemmaThreeLnOther\%} & \WsFourBCompInDistWhisperGemmaTwoToGemmaThreeLnEmptyClean\ / \WsFourBCompInDistWhisperGemmaTwoToGemmaThreeLnEmptyOther & \WsFourESingleSameWhisperMuonMedClean\% / \WsFourESingleSameWhisperMuonMedOther\% \\
Moonshine-Base (Muon, fitted) & Single-adapter (Moonshine-fit) & \textbf{\WsFourBCompInDistMoonshineGemmaTwoToGemmaThreeLnClean\% / \WsFourBCompInDistMoonshineGemmaTwoToGemmaThreeLnOther\%} & \WsFourBCompInDistMoonshineGemmaTwoToGemmaThreeLnEmptyClean\ / \WsFourBCompInDistMoonshineGemmaTwoToGemmaThreeLnEmptyOther & \WsFourESingleSameMoonshineMuonMedClean\% / \WsFourESingleSameMoonshineMuonMedOther\% \\
\midrule
\multicolumn{5}{@{}l}{\textit{Single-adapter bridges evaluated with the held-out AdamW adapter of the same encoder}} \\
Whisper-Small (AdamW, held-out) & Single-adapter (Whisper-fit) & \WsFourESingleCtrlWhisperAdamwClean\% / \WsFourESingleCtrlWhisperAdamwOther\% & \WsFourESingleCtrlWhisperAdamwEmptyClean\ / \WsFourESingleCtrlWhisperAdamwEmptyOther & \WsFourESingleCtrlWhisperAdamwMedClean\% / \WsFourESingleCtrlWhisperAdamwMedOther\% \\
Moonshine-Base (AdamW, held-out) & Single-adapter (Moonshine-fit) & \WsFourESingleCtrlMoonshineAdamwClean\% / \WsFourESingleCtrlMoonshineAdamwOther\% & \WsFourESingleCtrlMoonshineAdamwEmptyClean\ / \WsFourESingleCtrlMoonshineAdamwEmptyOther & \WsFourESingleCtrlMoonshineAdamwMedClean\% / \WsFourESingleCtrlMoonshineAdamwMedOther\% \\
\midrule
\multicolumn{5}{@{}l}{\textit{Two-adapter bridge ($\GemmaTwoDim \to \GemmaThreeDim$, layer norm) with each fitted Muon adapter and its midpoint checkpoint}} \\
Whisper-Small (Muon, fitted) & Two-adapter (Whisper + Moonshine) & \textbf{\WsFourEJointWhisperMuonClean\% / \WsFourEJointWhisperMuonOther\%} & \WsFourEJointWhisperMuonEmptyClean\ / \WsFourEJointWhisperMuonEmptyOther & \WsFourEJointWhisperMuonMedClean\% / \WsFourEJointWhisperMuonMedOther\% \\
Moonshine-Base (Muon, fitted) & Two-adapter (Whisper + Moonshine) & \textbf{\WsFourEJointMoonshineMuonClean\% / \WsFourEJointMoonshineMuonOther\%} & \WsFourEJointMoonshineMuonEmptyClean\ / \WsFourEJointMoonshineMuonEmptyOther & \WsFourEJointMoonshineMuonMedClean\% / \WsFourEJointMoonshineMuonMedOther\% \\
Whisper-Small (Muon, fitted) & Two-adapter midpoint (step \WsFourEFitJointHalfStep) & \WsFourEJointHalfWhisperMuonClean\% / \WsFourEJointHalfWhisperMuonOther\% & \WsFourEJointHalfWhisperMuonEmptyClean\ / \WsFourEJointHalfWhisperMuonEmptyOther & \WsFourEJointHalfWhisperMuonMedClean\% / \WsFourEJointHalfWhisperMuonMedOther\% \\
Moonshine-Base (Muon, fitted) & Two-adapter midpoint (step \WsFourEFitJointHalfStep) & \WsFourEJointHalfMoonshineMuonClean\% / \WsFourEJointHalfMoonshineMuonOther\% & \WsFourEJointHalfMoonshineMuonEmptyClean\ / \WsFourEJointHalfMoonshineMuonEmptyOther & \WsFourEJointHalfMoonshineMuonMedClean\% / \WsFourEJointHalfMoonshineMuonMedOther\% \\
\midrule
\multicolumn{5}{@{}l}{\textit{Two-adapter bridge evaluated with each held-out AdamW adapter}} \\
Whisper-Small (AdamW, held-out) & Two-adapter (Whisper + Moonshine) & \WsFourEJointCtrlWhisperAdamwClean\% / \WsFourEJointCtrlWhisperAdamwOther\% & \WsFourEJointCtrlWhisperAdamwEmptyClean\ / \WsFourEJointCtrlWhisperAdamwEmptyOther & \WsFourEJointCtrlWhisperAdamwMedClean\% / \WsFourEJointCtrlWhisperAdamwMedOther\% \\
Moonshine-Base (AdamW, held-out) & Two-adapter (Whisper + Moonshine) & \WsFourEJointCtrlMoonshineAdamwClean\% / \WsFourEJointCtrlMoonshineAdamwOther\% & \WsFourEJointCtrlMoonshineAdamwEmptyClean\ / \WsFourEJointCtrlMoonshineAdamwEmptyOther & \WsFourEJointCtrlMoonshineAdamwMedClean\% / \WsFourEJointCtrlMoonshineAdamwMedOther\% \\
\bottomrule
\end{tabular}%
\end{adjustbox}
\end{table}
\end{landscape}}

%% file: tables/tab_ws4c.tex
\afterpage{\clearpage\begin{landscape}\thispagestyle{plain}
\begin{table}[p]
\caption{\textbf{Natural-text (WikiText-103) bridges and mixed (\MixtureTextPct\% text, \MixtureAudioPct\% soft-prompt) bridges (layer norm, Muon).} Top: training on \LibriTrainUtts\ WikiText-103 chunks transfers to LibriSpeech text (NLL $\downarrow$ \WsFourCWikiMinLsVal--\WsFourCWikiMaxLsVal, P@1 $\uparrow$ \WsFourCWikiMinPatOne\%--\WsFourCWikiMaxPatOne\%) yet fails on all \WsFourCWikiCompDone/\WsFourCWikiCompTotal\ speech compositions (\WsFourCWikiMinClean\%--\WsFourCWikiMaxClean\% \texttt{test-clean} WER $\downarrow$). Bottom: co-training a single linear bridge on half WikiText-103 text and half speech soft prompts preserves low text NLL (\WsFourCMixMinWikiVal--\WsFourCMixMaxWikiVal) while tying soft-prompt-only bridges within $\le \MixVsAudioTiedMaxAbsDeltaClean$\,pp on \MixVsAudioSameTriplesTied\ of \MixVsAudioSameTriplesTotal\ same-adapter triples (\textbf{\WsFourCMixCompBestClean\%}--\textbf{\WsFourCMixInDistMaxClean\%} \texttt{test-clean} WER), yet still fails on the other speech adapter (\WsFourCMixCrossMinClean\%--\WsFourCMixCrossMaxClean\% \texttt{test-clean} WER).}
\label{tab:ws4c}
\vspace{4pt}
\centering
\begin{adjustbox}{max width=\linewidth, max totalheight=395pt}%
\setlength{\tabcolsep}{2.2pt}
\begin{tabular}{@{}llrrrrrr@{}}
\toprule
\textbf{Bridge Training Input} & \textbf{Port ($A \Rightarrow B$)} & \shortstack[r]{\strut\textbf{Wiki}\\\textbf{NLL $\downarrow$}} & \shortstack[r]{\strut\textbf{Transcript}\\\textbf{NLL $\downarrow$}} & \shortstack[r]{\strut\textbf{Audio}\\\textbf{NLL $\downarrow$}} & \shortstack[r]{\strut\textbf{Anchor}\\\textbf{$\cos$ / P@1 $\uparrow$}} & \shortstack[r]{\strut\textbf{WER, same adapter}\\\textbf{(\texttt{clean} / \texttt{other}) $\downarrow$}} & \shortstack[r]{\strut\textbf{WER, other adapter}\\\textbf{(\texttt{clean} / \texttt{other}) $\downarrow$}} \\
\midrule
\multicolumn{8}{@{}l}{\textit{Text-trained bridges on natural WikiText-103 (\ZeroAudioPct\% audio; evaluated with Moonshine and Whisper adapters)}} \\
WikiText only & Qwen2.5-1.5B-IT $\Rightarrow$ Gemma-2-2B-IT & \WsFourCWikiFitQwenToGemmaTwoLnWikiVal & \WsFourCWikiFitQwenToGemmaTwoLnLsVal & --- & \WsFourCWikiFitQwenToGemmaTwoLnCos / \WsFourCWikiFitQwenToGemmaTwoLnPatOne\% & MS: \WsFourCWikiCompMoonshineQwenToGemmaTwoLnCleanCell / \WsFourCWikiCompMoonshineQwenToGemmaTwoLnOtherCell & Wh: \WsFourCWikiCompWhisperQwenToGemmaTwoLnCleanCell / \WsFourCWikiCompWhisperQwenToGemmaTwoLnOtherCell \\
WikiText only & Gemma-2-2B-IT $\Rightarrow$ Qwen2.5-1.5B-IT & \WsFourCWikiFitGemmaTwoToQwenLnWikiVal & \WsFourCWikiFitGemmaTwoToQwenLnLsVal & --- & \WsFourCWikiFitGemmaTwoToQwenLnCos / \WsFourCWikiFitGemmaTwoToQwenLnPatOne\% & MS: \WsFourCWikiCompMoonshineGemmaTwoToQwenLnCleanCell / \WsFourCWikiCompMoonshineGemmaTwoToQwenLnOtherCell & Wh: \WsFourCWikiCompWhisperGemmaTwoToQwenLnCleanCell / \WsFourCWikiCompWhisperGemmaTwoToQwenLnOtherCell \\
WikiText only & Qwen2.5-1.5B-IT $\Rightarrow$ Gemma-3-4B-IT & \WsFourCWikiFitQwenToGemmaThreeLnWikiVal & \WsFourCWikiFitQwenToGemmaThreeLnLsVal & --- & \WsFourCWikiFitQwenToGemmaThreeLnCos / \WsFourCWikiFitQwenToGemmaThreeLnPatOne\% & MS: \WsFourCWikiCompMoonshineQwenToGemmaThreeLnCleanCell / \WsFourCWikiCompMoonshineQwenToGemmaThreeLnOtherCell & Wh: \WsFourCWikiCompWhisperQwenToGemmaThreeLnCleanCell / \WsFourCWikiCompWhisperQwenToGemmaThreeLnOtherCell \\
WikiText only & Gemma-2-2B-IT $\Rightarrow$ Gemma-3-4B-IT & \WsFourCWikiFitGemmaTwoToGemmaThreeLnWikiVal & \WsFourCWikiFitGemmaTwoToGemmaThreeLnLsVal & --- & \WsFourCWikiFitGemmaTwoToGemmaThreeLnCos / \WsFourCWikiFitGemmaTwoToGemmaThreeLnPatOne\% & MS: \WsFourCWikiCompMoonshineGemmaTwoToGemmaThreeLnCleanCell / \WsFourCWikiCompMoonshineGemmaTwoToGemmaThreeLnOtherCell & Wh: \WsFourCWikiCompWhisperGemmaTwoToGemmaThreeLnCleanCell / \WsFourCWikiCompWhisperGemmaTwoToGemmaThreeLnOtherCell \\
\midrule
\multicolumn{8}{@{}l}{\textit{Mixed bridges (\MixtureTextPct\% WikiText-103 text + \MixtureAudioPct\% frozen speech soft prompts; same adapter vs.\ other adapter)}} \\
WikiText + Moonshine & Qwen2.5-1.5B-IT $\Rightarrow$ Gemma-2-2B-IT & \WsFourCMixFitMoonshineQwenToGemmaTwoLnWikiVal & \WsFourCMixFitMoonshineQwenToGemmaTwoLnLsVal & \WsFourCMixFitMoonshineQwenToGemmaTwoLnAudioVal & \WsFourCMixFitMoonshineQwenToGemmaTwoLnCos / \WsFourCMixFitMoonshineQwenToGemmaTwoLnPatOne\% & \textbf{\WsFourCMixCompInDistMoonshineQwenToGemmaTwoLnCleanCell / \WsFourCMixCompInDistMoonshineQwenToGemmaTwoLnOtherCell} & Wh: \WsFourCMixCompCrossWhisperOnMoonshineQwenToGemmaTwoLnCleanCell / \WsFourCMixCompCrossWhisperOnMoonshineQwenToGemmaTwoLnOtherCell \\
WikiText + Whisper & Qwen2.5-1.5B-IT $\Rightarrow$ Gemma-2-2B-IT & \WsFourCMixFitWhisperQwenToGemmaTwoLnWikiVal & \WsFourCMixFitWhisperQwenToGemmaTwoLnLsVal & \WsFourCMixFitWhisperQwenToGemmaTwoLnAudioVal & \WsFourCMixFitWhisperQwenToGemmaTwoLnCos / \WsFourCMixFitWhisperQwenToGemmaTwoLnPatOne\% & \textbf{\WsFourCMixCompInDistWhisperQwenToGemmaTwoLnCleanCell / \WsFourCMixCompInDistWhisperQwenToGemmaTwoLnOtherCell} & MS: \WsFourCMixCompCrossMoonshineOnWhisperQwenToGemmaTwoLnCleanCell / \WsFourCMixCompCrossMoonshineOnWhisperQwenToGemmaTwoLnOtherCell \\
WikiText + Moonshine & Gemma-2-2B-IT $\Rightarrow$ Qwen2.5-1.5B-IT & \WsFourCMixFitMoonshineGemmaTwoToQwenLnWikiVal & \WsFourCMixFitMoonshineGemmaTwoToQwenLnLsVal & \WsFourCMixFitMoonshineGemmaTwoToQwenLnAudioVal & \WsFourCMixFitMoonshineGemmaTwoToQwenLnCos / \WsFourCMixFitMoonshineGemmaTwoToQwenLnPatOne\% & \textbf{\WsFourCMixCompInDistMoonshineGemmaTwoToQwenLnCleanCell / \WsFourCMixCompInDistMoonshineGemmaTwoToQwenLnOtherCell} & Wh: \WsFourCMixCompCrossWhisperOnMoonshineGemmaTwoToQwenLnCleanCell / \WsFourCMixCompCrossWhisperOnMoonshineGemmaTwoToQwenLnOtherCell \\
WikiText + Whisper & Gemma-2-2B-IT $\Rightarrow$ Qwen2.5-1.5B-IT & \WsFourCMixFitWhisperGemmaTwoToQwenLnWikiVal & \WsFourCMixFitWhisperGemmaTwoToQwenLnLsVal & \WsFourCMixFitWhisperGemmaTwoToQwenLnAudioVal & \WsFourCMixFitWhisperGemmaTwoToQwenLnCos / \WsFourCMixFitWhisperGemmaTwoToQwenLnPatOne\% & \textbf{\WsFourCMixCompInDistWhisperGemmaTwoToQwenLnCleanCell / \WsFourCMixCompInDistWhisperGemmaTwoToQwenLnOtherCell} & MS: \WsFourCMixCompCrossMoonshineOnWhisperGemmaTwoToQwenLnCleanCell / \WsFourCMixCompCrossMoonshineOnWhisperGemmaTwoToQwenLnOtherCell \\
WikiText + Moonshine & Qwen2.5-1.5B-IT $\Rightarrow$ Gemma-3-4B-IT & \WsFourCMixFitMoonshineQwenToGemmaThreeLnWikiVal & \WsFourCMixFitMoonshineQwenToGemmaThreeLnLsVal & \WsFourCMixFitMoonshineQwenToGemmaThreeLnAudioVal & \WsFourCMixFitMoonshineQwenToGemmaThreeLnCos / \WsFourCMixFitMoonshineQwenToGemmaThreeLnPatOne\% & \textbf{\WsFourCMixCompInDistMoonshineQwenToGemmaThreeLnCleanCell / \WsFourCMixCompInDistMoonshineQwenToGemmaThreeLnOtherCell} & Wh: \WsFourCMixCompCrossWhisperOnMoonshineQwenToGemmaThreeLnCleanCell / \WsFourCMixCompCrossWhisperOnMoonshineQwenToGemmaThreeLnOtherCell \\
WikiText + Whisper & Qwen2.5-1.5B-IT $\Rightarrow$ Gemma-3-4B-IT & \WsFourCMixFitWhisperQwenToGemmaThreeLnWikiVal & \WsFourCMixFitWhisperQwenToGemmaThreeLnLsVal & \WsFourCMixFitWhisperQwenToGemmaThreeLnAudioVal & \WsFourCMixFitWhisperQwenToGemmaThreeLnCos / \WsFourCMixFitWhisperQwenToGemmaThreeLnPatOne\% & \textbf{\WsFourCMixCompInDistWhisperQwenToGemmaThreeLnCleanCell / \WsFourCMixCompInDistWhisperQwenToGemmaThreeLnOtherCell} & MS: \WsFourCMixCompCrossMoonshineOnWhisperQwenToGemmaThreeLnCleanCell / \WsFourCMixCompCrossMoonshineOnWhisperQwenToGemmaThreeLnOtherCell \\
WikiText + Moonshine & Gemma-2-2B-IT $\Rightarrow$ Gemma-3-4B-IT & \WsFourCMixFitMoonshineGemmaTwoToGemmaThreeLnWikiVal & \WsFourCMixFitMoonshineGemmaTwoToGemmaThreeLnLsVal & \WsFourCMixFitMoonshineGemmaTwoToGemmaThreeLnAudioVal & \WsFourCMixFitMoonshineGemmaTwoToGemmaThreeLnCos / \WsFourCMixFitMoonshineGemmaTwoToGemmaThreeLnPatOne\% & \textbf{\WsFourCMixCompInDistMoonshineGemmaTwoToGemmaThreeLnCleanCell / \WsFourCMixCompInDistMoonshineGemmaTwoToGemmaThreeLnOtherCell} & Wh: \WsFourCMixCompCrossWhisperOnMoonshineGemmaTwoToGemmaThreeLnCleanCell / \WsFourCMixCompCrossWhisperOnMoonshineGemmaTwoToGemmaThreeLnOtherCell \\
WikiText + Whisper & Gemma-2-2B-IT $\Rightarrow$ Gemma-3-4B-IT & \WsFourCMixFitWhisperGemmaTwoToGemmaThreeLnWikiVal & \WsFourCMixFitWhisperGemmaTwoToGemmaThreeLnLsVal & \WsFourCMixFitWhisperGemmaTwoToGemmaThreeLnAudioVal & \WsFourCMixFitWhisperGemmaTwoToGemmaThreeLnCos / \WsFourCMixFitWhisperGemmaTwoToGemmaThreeLnPatOne\% & \textbf{\WsFourCMixCompInDistWhisperGemmaTwoToGemmaThreeLnCleanCell / \WsFourCMixCompInDistWhisperGemmaTwoToGemmaThreeLnOtherCell} & MS: \WsFourCMixCompCrossMoonshineOnWhisperGemmaTwoToGemmaThreeLnCleanCell / \WsFourCMixCompCrossMoonshineOnWhisperGemmaTwoToGemmaThreeLnOtherCell \\
\bottomrule
\end{tabular}%
\end{adjustbox}
\end{table}
\end{landscape}}

%% file: tables/tab_ws4d.tex
\afterpage{\clearpage\begin{landscape}\thispagestyle{plain}
\begin{table}[p]
\caption{\textbf{Bridges evaluated as vocabularies: transplanting Gemma-2-2B-IT ($\GemmaTwoDim$) into frozen Gemma-3-4B-IT ($\GemmaThreeDim$) across closed-form solvers and gradient-trained bridges.} Each transplant overwrites \WsFourDConeLstsqScaledRowsOverwritten\ of the \GemmaThreeVocab\ input rows (the shared exact-string tokens other than the matched chat-control tokens, which stay native), keeps the output head native, and is evaluated zero-shot on PIQA (\PiqaEvalItems\ items, accuracy $\uparrow$, \PiqaChancePct\% chance) and HellaSwag (\HsEvalItems\ items, accuracy $\uparrow$, \HsChancePct\% chance). The rightmost column reports accuracy difference in percentage points relative to the unmodified Gemma-3-4B-IT baseline measured on the same host (\WsFourDBzeroNativePiqa\% / \WsFourDBzeroNativeHs\%, and $^\dagger$\WsFourDWoneBzeroNativePiqa\% / \WsFourDWoneBzeroNativeHs\% on the host running the two Whisper-bridge transplants).}
\label{tab:ws4d}
\vspace{4pt}
\centering
\begin{adjustbox}{max width=\linewidth, max totalheight=395pt}%
\setlength{\tabcolsep}{3.8pt}
\begin{tabular}{@{}llrrrrr@{}}
\toprule
\textbf{Bridge Method} & \textbf{What the Map Was Fit On} & \shortstack[r]{\strut\textbf{Held-out}\\\textbf{$\cos$ / P@1 $\uparrow$}} & \shortstack[r]{\strut\textbf{Radius}\\\textbf{Ratio}} & \textbf{PIQA $\uparrow$} & \textbf{HellaSwag $\uparrow$} & \shortstack[r]{\strut\textbf{$\Delta$ vs.}\\\textbf{Native (pp)}} \\
\midrule
\multicolumn{7}{@{}l}{\textit{Same-host native reference (unmodified Gemma-3-4B-IT)}} \\
Native Gemma-3-4B-IT (no transplant) & --- & --- & \WsFourDBzeroNativeRr & \textbf{\WsFourDBzeroNativePiqaCell} & \textbf{\WsFourDBzeroNativeHsCell} & --- \\
\midrule
\multicolumn{7}{@{}l}{\textit{Closed-form vocabulary solvers and Haar-random orthogonal control (Gemma-2-2B-IT $\to$ Gemma-3-4B-IT)}} \\
Least squares + target-radius norm & \AnchorTrainPct\% shared vocabulary tokens & \WsFourDConeLstsqScaledCos / \WsFourDConeLstsqScaledPatOne\% & \WsFourDConeLstsqScaledRr & \textbf{\WsFourDConeLstsqScaledPiqaCell} & \textbf{\WsFourDConeLstsqScaledHsCell} & \WsFourDConeLstsqScaledDeltaBzeroPiqa / \WsFourDConeLstsqScaledDeltaBzeroHs \\
Least squares + no output norm & \AnchorTrainPct\% shared vocabulary tokens & \WsFourDCtwoLstsqNoneCos / \WsFourDCtwoLstsqNonePatOne\% & \WsFourDCtwoLstsqNoneRr & \WsFourDCtwoLstsqNonePiqaCell & \WsFourDCtwoLstsqNoneHsCell & \WsFourDCtwoLstsqNoneDeltaBzeroPiqa / \WsFourDCtwoLstsqNoneDeltaBzeroHs \\
Orthogonal Procrustes + target-radius norm & \AnchorTrainPct\% shared vocabulary tokens & \WsFourDCthreeProcScaledCos / \WsFourDCthreeProcScaledPatOne\% & \WsFourDCthreeProcScaledRr & \WsFourDCthreeProcScaledPiqaCell & \WsFourDCthreeProcScaledHsCell & \WsFourDCthreeProcScaledDeltaBzeroPiqa / \WsFourDCthreeProcScaledDeltaBzeroHs \\
Random orthogonal + target-radius norm & Haar random (negative control) & \WsFourDCfourRandScaledCos / \WsFourDCfourRandScaledPatOne\% & \WsFourDCfourRandScaledRr & \WsFourDCfourRandScaledPiqaCell & \WsFourDCfourRandScaledHsCell & \WsFourDCfourRandScaledDeltaBzeroPiqa / \WsFourDCfourRandScaledDeltaBzeroHs \\
\midrule
\multicolumn{7}{@{}l}{\textit{Gradient-trained linear bridges applied as static vocabulary tables (layer norm, zero anchor supervision)}} \\
Text-trained bridge (transcripts) & \FullSignalPct\% LibriSpeech transcripts & \WsFourDToneWteLibriCos / \WsFourDToneWteLibriPatOne\% & \WsFourDToneWteLibriRr & \WsFourDToneWteLibriPiqaCell & \WsFourDToneWteLibriHsCell & \WsFourDToneWteLibriDeltaBzeroPiqa / \WsFourDToneWteLibriDeltaBzeroHs \\
Text-trained bridge (WikiText-103) & \FullSignalPct\% WikiText-103 text & \WsFourDTtwoWteWikiCos / \WsFourDTtwoWteWikiPatOne\% & \WsFourDTtwoWteWikiRr & \textbf{\WsFourDTtwoWteWikiPiqaCell} & \WsFourDTtwoWteWikiHsCell & \WsFourDTtwoWteWikiDeltaBzeroPiqa / \WsFourDTtwoWteWikiDeltaBzeroHs \\
Mixed bridge (text + Moonshine) & \MixtureTextPct\% WikiText + \MixtureAudioPct\% Moonshine soft prompts & \WsFourDTthreeWteMixMoonshineCos / \WsFourDTthreeWteMixMoonshinePatOne\% & \WsFourDTthreeWteMixMoonshineRr & \WsFourDTthreeWteMixMoonshinePiqaCell & \textbf{\WsFourDTthreeWteMixMoonshineHsCell} & \WsFourDTthreeWteMixMoonshineDeltaBzeroPiqa / \WsFourDTthreeWteMixMoonshineDeltaBzeroHs \\
Mixed bridge (text + Whisper)$^\dagger$ & \MixtureTextPct\% WikiText + \MixtureAudioPct\% Whisper soft prompts & \WsFourDTsixWteMixWhisperCos / \WsFourDTsixWteMixWhisperPatOne\% & \WsFourDTsixWteMixWhisperRr & \WsFourDTsixWteMixWhisperPiqaCell & \WsFourDTsixWteMixWhisperHsCell & \WsFourDTsixWteMixWhisperDeltaBzeroPiqa / \WsFourDTsixWteMixWhisperDeltaBzeroHs \\
Soft-prompt bridge (Moonshine) & \FullSignalPct\% Moonshine soft prompts & \WsFourDTfourWteAudioMoonshineCos / \WsFourDTfourWteAudioMoonshinePatOne\% & \WsFourDTfourWteAudioMoonshineRr & \WsFourDTfourWteAudioMoonshinePiqaCell & \WsFourDTfourWteAudioMoonshineHsCell & \WsFourDTfourWteAudioMoonshineDeltaBzeroPiqa / \WsFourDTfourWteAudioMoonshineDeltaBzeroHs \\
Soft-prompt bridge (Whisper)$^\dagger$ & \FullSignalPct\% Whisper soft prompts & \WsFourDTfiveWteAudioWhisperCos / \WsFourDTfiveWteAudioWhisperPatOne\% & \WsFourDTfiveWteAudioWhisperRr & \WsFourDTfiveWteAudioWhisperPiqaCell & \WsFourDTfiveWteAudioWhisperHsCell & \WsFourDTfiveWteAudioWhisperDeltaBzeroPiqa / \WsFourDTfiveWteAudioWhisperDeltaBzeroHs \\
Two-adapter soft-prompt bridge & \FullSignalPct\% Whisper + Moonshine soft prompts & \WsFourDTsevenWteJointCos / \WsFourDTsevenWteJointPatOne\% & \WsFourDTsevenWteJointRr & \WsFourDTsevenWteJointPiqaCell & \WsFourDTsevenWteJointHsCell & \WsFourDTsevenWteJointDeltaBzeroPiqa / \WsFourDTsevenWteJointDeltaBzeroHs \\
\bottomrule
\end{tabular}%
\end{adjustbox}
\end{table}
\end{landscape}}

%% file: sections/91_protocol.tex
\subsection{Protocol Details}
\label{app:protocol}

{\small
\begin{itemize}[leftmargin=1.1em,itemsep=1pt,topsep=1pt,parsep=0pt]
\item \textbf{Anchor matching.}
Token IDs $0, \dots, |V_{\text{tok}}|{-}1$ are decoded via \texttt{convert\_tokens\_to\_string}, preserving leading whitespace so SentencePiece (\texttt{U+2581}) and byte-level BPE (\texttt{U+0120}) space markers both render as ASCII space.
Specials (\texttt{all\_special\_ids}), unrenderable pieces, $<2$-char strings, and whitespace-only strings are excluded; duplicate strings keep the lowest target ID, matched 1:1 to the lowest donor ID.
Gemma-3-4B-IT has \GemmaThreeVocab\ rows and 262{,}145 tokenizer entries (63 padded rows unvisited); excluding 8 specials, 19{,}483 1-char strings, and 90 whitespace strings leaves \WsOneGateAnchors\ identity-gate anchors (0 duplicates; 8 matched controls stay native under surgery, writing 242{,}556 rows).
Anchors split 80\% train / 20\% held-out test via \texttt{torch.randperm} with seed 42 (first $\max(1,\lfloor 0.2N\rfloor)$ indices held out).

\item \textbf{Surface matcher (Table~\ref{tab:ws1_s2}).}
Each raw target piece queries the donor vocabulary in order with \texttt{U+2581}/space$\to$\texttt{U+0120}, \texttt{U+2581}$\to$space, \texttt{U+2581}$\to\emptyset$, raw piece, and stripped piece, taking the first non-\texttt{unk} hit (many-to-one on donor: 63{,}797 unique donor IDs across \WsOneQwenClReplicaAnchors\ Qwen3-4B target anchors).
It yields fewer anchors than exact matching (\WsOneQwenClReplicaAnchors\ vs.\ \WsOneQwenLstsqScaledAnchors) because raw strings miss byte-level BPE non-ASCII pieces (19{,}274 of 19{,}275 exact-only anchors are non-ASCII).
Control exemptions are disabled, so all \WsOneQwenClReplicaAnchors\ anchor rows are overwritten while the remaining 196{,}802 non-anchor rows stay native Gemma-3-4B-IT rows.
The match is many-to-one, so 510 of the 13{,}081 held-out pairs share their donor row with a training pair, and this row's held-out metrics are to that extent not out of sample.

\item \textbf{Retrieval P@1 and radius ratio.}
Top-1 cosine retrieval between $\ell_2$-normalized mapped embeddings and $\ell_2$-normalized native target embeddings over the $N_{\text{test}}$ held-out anchors only (not the full target vocabulary).
The radius ratio is the median over the held-out anchors of the norm of a mapped row over the norm of the native row of the same token (all shared anchors for the gradient-trained rows of Table~\ref{tab:ws4d}).

\item \textbf{PIQA and HellaSwag.}
PIQA uses all \PiqaEvalItems\ dev items (2 choices); HellaSwag uses the first \HsEvalItems\ items in dataset order (unshuffled) of the \HsTotalEvalItems-item \texttt{Rowan/hellaswag} validation split (4 choices).
Stripped prefixes (\texttt{goal}/\texttt{ctx}) use \texttt{add\_special\_tokens=True} (prepending \texttt{<bos>}); continuations (\texttt{" "}\,+\,choice) use \texttt{add\_special\_tokens=False}, concatenated, length-sorted, and right-padded (batch 8), scored by length-normalized mean next-token log-probability (\texttt{float32} \texttt{log\_softmax}) over continuation tokens.
Mapped \texttt{inputs\_embeds} rows are divided by the measured \texttt{bfloat16} scale (\GemmaThreeEmbedScaleBf\ on Gemma-3-4B-IT) and cast to \texttt{bfloat16} into \texttt{embed\_tokens.weight}; 17 Gemma control IDs stay native by default (5 matched for cross-family donors, 6 for Gemma-2-2B-IT).
Because Gemma-3-4B-IT ties \texttt{embed\_tokens} and \texttt{lm\_head} by default, \texttt{lm\_head.weight} is cloned into a separate frozen parameter before surgery except in the tied ablation.

\item \textbf{Speech preprocessing, templates, decoding, WER, and validation.}
Training uses \LibriTrainUtts\ 16\,kHz LibriSpeech utterances (\texttt{openslr/librispeech\_asr}, \texttt{all}, lowercased), dropping clips $>30$\,s; evaluation decodes all \LibriTestCleanUtts\ \texttt{test-clean} and \LibriTestOtherUtts\ \texttt{test-other} clips.
Whisper-Small computes 80-bin log-mel features (\texttt{n\_fft}=400, \texttt{hop\_length}=160) padded/truncated to 30\,s (3{,}000 mel frames), runs the frozen encoder (1{,}500 frames), and trims \emph{after} the encoder to $\min(1500,\max(1,\lceil n_{\text{samples}}/320\rceil))$ frames (50\,Hz).
Moonshine-Base passes raw unnormalized 16\,kHz waveforms directly to its frozen encoder (41.67\,Hz, stride 384; no padding or trimming).
Soft prompts sit before \texttt{"Transcribe this audio."} in each native chat template (\texttt{add\_special\_tokens=False}): on Gemma-2/3, prefix \texttt{<bos><start\_of\_turn>user\textbackslash n} (4 tok) and suffix \texttt{Transcribe this audio.}\allowbreak\texttt{<end\_of\_turn>\textbackslash n}\allowbreak\texttt{<start\_of\_turn>model\textbackslash n} (10 tok; stops \texttt{<eos>}, \texttt{<end\_of\_turn>}); on Qwen2.5, prefix \texttt{<|im\_start|>system\textbackslash n}\allowbreak\texttt{You are Qwen, created by Alibaba Cloud.\ You are a helpful assistant.}\allowbreak\texttt{<|im\_end|>\textbackslash n}\allowbreak\texttt{<|im\_start|>user\textbackslash n} (24 tok) and suffix \texttt{Transcribe this audio.}\allowbreak\texttt{<|im\_end|>\textbackslash n}\allowbreak\texttt{<|im\_start|>assistant\textbackslash n} (10 tok; stop \texttt{<|im\_end|>}).
Loss masks (\texttt{-100}) all non-target tokens.
Decoding is greedy (left-padded batch 8, \texttt{max\_new\_tokens}=\EvalMaxNewTokens; native decoders: greedy, \texttt{num\_beams}=1, \texttt{bfloat16}; English transcribe for Whisper; \texttt{max\_new\_tokens}$=\max(16,\lceil 6.5\,\text{dur}\rceil+16)$ for Moonshine, with dur the clip length in seconds).
Before \texttt{jiwer} WER, both strings are lowercased, hyphens become spaces, characters matching \texttt{[\textasciicircum\textbackslash w\textbackslash s']} are stripped, and spaces are collapsed (Appendix~\ref{app:ws2-details} gives WER under Whisper's own text normalizer).
Validation concatenates \texttt{validation.clean} (2{,}703) before \texttt{validation.other} (2{,}864) and takes the first \LibriValUtts\ rows unshuffled, so checkpoint selection uses strictly \texttt{validation.clean}.

\item \textbf{Text, mixed, and noise-control bridges.}
Text bridges map source \texttt{inputs\_embeds} of text $t$ into target $B$'s ASR slot and minimize $B$'s token-mean masked NLL on the same text $t$ plus end-of-turn.
Transcript bridges use all \LibriTrainUtts\ lowercased LibriSpeech transcripts (no duration filter) and the first \LibriValUtts\ \texttt{validation.clean} transcripts.
WikiText-103 bridges use \texttt{Salesforce/wikitext} (\texttt{wikitext-103-raw-v1}, natural casing/punctuation), drop empty lines and \texttt{=} headers, split paragraphs into 48-word chunks, keep 8--48-word chunks with $\le 96$ source tokens, shuffle (seed 42), and cap at \LibriTrainUtts\ train and \LibriValUtts\ val chunks.
Mixed bridges sample 8 WikiText chunks (seed $42+\text{epoch}$) and 8 LibriSpeech utterances via the frozen adapter (seed $1042+\text{epoch}$) per step (0.5 weight each), running $\lceil 281{,}241/8\rceil = 35{,}156$ steps (one pass over each dataset) and selecting checkpoints on $0.5\,\text{wiki\_val} + 0.5\,\text{audio\_val}$.
Noise controls add i.i.d.\ per-coordinate $\mathcal{N}(0,\sigma^2)$ noise with $\sigma=k\,\bar r_A/\sqrt{d_A}$ ($\sigma\in\{0.8401,1.6803,12.9963\}$ for $k\in\{0.50,1.00,7.73\}$) to source embeddings during training only.
All bridges keep default $\mathrm{Norm}_{\text{in}}$ initialization (Table~\ref{tab:protocol_optim}), not source radius $\bar r_A$.

\item \textbf{Bridges as static vocabularies.}
Exported bridge modules ($\mathrm{Norm}_{\text{in}}$, matrix, bias, $\mathrm{Norm}_{\text{out}}$) run in \texttt{eval} mode (\texttt{float32}, no grad) on donor \texttt{inputs\_embeds} rows of all non-control shared anchors in chunks of 4{,}096 with no external norm; outputs are divided by the measured \texttt{bfloat16} scale (\GemmaThreeEmbedScaleBf), cast to \texttt{bfloat16}, and written into \texttt{embed\_tokens.weight} with \texttt{lm\_head} untied and native.
\item \textbf{Muon and batching.}
Muon divides the Nesterov momentum update by its Frobenius norm, applies 5 Newton--Schulz steps $\mathbf{X} \leftarrow a\mathbf{X} + b\,\mathbf{X}\mathbf{X}^{\top}\mathbf{X} + c\,(\mathbf{X}\mathbf{X}^{\top})^{2}\mathbf{X}$ with $(a, b, c) = (3.4445, -4.7750, 2.0315)$, and moves the matrix by the learning rate times $\sqrt{\max(1, d_{\text{out}}/d_{\text{in}})}$ times the result.
The loss of an optimizer step is the mean over its micro-batches (Table~\ref{tab:protocol_optim}) of each micro-batch's token-mean NLL.
Inside adapters and bridges, the frozen speech encoders run in \texttt{float32}.
\end{itemize}
}

\begin{table}[!ht]
\centering
\vspace{-4pt}
\caption{\textbf{Optimization and initialization for adapters and bridges.} All runs use seed 42, \texttt{bfloat16} frozen LLMs, \texttt{float32} modules, 100-step linear warm-up, and plateau halving ($\times 0.5$ after 2 checks every 2{,}500 steps and epoch ends without $>0.5\%$ relative drop; floor $1/16$). Matrix $\mathbf{W}$ and bias $\mathbf{b}$ use \texttt{nn.Linear} $\mathcal{U}(-d_{\text{in}}^{-1/2}, d_{\text{in}}^{-1/2})$; $\mathrm{Norm}_{\text{in}}$ keeps defaults ($\gamma{=}\mathbf{1}$ for RMS/LN, $\beta{=}\mathbf{0}$ for LN; $g{=}\sqrt{d_{\text{in}}}$ for target-radius); $\mathrm{Norm}_{\text{out}}$ starts at target radius $\bar r_M$.}
\label{tab:protocol_optim}
\vspace{1pt}
\renewcommand{\arraystretch}{0.92}
\resizebox{\textwidth}{!}{%
\setlength{\tabcolsep}{3.5pt}
\begin{tabular}{@{}llllllrr@{}}
\toprule
\textbf{Run family} & \textbf{Optimizer (2D / 1D)} & \textbf{LR (2D / 1D)} & \textbf{Mom.\ / $\beta$} & \textbf{WD (2D / 1D)} & \textbf{Micro-batch $\times$ accum} & \textbf{Batch} & \textbf{Steps} \\
\midrule
Speech adapter (Muon) & Muon (5 NS) / AdamW & $0.02$ / $5{\times}10^{-4}$ & $0.95$ (Nest.) / $(0.9, 0.999)$ & $0$ / $0$ & $8{\times}2$, $4{\times}4$, or $2{\times}8$ & 16 & 87{,}890 \\
Speech adapter (AdamW) & AdamW / AdamW & $5{\times}10^{-4}$ / $5{\times}10^{-4}$ & $(0.9, 0.999)$ & $0.01$ / $0$ & $8{\times}2$, $4{\times}4$, or $2{\times}8$ & 16 & 87{,}890 \\
1-epoch bridges (text, 1-adapter) & Muon (5 NS) / AdamW & $0.02$ / $5{\times}10^{-4}$ & $0.95$ (Nest.) / $(0.9, 0.999)$ & $0$ / $0$ & $16{\times}1$, $8{\times}2$, $4{\times}4$, or $2{\times}8$ & 16 & 17{,}578 \\
2-stream bridges (mixed, 2-adapter) & Muon (5 NS) / AdamW & $0.02$ / $5{\times}10^{-4}$ & $0.95$ (Nest.) / $(0.9, 0.999)$ & $0$ / $0$ & $2{\times}4 + 2{\times}4$ (8/stream) & 16 & 35{,}156 \\
\bottomrule
\end{tabular}%
}
\end{table}

%% file: main.bbl
\begin{thebibliography}{50}
\providecommand{\natexlab}[1]{#1}
\providecommand{\url}[1]{\texttt{#1}}
\expandafter\ifx\csname urlstyle\endcsname\relax
  \providecommand{\doi}[1]{doi: #1}\else
  \providecommand{\doi}{doi: \begingroup \urlstyle{rm}\Url}\fi

\bibitem[Alayrac et~al.(2022)Alayrac, Donahue, Luc, Miech, Barr, Hasson, Lenc,
  Mensch, Millican, Reynolds, Ring, Rutherford, Cabi, Han, Gong, Samangooei,
  Monteiro, Menick, Borgeaud, Brock, Nematzadeh, Sharifzadeh, Bi{\'n}kowski,
  Barreira, Vinyals, Zisserman, and Simonyan]{alayrac2022flamingo}
Alayrac, J.-B., Donahue, J., Luc, P., Miech, A., Barr, I., Hasson, Y., Lenc,
  K., Mensch, A., Millican, K., Reynolds, M., Ring, R., Rutherford, E., Cabi,
  S., Han, T., Gong, Z., Samangooei, S., Monteiro, M., Menick, J.~L., Borgeaud,
  S., Brock, A., Nematzadeh, A., Sharifzadeh, S., Bi{\'n}kowski, M., Barreira,
  R., Vinyals, O., Zisserman, A., and Simonyan, K.
\newblock {Flamingo: A Visual Language Model for Few-Shot Learning}.
\newblock In \emph{Advances in Neural Information Processing Systems
  (NeurIPS)}, volume~35, pp.\  23716--23736, 2022.
\newblock \doi{10.52202/068431-1723}.

\bibitem[Bansal et~al.(2021)Bansal, Nakkiran, and Barak]{bansal2021revisiting}
Bansal, Y., Nakkiran, P., and Barak, B.
\newblock {Revisiting Model Stitching to Compare Neural Representations}.
\newblock In \emph{Advances in Neural Information Processing Systems
  (NeurIPS)}, volume~34, pp.\  225--236, 2021.

\bibitem[{Ben Allal} et~al.(2025){Ben Allal}, Lozhkov, Bakouch, {Mart{\'i}n
  Bl{\'a}zquez}, Penedo, Tunstall, Marafioti, {Piqueres Lajar{\'i}n},
  Kydl{\'i}{\v{c}}ek, Srivastav, Lochner, Fahlgren, Nguyen, Burtenshaw,
  Fourrier, Zhao, Larcher, Morlon, Zakka, Raffel, {von Werra}, and
  Wolf]{allal2025smollm2}
{Ben Allal}, L., Lozhkov, A., Bakouch, E., {Mart{\'i}n Bl{\'a}zquez}, G.,
  Penedo, G., Tunstall, L., Marafioti, A., {Piqueres Lajar{\'i}n}, A.,
  Kydl{\'i}{\v{c}}ek, H., Srivastav, V., Lochner, J., Fahlgren, C., Nguyen,
  X.-S., Burtenshaw, B., Fourrier, C., Zhao, H., Larcher, H., Morlon, M.,
  Zakka, C., Raffel, C., {von Werra}, L., and Wolf, T.
\newblock {SmolLM2: When Smol Goes Big --- Data-Centric Training of a Fully
  Open Small Language Model}.
\newblock In \emph{Second Conference on Language Modeling (COLM)}, 2025.

\bibitem[Bisk et~al.(2020)Bisk, Zellers, {Le Bras}, Gao, and
  Choi]{bisk2020piqa}
Bisk, Y., Zellers, R., {Le Bras}, R., Gao, J., and Choi, Y.
\newblock {PIQA: Reasoning About Physical Commonsense in Natural Language}.
\newblock In \emph{Proceedings of the AAAI Conference on Artificial
  Intelligence (AAAI)}, volume~34, pp.\  7432--7439, 2020.
\newblock \doi{10.1609/aaai.v34i05.6239}.

\bibitem[Chen et~al.(2025)Chen, Merullo, Stolfo, and
  Pavlick]{chen2025transferring}
Chen, A., Merullo, J., Stolfo, A., and Pavlick, E.
\newblock {Transferring Linear Features Across Language Models With Model
  Stitching}.
\newblock In \emph{Advances in Neural Information Processing Systems
  (NeurIPS)}, volume~38, 2025.
\newblock \doi{10.52202/085713-1621}.

\bibitem[Chen et~al.(2023)Chen, Wang, Changpinyo, Piergiovanni, Padlewski,
  Salz, Goodman, Grycner, Mustafa, Beyer, Kolesnikov, Puigcerver, Ding, Rong,
  Akbari, Mishra, Xue, Thapliyal, Bradbury, Kuo, Seyedhosseini, Jia,
  Karagol~Ayan, Riquelme, Steiner, Angelova, Zhai, Houlsby, and
  Soricut]{chen2023pali}
Chen, X., Wang, X., Changpinyo, S., Piergiovanni, A.~J., Padlewski, P., Salz,
  D., Goodman, S., Grycner, A., Mustafa, B., Beyer, L., Kolesnikov, A.,
  Puigcerver, J., Ding, N., Rong, K., Akbari, H., Mishra, G., Xue, L.,
  Thapliyal, A., Bradbury, J., Kuo, W., Seyedhosseini, M., Jia, C.,
  Karagol~Ayan, B., Riquelme, C., Steiner, A., Angelova, A., Zhai, X., Houlsby,
  N., and Soricut, R.
\newblock {PaLI: A Jointly-Scaled Multilingual Language-Image Model}.
\newblock In \emph{International Conference on Learning Representations
  (ICLR)}, 2023.

\bibitem[Chu et~al.(2023)Chu, Xu, Zhou, Yang, Zhang, Yan, Zhou, and
  Zhou]{chu2023qwenaudio}
Chu, Y., Xu, J., Zhou, X., Yang, Q., Zhang, S., Yan, Z., Zhou, C., and Zhou, J.
\newblock {Qwen-Audio: Advancing Universal Audio Understanding via Unified
  Large-Scale Audio-Language Models}.
\newblock \emph{arXiv preprint arXiv:2311.07919}, 2023.
\newblock \doi{10.48550/arXiv.2311.07919}.

\bibitem[Conneau et~al.(2018)Conneau, Lample, Ranzato, Denoyer, and
  J{\'e}gou]{conneau2018word}
Conneau, A., Lample, G., Ranzato, M., Denoyer, L., and J{\'e}gou, H.
\newblock {Word Translation Without Parallel Data}.
\newblock In \emph{International Conference on Learning Representations
  (ICLR)}, 2018.

\bibitem[Dobler \& de~Melo(2023)Dobler and de~Melo]{dobler2023focus}
Dobler, K. and de~Melo, G.
\newblock {FOCUS: Effective Embedding Initialization for Monolingual
  Specialization of Multilingual Models}.
\newblock In \emph{Proceedings of the 2023 Conference on Empirical Methods in
  Natural Language Processing (EMNLP)}, pp.\  13440--13454. Association for
  Computational Linguistics, 2023.
\newblock \doi{10.18653/v1/2023.emnlp-main.829}.

\bibitem[Fathullah et~al.(2024)Fathullah, Wu, Lakomkin, Jia, Shangguan, Li,
  Guo, Xiong, Mahadeokar, Kalinli, Fuegen, and Seltzer]{fathullah2024prompting}
Fathullah, Y., Wu, C., Lakomkin, E., Jia, J., Shangguan, Y., Li, K., Guo, J.,
  Xiong, W., Mahadeokar, J., Kalinli, O., Fuegen, C., and Seltzer, M.
\newblock {Prompting Large Language Models with Speech Recognition Abilities}.
\newblock In \emph{2024 IEEE International Conference on Acoustics, Speech and
  Signal Processing (ICASSP)}, pp.\  13351--13355. IEEE, 2024.
\newblock \doi{10.1109/icassp48485.2024.10447605}.

\bibitem[{Gemma Team}(2024)]{gemma2024gemma2}
{Gemma Team}.
\newblock {Gemma 2: Improving Open Language Models at a Practical Size}.
\newblock \emph{arXiv preprint arXiv:2408.00118}, 2024.
\newblock \doi{10.48550/arXiv.2408.00118}.

\bibitem[{Gemma Team}(2025)]{gemma2025gemma3}
{Gemma Team}.
\newblock {Gemma 3 Technical Report}.
\newblock \emph{arXiv preprint arXiv:2503.19786}, 2025.
\newblock \doi{10.48550/arXiv.2503.19786}.

\bibitem[Gorbett \& Jana(2026)Gorbett and Jana]{gorbett2026characterizing}
Gorbett, M. and Jana, S.
\newblock {Ventriloquist LLMs: Linear Alignment of Late-Stage Representations}.
\newblock \emph{arXiv preprint arXiv:2603.18908}, 2026.
\newblock \doi{10.48550/arXiv.2603.18908}.

\bibitem[Huh et~al.(2024)Huh, Cheung, Wang, and Isola]{huh2024platonic}
Huh, M., Cheung, B., Wang, T., and Isola, P.
\newblock {Position: The Platonic Representation Hypothesis}.
\newblock In \emph{Proceedings of the 41st International Conference on Machine
  Learning (ICML)}, volume 235 of \emph{Proceedings of Machine Learning
  Research}, pp.\  20617--20642. PMLR, 2024.

\bibitem[Javaheripi et~al.(2023)Javaheripi, Bubeck, Abdin, Aneja, Mendes, Chen,
  {Del Giorno}, Eldan, Gopi, Gunasekar, Kauffmann, Lee, Li, Nguyen, {de Rosa},
  Saarikivi, Salim, Shah, Santacroce, Behl, Kalai, Wang, Ward, Witte, Zhang,
  and Zhang]{li2023phi2}
Javaheripi, M., Bubeck, S., Abdin, M., Aneja, J., Mendes, C. C.~T., Chen, W.,
  {Del Giorno}, A., Eldan, R., Gopi, S., Gunasekar, S., Kauffmann, P., Lee,
  Y.~T., Li, Y., Nguyen, A., {de Rosa}, G., Saarikivi, O., Salim, A., Shah, S.,
  Santacroce, M., Behl, H.~S., Kalai, A.~T., Wang, X., Ward, R., Witte, P.,
  Zhang, C., and Zhang, Y.
\newblock {Phi-2: The Surprising Power of Small Language Models}.
\newblock Microsoft Research Blog, December 2023.
\newblock URL
  \url{https://www.microsoft.com/en-us/research/blog/phi-2-the-surprising-power-of-small-language-models/}.

\bibitem[Jeffries et~al.(2024)Jeffries, King, Kudlur, Nicholson, Wang, and
  Warden]{jeffries2024moonshine}
Jeffries, N., King, E., Kudlur, M., Nicholson, G., Wang, J., and Warden, P.
\newblock {Moonshine: Speech Recognition for Live Transcription and Voice
  Commands}.
\newblock \emph{arXiv preprint arXiv:2410.15608}, 2024.
\newblock \doi{10.48550/arXiv.2410.15608}.

\bibitem[Jiang \& Li(2024)Jiang and Li]{jiang2024stitching}
Jiang, J.~J. and Li, X.
\newblock {Look Ahead Text Understanding and LLM Stitching}.
\newblock In \emph{Proceedings of the International AAAI Conference on Web and
  Social Media (ICWSM)}, volume~18, pp.\  751--760, 2024.
\newblock \doi{10.1609/icwsm.v18i1.31349}.

\bibitem[Jordan et~al.(2024)Jordan, Jin, Boza, You, Cesista, Newhouse, and
  Bernstein]{jordan2024muon}
Jordan, K., Jin, Y., Boza, V., You, J., Cesista, F., Newhouse, L., and
  Bernstein, J.
\newblock {Muon: An Optimizer for Hidden Layers in Neural Networks}.
\newblock Keller Jordan blog, 2024.
\newblock URL \url{https://kellerjordan.github.io/posts/muon/}.

\bibitem[Khashabi et~al.(2022)Khashabi, Lyu, Min, Qin, Richardson, Welleck,
  Hajishirzi, Khot, Sabharwal, Singh, and Choi]{khashabi2022prompt}
Khashabi, D., Lyu, X., Min, S., Qin, L., Richardson, K., Welleck, S.,
  Hajishirzi, H., Khot, T., Sabharwal, A., Singh, S., and Choi, Y.
\newblock {Prompt Waywardness: The Curious Case of Discretized Interpretation
  of Continuous Prompts}.
\newblock In \emph{Proceedings of the 2022 Conference of the North American
  Chapter of the Association for Computational Linguistics: Human Language
  Technologies (NAACL-HLT)}, pp.\  3631--3643. Association for Computational
  Linguistics, 2022.
\newblock \doi{10.18653/v1/2022.naacl-main.266}.

\bibitem[Lenc \& Vedaldi(2015)Lenc and Vedaldi]{lenc2015understanding}
Lenc, K. and Vedaldi, A.
\newblock {Understanding Image Representations by Measuring Their Equivariance
  and Equivalence}.
\newblock In \emph{2015 IEEE Conference on Computer Vision and Pattern
  Recognition (CVPR)}, pp.\  991--999. IEEE, 2015.
\newblock \doi{10.1109/cvpr.2015.7298701}.

\bibitem[Lester et~al.(2022)Lester, Yurtsever, Shakeri, and
  Constant]{lester2022reducing}
Lester, B., Yurtsever, J., Shakeri, S., and Constant, N.
\newblock {Reducing Retraining by Recycling Parameter-Efficient Prompts}.
\newblock \emph{arXiv preprint arXiv:2208.05577}, 2022.
\newblock \doi{10.48550/arXiv.2208.05577}.

\bibitem[Li et~al.(2026)Li, Xue, Yang, Shi, Chen, Guan, Zhang, and
  Zhang]{li2026unseen}
Li, B., Xue, X., Yang, S., Shi, Y., Chen, X., Guan, Y., Zhang, Y., and Zhang,
  W.
\newblock {The Unseen Bias: How Norm Discrepancy in Pre-Norm MLLMs Leads to
  Visual Information Loss}.
\newblock In \emph{International Conference on Learning Representations
  (ICLR)}, 2026.

\bibitem[Li et~al.(2016)Li, Yosinski, Clune, Lipson, and
  Hopcroft]{li2015convergent}
Li, Y., Yosinski, J., Clune, J., Lipson, H., and Hopcroft, J.
\newblock {Convergent Learning: Do Different Neural Networks Learn the Same
  Representations?}
\newblock In \emph{International Conference on Learning Representations
  (ICLR)}, 2016.

\bibitem[Liao et~al.(2025)Liao, Niu, Meng, Li, Tian, Du, Xiong, Li, Zhu, Yuan,
  Dai, and Cheng]{liao2025langbridge}
Liao, J., Niu, Y., Meng, F., Li, H., Tian, C., Du, Y., Xiong, Y., Li, D., Zhu,
  X., Yuan, L., Dai, J., and Cheng, Y.
\newblock {LangBridge: Interpreting Image as a Combination of Language
  Embeddings}.
\newblock In \emph{Proceedings of the IEEE/CVF International Conference on
  Computer Vision (ICCV)}, pp.\  23752--23762. IEEE, 2025.
\newblock \doi{10.1109/iccv51701.2025.02205}.

\bibitem[Liu et~al.(2023)Liu, Li, Wu, and Lee]{liu2023llava}
Liu, H., Li, C., Wu, Q., and Lee, Y.~J.
\newblock {Visual Instruction Tuning}.
\newblock In \emph{Advances in Neural Information Processing Systems
  (NeurIPS)}, volume~36, pp.\  34892--34916, 2023.
\newblock \doi{10.52202/075280-1516}.

\bibitem[Loshchilov \& Hutter(2019)Loshchilov and
  Hutter]{loshchilov2018decoupled}
Loshchilov, I. and Hutter, F.
\newblock {Decoupled Weight Decay Regularization}.
\newblock In \emph{International Conference on Learning Representations
  (ICLR)}, 2019.

\bibitem[Ma et~al.(2025)Ma, Yang, Yang, Gao, Wang, Du, Yu, Chen, Zheng, Zhang,
  and Chen]{ma2024embarrassingly}
Ma, Z., Yang, G., Yang, Y., Gao, Z., Wang, J., Du, Z., Yu, F., Chen, Q., Zheng,
  S., Zhang, S., and Chen, X.
\newblock {Speech Recognition Meets Large Language Model: Benchmarking, Models,
  and Exploration}.
\newblock In \emph{Proceedings of the AAAI Conference on Artificial
  Intelligence (AAAI)}, volume~39, pp.\  24840--24848, 2025.
\newblock \doi{10.1609/aaai.v39i23.34666}.

\bibitem[Merity et~al.(2017)Merity, Xiong, Bradbury, and
  Socher]{merity2017pointer}
Merity, S., Xiong, C., Bradbury, J., and Socher, R.
\newblock {Pointer Sentinel Mixture Models}.
\newblock In \emph{International Conference on Learning Representations
  (ICLR)}, 2017.

\bibitem[Merullo et~al.(2023)Merullo, Castricato, Eickhoff, and
  Pavlick]{merullo2023linearly}
Merullo, J., Castricato, L., Eickhoff, C., and Pavlick, E.
\newblock {Linearly Mapping from Image to Text Space}.
\newblock In \emph{International Conference on Learning Representations
  (ICLR)}, 2023.

\bibitem[Mikolov et~al.(2013)Mikolov, Le, and Sutskever]{mikolov2013exploiting}
Mikolov, T., Le, Q.~V., and Sutskever, I.
\newblock {Exploiting Similarities Among Languages for Machine Translation}.
\newblock \emph{arXiv preprint arXiv:1309.4168}, 2013.
\newblock \doi{10.48550/arXiv.1309.4168}.

\bibitem[Minixhofer et~al.(2022)Minixhofer, Paischer, and
  Rekabsaz]{minixhofer2022wechsel}
Minixhofer, B., Paischer, F., and Rekabsaz, N.
\newblock {WECHSEL: Effective Initialization of Subword Embeddings for
  Cross-Lingual Transfer of Monolingual Language Models}.
\newblock In \emph{Proceedings of the 2022 Conference of the North American
  Chapter of the Association for Computational Linguistics: Human Language
  Technologies (NAACL-HLT)}, pp.\  3992--4006. Association for Computational
  Linguistics, 2022.
\newblock \doi{10.18653/v1/2022.naacl-main.293}.

\bibitem[Minixhofer et~al.(2024)Minixhofer, Ponti, and
  Vuli{\'c}]{minixhofer2024zeroshot}
Minixhofer, B., Ponti, E.~M., and Vuli{\'c}, I.
\newblock {Zero-Shot Tokenizer Transfer}.
\newblock In \emph{Advances in Neural Information Processing Systems
  (NeurIPS)}, volume~37, pp.\  46791--46818, 2024.
\newblock \doi{10.52202/079017-1484}.

\bibitem[Moschella et~al.(2023)Moschella, Maiorca, Fumero, Norelli, Locatello,
  and Rodol{\`a}]{moschella2023relative}
Moschella, L., Maiorca, V., Fumero, M., Norelli, A., Locatello, F., and
  Rodol{\`a}, E.
\newblock {Relative Representations Enable Zero-Shot Latent Space
  Communication}.
\newblock In \emph{International Conference on Learning Representations
  (ICLR)}, 2023.

\bibitem[Nelson et~al.(2026)Nelson, Fumero, Karaletsos, and
  Locatello]{nelson2026statistical}
Nelson, W., Fumero, M., Karaletsos, T., and Locatello, F.
\newblock {Statistical and Structural Identifiability in Representation
  Learning}.
\newblock In \emph{International Conference on Learning Representations
  (ICLR)}, 2026.

\bibitem[Panayotov et~al.(2015)Panayotov, Chen, Povey, and
  Khudanpur]{panayotov2015librispeech}
Panayotov, V., Chen, G., Povey, D., and Khudanpur, S.
\newblock {LibriSpeech: An ASR Corpus Based on Public Domain Audio Books}.
\newblock In \emph{2015 IEEE International Conference on Acoustics, Speech and
  Signal Processing (ICASSP)}, pp.\  5206--5210. IEEE, 2015.
\newblock \doi{10.1109/icassp.2015.7178964}.

\bibitem[Pandey et~al.(2026)Pandey, Mittal, Bahuguna, and
  Ramakrishnan]{pandey2026dynamic}
Pandey, I., Mittal, A., Bahuguna, V., and Ramakrishnan, G.
\newblock {Dynamic Multi-Expert Projectors with Stabilized Routing for
  Multilingual Speech Recognition}.
\newblock In \emph{2026 IEEE International Conference on Acoustics, Speech and
  Signal Processing (ICASSP)}, pp.\  18737--18741. IEEE, 2026.
\newblock \doi{10.1109/icassp55912.2026.11463560}.

\bibitem[{Qwen Team}(2024)]{yang2024qwen25}
{Qwen Team}.
\newblock {Qwen2.5 Technical Report}.
\newblock \emph{arXiv preprint arXiv:2412.15115}, 2024.
\newblock \doi{10.48550/arXiv.2412.15115}.

\bibitem[{Qwen Team}(2025)]{qwen2025qwen3}
{Qwen Team}.
\newblock {Qwen3 Technical Report}.
\newblock \emph{arXiv preprint arXiv:2505.09388}, 2025.
\newblock \doi{10.48550/arXiv.2505.09388}.

\bibitem[Radford et~al.(2023)Radford, Kim, Xu, Brockman, McLeavey, and
  Sutskever]{radford2023whisper}
Radford, A., Kim, J.~W., Xu, T., Brockman, G., McLeavey, C., and Sutskever, I.
\newblock {Robust Speech Recognition via Large-Scale Weak Supervision}.
\newblock In \emph{Proceedings of the 40th International Conference on Machine
  Learning (ICML)}, volume 202 of \emph{Proceedings of Machine Learning
  Research}, pp.\  28492--28518. PMLR, 2023.

\bibitem[Rios et~al.(2026)Rios, Sourabh, Zaruma, Ewoorkar, Mack, and
  Lai]{rios2026zeroshot}
Rios, E.~A., Sourabh, S., Zaruma, J., Ewoorkar, G., Mack, J., and Lai, B.-C.
\newblock {Zero-Shot Spoken Language Understanding with SLAM-ASR: What Do
  Monolingual ASR-Trained Linear Projectors Enable?}
\newblock \emph{arXiv preprint arXiv:2610.20844}, 2026.
\newblock \doi{10.48550/arXiv.2610.20844}.

\bibitem[Sch{\"o}nemann(1966)]{schonemann1966generalized}
Sch{\"o}nemann, P.~H.
\newblock {A Generalized Solution of the Orthogonal Procrustes Problem}.
\newblock \emph{Psychometrika}, 31\penalty0 (1):\penalty0 1--10, 1966.
\newblock \doi{10.1007/bf02289451}.

\bibitem[Smith et~al.(2017)Smith, Turban, Hamblin, and
  Hammerla]{smith2017offline}
Smith, S.~L., Turban, D. H.~P., Hamblin, S., and Hammerla, N.~Y.
\newblock {Offline Bilingual Word Vectors, Orthogonal Transformations and the
  Inverted Softmax}.
\newblock In \emph{International Conference on Learning Representations
  (ICLR)}, 2017.

\bibitem[Su et~al.(2022)Su, Wang, Qin, Chan, Lin, Wang, Wen, Liu, Li, Li, Hou,
  Sun, and Zhou]{su2022transferability}
Su, Y., Wang, X., Qin, Y., Chan, C.-M., Lin, Y., Wang, H., Wen, K., Liu, Z.,
  Li, P., Li, J., Hou, L., Sun, M., and Zhou, J.
\newblock {On Transferability of Prompt Tuning for Natural Language
  Processing}.
\newblock In \emph{Proceedings of the 2022 Conference of the North American
  Chapter of the Association for Computational Linguistics: Human Language
  Technologies (NAACL-HLT)}, pp.\  3949--3969. Association for Computational
  Linguistics, 2022.
\newblock \doi{10.18653/v1/2022.naacl-main.290}.

\bibitem[Tang et~al.(2024)Tang, Yu, Sun, Chen, Tan, Li, Lu, Ma, and
  Zhang]{tang2024salmonn}
Tang, C., Yu, W., Sun, G., Chen, X., Tan, T., Li, W., Lu, L., Ma, Z., and
  Zhang, C.
\newblock {SALMONN: Towards Generic Hearing Abilities for Large Language
  Models}.
\newblock In \emph{International Conference on Learning Representations
  (ICLR)}, 2024.

\bibitem[Tsimpoukelli et~al.(2021)Tsimpoukelli, Menick, Cabi, Eslami, Vinyals,
  and Hill]{tsimpoukelli2021multimodal}
Tsimpoukelli, M., Menick, J.~L., Cabi, S., Eslami, S. M.~A., Vinyals, O., and
  Hill, F.
\newblock {Multimodal Few-Shot Learning with Frozen Language Models}.
\newblock In \emph{Advances in Neural Information Processing Systems
  (NeurIPS)}, volume~34, pp.\  200--212, 2021.

\bibitem[Vu et~al.(2022)Vu, Lester, Constant, Al-Rfou{'}, and Cer]{vu2022spot}
Vu, T., Lester, B., Constant, N., Al-Rfou{'}, R., and Cer, D.
\newblock {SPoT: Better Frozen Model Adaptation through Soft Prompt Transfer}.
\newblock In \emph{Proceedings of the 60th Annual Meeting of the Association
  for Computational Linguistics (Volume 1: Long Papers)}, pp.\  5039--5059.
  Association for Computational Linguistics, 2022.
\newblock \doi{10.18653/v1/2022.acl-long.346}.

\bibitem[Yang et~al.(2025)Yang, Chen, Xie, and Hansen]{yang2025bridging}
Yang, M., Chen, S.-J., Xie, J., and Hansen, J. H.~L.
\newblock {Bridging the Modality Gap: Softly Discretizing Audio Representation
  for LLM-based Automatic Speech Recognition}.
\newblock In \emph{2025 IEEE Automatic Speech Recognition and Understanding
  Workshop (ASRU)}. IEEE, 2025.
\newblock \doi{10.1109/asru65441.2025.11434631}.

\bibitem[Yu et~al.(2024)Yu, Tang, Sun, Chen, Tan, Li, Lu, Ma, and
  Zhang]{yu2024connecting}
Yu, W., Tang, C., Sun, G., Chen, X., Tan, T., Li, W., Lu, L., Ma, Z., and
  Zhang, C.
\newblock {Connecting Speech Encoder and Large Language Model for ASR}.
\newblock In \emph{2024 IEEE International Conference on Acoustics, Speech and
  Signal Processing (ICASSP)}, pp.\  12637--12641. IEEE, 2024.
\newblock \doi{10.1109/icassp48485.2024.10445874}.

\bibitem[Zellers et~al.(2019)Zellers, Holtzman, Bisk, Farhadi, and
  Choi]{zellers2019hellaswag}
Zellers, R., Holtzman, A., Bisk, Y., Farhadi, A., and Choi, Y.
\newblock {HellaSwag: Can a Machine Really Finish Your Sentence?}
\newblock In \emph{Proceedings of the 57th Annual Meeting of the Association
  for Computational Linguistics (ACL)}, pp.\  4791--4800. Association for
  Computational Linguistics, 2019.
\newblock \doi{10.18653/v1/p19-1472}.

\bibitem[Zhang et~al.(2023)Zhang, Fei, Yao, Ji, Li, Liu, and
  Chua]{zhang2023vpgtrans}
Zhang, A., Fei, H., Yao, Y., Ji, W., Li, L., Liu, Z., and Chua, T.-S.
\newblock {VPGTrans: Transfer Visual Prompt Generator across LLMs}.
\newblock In \emph{Advances in Neural Information Processing Systems
  (NeurIPS)}, volume~36, pp.\  20299--20319, 2023.
\newblock \doi{10.52202/075280-0891}.

\end{thebibliography}
